\documentclass[10pt,logo,copyright]{nvidiatechreport}
\usepackage[authoryear,sort&compress,round]{natbib}

\usepackage[utf8]{inputenc} %
\usepackage[T1]{fontenc}    %

\usepackage{parskip}        %
\usepackage{url}            %
\usepackage{booktabs}       %
\usepackage{amsfonts}       %
\usepackage{nicefrac}       %
\usepackage{microtype}      %
\usepackage{xcolor}         %
\usepackage[dvipsnames]{xcolor} %
\usepackage{graphicx}
\usepackage{animate}        %
\usepackage{subcaption}
\usepackage{tabularx}
\usepackage{makecell}
\usepackage{adjustbox}
\usepackage{setspace}
\newcolumntype{M}[1]{>{\centering\arraybackslash}m{#1}}
\usepackage{float}
\usepackage{tikz}
\usetikzlibrary{positioning,shapes,arrows}
\usepackage{amsmath,amsfonts,bm, bbm,leftindex}
\usepackage{multirow}
\usepackage{comment}
\usepackage{gensymb}
\usepackage{lipsum}
\usetikzlibrary{arrows.meta, positioning, fit}
\usepackage[para]{threeparttable}
\usepackage{tikz}
\usetikzlibrary{tikzmark}

\newenvironment{multicoltextblock}{%
    \hspace{16pt} %
    \begin{minipage}[t]{\dimexpr\textwidth-16pt\relax} %
}{%
    \end{minipage}%
    \unskip\ignorespacesafterend %
}

\newenvironment{multicoltextblockhalf}{%
    \hspace{4pt} %
    \begin{minipage}[t]{\dimexpr\textwidth/2-16pt\relax} %
}{%
    \end{minipage}%
    \unskip\ignorespacesafterend %
}

\newcommand{\prompttext}[1]{\textbf{#1}}
\newcommand{\prompt}[1]{\textsf{#1}}

\def\eqref#1{equation~\ref{#1}}

\def\1{\bm{1}}

\def\x{\mathbf{x}}
\def\y{\mathbf{y}}

\def\rvn{{\mathbf{n}}}

\def\rvx{{\mathbf{x}}}

\def\rmI{{\mathbf{I}}}

\DeclareMathAlphabet{\mathsfit}{\encodingdefault}{\sfdefault}{m}{sl}
\SetMathAlphabet{\mathsfit}{bold}{\encodingdefault}{\sfdefault}{bx}{n}

\def\gL{{\mathcal{L}}}

\def\gN{{\mathcal{N}}}

\newcommand{\pdata}{p_{\rm{data}}}

\newcommand{\E}{\mathbb{E}}

\newcommand{\R}{\mathbb{R}}

\makeatletter
\let\save@mathaccent\mathaccent
\newcommand*\if@single[3]{%
  \setbox0\hbox{${\mathaccent"0362{#1}}^H$}%
  \setbox2\hbox{${\mathaccent"0362{\kern0pt#1}}^H$}%
  \ifdim\ht0=\ht2 #3\else #2\fi
  }
\newcommand*\rel@kern[1]{\kern#1\dimexpr\macc@kerna}
\newcommand*\widebar[1]{\@ifnextchar^{{\wide@bar{#1}{0}}}{\wide@bar{#1}{1}}}
\newcommand*\wide@bar[2]{\if@single{#1}{\wide@bar@{#1}{#2}{1}}{\wide@bar@{#1}{#2}{2}}}
\newcommand*\wide@bar@[3]{%
  \begingroup
  \def\mathaccent##1##2{%
    \let\mathaccent\save@mathaccent
    \if#32 \let\macc@nucleus\first@char \fi
    \setbox\z@\hbox{$\macc@style{\macc@nucleus}_{}$}%
    \setbox\tw@\hbox{$\macc@style{\macc@nucleus}{}_{}$}%
    \dimen@\wd\tw@
    \advance\dimen@-\wd\z@
    \divide\dimen@ 3
    \@tempdima\wd\tw@
    \advance\@tempdima-\scriptspace
    \divide\@tempdima 10
    \advance\dimen@-\@tempdima
    \ifdim\dimen@>\z@ \dimen@0pt\fi
    \rel@kern{0.6}\kern-\dimen@
    \if#31
      \overline{\rel@kern{-0.6}\kern\dimen@\macc@nucleus\rel@kern{0.4}\kern\dimen@}%
      \advance\dimen@0.4\dimexpr\macc@kerna
      \let\final@kern#2%
      \ifdim\dimen@<\z@ \let\final@kern1\fi
      \if\final@kern1 \kern-\dimen@\fi
    \else
      \overline{\rel@kern{-0.6}\kern\dimen@#1}%
    \fi
  }%
  \macc@depth\@ne
  \let\math@bgroup\@empty \let\math@egroup\macc@set@skewchar
  \mathsurround\z@ \frozen@everymath{\mathgroup\macc@group\relax}%
  \macc@set@skewchar\relax
  \let\mathaccentV\macc@nested@a
  \if#31
    \macc@nested@a\relax111{#1}%
  \else
    \def\gobble@till@marker##1\endmarker{}%
    \futurelet\first@char\gobble@till@marker#1\endmarker
    \ifcat\noexpand\first@char A\else
      \def\first@char{}%
    \fi
    \macc@nested@a\relax111{\first@char}%
  \fi
  \endgroup
}
\makeatother

\newcommand{\Bigo}{\mathcal{O}}

\newcommand{\diffusionsevenbvideotoworld}{Cosmos-Predict1-7B-Video2World}

\definecolor{darkred}{rgb}{0.7, 0.0, 0.0}

\usepackage{pifont}
\newcommand{\cmark}{\ding{51}} %
\newcommand{\xmark}{\ding{55}} %

\usepackage[nameinlink]{cleveref}
\crefname{equation}{Eq.}{Eqs.}
\crefname{figure}{Fig.}{Figs.}
\crefname{section}{Sec.}{Sec.}
\crefname{appendix}{App.}{App.}
\crefname{table}{Tab.}{Tabs.}
\crefname{algorithm}{Algo}{Algo}
\crefname{thm}{Thm}{Thm}
\Crefname{thm}{Thm}{Thm}
\crefname{prop}{Prop}{Prop}

\newcommand{\crefnames}[3]{%
  \@for\next:=#1\do{%
    \expandafter\crefname\expandafter{\next}{#2}{#3}%
  }%
}

\title{Cosmos World Foundation Model Platform for~Physical~AI}

\author{NVIDIA\footnote{A detailed list of contributors and acknowledgments can be found in~\cref{sec:contributors} of this paper.}}

\begin{abstract}
Physical AI needs to be trained digitally first. It needs a digital twin of itself, the policy model, and a digital twin of the world, the world model. In this paper, we present the Cosmos World Foundation Model Platform to help developers build customized world models for their Physical AI setups. We position a world foundation model as a general-purpose world model that can be fine-tuned into customized world models for downstream applications. Our platform covers a video curation pipeline, pre-trained world foundation models, examples of post-training of pre-trained world foundation models, and video tokenizers. To help Physical AI builders solve the most critical problems of our society, we make Cosmos open-source and our models open-weight with permissive licenses available via \href{https://github.com/nvidia-cosmos/cosmos-predict1}{NVIDIA Cosmos-Predict1}.
\end{abstract}

\begin{document}

\maketitle

\abscontent

\section{Introduction}\label{sec::intro}

Physical AI is an AI system equipped with sensors and actuators: the sensors allow it to observe the world, and the actuators allow it to interact with and modify the world. It holds the promise of freeing human workers from physical tasks that are dangerous, laborious, or tedious. While several fields of AI have advanced significantly thanks to data and compute scaling in the recent decade, Physical AI only inches forward. This is largely because scaling training data for Physical AI is much more challenging, as the desired data must contain sequences of interleaved observations and actions. These actions perturb the physical world and may cause severe damage to the system and the world. This is especially true when the AI is still in its infancy when exploratory actions are essential. A World Foundation Model (WFM), a digital twin of the physical world that a Physical AI can safely interact with, has been a long-sought remedy to the data scaling problem.

\begin{figure*}[!htp]
    \centering
    \setlength{\tabcolsep}{1pt}
    \renewcommand{\arraystretch}{0}
    \begin{subfigure}{\linewidth}
        \centering
        \textbf{Pre-training: Diffusion WFM}\\[2pt]
        \begin{tabular}{ccccc} %

        \includegraphics[width=0.196\textwidth]{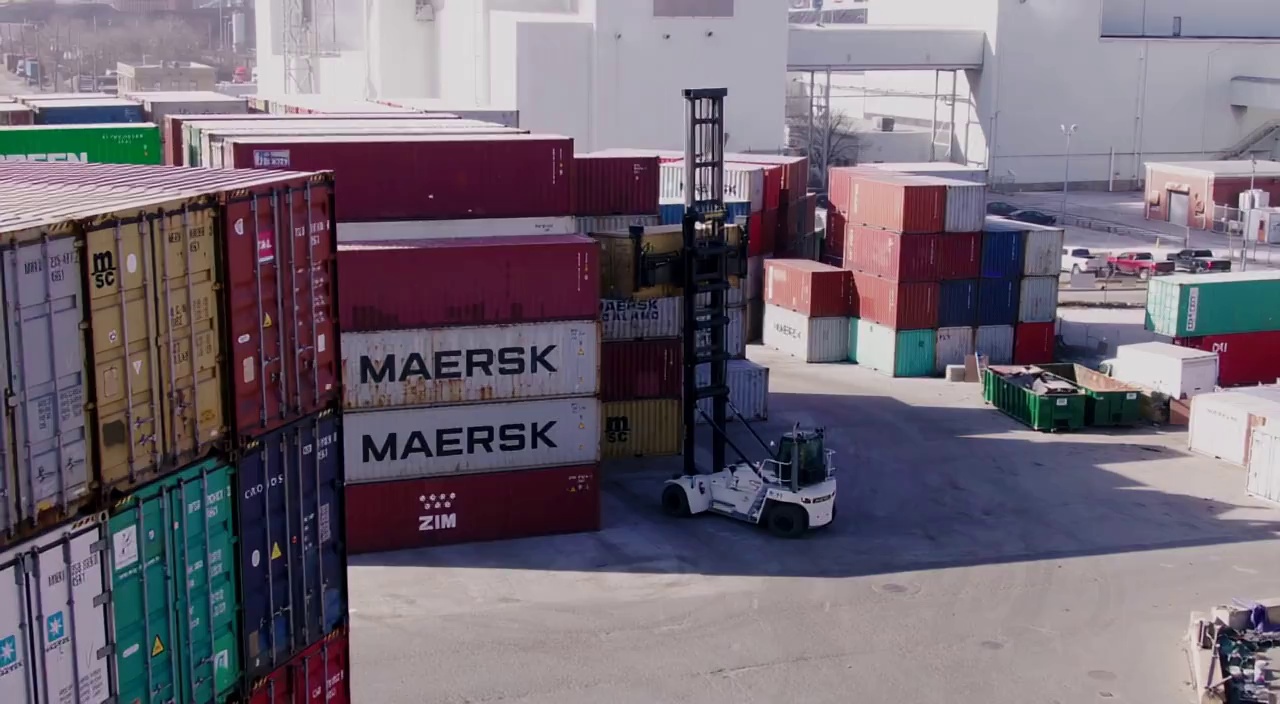} &
        \includegraphics[width=0.196\textwidth]{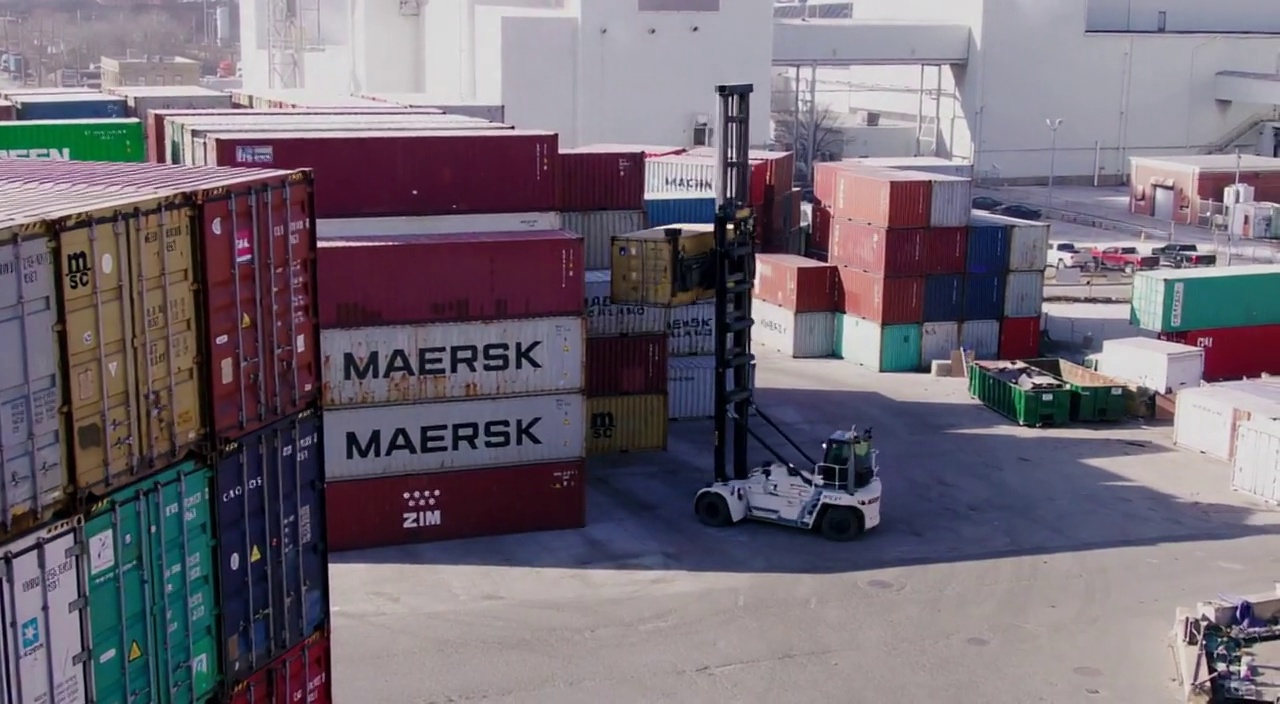} &
        \includegraphics[width=0.196\textwidth]{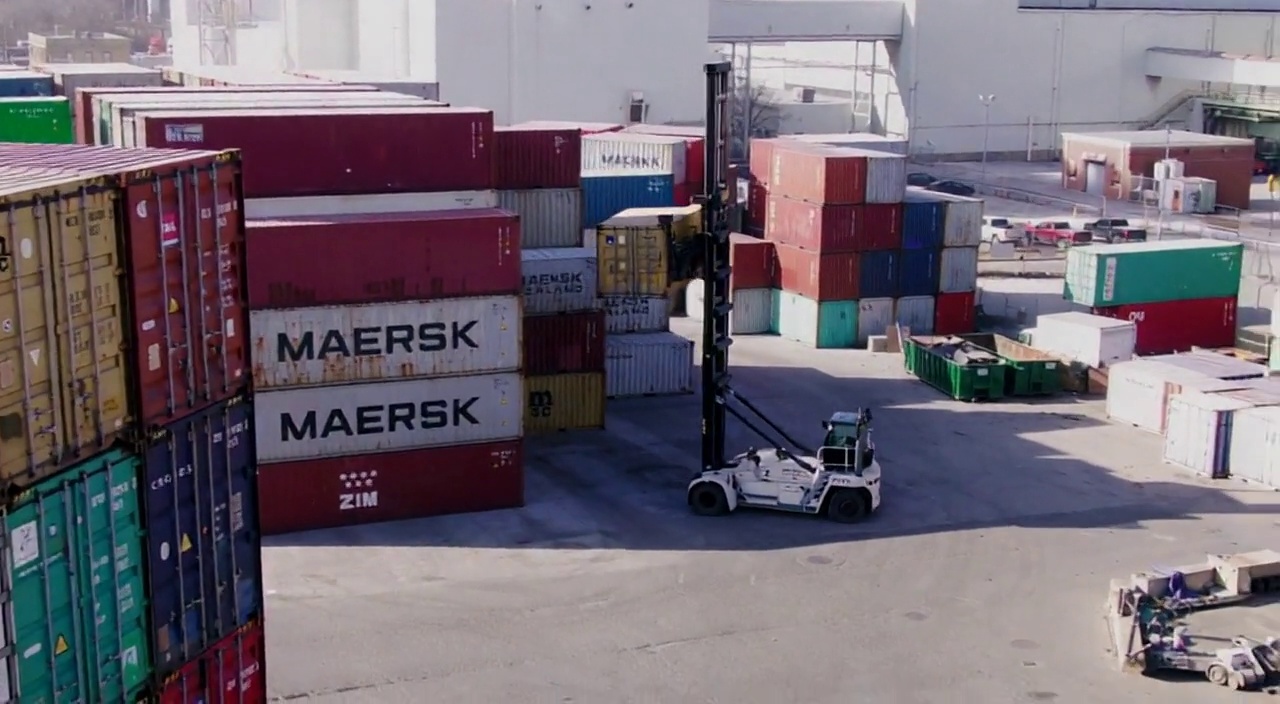} &
        \includegraphics[width=0.196\textwidth]{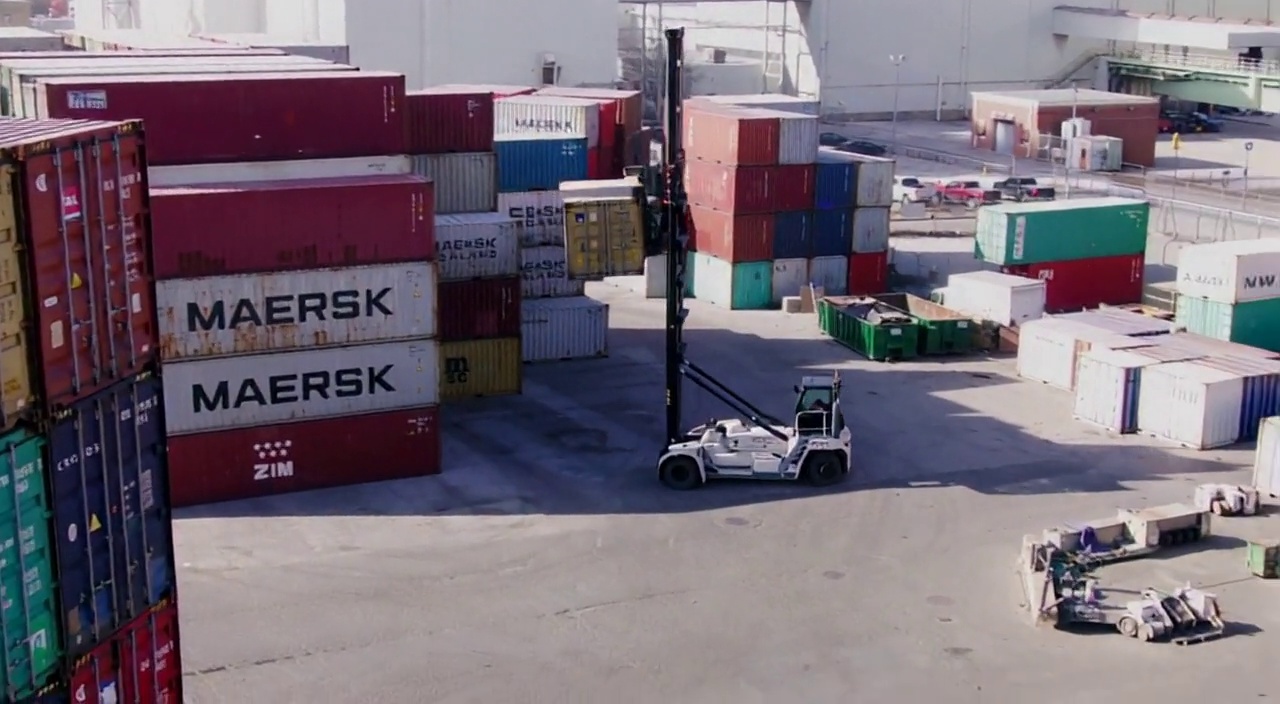} &
        \includegraphics[width=0.196\textwidth]{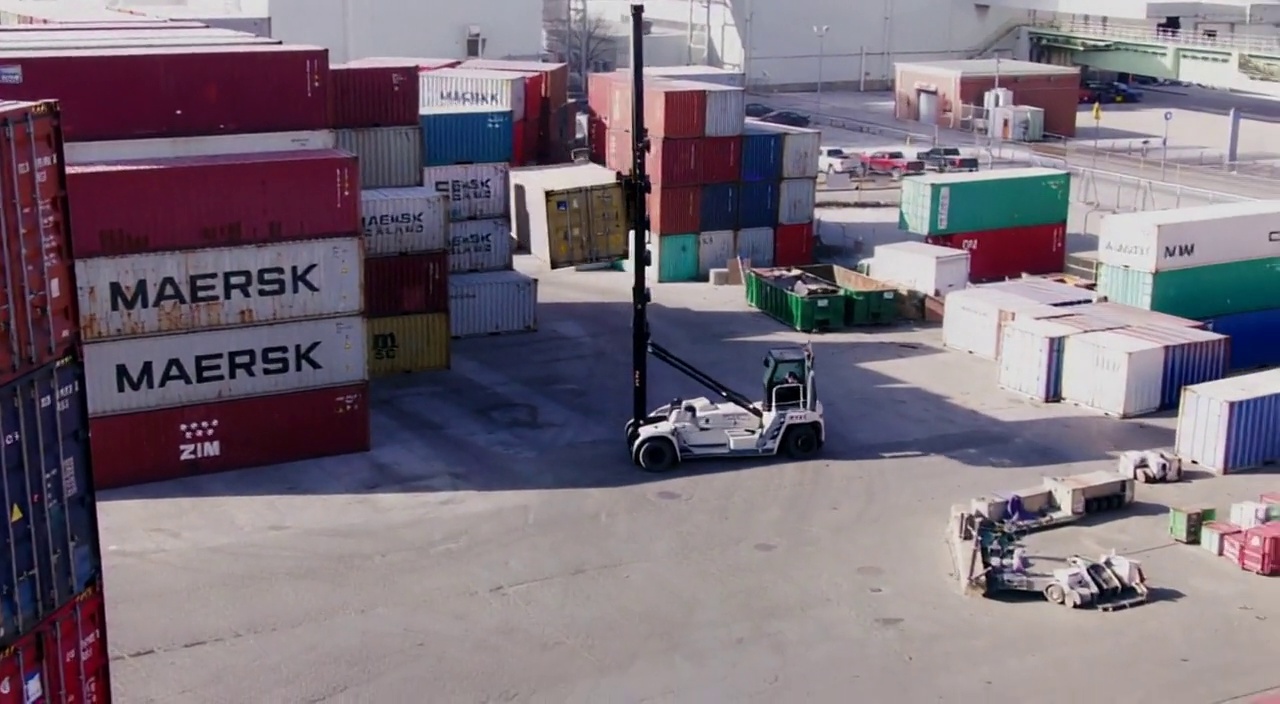}\\[10pt]
        \end{tabular}
    \end{subfigure}

    \begin{subfigure}{\linewidth}
        \centering
        \textbf{Pre-training: Autoregressive WFM}\\[2pt]
        \begin{tabular}{ccccc} %

        \includegraphics[width=0.196\textwidth]{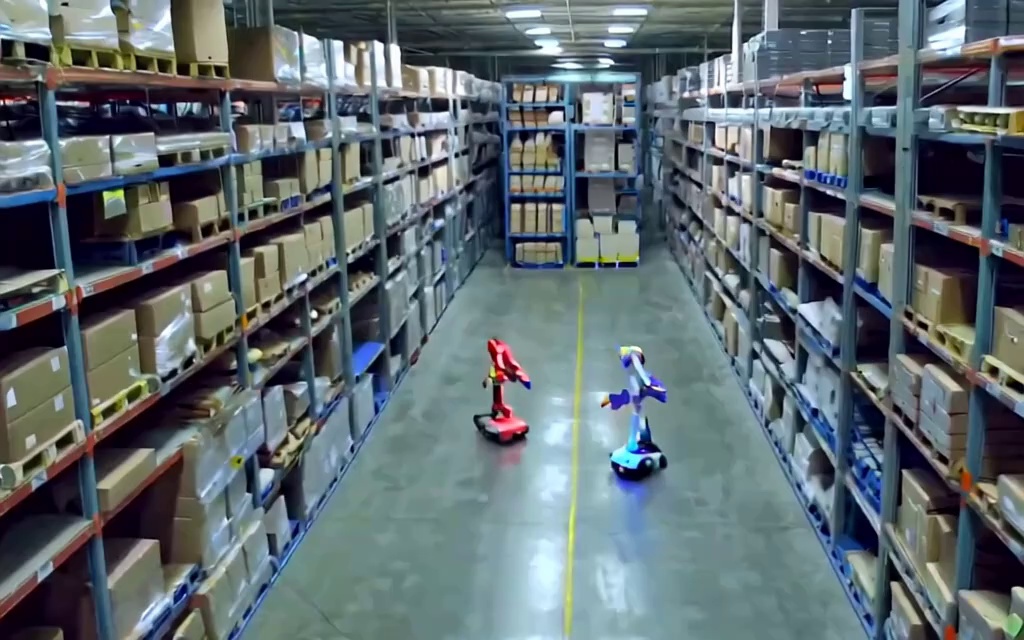} &
        \includegraphics[width=0.196\textwidth]{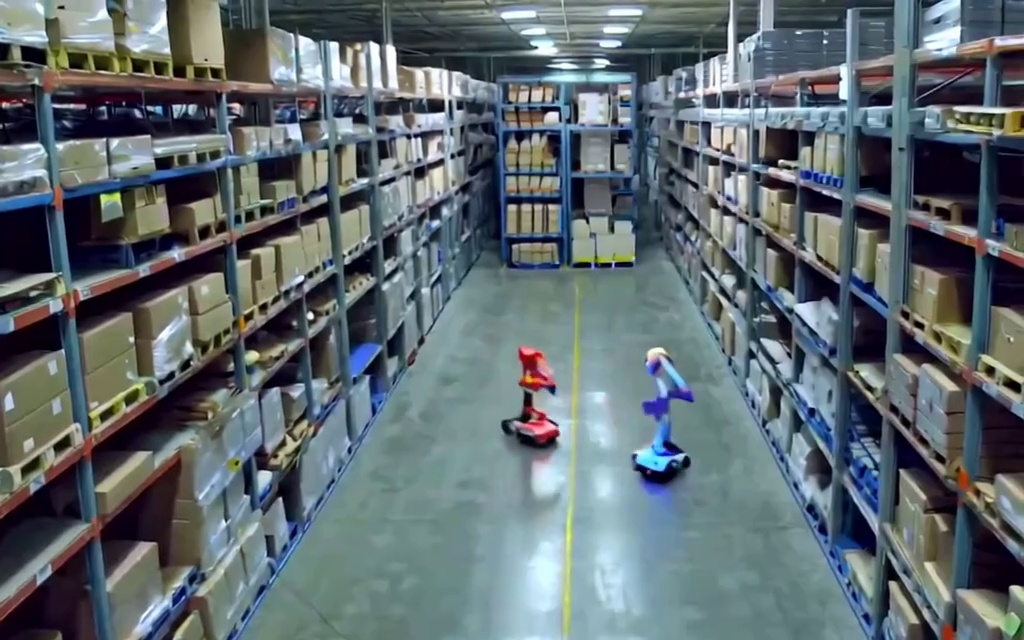} &
        \includegraphics[width=0.196\textwidth]{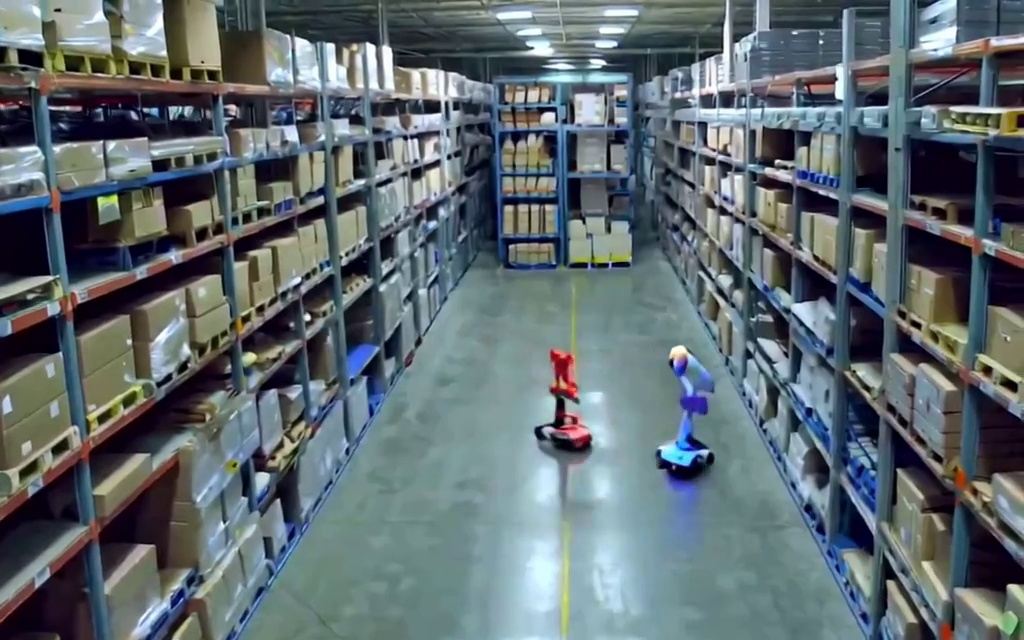} &
        \includegraphics[width=0.196\textwidth]{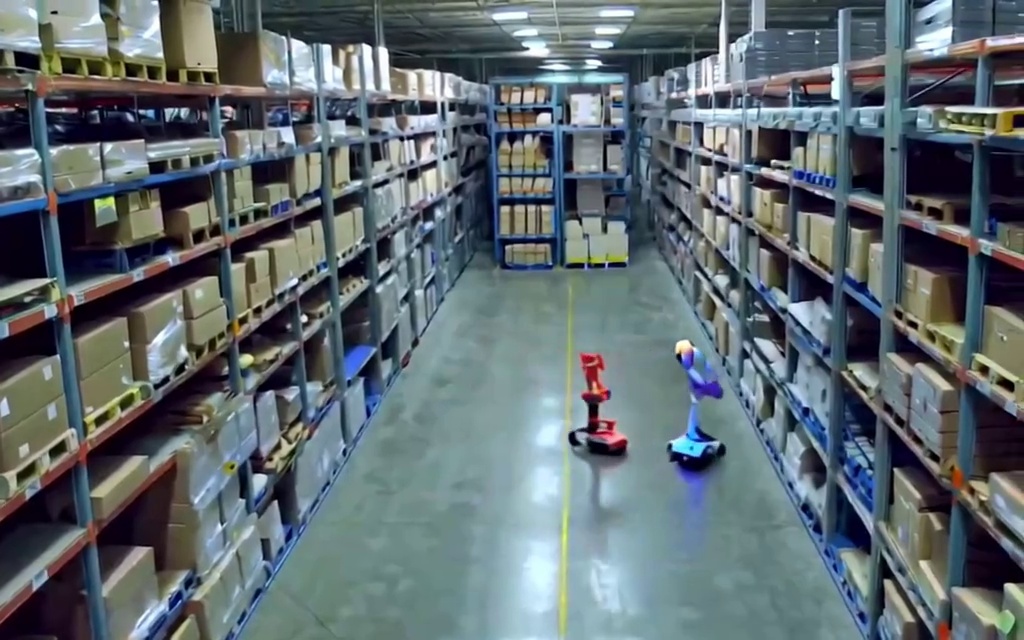} &
        \includegraphics[width=0.196 \textwidth]{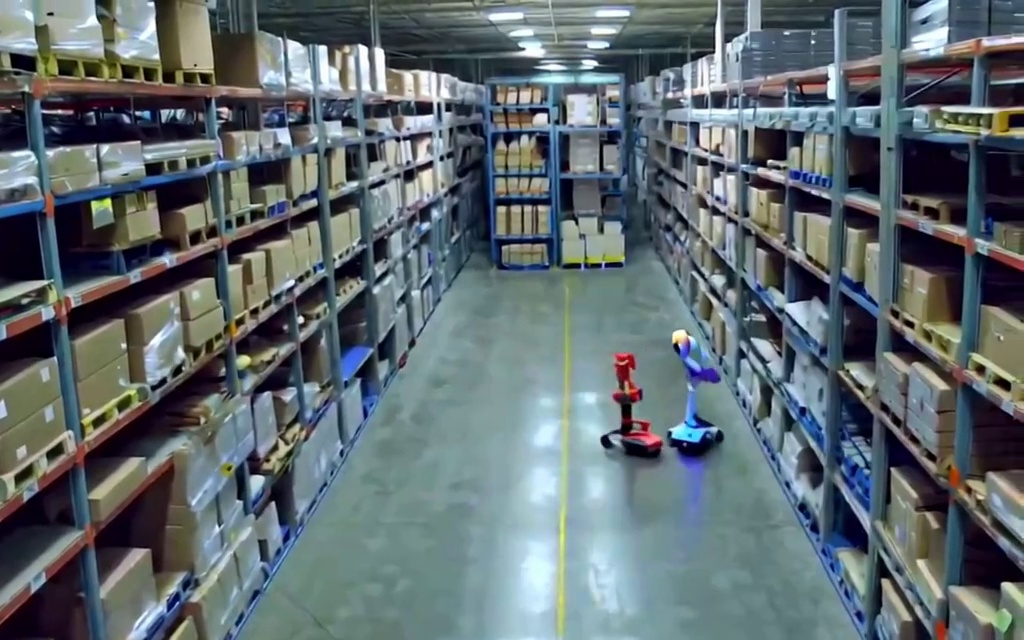}\\[10pt]
        \end{tabular}
    \end{subfigure}

    \begin{subfigure}{\linewidth}
        \centering
        \textbf{Post-training: Camera Control}
        \vspace{2pt}
    \end{subfigure}

    \begin{subfigure}[c]{0.185\textwidth}
        \centering
        \includegraphics[width=\textwidth]{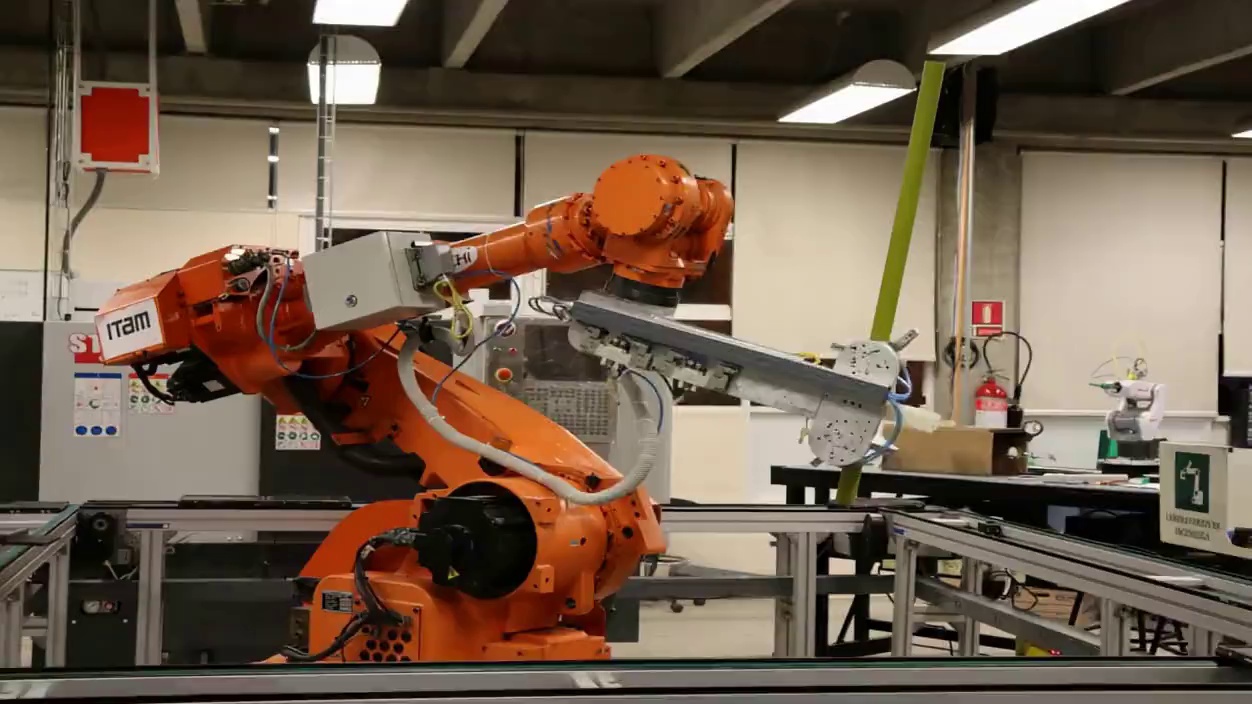}
    \end{subfigure}
    \hfill
    \begin{subfigure}[c]{0.048\textwidth}
        \centering
        \includegraphics[width=0.8\textwidth]{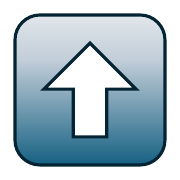}
    \end{subfigure}
    \hfill
    \begin{subfigure}[c]{0.185\textwidth}
        \centering
        \includegraphics[width=\textwidth]{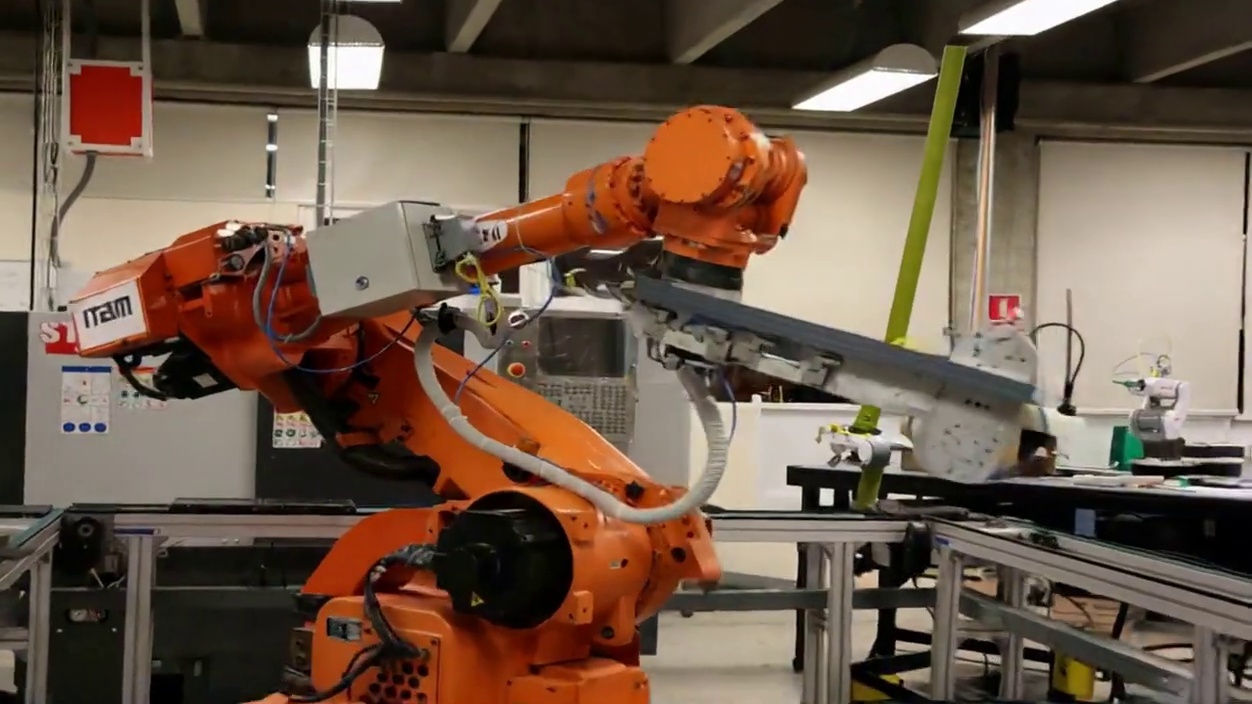}
    \end{subfigure}
    \hfill
    \begin{subfigure}[c]{0.185\textwidth}
        \centering
        \includegraphics[width=\textwidth]{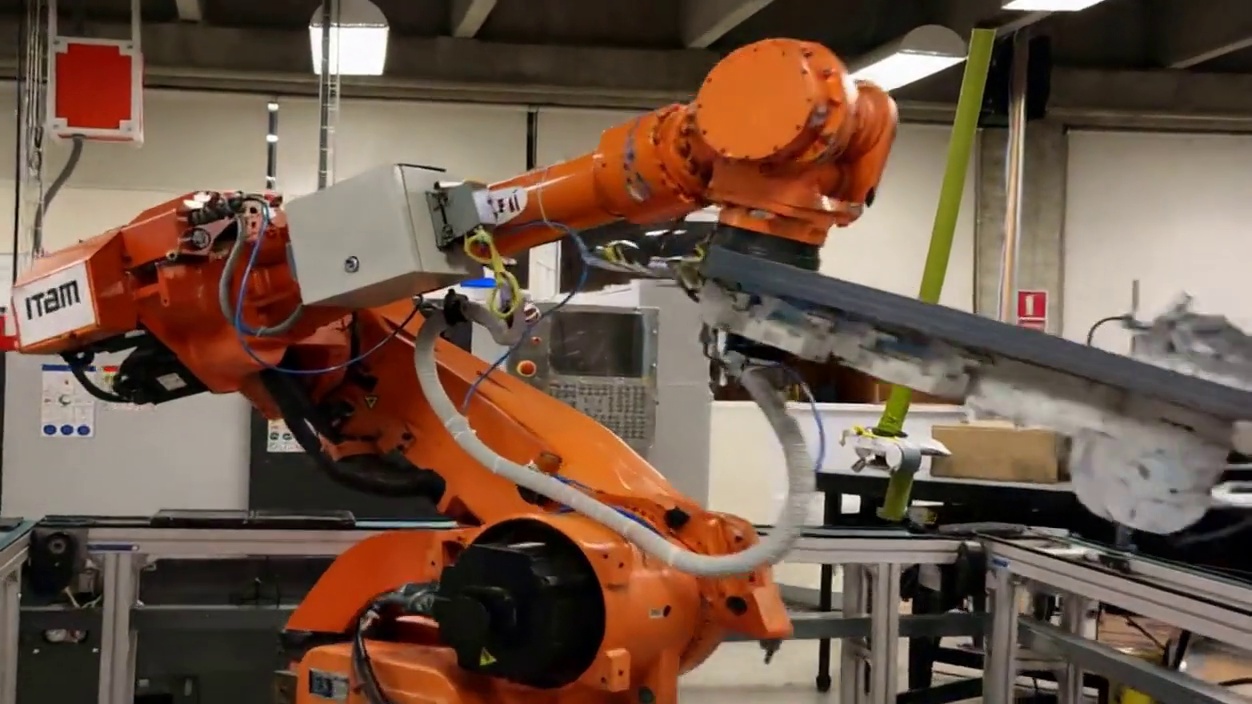}
    \end{subfigure}
    \hfill
    \begin{subfigure}[c]{0.185\textwidth}
        \centering
        \includegraphics[width=\textwidth]{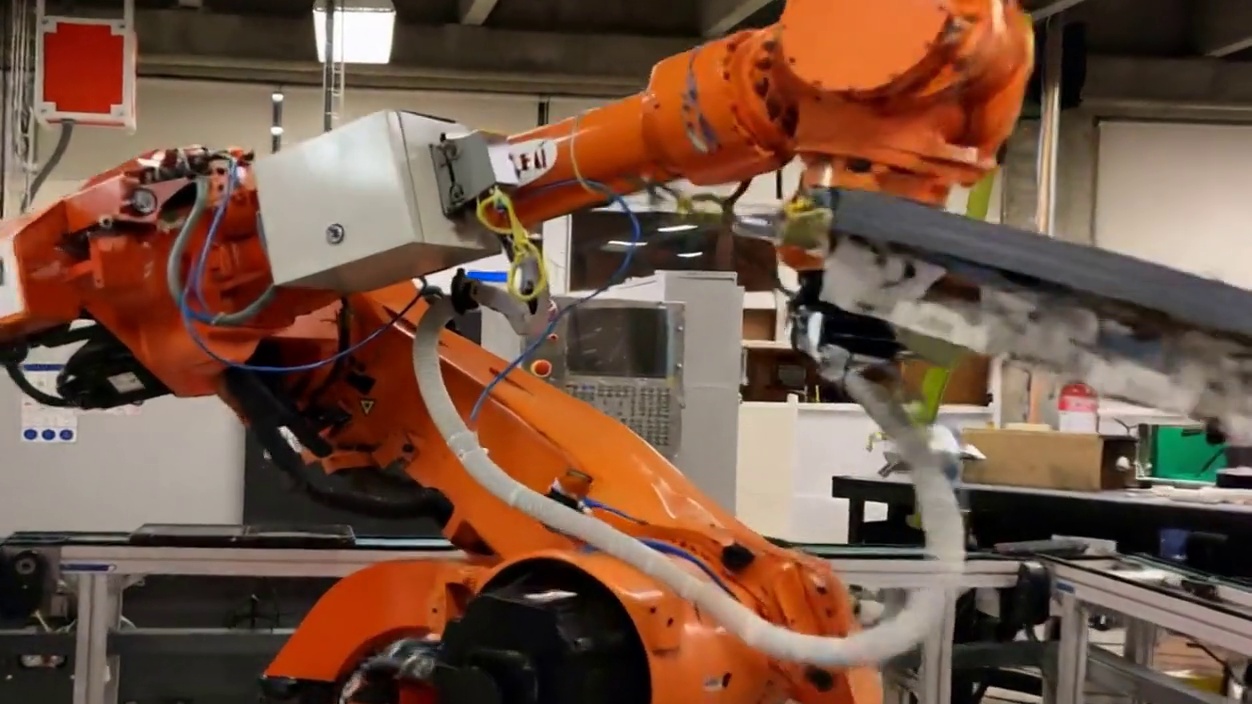}
    \end{subfigure}
    \hfill
    \begin{subfigure}[c]{0.185\textwidth}
        \centering
        \includegraphics[width=\textwidth]{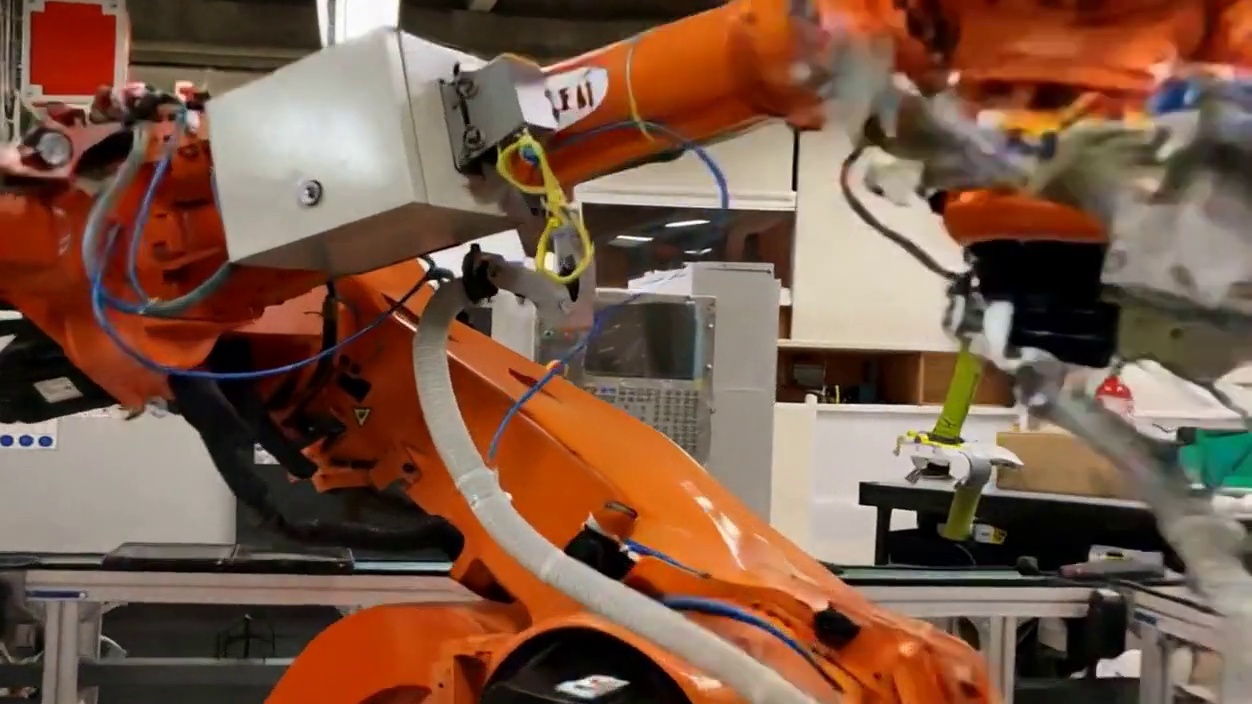}
    \end{subfigure}\\[1pt]

    \begin{subfigure}[c]{0.185\textwidth}
        \centering
        \includegraphics[width=\textwidth]{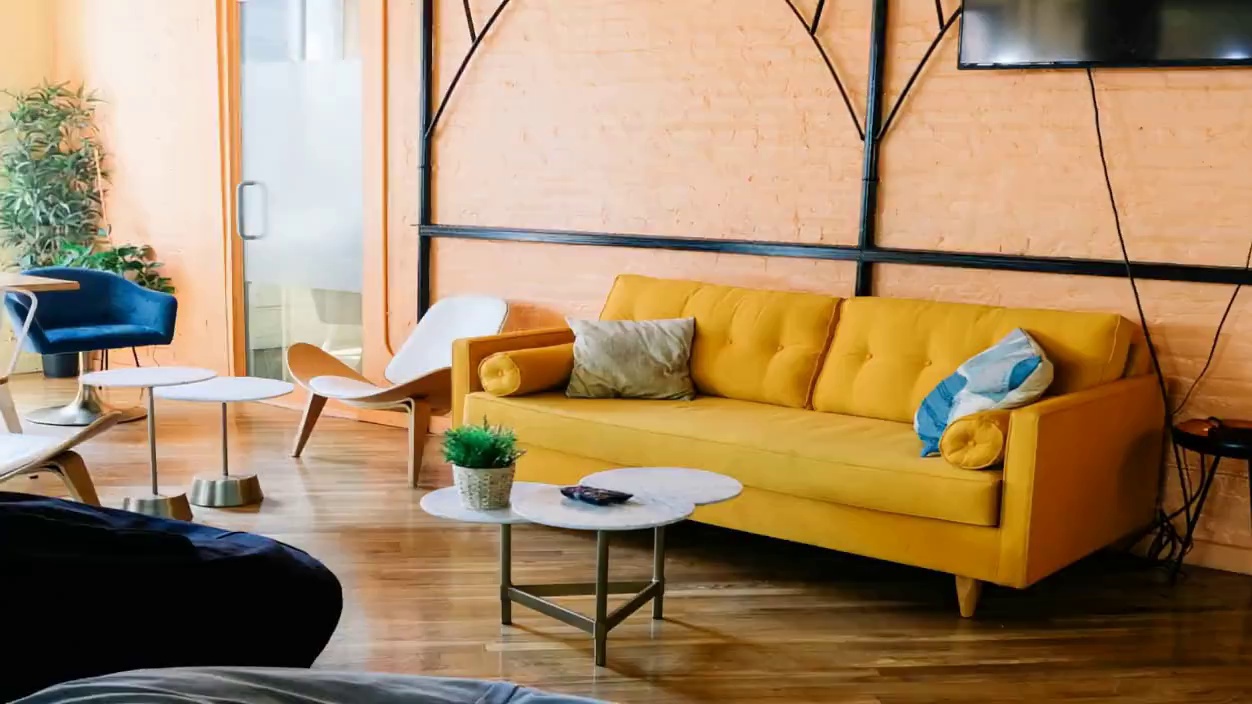}
    \end{subfigure}
    \hfill
    \begin{subfigure}[c]{0.048\textwidth}
        \centering
        \includegraphics[width=0.8\textwidth]{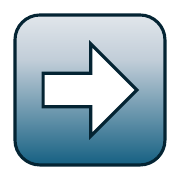}
    \end{subfigure}
    \hfill
    \begin{subfigure}[c]{0.185\textwidth}
        \centering
        \includegraphics[width=\textwidth]{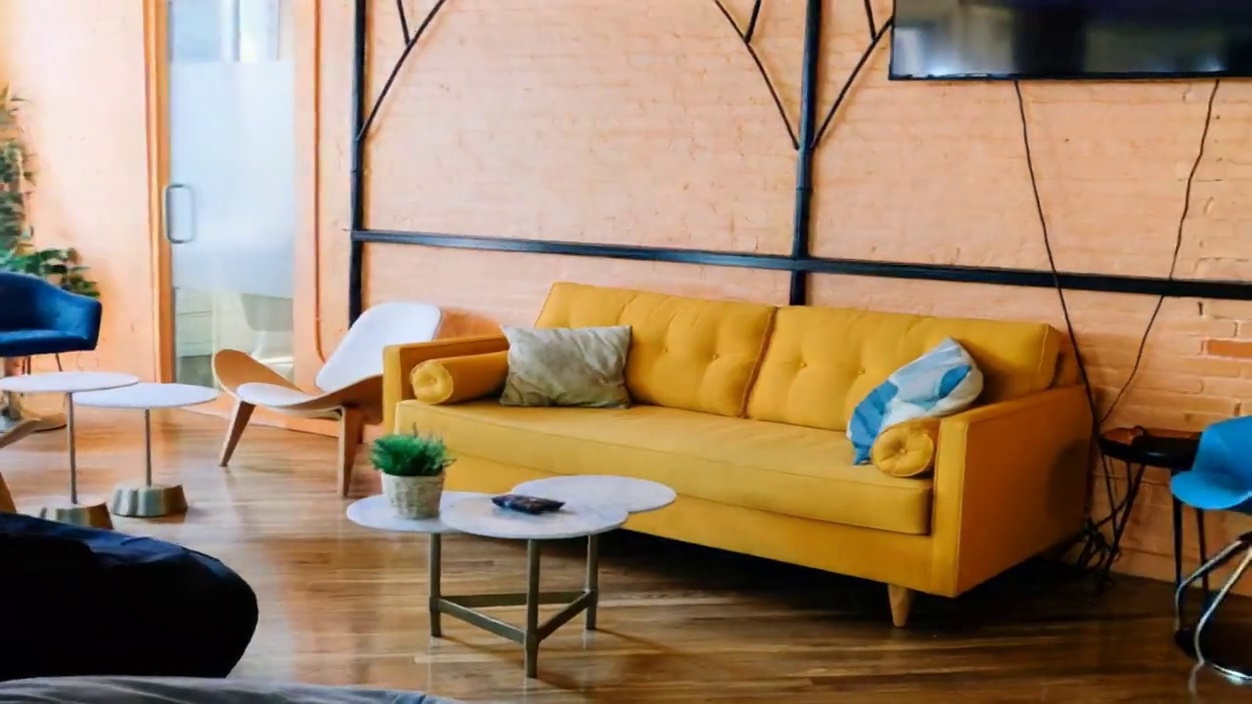}
    \end{subfigure}
    \hfill
    \begin{subfigure}[c]{0.185\textwidth}
        \centering
        \includegraphics[width=\textwidth]{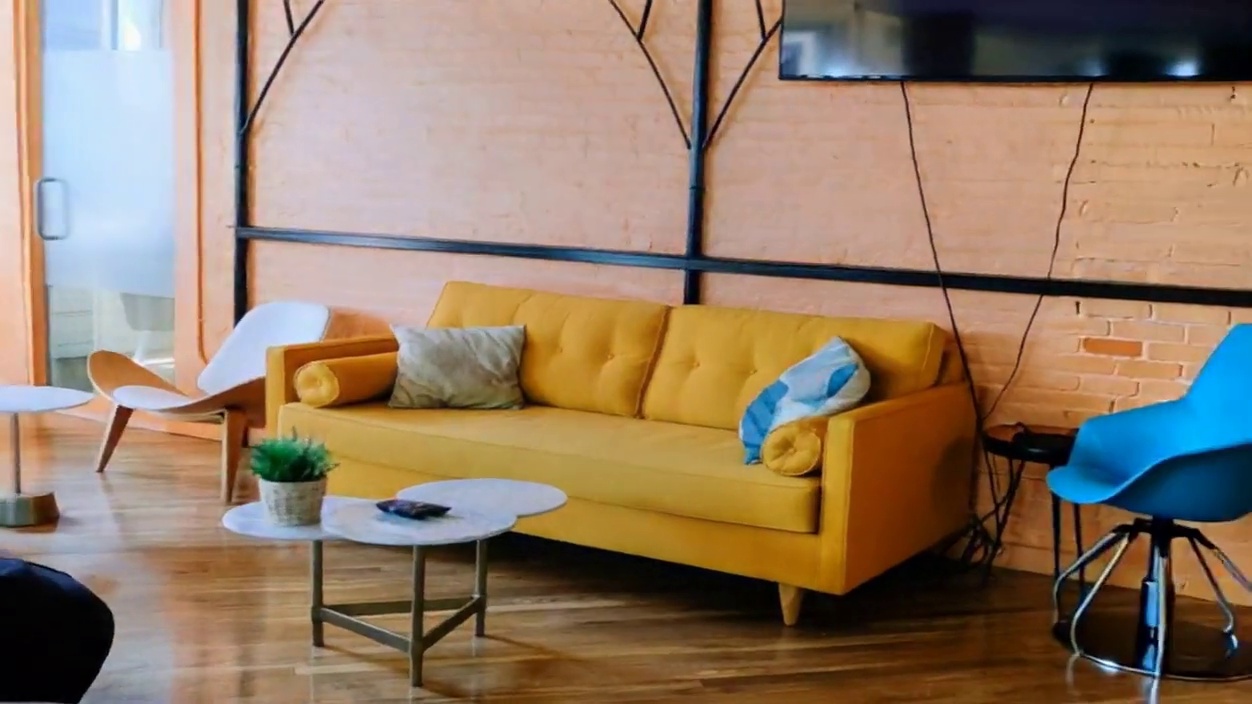}
    \end{subfigure}
    \hfill
    \begin{subfigure}[c]{0.185\textwidth}
        \centering
        \includegraphics[width=\textwidth]{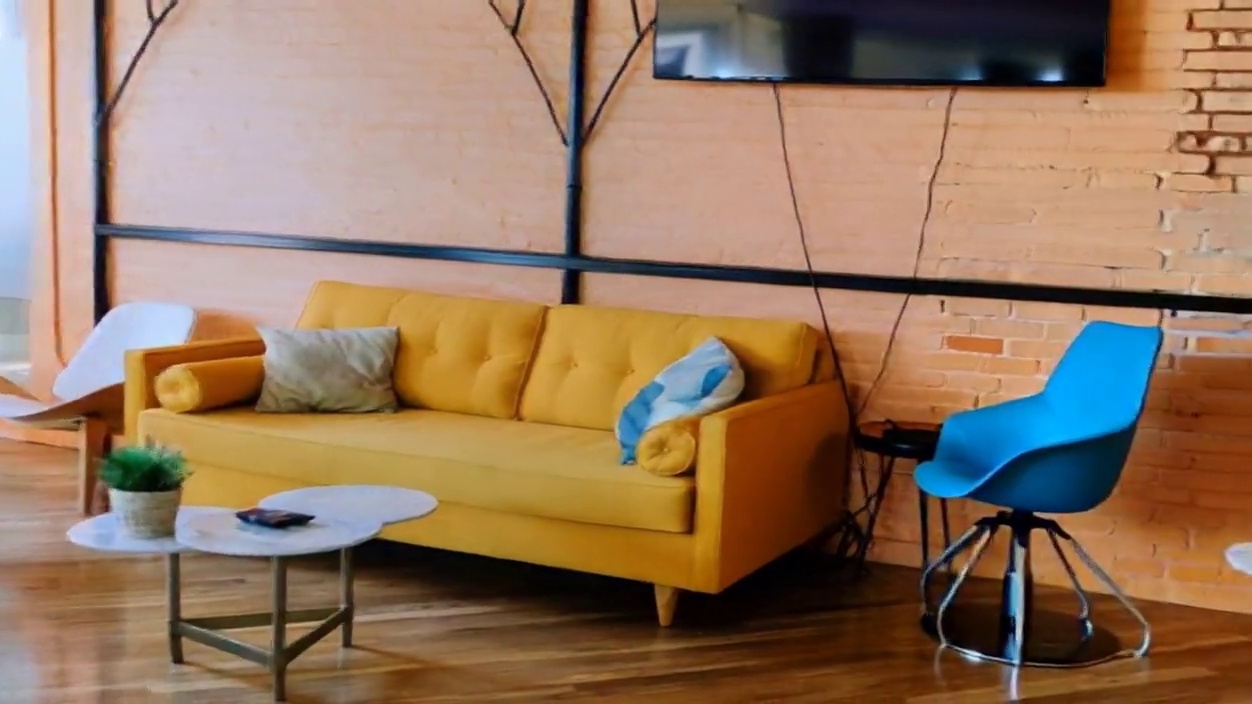}
    \end{subfigure}
    \hfill
    \begin{subfigure}[c]{0.185\textwidth}
        \centering
        \includegraphics[width=\textwidth]{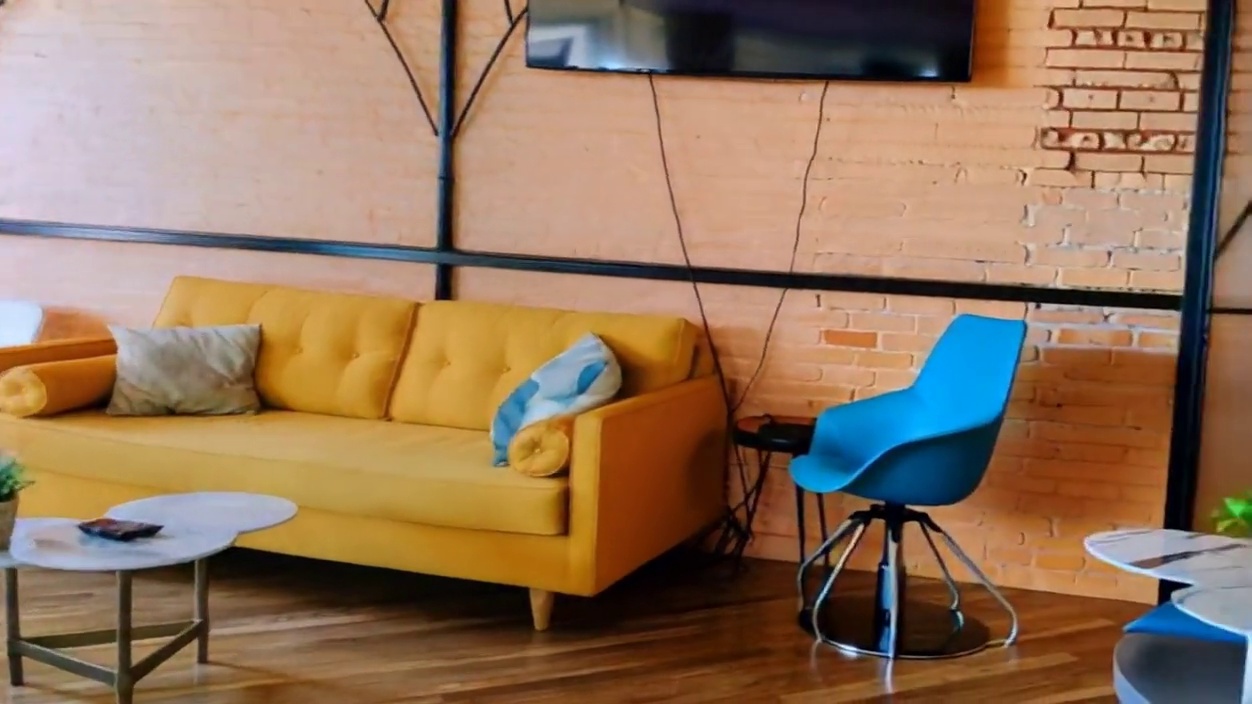}
    \end{subfigure}\\[10pt]

    \begin{subfigure}{\linewidth}
        \centering
        \textbf{Post-training: Robotic Manipulation}\\[2pt]
        \begin{tabular}{ccccc} %

        \includegraphics[width=0.196\textwidth]{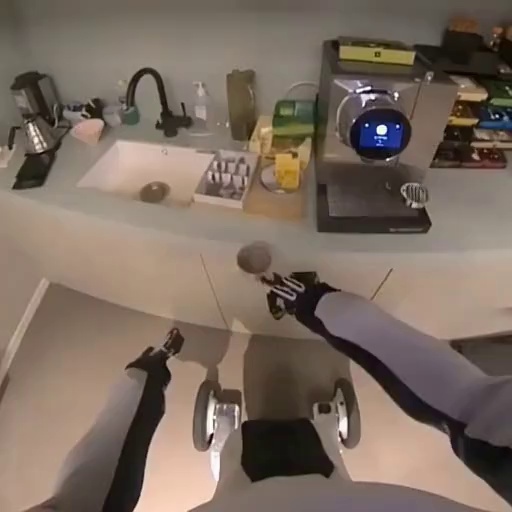} &
        \includegraphics[width=0.196\textwidth]{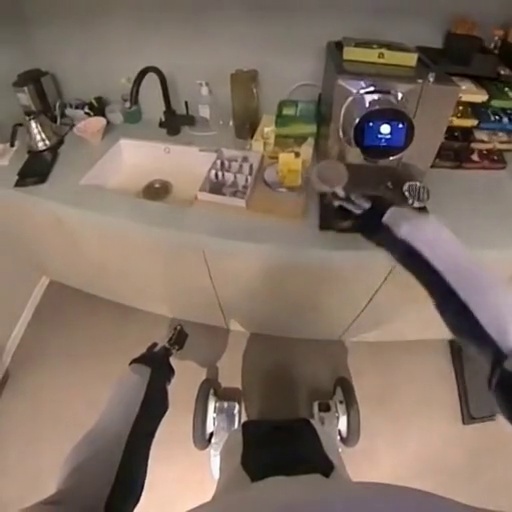} &
        \includegraphics[width=0.196\textwidth]{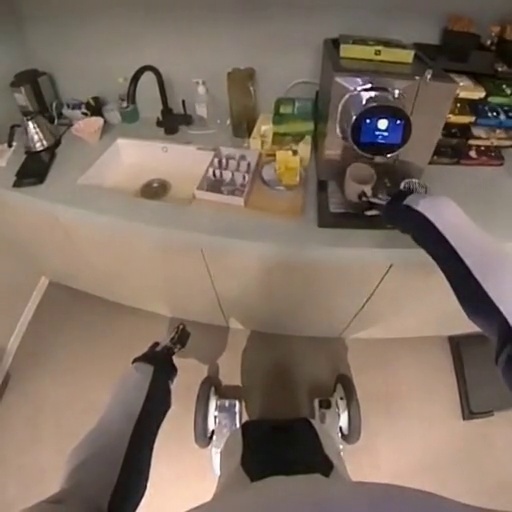} &
        \includegraphics[width=0.196\textwidth]{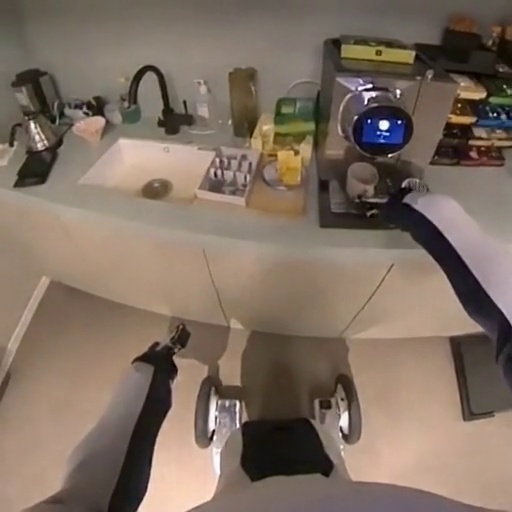} &
        \includegraphics[width=0.196\textwidth]{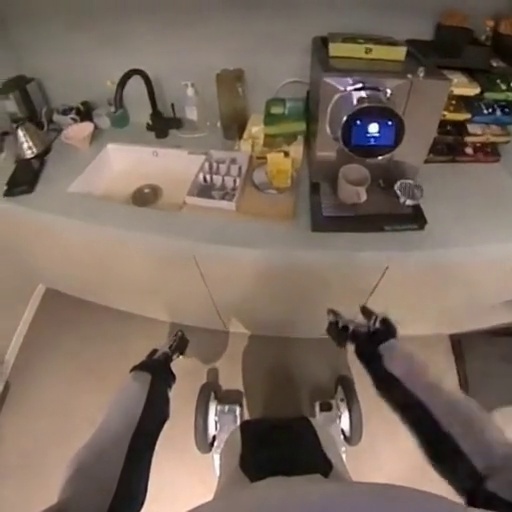}
        \\[2pt]
        \includegraphics[width=0.196\textwidth]{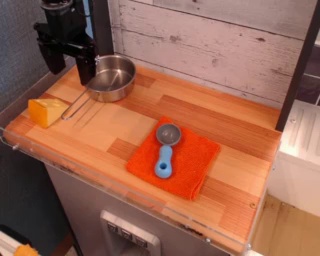} &
        \includegraphics[width=0.196\textwidth]{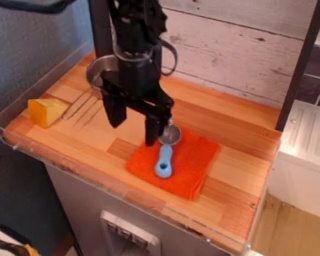} &
        \includegraphics[width=0.196\textwidth]{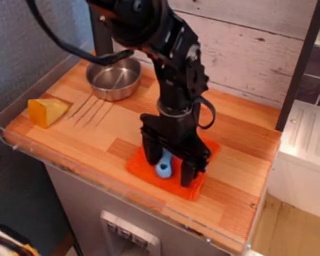} &
        \includegraphics[width=0.196\textwidth]{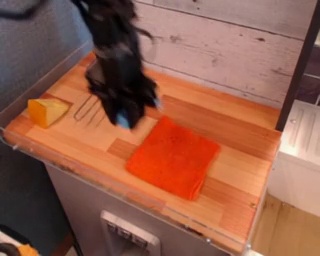} &
        \includegraphics[width=0.196\textwidth]{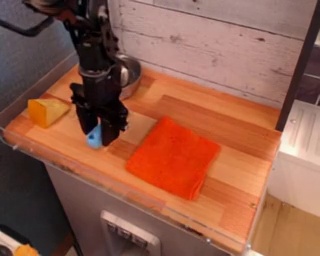}
        \\[10pt]
        \end{tabular}
    \end{subfigure}

    \begin{subfigure}{\linewidth}
        \centering
        \textbf{Post-training: Autonomous Driving}\\[2pt]
        \begin{tabular}{ccccc} %

        \includegraphics[width=0.196\textwidth]{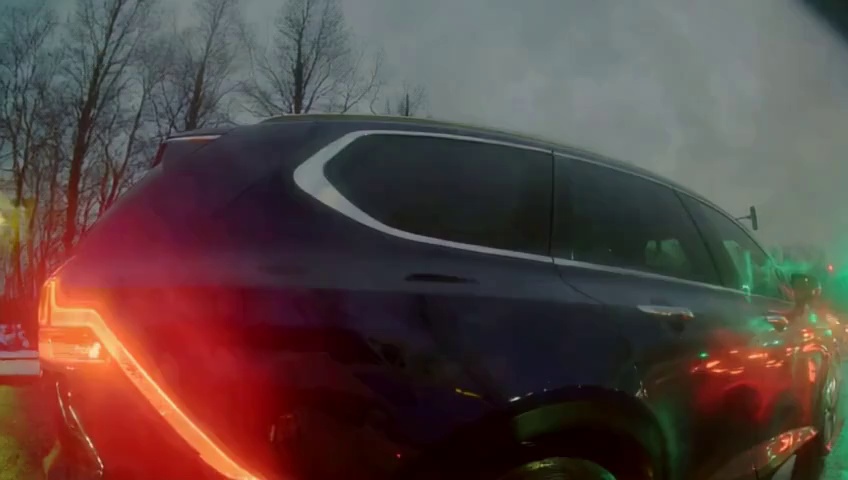} &
        \includegraphics[width=0.196\textwidth]{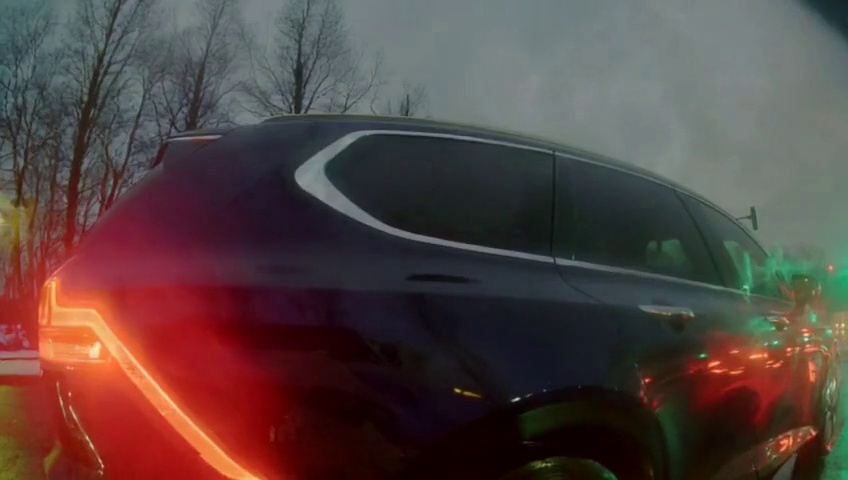} &
        \includegraphics[width=0.196\textwidth]{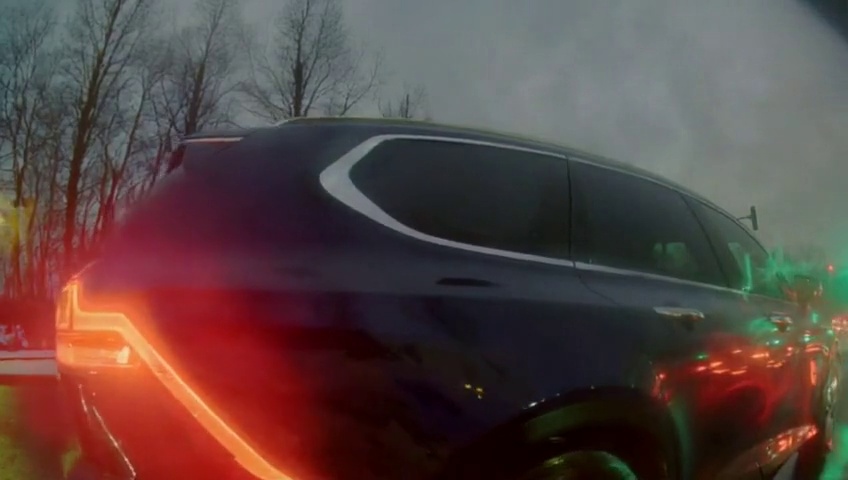} &
        \includegraphics[width=0.196\textwidth]{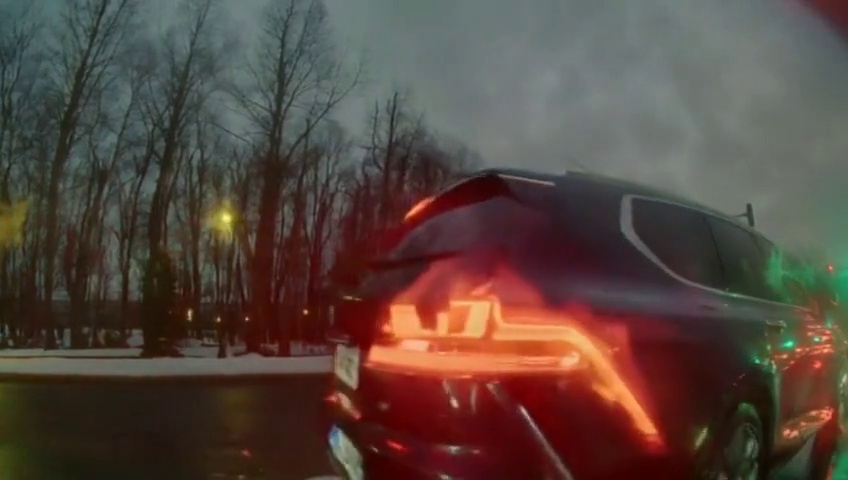} &
        \includegraphics[width=0.196\textwidth]{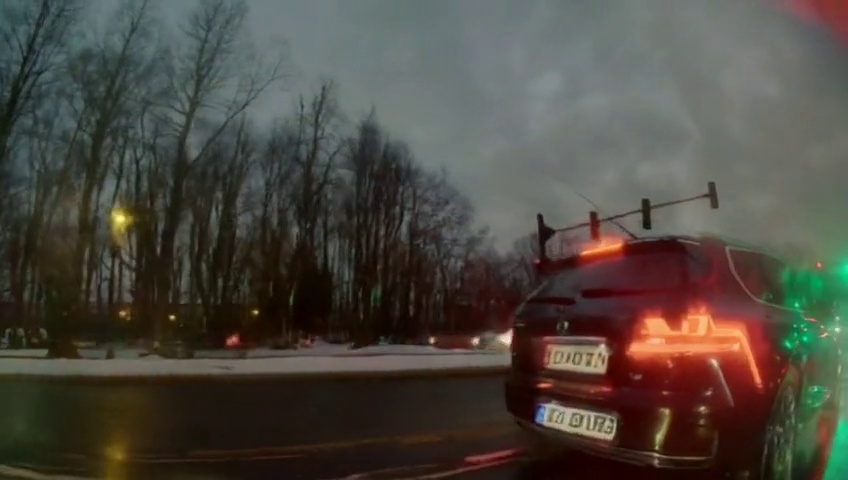}
        \\[2pt]
        \includegraphics[width=0.196\textwidth]{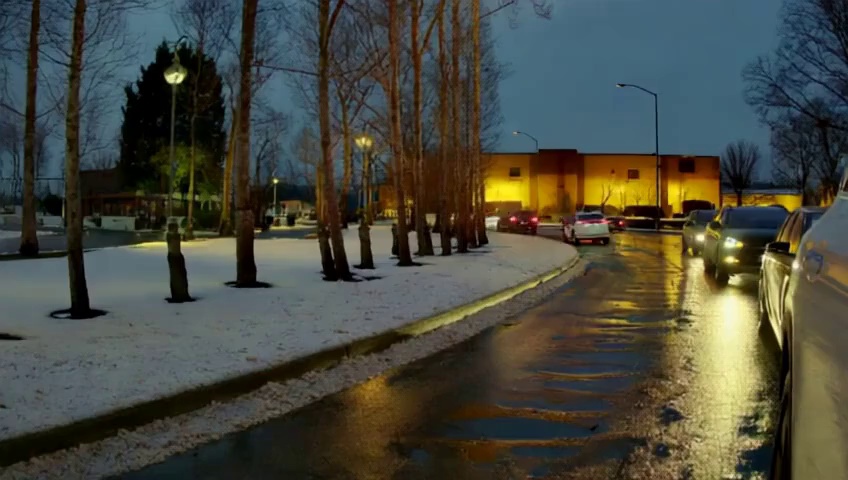} &
        \includegraphics[width=0.196\textwidth]{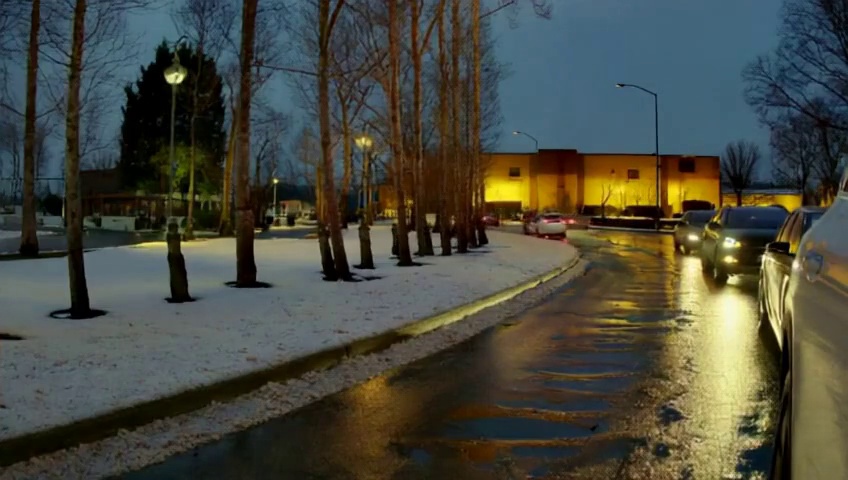} &
        \includegraphics[width=0.196\textwidth]{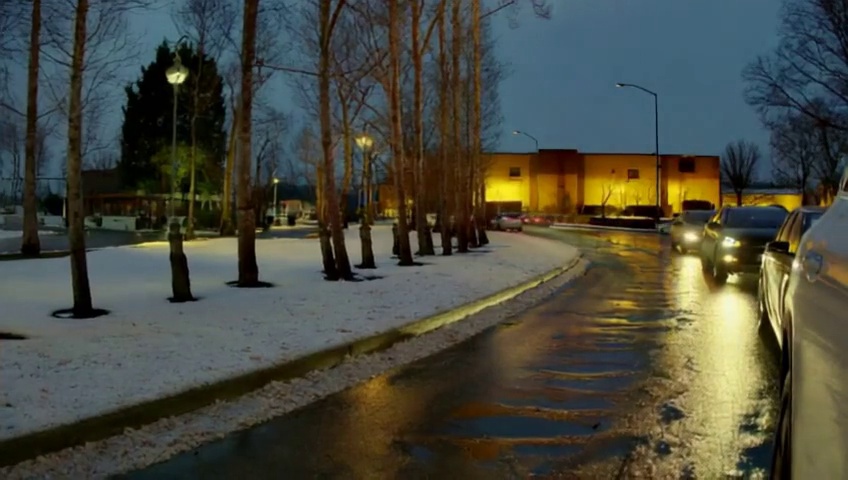} &
        \includegraphics[width=0.196\textwidth]{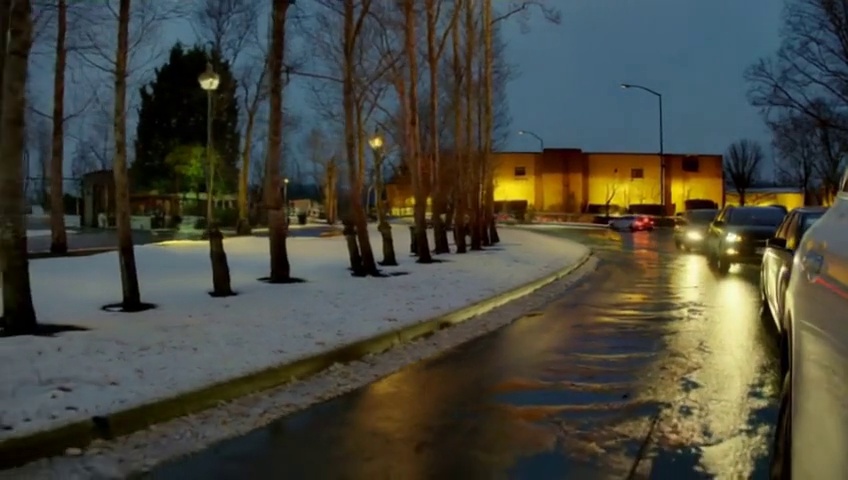} &
        \includegraphics[width=0.196\textwidth]{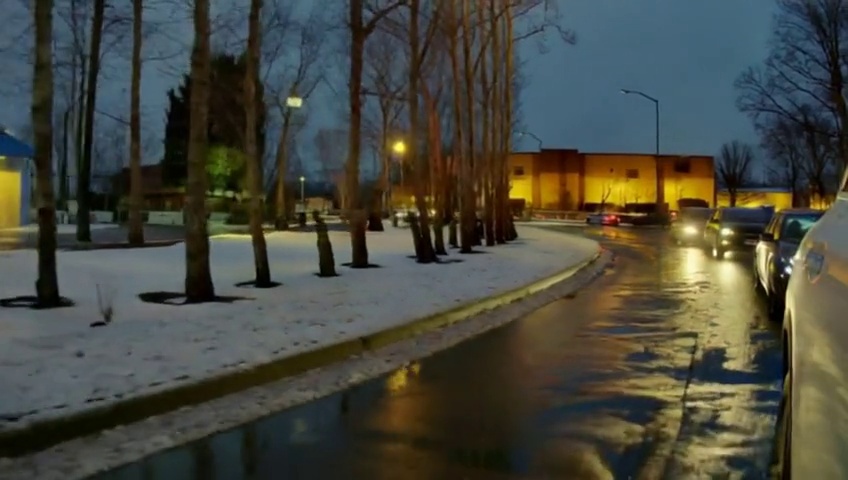} \\[4pt]
        \end{tabular}
    \end{subfigure}

    \caption{\textbf{Cosmos World Foundation Models}. Pre-trained Cosmos WFMs generate high-quality 3D consistent videos with accurate physics. The Cosmos suite of models includes both diffusion and autoregressive transformer models, which are trained using continuous and discrete latent representations of videos, respectively. Post-training these WFMs with specialized datasets enables them to be utilized in a wide range of Physical AI setups. Specifically, we present models with camera controllability, models capable of instruction-following for robotic manipulation, and models for autonomous driving scenarios. To check full videos and more video examples, please visit our \href{https://research.nvidia.com/labs/dir/cosmos1/}{website}.}
    \label{fig:teaser}
\end{figure*}

\begin{figure}[!ht]
\centering
    \includegraphics[width=0.99\textwidth]{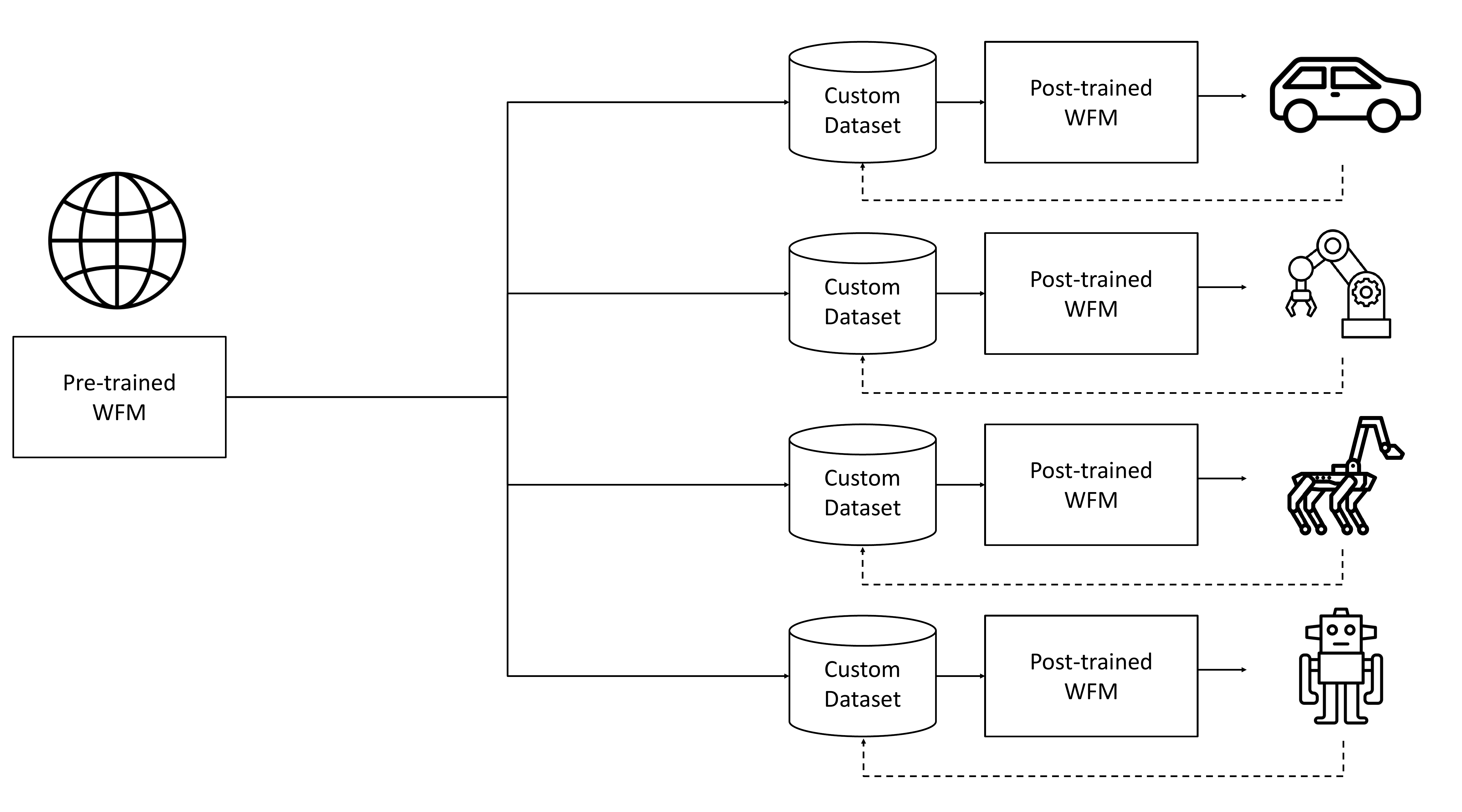}
    \caption{Pre-trained WFMs are world model generalists that are trained with large-scale, diverse video datasets capturing different aspects of real-world physics. These pre-trained world foundation models can be specialized to a target Physical AI setup through post-training. Usually, the datasets for post-training are ``prompt''-video pairs collected from the target Physical AI setup. The prompt can be in the form of action commands, trajectory, instructions, \etc. As the pre-trained WFM provides a great foundation, the dataset for post-training can be much smaller. This pre-training-and-post-training yields an efficient strategy for building a Physical AI system. In the figure, the dashed lines represent the data loop.}
    \label{fig:intro}
\end{figure}

In this paper, we introduce the Cosmos World Foundation Model (WFM) Platform for building Physical AI. We are mainly concerned with the visual world foundation model, where the observations are presented as videos, and the perturbations can exist in various forms. As illustrated in~\cref{fig:intro}, we present a pre-training-and-then-post-training paradigm, where we divide WFMs into pre-trained and post-trained WFMs. To build a pre-trained WFM, we leverage a large-scale video training dataset to expose the model to a diverse set of visual experiences to become a generalist. To build a post-trained WFM, we fine-tune the pre-trained WFM to arrive at a specialized WFM using a dataset collected from a particular Physical AI environment for the targeted, specialized Physical AI setup. \cref{fig:teaser} shows example results from our pre-trained and post-trained WFMs.

Data determines the ceiling of an AI model. To build a high-ceiling pre-trained WFM, we develop a video data curation pipeline. We use it to locate portions of videos with rich dynamics and high visual quality that facilitate learning of physics encoded in visual content. We use the pipeline to extract about 100M clips of videos ranging from 2 to 60 seconds from a 20M hour-long video collection. For each clip, we use a visual language model (VLM) to provide a video caption per 256 frames. Video processing is computationally intensive. We leverage hardware implementations of the H.264 video encoder and decoder available in modern GPUs for decoding and transcoding. Our video data curation pipeline leverages many pre-trained image/video understanding models. These models have different throughputs. To maximize the overall throughput for generating trainable video data, we build a Ray-based orchestration pipeline~\citep{moritz2017ray}. The details are described in~\cref{sec::curation}.

We explore two scalable approaches for building pre-trained WFMs discussed in~\cref{sec::pretrained_world_model}. These approaches are transformer-based diffusion models and transformer-based autoregressive models. A diffusion model generates videos by gradually removing noise from a Gaussian noise video. An autoregressive model generates videos piece by piece, conditioned on the past generations following a preset order. Both approaches decompose a difficult video generation problem into easier sub-problems, making it more tractable. We leverage state-of-the-art transformer architectures for their scalability. In~\cref{sec::diffusion_model}, we present a transformer-based diffusion model design that exhibits strong world-generation capabilities. In~\cref{sec::autoregress}, we present a transformer-based autoregressive model design for world generation.

Both the transformer-based diffusion model and transformer-based autoregressive model use tokens as representations of videos, where the former uses continuous tokens in the form of vectors, and the latter uses discrete tokens in the form of integers. We note that tokenization for videos---a process that transforms videos into a set of tokens---is highly nontrivial. Video contains rich information about the visual world. However, to facilitate learning of the WFMs, we need to compress videos into sequences of compact tokens while maximally preserving the original contents in the videos as the computation complexity of world foundation model training grows with the token counts. In many ways, building a video tokenizer is similar to building a video codec. We develop an attention-based encoder-decoder architecture to learn video tokenization for both continuous and discrete tokens described in~\cref{sec::token}.

We fine-tune the pre-trained WFMs to arrive at post-trained WFMs for various Physical AI tasks in~\cref{sec::posttrained_world_models}. In~\cref{sec::camera}, we fine-tune our pre-trained diffusion WFM to make it camera pose conditional. This post-training creates a navigable virtual world where users can explore the created world by moving the virtual viewpoint around. In~\cref{sec::robo}, we fine-tune our WFMs on various robotic tasks, which consist of video-action sequences. We show that by leveraging the pre-trained WFMs, we can better predict the future state of the world based on the action taken by the robot. In~\cref{sec::av}, we demonstrate how the pre-trained WFMs can be fine-tuned for various autonomous driving-related tasks.

Our intended use of the developed WFMs is for Physical AI builders.
To better protect the developers when using the world foundation models, we develop a powerful guardrail system that consists of a pre-Guard to block harmful inputs and a post-Guard to block harmful outputs. The details are described in~\cref{sec::guardrails}.

We aim to build a world foundation model platform to help Physical AI builders advance their systems. To achieve this goal, we make our pre-trained world foundation models and tokenizers available under the NVIDIA Open Model License at \href{https://github.com/nvidia-cosmos/cosmos-predict1}{NVIDIA Cosmos}. While this paper makes several improvements in world foundation model design, the world foundation model problem is still far from being solved. Additional research is required to advance the state-of-the-art further.

\section{World Foundation Model Platform}\label{sec::world}

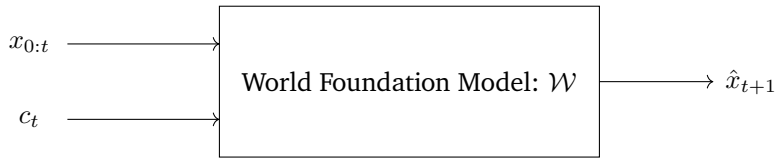
\begin{figure}[ht]
\centering
    \begin{tikzpicture}
        \node at (-3,1.5) {$x_{0:t}$};
        \draw (-0.5,0) rectangle (4.5,2);
        \node at (2,1) {World Foundation Model: $\mathcal{W}$};
        \node at (-3,0.5) {$c_t$};
        \node at (6.5,1) {$\hat{x}_{t+1}$};
        \draw[->] (-2.5,1.5) -- (-0.5,1.5); %
        \draw[->] (-2.5,0.5) -- (-0.5,0.5); %
        \draw[->] (4.5,1) -- (6.0,1); %
    \end{tikzpicture}
    \caption{A world foundation model (WFM) $\mathcal{W}$ is a model that generates the future state of the world $x_{t+1}$ based on the past observations $x_{0:t}$ and current perturbation $c_t$.}
    \label{fig:world_model}
\end{figure}

Let $x_{0:t}$ be a sequence of visual observations of the real world from time $0$ to $t$. Let $c_t$ be the perturbation to the world. As illustrated in~\cref{fig:world_model}, a WFM is a model $\mathcal{W}$ that predicts the future observation at time $t+1$, $\hat{x}_{t+1}$, based on the past observation $x_{0:t}$ and the current perturbation $c_t$. In our case, $x_{0:t}$ is an RGB video, while $c_t$ is a perturbation that can take many forms. It can be an action taken by the Physical AI, a random perturbation, a text description of the perturbation, etc.

\subsection{Future Cosmos}\label{sec::future_wfm}

We believe a WFM is useful to Physical AI builders in many ways, including (but not limited to)
\begin{itemize}
\item {\bf Policy evaluation.} This refers to evaluating the quality of a policy model in a Physical AI system. Instead of evaluating a trained policy by deploying it to a Physical AI system operating in the real world, one could instead let the digital copy of the Physical AI system interact with the world foundation model. The WFM-based evaluation is more cost-effective and time-efficient. With the WFM, builders can deploy the policy model in unseen environments that are otherwise unavailable. WFMs can help developers rule out incapable policies quickly and focus the physical resources on a few promising ones.

\item {\bf Policy initialization.} A policy model generates actions to be taken by the Physical AI system based on the current observations and the given task. A well-trained WFM, which models the dynamic patterns of the world based on the input perturbations, can serve as a good initialization of the policy model. This helps address the data scarcity problem in Physical AI.

\item {\bf Policy training.} A WFM paired with a reward model can be a proxy for the physical world to provide feedback to the policy model in a reinforcement learning setup. The agent can gain proficiency in solving tasks by interacting with the WFM.

\item {\bf Planning or model-predictive control.} A WFM can be used to simulate different future states following different action sequences taken by a Physical AI system. A cost/reward module can then be used to quantify the performance of these different action sequences based on the outcomes. The Physical AI can then execute the best action sequence based on the simulation results as a whole, as in planning algorithms or in a receding horizon manner, as in model-predictive control. The accuracy of the world model upper-bounds the performance of these decision-making strategies.

\item {\bf Synthetic data generation.} A WFM can be used to generate synthetic data for training. It can also be fine-tuned to be conditioned on rendering metadata such as depth or semantic maps. One can use the conditional WFM for the Sim2Real use case.
\end{itemize}

While we list the possibilities, this paper does not include empirical results in applying Cosmos WFMs to them. We are eager to verify the claims in future work.

\subsection{Current  Cosmos}\label{sec::current_wfm}
\begin{figure}[ht]
    \centering
    \includegraphics[width=0.99\textwidth]{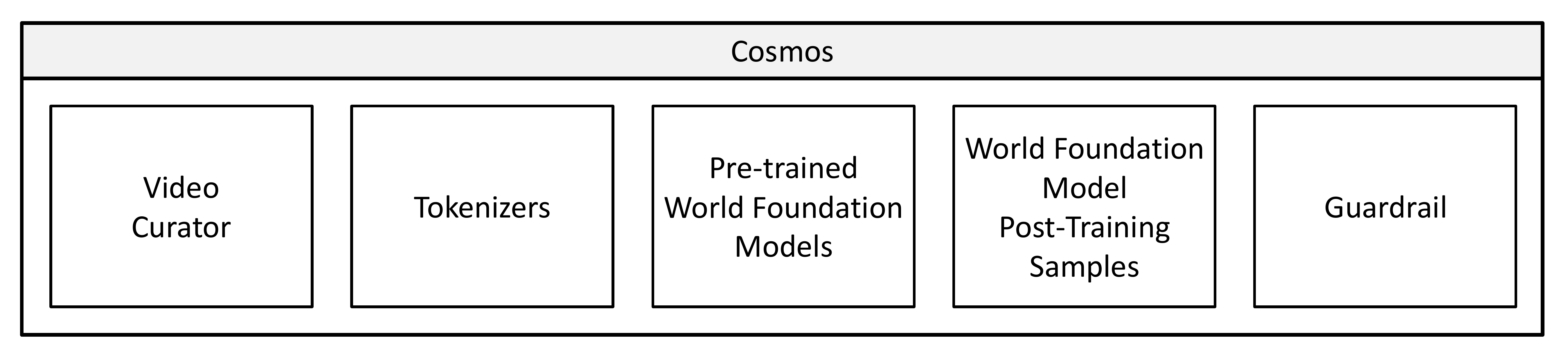}
    \caption{Cosmos World Foundation Model Platform consists of several major components: video curator, video tokenizer, pre-trained world foundation model, world foundation model post-training samples, and guardrail.}
    \label{fig:cosmos_platform}
\end{figure}

\cref{fig:cosmos_platform} visualizes what is available in the Cosmos WFM platform in this paper, including video curator, video tokenization, world foundation model pre-training, world foundation model post-training, and guardrail.

{\bf Video curator.} We develop a scalable video data curation pipeline. Each video is split into individual shots without scene changes. A sequence of filtering steps is then applied to the clips to locate high-quality and dynamic information-rich subsets for training. These high-quality shots are then annotated using a VLM. We then perform semantic de-duplication to construct a diverse but compact dataset.

{\bf Video tokenization.} We develop a family of video tokenizers of different compression ratios. These tokenizers are causal. The token computation for the current frames is not based on future observation. This causal design has several benefits. On the training side, it makes joint image and video training possible since a causal video tokenizer is also an image tokenizer when the input is a single image. This is important for the video model to leverage image datasets for training, which contain rich appearance information of the worlds and tend to be more diverse. On the application side, causal video tokenizers are better aligned with Physical AI systems that live in the causal world.

{\bf WFM pre-training.} We explore two scalable approaches for building pre-trained world foundation models---the diffusion model and the autoregressive model. We use the transformer architecture for its scalability.

For the diffusion-based WFM, the pre-training consists of two steps: 1) Text2World generation pre-training and 2) Video2World generation pre-training. Specifically, we train the model to generate a video world based on the input text prompt. We then fine-tune it to generate a future video world based on the past video and an input text prompt, which we refer to as the Video2World generation task.

For the autoregressive-based WFM, the pre-training consists of two steps: 1) vanilla next token generation and 2) text-conditioned Video2World generation. We first train the model to generate a future video world based on the input of past video---foresight generation. We then fine-tune it to generate a future video world based on the past video and a text prompt.

The video2world generation model is a pre-trained world model that generates the future based on the current observation (the past video) and control input (prompt). For both diffusion-based and autoregressive-based WFMs, we build a family of models with different capacities and study their effectiveness on various downstream applications.

We further fine-tune our pre-trained diffusion WFM to arrive at a diffusion decoder to enhance the generation results of the autoregressive model. To better control the WFM, we also built a prompt upsampler based on a Large Language Model (LLM).

{\bf World model post-training.} We show applications of the pre-trained WFMs on several downstream Physical AI applications. We fine-tune a pre-trained WFM with the camera pose as the input prompt. This allows us to navigate freely in the created world. We also demonstrate how our pre-trained WFMs might be fine-tuned for humanoid and autonomous driving tasks.

{\bf Guardrail.} For safe usage of the developed world foundation models, we develop a guardrail system where harmful inputs and outputs are blocked.

\section{Data Curation} \label{sec::curation}

\begin{figure}[ht]
    \centering
    \includegraphics[width=0.99\textwidth]{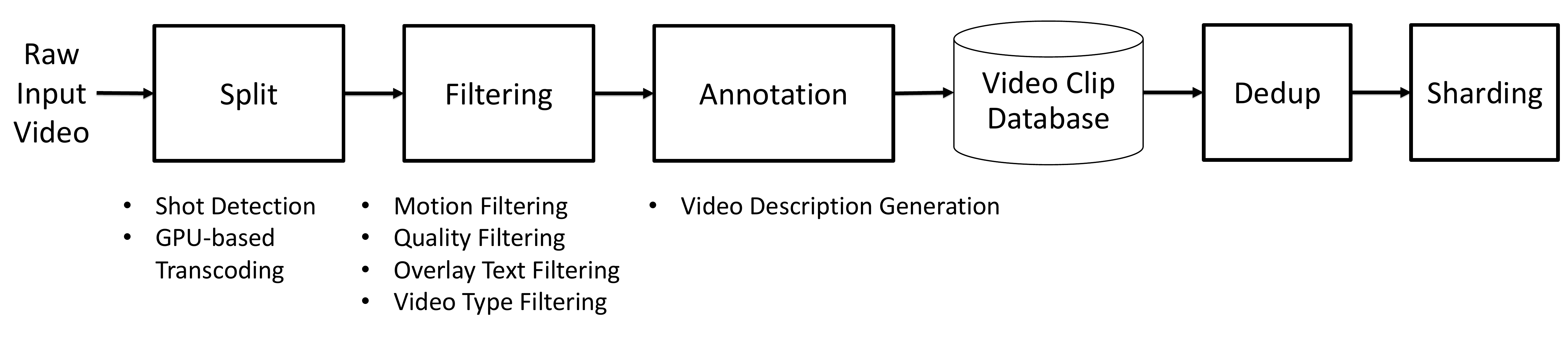}
    \caption{\textbf{Cosmos Video Curator contains five major steps: 1) split, 2) filtering, 3) annotation, 4) dedup, and 5) sharding}. The split step divides a long video into shots and transcribes them into clips. The filtering step removes clips that are of little value to world foundation model building. The annotation step adds a video description to each clip. The clips are then stored in a video clip database. To obtain the training dataset, one first performs semantic deduplication and then shards the video clips based on their resolutions and aspect ratios.}
    \label{fig:video_pipeline}
\end{figure}

We describe our video curation pipeline, which produces high-quality training datasets for both tokenizers and WFMs. As shown in~\cref{fig:video_pipeline}, our pipeline consists of 5 main steps: 1) splitting, 2) filtering, 3) annotation, 4) deduplication, and 5) sharding. Every step is tailored to improve the data quality and accommodate the requirements of model training. We first present our raw dataset and then describe each step in detail.

\subsection{Dataset}

We use both proprietary video datasets and publicly available open-domain Internet videos to train our models. Our goal is to enable Physical AI developers. To this end, we curate the video training dataset to cover various Physical AI applications and target the following video categories:
\begin{enumerate}
    \item Driving (11\%),
    \item Hand motion and object manipulation (16\%),
    \item Human motion and activity (10\%),
    \item Spatial awareness and navigation (16\%),
    \item First person point-of-view (8\%),
    \item Nature dynamics (20\%),
    \item Dynamic camera movements (8\%),
    \item Synthetically rendered (4\%), and
    \item Others (7\%).
\end{enumerate}
These videos offer a broad coverage of different visual objects and actions. Their diversity improves the generalization of our WFMs and helps the models handle different downstream tasks. The unstructured nature of these videos and their sheer volume creates many challenges to processing them efficiently from both an algorithmic and an infrastructural perspective. The videos can be encoded with a wide variety of codecs and have different aspect ratios, resolutions, lengths, \etc. Many videos have also been post-processed or edited with different visual effects, which may induce unwanted artifacts in the generated videos and hurt the performance of the world models if not appropriately handled.

In total, we accumulate about 20M hours of raw videos with resolutions from 720p to 4k. However, a significant amount of the video data is either semantically redundant or does not contain useful information for learning the physics of the world. Hence, we design a sequence of data processing steps to find the most valuable parts of the raw videos for training. We also collect image data as joint-image-and-video training has been shown to improve the visual quality of the generated videos and accelerate the model training. Thanks to the modular design of our data curation pipeline, we can use it to process both image and video data and generate datasets for both pre-training and fine-tuning. We generate about $10^8$ video clips for pre-training and about $10^7$ for fine-tuning.

\subsection{Splitting}

Our videos have arbitrary lengths, and modern deep-learning models cannot directly consume very long videos. Also, many videos contain shot transitions. They can start from one scene and then transition to a different scene where the two scenes can be disconnected entirely, \eg, from two people talking in a modern kitchen in New York City to a scene of lions chasing zebra in an African savanna. It is important to segment each video based on its shot changes and generate visually consistent video clips so that the model can learn visual content transitions that are physically plausible instead of artificially edited.

\subsubsection{Shot Detection}

Splitting aims to temporally segment raw videos of arbitrary lengths into clips without shot changes. It takes the raw videos as input and generates each shot's start and end frame indices. Clips shorter than 2s are discarded, as they could be shot transitions or visual effects. Clips longer than 60s are further split to have a maximal length of 60s. The subsequent filtering steps can then determine whether a clip contains useful information for learning the physics of the world.

Shot boundary detection is a classical computer vision problem. Existing methods detect shot boundaries based on changes in the visual feature space, but they differ in how to learn visual features from video frames. We evaluate several algorithms for the task in ~\cref{tab:splitting_algorithm_metrics}: PySceneDetect \citep{PySceneDetect}, Panda70M \citep{chen2024panda70m}, TransNetV2 \citep{soucek2020transnetv2}, and AutoShot \citep{zhu2023autoshot}.

PySceneDetect is a popular library that detects shot changes by thresholding the temporal change of color histogram in HSV space. Note that it is also adopted by the recent MovieGen work~\citep{polyak2024movie}. Panda70M augments PySceneDetect with CLIP-embedding-based stitching and filtering. TransNetV2 and AutoShot, on the other hand, are neural network-based, predicting a probability of each frame being a transition frame given a 100-frame rolling input window.

It is critical to select an algorithm that can handle heavily edited videos well, as they often have complex shot changes compounded with various visual effects. This motivates us to build a dedicated benchmark to evaluate whether the method can generate clips with clean shot cuts from videos. Our benchmark (named ShotBench\footnote{ShotBench is available at \url{https://github.com/NVlabs/ShotBench}.}) includes existing datasets, such as RAI, BBC Planet Earth \citep{bbcplanetearthdataset}, ClipShots \citep{tang2018fastvideoshottransition} and SHOT \citep{zhu2023autoshot}. For ClipShots, we define the transition frame as the midpoint of the start and end of each shot annotation to be consistent with other datasets.

\begin{table}[ht]
    \setlength{\tabcolsep}{17.6pt} %
    \small %
    \captionsetup{justification=centering} %
    \caption{Comparison of splitting algorithms on different datasets.}
    \centering
    \vspace{-1em}
    \begin{tabular}{rccccc}
        \toprule
        Dataset & Metric & PySceneDetect & Panda70M & TransNetV2 & AutoShot \\
        \midrule
        BBC & Precision $\uparrow$ & 0.894 & 0.959 & 0.983 & \textbf{0.984} \\
        & Recall $\uparrow$ & 0.884 & 0.653 & \textbf{0.951} & 0.922 \\
        & F1 $\uparrow$ & 0.889 & 0.777 & \textbf{0.967} & 0.952 \\
        \midrule
        RAI & Precision $\uparrow$ & 0.856 & \textbf{0.933} & 0.918 & 0.889 \\
        & Recall $\uparrow$ & 0.807 & 0.746 & 0.921 & \textbf{0.923} \\
        & F1 $\uparrow$ & 0.831 & 0.829 & \textbf{0.919} & 0.906 \\
        \midrule
        SHOT & Precision $\uparrow$ & 0.769 & \textbf{0.949} & 0.883 & 0.866 \\
        & Recall $\uparrow$ & 0.673 & 0.462 & 0.767 & \textbf{0.804} \\
        & F1 $\uparrow$ & 0.718 & 0.622 & 0.821 & \textbf{0.834} \\
        \midrule
        ClipShots & Precision $\uparrow$ & 0.395 & 0.649 & \textbf{0.685} & 0.653 \\
        & Recall $\uparrow$ & 0.602 & 0.424 & 0.772 & \textbf{0.781} \\
        & F1 $\uparrow$ & 0.477 & 0.513 & \textbf{0.726} & 0.711 \\
        \bottomrule
    \end{tabular}
    \label{tab:splitting_algorithm_metrics}
\end{table}

\cref{tab:splitting_algorithm_metrics} compares the different methods on ShotBench. We set the confidence threshold to 0.4 for both TransNetV2 and AutoShot. For Panda70M, we follow their implementation for splitting, excluding the filtering steps, for a fair comparison. End-to-end learning-based approaches (\eg, TransNetV2 and AutoShot) perform much better than methods using hand-crafted features or heuristic rules (\eg, PySceneDetect and Panda70M). Though TransNetV2 and AutoShot perform comparably on existing datasets, we found TransNetV2 works better on more challenging shot changes. Using an end-to-end neural network (\ie, TransNetV2) also allows us to increase the throughput of splitting by leveraging modern GPUs for acceleration without the hurdle of hybrid approaches (such as Panda70M) that use complicated logic to combine PySceneDetect and ImageBind embeddings \citep{girdhar2023imagebind}.

\subsubsection{Transcoding}

Our videos use many different codecs with various settings, which poses challenges to data curation. We re-encode each video clip from shot detection into a consistent, high-quality mp4 format. This simplifies the subsequent data curation process. With a unified video codec, the stability and efficiency of our dataloader for model training are also greatly improved. We use the h264\_nvenc codec with a high bitrate and stress test our setting using videos with fast motion and high-frequency texture to ensure no perceptible visual degradation.

We thoroughly evaluate different hardware and software configurations for transcoding to maximize the throughput in~\cref{tab:transcoding_speed}. Modern GPUs provide hardware-accelerated video encoding and decoding capabilities. NVIDIA L40S has hardware accelerators for both decoding (NVDEC) and encoding (NVENC), whereas NVIDIA H100 only has NVDEC. We compensate H100 with the maximum available CPU cores (28 instead of 1) for a fair comparison with L40S in~\cref{tab:transcoding_speed}. L40S has about 17\% higher throughput than H100 (0.0674 \vs 0.0574). For software configurations, switching from libx264 to h264\_nvenc and transcoding multiple clips from the same video in batches significantly boost the throughput. We observe issues with ffmpeg fully utilizing NVDEC/NVENC accelerators, especially on multi-GPU nodes. Replacing ffmpeg with PyNvideoCodec for video stream transcoding leads to much higher accelerator utilization and the biggest throughput improvement (0.3702 \vs 0.1026). We only keep ffmpeg for audio remixing and use PyNvideoCodec to better leverage the computing power in the GPUs. We achieve a $\sim6.5\times$ increase in throughput when combining all the improvements together.

\begin{table}[ht]
    \small %
    \captionsetup{justification=centering} %
    \caption{Transcoding performance with different software settings.}
    \centering
    \vspace{-1em}
    \resizebox{\textwidth}{!}{
    \begin{tabular}{rcccccccc}
        \toprule
        Method & GPU & CPU & Codec & Batch & NVDEC (\#accelerator) & NVENC (\#accelerator)  & Throughput (videos/s) \\
        \midrule
        ffmpeg & H100 & 28 & libx264 & 1 & 7 & 0 & 0.0574 \\
        ffmpeg & L40S & 1 & h264\_nvenc & 1 & 3 & 3 & 0.0674 \\
        \midrule
        ffmpeg & L40S & 1 & h264\_nvenc & 16 & 3 & 3 & 0.1026 \\
        pynvc+ffmpeg & L40S & 1 & h264\_nvenc & 1 & 3 & 3 & \textbf{0.3702} \\
        \bottomrule
    \end{tabular}
    }
    \label{tab:transcoding_speed}
\end{table}

\subsection{Filtering}

The video clips produced from the splitting step are noisy, with vastly different qualities covering various topics. We design the filtering step to 1) remove video clips whose visual quality fails to meet our minimal requirements, 2) select high-quality video clips suitable for fine-tuning, and 3) tailor the data distribution for building WFMs. We achieve the above goal by doing motion filtering, visual quality filtering, text filtering, and video type filtering.

\subsubsection{Motion Filtering}

We have two main goals in motion filtering: 1) remove videos that are static or with random abrupt camera motion (usually from hand-held cameras) and 2) tag videos with different types of camera motion (\eg, pan, zoom, tilt, \etc), which can provide additional information to guide model training.

We build a lightweight classifier for motion filtering. The input to the classifier is a sequence of motion vectors or optical flow extracted from a video clip. The classifier is based on the ViT architecture and is trained with labeled videos. We experiment with motion vectors from h264 codec, the Farneback optical flow algorithm\citep{farneback2003two}, and an NVIDIA TensorRT-accelerated optical flow estimation network. We find that the classifier built on top of the NVIDIA TensorRT-accelerated optical flow estimation works the best, producing high classification accuracy for motion filtering.

\subsubsection{Visual Quality Filtering}

We consider two criteria, distortion and appearance quality, for visual quality-based filtering. First, we remove video clips with distortions, such as artifacts, noise, blur, low sharpness, overexposure, underexposure, \etc. We use a video quality assessment model trained on human-rated videos based on DOVER~\citep{wu2023exploring}. This gives a perceptual quality score per clip, and we use the scores to remove clips that are in the bottom $15\%$. Second, we filter out video clips with low appearance quality. We apply an image aesthetic model~\citep{schuhmann2022} on sampled frames from an input clip. We set a conservative threshold, \ie, $3.5$, since aesthetics are less important for Physical AI.

\subsubsection{Text Overlay Filtering}

Some of our videos are post-processed to add text to include additional information for the viewer. We also find that text tends to co-occur with different visual effects. Our goal is to learn the physics of the world. It is crucial to remove videos with such excessive text. Note that we focus on text added in post-processing instead of text in the original scene from which the video is created, such as the street names in driving videos.

We train an MLP-based binary classifier to detect such videos. The input to the classifier is a video embedding extracted using InternVideo2 \citep{wang2024internvideo2}. We use a proprietary VLM to build the training set to label positive and negative videos. Our trained model achieves high prediction accuracy in the validation set.

\subsubsection{Video Type Filtering}

To adjust the training data distribution and filter out unwanted video types, we design a comprehensive taxonomy that categorizes videos based on their content type and visual style. We train a classifier to label each video clip with categories from the taxonomy. We refine our data by excluding specific video types that could lead to poor generation quality or unrealistic dynamics, such as abstract visual patterns, video game footage, animated content, \etc. We further adjust the data distribution by upsampling from categories that are more relevant to WFMs (\eg, human action, human and object interaction, \etc) and downsampling on categories that are less important (\eg, nature or landscape videos).

Given the absence of pre-existing labeled datasets matching our taxonomy, we leverage a proprietary VLM to create training and evaluation data for the classifier. For each video clip, we prompt the VLM with eight uniformly sampled frames and query for the most appropriate taxonomy label. Using the annotated data, we train an MLP classifier on the same InternVideo2 embeddings from text filtering.

\subsection{Annotation}

Text descriptions are usually paired with image and video data to provide supervision and conditions for world model training. We use a VLM to generate high-quality and consistent captions for each video clip. We configure the VLM in a way such that it focuses on the material facts and details in the videos. Using this approach to provide descriptions of videos instead of relying on Alt text also eases the burden of learning for world models as we do not need to adapt to different text styles or formats during training.

We test several SOTA methods (\ie, VFC \citep{ge2024visual}, Qwen2-VL \citep{wang2024qwen2}, VILA \citep{lin2024vilapretrainingvisuallanguage, xue2024longvilascalinglongcontextvisual}) for caption generation on our videos, and find VILA generates more accurate descriptions based on a small-scale human evaluation. We use an internal VILA model with 13B parameters, fine-tuned for video captioning. It has an enlarged context window suitable for processing long, multi-frame contexts, with a max input and output token length of 5904 and 256, respectively. To improve the inference efficiency, we use an FP8-quantized TensorRT-LLM engine, resulting in a 10 $\times$ speed-up in throughput compared to a PyTorch half-precision baseline, as shown in \cref{tab:vila_captioning_speed}. We prompt VILA with ``\textit{Elaborate on the visual and narrative elements of the video in detail}'' and feed it 8 uniformly sampled frames from the input clip. The average length of captions is 559 characters or 97 words.

\begin{table}[ht]
    \setlength{\tabcolsep}{18.8pt} %
    \small
    \captionsetup{justification=centering}
    \caption{Inference throughput comparison of VILA on a single H100 GPU.}
    \centering
    \vspace{-1em}
    \begin{tabular}{rcccc}
        \toprule
        Engine & Precision & Batch Size & Throughput (clips/s) & Throughput (tokens/s)  \\
        \midrule
        PyTorch & FP16 & 1 & 0.21 & 49.6 \\
        TRT-LLM & FP16 & 1 & 0.40 & 95.6 \\
        \midrule
        TRT-LLM & FP16 & 16 & 1.09 & 260.9 \\
        TRT-LLM & FP8 & 16 & 1.96 & 470.6 \\
        \bottomrule
    \end{tabular}
    \label{tab:vila_captioning_speed}
\end{table}

\subsection{Deduplication}
Given the sheer volume of our videos, there could be duplicated or near-duplicated samples in the training set. It is critical to deduplicate the data to create a more balanced and diverse data distribution. It also improves the efficiency of training and reduces the chance of memorizing specific training samples.

We adopt the approach from SemDeDup~\citep{abbas2023semdedup} and DataComp~\citep{gadre2023datacomp} for scalable semantic deduplication. We reuse the InternVideo2 embeddings computed during filtering and cluster the embeddings using a multi-node GPU-accelerated implementation of k-means~\citep{RAPIDS} with $k=10,000$. We compute the pairwise distances within each cluster of embeddings to identify duplicates. When duplicated videos are detected, we choose the video with the highest resolution to ensure no quality is lost due to deduplication. To avoid storing the entire pairwise distance matrix in GPU memory, we calculate on-the-fly the necessary upper-triangular matrix and argmax reduction in blocks of 256. We remove about $30\%$ of training data during deduplication.

We also leverage the extracted InternVideo2 embeddings and clustering results to build a visual search engine that supports querying the whole training dataset with free-form text and videos. The search engine is useful for debugging issues in our data and understanding the gap between the pre-training dataset and downstream applications.

\subsection{Sharding}

This step aims to package the processed video clips into webdatasets that our model trainer can directly consume for training. We shard the videos based on their resolution, aspect ratio, and length to align with our training curriculum. Besides pre-training datasets, we also create fine-tuning datasets with even higher quality by leveraging the different filters described above.

\subsection{Infrastructure}

Our data processing infrastructure uses AnyScale Ray~\citep{moritz2017ray} to implement a streaming pipeline system for geographically distributed clusters, addressing two key challenges in large-scale ML workflows: efficient resource utilization across homogeneous nodes and robust operation over high-latency connections to data sources. By decoupling data transfer from computation, pipelines operate efficiently with remote data storage while maintaining memory requirements that scale with pipeline complexity rather than dataset size, enabling unbounded stream processing.

Our architecture enables concurrent utilization of complementary hardware resources through parallel pipeline stages, for instance, simultaneously using network bandwidth for data ingestion, NVDEC units for video decoding, and GPUs for compute-intensive transformations. We extend the Fragmentation Gradient Descent algorithm \citep{weng2023beware} to optimize this multi-resource allocation, with our scheduler automatically scaling individual stages to maintain balanced throughput across specialized hardware accelerators.

\section{Tokenizer}\label{sec::token}

\begin{figure}[ht]
    \centering
    \includegraphics[width=0.99\textwidth]{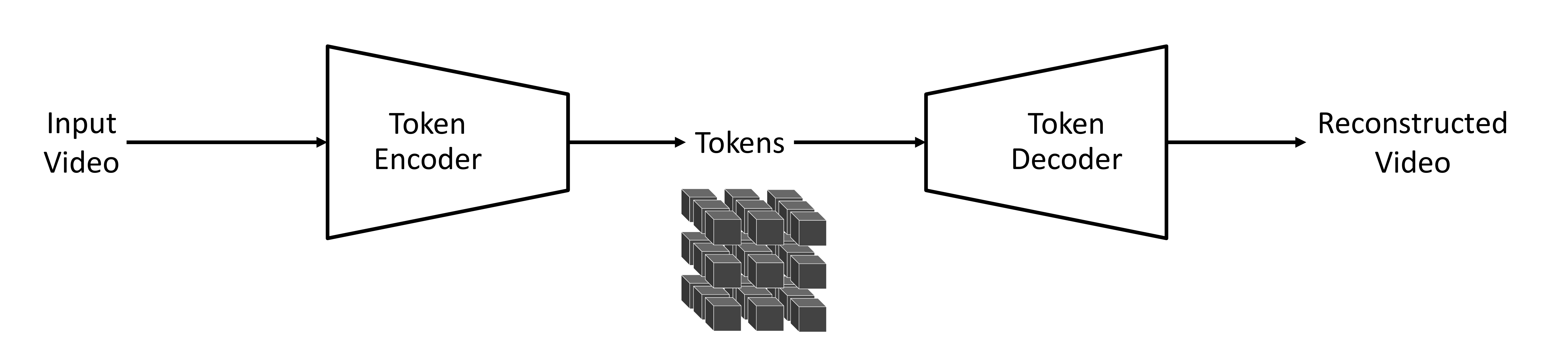}
    \caption{\textbf{Video tokenization pipeline}. An input video is encoded into tokens, which are usually much more compact than the input video. The decoder then reconstructs the input video from the tokens. Tokenizer training is about learning the encoder and decoder to maximally preserve the visual information in the tokens.}
    \label{fig:tokenization_flowchart}
\end{figure}

Tokenizers are fundamental building blocks of modern large-scale models. They transform raw data into more efficient representations by learning a bottle-necked latent space discovered in an unsupervised manner. Specifically, visual tokenizers map raw and redundant visual data---such as images and videos---into compact semantic tokens, making them crucial for handling high-dimensional visual data. This ability not only enables efficient training of large-scale transformer models but also democratizes their inference on limited computational resources. \cref{fig:tokenization_flowchart} schematically illustrates the tokenization training pipeline where the goal is to train the encoder and decoder so that the bottleneck token representation maximally preserves visual information in the input.

\begin{figure}[ht]
    \centering
    \begin{subfigure}{0.3\textwidth}
        \centering
        \includegraphics[width=\textwidth]{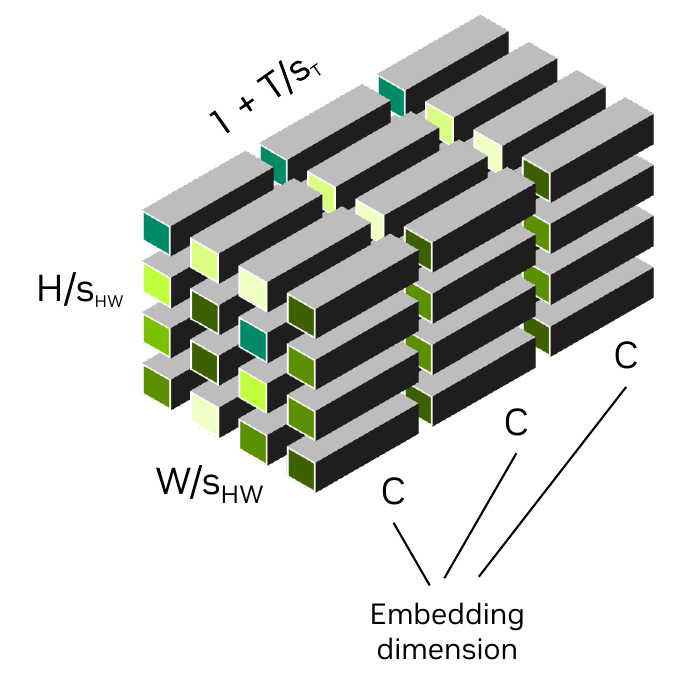}
        \caption{\centering Continuous tokens}
        \label{fig:continuous_tokens}
    \end{subfigure}%
    \hspace{0.15\textwidth} %
    \begin{subfigure}{0.3\textwidth}
        \centering
        \includegraphics[width=\textwidth]{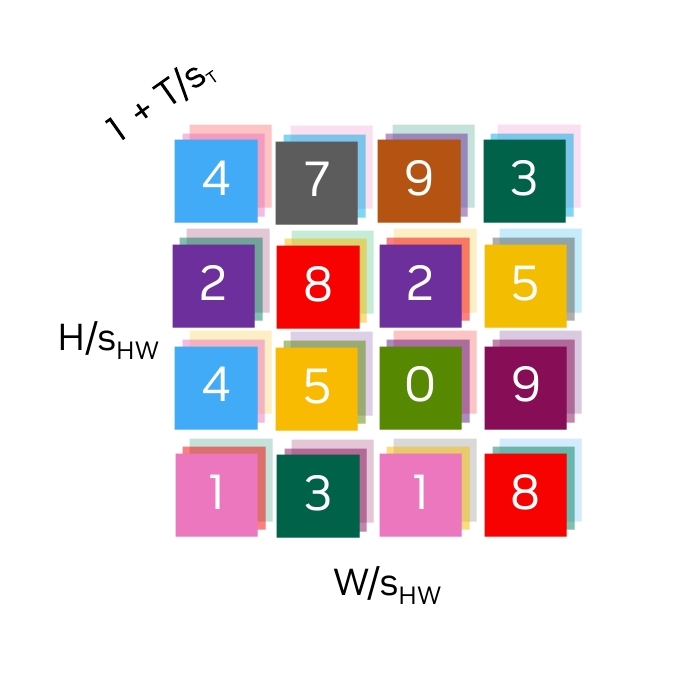}
        \caption{\centering Discrete tokens}
        \label{fig:discrete_tokens}
    \end{subfigure}
    \caption{\textbf{Visualization of continuous and discrete tokenizers}. Tokens along spatial ($\frac{H}{s_{HW}} \times \frac{W}{s_{HW}}$) and temporal ($1+\frac{T}{s_{T}}$) dimensions, with a spatial compression factor of $s_{HW}$ and a temporal compression factor of $s_{T}$. The first temporal token represents the first input frame, enabling joint image ($T=0$) and video ($T>0$) tokenization in a shared latent space. Left: Continuous latent embeddings with an embedding size of $C$. Right: Quantized indices, each color representing a discrete latent code.}

    \label{fig:continuous_vs_discrete}
\end{figure}

Tokenizers come in two types: continuous and discrete (see~\cref{fig:continuous_vs_discrete} for illustrations). Continuous tokenizers encode visual data into continuous latent embeddings, as in latent diffusion models like Stable Diffusion~\citep{rombach2022high} or VideoLDM~\citep{Blattmann2023Align}. These embeddings are suitable for models that generate data by sampling from continuous distributions. Discrete tokenizers encode visual data into discrete latent codes, mapping them into quantized indices, as seen in autoregressive transformers such as VideoPoet~\citep{videopoet}. This discrete representation is necessary for models such as GPT that are trained with the cross-entropy loss. \cref{fig:continuous_vs_discrete} illustrates the two types of tokens.

The success of tokenizers largely relies on their ability to deliver high compression rates without compromising their subsequent visual reconstruction quality. On one hand, high compression reduces storage and computational demands. On the other hand, excessive compression can lead to the loss of essential visual details. This trade-off presents a significant challenge in tokenizer design.

We present Cosmos Tokenizer, a suite of visual tokenizers that includes both continuous and discrete tokenizers for images and videos. Cosmos Tokenizer offers exceptional visual reconstruction quality and inference efficiency. It offers a range of compression rates to accommodate diverse computational constraints and application needs. \cref{tab:tokenizer_features_comparisons} presents a comparison of different visual tokenizers and their capabilities.

\begin{table}[ht]
    \small
    \captionsetup{justification=centering}
    \caption{Comparison of different visual tokenizers and their capabilities.
}
    \label{tab:tokenizer_features_comparisons}
    \vspace{-1em}
    \resizebox{1\linewidth}{!}{
    \begin{tabular*}{\linewidth}{@{\extracolsep{\fill}}l *{6}{c}}
    \toprule
    Model & Causal & Image & Video & Joint & Discrete & Continuous \\
    \midrule
    FLUX-Tokenizer~\citep{flux2024} & - & \textcolor{green}{\ding{51}} & \textcolor{red}{\ding{55}} & \textcolor{red}{\ding{55}} & \textcolor{red}{\ding{55}} & \textcolor{green}{\ding{51}} \\
    Open-MAGVIT2-Tokenizer~\citep{luo2024open_magvit2} & - & \textcolor{green}{\ding{51}} & \textcolor{red}{\ding{55}} & \textcolor{red}{\ding{55}} & \textcolor{green}{\ding{51}} & \textcolor{red}{\ding{55}} \\
    LlamaGen-Tokenizer~\citep{llamagen} & - & \textcolor{green}{\ding{51}} & \textcolor{red}{\ding{55}} & \textcolor{red}{\ding{55}} & \textcolor{green}{\ding{51}} & \textcolor{red}{\ding{55}} \\
    VideoGPT-Tokenizer~\citep{yan2021videogpt} & \textcolor{red}{\ding{55}} & \textcolor{red}{\ding{55}} & \textcolor{green}{\ding{51}} & \textcolor{red}{\ding{55}} & \textcolor{green}{\ding{51}} & \textcolor{red}{\ding{55}} \\
    Omni-Tokenizer~\citep{wang2024omnitokenizer} & \textcolor{red}{\ding{55}} & \textcolor{green}{\ding{51}} & \textcolor{green}{\ding{51}} & \textcolor{green}{\ding{51}} & \textcolor{green}{\ding{51}} & \textcolor{green}{\ding{51}} \\
    CogVideoX-Tokenizer~\citep{yang2024cogvideox} & \textcolor{green}{\ding{51}} &\textcolor{green}{\ding{51}}  &\textcolor{green}{\ding{51}} & \textcolor{green}{\ding{51}} & \textcolor{red}{\ding{55}} & \textcolor{green}{\ding{51}} \\
    \midrule
    \textcolor[HTML]{76B900}{\textbf{Cosmos-Tokenize1}}  & \textcolor{green}{\ding{51}} & \textcolor{green}{\ding{51}} & \textcolor{green}{\ding{51}} & \textcolor{green}{\ding{51}} & \textcolor{green}{\ding{51}} & \textcolor{green}{\ding{51}} \\
    \bottomrule
    \end{tabular*}
    }
\end{table}

We design Cosmos Tokenizer using a lightweight and computationally efficient architecture with a temporally causal mechanism. Specifically, we employ causal temporal convolution layers and causal temporal attention layers to preserve the natural temporal order of video frames, ensuring seamless tokenization of images and videos using a single unified network architecture.

We train our tokenizers directly on high-resolution images and long-duration videos without limiting the categories or aspect ratios. Unlike existing tokenizers that focus on specific data categories and sizes, the Cosmos Tokenizer operates across various aspect ratios—including 1:1, 3:4, 4:3, 9:16, and 16:9. They are temporally length-agnostic during inference, capable of tokenizing beyond the temporal length on which it was trained.

We also evaluate our tokenizers on standard image and video benchmarking datasets, including MS-COCO 2017~\citep{lin2015mscoco}, ImageNet-1K~\citep{deng2009imagenet}, and DAVIS~\citep{perazzi2016benchmark}. To facilitate the video tokenization study for Physical AI applications, we curate a video dataset that covers many video categories for Physical AI, ranging from fish-eye, robotics, driving, human activities, and spatial navigation. The dataset is available at \href{https://github.com/NVlabs/TokenBench}{github.com/NVlabs/TokenBench}.

\begin{figure}[ht]
    \centering
    \begin{subfigure}{0.49\textwidth} %
        \centering
        \includegraphics[width=\textwidth]{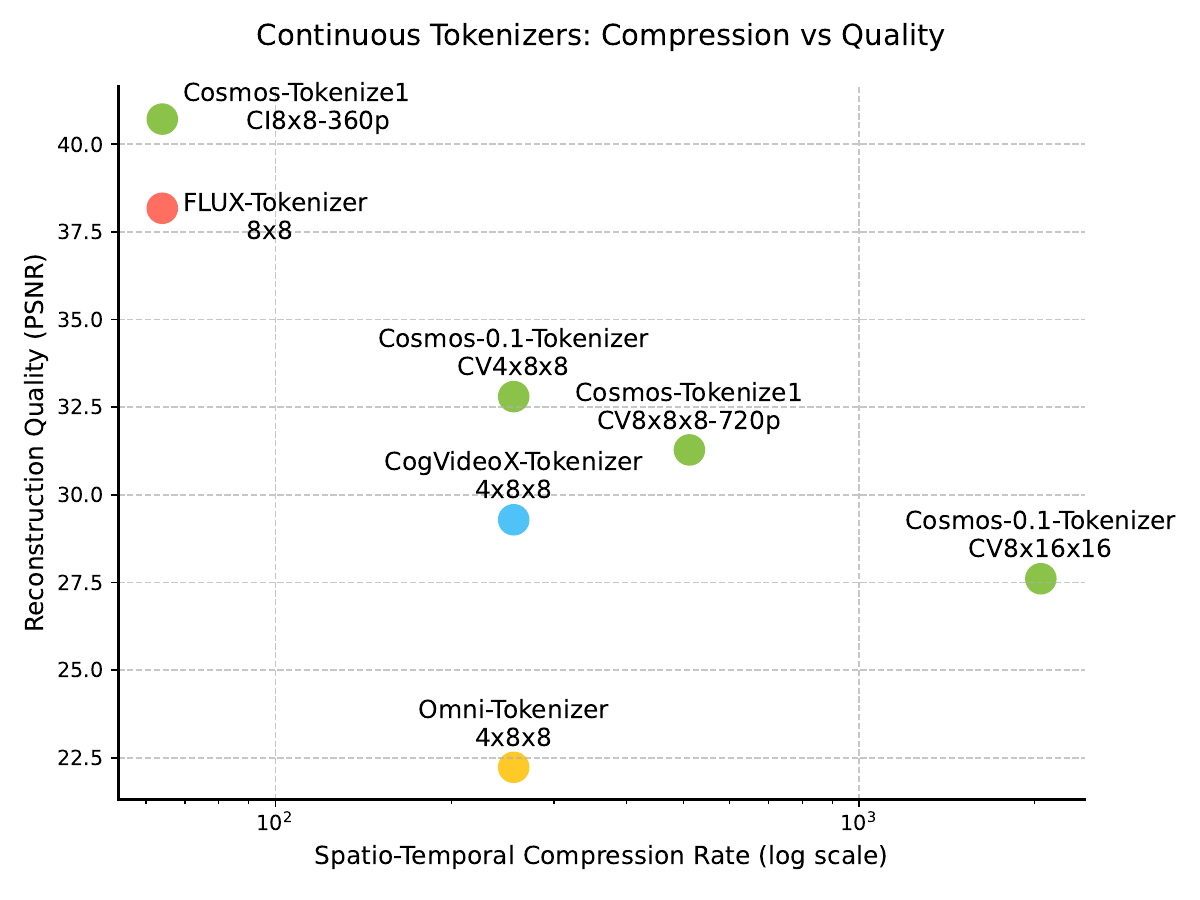}
        \caption{\centering Continuous tokenizers}
        \label{fig:continuous_tokens_compression}
    \end{subfigure}%
    \hspace{0.005\textwidth} %
    \begin{subfigure}{0.49\textwidth} %
        \centering
        \includegraphics[width=\textwidth]{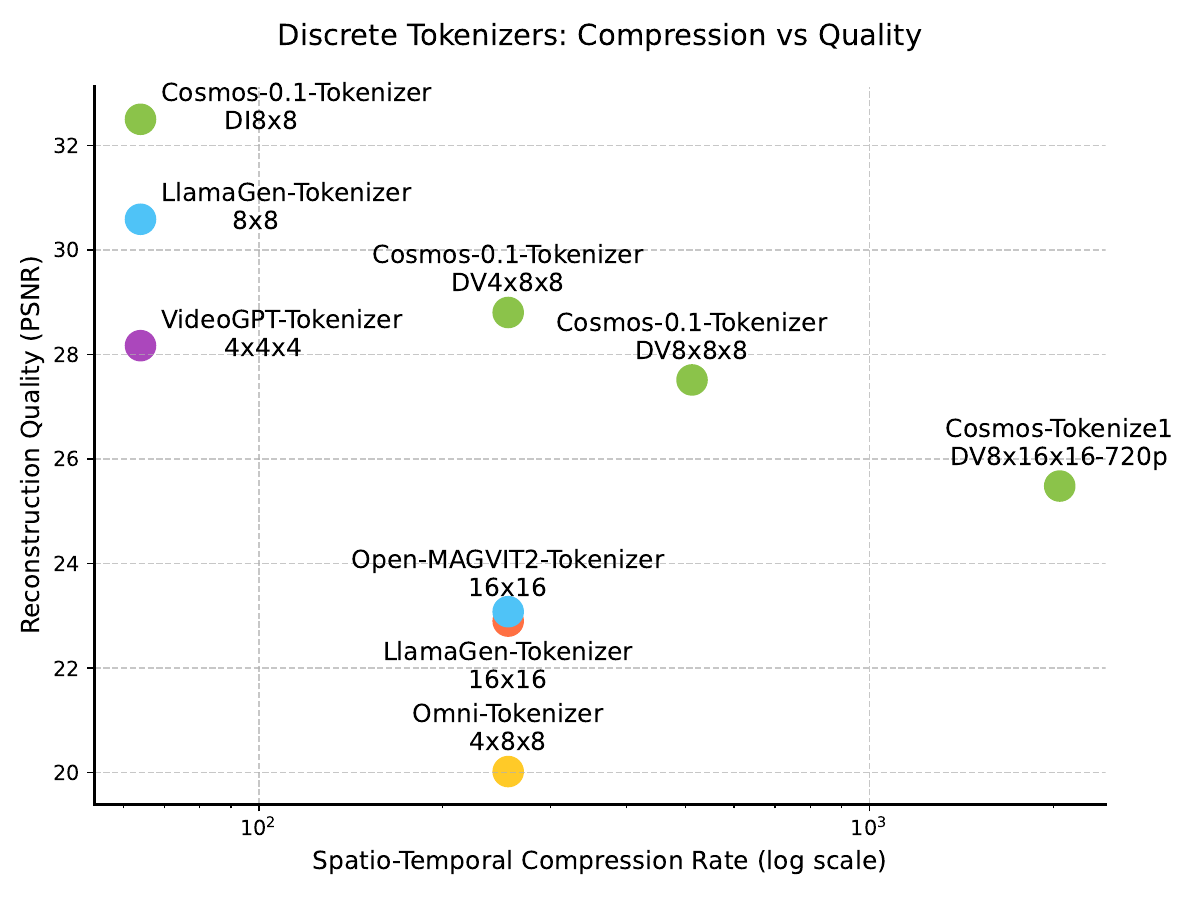}
        \caption{\centering Discrete tokenizers}
        \label{fig:discrete_tokens_compression}
    \end{subfigure}
    \caption{\textbf{Comparison of continuous (left) and discrete (right) tokenizers in terms of spatio-temporal compression rate (log scale) versus reconstruction quality (PSNR)}. Each solid point represents a tokenizer configuration, illustrating the trade-off between compression rate and quality. Notably, our tokenizer demonstrates an excellent compression-quality trade-off, delivering superior quality even at higher compression rates compared to other methods. The evaluation was performed on the DAVIS dataset. We calculate the PSNR of image tokenizers on all the individual frames.}
    \label{fig:compression_quality_scatter_plots}
\end{figure}

As shown in~\cref{fig:compression_quality_scatter_plots}, our evaluation results demonstrate Cosmos Tokenizer significantly outperforms existing tokenizers by a large margin---for instance, achieving a +4 dB PSNR improvement in reconstruction quality on DAVIS videos. It runs up to $12\times$ faster and can encode videos up to 8 seconds at 1080p and 10 seconds at 720p in one shot without running out of memory on a single NVIDIA A100 GPU with 80GB memory.

\subsection{Architecture}

Cosmos Tokenizer is designed as an encoder-decoder architecture. Given an input video $x_{0:T} \in \mathbb{R}^{(1+T) \times H \times W \times 3}$, with $H$, $W$, $T$ being the height, width, and number of frames, the encoder ($\mathcal{E}$) tokenizes the inputs into a token video $z_{0:T'} \in \mathbb{R}^{(1+T') \times H' \times W' \times C}$, with a spatial compression factor of $s_{HW} = \frac{H}{H'} = \frac{W}{W'}$ and a temporal compression factor of $ s_T = \frac{T}{T'}$. The decoder ($\mathcal{D}$) then reconstructs the input video from these tokens, resulting in the reconstructed video $\hat{x}_{0:T} \in \mathbb{R}^{(1+T) \times H \times W \times 3}$, mathematically given by:

\begin{equation}
    \hat{x}_{0:T} = \mathcal{D}\Big(\mathcal{E}\big(x_{0:T}\big)\Big).
\end{equation}

Our architecture employs a temporally causal design, ensuring that each stage processes only current and past frames, independent of future frames. Unlike common approaches, our tokenizer operates in the wavelet space, where inputs are first processed by a 2-level wavelet transform. Specifically, the wavelet transform maps the input video $x_{0:T}$ in a group-wise manner to downsample the inputs by a factor of four along $x$, $y$, and $t$. The groups are formed as: $\{x_0, x_{1:4}, x_{5:8}, ..., x_{(T-3):T} \}\rightarrow\{g_0, g_1, g_2, ..., g_{T/4}\}$. Subsequent encoder stages process the frames in a temporally causal manner as $\{g_0, g_{0:1}, g_{0:2}, ...\}\rightarrow\{\xi_0, \xi_1, \xi_2, ...\}$. Successive encoder stages follow a similar scheme, finally outputting the tokens $z_{0:T'}$. The causal design helps adapt models built on top of the tokenizer to downstream Physical AI applications that often operate on the temporal causal setting. The wavelet transform allows us to operate on a more compact video representation that eliminates redundancies in pixel information, allowing the remaining layers to focus on more semantic compression.

\begin{figure}[ht]
    \centering
    \begin{subfigure}{0.57\textwidth}
        \centering
        \includegraphics[width=\textwidth]{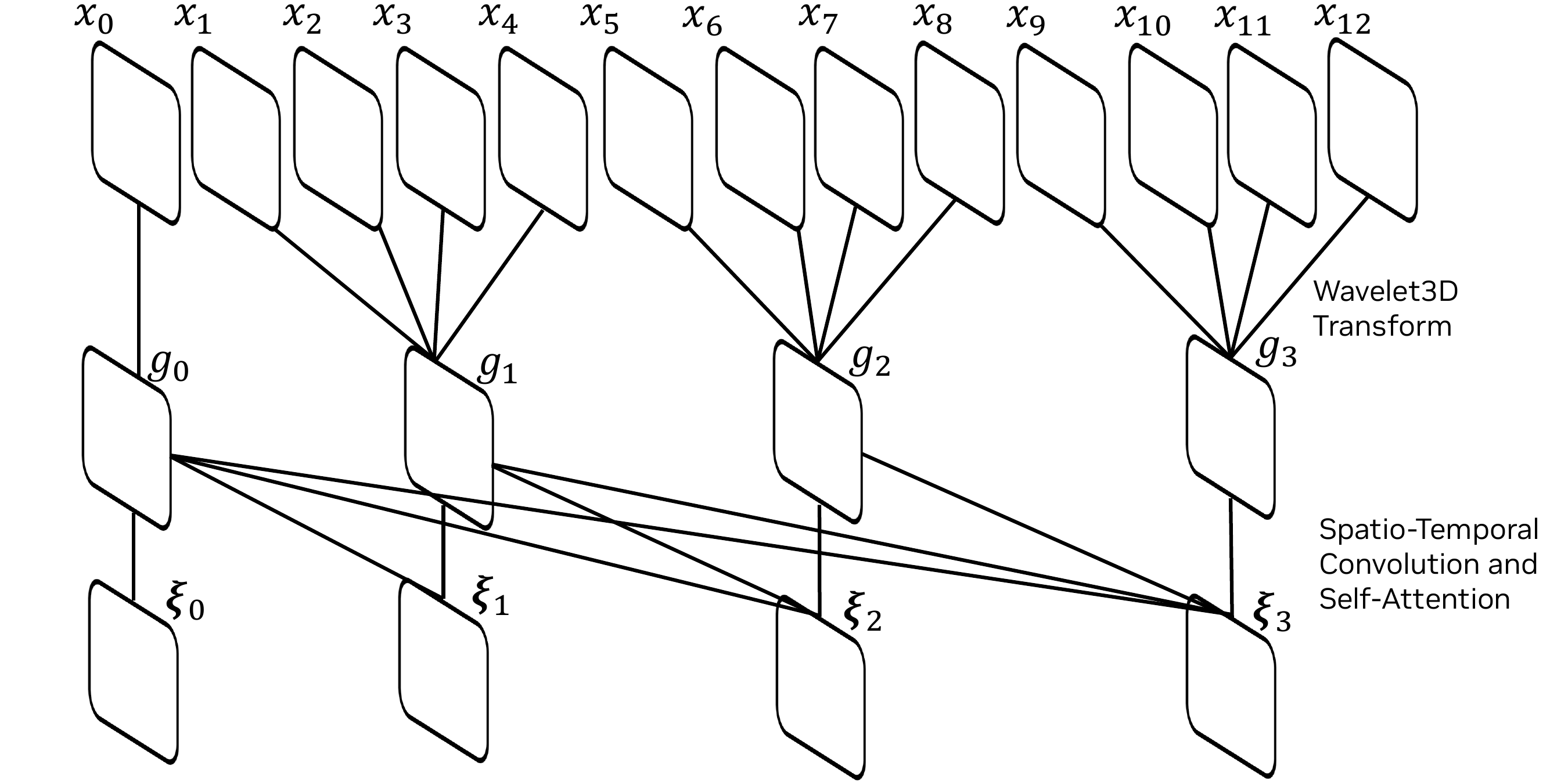}
        \caption{Temporal Causality: Illustration of the temporal causality mechanism, where inputs $x_0, x_1, \dots, x_{12}$ are processed through grouped intermediate outputs $g_0, g_1, \dots$, and further refined by spatio-temporal convolution and attention operations.
        }
        \label{fig:continuous_tokens_architecture}
    \end{subfigure}%
    \hspace{0.02\textwidth} %
    \begin{subfigure}{0.4\textwidth}
        \centering
        \includegraphics[width=\textwidth]{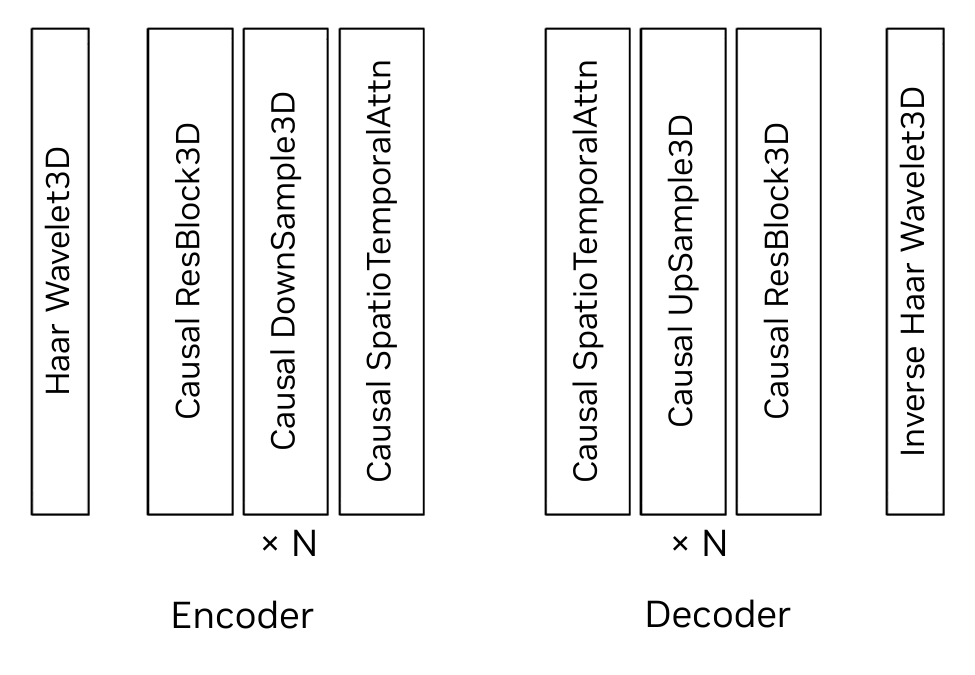}
        \caption{Network Architecture: The encoder-decoder network structure includes a 3D Haar wavelet, causal residual, causal downsampling, and causal spatio-temporal attention blocks. The decoder mirrors the encoder’s structure, replacing downsampling with upsampling.}
        \label{fig:discrete_tokens_architecture}
    \end{subfigure}
    \caption{\textbf{Overall Cosmos Tokenizer architecture illustrating the integration of temporal causality and an encoder-decoder structure}. Temporal causality (left) processes sequential inputs, while the encoder-decoder (right) leverages wavelet transforms and causal operations to capture spatial and temporal dependencies in the data.}
    \label{fig:network_architecture}
\end{figure}

Our encoder stages (post wavelet transform) are implemented using a series of residual blocks interleaved with downsampling blocks. In each block, we employ a spatio-temporal factorized 3D convolution, where we first apply a 2D convolution with a kernel size of $1 \times k \times k$ to capture spatial information, followed by a temporal convolution with a kernel size of $k \times 1 \times 1$ to capture temporal dynamics. We use left padding of $k-1$ to ensure causality. To capture long-range dependencies, we utilize a spatio-temporal factorized causal self-attention with a global support region---for instance, $1+T'$ for the last encoder block. We use the Swish activation function~\citep{ramachandran2017searching} for non-linearity. We leverage Layer Normalization (LayerNorm)~\citep{lei2016layer} instead of Group Normalization (GroupNorm)~\citep{wu2018group}, which prevents large magnitudes from appearing in specific regions of the latent space or reconstructed outputs~\citep{karras2020analyzing, sadat2024litevae}. The decoder mirrors the encoder, replacing the downsampling blocks with an upsampling block. \cref{fig:network_architecture} depicts an overview of the overall Cosmos Tokenizer architecture.

We employ the vanilla autoencoder (AE) formulation to model the continuous tokenizer's latent space. For discrete tokenizers, we adopt the Finite-Scalar-Quantization (FSQ)~\citep{mentzer2023finite} as the latent space quantizer. The latent dimension for the continuous tokenizers is $16$, whereas for the discrete tokenizers, it is $6$, which represents the number of the FSQ levels, which are $(8,8,8,5,5,5)$. This configuration corresponds to a vocabulary size of $64{,}000$.

\subsection{Training Strategy}

We employ a joint training strategy by alternating mini-batches of images and videos at a preset frequency. We only supervise the final output of our tokenizer's decoder. We do not use auxiliary losses tapped into the latent spaces, such as commitment or KL prior losses. For example, if a VAE~\citep{kingma2013auto} formulation were used for continuous tokenizers instead of the vanilla AE, one would need to have the KL prior loss. If a VQ-VAE~\citep{van2017neural} were used for discrete quantization instead of the FSQ, one would need to have the commitment loss.

We employ a two-stage training scheme. In the first stage, we optimize with the L1 loss that minimizes the pixel-wise RGB difference between the input and reconstructed video ($\hat{x}_{0:T}$), given by
\begin{equation}
    \mathcal{L}_1 = \left\lVert \hat{x}_{0:T} - x_{0:T} \right\rVert_1 ,
\end{equation}
and the perceptual loss based on the VGG-19 features~\citep{simonyan2014very}, given by,
\begin{equation}
    \mathcal{L}_{\text{Perceptual}} = \frac{1}{L} \sum_{l=1}^{L} \sum_{t} \alpha_l \left\lVert \texttt{VGG}_l (\hat{x}_t) - \texttt{VGG}_l (x_t) \right\rVert_1,
\end{equation}
where $\texttt{VGG}_l (\cdot)\in \R^{H \times W \times C}$ is the features from the $l$-th layer of a pre-trained VGG-19 network, $L$ is the number of layers considered, and $\alpha_{l}$ is the weight of the $l$-th layer.

\begin{figure}[ht]
    \centering
    \includegraphics[width=\textwidth]{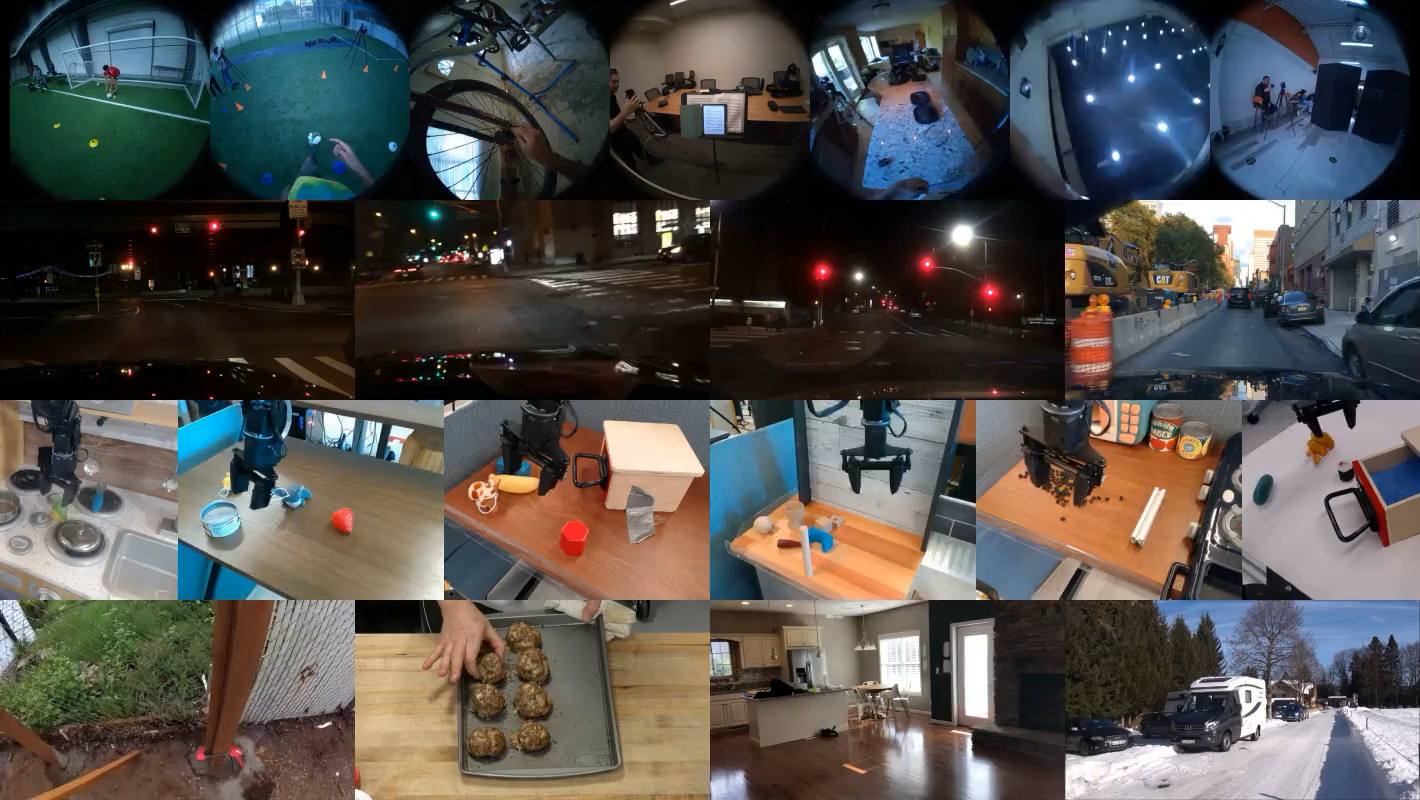}
    \caption{\textbf{Example videos from \emph{TokenBench}}. This figure shows diverse examples, including egocentric, driving, robotic manipulation, and web videos.}
    \label{fig:tokenbench}
\end{figure}

\begin{table}[t]
    \setlength{\tabcolsep}{4.82pt}
    \small
    \captionsetup{justification=centering}
    \caption{Evaluation of continuous video (CV) tokenizers on DAVIS and TokenBench.}
    \centering
    \vspace{-1em}
    \begin{tabular}{rcccccccc}
        \toprule
        &&& \multicolumn{3}{c}{DAVIS} & \multicolumn{3}{c}{TokenBench}
         \\
        \cmidrule(r){4-6} \cmidrule(r){7-9}
        Tokenizer & Frames  & Formulation & PSNR $\uparrow$ & SSIM $\uparrow$ & rFVD $\downarrow$ & PSNR $\uparrow$ & SSIM $\uparrow$ & rFVD $\downarrow$  \\
        \midrule
        CogVideoX-Tokenizer4$\times$8$\times$8 & 17 & VAE & 29.29 & 0.864 & 19.58 & 32.06 & 0.909 & 6.97 \\
        Omni-Tokenizer4$\times$8$\times$8 & 17 & VAE & 22.23 & 0.713 & 117.66 & 24.48 & 0.830 & 35.86 \\        
        Cosmos-0.1-Tokenizer-CV4$\times$8$\times$8 & 49 & AE & 32.80 & 0.900 & 15.93 & 35.45 & 0.928 & 6.85 \\        
        Cosmos-0.1-Tokenizer-CV8$\times$8$\times$8 & 49 & AE & 30.61 & 0.856 & 30.16 & 34.44 & 0.917 & 11.62 \\
        Cosmos-0.1-Tokenizer-CV8$\times$16$\times$16 & 49 & AE & 27.60 & 0.779 & 93.82 & 31.61 & 0.875 & 43.08 \\
        \midrule
        Cosmos-Tokenize1-CV4$\times$8$\times$8-360p & 49 & AE & \textbf{35.85} & \textbf{0.920} & \textbf{10.057} & \textbf{38.42} & \textbf{0.950} & \textbf{3.34} \\
        Cosmos-Tokenize1-CV8$\times$8$\times$8-720p & 121 & AE & 31.28 & 0.868 & 23.49 & 35.13 & 0.926 & 9.82 \\
        \bottomrule
    \end{tabular}
    \vspace{1em}
    \label{tab:token_res_CV}
    \small
    \captionsetup{justification=centering}
    \caption{Evaluation of discrete video (DV) tokenizers on DAVIS and TokenBench.}
    \centering
    \vspace{-1em}
    \setlength{\tabcolsep}{4.2pt}
    \begin{tabular}{rcccccccc}
        \toprule
        &&& \multicolumn{3}{c}{DAVIS} & \multicolumn{3}{c}{TokenBench} \\
        \cmidrule(r){4-6} \cmidrule(r){7-9}
        Tokenizer & Frames & Quantization & PSNR $\uparrow$ & SSIM $\uparrow$ & rFVD $\downarrow$ & PSNR $\uparrow$ & SSIM $\uparrow$ & rFVD $\downarrow$  \\
        \midrule
        VideoGPT-Tokenizer4$\times$4$\times$4 & - & VQ & 28.17 & \textbf{0.850} & 72.33  & 33.66 & \textbf{0.914} & \textbf{13.85} \\
        Omni-Tokenizer4$\times$8$\times$8 & 17 & VQ & 20.02 & 0.703 & 188.60  & 25.31 & 0.827 & 53.55 \\
        Cosmos-0.1-Tokenizer-DV4$\times$8$\times$8 & 17 & FSQ & 28.81 & 0.818 & \textbf{37.36} & 31.97 & 0.888 & 19.67 \\
        Cosmos-0.1-Tokenizer-DV8$\times$8$\times$8 & 17 & FSQ & 27.51 & 0.789 & 100.15 & 30.95	& 0.873 & 43.86\\
        Cosmos-0.1-Tokenizer-DV8$\times$16$\times$16 & 17 & FSQ & 25.09 & 0.714 & 241.52 & 28.91 & 0.829 & 113.48\\
        \midrule
        Cosmos-Tokenize1-DV4$\times$8$\times$8-360p & 49 & FSQ & \textbf{32.97} & 0.840 & 53.44 & \textbf{35.74} & 0.910 & 19.25\\
        Cosmos-Tokenize1-DV8$\times$16$\times$16-720p & 49 & FSQ & 25.49 & 0.719 & 259.33 & 29.33 & 0.838 & 107.43\\
        \bottomrule
    \end{tabular}
    \label{tab:token_res_DV}
\end{table}

\begin{table}[t]
    \setlength{\tabcolsep}{6.05pt}
    \small
    \captionsetup{justification=centering}
    \caption{Evaluation of continuous image (CI) tokenizers on various image datasets.}
    \centering
    \vspace{-1em}
    \begin{tabular}{rcccccccc}
        \toprule
        \multicolumn{3}{c}{} & \multicolumn{3}{c}{(a) MS-COCO 2017} & \multicolumn{3}{c}{(b) ImageNet-1K} \\
        \cmidrule(r){4-6} \cmidrule(r){7-9}
        Tokenizer & Height & Formulation & PSNR $\uparrow$ & SSIM $\uparrow$ & rFID $\downarrow$ & PSNR $\uparrow$ & SSIM $\uparrow$ & rFID $\downarrow$ \\
        \midrule
        FLUX-Tokenizer8$\times$8 & - & VAE & 24.00 & 0.682 & 2.501 & 20.09 & 0.518 & 1.229\\
        Cosmos-0.1-Tokenizer-CI8$\times$8 & 1024 & AE & 28.66 & \textbf{0.836} & \textbf{1.760} & 28.83 & \textbf{0.837} & \textbf{0.689}\\
        Cosmos-0.1-Tokenizer-CI16$\times$16 & 1024 & AE & 23.63 & 0.663 & 3.823 & 23.72 & 0.655 & 1.031\\
        \midrule
        Cosmos-Tokenize1-CI8$\times$8-360p & 1024 & AE & \textbf{32.79} & 0.824 & 1.874 & \textbf{32.91} & 0.824 & 0.785\\
        Cosmos-Tokenize1-CI16$\times$16-360p & 1024 & AE & 31.28 & 0.682 & 4.294 & 31.29 & 0.675 & 0.701\\
        \bottomrule
    \end{tabular}
    \label{tab:token_res_CI}
    \setlength{\tabcolsep}{5.55pt}
    \small
    \captionsetup{justification=centering}
    \vspace{2em}
    \caption{Evaluation of discrete image (DI) tokenizers on various image datasets.}
    \centering
    \vspace{-1em}
    \begin{tabular}{rcccccccc}
        \toprule
        \multicolumn{3}{c}{} & \multicolumn{3}{c}{(a) MS-COCO 2017} & \multicolumn{3}{c}{(b) ImageNet-1K} \\
        \cmidrule(r){4-6} \cmidrule(r){7-9}
        Tokenizer & Height & Quantization & PSNR $\uparrow$ & SSIM $\uparrow$ & rFID $\downarrow$ & PSNR $\uparrow$ & SSIM $\uparrow$ & rFID $\downarrow$ \\
        \midrule
        Open-MAGVIT2-Tokenizer16$\times$16 & - & LFQ & 19.50 & 0.502 & 6.649 & 17.00 & 0.398 & 2.701 \\
        LlamaGen-Tokenizer8$\times$8 & - & VQ & 21.99 & 0.616 & 4.123 & 19.64 & 0.498 & 1.403 \\
        LlamaGen-Tokenizer16$\times$16 & - & VQ & 19.11 & 0.491 & 6.077 & 18.38 & 0.448 & 1.657 \\
        Cosmos-0.1-Tokenizer-DI8$\times$8 & 1024 & FSQ & 24.40 & 0.704 & \textbf{3.710}  & 24.48 & 0.701 & \textbf{1.265} \\
        Cosmos-0.1-Tokenizer-DI16$\times$16 & 1024 & FSQ & 20.45 & 0.529 & 7.234 & 20.49 & 0.518 & 2.518 \\
        \midrule        
        Cosmos-Tokenize1-DI8$\times$8-360p & 1024 & AE & \textbf{31.36} & \textbf{0.714} & 4.133 & \textbf{31.34} & \textbf{0.707} & 1.324 \\
        Cosmos-Tokenize1-DI16$\times$16-360p & 1024 & AE & 30.54 & 0.565 & 7.557 & 30.46 & 0.549 & 2.147 \\
        \bottomrule
    \end{tabular}
    \label{tab:token_res_DI}
\end{table}

\begin{table}[t]
    \setlength{\tabcolsep}{9.2pt} %
    \small
    \captionsetup{justification=centering}
    \caption{Runtime performance comparison of tokenizers. Time is reported per image or per video frame.}
    \centering
    \vspace{-1em}
    \begin{tabular}{rccccc}
        \toprule
        Tokenizer & Type & Resolution & Frames & Parameters & Time (ms)  \\
        \midrule
        FLUX-Tokenizer8$\times$8 & Continuous-Image & $1024 \times 1024$ & - &  84M & 242 \\
        Cosmos-0.1-Tokenizer-CI8$\times$8 & Continuous-Image & $1024 \times 1024$ & - & 77M & \textbf{62.7} \\
        \midrule
        LlamaGen-Tokenizer8$\times$8 & Discrete-Image & $1024 \times 1024$ & - & 70M & 475 \\
        Cosmos-0.1-Tokenizer-DI8$\times$8 & Discrete-Image & $1024 \times 1024$ & - & 79M & \textbf{64.2} \\
        \midrule
        CogVideoX-Tokenizer4$\times$8$\times$8 & Continuous-Video & $720 \times 1280$ & 17 & 216M & 414 \\
        Omni-Tokenizer4$\times$8$\times$8 & Continuous-Video & $720 \times 1280$ & 17 & 54M & 82.9 \\
        Cosmos-0.1-Tokenizer-CV4$\times$8$\times$8 & Continuous-Video & $720 \times 1280$ & 49 & 105M & \textbf{34.8} \\
        \midrule
        Omni-Tokenizer4$\times$8$\times$8 & Discrete-Video & $720 \times 1280$ & 17 & 54M & 53.2 \\
        Cosmos-0.1-Tokenizer-DV4$\times$8$\times$8 & Discrete-Video & $720 \times 1280$ & 17 & 105M & \textbf{51.5} \\
        \bottomrule
    \end{tabular}
    \label{tab:token_res_perf}
\end{table}

In the second stage, we use the optical flow ($\texttt{OF}$) loss~\citep{teed2020raft} to handle the temporal smoothness of reconstructed videos,
\begin{equation}\nonumber
    \mathcal{L}_{\text{Flow}} = \frac{1}{T} \sum_{t=1}^{T} \left\lVert \texttt{OF}(\hat{x}_{t}, \hat{x}_{t-1}) - \texttt{OF}(x_{t}, x_{t-1}) \right\rVert_1 + \frac{1}{T} \sum_{t=0}^{T-1} \left\lVert \texttt{OF}(\hat{x}_{t}, \hat{x}_{t+1}) - \texttt{OF}(x_{t}, x_{t+1}) \right\rVert_1,
\end{equation}
and the Gram-matrix ($\texttt{GM}$) loss~\citep{gatys2016image} to enhance the sharpness of reconstructed images,
\begin{equation}
    \mathcal{L}_{\text{Gram}} = \frac{1}{L} \sum_{l=1}^{L} \sum_t \alpha_l \left\lVert \texttt{GM}_l (\hat{x}_t) - \texttt{GM}_l (x_t) \right\rVert_1.
\end{equation}

Additionally, we use adversarial loss in the fine-tuning stage to further enhance reconstruction details, particularly at large compression rates.

We train the image tokenizers (denoted as \textbf{CI} and \textbf{DI}) at two compression rates: \( 8 \times 8 \) and \( 16 \times 16 \). Similarly, we train the video tokenizers (denoted as \textbf{CV} and \textbf{DV}) at three compression rates: \( 4 \times 8 \times 8 \), \( 8 \times 8 \times 8 \), and \( 8 \times 16 \times 16 \). Here, the compression rates are expressed as \( H \times W \) for images and \( T \times H \times W \) for videos, where \( T \) represents the temporal dimension, and \( H \) and \( W \) represent the spatial dimensions.

For the video tokenizers, we create two variants:
\begin{enumerate}
    \item \textbf{Cosmos-0.1-Tokenizer}: Trained using mini-batches sampling a smaller number of 720p video frames (49 frames for \textbf{CV} and 17 frames for \textbf{DV}).
    \item \textbf{Cosmos-Tokenize1}: Trained using mini-batches sampling a larger number of 360p or 720p video frames (121 frames for \textbf{CV} and 49 frames for \textbf{DV}).
\end{enumerate}

This approach ensures flexibility in handling varying temporal and spatial resolutions for image and video data. Our experiments indicate that the tokenizers generalize well to resolutions they were trained on and maintain strong quality at higher resolutions.

\subsection{Results}\label{subsec:token_evaluation}

We extensively evaluate our Cosmos Tokenizer suite on various image and video benchmark datasets. For the evaluation of image tokenizers, we follow prior art to evaluate MS-COCO 2017~\citep{lin2015mscoco} and ImageNet-1K~\citep{deng2009imagenet}. We use the MS-COCO 2017 validation subset of $5{,}000$ images, and ImageNet-1K validation subset of $50{,}000$ images as image evaluation benchmark.

\noindent \textbf{TokenBench.} For video tokenizer evaluation, there is not yet a standard benchmark for high-resolution and long-duration videos. To this end, we introduce a benchmark called \emph{TokenBench} to cover a wide variety of domains, including robotic manipulation, driving, egocentric, and web videos, and standardize the evaluation. We resort to existing video datasets that are commonly used for various tasks, including BDD100K~\citep{yu2020bdd100k}, EgoExo-4D~\citep{grauman2024egoexo4d}, BridgeData V2~\citep{walke2023bridgev2}, and Panda-70M~\citep{chen2024panda70m}. We randomly sample $100$ videos from each dataset and preprocess them by taking the first $10$ seconds and resizing the short size to $1080$. For Panda-70M, we manually filter out the videos with low-quality content and small motions. For EgoExo-4D, we randomly pick $100$ scenes and sample one egocentric video and one exocentric video. This results in a total of $500$ videos. Some examples of \emph{TokenBench} can be found in~\cref{fig:tokenbench}. We release \emph{TokenBench} at the \href{https://github.com/NVlabs/TokenBench}{github.com/NVlabs/TokenBench}.

In addition to \emph{TokenBench}, we also evaluate our video tokenizers on the DAVIS dataset at $1080$p resolution.

\noindent {\textbf{Baselines and evaluation metrics.}}  We evaluate our tokenizers at various compression rates to showcase their effectiveness for different computational needs. We compare each of these tokenizers with state-of-the-art image and video tokenizers. \cref{tab:tokenizer_features_comparisons} presents the specific SOTA tokenizers we compared against in various settings. The evaluation metrics include Peak Signal-to-Noise Ratio (PSNR), Structural Similarity (SSIM), reconstruction Fr\'{e}chet Inception Distance (rFID)~\citep{FID} for images, and reconstruction Fr\'{e}chet Video Distance (rFVD)~\citep{FVD} for videos.

\noindent {\textbf{Quantitative results.}} \cref{tab:token_res_CV,tab:token_res_DV} summarize the average quantitative metrics of continuous and discrete video tokenizers on various benchmarks. As shown in both tables, Cosmos Tokenizer achieves state-of-the-art performance in all the metrics compared to prior arts on both the DAVIS video dataset and \emph{TokenBench}, with a spatial-temporal compression ratio of $4\times 8 \times 8$. Moreover, even with $2\times$ and $8\times$ higher compression ratios (\ie, $8\times 8\times 8$ and $8\times 16 \times 16$), Cosmos Tokenizer still achieves better quality than prior art, showcasing an excellent compression-quality trade-off.

\cref{tab:token_res_CI,tab:token_res_DI} summarize the average quantitative metrics of continuous and discrete image tokenizers on various image benchmarks, covering a wide range of image types. As shown, compared to prior arts, Cosmos Tokenizer consistently achieves state-of-the-art results with a compression ratio of $8\times 8$. More importantly, at a $4\times$ larger compression ratio of $16\times 16$, the image quality of Cosmos Tokenizer is often comparable or even better than prior art at $8\times 8$ compression ratio, as shown in \cref{tab:token_res_CI,tab:token_res_DI}.

These quantitative results on a variety of image and video benchmark datasets confirm that Cosmos Tokenizer is able to better represent visual content with large spatial-temporal compression.

\noindent {\textbf{Runtime performance.}} \cref{tab:token_res_perf} shows the number of parameters and the averaged encoding and decoding times per image or per video frame, measured on a single A100 80GB GPU. In comparison, we also list the parameters and the average speeds of prior state-of-the-art tokenizers. As shown, for both image and video tokenizers,  Cosmos Tokenizer is $2\times \sim 12\times$ faster while maintaining the smallest model size compared to prior arts, showing that Cosmos Tokenizer has high efficiency for encoding and decoding visual content.

\section{World Foundation Model Pre-training}\label{sec::pretrained_world_model}

Pre-trained WFMs are generalists that capture general knowledge of real-world physics and natural behaviors. We exploit two different scalable deep learning paradigms, diffusion models and autoregressive models, to build two families of WFMs. Both diffusion models and autoregressive models break a difficult generation problem into a sequence of easier sub-problems and have been turbo-charging the development of generative models. In the case of diffusion models, the difficult generation problem is divided into a sequence of denoising problems. In the case of autoregressive models, the difficult generation problem is divided into a sequence of next-token prediction problems. We discuss how we scale these deep learning paradigms using various parallelization techniques tailored for modern GPUs in our endeavor of building pre-trained WFMs. We train all of the WFM models reported in the paper using a cluster of $10{,}000$ NVIDIA H100 GPUs in a time span of three months.

\begin{table}[ht!]
    \centering
    \setlength{\tabcolsep}{8pt} %
    \captionsetup{justification=centering}
    \begin{threeparttable}
        \caption{Maps of Cosmos World Foundation Model. We have two sets of WFMs. One is based on diffusion models, while the other is based on autoregressive models. For each family, we build two base models and two derivative models. To achieve the best generation quality, we also build a prompt upsampler for the diffusion models and a diffusion decoder for the autoregressive models.}
        \label{tab:model_map}
        \begin{tabular}{c|cc}
            \toprule
            \textbf{Type} & Diffusion & Autoregressive \\
            \midrule
            \multirow{4}{*}{\makecell[c]{\vspace{8pt}\textbf{Models}}} & \makecell[c]{1.\\} \makecell[c]{Cosmos-Predict1-\\7B-Text2World} $\rightarrow$ \makecell[c]{Cosmos-Predict1-\\7B-Video2World} & \makecell[c]{1.\;\;\\} \makecell[c]{Cosmos-Predict1-\\4B} $\rightarrow$ \makecell[c]{Cosmos-Predict1-\\5B-Video2World} \vspace{4pt} \\
             & \makecell[c]{2.\\} \makecell[c]{Cosmos-Predict1-\\14B-Text2World} $\rightarrow$ \makecell[c]{Cosmos-Predict1-\\14B-Video2World} & \makecell[c]{2.\;\;\\} \makecell[c]{Cosmos-Predict1-\\12B} $\rightarrow$ \makecell[c]{Cosmos-Predict1-\\13B-Video2World} \\
            \midrule
            \textbf{Tokenizer} & Cosmos-Tokenize1-CV8$\times$8$\times$8-720p & Cosmos-Tokenize1-DV8$\times$16$\times$16-720p \\
            \midrule
            \multirow{2}{*}{\makecell[c]{\vspace{12pt}\textbf{Enhancer}}} & \multirow{2}{*}{\makecell[c]{\vspace{12pt}Cosmos-UpsamplePrompt1-12B-Text2World}} & \makecell[c]{Cosmos-Predict1-7B-Decoder-\\DV8$\times$16$\times$16ToCV8$\times$8$\times$8-720p} \\
            \bottomrule
        \end{tabular}
    \end{threeparttable}
\end{table}

In \cref{tab:model_map}, we present a map of our pre-trained WFMs and their companions. For the diffusion-based WFM family, we start by building two Text2World models of 7B and 14B, respectively, which render Cosmos-Predict1-7B-Text2World and Cosmos-Predict1-14B-Text2World. These models can map text prompts to videos of visual worlds. We then fine-tune the Text2World models to take additional video input, representing the current observation. The result is a Video2World model where the future video is predicted based on the current observation (input video) and the perturbation (text prompt). These diffusion models are latent diffusion models that take continuous tokens. We use Cosmos-Tokenize1-CV8$\times$8$\times$8-720p to produce the visual tokens. The training text prompts for the WFMs are produced by a VLM through video description generation. These descriptions follow a different distribution of human descriptions of videos. To mitigate the domain gap, we build Cosmos-UpsamplePrompt1-12B-Text2World based on the Mistral-NeMo-12B-Instruct model~\citep{mistral_nemo_2024} to help convert human text prompts to those preferred by our diffusion-based WFMs.

For the autoregressive-based WFM family, we first build two base models that are 4B and 12B in size, respectively, to predict future videos purely based on the current video observation. We name them Cosmos-Predict1-4B and Cosmos-Predict1-12B, respectively. These are Llama3-style GPT models trained from scratch for the video prediction task and bear no language understanding. To enable autoregressive-based WFMs to utilize textual information for next token prediction, we incorporate T5 embeddings of the input text prompt into the WFMs through cross-attention layers added to the transformer blocks. These autoregressive WFMs use Cosmos-Tokenize1-DV8$\times$16$\times$16-720p, which maps an input video to a few integers. The heavy compression of the tokenizer can sometimes lead to undesired distortions. To address the problem, we build a diffusion decoder (Cosmos-Predict1-7B-Decoder-DV8$\times$16$\times$16ToCV8$\times$8$\times$8-720p) through fine-tuning the Cosmos-Predict1-7B-Text2World model to map discrete tokens in the DV8$\times$16$\times$16 space to continuous tokens in the CV8$\times$8$\times$8 space.

\subsection{Diffusion-based World Foundation Model}\label{sec::diffusion_model}

Our diffusion-based WFMs are latent diffusion models that operate within a learned latent space of a tokenizer, enabling a compact, reduced-dimensional representation of videos. This design choice offers several advantages: it reduces computational costs during both training and inference while simplifying the denoising task~\citep{rombach2022high,hoogeboom2024simpler}. To tokenize videos into latent representations, we employ Cosmos-Tokenize1-CV8$\times$8$\times$8-720p.

\subsubsection{Formulation}\label{sec::diffusion_model_formulation}

To train our diffusion WFMs, we adopt the approach outlined in EDM~\citep{Karras2022edm, Karras2024edm2}. The denoising score matching loss for the denoiser $D_\theta$, evaluated at a noise level $\sigma$, is defined as
\begin{equation}
  \gL(D_\theta, \sigma) ~=~ \E_{\rvx_0, \rvn} \Big[ {\big\lVert D_\theta(\rvx_0 + \rvn; \sigma) - \rvx_0 \big\rVert}^2_2 \Big] \label{eq:lossPerNoiseLevel}
  \text{,}
\end{equation}
where $\rvx_0 \sim \pdata$ is a clean image or video sampled from the training set, $\rvn \sim \gN\big( \mathbf{0}, \sigma^2 \rmI \big)$ is i.i.d.~Gaussian noise, and $D_\theta$ is a noise-conditioned neural network tasked with denoising the corrupted sample $\rvx_0 + \rvn$. We adhere to the preconditioning design introduced in EDM for parameterizing \( D_\theta \). The overall training loss is defined as a weighted expectation of $\gL(D_\theta; \sigma)$ over the noise levels:
\begin{align}
    \gL(D_\theta) &= \E_{\sigma} \left[\frac{\lambda(\sigma)}{e^{u(\sigma)}} \gL(D_\theta, \sigma) + u(\sigma)\right] \label{eq:loss} \text{,}\\
    \lambda(\sigma) &= \big( \sigma^2 + \sigma_{\text{data}}^2 \big) \,/\, (\sigma \cdot \sigma_{\text{data}})^2 \label{eq:edmLambda} \text{,}\\
    \ln(\sigma) &\sim~ \gN\big( P_\text{mean}, P_\text{std}^2 \big) \label{eq:edmLogNormal}
    \text{,}
\end{align}
where the distribution of noise levels $\sigma$ is controlled by hyperparameters $P_\text{mean}$ and $P_\text{std}$. \( \sigma_{\text{data}} \) is the standard deviation of the training data, and the weighting function $\lambda(\sigma)$ ensures equal contribution of each noise level at the beginning of the training. However, as training progresses, this balance may deteriorate. To mitigate this issue, we treat the optimization over various noise levels as a form of multi-task learning. We utilize the uncertainty-based weighting approach by introducing $u(\sigma)$ as a continuous uncertainty function quantifying the uncertainty for the denoising objective $\gL(D_\theta, \sigma)$ at noise level $\sigma$. We use a simple MLP to parameterize $u(\sigma)$ and minimize the overall loss $\gL(D_\theta)$ during training. Intuitively, the contribution of loss at noise level $\sigma$ is weighted down if the model is uncertain about the task, \ie, if $u(\sigma)$ is high. At the same time, the model is penalized for this uncertainty, encouraging $u(\sigma)$ to be as low as possible.

Compared to recent video generative models that adopt the Gaussian flow matching formulation~\citep{polyak2024movie, kong2024hunyuanvideo}, our work is derived from the diffusion score matching perspective~\citep{ho2020denoising,song2020score}. However, as shown by \citet{gao2025diffusionmeetsflow}, these frameworks are theoretically equivalent, sharing fundamental similarities in their objectives and training procedures. Our EDM-based formulation aligns with these insights, mainly differing in the choice of preconditioning designs and hyperparameters. In practice, we have not encountered any performance limitations with the EDM formulation.

\subsubsection{Architecture}\label{sec::diffusion_model_backbone}

\begin{figure}[tbh!]
    \centering
    \includegraphics[width=\textwidth]{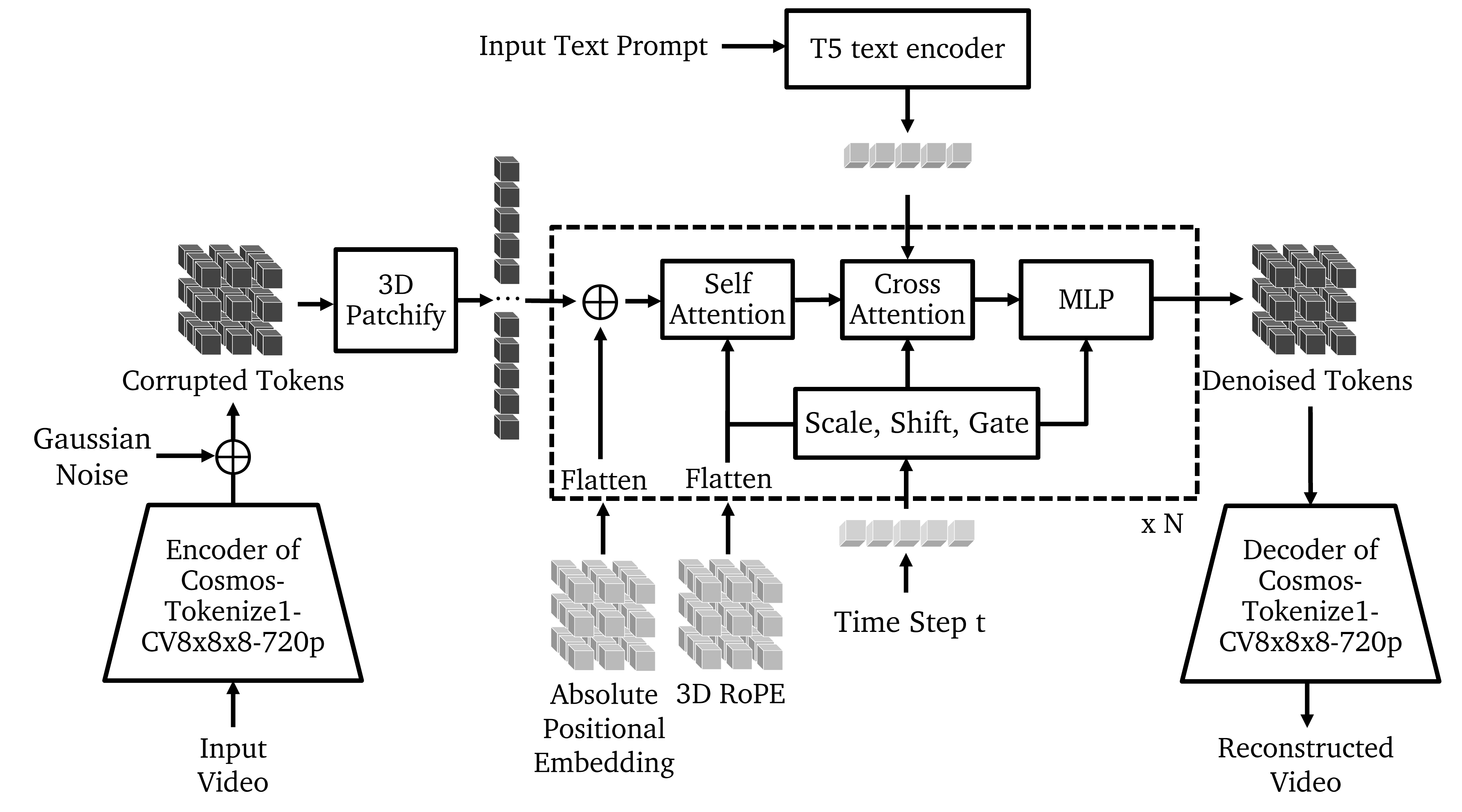}
    \caption{\textbf{Overall architecture of Cosmos-Predict1 World Foundation Model}. The model processes an input video through the encoder of the Cosmos-Tokenize1-CV8$\times$8$\times$8-720p to obtain latent representations, which are subsequently perturbed with Gaussian noise. These representations are then transformed using a 3D patchification process. In the latent space, the architecture applies repeated blocks of self-attention, cross-attention (integrating input text), and feed-forward MLP layers, modulated by adaptive layer normalization (scale, shift, gate) for a given time step $t$. The decoder of Cosmos-Tokenize1-CV8$\times$8$\times$8-720p reconstructs the final video output from the refined latent representation.}
    \label{fig:diffusion_architecture}
\end{figure}

In this section, we describe the design of our denoiser network $D_\theta$ that builds upon DiT~\citep{peebles2023scalable}, which was originally designed for label-conditioned image generation. We adapt its architecture to better suit our goal of controllable video generation. We visualize the overall network design in~\cref{fig:diffusion_architecture}.

\noindent \textbf{3D patchification.} The input to our network is a latent representation of shape $T \times C \times H \times W$ for both image and video data, with images differentiated by a video with a single frame. To prepare inputs for our denoiser network, we first ``patchify'' the state using a linear layer and subsequently flatten it. This process involves projecting non-overlapping cubes of shape $(p_t, p_h, p_w)$ into individual token inputs for the network. Consequently, after patchification, an image or video is reshaped into a one-dimensional, spatiotemporal sequence of length $THW/(p_t p_h p_w)$.
We use $p_t=1, p_h=p_w=2$ for our denoiser network.

\noindent \textbf{Hybrid positional embedding with FPS-aware 3D RoPE and learnable embedding.} We employ a 3D-factorized Rotary Position Embedding (RoPE)~\citep{su2024roformer} to allow the generation of arbitrary size, aspect ratio, and video length. Specifically, we partition the feature dimension into three approximately equal chunks, each applying RoPE with positional information along the temporal, height, and width axes, respectively. In practice, this can be implemented efficiently without splitting and concatenation in each block by concatenating frequency embeddings in their respective axes and reusing RoPE kernels optimized for Large Language Models (LLMs). To further support video synthesis with varying frame rates, we rescale temporal frequencies based on the training video's Frames Per Second (FPS). Due to RoPE's relative positional encoding property and our 3D factorization design, the FPS-aware design is compatible with our joint image-video training. An additional benefit of RoPE is evident during progressive training when we alter resolution or video length. By leveraging Neural Tangent Kernel (NTK)-RoPE~\citep{bloc97}, we observe rapid model convergence, achieving reasonable performance even within $5{,}000$ training steps. Additionally, we find that adding an extra learnable absolute positional embedding per transformer block can further enhance the model, reduce training loss, and reduce morphing artifacts in generated videos.

\noindent \textbf{Cross-attention for text conditioning.} We rely on cross-attention layers in our network for incorporating linguistic information. Each transformer block consists of sequential self-attention, cross-attention, and feed-forward layers. While self-attention operates over spatiotemporal tokens, cross-attention integrates semantic context using T5-XXL~\citep{raffel2020exploring} embeddings as keys and values, enabling effective text conditioning.

\noindent \textbf{Query-key normalization.}
In the early stages of training, we observe instability in the growth of attention logits, leading to a collapse of attention entropy. We follow existing literature~\citep{dehghani2023scaling,wortsman2023small,esser2024scaling} to normalize query $Q$ and key $K$ before the attention operation. We use Root Mean Square Normalization (RMSNorm)~\citep{zhang2019root} with learnable scales for all self-attention and cross-attention layers within our network.

\noindent \textbf{AdaLN-LoRA.} We find that DiT's adaptive layer normalization (AdaLN) layers~\citep{xu2019understanding,peebles2023scalable} account for a significant portion of the model parameters while contributing negligibly to the computational complexity in terms of FLOPs. Inspired by W.A.L.T~\citep{gupta2023photorealistic}, we implement Low-Rank Adaptation (LoRA)~\citep{hu2021lora} to decompose the dense linear projections in these layers into low-rank approximations. For Cosmos-Predict1-7B, this architectural optimization achieves a 36\% reduction in parameter count (from 11B to 7B parameters) while maintaining performance parity across all evaluation metrics, demonstrating the effectiveness of our parameter-efficient design.
\begin{table}[ht]
    \setlength{\tabcolsep}{8pt} %
    \small
    \captionsetup{justification=centering}
    \caption{Configuration details of Cosmos-Predict1 models.}
    \centering
    \begin{tabular}{l|cccc}
        \toprule
         \textbf{Configuration} & \textbf{7B-Text2World} & \textbf{14B-Text2World} & \textbf{7B-Video2World} & \textbf{14B-Video2World} \\
        \midrule
        Number of Layers & $28$ & $36$ & $28$ & $36$ \\
        Model Dimension & $4{,}096$ & $5{,}120$ & $4{,}096$ & $5{,}120$ \\
        FFN Hidden Dimension & $16{,}384$ & $20{,}480$ & $16{,}384$ & $20{,}480$ \\
        AdaLN-LoRA Dimension & $256$ & $256$ & $256$ & $256$ \\
        Number of Attention Heads & $32$ & $40$ & $32$ & 40 \\
        Number of Key / Value Heads & $32$ & $40$ & $32$ & $40$ \\
        MLP Activation & \multicolumn{4}{c}{GELU} \\
        Positional Embedding & \multicolumn{4}{c}{Hybrid positional embedding} \\
        Conditional Information & Text; FPS & Text; FPS & Text; FPS; Frames & Text; FPS; Frames \\
        Base Learning Rate & $2^{-15}$ & $2^{-16}$ & $2^{-15}$ & $2^{-16}$ \\
        Weight decay & $0.1$ & $0.2$ & $0.1$ & $0.2$ \\
        Learning Rate Warmup &  \multicolumn{4}{c}{Linear scheduler with $2{,}500$ iterations} \\
        AdamW momentum and $\epsilon$&  \multicolumn{4}{c}{$\beta_1, \beta_2 = 0.9, 0.99$; $\epsilon = 10^{-10}$} \\
        \bottomrule
    \end{tabular}
    \label{tab:diffusion_model_specs}
\end{table}

\subsubsection{Training Strategy}\label{sec::diffusion_model_training}
This section outlines the methodologies employed to train our models on datasets spanning multiple modalities, resolutions, aspect ratios, and conditioning inputs.

\noindent \textbf{Joint image and video training.} To leverage the vast abundance of high-quality, diverse image datasets in model training, we implement an alternating optimization strategy that interleaves batches of image and video data. To facilitate cross-modal knowledge transfer between image and video domains, we adopt a domain-specific normalization scheme that aligns the latent distributions using sufficient statistics estimated independently for image and video data. This approach is motivated by the observation that reducing the distributional shift between image and video latent representations improves generation quality. Furthermore, we observe non-stationary statistics across temporal and channel dimensions in video latent representations. To address this heterogeneity, we employ a normalization strategy that applies frame-wise and channel-wise standardization to video latent representations, effectively encouraging them to better approximate an isotropic Gaussian prior distribution.

Beyond cross-modality knowledge transfer, our normalization scheme provides an important theoretical benefit: scale invariance in the signal-to-noise ratio during training. Consider two zero-mean latent representations with different scales: one standardized to unit variance, and another with variance 4. When adding Gaussian noise $\gN(0, \sigma^2)$ to achieve a desired signal-to-noise ratio for the standardized representation, we must scale the noise to $\gN(0, 4\sigma^2)$ for the unnormalized representation to maintain the same ratio. By standardizing all latent representations, we ensure consistent signal-to-noise ratios across different scales, facilitating model adaptation even when the underlying tokenizer is updated during training.

To maintain computational efficiency, we balance image and video batch sizes to ensure comparable memory utilization across GPUs. However, we observe that the video batch denoising loss exhibits slower convergence compared to the image batch loss. We attribute this to the inherent temporal redundancy in video frames, which results in smaller gradient magnitudes for video batches. Drawing inspiration from recent advances in multi-resolution image training~\citep{chen2023importance,hoogeboom2023simple,atzmon2024edify}, we address this convergence discrepancy by scaling the video batch noise levels by the square root of the frame count relative to image batch noise levels.

\begin{table}[ht!]
    \centering
    \setlength{\tabcolsep}{6pt} %
    \small
    \captionsetup{justification=centering}
    \begin{threeparttable}
        \caption{Stages of progressive training and their specifications.}
        \label{tab:diffusion_training_strategy}
        \begin{tabular}{l|cccccc}
            \toprule
            \textbf{Stage} & \textbf{Resolution} & \textbf{Number of Frames} & \textbf{Context Length} & \textbf{FSDP Size} & \textbf{CP Size} \\
            \midrule
            Low-resolution Pre-training & 512p (640$\times$512) & 57 & $10{,}240$~\tnote{a} & 64 & 2 \\
            High-resolution Pre-training & 720p (1280$\times$704) & 121 & $56{,}320$~\tnote{b} & 64 & 8 \\
            High-quality Fine-tuning & 720p (1280$\times$704) & 121 & $56{,}320$~\tnote{b} & 64 & 8 \\
            \bottomrule
        \end{tabular}
        \begin{tablenotes}
            \small
            \item[a] $10{,}240$ (the context length) is computed as: $640$ (width) $\div 8$ (tokenize) $\div 2$ (patchify) $\times 512$ (height) $\div 8$ (tokenize) $\div 2$ (patchify) $\times [(57-1) \div 8 + 1]$ (tokenize frames). \\
            \item[b] $56{,}320$ (the context length) is computed as: $1280$ (width) $\div 8$ (tokenize) $\div 2$ (patchify) $\times 704$ (height) $\div 8$ (tokenize) $\div 2$ (patchify) $\times [(121-1) \div 8 + 1]$ (tokenize frames).
        \end{tablenotes}
    \end{threeparttable}
\end{table}

\noindent \textbf{Progressive training.} We adopt a progressive training strategy, with the specifics of each stage detailed in~\cref{tab:diffusion_training_strategy}. The initial stage involves training on videos and images at a resolution of 512 pixels, using videos composed of 57 frames. Subsequently, we transition to the target resolution of 720 pixels, increasing the video length to 121 frames. After pre-training on massive data, we fine-tune the model on a high-quality subset for $\Bigo(10k)$ iterations with a linearly decaying learning rate. Consistent with findings from~\cite {dai2023emu}, we also find that fine-tuning can improve the quality of the generated videos.

\noindent \textbf{Multi-aspect training.} To accommodate content with varying aspect ratios, we organize the data into five distinct buckets corresponding to ratios of 1:1, 3:4, 4:3, 9:16, and 16:9, assigning each image or video to the bucket with the closest aspect ratio. During training, each data parallel process group samples from one bucket, allowing different buckets across different parallel process groups. We implement longest-side resizing to maximally preserve the original content information described in the prompt. For batch processing, we apply reflection padding to missing pixels and supply the padding mask to the diffusion backbone, enabling precise control during inference.

\noindent \textbf{Mixed-precision training.} We maintain two copies of the model weights: one in BF16 and another in FP32. During the forward and backward passes, the BF16 weights are used to improve training efficiency, resulting in gradients and activations also in BF16 format. For parameter updates, the weights are updated in FP32 to ensure numerical stability. The updated FP32 parameters are then copied and cast to BF16 for the next iteration. To further stabilize training, we scale the loss of denoising score matching in~\cref{eq:lossPerNoiseLevel} by a factor of 10. We also find that lower betas and eps coefficients in AdamW significantly reduce loss spikes. For our 14B diffusion model training, we rarely encountered loss spikes, and there were no non-recoverable loss spikes.

\noindent \textbf{Text conditioning.} For our Text2World models, we employ T5-XXL~\citep{raffel2020exploring} as the text encoder. We zero-pad T5 embeddings to maintain a fixed sequence length of 512. To enhance text-context alignment, we adopt classifier-free guidance~\citep{ho2022classifier}. Unlike prior works~\citep{ediffI,saharia2022photorealistic} that randomly zero out text embeddings, we omit this step due to the effectiveness of negative prompts during inference. Notably, as a text-to-image generator, our model excels in generating high-fidelity images even without guidance, a capability we attribute to the high-quality training dataset. While classifier-free guidance typically promotes mode-seeking behavior for preferred visual content, we find that careful data selection achieves a similar effect. However, for video generation, the lack of comparable high-quality data leads to suboptimal results under low guidance settings. Consequently, higher guidance values are required to produce satisfactory content in video-generation tasks.

\noindent \textbf{Image and video conditioning.} We extend our Text2World models to build Video2World models that support image and video conditioning by incorporating previous frame(s) into the generation process. Specifically, the conditional frame(s) are concatenated with the generated frames along the temporal dimension. To improve robustness against variations in input frame(s) during inference, we introduce augmented noise to the conditional frames during training. The sigma value for this augmented noise is sampled with $P_\text{mean}=-3.0, P_\text{std}=2.0$. Additionally, the input to the diffusion model is concatenated along the channel dimension with a binary mask that distinguishes conditional frames from generated frames. The loss function excludes contributions from the locations of conditional frames, focusing exclusively on the generated output. To improve generalization, we randomly vary the number of conditional frames during training. During inference, the model can flexibly operate with either a single conditional frame (image) or multiple previous frames as input.

\subsubsection{Scaling Up}\label{sec::diffusion_model_scaling}

Here, we outline the techniques that enable efficient scaling of our diffusion WFMs. We analyze the memory requirements of our models, discuss parallelism strategies, and compare our training setup against other video diffusion models and state-of-the-art LLMs.

\begin{table}[t!]
    \centering
    \setlength{\tabcolsep}{10pt} %
    \small
    \captionsetup{justification=centering}
    \begin{threeparttable}
        \caption{Cosmos-Diffusion transformer FLOPs and activation memory. The table provides the computational cost (FLOPs) and activation memory requirements for each operation. For FLOPs, we use a factor of 2 to describe the multiply accumulate cost. A ``\textemdash''~denotes cases where the value is either negligible due to its small magnitude or omitted because activation checkpointing is employed to recompute values instead of storing them, thus saving memory.}
        \label{tab:diffusion_transformer_flops_memory}
        \begin{tabular}{l|ccc}
            \toprule
            \textbf{Layer} & \textbf{Operation} & \textbf{FLOPs} & \textbf{Activations (Tensor Shape)} \\
            \midrule
            \multirow{6}{*}{Self-attention} &
            $Q, K, V$ Projections & $2 \times 3 \times \text{seq\_len} \times \text{d\_model}^2$ &
            $\text{seq\_len} \times \text{batch\_size} \times \text{d\_model}$~\tnote{a} \\ &
            $QK$ Norm & \textemdash &
            $2 \times \text{seq\_len} \times \text{batch\_size} \times \text{d\_model}$~\tnote{b} \\ &
            $A = Q @ K^T$ & $2 \times \text{seq\_len}^2 \times \text{d\_model}$ &
            \textemdash~\tnote{c} \\ &
            $A' = \text{Softmax}(A)$ & \textemdash &
            \textemdash~\tnote{d} \\ &
            $A' @ V$ & $2 \times \text{seq\_len}^2 \times \text{d\_model}$ &
            $\text{seq\_len} \times \text{batch\_size} \times \text{d\_model}$~\tnote{e} \\ &
            Final Projection & $2 \times \text{seq\_len} \times \text{d\_model}^2$ &
            $\text{seq\_len} \times \text{batch\_size} \times \text{d\_model}$ \\
            \midrule
            \multirow{6}{*}{Cross-attention} &
            $Q, K, V$ Projections & $2 \times \text{seq\_len} \times \text{d\_model}^2$ &
            $\text{seq\_len} \times \text{batch\_size} \times \text{d\_model}$ \\ &
            $QK$ Norm & \textemdash &
            $\text{seq\_len} \times \text{batch\_size} \times \text{d\_model}$~\tnote{f} \\ &
            $A = Q @ K^T$ & \textemdash &
            \textemdash~\tnote{c} \\ &
            $A' = \text{Softmax}(A)$ & \textemdash &
            \textemdash~\tnote{d} \\ &
            $A' @ V$ & \textemdash &
            \textemdash~\tnote{g} \\ &
            Final Projection & $2 \times \text{seq\_len} \times \text{d\_model}^2$ &
            $\text{seq\_len} \times \text{batch\_size} \times \text{d\_model}$ \\
            \midrule
            \multirow{3}{*}{Feedforward} &
            Up Projection & $4 \times \text{seq\_len} \times \text{d\_model}^2$ &
            $\text{seq\_len} \times \text{batch\_size} \times \text{d\_model}$ \\ &
            GELU & \textemdash &
            $4 \times \text{seq\_len} \times \text{batch\_size} \times \text{d\_model}$ \\ &
            Down Projection & $4 \times \text{seq\_len} \times \text{d\_model}^2$ &
            \textemdash~\tnote{h} \\
            \midrule
            \multirow{4}{*}{AdaLN} &
            LayerNorm & \textemdash &
            $\text{seq\_len} \times \text{batch\_size} \times \text{d\_model}$ \\ &
            Scale & \textemdash &
            \textemdash~\tnote{i} \\ &
            Shift & \textemdash &
            \textemdash \\ &
            Gate & \textemdash &
            $\text{seq\_len} \times \text{batch\_size} \times \text{d\_model}$ \\
            \bottomrule
        \end{tabular}
        \begin{tablenotes}
            \small
            \item[a] The shared input is stored.
            \item[b] The query $Q$ and key $K$ are stored.
            \item[c] The normalized query $Q$ and key $K$ are recomputed.
            \item[d] The attention scores ($A = Q @ K^T$) are recomputed.
            \item[e] The value $V$ is stored. The normalized attention weights ($A' = \text{Softmax}(A)$) are recomputed.
            \item[f] In cross-attention, only query $Q$ is counted; key $K$ has much shorter sequence length and is thus negligible.
            \item[g] In cross-attention, the value $V$ has much shorter sequence length and is thus negligible.
            \item[h] The input is recomputed from GELU.
            \item[i] The input is recomputed from LayerNorm.
        \end{tablenotes}
    \end{threeparttable}
\end{table}
\vspace{-1em}

\noindent \textbf{Memory requirements.}
The four major components that consume the GPU memory are:
\begin{itemize}
    \item \textbf{Model parameters}: 10 bytes per parameter. Our mixed precision training stores model parameters in both FP32 and BF16, alongside Exponential Moving Average (EMA) weights in FP32.
    \item \textbf{Gradients}: 2 bytes per parameter. We store the gradients in BF16.
    \item \textbf{Optimizer states}: 8 bytes per parameter. We use AdamW~\citep{loshchilov2019decoupledweightdecayregularization} as our optimizer and store the optimizer states (\ie, first and second moments) in FP32.
    \item \textbf{Activations}: $(2 \times \text{number\_of\_layers} \times 15 \times \text{seq\_len} \times \text{batch\_size} \times \text{d\_model})$ bytes. We store the activations in BF16. \cref{tab:diffusion_transformer_flops_memory} provides details of the stored activations for major operations within the network. To optimize memory usage, we implement selective activation checkpointing~\citep{chen2016training,korthikanti2023reducing}, recomputing activations for memory-limited layers such as normalization functions.
\end{itemize}

For instance, our 14B model (Cosmos-Predict1-14B-Text2World) requires approximately 280 GB for model parameters, gradients, and optimizer states, alongside 310 GB for activations during high-resolution pre-training. Given the 80GB HBM3 limit of NVIDIA H100 GPUs, we employ Fully Sharded Data Parallelism (FSDP) and Context Parallelism (CP) to distribute memory demands across multiple GPUs.

\noindent \textbf{Fully Sharded Data Parallelism (FSDP).} FSDP improves memory efficiency by sharding model parameters, gradients, and optimizer states across devices. It gathers parameters only when needed during computation and releases them afterward. Unlike standard data parallelism, which duplicates parameters across devices, FSDP distributes parameters, gradients, and optimizer states, with each device managing only its shard. This approach minimizes memory usage to the largest temporarily unsharded parameter set alongside its shard of parameters, gradients, and optimizer states. For our implementation, we utilize a sharding factor of 32 for the 7B model and 64 for the 14B model to balance memory and communication latency.

\noindent \textbf{Context Parallelism (CP).} Scaling transformers for long-context settings introduces challenges with increased FLOPs and activation memory. CP addresses these challenges by distributing computation and activations across multiple GPUs. It works by splitting both the query $Q$ and the key-value $(K, V)$ along their sequence dimensions into $\text{CP\_SIZE}$ chunks, where $\text{CP\_SIZE}$ is the number of GPUs within a CP group. Each GPU processes one chunk of $Q$ and iteratively accumulates partial attention outputs using blocks of $(K,V)$ stored in the same CP group. Different implementations of CP utilize different communication primitives, including all-gather~\citep{dubey2024llama}, P2P~\citep{liu2023ring}, and all-to-all~\citep{jacobs2023deepspeed}. We employ the P2P variant from TransformerEngine~\citep{nvidia_transformer_engine}, which overlaps computation and communication by transferring $(K, V)$ blocks between GPUs while simultaneously processing attention. When block sizes are carefully chosen, this overlap effectively hides data transfer latency. We organize CP groups within NVLink-connected GPUs and overlap CP ranks with FSDP ranks for optimal utilization. For image iterations with shorter contexts, CP is disabled to improve throughput. Cross-attention layers do not use CP due to the shorter sequence lengths of $(K, V)$, which results in insufficient computation to mask communication latency.

Using Cosmos-Predict1-14B as an instance, employing FSDP with a sharding factor of 64 reduces memory requirements for parameters, gradients, and optimizer states, bringing them down from 280~GB to approximately $280 \mathbin{/} 64 \approx 4~\text{GB per GPU}$. Similarly, employing CP with \(\text{CP\_SIZE} = 8\) decreases activation memory from 310~GB to roughly $310 \mathbin{/} 8 \approx 40~\text{GB per GPU}$. It is important to note that these calculations are underestimations; in practice, additional memory is consumed by the tokenizer and unsharded parameters. Overlapping communication and computation in CP also necessitates each GPU to retain multiple chunks of $(K, V)$.

\noindent \textbf{Comparison with other video generative models.} Our parallelism strategy is deliberately streamlined compared to approaches outlined in HunyuanVideo~\citep{kong2024hunyuanvideo} and MovieGen~\citep{polyak2024movie}, which incorporate Tensor Parallelism (TP) and its extension, Sequence Parallelism (SP). Despite excluding TP/SP, our setup achieves comparable Model FLOPs Utilization (MFU). While TP/SP remains valuable in certain scenarios, such as larger models or alternative network topologies, a detailed analysis of tradeoffs is left for future work.

\begin{figure*}[ht!]
    \centering
    \setlength{\tabcolsep}{1.5pt}
    \renewcommand{\arraystretch}{0}
    \resizebox{\textwidth}{!}{%
    \begin{tabular}{cccccc} %
    & Frame 0 & Frame 29 & Frame 59 & Frame 89 & Frame 120 \\[4pt]
        {\rotatebox{90}{\hspace{16pt} 7B}} &
        \includegraphics[width=0.195\textwidth]{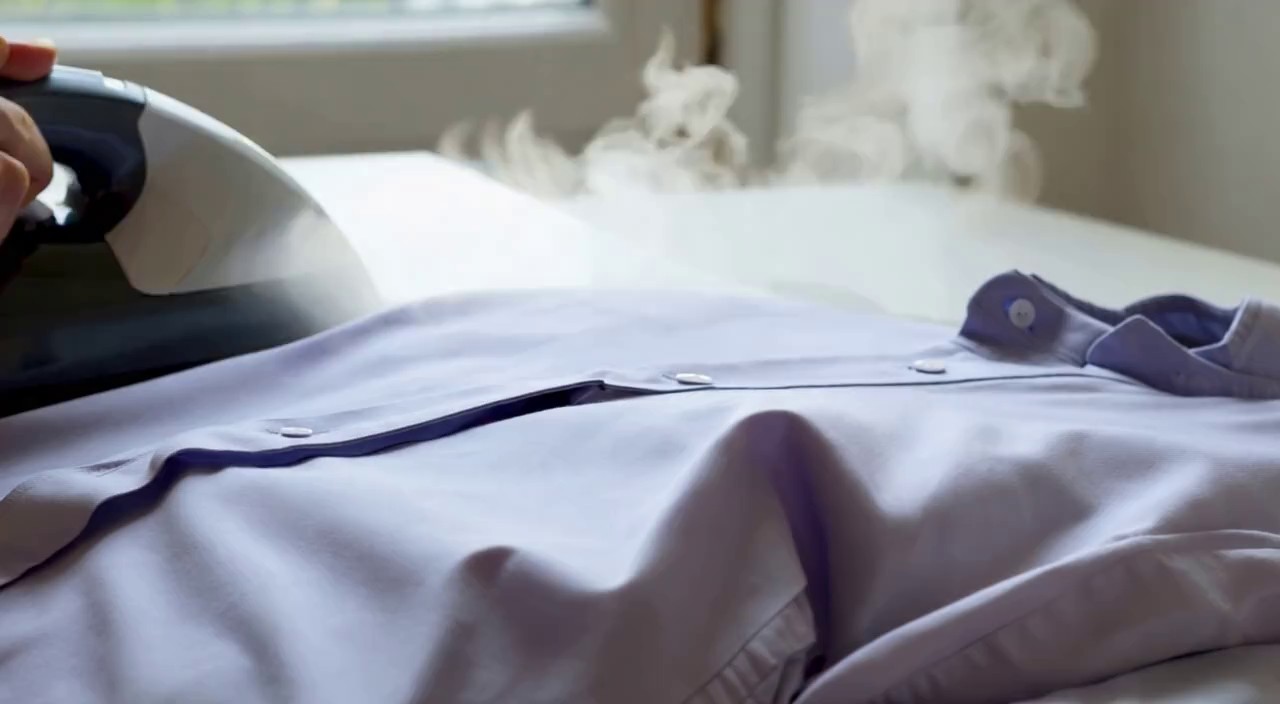} &
        \includegraphics[width=0.195\textwidth]{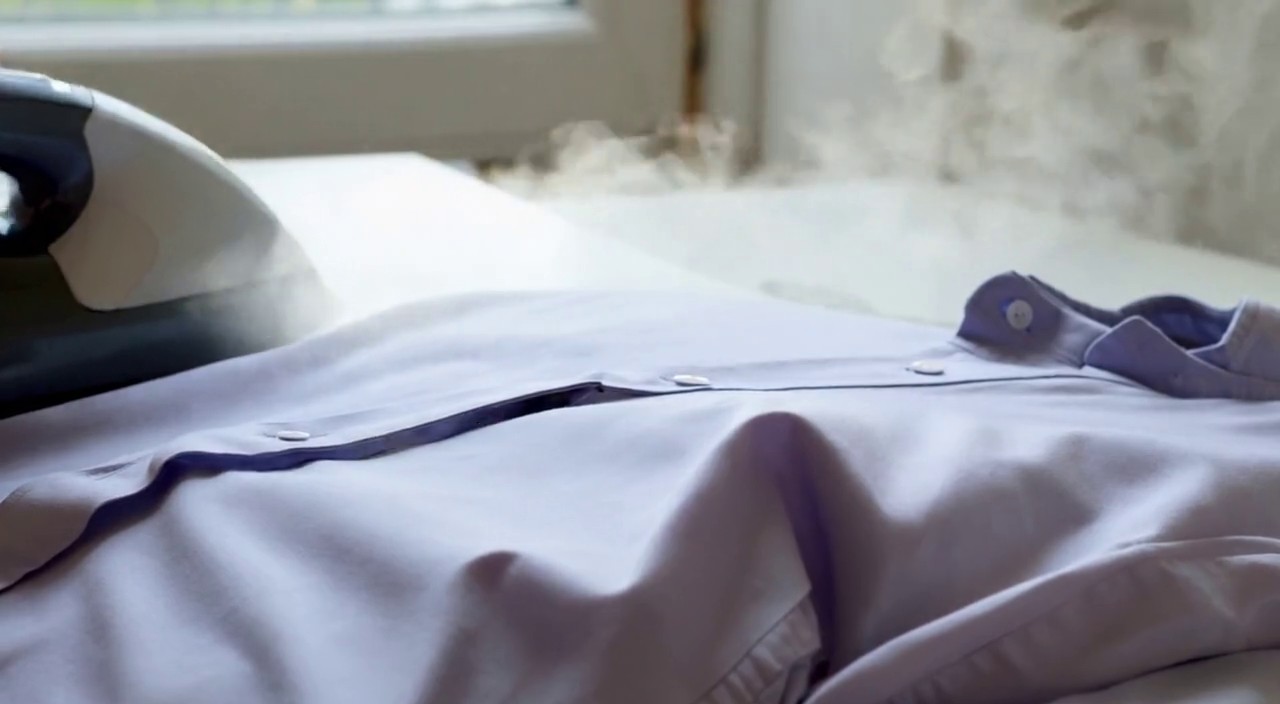} &
        \includegraphics[width=0.195\textwidth]{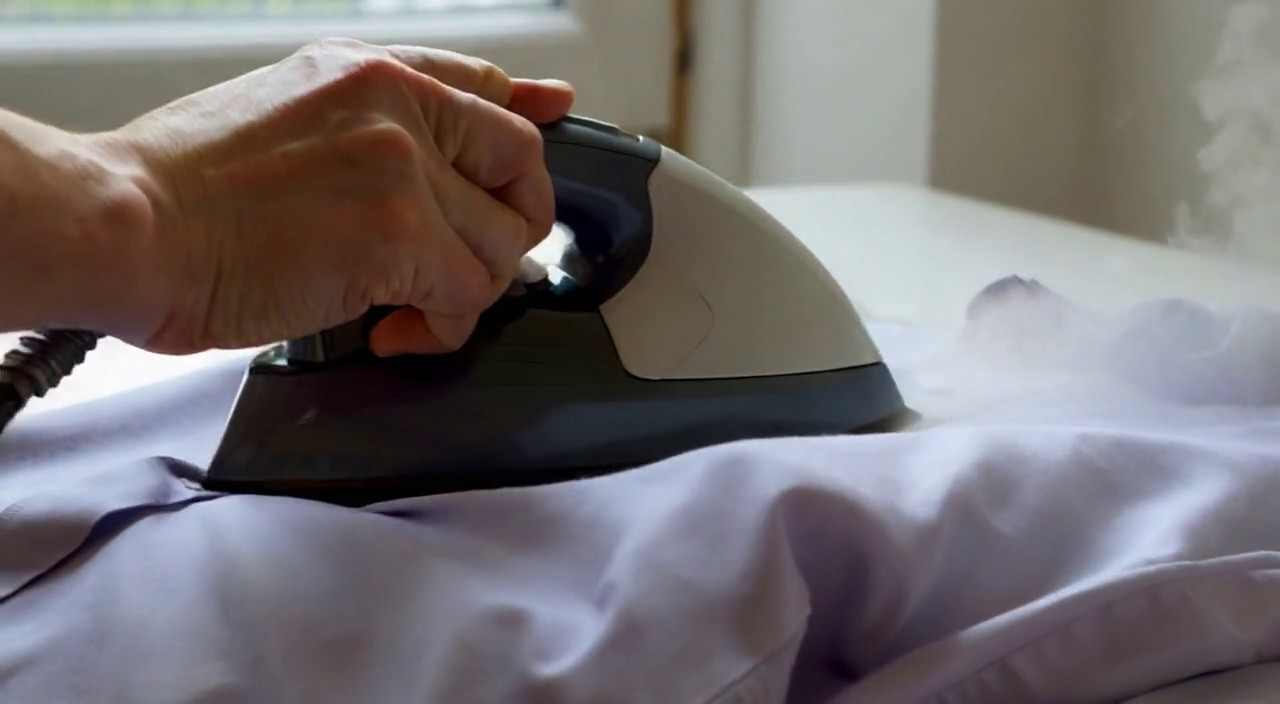} &
        \includegraphics[width=0.195\textwidth]{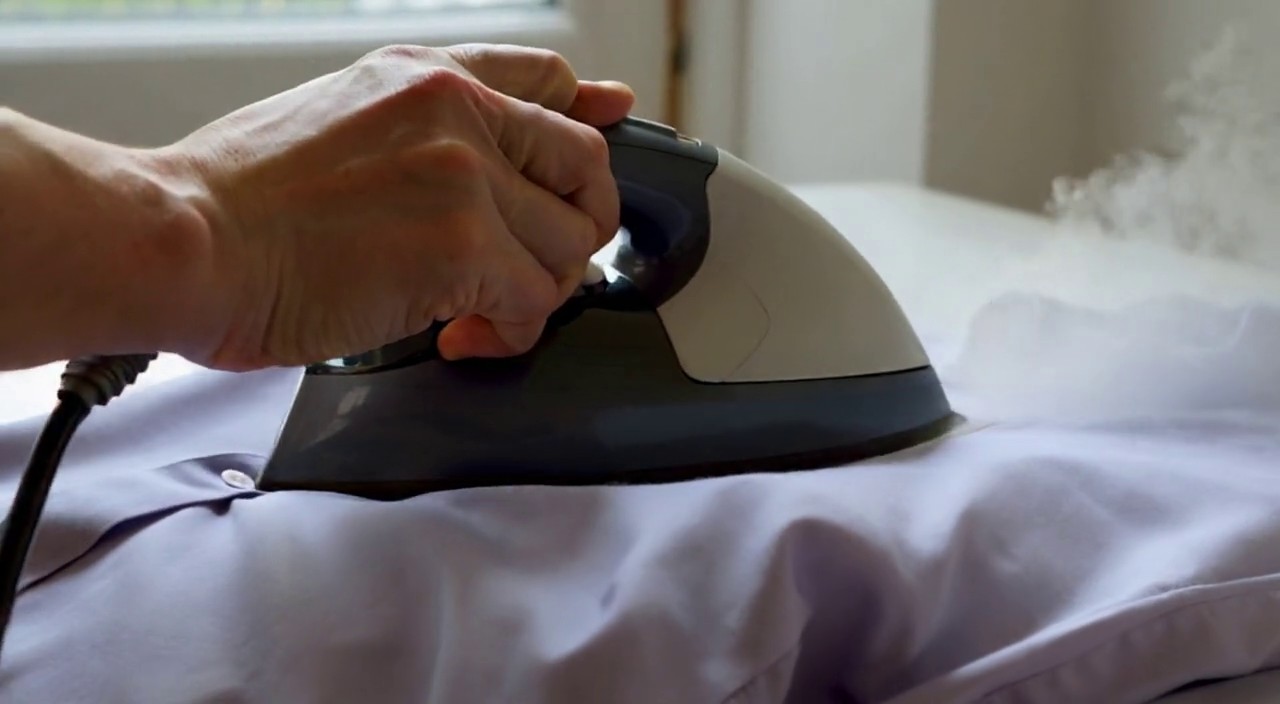} &
        \includegraphics[width=0.195\textwidth]{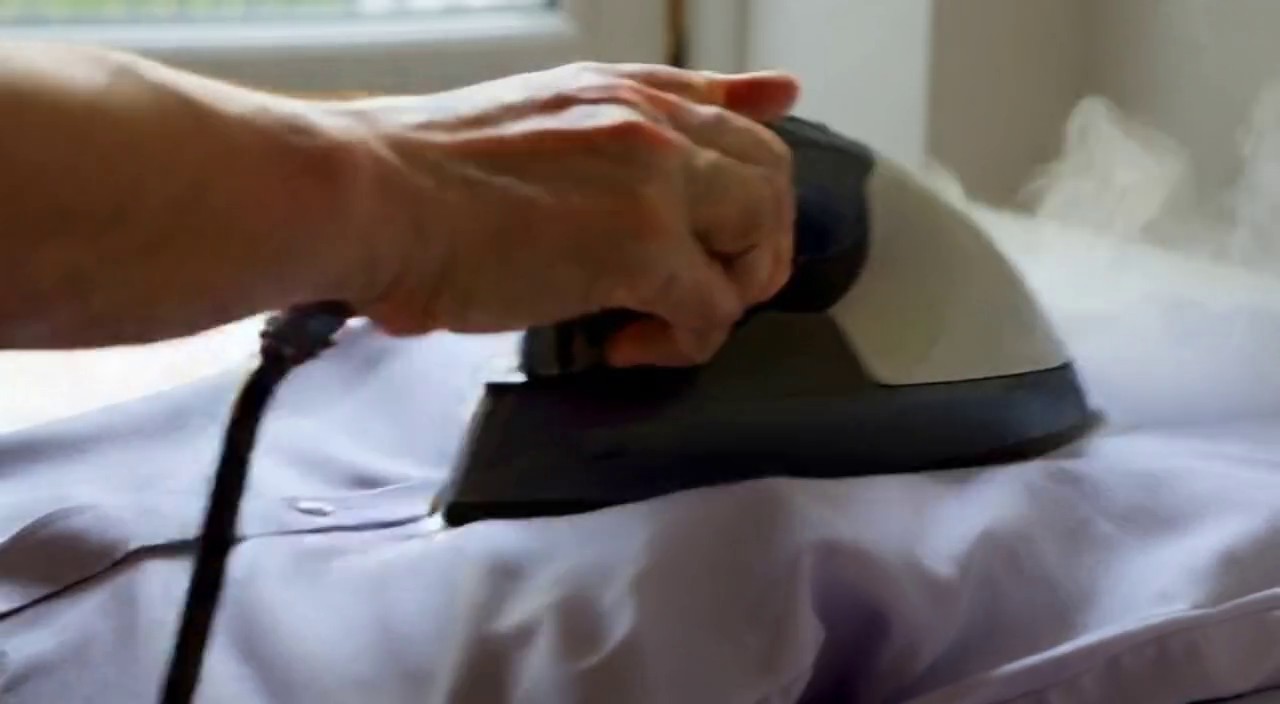} \\[4pt]

        {\rotatebox{90}{\hspace{12pt} 14B}} &
        \includegraphics[width=0.195\textwidth]{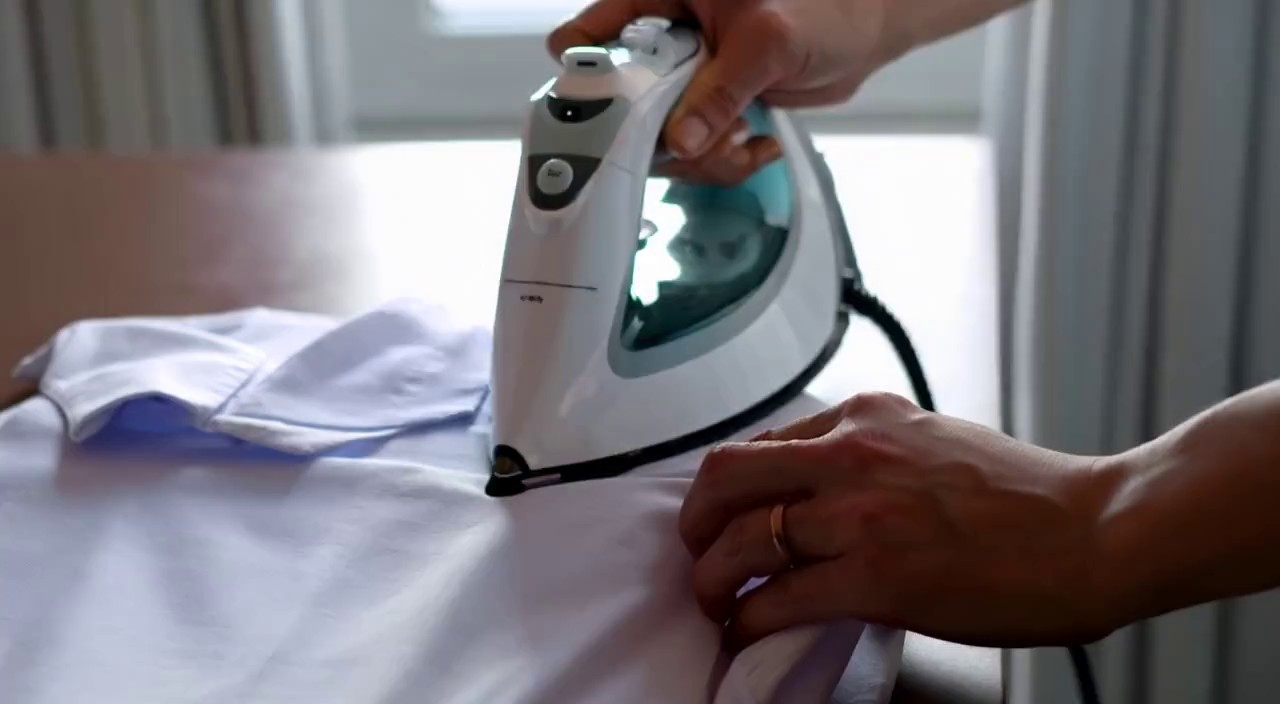} &
        \includegraphics[width=0.195\textwidth]{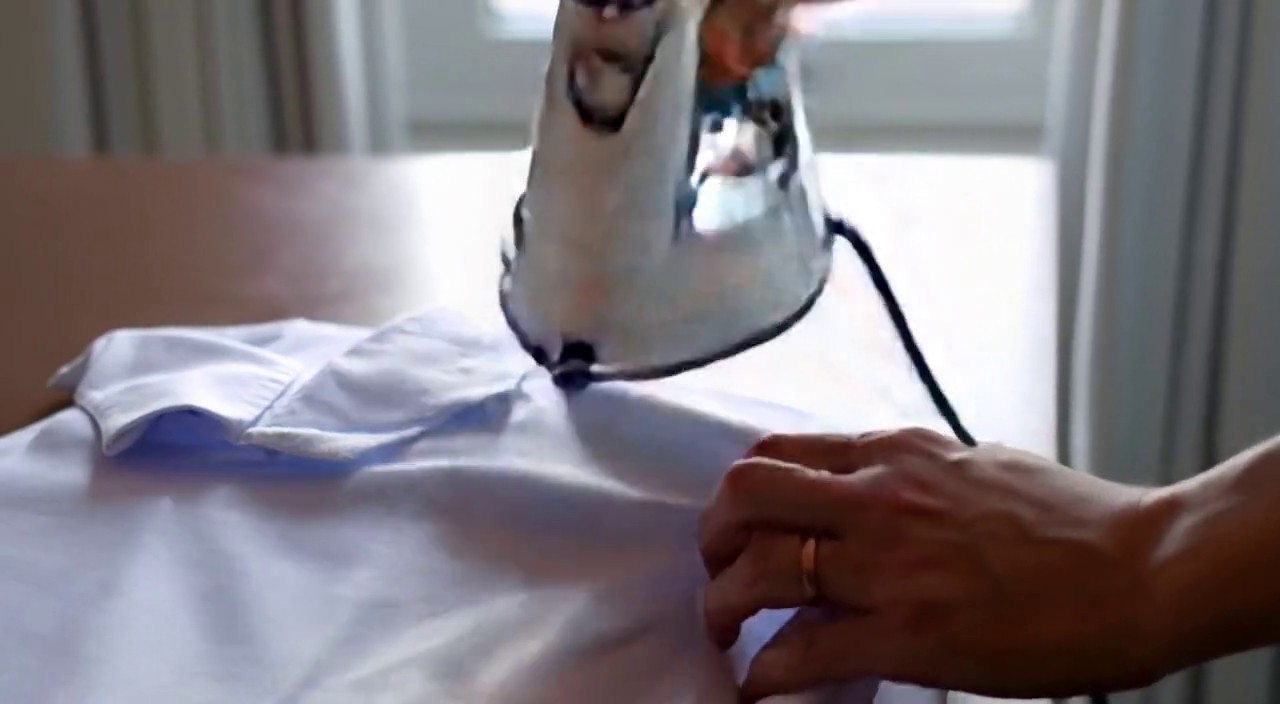} &
        \includegraphics[width=0.195\textwidth]{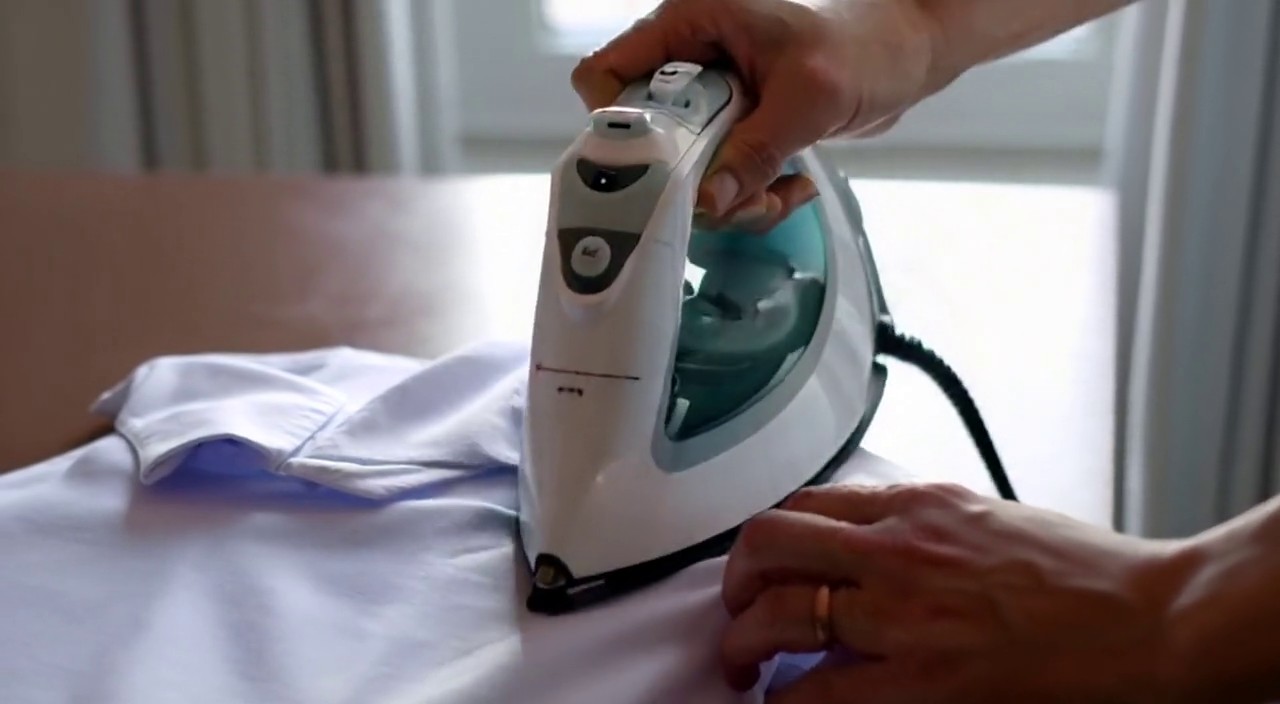} &
        \includegraphics[width=0.195\textwidth]{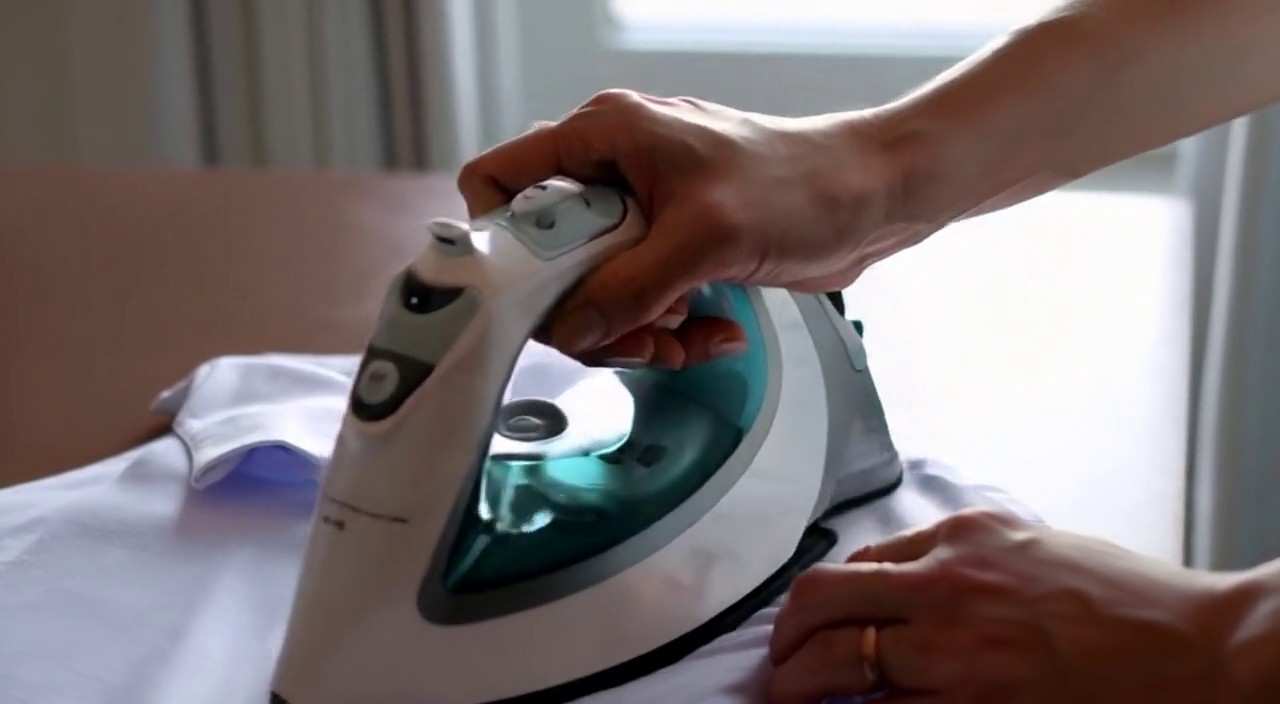} &
        \includegraphics[width=0.195\textwidth]{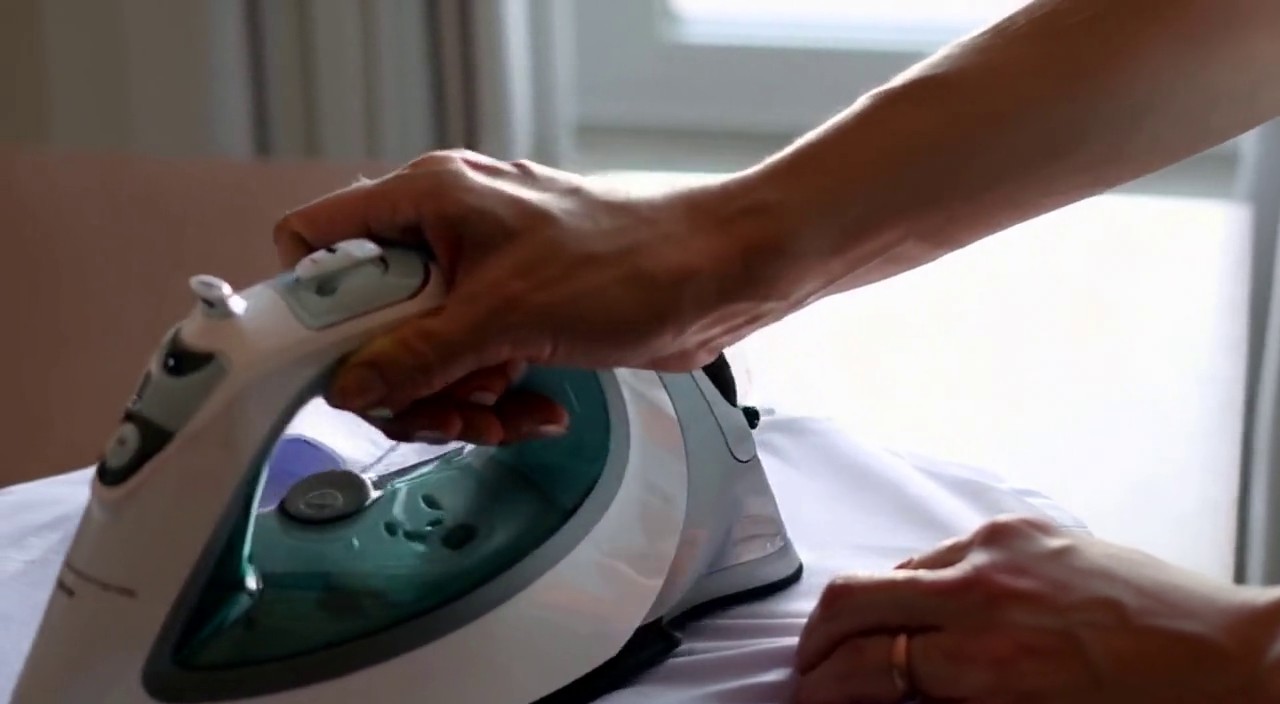} \\[4pt]

        \multicolumn{6}{p{\textwidth}}{
        \begin{multicoltextblock}
        \prompttext{Prompt}:
        \prompt{Hands firmly grasp the handle of a steam iron, expertly gliding it over a wrinkled shirt. With each pass, the iron releases gentle clouds of steam, effortlessly smoothing the fabric and erasing wrinkles to reveal a crisp, neat finish. The iron moves with precision and care, transforming the shirt with each stroke. A subtle scent of fresh linen permeates the air, adding to the serene ambiance. Soft light filters through a nearby window, highlighting the fabric's newly smooth texture and creating a tranquil atmosphere as this meticulous task unfolds.}
        \end{multicoltextblock}
        } \\
        & \\ [1pt]
    \end{tabular}}
    \caption{\textbf{Generated videos from Cosmos-Predict1-7B-Text2world and Cosmos-Predict1-14B-Text2world.} Both the Text2World models produce videos of high visual quality, motion dynamics and text alignment. Notably, compared to the 7B model, the 14B model demonstrates an enhanced ability to capture finer visual details and more intricate motion patterns. To check full videos and more video examples, please visit our \href{https://research.nvidia.com/labs/dir/cosmos1/}{website}.}
    \label{fig:cosmos_diffusion_text2world_qualitative_results}
\end{figure*}

\begin{figure*}[ht!]
    \centering
    \setlength{\tabcolsep}{1.5pt}
    \renewcommand{\arraystretch}{0}
    \resizebox{\textwidth}{!}{%
    \begin{tabular}{cccccc} %
    & Condition frame 0 & Frame 29 & Frame 59 & Frame 89 & Frame 120 \\[4pt]
        {\rotatebox{90}{\hspace{18pt} 7B}} &
        \includegraphics[width=0.195\textwidth]{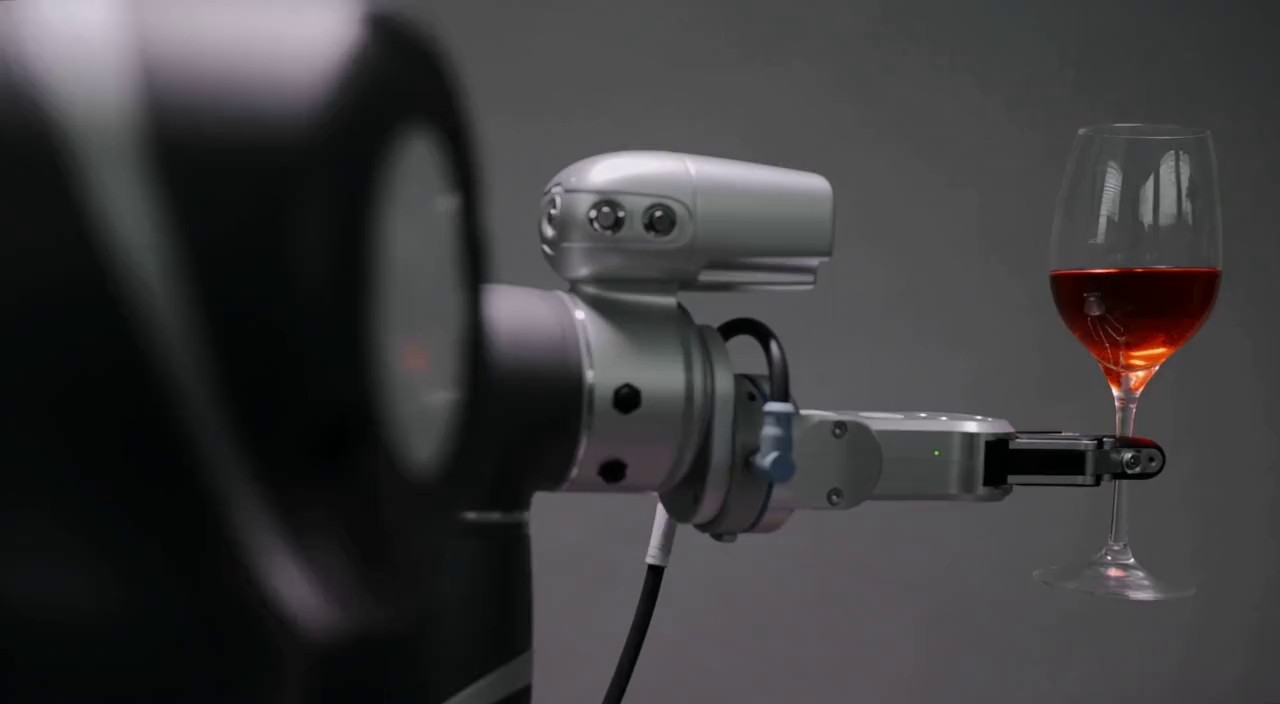} &
        \includegraphics[width=0.195\textwidth]{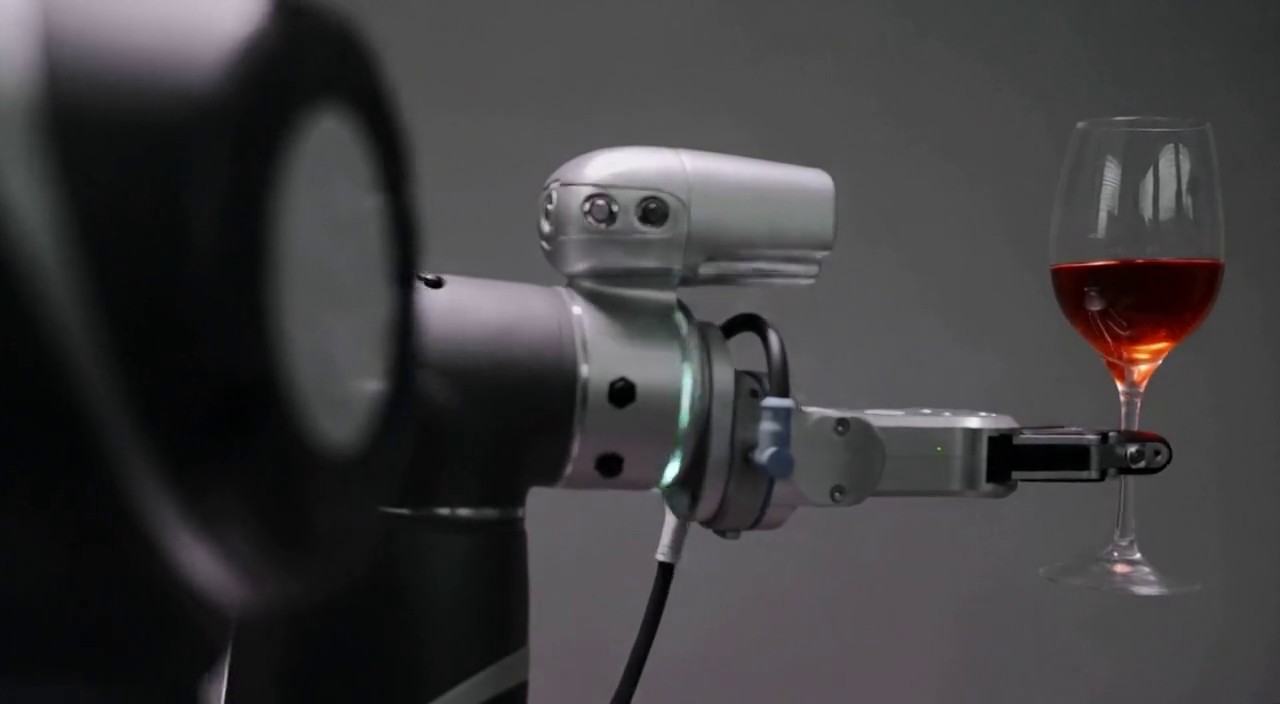} &
        \includegraphics[width=0.195\textwidth]{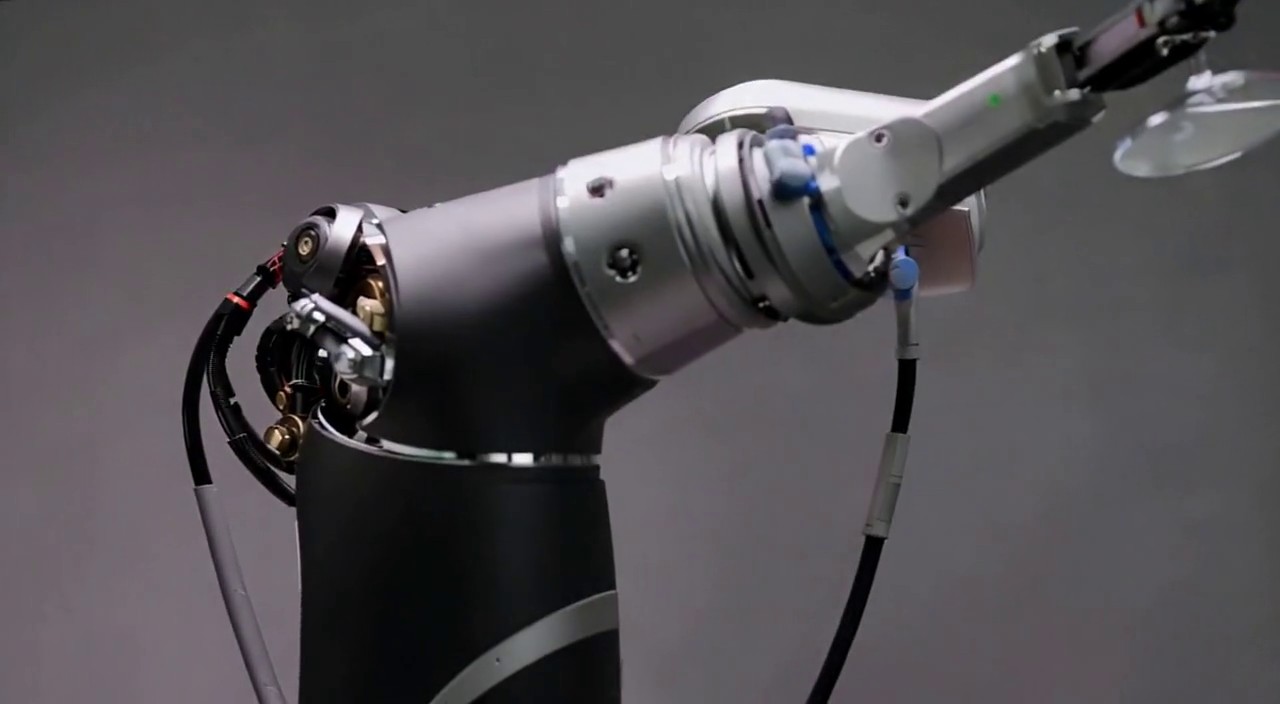} &
        \includegraphics[width=0.195\textwidth]{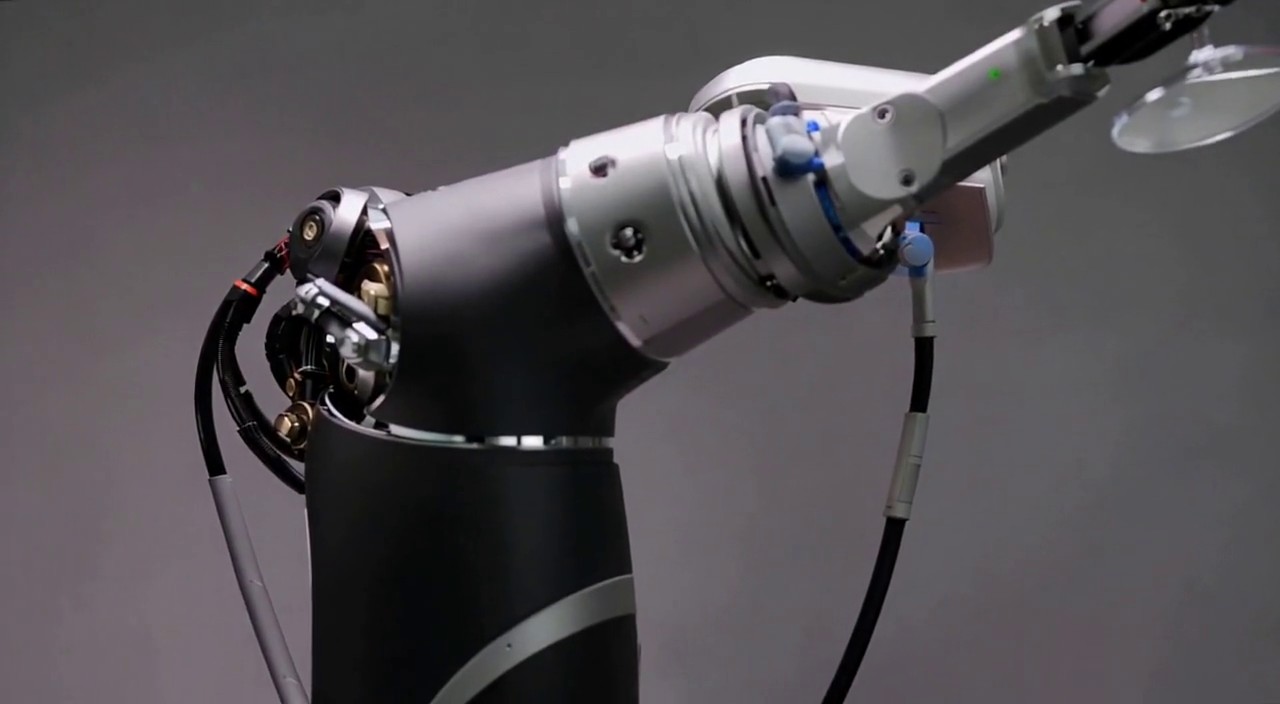} &
        \includegraphics[width=0.195\textwidth]{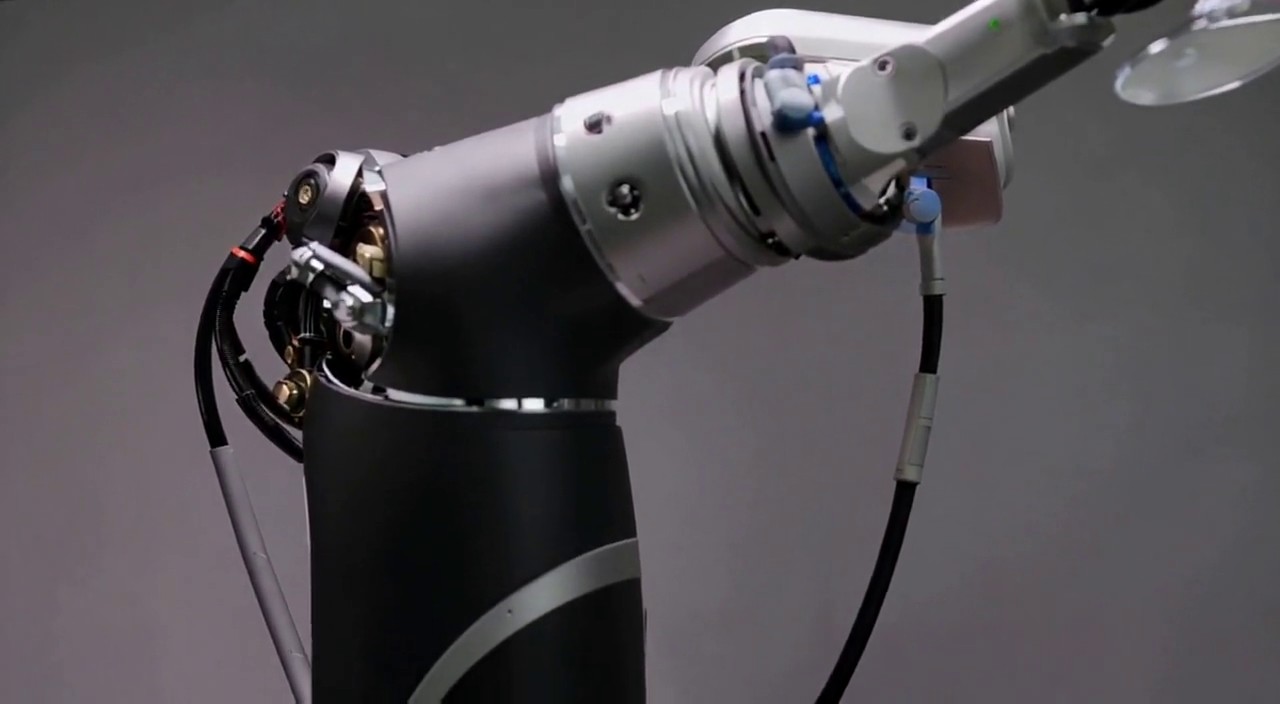} \\[4pt]

        {\rotatebox{90}{\hspace{14pt} 14B}} &
        \includegraphics[width=0.195\textwidth]{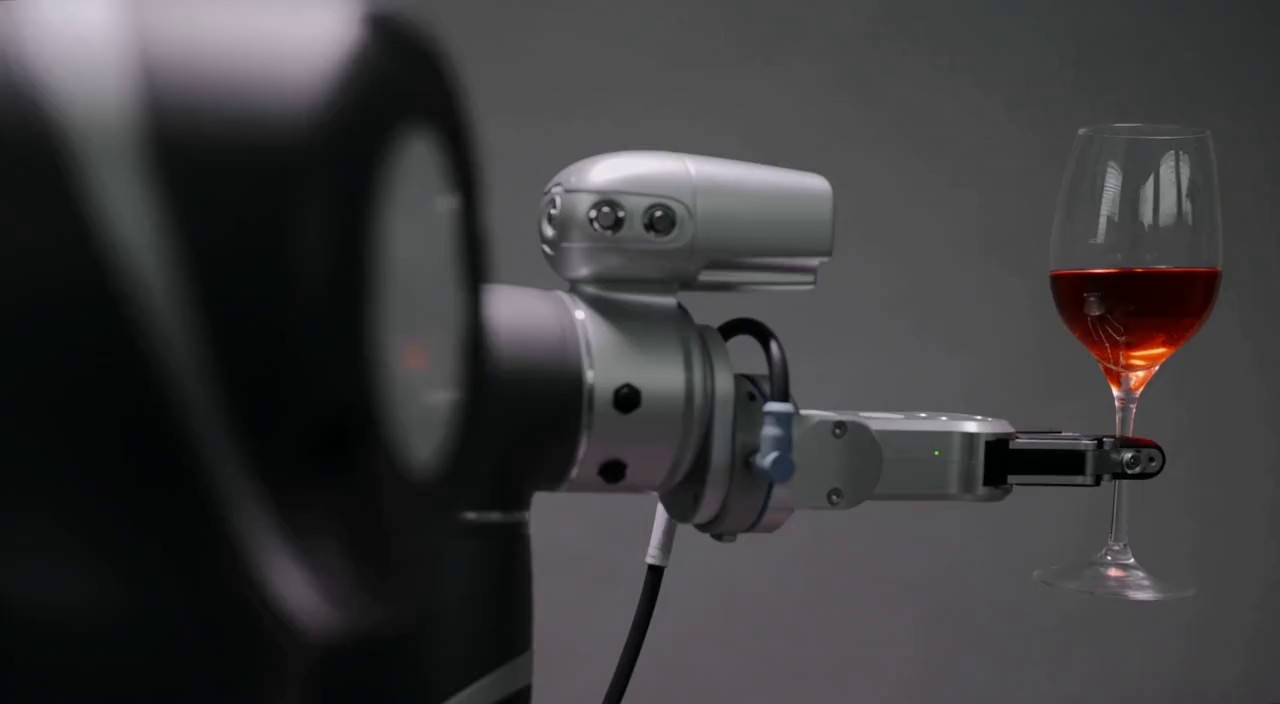} &
        \includegraphics[width=0.195\textwidth]{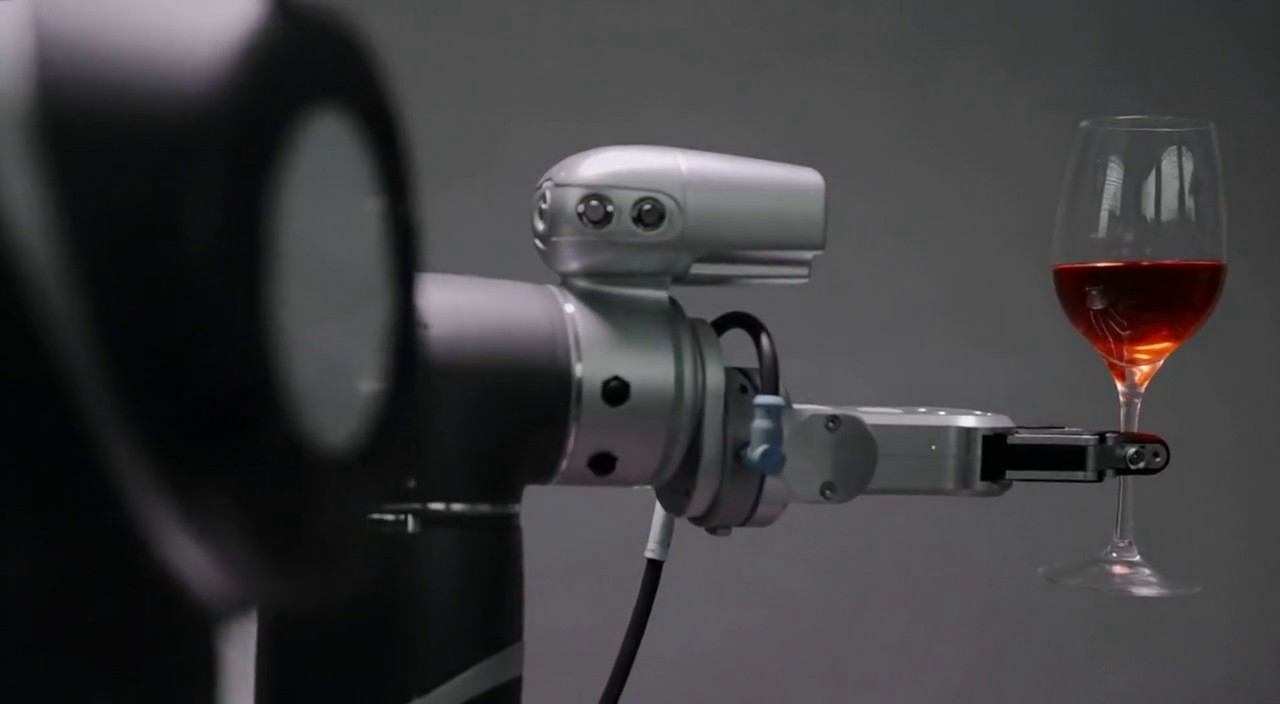} &
        \includegraphics[width=0.195\textwidth]{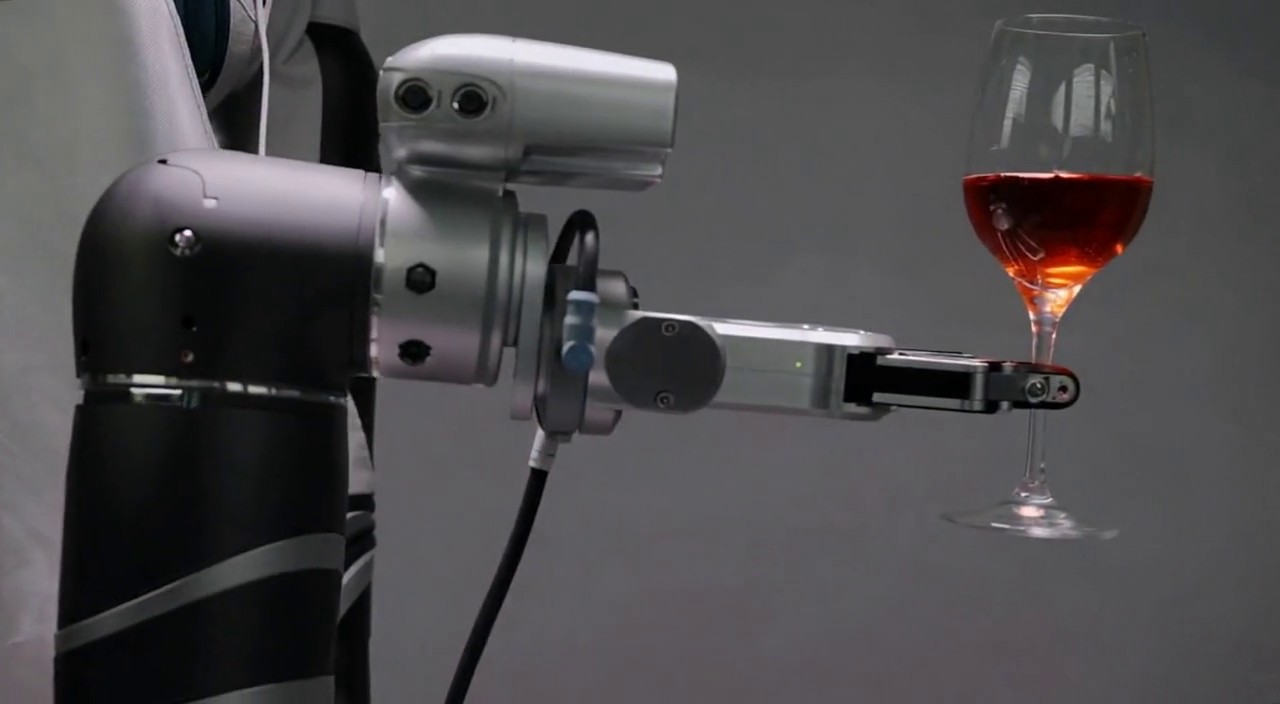} &
        \includegraphics[width=0.195\textwidth]{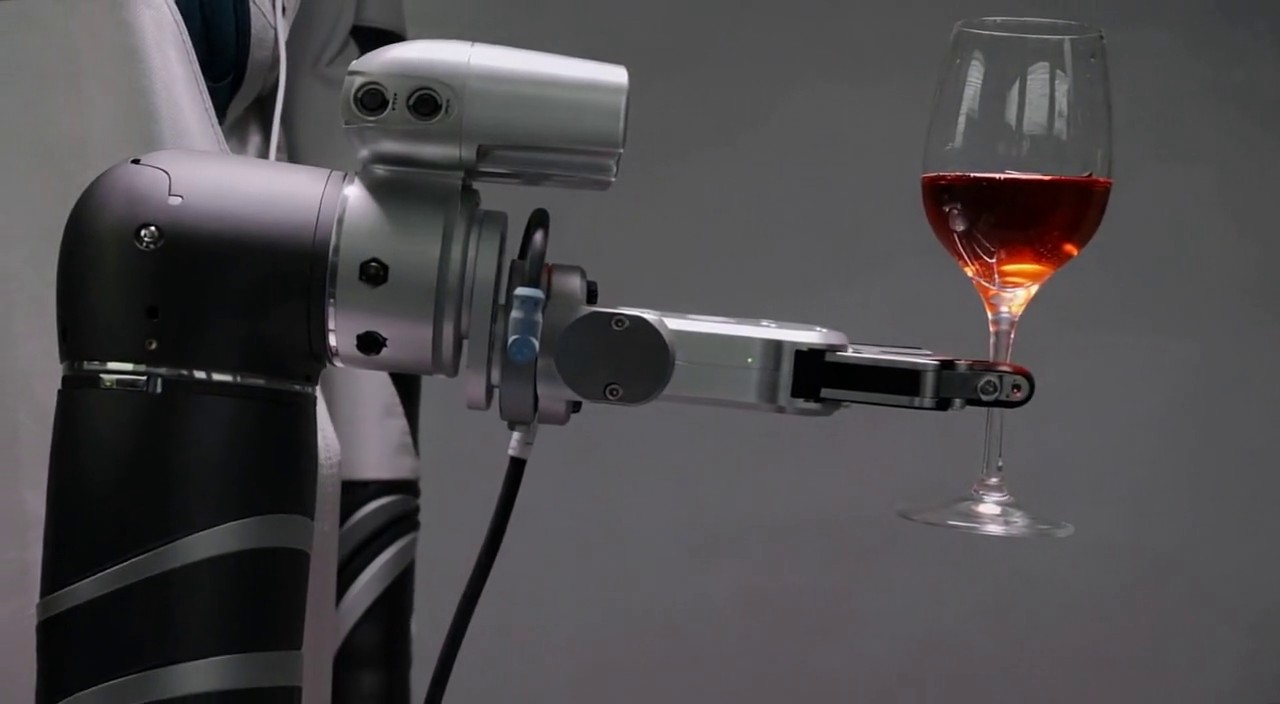} &
        \includegraphics[width=0.195\textwidth]{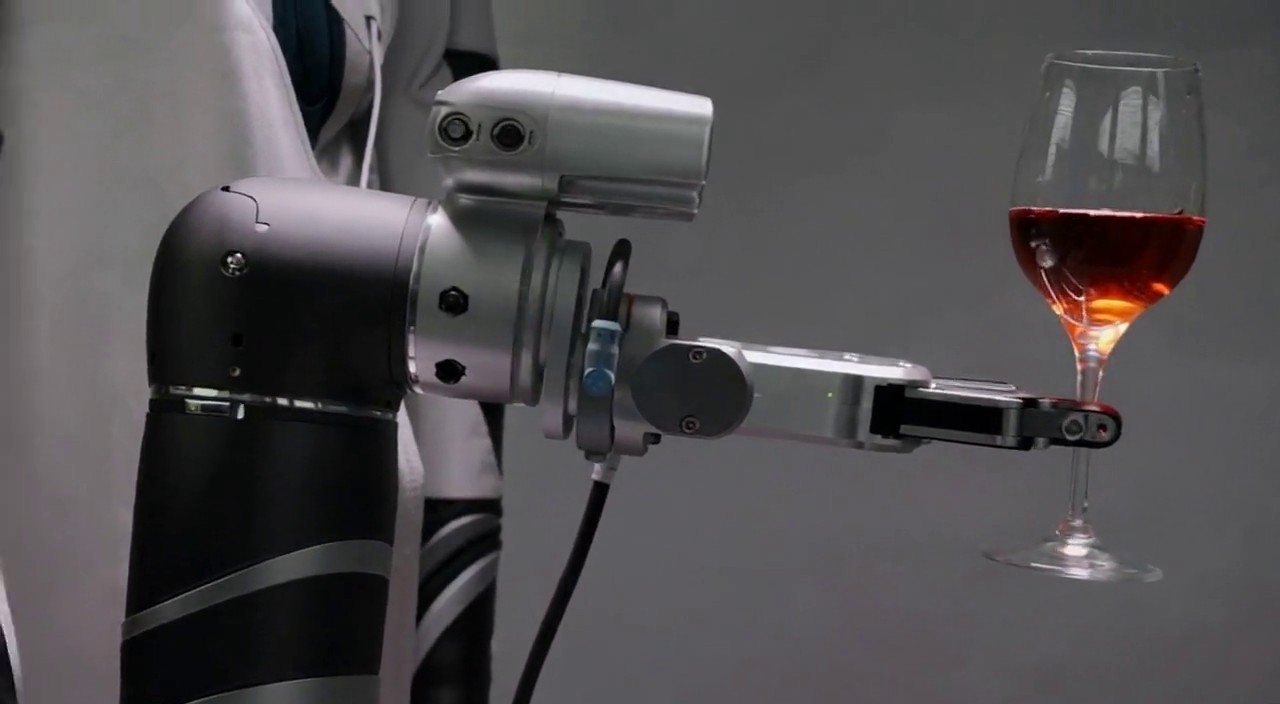} \\[4pt]

        \multicolumn{6}{p{\textwidth}}{
        \begin{multicoltextblock}
        \prompttext{Prompt}:
        \prompt{The video depicts a robotic arm holding a wine glass filled with red wine. The robotic arm, equipped with multiple joints and mechanical components, appears to be designed for precision tasks. The glass is held delicately, showcasing the robot's capability to handle fragile objects. The background is minimalistic, emphasizing the interaction between the robot and the wine glass.}
        \end{multicoltextblock}
        } \\
        & \\ [18pt]

    & Condition frame 0 & Frame 150 & Frame 300 & Frame 450 & Frame 680 \\[4pt]
        {\rotatebox{90}{\hspace{18pt} 7B}} &
        \includegraphics[width=0.195\textwidth]{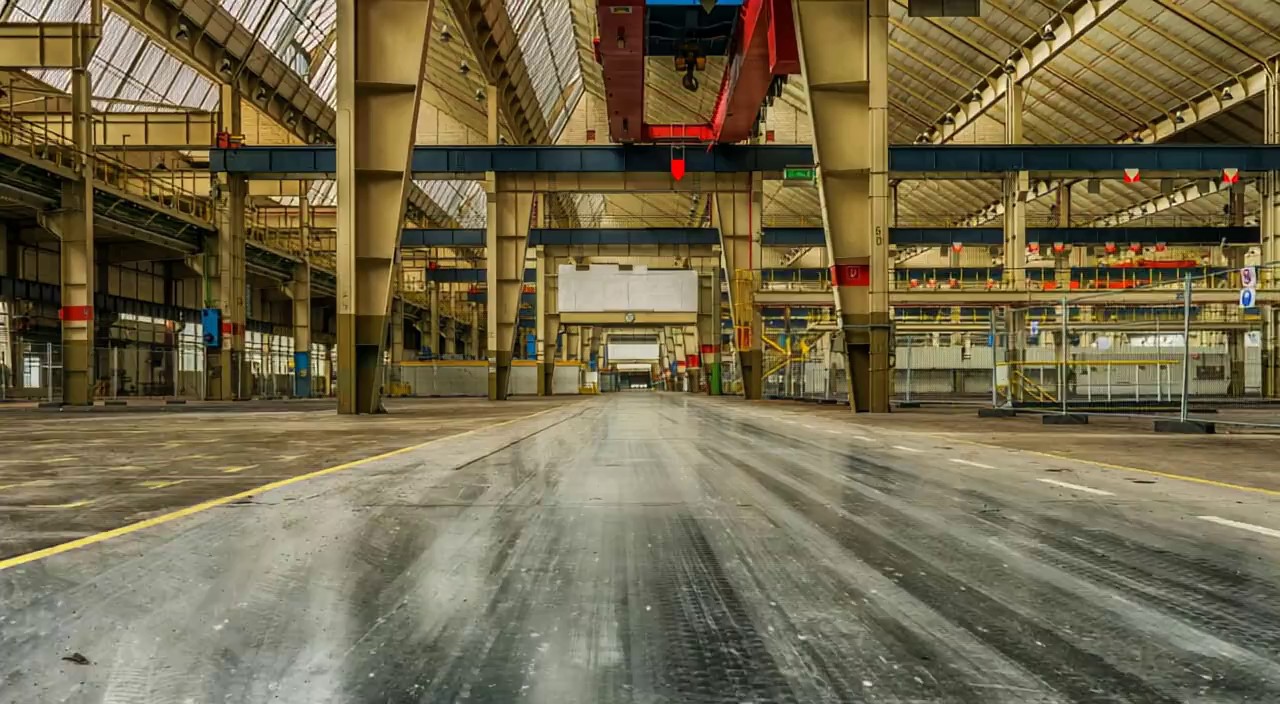} &
        \includegraphics[width=0.195\textwidth]{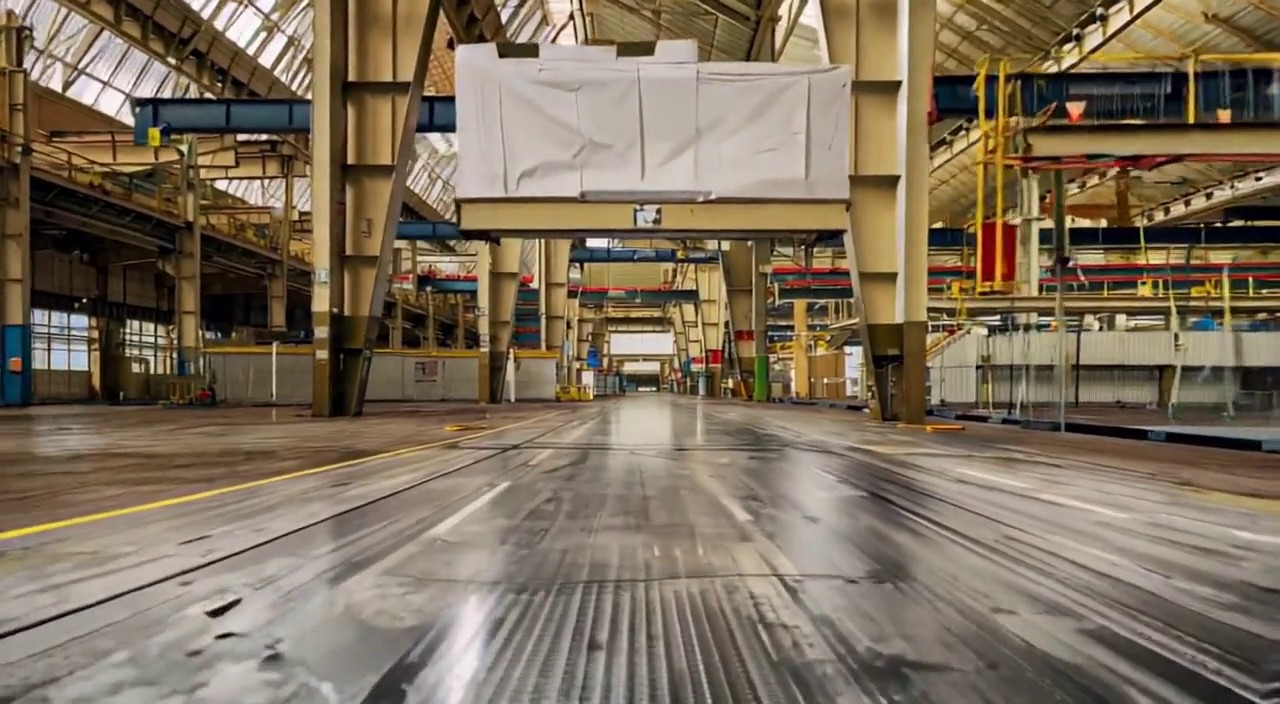} &
        \includegraphics[width=0.195\textwidth]{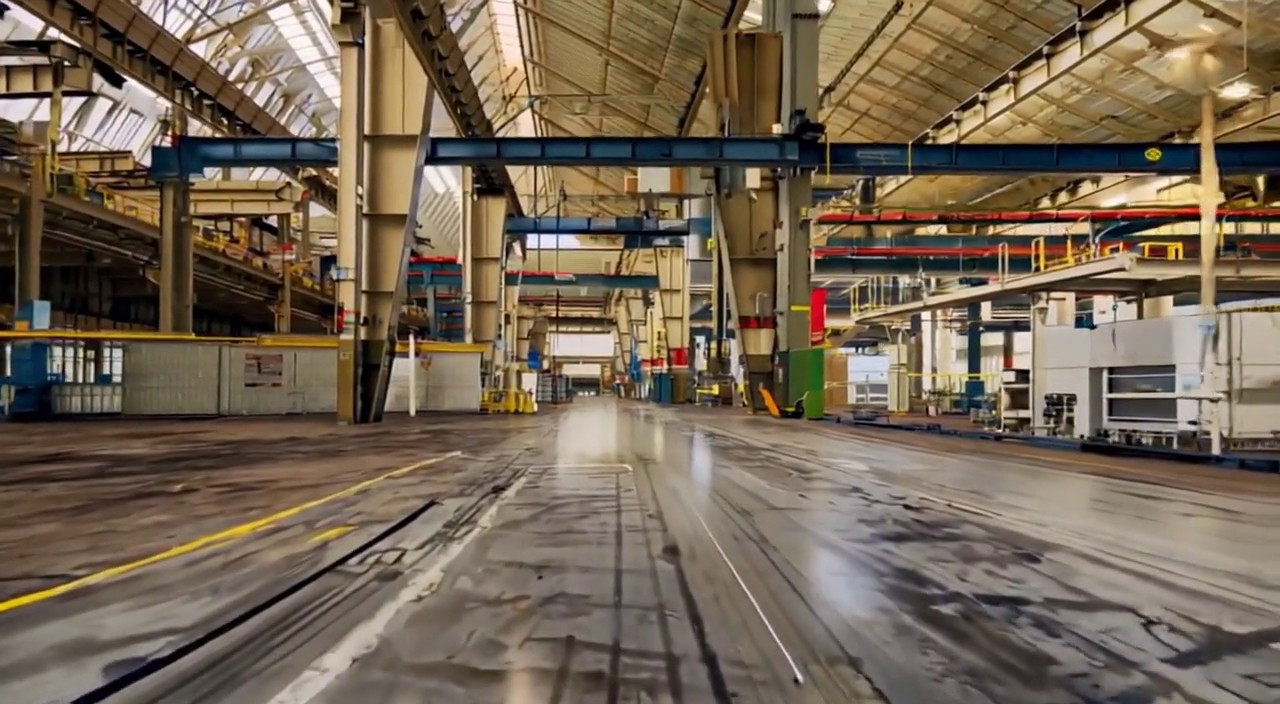} &
        \includegraphics[width=0.195\textwidth]{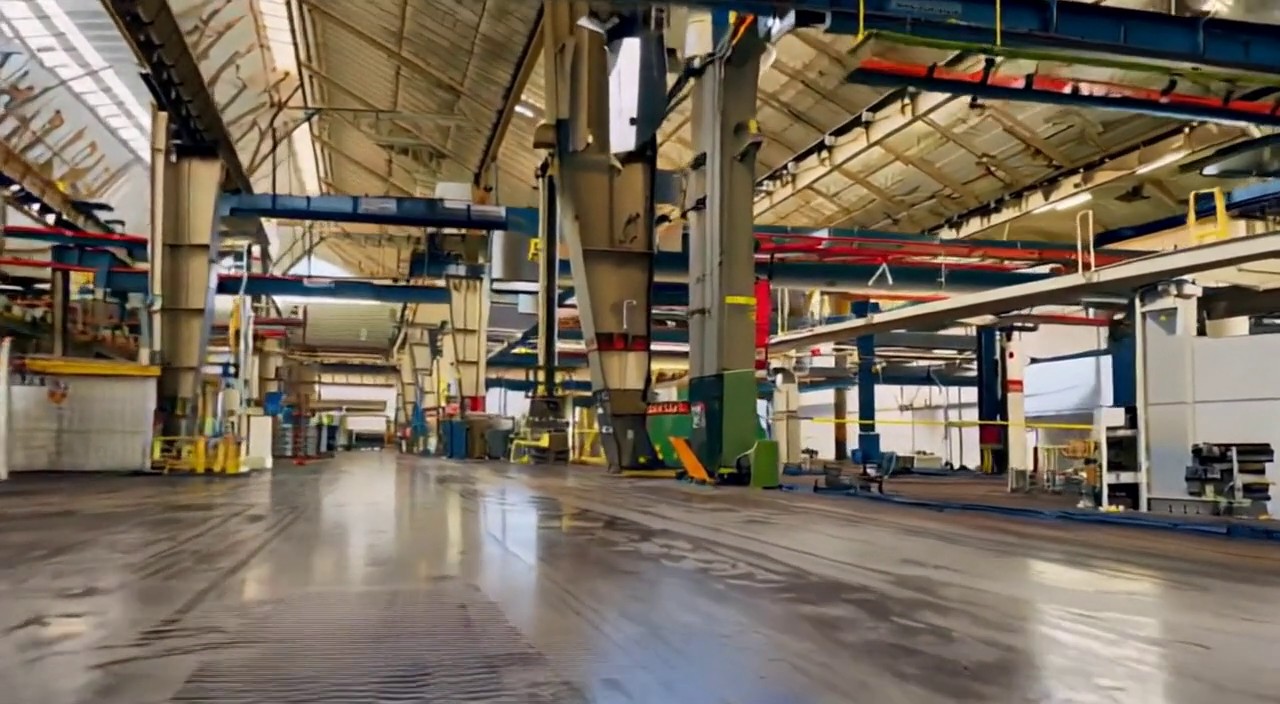} &
        \includegraphics[width=0.195\textwidth]{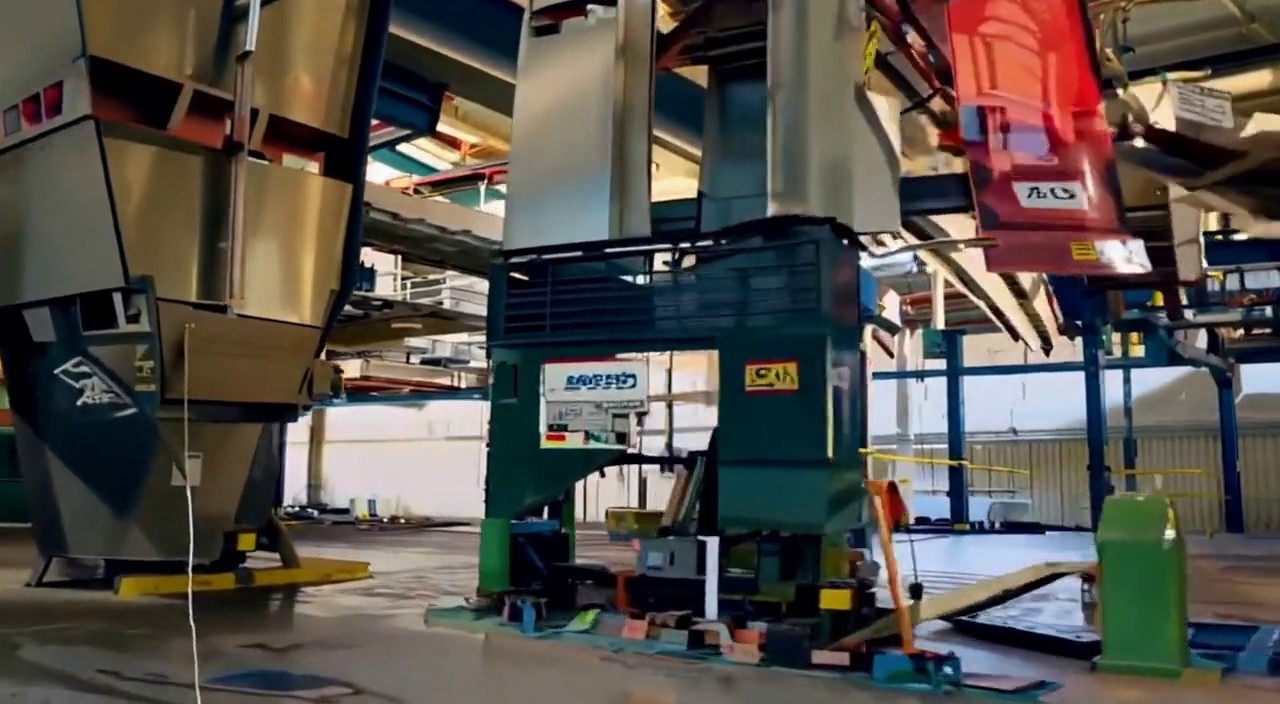} \\[4pt]

        {\rotatebox{90}{\hspace{14pt} 14B}} &
        \includegraphics[width=0.195\textwidth]{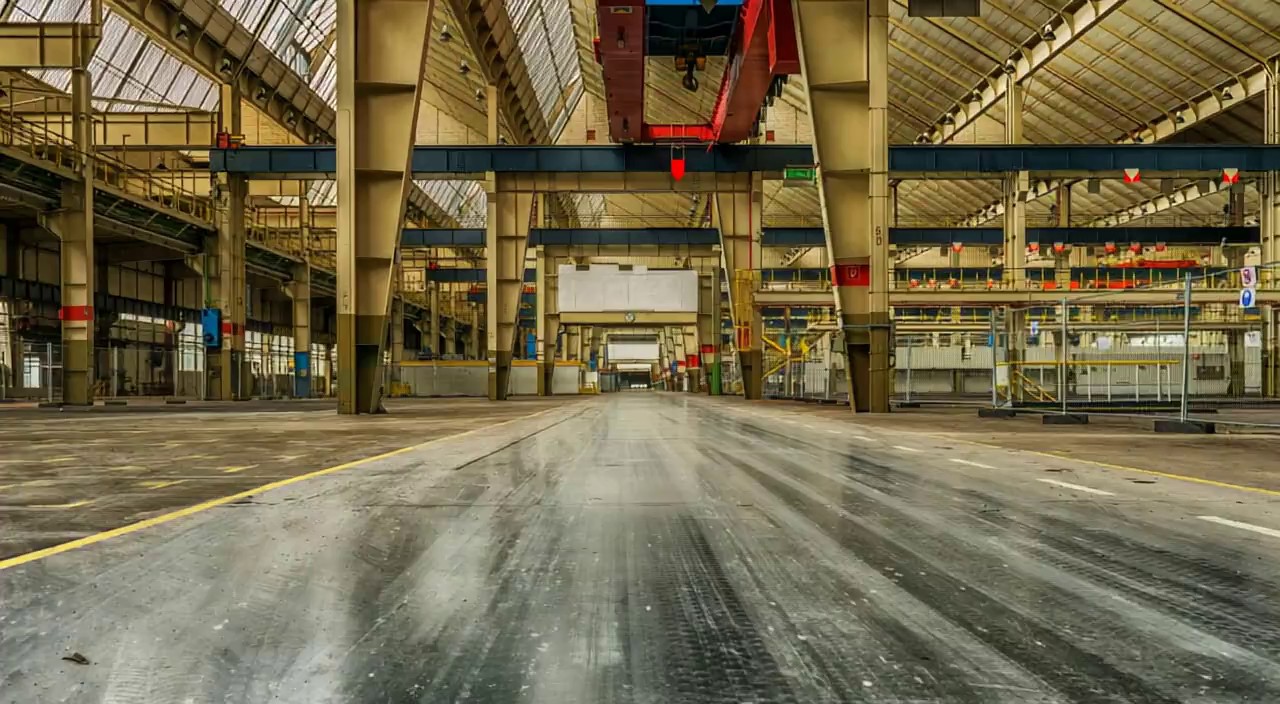} &
        \includegraphics[width=0.195\textwidth]{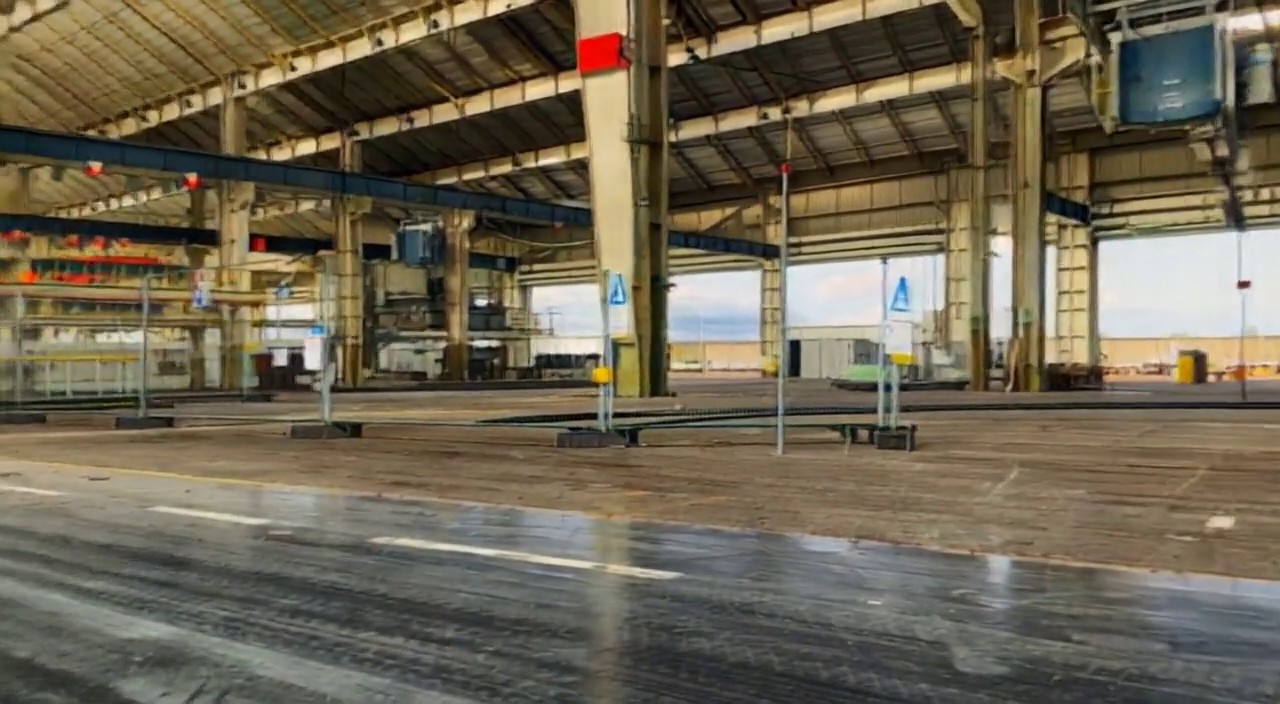} &
        \includegraphics[width=0.195\textwidth]{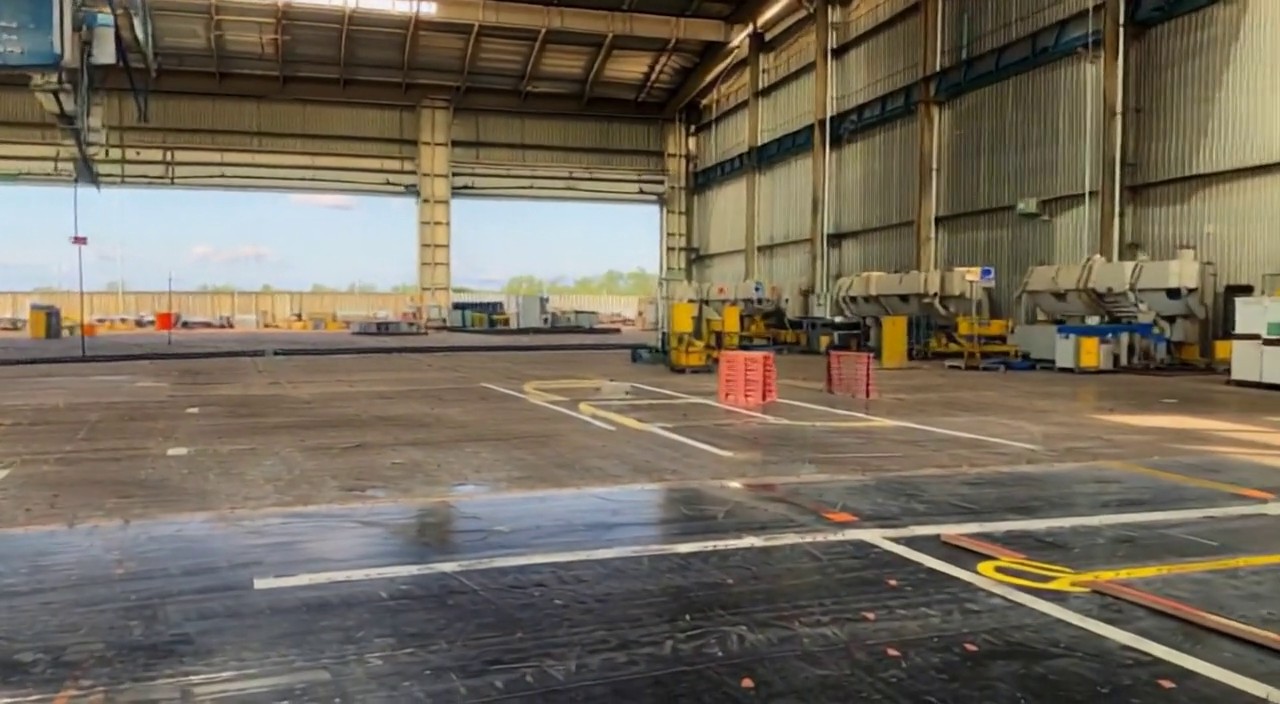} &
        \includegraphics[width=0.195\textwidth]{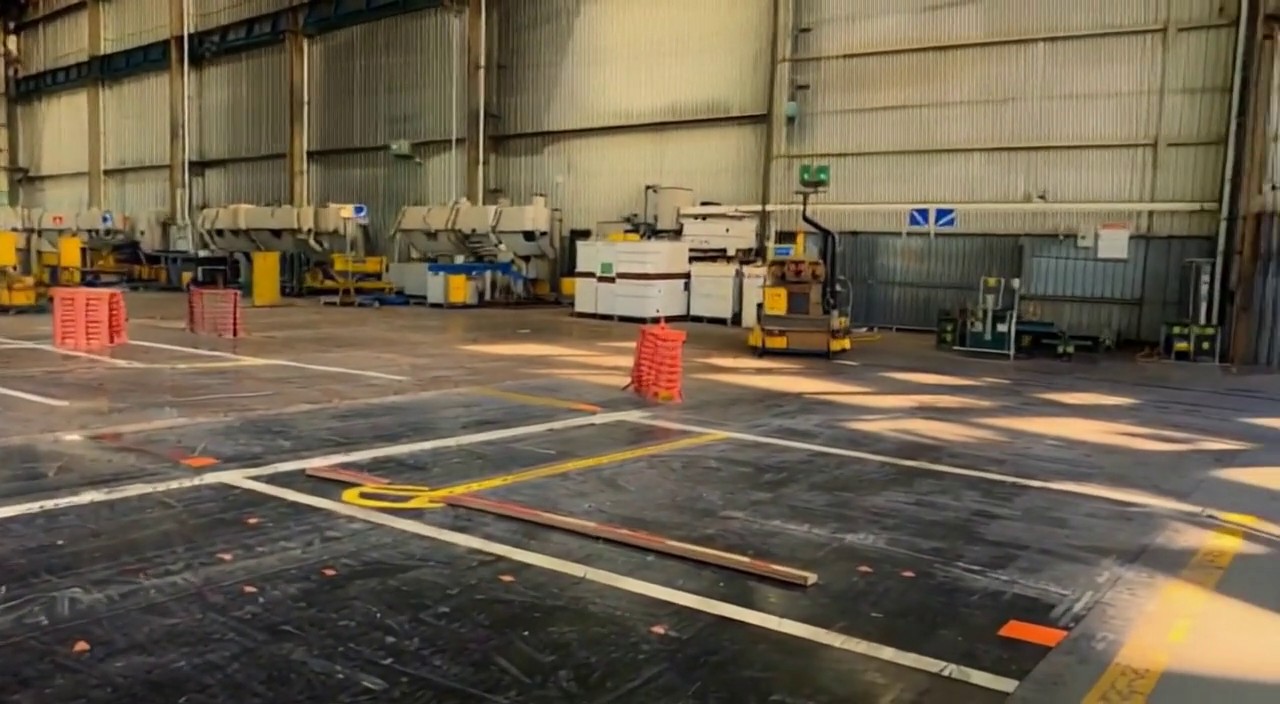} &
        \includegraphics[width=0.195\textwidth]{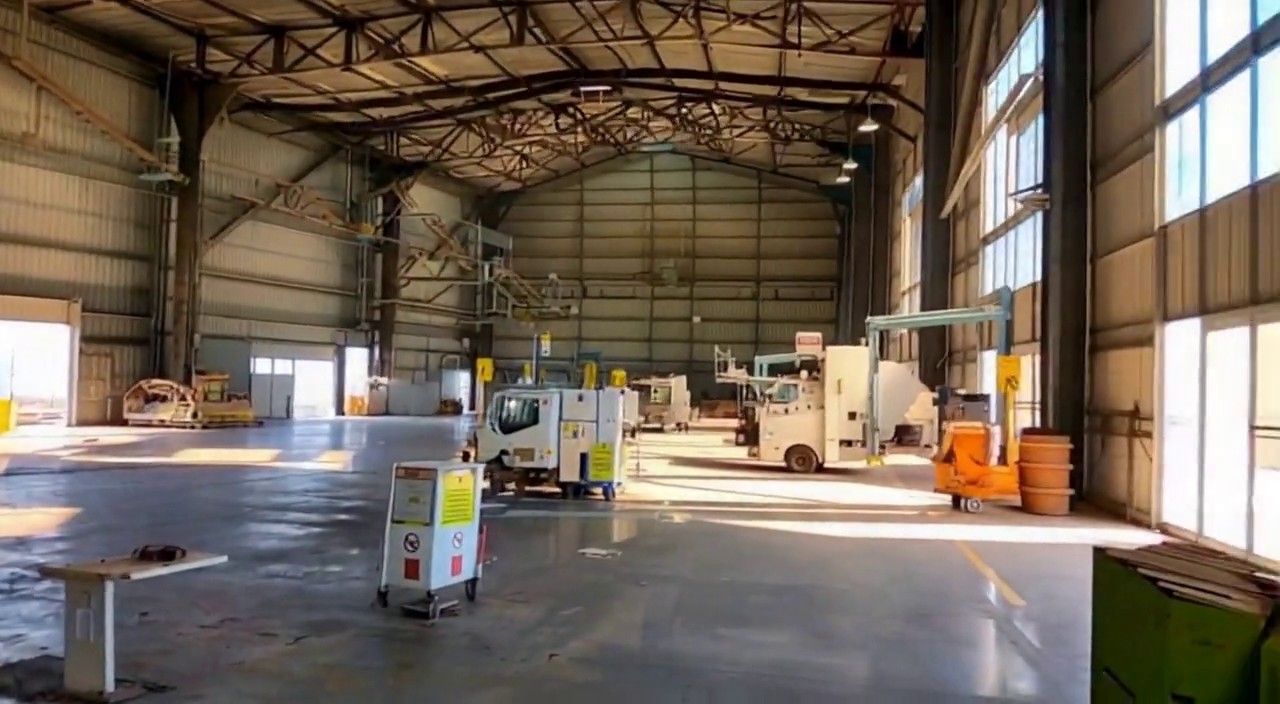} \\[4pt]

        \multicolumn{6}{p{\textwidth}}{
        \begin{multicoltextblock}
        \prompttext{Prompt}:
        \prompt{The video depicts the interior of a large industrial facility, likely a factory or warehouse. The space is expansive with high ceilings and metal framework. Overhead cranes and various machinery are visible, indicating a setting for heavy manufacturing or assembly. The floor is mostly empty, with some scattered debris and marked lines. Safety signs and barriers are present, emphasizing the industrial environment. The lighting is natural, streaming through the high windows, illuminating the workspace.}
        \end{multicoltextblock}
        } \\
        & \\ [4pt]

    \end{tabular}}
    \caption{\textbf{Generated videos from Cosmos-Predict1-7B-Video2world and Cosmos-Predict1-14B-Video2world.} The top two rows show 5-second videos generated by the models, conditioned on the first 9 frames. The bottom two rows show long video generation results. We generate long videos in an autoregressive manner, where the first generated video is conditioned on a single input image, and the subsequent five videos are conditioned on their previous nine frames. Both of the 7B and 14B models produce photorealistic videos with high visual fidelity. The 14B model demonstrates the ability to generate more complex scenes and exhibits superior motion stability. To check full videos and more video examples, please visit our \href{https://research.nvidia.com/labs/dir/cosmos1/}{website}.}
    \label{fig:cosmos_diffusion_video2world_qualitative_results}
\end{figure*}

\noindent \textbf{Comparison with large language models.} Unlike LLMs, which are typically pre-trained with shorter context lengths, long-context settings significantly increase FLOPs due to the quadratic cost of self-attention.
While FLOPs for LLMs are commonly calculated as $6 \times \text{seq\_len} \times P$, where $P$ is the number of parameters~\citep{kaplan2020scaling}, we note that this formula is inaccurate for our diffusion WFMs. We provide the forward pass FLOPs of each key operation in \cref{tab:diffusion_transformer_flops_memory}.\subsubsection{Prompt Upsampler}\label{sec::prompt_upsampler}

During training, our WFMs use detailed video descriptions as input text prompts to produce high-quality videos. However, during inference, user prompts may vary in length, structure, and style, often being much shorter. To bridge this gap between training and inference text prompts, we develop a prompt upsampler to transform original input prompts into more detailed and enriched versions. It can improve the prompts by adding more details and maintaining a consistent description structure, which leads to higher quality output.

The main requirements for the prompt upsampler are:
\begin{itemize}
\item \textbf{Fidelity to the input prompts}: The upsampled prompt must faithfully preserve the key elements of the original user input, including the main characters, actions or motions, key attributes, and overall intent.
\item \textbf{Alignment with training distribution}: The upsampled prompt should closely resemble the distribution of training prompts of WFMs in terms of length, language structure, and style.
\item \textbf{Enhanced visual details}: The upsampled prompt should be designed to prompt the WFMs to generate more accurate imagery.
\end{itemize}

\noindent \textbf{Prompt upsampler for Text2World model.} We fine-tune Mistral-NeMo-12B-Instruct~\citep{mistral_nemo_2024} to build our prompt upsampler. To obtain paired data, that is, short prompts simulating user input and the corresponding long prompts reflecting the distribution of training prompts, we use a VLM to generate short captions based on our training long prompts and corresponding videos. This long-to-short data creation strategy is effective in (1) preserving the authentic video content and distribution from detailed training prompts of WFMs and (2) ensuring fidelity between the short and long prompts. The resulting prompt upsampler is termed Cosmos-UpsamplePrompt1-12B-Text2World.

\noindent \textbf{Prompt upsampler for Video2World model.} For the Video2World model, the input consists of video conditions and a user text prompt. To enhance the user prompt, we utilize an open-source VLM, Pixtral-12B~\citep{agrawal2024pixtral}, combined with zero-shot prompt engineering, to upsample the prompt into a detailed description that considers both the video conditions and the user prompt. We found the vanilla Pixtral-12B model works well out of the box and did not proceed to perform a similar fine-tuning described above.

\subsubsection{Results}

In~\cref{fig:cosmos_diffusion_text2world_qualitative_results}, we present qualitative results generated by our Cosmos-Predict1-7B-Text2World and Cosmos-Predict1-14B-Text2World models. Both models produce videos of high visual quality, motion dynamics, and text alignment. Compared to the 7B model, the 14B model is able to generate videos capturing more complex visual details and intricate motions.

We show generated videos from Video2World 7B and 14B models in~\cref{fig:cosmos_diffusion_video2world_qualitative_results}. The Video2World models support both image and video conditioning and can generate extended videos in an autoregressive manner. As demonstrated in~\cref{fig:cosmos_diffusion_video2world_qualitative_results}, our Video2World models produce photorealistic videos with good motion dynamics and visual fidelity. The 14B model, again, generates better videos in terms of scene richness and motion stability.

\subsection{Autoregressive-based World Foundation Model}\label{sec::autoregress}

In autoregressive WFMs, we formulate world simulation generation as a next-token prediction task similar to language modeling. We start by converting a video into a sequence of discrete video tokens $\mathcal{V} = \{v_1, v_2, \dots, v_n\}$ using the Cosmos Discrete Tokenizer introduced in \cref{sec::token}. Then we train a Transformer decoder~\citep{vaswani2017attention} to predict the next video token using past video tokens as context, similar to large language models (LLMs)~\citep{gpt3,jiang2023mistral,dubey2024llama}. Specifically, the training objective is to minimize the following negative log-likelihood (NLL) loss:
\begin{equation}\label{eq:loss_nll}
    \mathcal{L}_{NLL} = \sum_{i} -\log P(v_i | v_1, v_2, \dots, v_{i-1}; \Theta),
\end{equation}
where the conditional probability $P$ of the predicted next video token $v_i$ is modeled by a Transformer decoder with parameters $\Theta$.

\subsubsection{Architecture} \label{sec:autoregressive:architecture}

\begin{figure}[tbh!]
    \centering
    \includegraphics[width=0.95\textwidth]{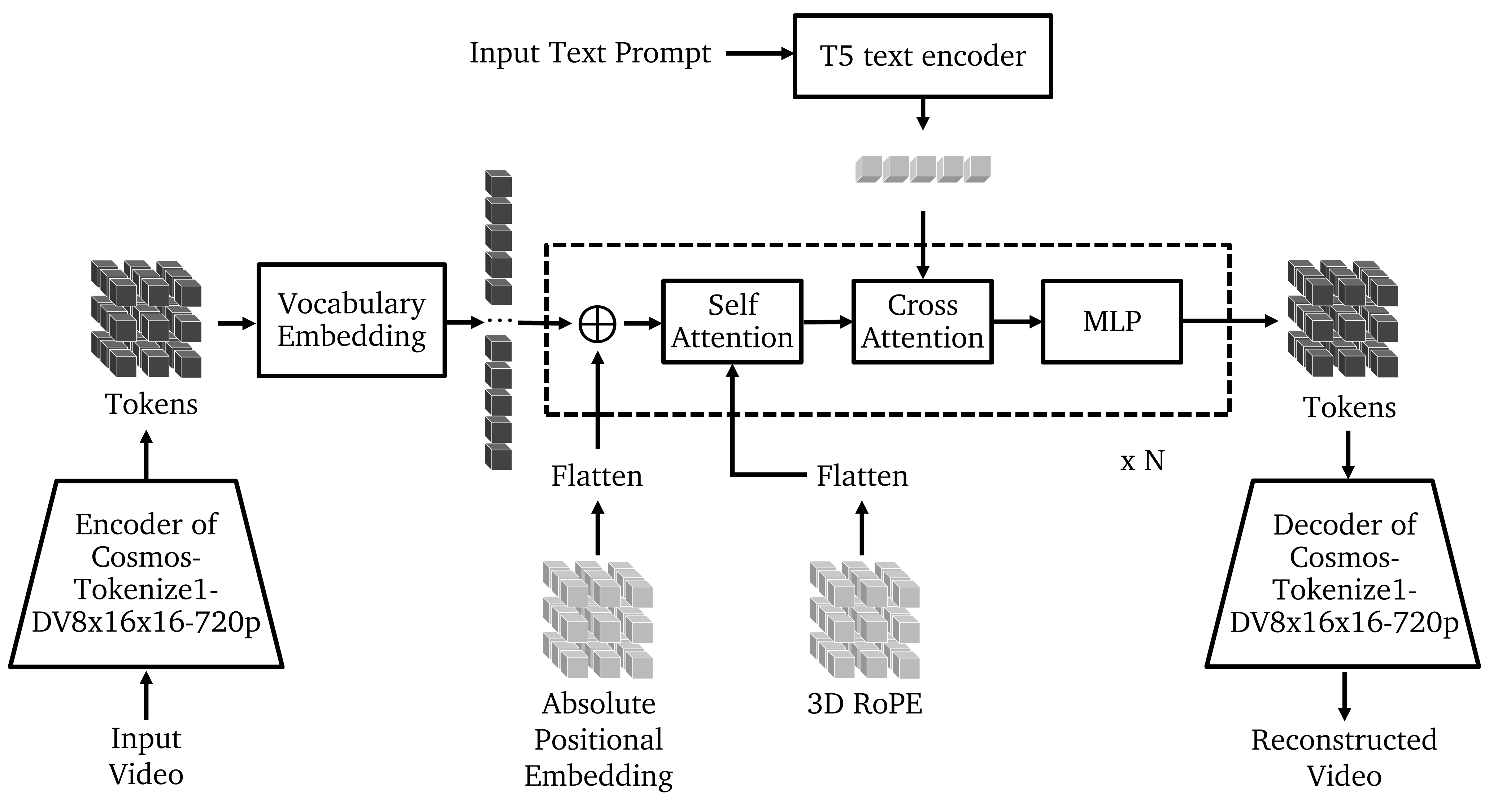}
        \caption{\textbf{Architecture of Cosmos-Predict1-Video2World Model.} The pipeline begins by encoding input video through the encoder of Cosmos-Tokenize1-DV8$\times$16$\times$16-720p to generate discrete tokens, which are transformed into learned embeddings. These embeddings are processed through repeated transformer blocks, each consisting of absolute positional embedding and 3D RoPE components that are flattened before entering the self-attention module. Each block also includes a cross-attention module that incorporates encoded text prompts (processed via a T5 text encoder), followed by a two-layer MLP. Finally, the decoder of Cosmos-Tokenize1-DV8$\times$16$\times$16-720p reconstructs the video from the output tokens.}
    \label{fig:ar_architecture}
\end{figure}

Our autoregressive-based WFM architecture is illustrated in~\cref{fig:ar_architecture}. We make several modifications to the standard transformer model architecture tailored for our video generation task, including adding 1) 3D-aware positional embeddings, 2) cross-attention to enable textual inputs for better control, and 3) QK-Normalization~\citep{wortsman2023small}.

\noindent \textbf{3D positional embeddings.} Similar to our diffusion-based WFM (\cref{sec::diffusion_model_backbone}), we incorporate two complementary positional embedding mechanisms: 3D factorized Rotary Position Embedding (RoPE) for relative positions and 3D factorized absolute positional embedding (APE) for absolute coordinates. These mechanisms work in concert to provide comprehensive spatial and temporal information throughout the network.
\begin{itemize}
    \item \textbf{3D Rotary Position Embedding (RoPE).} We apply 3D RoPE to our model to encode relative positional information across the temporal, height, and width dimensions. During training, we adopt a multi-stage training strategy in which the sequence length of videos increases as the training progresses. To adapt the 3D RoPE to the changing temporal duration, we use YaRN~\citep{peng2023yarn}, a compute-efficient technique designed to extend the context window of RoPE. We apply YaRN extension only along the temporal axis as the video sequence length increases only along the temporal dimension. By utilizing YaRN, our model can extrapolate to context lengths longer than those encountered during the initial stages of training.

    \item \textbf{3D Absolute Positional Embedding (APE).} In addition to 3D RoPE, we incorporate a 3D APE within each transformer block to complement the relative positional encoding. This APE encodes positional information using sinusoidal embeddings factorized across temporal, height, and width dimensions, ensuring the model is aware of absolute positions. The embedding is added directly to the input tensor at each stage, enriching the positional context for the transformer. We find combining absolute and relative positional encodings enhances model performance, reduces training loss, and minimizes morphing artifacts in generated videos. Notably, while our diffusion-based WFM (\cref{sec::diffusion_model_backbone}) employs learnable embeddings, we adopt sinusoidal-based embeddings for APE in our autoregressive-based WFM.
\end{itemize}

\noindent \textbf{Vocabulary.} Tokenization is a crucial step that turns input text into a sequence of discrete tokens in large language models (LLMs). In LLMs, the vocabulary of possible tokens is determined by the LLM's tokenizer (\eg, tiktoken introduced by~\citet{tiktoken}) trained on a large corpus of text with algorithms such as Byte Pair Encoding (BPE)~\citep{gage1994new}.

For our autoregressive models, we use our Cosmos-Tokenize1-DV8$\times$16$\times$16-720p as the tokenizer. As introduced in \cref{sec::token}, we leverage the Finite-Scalar-Quantization (FSQ)~\citep{mentzer2023finite} to quantize the $6$-dimensional latent space into $(8,8,8,5,5,5)$ levels. This quantization leads to a vocabulary size of $8\times8\times8\times5\times5\times5 = 64{,}000$.

\noindent \textbf{Cross-attention for text conditioning.} In addition to the self-attention blocks present in the transformer architecture, we add cross-attention layers to enable the model to condition on input text. Similar to diffusion-based WFM(\cref{sec::diffusion_model_backbone}), cross-attention is applied between the features of the transformer model and text embeddings obtained from a pre-trained text encoder (T5-XXL). In our experiments, we add cross-attention blocks after every self-attention layer.

\noindent \textbf{Query-key normalization.} In order to enhance training stability, we incorporate Query-Key Normalization (QKNorm)~\citep{wortsman2023small}. QKNorm addresses instability in attention mechanisms by normalizing the query ($Q$) and key ($K$) vectors before computing their dot product, thereby preventing the softmax function from saturating and ensuring more effective learning. After normalization, the dot product is scaled by a learnable parameter $\gamma$ instead of the fixed $1/\sqrt{d_k}$. This learnable scaling factor allows the model to adaptively control the magnitude of the attention scores, enhancing flexibility and expressivity.

\noindent \textbf{Z-loss.} To further improve training stability, we introduce a stabilization term known as the z-loss~\citep{de2016z} into our training objective. The z-loss penalizes deviations of the logits from zero, effectively discouraging the model from generating excessively large logit values that could result in numerical instability or gradient explosions. The z-loss is defined as the sum of the squared logits as $\mathcal{L}_{\text{z-loss}} = \lambda \cdot \sum_{i} z_i^2$. We found z-loss to be critical in maintaining gradient norms to a healthy range, especially when scaling the training to a large number of GPU nodes. Empirically, we found that the z-loss coefficient $\lambda = 3 \times 10^{-4}$ strikes an optimal balance, effectively stabilizing training without adversely affecting model performance.

\subsubsection{Scaling Up}

This section describes the techniques that enable efficient scaling of our autoregressive WFMs. We briefly analyze the memory consumption of our models, discuss parallelism strategies, and compare our training setup with other autoregressive models.

\noindent \textbf{Memory requirements.} During training, GPU memory is mainly consumed by:
\begin{itemize}
    \item \textbf{Model parameters}: 6 bytes per parameter. We store the model parameters in both BF16 and FP32.
    \item \textbf{Gradients}: 2 bytes per parameter. We store the gradients in BF16.
    \item \textbf{Optimizer states}: 8 bytes per parameter. We store the first and second moments of AdamW~\citep{loshchilov2019decoupledweightdecayregularization} both in FP32.
    \item \textbf{Activations}: Approximately $(2 \times \text{number\_of\_layers} \times 17 \times \text{seq\_len} \times \text{batch\_size} \times \text{d\_model})$ bytes. We refer readers to~\citet{korthikanti2023reducing} for a detailed analysis of activation memory of state-of-the-art autoregressive models.
\end{itemize}
For instance, our 12B model (Cosmos-Predict1-12B) demands approximately 192 GB of memory for its parameters, gradients, and optimizer states combined. As this is beyond a single NVIDIA H100 GPU's 80GB HBM3 capacity, we leverage tensor parallelism (TP)~\citep{shoeybi2019megatron} and its extension, sequence parallelism (SP)~\citep{korthikanti2023reducing}, to distribute the memory requirements and computation across multiple GPUs.

\noindent \textbf{Tensor Parallelism (TP).} Tensor Parallelism (TP)~\citep{shoeybi2019megatron} splits the weights of linear layers along either the input or output feature dimensions, with the choice guided by the goal of minimizing inter-GPU communication. For example, in a two-layer feedforward network, the weights of the first layer are partitioned along the output feature dimension, while those of the second layer are partitioned along the input feature dimension. This arrangement allows intermediate activations to be processed locally without requiring communication between GPUs. The final outputs are then combined using all-reduce communication. By employing TP, each GPU stores only a fraction, specifically $1 \mathbin{/} \text{TP\_SIZE}$, of the weights for linear layers. However, the default implementation of TP still replicates activations along the sequence dimension for operations like LayerNorm, resulting in redundancy.

\noindent \textbf{Sequence Parallelism (SP).} SP~\citep{korthikanti2023reducing} extends Tensor Parallelism by further partitioning the context along the sequence dimension. This approach is applicable to operators, such as LayerNorm and Dropout in self-attention layers, where each element in the sequence can be processed independently. With SP enabled. Each GPU stores only a fraction, specifically $1 \mathbin{/} \text{TP\_SIZE}$, of the activations.

\noindent \textbf{Comparison with other autoregressive models.} Compared to popular LLMs, our model doesn't leverage memory-saving attention variants such as MQA or GQA. Otherwise, our autoregressive model is deliberately designed to closely resemble the architecture of LLMs~\citep{gpt3,dubey2024llama,jiang2023mistral,adler2024nemotron,gemma2,qwen2p5}, as this alignment offers flexibility and scalability. Experiments that leverage more parallelisms, such as context parallelism and pipeline parallelism, to further scale up the model sizes and context lengths are left for future works.

\subsubsection{Training Strategy} \label{sec:autoregressive:training}

We perform pre-training of our autoregressive WFMs in multiple stages.
\begin{itemize}
    \item \textbf{Stage 1}: In the first stage, the model is trained using the video prediction objective. Given the first frame as the input condition, the model is trained to predict future video frames. A context length of 17 frames is used for this task, \ie, the model predicts 16 future frames with the first frame as input.
    \item \textbf{Stage 1.1}: This stage performs video prediction but with an increased context length of 34 frames. We use the YaRN extension on the temporal dimension to increase the context length of RoPE.
    \item \textbf{Stage 2}: In stage 2 of our training, we introduce text conditioning to our model. Text embeddings are incorporated using newly initialized cross-attention layers. The model is trained with a 34-frame context. To improve text-to-video generation ability, the model is trained using joint image and video data as described in \cref{sec::diffusion_model_training}. When image batches are used, we use a larger batch size as the context length for images is much smaller than that of videos.
\end{itemize}
All our models are trained with a fixed spatial resolution of $640\times1024$.

\noindent \textbf{Cooling down.} After pre-training, we conduct a “cooling-down” phase with high-quality data, similar to LLM training practices \citep{dubey2024llama}. During this phase, we linearly decay the learning rate to $0$ while training on high-quality image-video pairs. The cooling-down phase is carried out over $30{,}000$ iterations.

\begin{table}[ht]
    \setlength{\tabcolsep}{14.12pt} %
    \small
    \captionsetup{justification=centering}
    \caption{Configuration details of Cosmos-Predict1 models.}
    \centering
    \begin{tabular}{l|cccc}
        \toprule
        \makecell[l]{\bf Configuration}
          & \textbf {4B} & \textbf {5B-Video2World} & \textbf{12B} & \textbf{13B-Video2World}  \\
        \midrule
        Number of Layers & $16$ & $16$ & $40$ & $40$ \\
        Model Dimension & $4{,}096$ & $4{,}096$ & $5{,}120$ & $5{,}120$ \\
        Cross Attention Layers & \xmark & \cmark & \xmark & \cmark \\
        Base Learning Rate & $1 \times 10^{-3}$&$3 \times 10^{-4}$&$1 \times 10^{-3}$&$5 \times 10^{-4}$ \\
        Weight Decay &\multicolumn{4}{c}{$0.01$}\\
        Learning Rate Warmup & \multicolumn{4}{c}{Linear scheduler with $5{,}000$ iterations}\\
        Activation Function & \multicolumn{4}{c}{SwiGLU} \\
        FFN Hidden Dimension & \multicolumn{4}{c}{$14{,}336$} \\
        Number of Attention Heads & \multicolumn{4}{c}{$32$} \\
        Number of Key / Value Heads & \multicolumn{4}{c}{$8$} \\
        Number of Tokens & \multicolumn{4}{c}{$12{,}800$} \\
        Vocabulary Size & \multicolumn{4}{c}{$64{,}000$} \\
        Positional Embedding & \multicolumn{4}{c}{3D RoPE ($\theta = 500{,}000$) + 3D APE} \\
        \bottomrule
    \end{tabular}
    \label{tab:ar_model_specs}
\end{table}

We train two sets of autoregressive-based WFMs. We start by building two base models: one with a 4B capacity and the other with a 12B capacity. These are pure next-video token predictors that do not take text prompts as input. We then derive a Video2World version from each of the base models, where we add cross-attention layers to them to leverage text prompt inputs for next video token prediction.

\begin{itemize}
    \item {\textbf{Cosmos-Predict1-4B}}: a 4B transformer model for next video token prediction. This model is trained using stage 1 and stage 1.1 of the multi-stage training objective.
    \item {\textbf{Cosmos-Predict1-5B-Video2World}}: a 5B transformer model derived from our Cosmos-Predict1-4B and trained additionally with stage 2 of the multi-stage training objective.
    \item {\textbf{Cosmos-Predict1-12B}}: a 12B transformer model for next video token prediction. This model is trained using stage 1 and stage 1.1 of the multi-stage training objective.
    \item {\textbf{Cosmos-Predict1-13B-Video2World}}: a 13B transformer model derived from  Cosmos-Predict1-12B and trained additionally with stage 2 of the multi-stage training objective.
\end{itemize}

\subsubsection{Inference Optimization Towards Real-Time Generation}

Our Cosmos Autoregressive WFMs share architectural similarities with LLMs, enabling us to leverage established LLM inference optimization techniques to address the sequential decoding bottleneck. We implement a combination of key-value caching, tensor parallelism, and torch.compile, following the gpt-fast\footnote{\url{https://github.com/pytorch-labs/gpt-fast}} implementation in PyTorch \citep{paszke2019pytorch}.

\noindent \textbf{Speculative decoding.}  To further accelerate our autoregressive WFMs, we apply the Medusa speculative decoding framework~\citep{cai2024medusa}. Unlike common speculative decoding approaches that require a separate draft model~\citep{leviathan2023fast} or training-free methods with limited speedup~\citep{teng2024accelerating}, Medusa extends the transformer backbone with extra decoding heads to predict multiple subsequent tokens in parallel. It then verifies these speculated tokens with rejection sampling. The inference is thus accelerated by alleviating the bottleneck of one-token-at-a-time processing. We demonstrate the potential of the Medusa technique in visual autoregressive acceleration without compromising the quality of generated outputs.

In our implementation, we fine-tune our pre-trained autoregressive WFMs by introducing Medusa heads into the architecture. These heads are strategically inserted after the last transformer hidden states, where all backbone parameters and the final unembedding layer are shared across different heads. Each Medusa head is a single-layer FFN with SiLU activation and residual connection. We further merge the weight matrices of multiple Medusa heads into a unified FFN to maximize parallelism during token prediction. Note that we do not use the tree-based attention mechanism from \citet{cai2024medusa}.

To investigate the optimal Medusa setup for our autoregressive WFMs, we conduct an in-depth study from two aspects: (1) which transformer layers to fine-tune and (2) how many Medusa heads to add. For the first problem, we compare between full fine-tuning and selective layer freezing. We observe that only fine-tuning the Medusa heads gives poor multi-token prediction, while full fine-tuning incurs quality degradation. We empirically identify that unfreezing the last two transformer layers and the final unembedding layer while keeping the backbone frozen yields the best performance. This strategy ensures our Medusa training achieves decent speculative decoding accuracy without suffering from catastrophic forgetting.

\begin{table}[ht]
    \setlength{\tabcolsep}{13pt}
    \small
    \captionsetup{justification=centering}
    \caption{Impact of Medusa head number on average token throughput and number of forward passes.\\ The experiments are conducted on 8 $\times$ H100 GPUs and 50 unseen test videos of $640\times1024$ resolution.}
    \centering
    \begin{tabular}{l|c|c|c|c|c|c}
        \toprule
        \textbf{Model} &\textbf{Medusa Head Number} & \textbf{0} & \textbf{3} & \textbf{6} & \textbf{9} & \textbf{12} \\
        \midrule
        4B  & Token Throughput (tokens/s) & 444.95 & 663.51 & 829.59 & \textbf{894.67} & 890.64 \\
             & $\#$ of Forward Passes & 7680 & 2860 & 2073 & 1812 & \textbf{1682} \\
        \midrule
        5B  & Token Throughput (tokens/s) & 303.61 & 659.94 & 758.58 & \textbf{982.77} & 978.80 \\
             & $\#$ of Forward Passes & 10240 & 2857 & 2382 & 1799 & \textbf{1673} \\
        \bottomrule
    \end{tabular}
    \label{tab:medusa_head_num}
\end{table}

To explore the optimal number of Medusa heads, we calculate the model token throughput and forward pass count with different numbers of Medusa heads. The ablation studies are conducted on 8 $\times$ H100 GPUs and evaluated on 50 unseen test videos of $640\times1024$ resolution. The results in~\cref{tab:medusa_head_num} suggest that our Medusa framework can effectively accelerate inference, with up to $2.0\times$ token throughput and $4.6\times$ less forward pass for the 4B model, and up to $3.2\times$ token throughput and $6.1\times$ less forward pass for the 5B model. We show that though more Medusa heads can reduce the number of forward passes needed to generate, it may slow down the overall token throughput. We find that $9$ Medusa heads yield the best trade-off between computational efficiency and model performance.

\begin{table}[ht]
    \setlength{\tabcolsep}{8pt}
    \small
    \captionsetup{justification=centering}
    \begin{threeparttable}
    \caption{Performance analysis of Cosmos Autoregressive Models on test videos of $640\times1024$ resolution.}
    \centering
    \begin{tabular}{l|c|c|c|c|c|c}
        \toprule
        \textbf{Model} & \textbf{GPU} & \textbf{No DD (s)} & \textbf{No DD+Medusa (s)} & \textbf{With DD (s)} & \textbf{With DD+Medusa (s)} & \textbf{VRAM (GB)} \\
        \midrule
          & 1 & 31.04 & 23.52 & 61.49 & 53.08 & 29 \\
        4B     & 4 & 18.20 & 13.87 & 29.60 & 25.63 & 31 \\
             & 8 & 17.62 & 9.91  & 30.30 & 22.83 & 34 \\
        \midrule
          & 1 & 39.68 & 24.97 & 70.39 & 54.72 & 59 \\
        5B     & 4 & 25.59 & 20.96 & 37.29 & 33.35 & 51 \\
             & 8 & 25.70 & 11.67 & 38.41 & 24.35 & 49 \\
        \midrule
         & 1 & 84.78 & \textemdash   & 116.66 & \textemdash  & 45 \\
        12B     & 4 & 47.49 & \textemdash   & 60.27  & \textemdash   & 36 \\
             & 8 & 45.69 & \textemdash   & 58.81  & \textemdash   & 37 \\
        \midrule
         & 1 & 109.18 & \textemdash   & 140.24 & \textemdash   & 77 \\
        13B     & 4 & 67.80  & \textemdash   & 80.76  & \textemdash   & 55 \\
             & 8 & 67.22  & \textemdash   & 80.93  & \textemdash   & 55 \\
        \bottomrule
    \end{tabular}
    \begin{tablenotes}
    The table reports the average inference time (in seconds) and VRAM utilization of various Cosmos Autoregressive WFMs under different settings. Inference time is reported for generating $32$ frames with a single conditioning frame as input. \textbf{No DD}: Time without diffusion decoder. \textbf{No DD+Medusa}: Time without diffusion decoder but with Medusa heads. \textbf{With DD}: Time with diffusion decoder. \textbf{With DD+Medusa}: Time with diffusion decoder and Medusa heads. \textbf{VRAM}: Video RAM usage in gigabytes.
    \end{tablenotes}
    \label{tab:medusa_throughput}
    \end{threeparttable}
\end{table}

In \cref{tab:medusa_throughput}, we show performance analysis of autoregressive WFMs with Medusa integration. This analysis was conducted on H100 GPUs and evaluated on test videos of $640\times1024$ resolution in the BF16 precision. Results show that the Medusa implementation consistently accelerates inference for both 4B and 5B models under different GPU configurations.

\noindent \textbf{Low-resolution adaptation for real-time inference.}  We pursue real-time inference by adapting our model to a lower spatial resolution of $320\times512$, which results in a lower number of tokens per video. Specifically, we first fine-tune the discrete video tokenizer (Cosmos-Tokenize1-DV8$\times$16$\times$16-720p in~\cref{sec::token}) on 320p low-resolution videos using videos from the target Physical AI domain. Then, we fine-tune our autoregressive WFM that is pre-trained in $640\times1024$ resolution (Cosmos-Predict1-4B in~\cref{sec:autoregressive:training}) with this low-resolution tokenizer on videos of $320\times512$ resolution from the target Physical AI domain. Finally, we add the Medusa heads to the fine-tuned low-resolution autoregressive WFM.

\begin{table}[ht]
    \setlength{\tabcolsep}{8pt}
    \small
    \captionsetup{justification=centering}
    \caption{Decoding throughput of Cosmos-Predict1-4B with low-resolution adaptation, benchmarked on 8 $\times$ H100 80GB GPUs using 10-FPS videos of $320 \times 512$ resolution from the Physical AI domain.}\label{tab:medusa_low_res}
    \centering
    \begin{tabular}{r|c|c}
        \toprule
        \textbf{Model ($320 \times 512$)} & \textbf{Token Throughput (tokens/s)} & \textbf{Video Throughput (frames/s)} \\
        \midrule
        Cosmos-Predict1-4B (with Medusa) & 806.61 & 10.08 \\
        \bottomrule
    \end{tabular}
\end{table}

We conducted inference benchmarking on 8 $\times$ H100 GPUs using torch.compile's ``max-autotune'' mode in BF16 precision, and evaluated with 10-FPS input videos from the target Physical AI domain. In~\cref{tab:medusa_low_res}, we report the average token throughput and frame generation speed achieved in this setup. We observe that our model can generate 10 video frames in less than 1 second, demonstrating that we can achieve real-time video generation at 10 FPS.

\subsubsection{Diffusion Decoder}\label{sec::diffusion_decoder}

Our Cosmos tokenizer uses a lightweight encoder-decoder architecture to perform aggressive compression, which reduces the number of tokens for our WFM training. As a result of aggressive compression, it could sometimes lead to blurriness and visible artifacts in video generation, especially in the autoregressive WFM setting, where only a few integers are used to represent a rich video through discrete tokenization. We resort to the diffusion decoder design~\citep{DALLE2, openai2024dalle3} to address the limitation. Specifically, we build a more powerful tokenizer decoder by fine-tuning Cosmos-Predict1-7B-Text2Video in \cref{sec::diffusion_model}.

\begin{figure}[ht]
    \centering
    \includegraphics[width=0.99\textwidth]{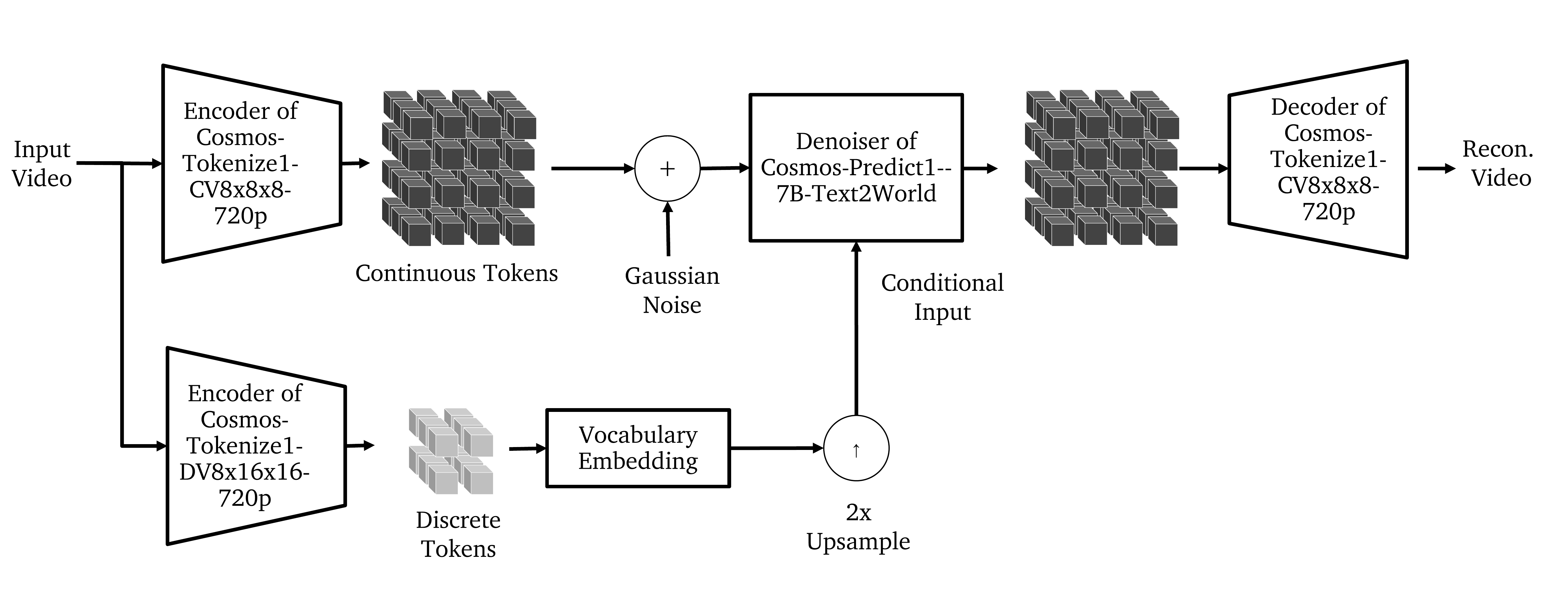}
    \caption{\textbf{Cosmos diffusion decoder training}. During training, each input video is tokenized twice: one through the target discrete tokenizer (DV8$\times$16$\times$16) and the other through a less-constrained continuous tokenizer (CV8$\times$8$\times$8). The discrete token video is used as the conditional input to the diffusion denoiser.}
    \label{fig:dd_arch}
    \vspace{2mm}
    \includegraphics[width=0.99\textwidth]{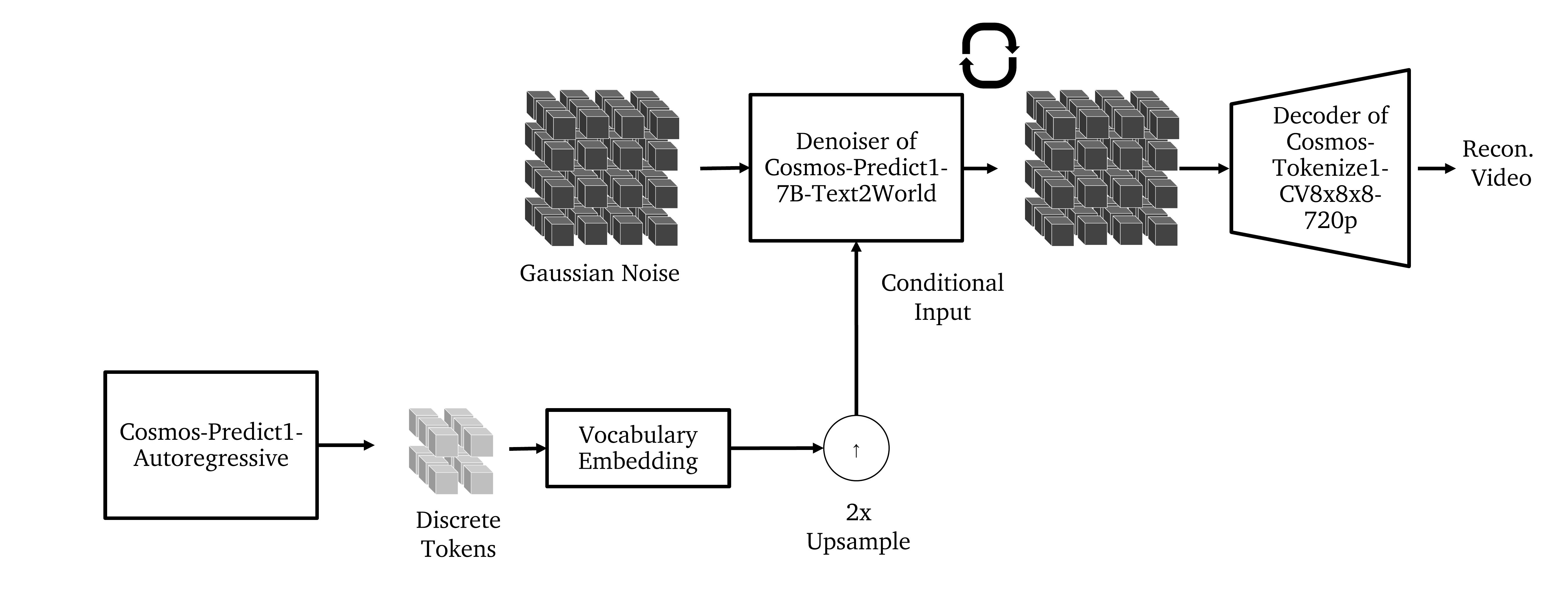}
    \caption{\textbf{Cosmos diffusion decoder inference}. During inference, the output video token from a Cosmos-Predict1 model is conditionally inputted to the denoiser.}
    \label{fig:dd_inference}

\end{figure}

\cref{fig:dd_arch} illustrates how we train a diffusion decoder for our autoregressive WFMs. For each training video, we use Cosmos-Tokenize1-CV8$\times$8$\times$8-720p and Cosmos-Tokenize1-DV8$\times$16$\times$16-720p to compute a continuous token video and a corresponding discrete token video, respectively. We note that Cosmos-Tokenize1-CV8$\times$8$\times$8-720p can produce higher quality video outputs than Cosmos-Tokenize1-DV8$\times$16$\times$16-720p thanks to the more gentle continuous tokenization process and the less aggressive compression scheme ($8\times8\times8$ instead of $8\times16\times16$).

The discrete token video is treated as the conditional input to the denoiser of the Cosmos-Predict1-7B model. To compute the conditional input, we first embed each discrete token of the discrete token video into a 16-dimensional vector based on a learnable vocabulary embedding layer. We then upsample the embedding $2\times$ along the $x$ and $y$ directions so that the conditional input will be of the same size as the noisy input to the denoiser from the continuous token video. We concatenate the noisy continuous inputs with the conditional inputs along the channel dimension, which becomes the input to the diffusion denoiser. The first layer of the denoiser is channel-dimension expanded to accommodate the new input shape. We fine-tune the updated Cosmos-Predict1-7B by removing the added noise. As the discrete token video is not noise-corrupted, the denoiser learns to leverage the residing information in the conditional input for denoising. The result is a higher-quality decoder for the tokenizer that decodes the discrete token by solving a reserve diffusion problem.

\cref{fig:dd_inference} illustrates the inference. The output discrete token video (under $8 \times 16 \times 16$ discrete compression) from our autoregressive WFM is decoded into a video through two steps. First, we roll out the conditional denoiser to generate a continuous token video (under $8 \times 8 \times 8$ continuous compression) based on the autoregressive WFM output. Next, the continuous token video is decoded by Cosmos-Tokenize1-CV8$\times$8$\times$8-720p to produce the resulting RGB video.

\begin{figure*}[tb!]
    \centering
    \setlength{\tabcolsep}{1.5pt}
    \renewcommand{\arraystretch}{0}
    \resizebox{\textwidth}{!}{%
    \begin{tabular}{ccccc} %
        & Condition frame $0$ & Frame $10$ & Frame $20$ & Frame $30$ \\[2pt]

        {\rotatebox{90}{\hspace{24pt} 4B}} & \includegraphics[width=0.245\textwidth]{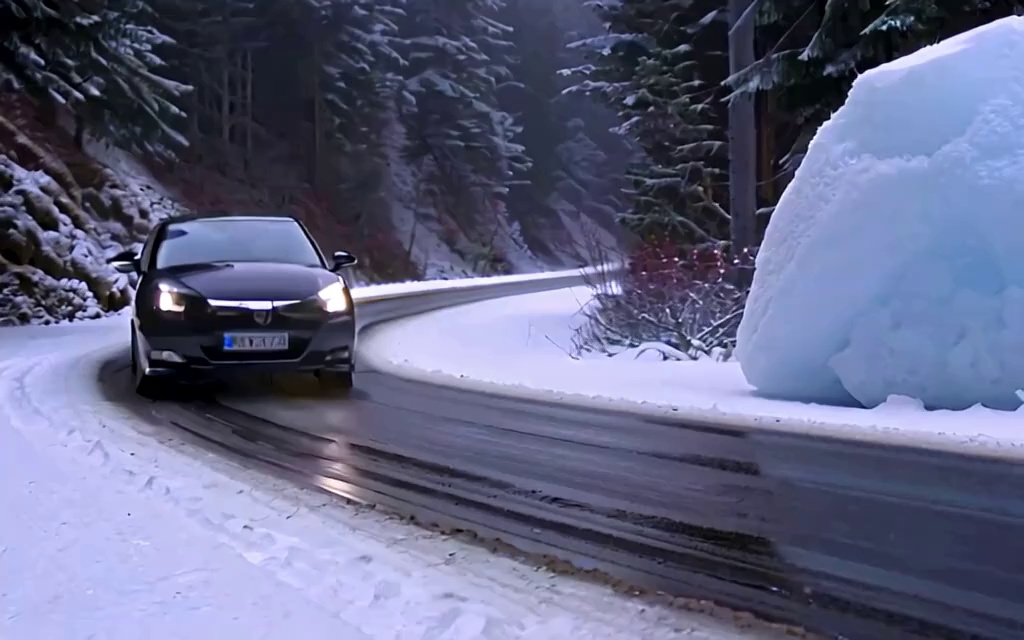} &
        \includegraphics[width=0.245\textwidth]{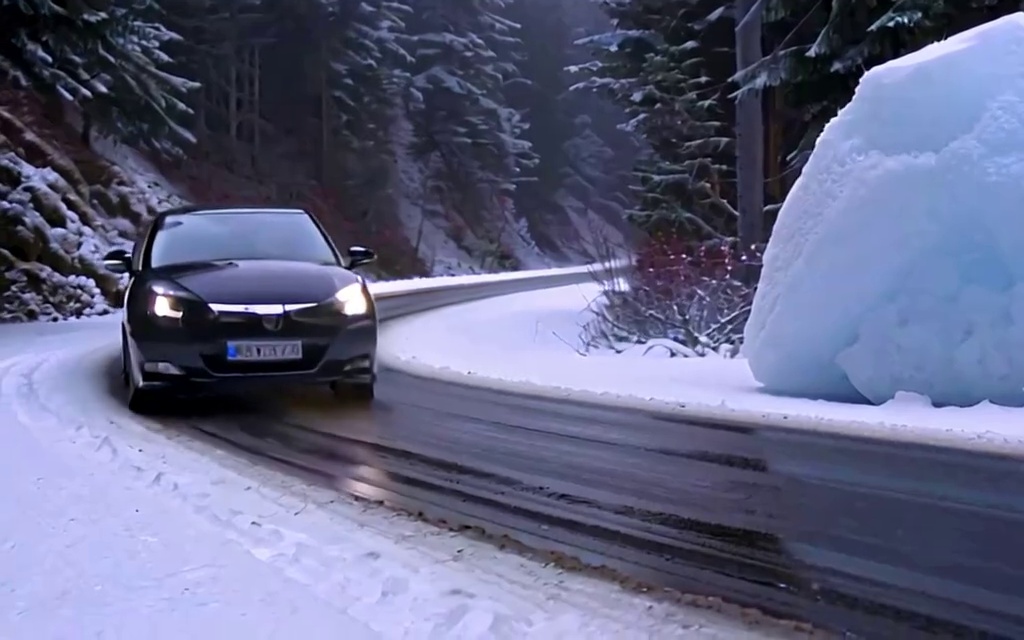} &
        \includegraphics[width=0.245\textwidth]{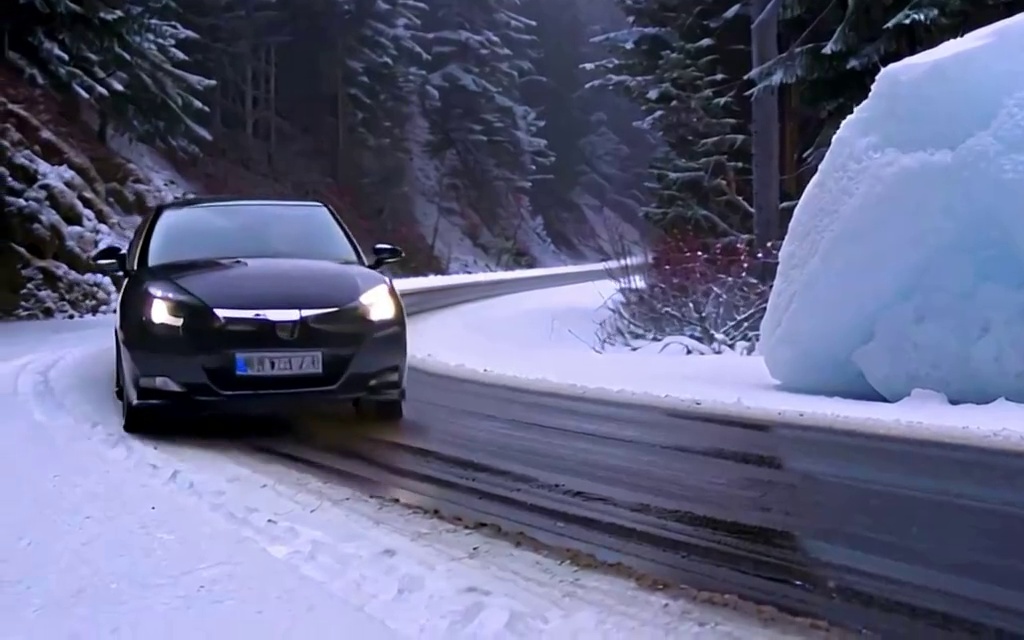} &
        \includegraphics[width=0.245\textwidth]{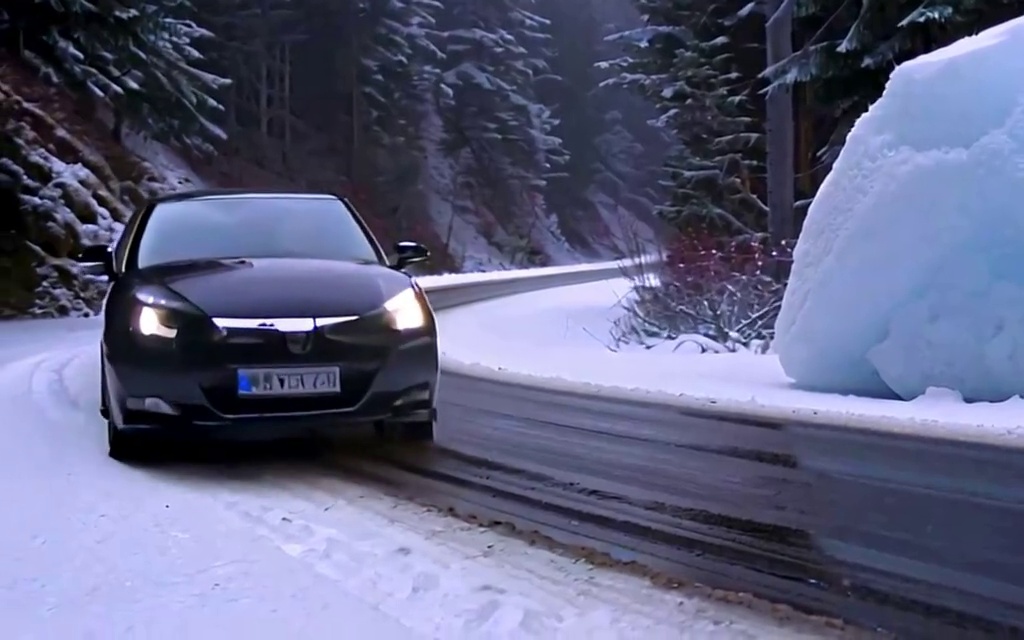} \\[2pt]

        {\rotatebox{90}{\hspace{24pt} 12B}} & \includegraphics[width=0.245\textwidth]{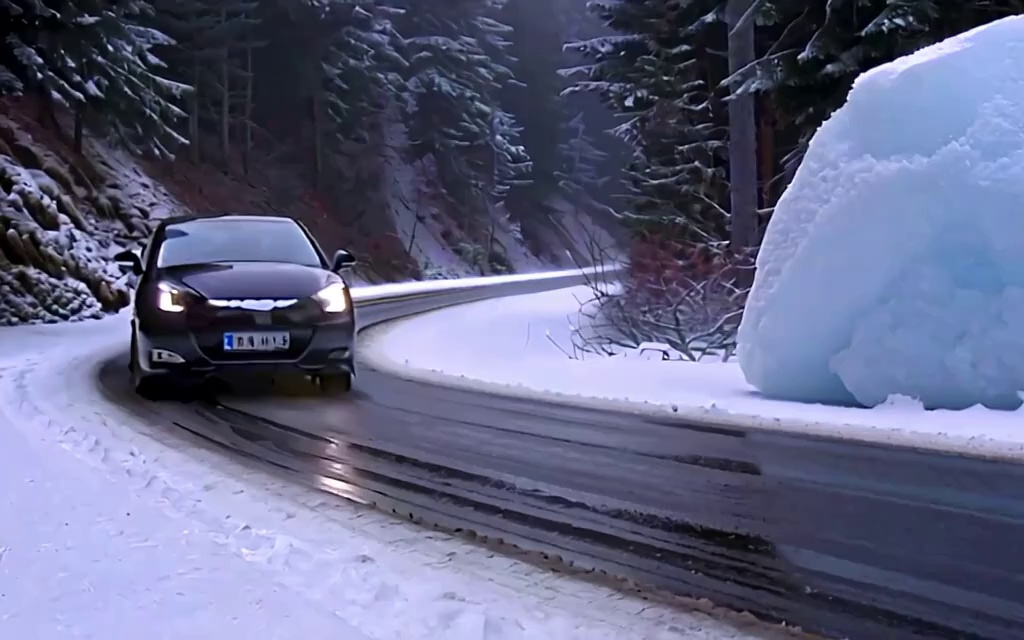} &
        \includegraphics[width=0.245\textwidth]{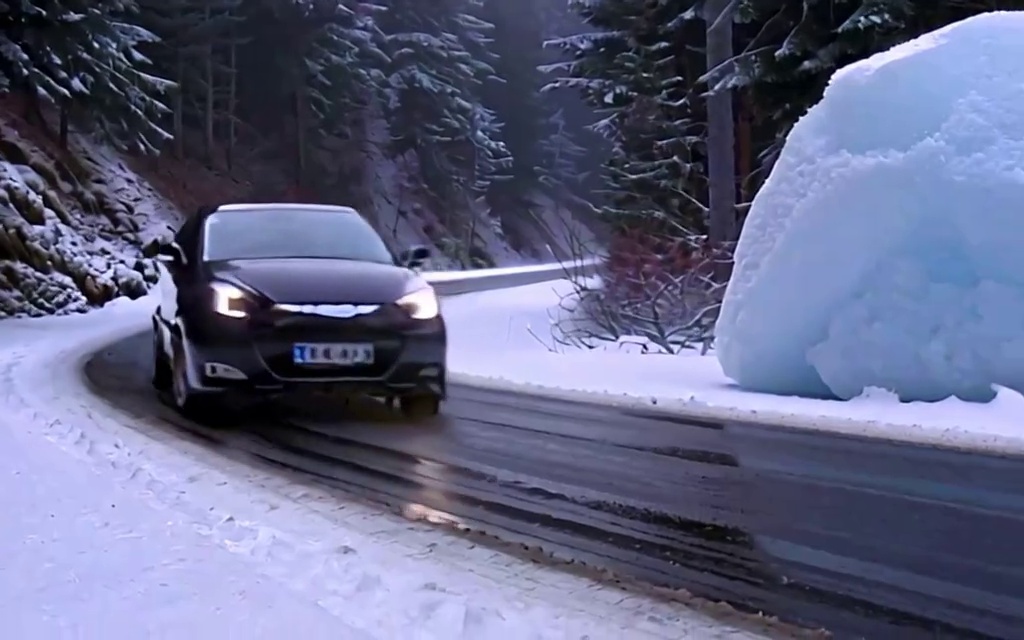} &
        \includegraphics[width=0.245\textwidth]{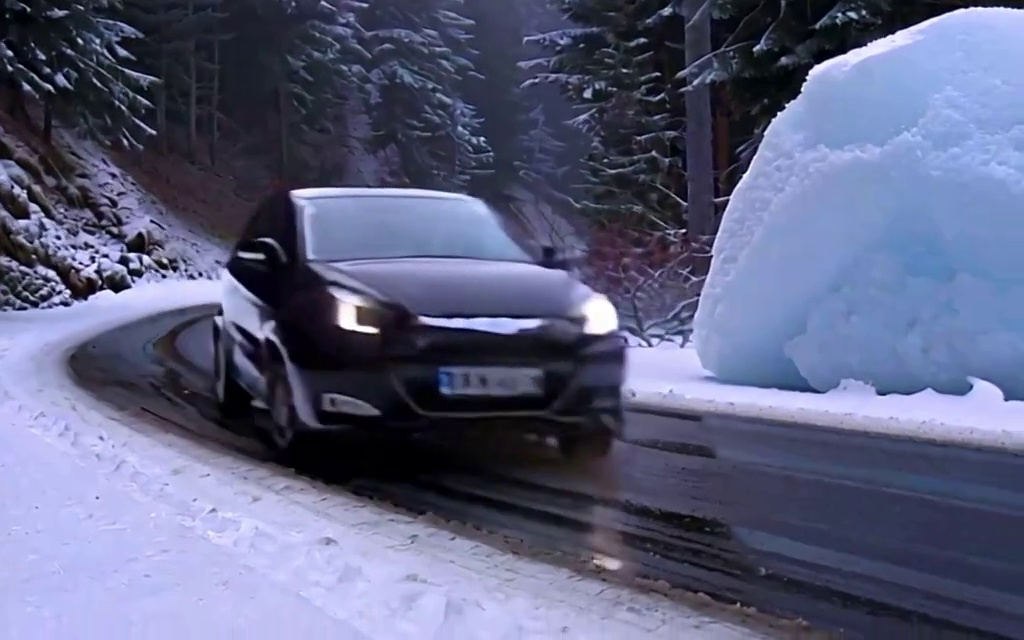} &
        \includegraphics[width=0.245\textwidth]{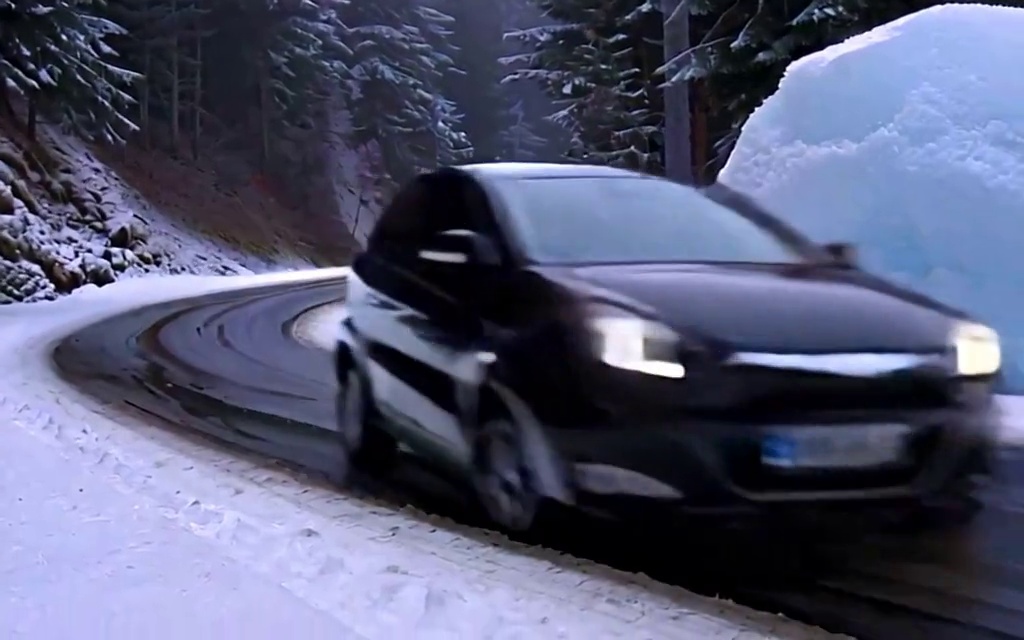} \\[4pt]

        \multicolumn{5}{p{\textwidth}}{
        \begin{multicoltextblock}
        \prompttext{Prompt}: 
        \prompt{None.}
        \end{multicoltextblock} 
        } \\
        & \\ [18pt]

        {\rotatebox{90}{\hspace{24pt} 5B}} & \includegraphics[width=0.245\textwidth]{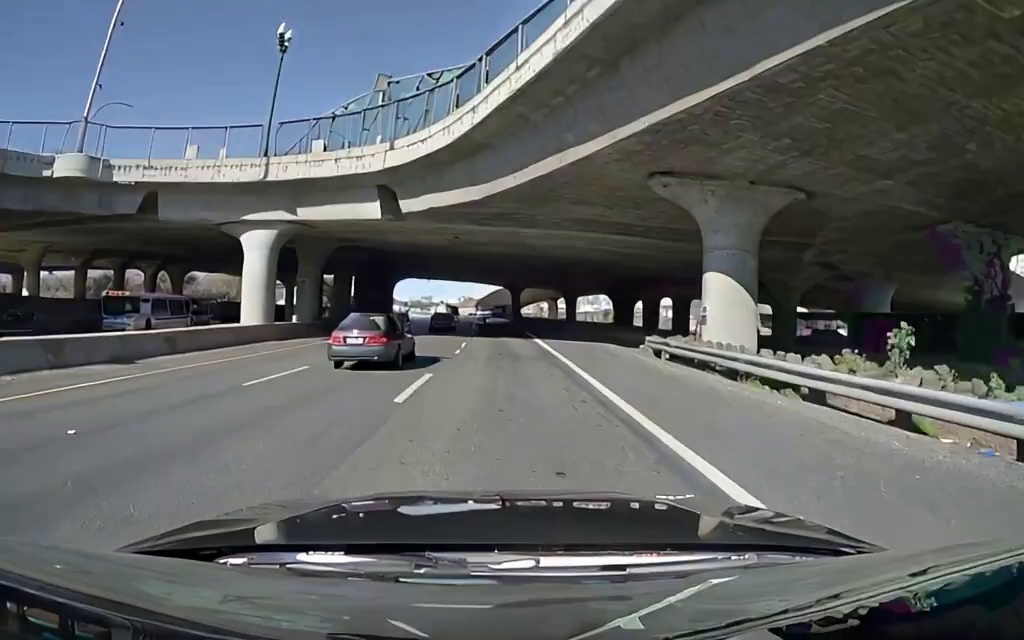} &
        \includegraphics[width=0.245\textwidth]{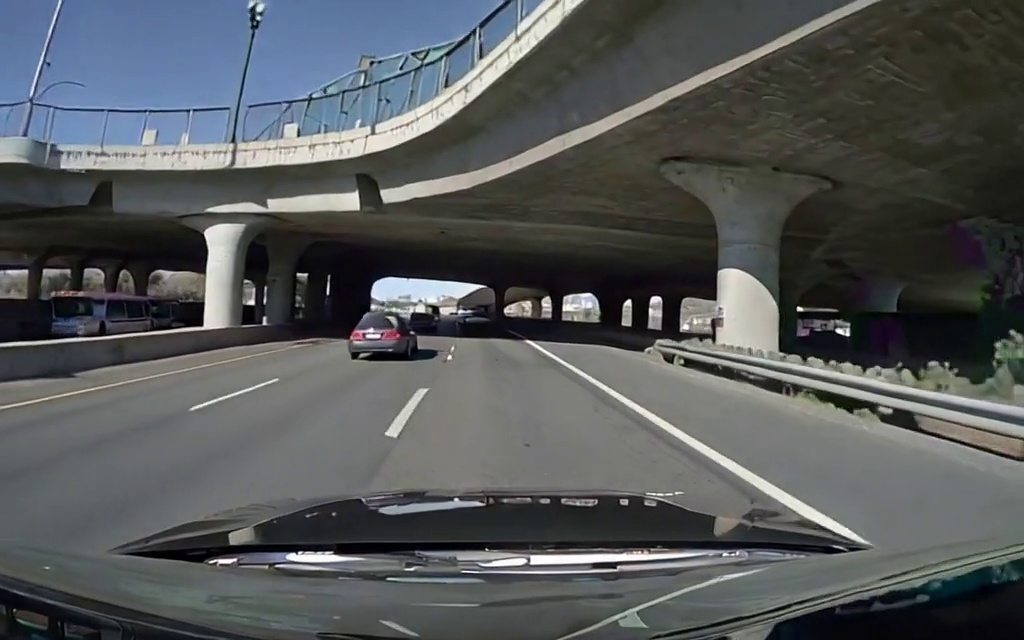} &
        \includegraphics[width=0.245\textwidth]{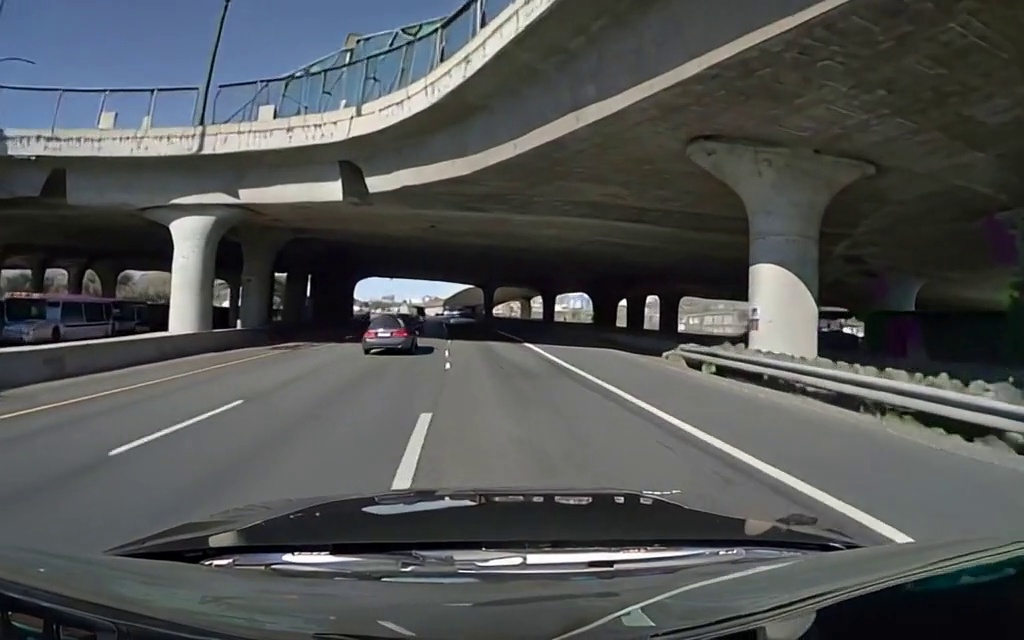} &
        \includegraphics[width=0.245\textwidth]{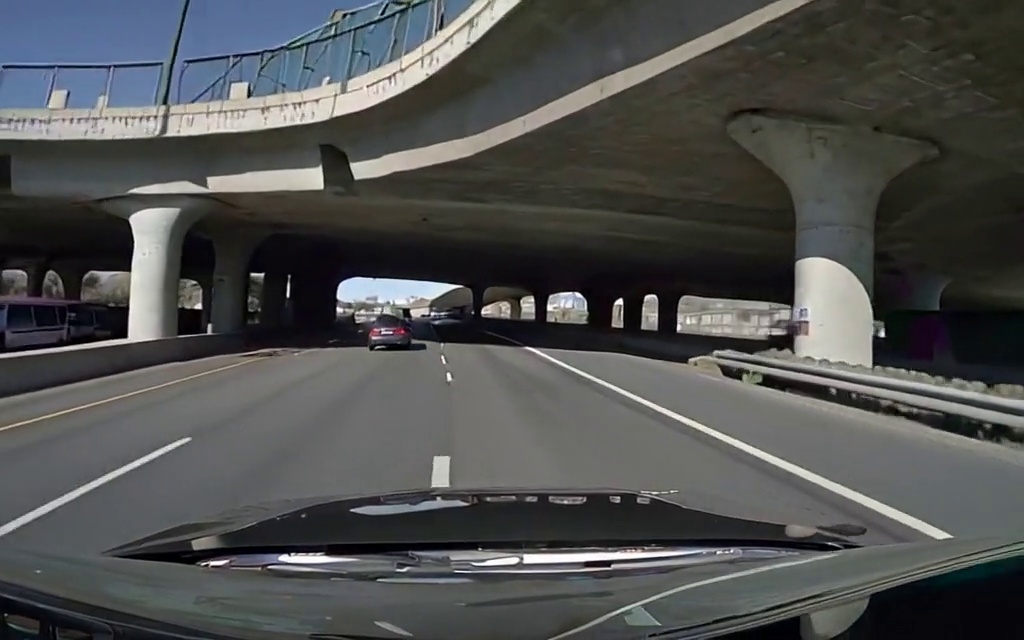} \\[2pt]

        {\rotatebox{90}{\hspace{24pt} 13B}} & \includegraphics[width=0.245\textwidth]{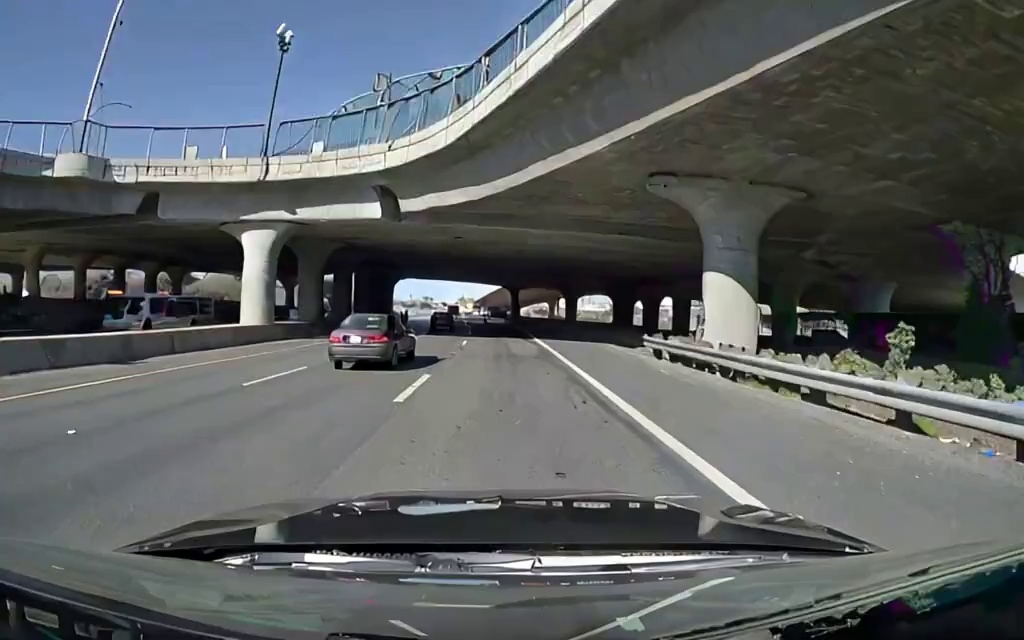} &
        \includegraphics[width=0.245\textwidth]{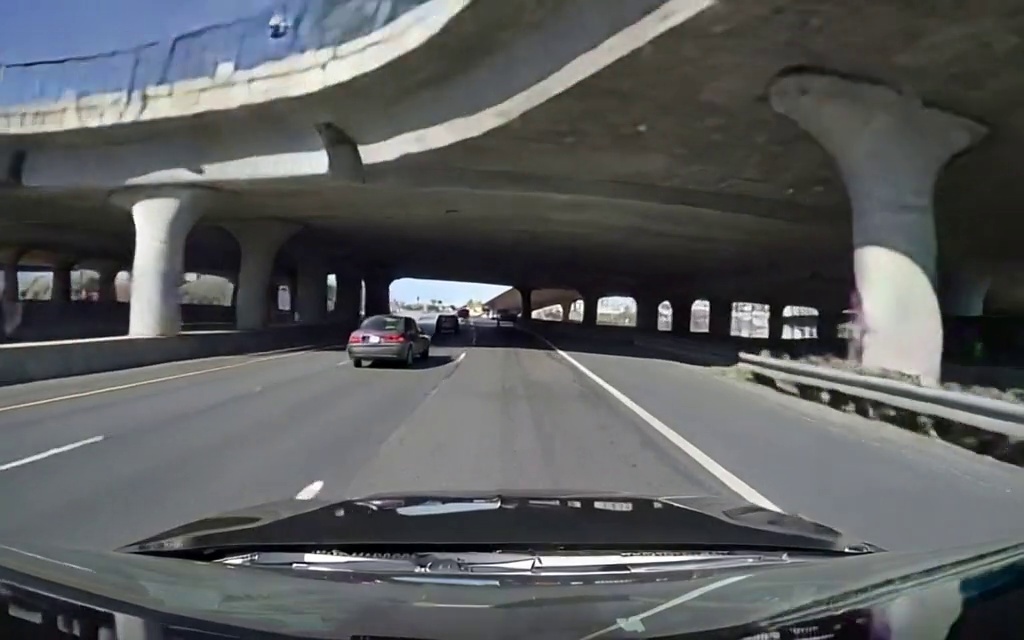} &
        \includegraphics[width=0.245\textwidth]{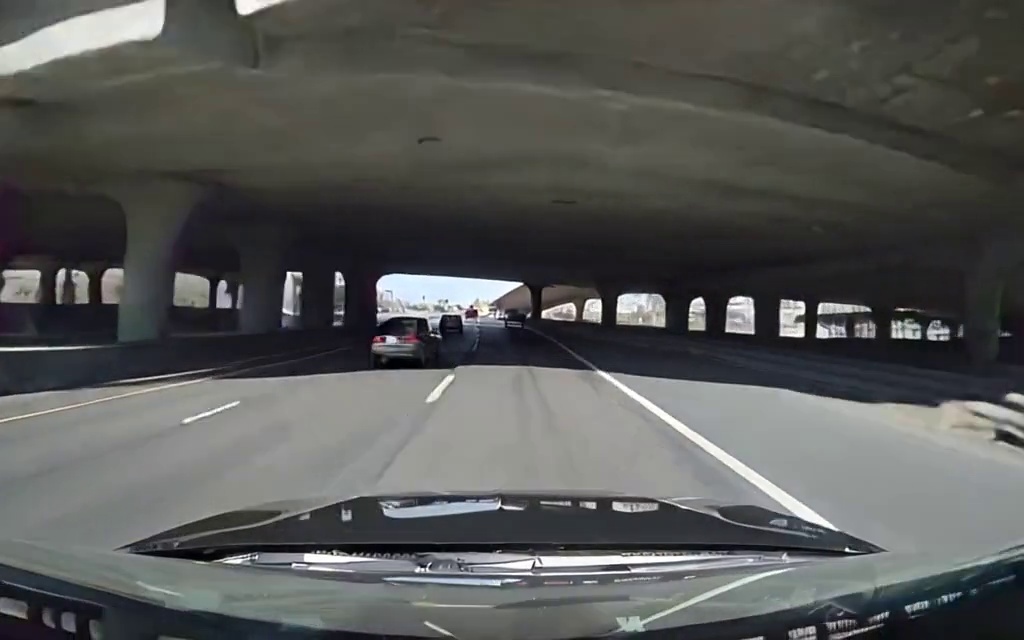} &
        \includegraphics[width=0.245\textwidth]{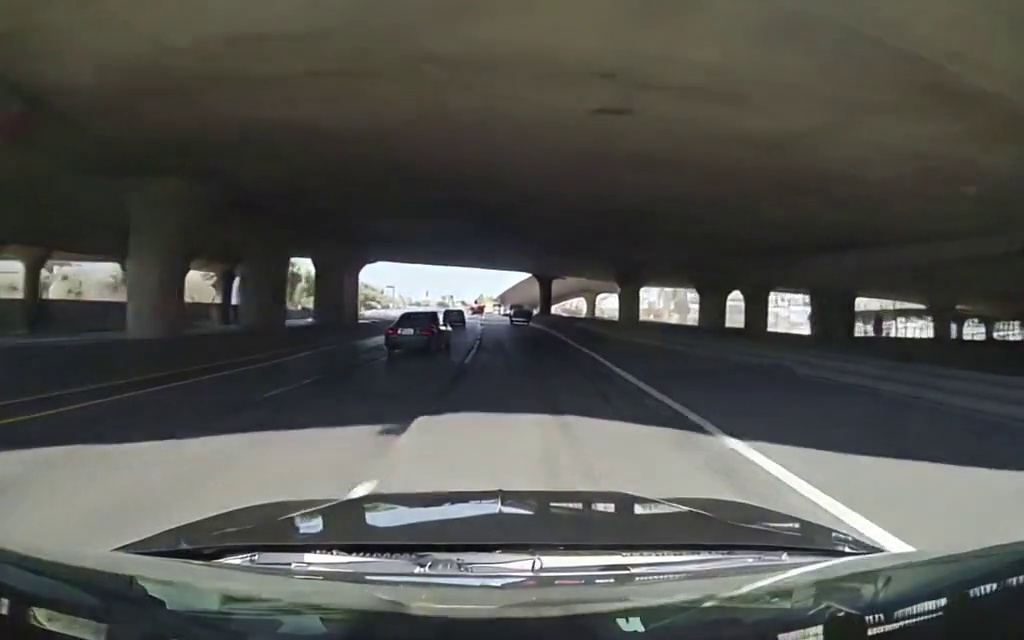} \\[4pt]
        \multicolumn{5}{p{\textwidth}}{
        \begin{multicoltextblock}
        \prompttext{Prompt}: 
        \prompt{The video of a car moving forward, passing under a large overpass. The road is clear, and there are a few other cars visible in the distance. The weather appears to be sunny, and the time of day is daytime. The scene is set on a busy highway with concrete structures and greenery on the sides.}
        \end{multicoltextblock} 
        } \\
        & \\ [4pt]
    \end{tabular}}
    \caption{\textbf{Generated videos from Cosmos Autoregressive World Foundation Models.} The top two rows are the video generation results with 4B and 12B models, while the bottom two rows are the Video2World results with a text prompt. We observe that the 12B and 13B models shows sharper videos and better motion than 4B and 5B models in both prompted and unprompted settings. To check full videos and more video examples, please visit our \href{https://research.nvidia.com/labs/dir/cosmos1/}{website}.}
    \label{fig:cosmos_ar_qualitative_results}
\end{figure*}

\begin{figure*}[tb!]
    \centering
    \setlength{\tabcolsep}{1.5pt}
    \renewcommand{\arraystretch}{0}
    \resizebox{\textwidth}{!}{%
    \begin{tabular}{cccc} %

        Condition frame $0$ & Frame $10$ & Frame $20$ & Frame $30$ \\[2pt]

       \includegraphics[width=0.249\textwidth]{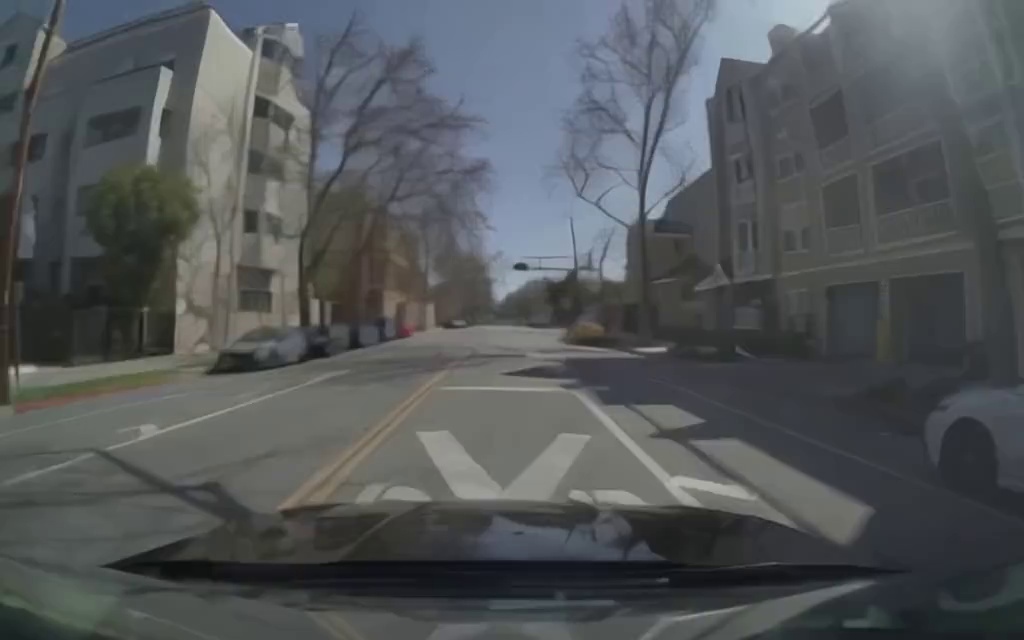} &
        \includegraphics[width=0.249\textwidth]{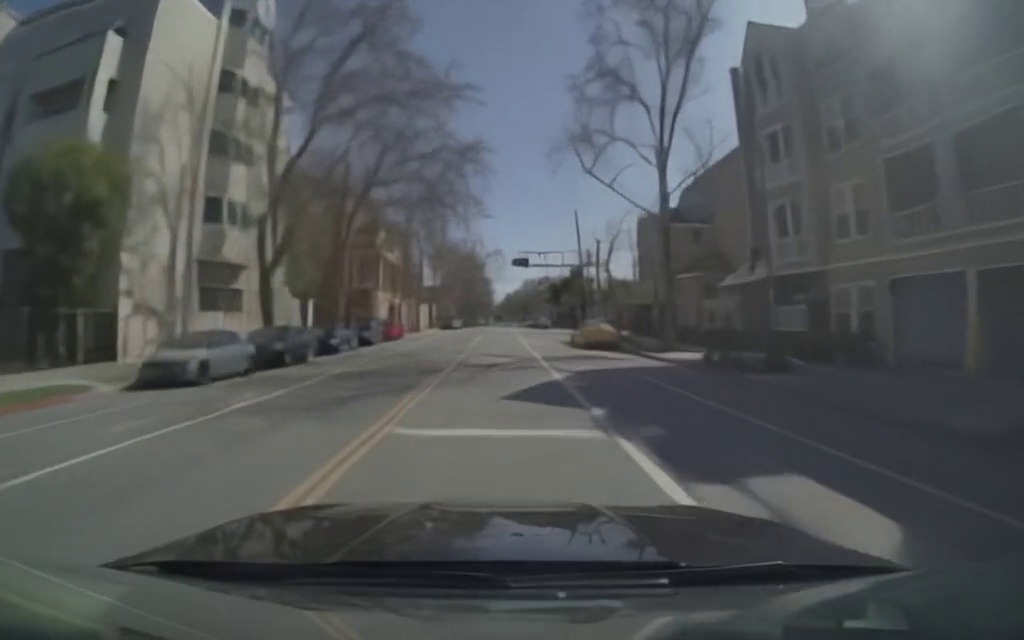} &
        \includegraphics[width=0.249\textwidth]{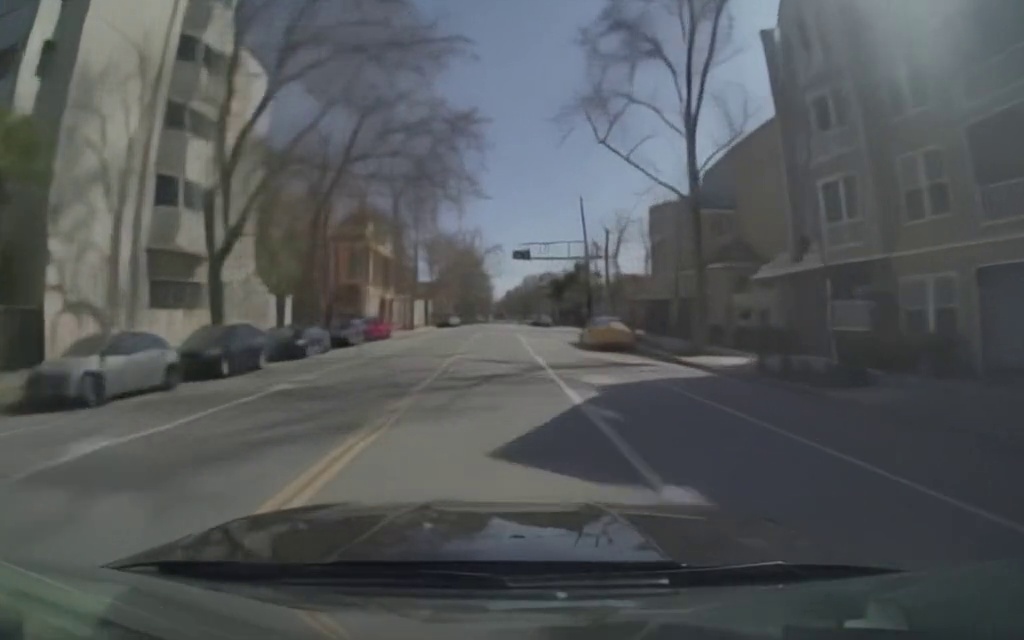} &
        \includegraphics[width=0.249\textwidth]{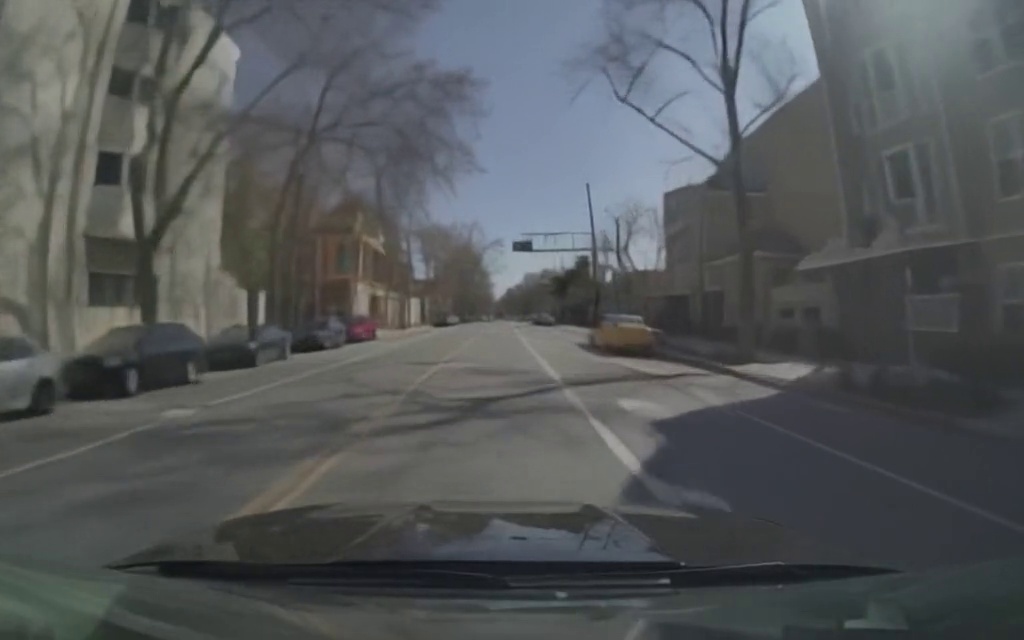} \\[4pt]
        \multicolumn{4}{c}{Output of Cosmos-Predict1-13B-Video2World} \\[4pt]

        \includegraphics[width=0.249\textwidth]{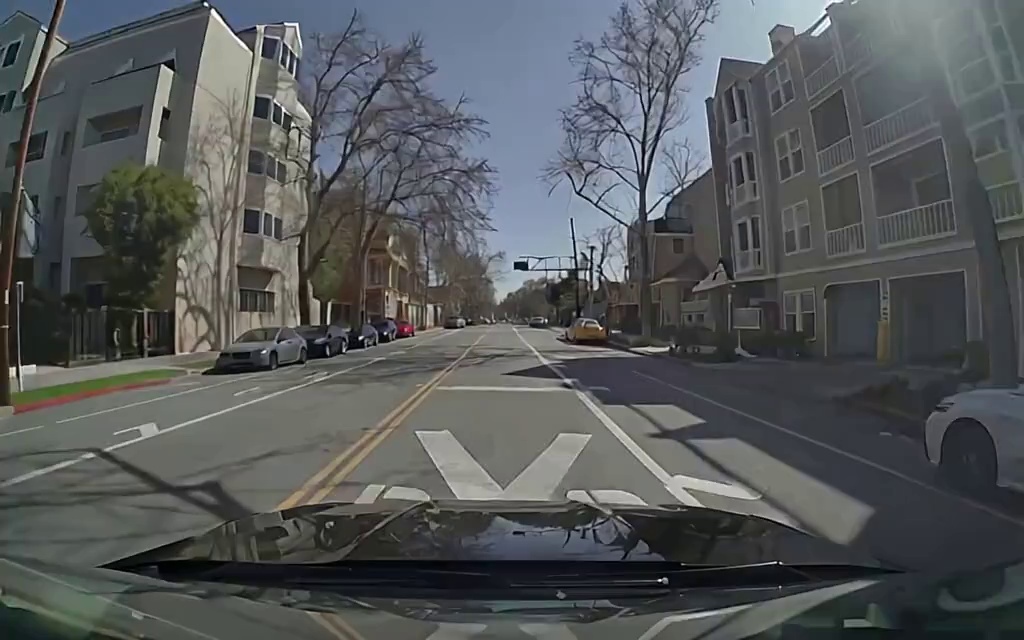} &
        \includegraphics[width=0.249\textwidth]{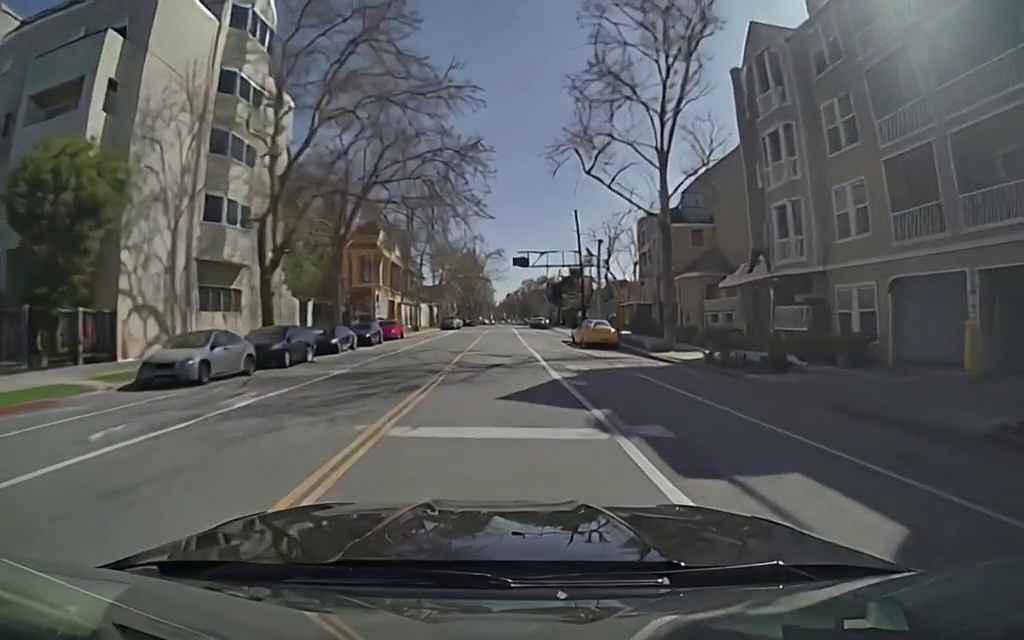} &
        \includegraphics[width=0.249\textwidth]{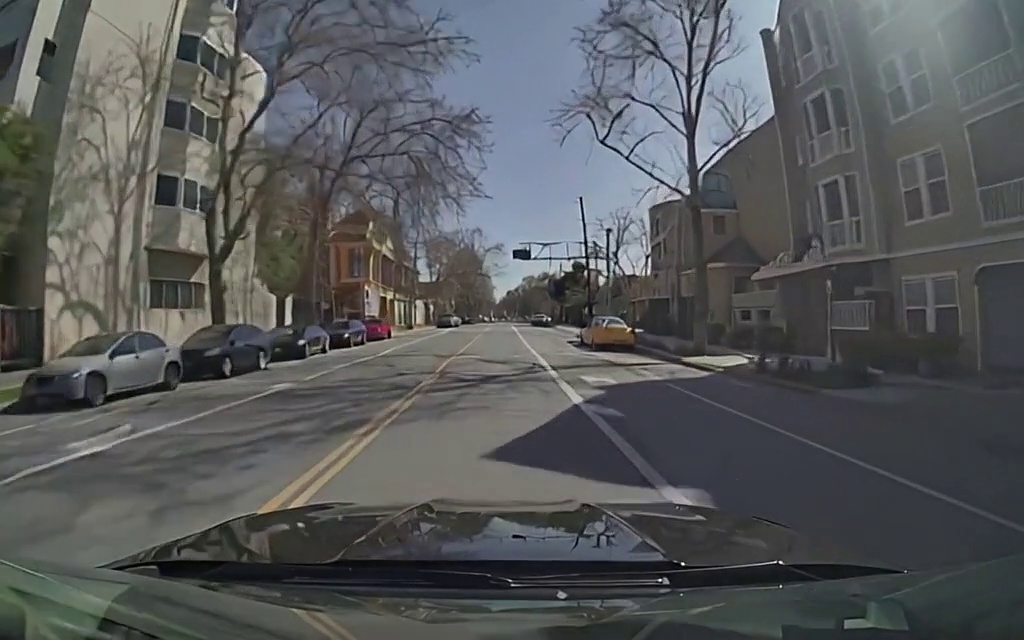} &
        \includegraphics[width=0.249\textwidth]{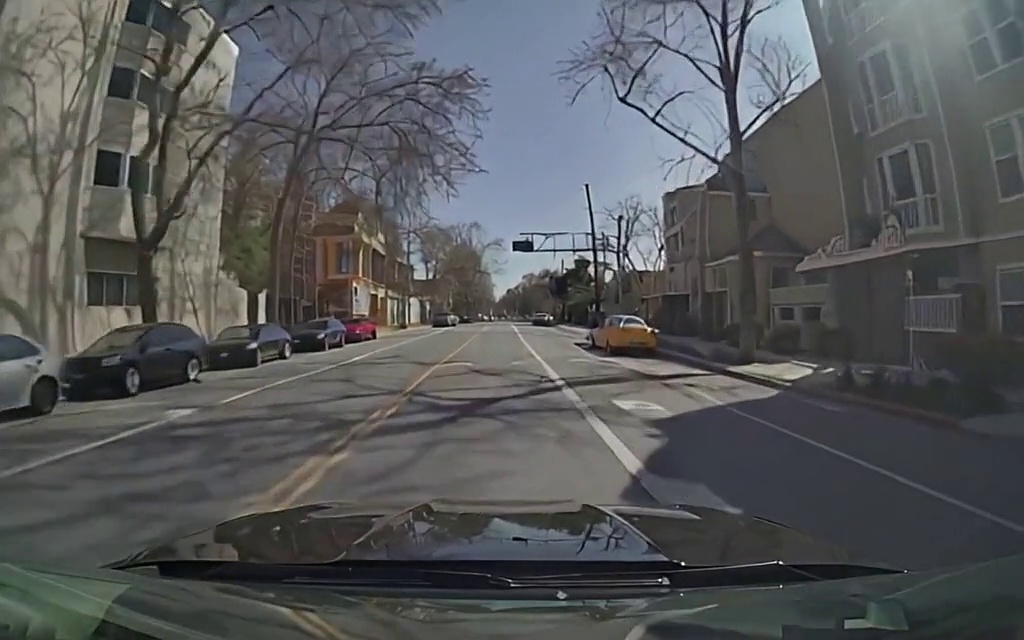} \\[4pt]
        \multicolumn{4}{c}{Output of Cosmos-Predict1-13B-Video2World + diffusion decoder} \\[4pt]
    \end{tabular}}
    \caption{\textbf{Diffusion decoder comparison}. In the top panel, we show the video generation results with Cosmos-Predict1-13B-Video2World model. In the bottom panel, we show the enhanced video after the output from the autoregressive model is passed through the diffusion decoder. We observe that the autoregressive model alone produces blurry results, while the diffusion decoder can enhance the sharpness of videos while preserving the content.}
    \label{fig:cosmos_ar_diffusion_decoder_comparison}
\end{figure*}

\subsubsection{Results}

In \cref{fig:cosmos_ar_qualitative_results}, we show qualitative results of our autoregressive WFMs using different model sizes. In the unprompted setting, comparing Cosmos-Predict1-4B and Cosmos-Predict1-12B model, we observe that the 12B model generates videos with better motion and sharper details. Similarly, in the prompted setting, comparing Cosmos-Predict1-5B-Video2World and Cosmos-Predict1-13B-Video2World reveals that the 13B model gets better motion than the 5B model.

In \cref{fig:cosmos_ar_diffusion_decoder_comparison}, we show the enhancements obtained when using the diffusion decoder. The outputs of the autoregressive model are blurry mainly due to the lossy compression in our discrete tokenizer. The use of the diffusion decoder can enhance details while preserving the content.

We empirically find the outputs of the autoregressive-based Text2World WFMs do not improve with upsampled prompts from the prompt upsampler discussed in~\cref{sec::prompt_upsampler}. We hypothesize this is possibly due to the fact that these WFMs are pretrained with pure video generation tasks for most of the training. They are not forced hard enough to leverage text inputs.

\begin{figure*}[tb!]
    \centering
    \setlength{\tabcolsep}{1.5pt}
    \renewcommand{\arraystretch}{0}
    \resizebox{\textwidth}{!}{%
    \begin{tabular}{cccc} %

        Condition frame $0$ & Frame $10$ & Frame $20$ & Frame $30$ \\[2pt]

       \includegraphics[width=0.249\textwidth]{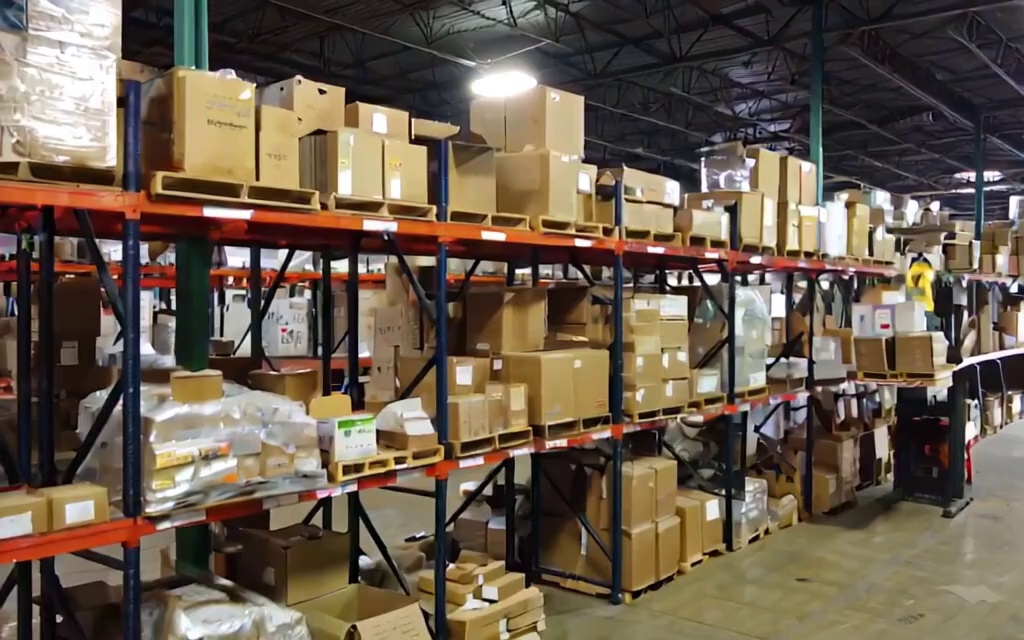} &
        \includegraphics[width=0.249\textwidth]{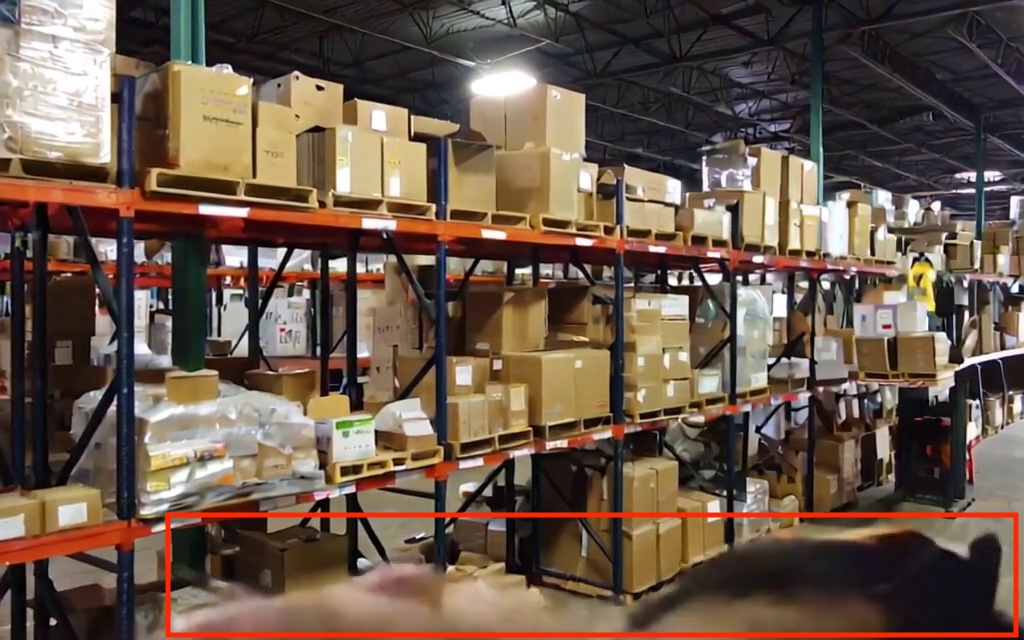} &
        \includegraphics[width=0.249\textwidth]{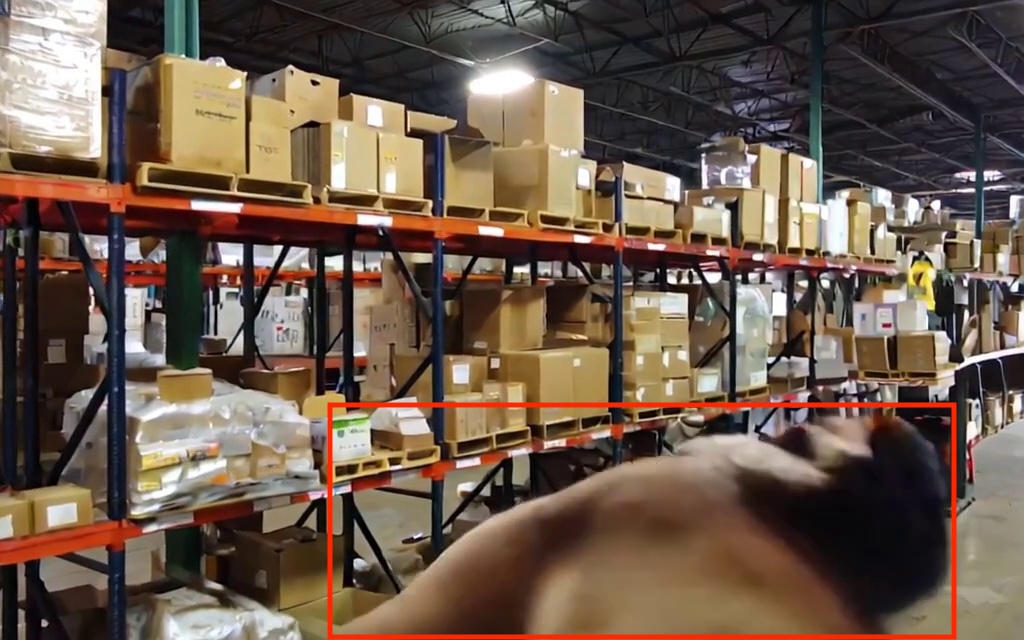} &
        \includegraphics[width=0.249\textwidth]{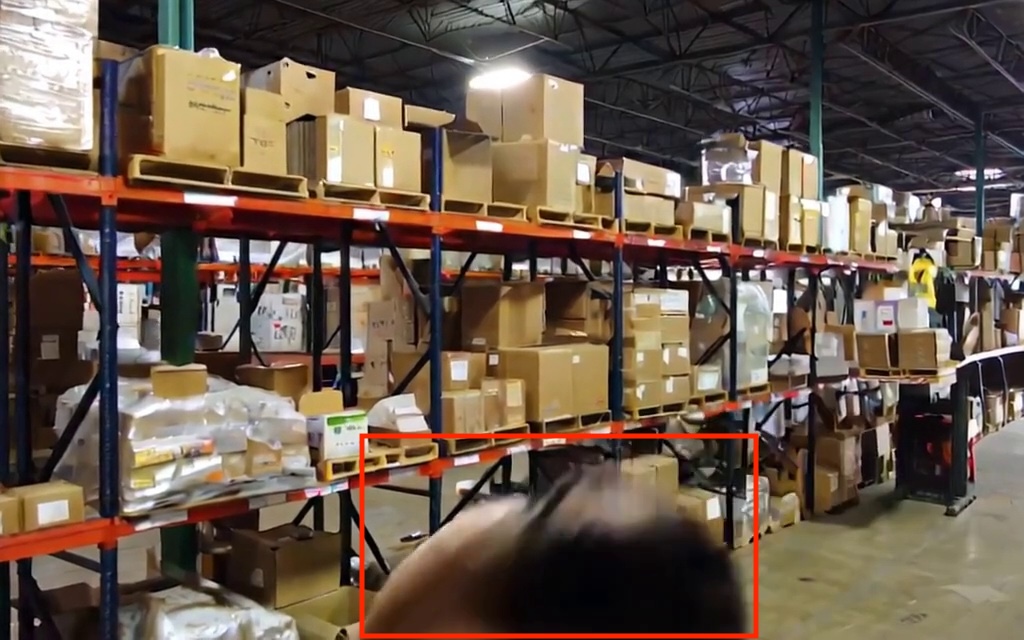} \\[4pt]
        \multicolumn{4}{c}{Output of Cosmos-Predict1-4B} \\[4pt]
    \end{tabular}
    }
    \caption{\textbf{Failure cases of Cosmos Autogressive WFMs}. We observe failure cases in generated videos in which some objects (shown in red) unexpectedly appear from below.}
    \label{fig:cosmos_ar_failure_cases}
\end{figure*}

\begin{table}[ht]
    \setlength{\tabcolsep}{15pt}
    \small
    \captionsetup{justification=centering}
    \caption{Failure rate analysis of Cosmos Autoregressive models.}
    \centering
    \begin{tabular}{r|c|c}
        \toprule
        \textbf{Model} &\textbf{Image Conditioning} & \textbf{Video Conditioning ($9$ frames)} \\
        \midrule
        Cosmos-Predict1-4B & $15\%$ & $1\%$ \\
        Cosmos-Predict1-5B-Video2World & $7\%$ & $2\%$ \\
        Cosmos-Predict1-12B & $2\%$ & $1\%$ \\
        Cosmos-Predict1-13B-Video2World & $3\%$ & $0\%$ \\

        \bottomrule
    \end{tabular}
    \label{tab:cosmos_ar_failure_rate}
\end{table}

\subsubsection{Limitations}
One notable failure case observed in the generated videos of our autoregressive WFMs is objects unexpectedly appearing from below. \cref{fig:cosmos_ar_failure_cases} illustrates an example of this issue. To understand the failure rate of our models, we conduct a systematic study by creating an evaluation set of $100$ Physical AI inputs to our autoregressive WFMs. We generate videos with all our models using two input modes---image (single-frame) conditioning and video (9-frame) conditioning. For all generated videos, we manually inspect the failure cases and report the failure rate in~\cref{tab:cosmos_ar_failure_rate}. We observe that the smaller models Cosmos-Predict1-4B and Cosmos-Predict1-5B-Video2World show a higher corruption rate in single frame conditioning, while the larger models Cosmos-Predict1-12B and Cosmos-Predict1-13B-Video2World are more robust. Generation with $9$-frame video conditioning is stable for all models, with a failure rate lower than $2\%$.

\subsection{Evaluation}

Pre-trained WFMs are generalists of visual world simulation. Their capabilities should be measured across multiple aspects. Here, we evaluate our models on two aspects. First, we evaluate the 3D consistency of the generated videos. An ideal WFM should generate video simulations from geometrically plausible 3D worlds. Second, we evaluate the physics alignment of the generated videos. We calculate how well the rendered dynamics adhere to the laws of physics. Evaluation of WFMs is a highly nontrivial task. We acknowledge that there are several other important aspects required for evaluation. We leave a more comprehensive evaluation as future work.

\subsubsection{3D Consistency}\label{sec::eval_3d_consistency}

WFMs are designed to simulate 3D worlds through video generation, and it is essential to evaluate how well the generated videos are consistent with the 3D structure of the visual world. In addition to appearing realistic, the generated videos should maintain coherence with the physical principles of scenes through time, a key requirement for downstream Physical AI applications.

\noindent \textbf{Test data and baseline model.}
We focus on the scenario of static scenes in order to effectively measure 3D consistency of videos with existing tools based on multi-view geometry. We curate a dataset of 500 videos randomly chosen from the test set of the RealEstate10K dataset~\citep{zhou2018stereo}. We additionally caption the videos using a proprietary VLM to obtain text prompts that describe the videos as static scenes, so one does not need to consider scene motions for metric computation. We compare against VideoLDM~\citep{Blattmann2023Align} as the baseline method.

\noindent \textbf{Metrics.} Generated videos are effectively 2D projections of the underlying 3D visual worlds. We design the following metrics to measure the 3D consistency of generated videos.
\begin{enumerate}
    \item \textbf{Geometric consistency}. We evaluate the 3D consistency of our generated worlds by quantifying how the epipolar geometry constraints are satisfied, including the Sampson error~\citep{sampson1982fitting,hartley2003multiple} and the success rate of camera pose estimation algorithms~\citep{schoenberger2016sfm,schonberger2016pixelwise} on the generated videos.
    \item \textbf{View synthesis consistency}. We evaluate the ability of world foundation models to synthesize images at interpolated novel viewpoints while maintaining coherence with the underlying 3D structure.
\end{enumerate}

The Sampson error is the first-order approximation of the distance from one interest point to its corresponding epipolar line in another view. Given $N$ point correspondences (represented in homogeneous coordinates) $\{\left(\bar{\x}_i, \bar{\y}_i\right)\}_{i=1}^N$ in a given frame pair, we define the Sampson error as
\begin{align}
    \epsilon_\text{samp} = \frac{1}{N} \sum_{i=1}^N \frac{|\bar{\y}_i^\top \mathbf{F} \bar{\x}_i|}{\sqrt{\left\|\mathbf{S}\mathbf{F} \bar{\x}_i\right\|_2^2 + \left\|\mathbf{S}\mathbf{F}^\top \bar{\y}_i\right\|_2^2}} \; , \quad \text{where} \; \mathbf{S} = \begin{bmatrix}
        1 & 0 & 0 \\
        0 & 1 & 0 \\
        0 & 0 & 0
    \end{bmatrix}, 
\end{align}
and $\mathbf{F}$ is the fundamental matrix estimated from the correspondences. We use the square root version of the error function to make the metric more intuitive in pixel units. We use a combination of SuperPoint~\citep{detone2018superpoint} and LightGlue~\citep{lindenberger2023lightglue} to detect and match keypoint correspondences from a frame pair and estimate $\mathbf{F}$ using OpenCV's 8-point RANSAC algorithm. We normalize the average error by the diagonal length of the frame with respect to a $960 \times 540$ canvas.

We also evaluate 3D consistency of a generated video with its ability to self-synthesize novel viewpoints. Following the common practice of novel view synthesis literature~\citep{mildenhall2020nerf}, we hold out every 8 frames as the test frames and fit a 3D Gaussian splatting model~\citep{kerbl20233d} with the rest of the training frames using the default settings from the Nerfstudio library~\citep{tancik2023nerfstudio}. We report the Peak Signal-to-Noise Ratio (PSNR), Structural Similarity (SSIM), and LPIPS~\citep{zhang2018unreasonable} as the metrics to quantify the quality of the synthesized test views.

\begin{table}[ht]
    \setlength{\tabcolsep}{5.3pt}
    \small
    \captionsetup{justification=centering}
    \caption{Evaluation of 3D consistency on base Cosmos models.}
    \centering
    \vspace{-1em}
    \setlength{\tabcolsep}{6pt}
    \begin{tabular}{rccccc}
        \toprule
        & \multicolumn{2}{c}{\bf Geometric Consistency} & \multicolumn{3}{c}{\bf View Synthesis Consistency} \\
        \cmidrule(r){2-3} \cmidrule(r){4-6}
        & \multirow{2}{*}{Sampson error $\downarrow$} & Pose estimation & \multirow{2}{*}{PSNR $\uparrow$} & \multirow{2}{*}{SSIM $\uparrow$} & \multirow{2}{*}{LPIPS $\downarrow$} \\
        Method & & success rate (\%) $\uparrow$ & & & \\
        \midrule
        VideoLDM~\citep{Blattmann2023Align}      & 0.841 & 4.4\%  & 26.23 & 0.783 & 0.135 \\
        \midrule
        Cosmos-Predict1-7B-Text2World       & \bf 0.355 & 62.6\% & \bf 33.02 & \bf 0.939 & \bf 0.070 \\
        Cosmos-Predict1-7B-Video2World      & 0.473 & \bf 68.4\% & 30.66 & 0.929 & 0.085 \\
        Cosmos-Predict1-4B & 0.433 & 35.6\% & 32.56 & 0.933 & 0.090 \\
        Cosmos-Predict1-5B-Video2World & 0.392 & 27.0\% & 32.18 & 0.931 & 0.090 \\
        \midrule
        Real Videos (Reference)                  & 0.431 & 56.4\% & 35.38 & 0.962 & 0.054 \\
        \bottomrule
    \end{tabular}
    \label{tab:3d-consistency}
\end{table}

\noindent \textbf{Results.} We present the quantitative evaluation results in~\cref{tab:3d-consistency}. The Cosmos WFMs achieve significantly better 3D consistency than our baseline model in terms of both geometric and view synthesis consistency. Not only are the interest points from Cosmos WFMs more 3D-consistent, but the camera pose estimation success rate is also notably higher, reflecting both improved overall quality and enhanced 3D consistency, even reaching the level of real-world videos. Among the cases where camera poses were successfully estimated, the synthesized held-out views demonstrate higher quality across all image synthesis metrics. These results highlight the capability of our Cosmos WFMs to generate 3D-consistent videos, establishing them as effective world simulators.

\subsubsection{Physics Alignment} \label{sec::eval_physical_alignment}

An ideal WFM should exhibit a strong understanding of the laws of physics and produce future observations that respect them. While our pre-trained WFMs exhibit a certain level of physics understanding and advance the state-of-the-art, one can still easily generate examples that do not obey the law of physics. We believe additional steps in data curation where physically implausible videos are removed are required, as well as improved model design. While we leave a strong physics-aligned WFM as future work, we are still interested in measuring how much intuitive physics naturally emerges from large-scale data-driven pre-training. 

To explore this, we design acontrolled benchmark dataset using a physics simulation engine, taking inspiration from \citep{kang2024how}. We generate physics-grounded simulations to test the adherence of our pre-trained WFMs to Newtonian physics and rigid body dynamics. Specifically, we use simulation to generate physically correct photorealistic videos of test scenarios specific to physical laws of interest. These reference ``ground truth'' videos are then compared with ``predicted'' videos produced by a WFM given shared context (past observations and perturbation).

\noindent \textbf{Synthetic data generation.} Using PhysX \citep{PhysX} and Isaac Sim \citep{IsaacSim}, we design eight 3D scenarios aimed at evaluating different physical effects:
\begin{enumerate}
    \item \textbf{Free-falling object(s)}: objects dropping on a plane (gravity, collision, \etc)
    \item \textbf{Tilted planar slope}: objects rolling down an incline (gravity, moment of inertia, \etc)
    \item \textbf{U-shaped slope}: objects rolling down a U-shaped slope (potential, kinetic energy, \etc)
    \item \textbf{Stable stack}: a stack of objects in equilibrium (balanced forces)
    \item \textbf{Unstable stack}: a stack of objects in imbalance (gravity, collision, \etc)
    \item \textbf{Dominoes}: sequence of rectangular bricks falling in sequence (transfer of momentum, collision, \etc)
    \item \textbf{Seesaw}: objects on either side of a seesaw (torque, rotational inertia, \etc)
    \item \textbf{Gyroscope}: a spinning top on a flat surface (angular momentum, precession, \etc)
\end{enumerate}

For each scenario, we randomize the number and type of dynamic objects (varying sizes, textures, shapes), selecting from Omniverse assets \citep{Omniverse}, as well as the background appearance. We simulate the kinematic state of objects over time and render the output videos from 4 different static camera views. In total, we render 800 1080p videos of 100 frames in length. The objects in each simulation roll-out are positioned so that they are all visible from the first frame to avoid any existence ambiguity.

\begin{figure}[!tp]
    \centering
    \setlength{\tabcolsep}{1.5pt}
    \renewcommand{\arraystretch}{0}
    \resizebox{\textwidth}{!}{%
    \begin{tabular}{ccccc} %
        & \multicolumn{4}{c}{\small{\prompt{\textbf{Tilted planar slope} - An object rolling down an inclined plane}}} \\[3pt]
        {\rotatebox{90}{\hspace{6pt}Simulated}} & \includegraphics[width=0.23\textwidth]{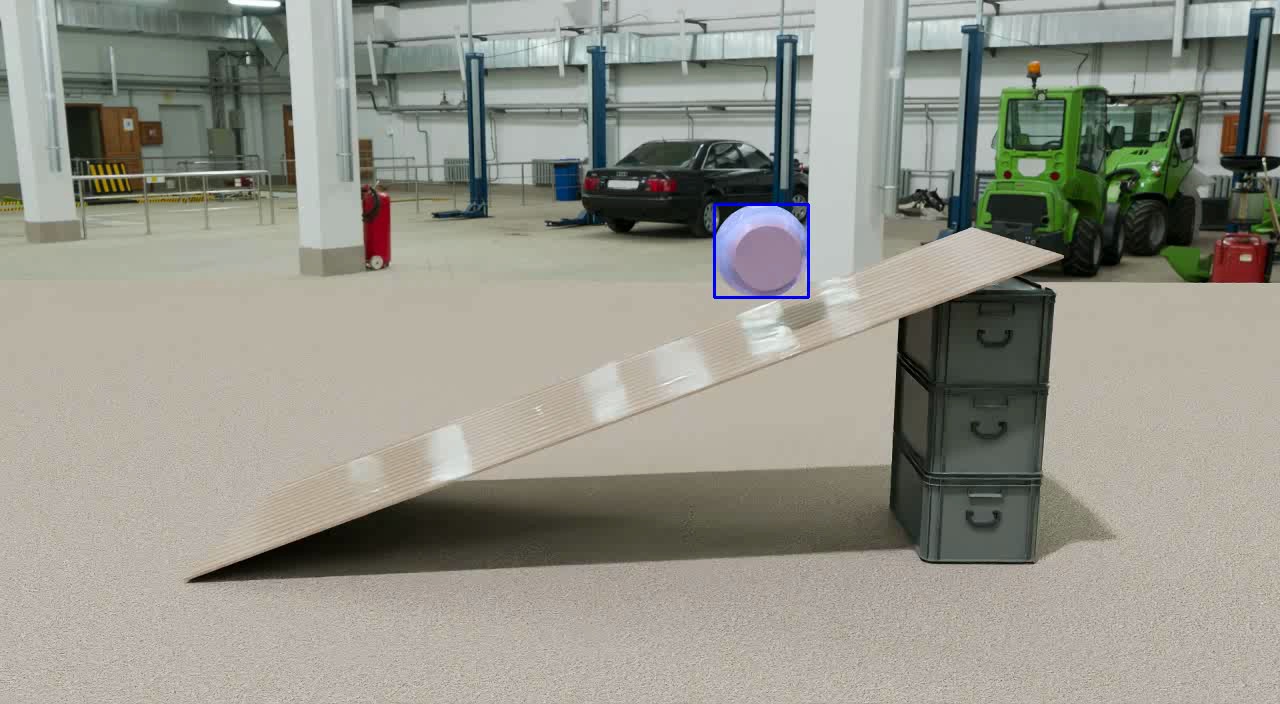} &
        \includegraphics[width=0.23\textwidth]{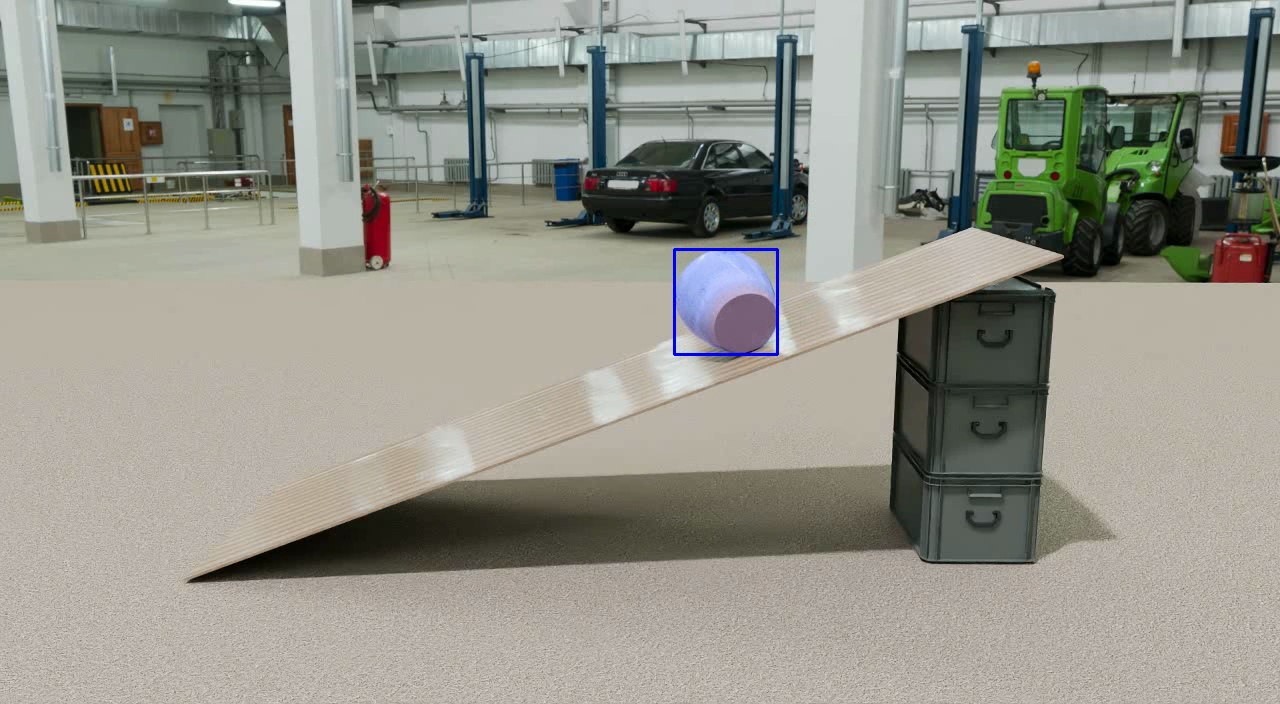} &
        \includegraphics[width=0.23\textwidth]{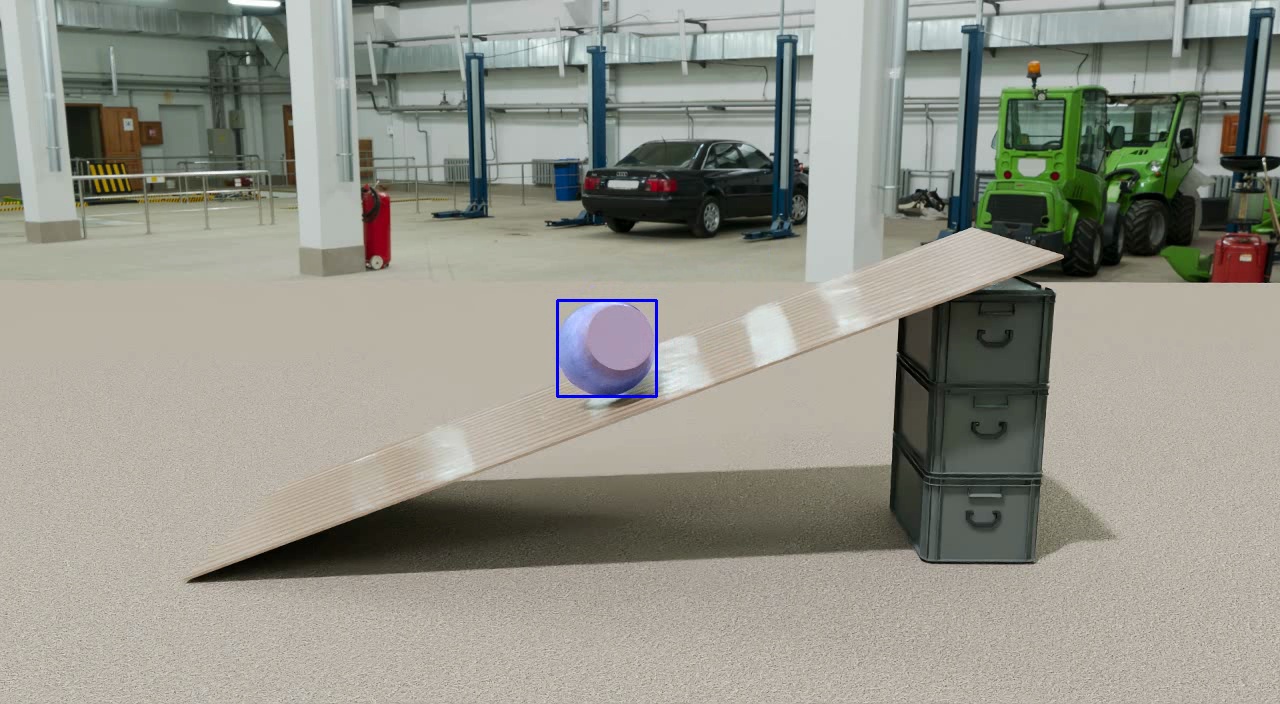} &
        \includegraphics[width=0.23\textwidth]{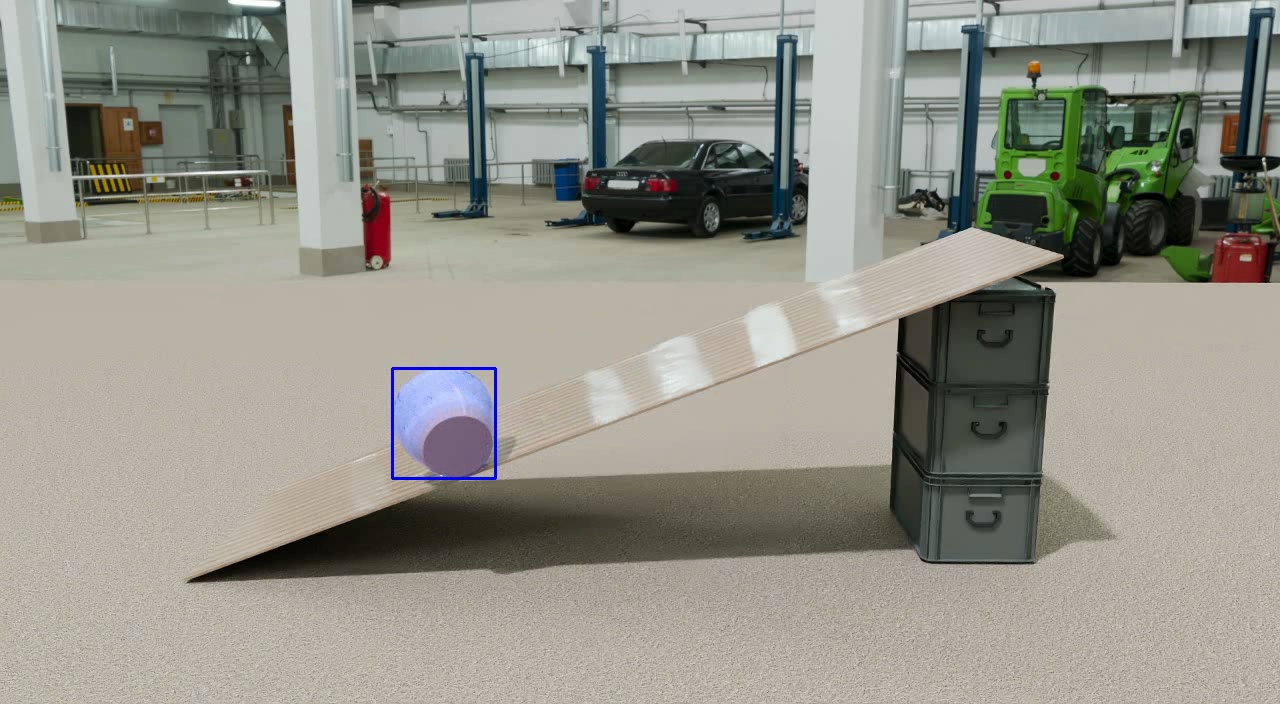} \\[1pt]
        {\rotatebox{90}{\hspace{15pt}WFM}} & \includegraphics[width=0.23\textwidth]{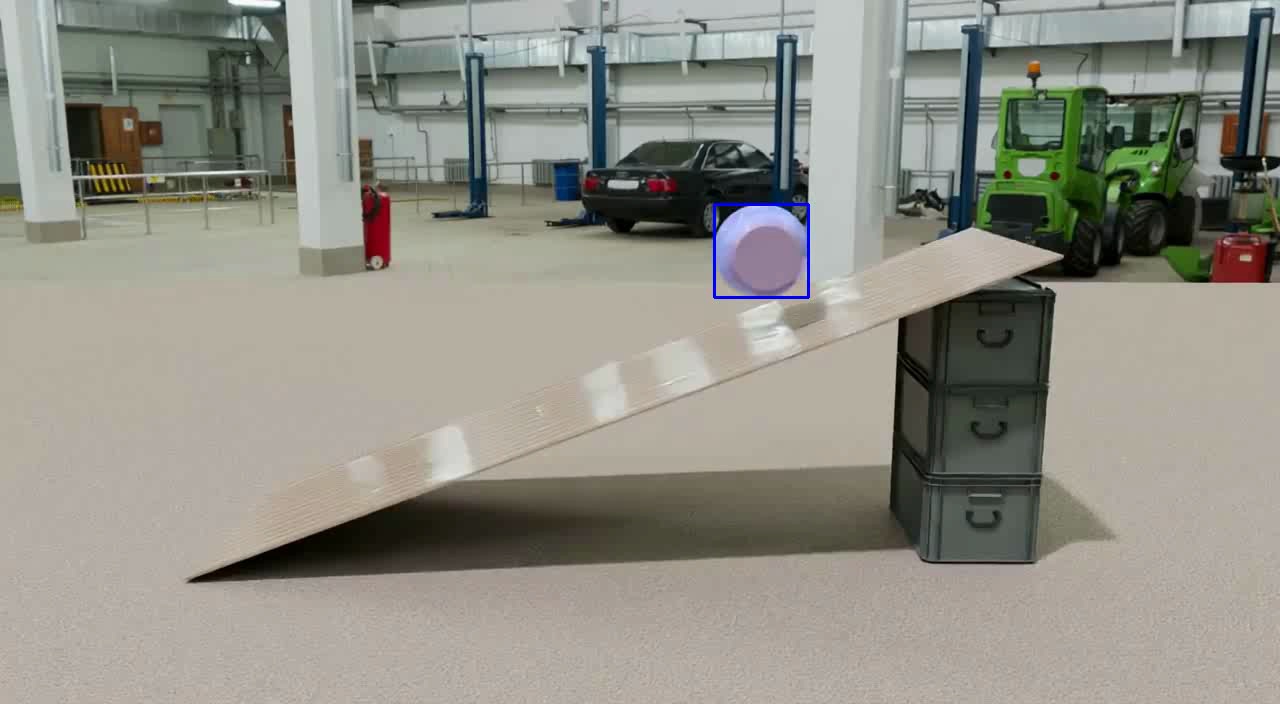} &
        \includegraphics[width=0.23\textwidth]{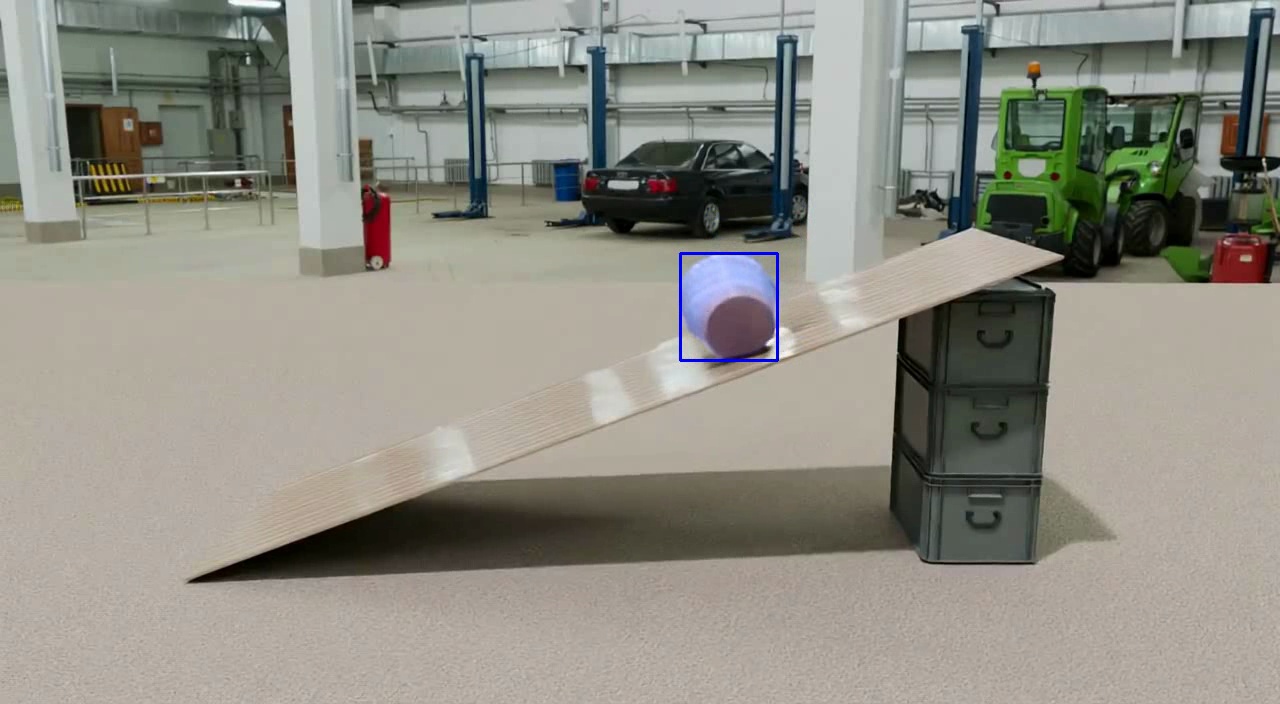} &
        \includegraphics[width=0.23\textwidth]{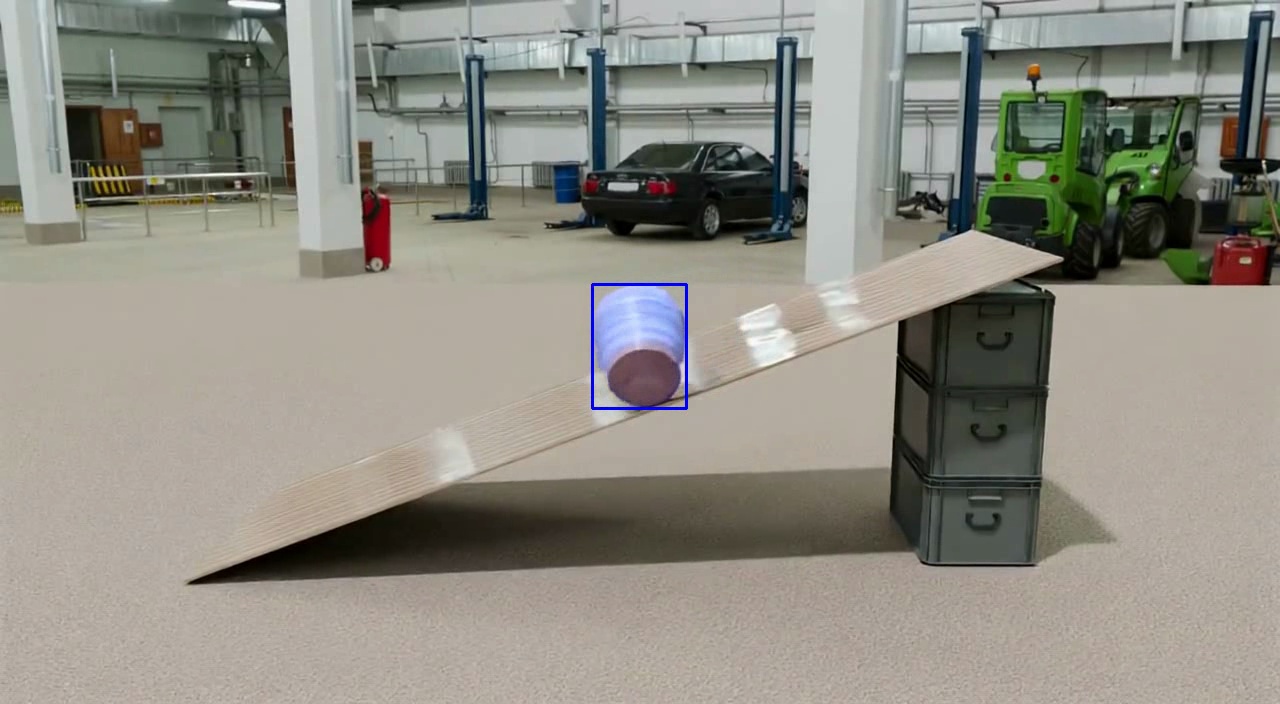} &
        \includegraphics[width=0.232\textwidth]{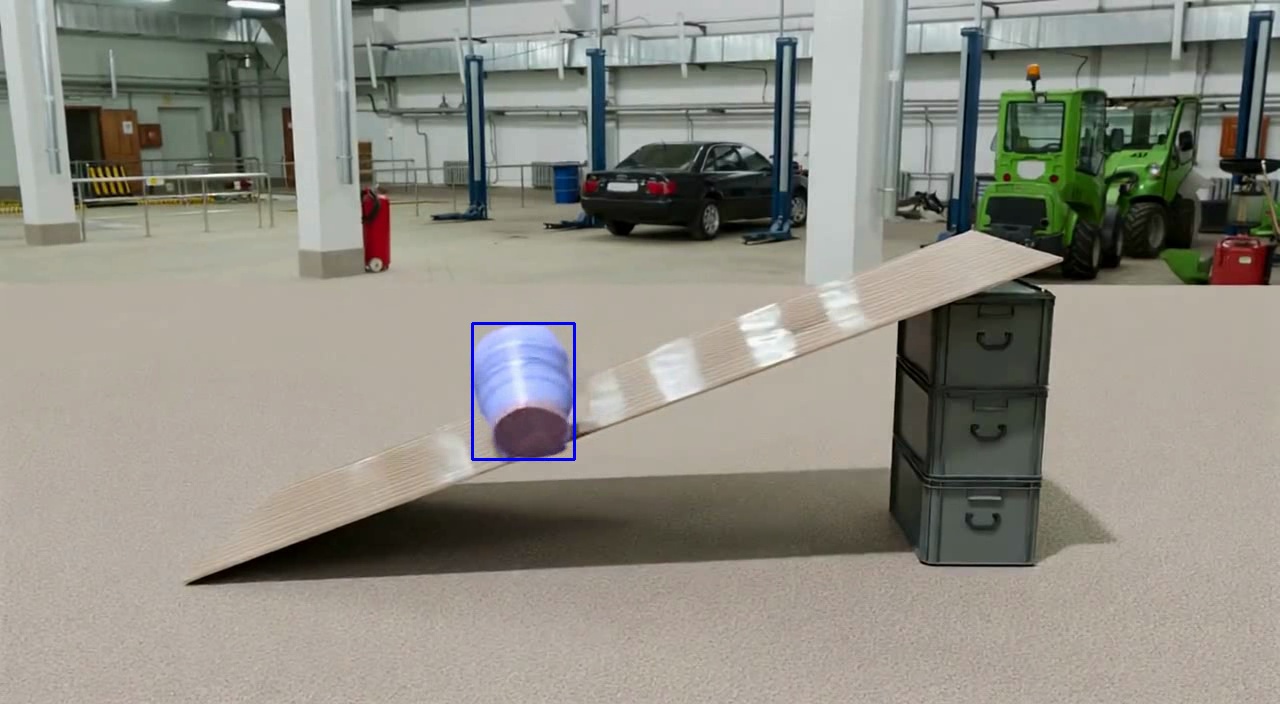} \\[2pt]

        & \multicolumn{4}{c}{\small{\prompt{\textbf{U-shaped slope} - Two objects rolling down from either ends of a curved slope}}} \\[3pt]
        {\rotatebox{90}{\hspace{6pt}Simulated}} &\includegraphics[width=0.23\textwidth]{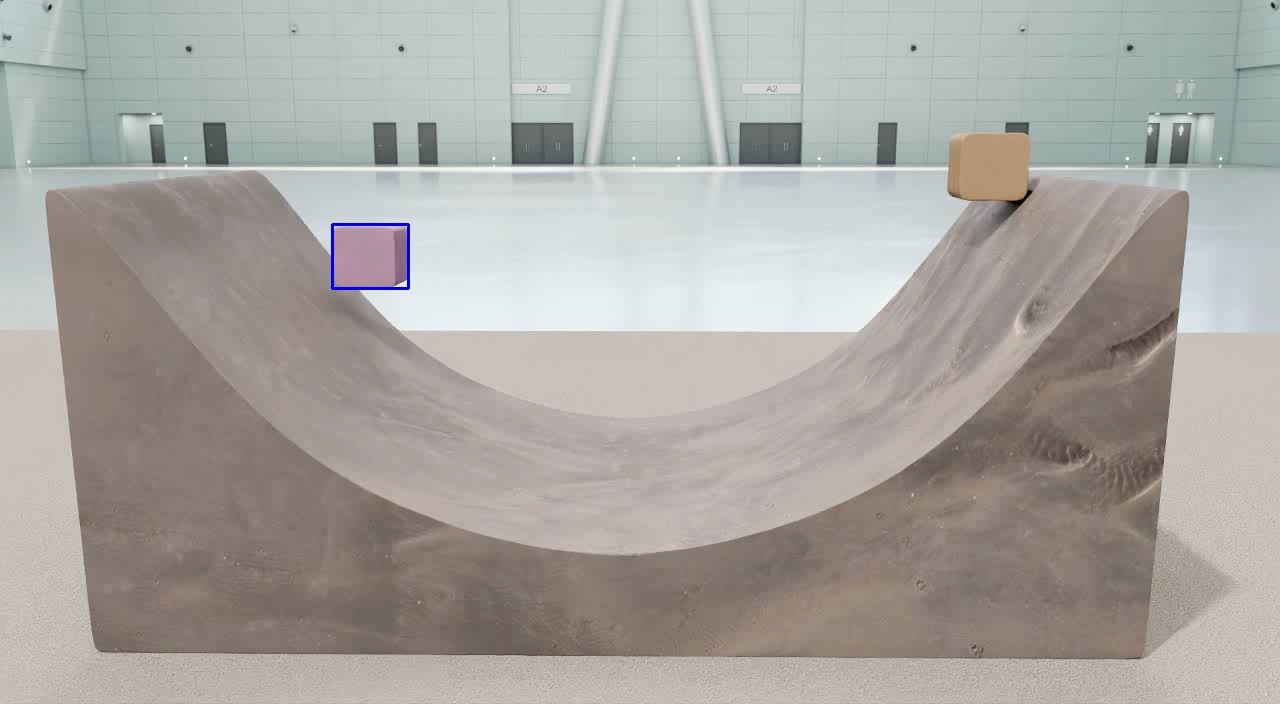} &
        \includegraphics[width=0.23\textwidth]{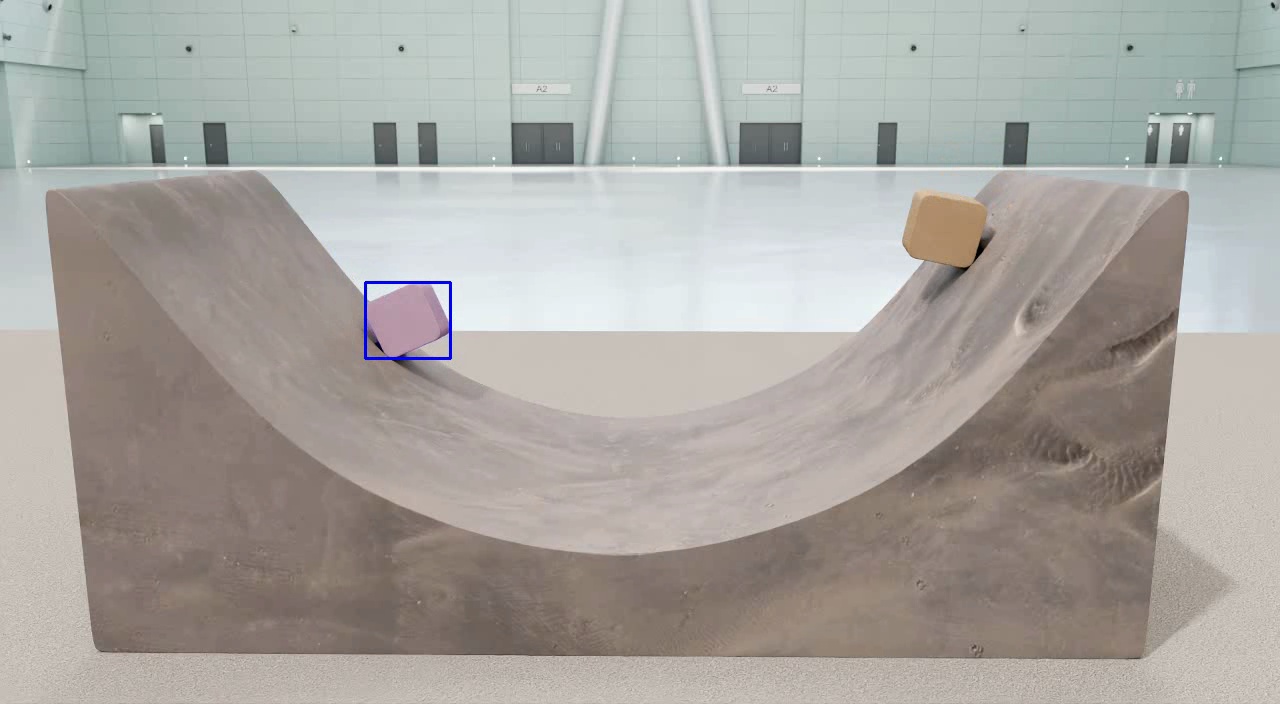} &
        \includegraphics[width=0.23\textwidth]{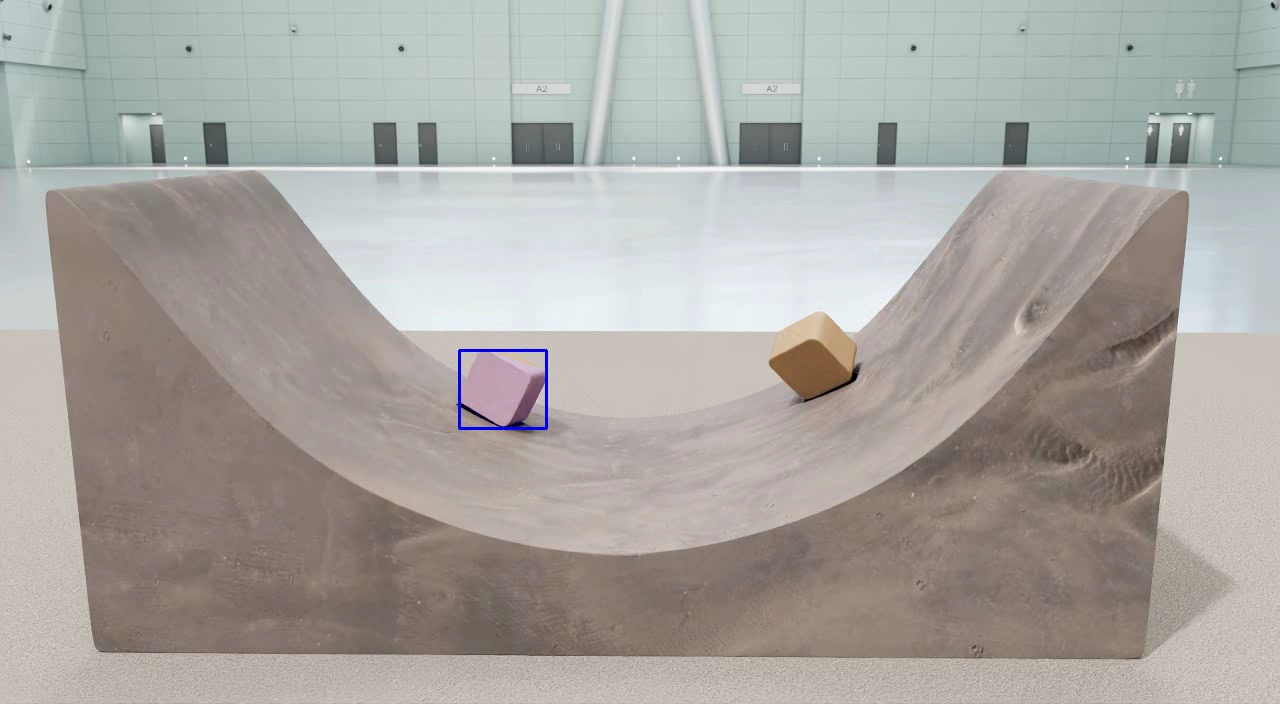} &
        \includegraphics[width=0.23\textwidth]{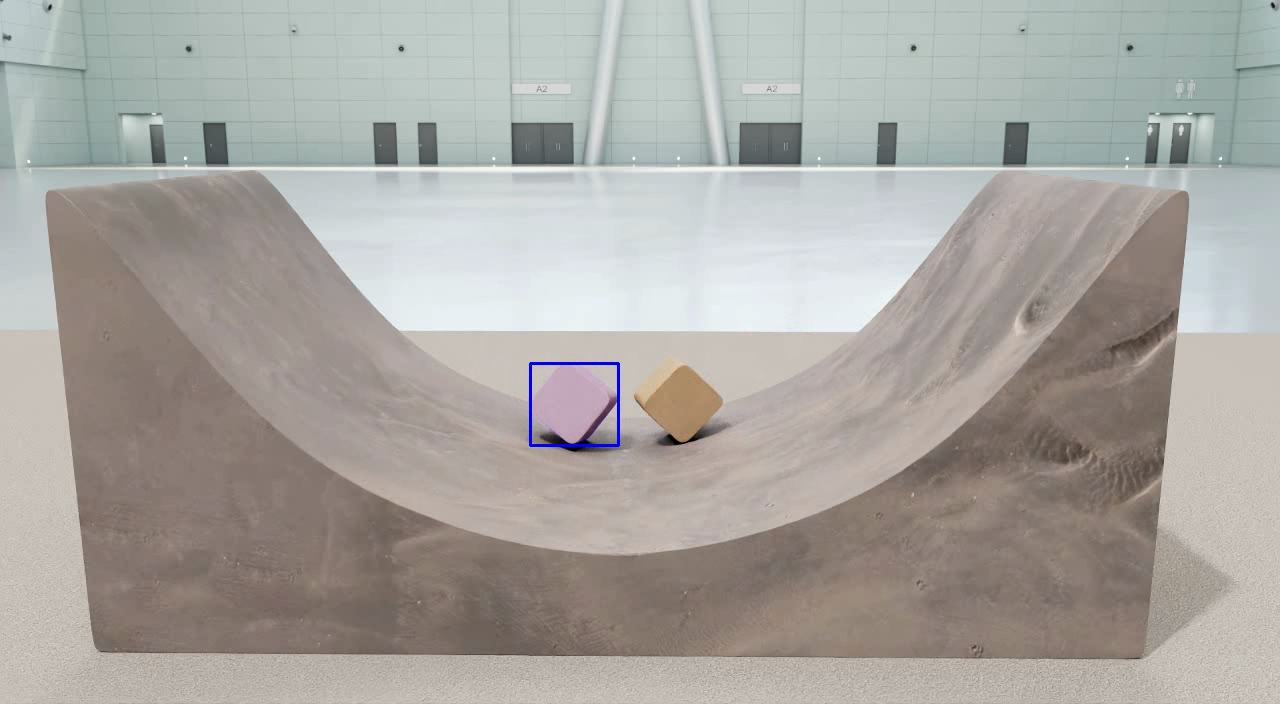} \\[1pt]
        {\rotatebox{90}{\hspace{15pt}WFM}} & \includegraphics[width=0.23\textwidth]{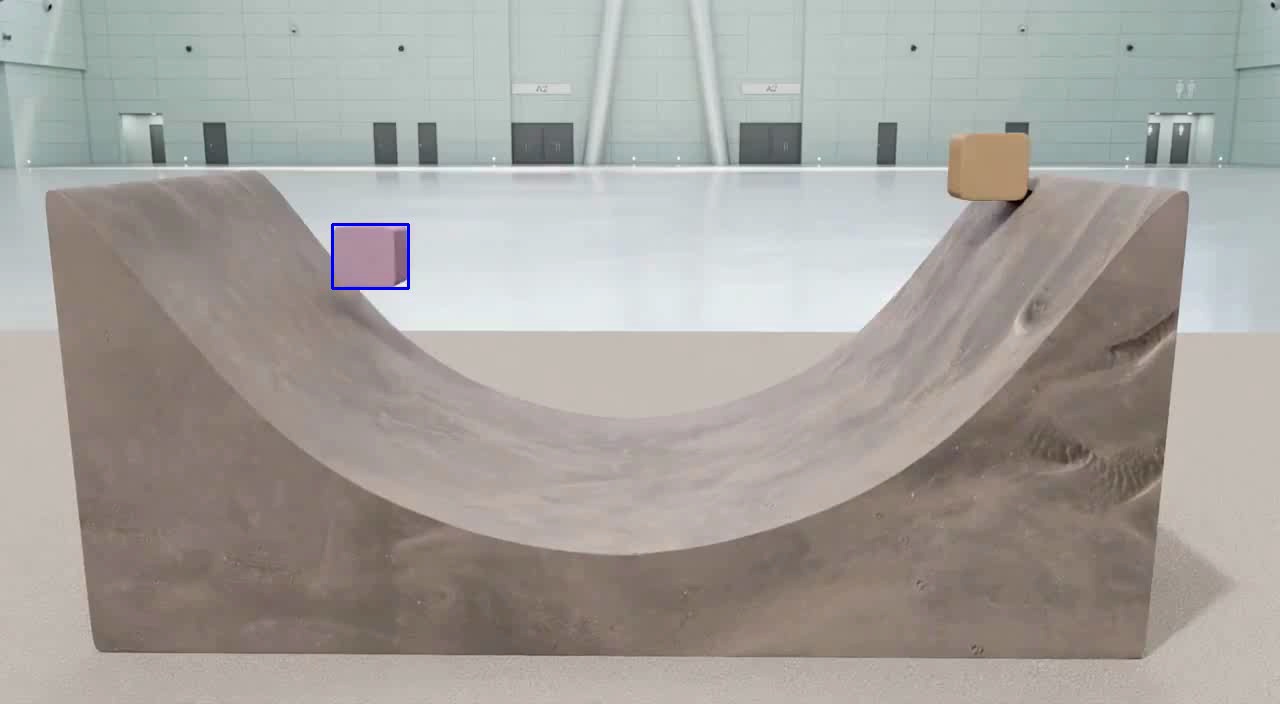} &
        \includegraphics[width=0.23\textwidth]{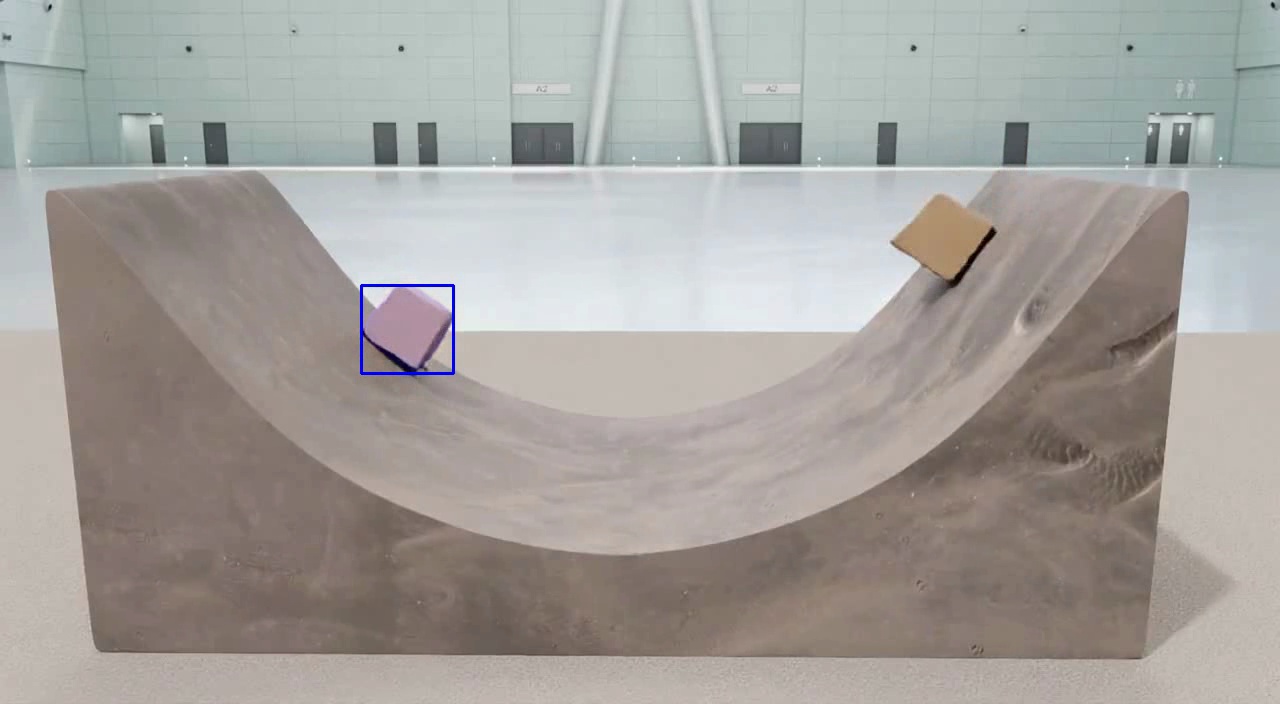} &
        \includegraphics[width=0.23\textwidth]{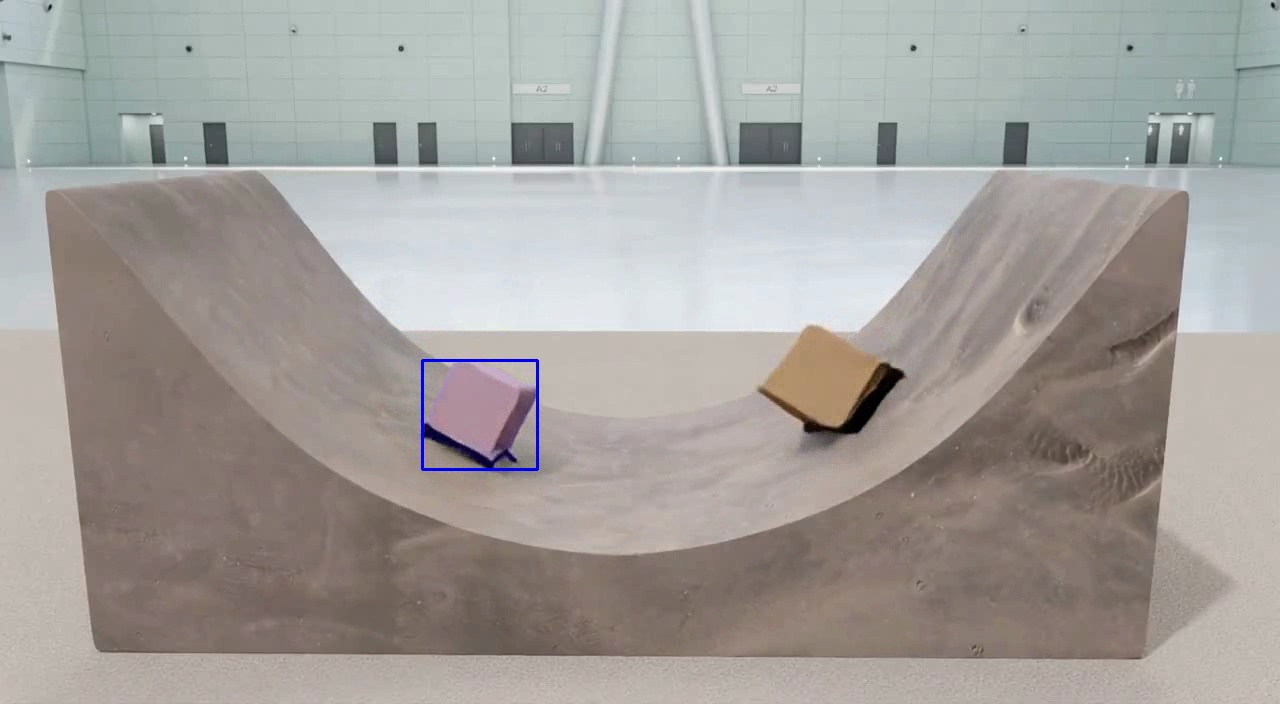} &
        \includegraphics[width=0.23\textwidth]{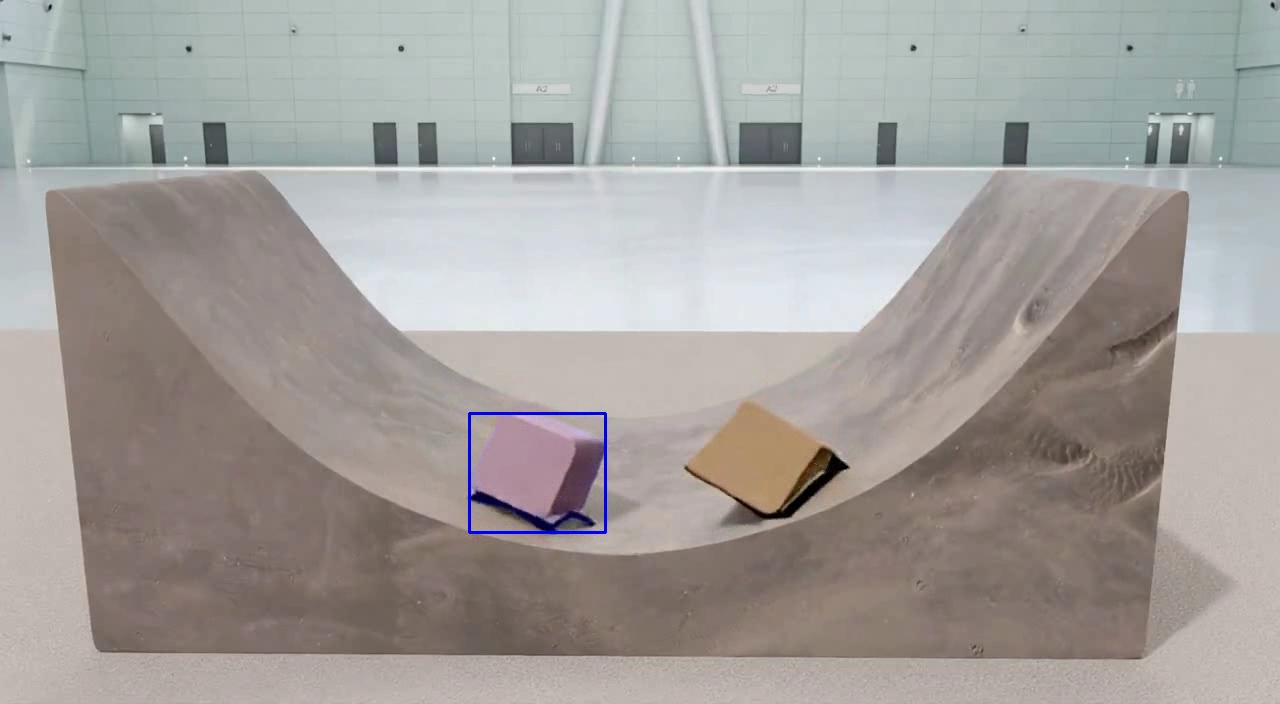} \\[2pt]

        & \multicolumn{4}{c}{\small{\prompt{\textbf{Unstable stack} - An unstable stack of objects falling down due to imbalanced forces}}} \\[3pt]
        {\rotatebox{90}{\hspace{6pt}Simulated}} &\includegraphics[width=0.23\textwidth]{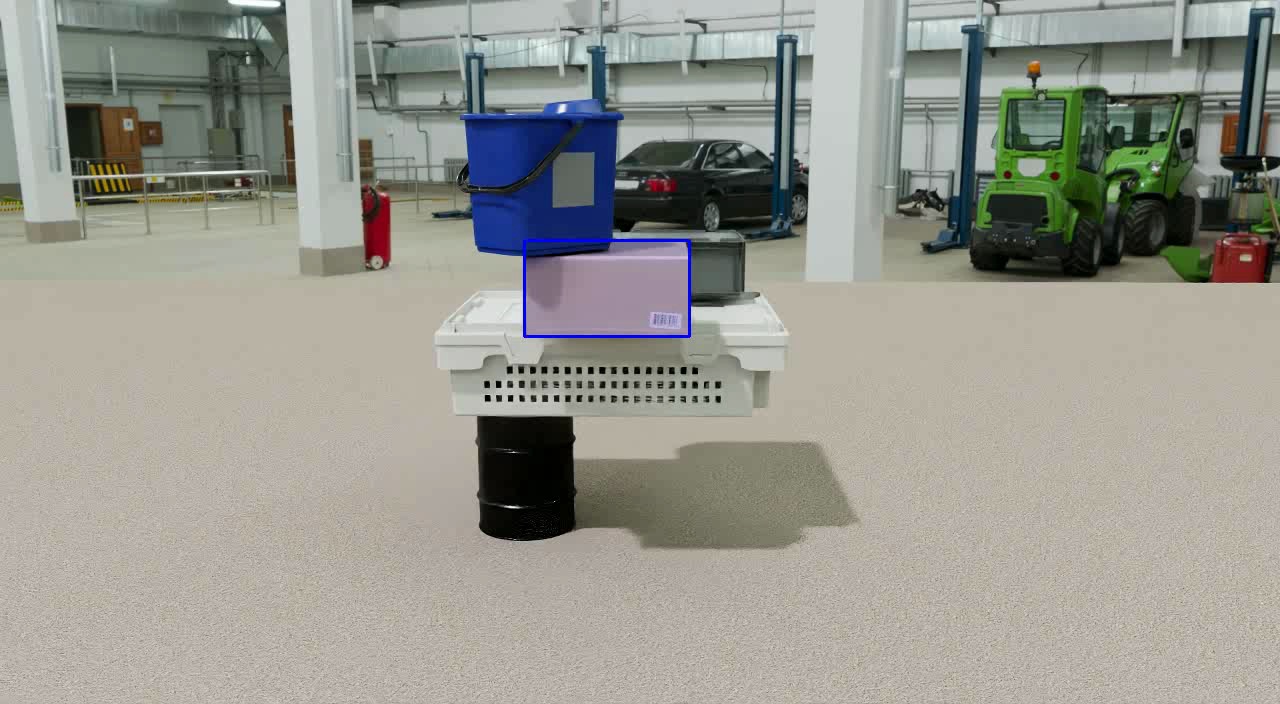} &
        \includegraphics[width=0.23\textwidth]{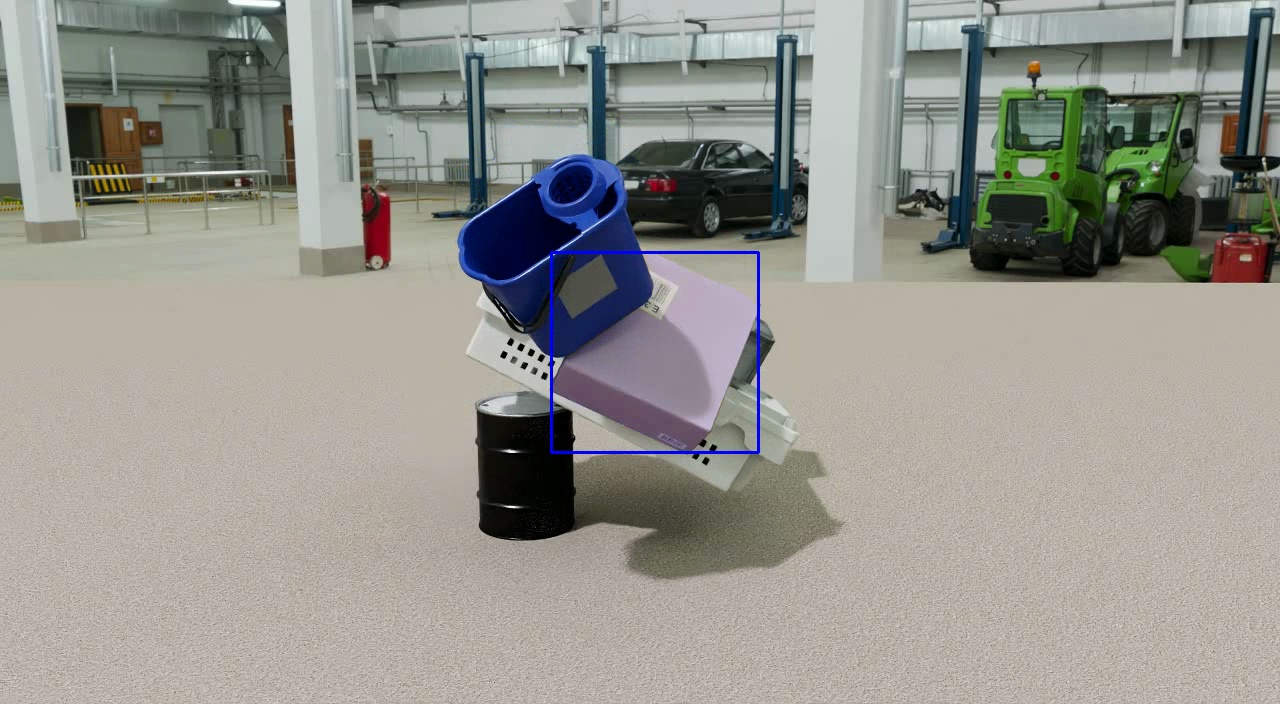} &
        \includegraphics[width=0.23\textwidth]{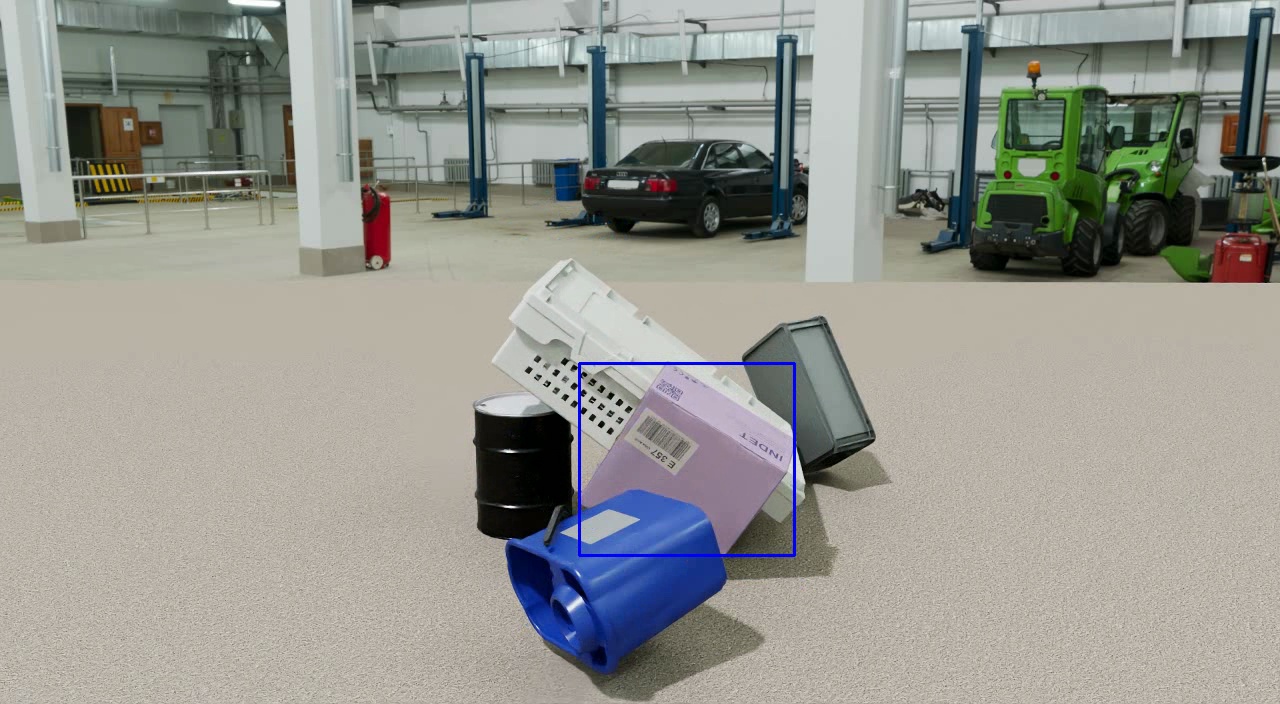} &
        \includegraphics[width=0.23\textwidth]{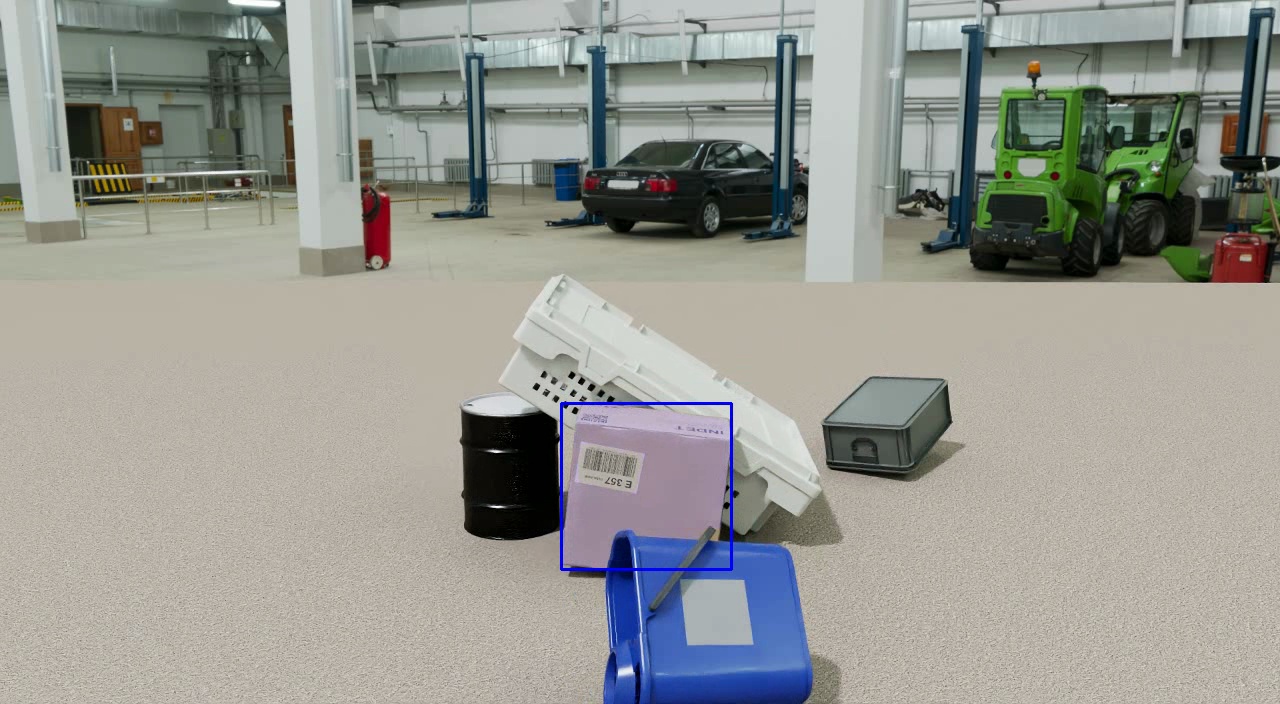} \\[1pt]
        {\rotatebox{90}{\hspace{15pt}WFM}} & \includegraphics[width=0.23\textwidth]{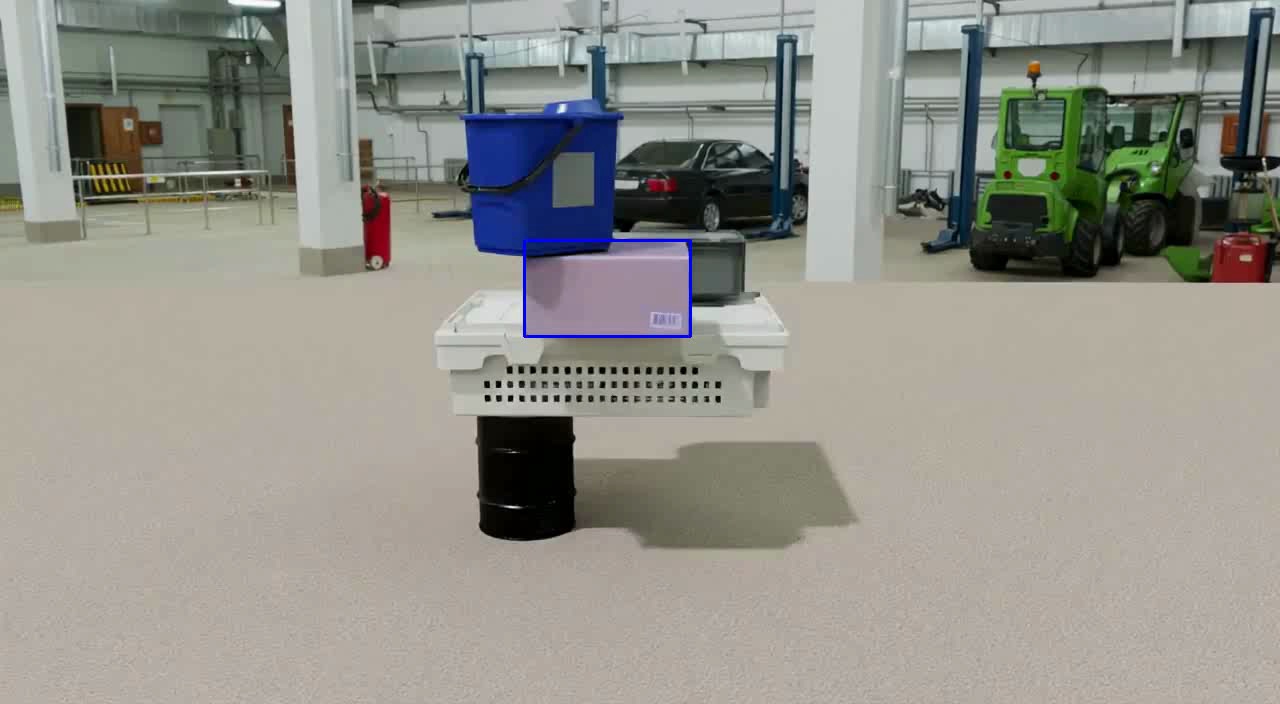} &
        \includegraphics[width=0.23\textwidth]{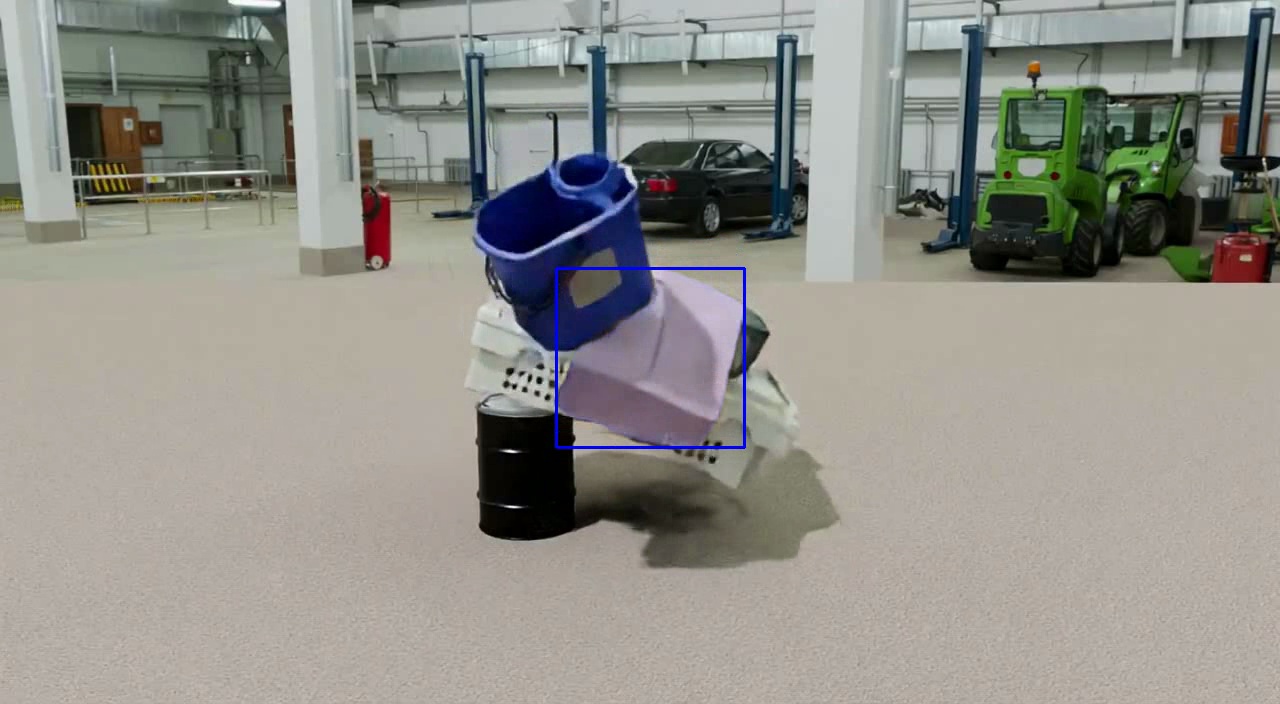} &
        \includegraphics[width=0.23\textwidth]{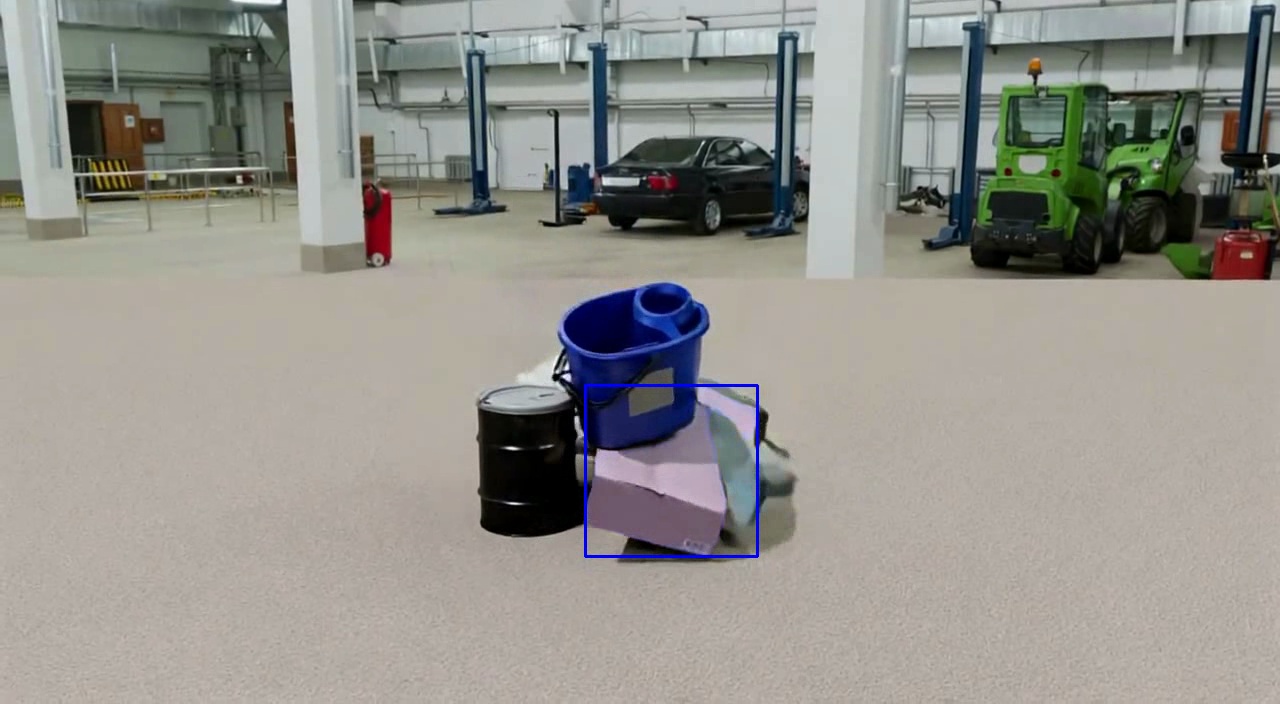} &
        \includegraphics[width=0.23\textwidth]{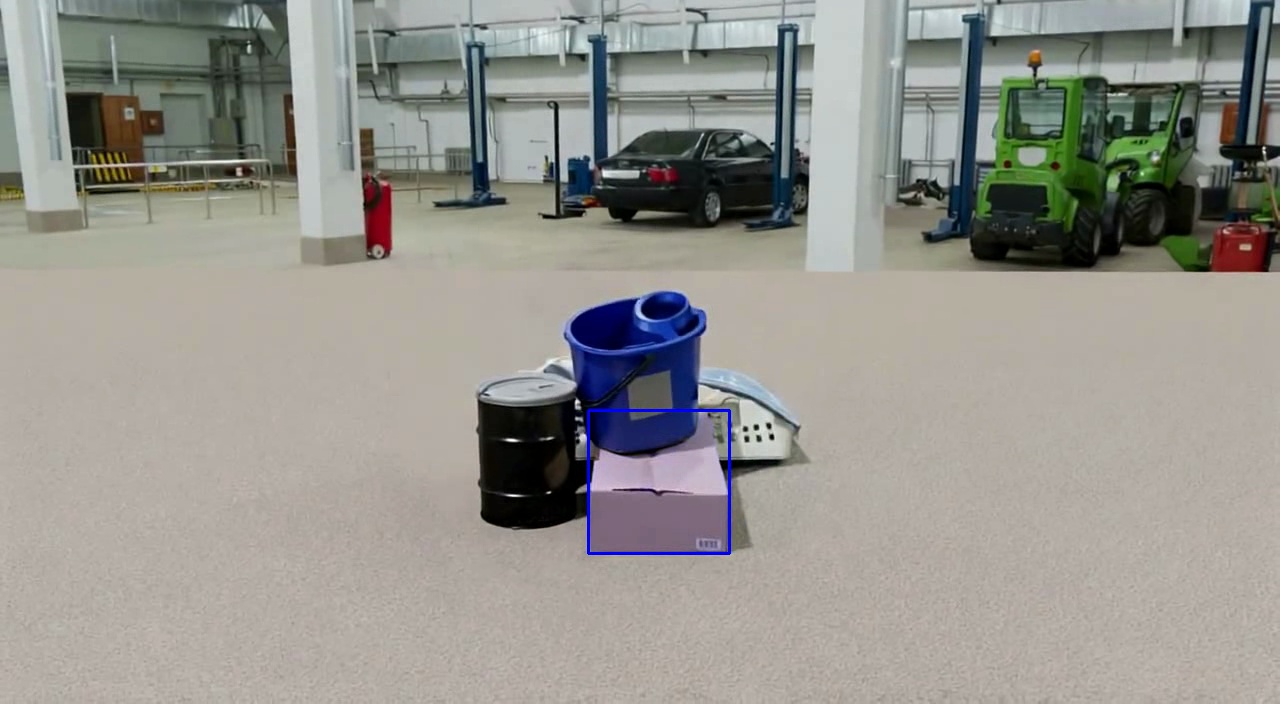} \\[2pt]

        & $t=0$ (Conditioning) & $t=11$ & $t=22$ & $t=32$
    \end{tabular}}
    \caption{\textbf{Physics-scenario rollouts in simulation \vs pre-trained WFM.} We demonstrate three exemplar scenarios of increasing complexity as obtained from the reference (physically correct) simulation (first row in each group) and Cosmos-Predict1-7B-Video2World rollouts (second row in each group). We condition the WFM on 9 frames and a prompt focusing on the kinematic state of the simulated objects. We show one tracked object (blue bounding box and mask) per example used to compute our object-level metrics (average IOU).}
    \label{tab:phyics_eval_examples}
\end{figure}

\noindent \textbf{Metrics.} We are interested in assessing the adherence to physical laws by comparing the simulated ground-truth video to the output directly generated by the WFM. Therefore, to produce future observations, we condition our WFMs on the first few frames (either 1 or 9 frames) of the ground truth video. When applicable, we additionally condition a WFM on a text prompt (obtained using a proprietary VLM by captioning the conditioning frames), focusing on the kinematic state of the objects being simulated in the past observations. Please refer to~\cref{tab:phyics_eval_examples} for some examples of simulated versus predicted scenarios. For evaluation, we use the following metrics:
\begin{enumerate}
    \item \textbf{Pixel-level metrics}. For a pixel-level comparison, we compute the Peak Signal-to-Noise Ratio (PSNR) and Structural Similarity Index Measure (SSIM) to compare a predicted frame from the WFM rollout with the reference frame from the ground truth video.
    \item \textbf{Feature-level metrics}. For a slightly higher-level semantic comparison, we calculate DreamSim similarity scores~\citep{fu2023dreamsim}, a feature similarity metric, between the predicted and reference frames.
    \item \textbf{Object-level metrics}. Finally, since we care most about how objects of interest are impacted by the ongoing physical phenomenon, we use tracking to compute object-level metrics that eliminate confounders (background changes, visual quality, \etc).
Since the test conditions are synthetically generated, we have access to the ground-truth instance segmentation masks of the dynamic objects in the scenes. Using SAMURAI \citep{yang2024samurai}, we propagate the ground-truth instance masks in the first frame through the rest of the predicted video frames to extract tracks, allowing us to quantify object-level metrics. We compute the intersection-over-union (IoU) between ground truth and predicted object masks for each frame and object of interest.
\end{enumerate}

We average these metrics across frames in a video, across videos in the evaluation set, and across four random seeds for rollouts. PSNR and SSIM are computed on all frames, excluding the ones used for conditioning.

\begin{table}[!tp]
    \centering
    \setlength{\tabcolsep}{4.7pt}
    \caption{Physics alignment results. We compare different variants of Cosmos WFMs in terms of accurate future prediction of a physical scenario using the pixel-level, feature-level, and object-level metrics. Metrics are calculated over 33 frames, the maximum length supported by the autoregressive variants of the Cosmos WFMs.}
    \resizebox{\textwidth}{!}{%
    \begin{tabular}{rccccc}
        \toprule
        & & \multicolumn{2}{c}{\textbf{Pixel-level}} & \multicolumn{1}{c}{\textbf{Feature-level}} & \multicolumn{1}{c}{\textbf{Object-level}} \\
        \cmidrule(r){3-4} \cmidrule(r){5-5} \cmidrule(r){6-6}
        Model & Condition(s) & PSNR $\uparrow$ & SSIM $\uparrow$ & DreamSim $\uparrow$ & Avg. IoU $\uparrow$ \\
        \midrule
        Cosmos-Predict1-7B-Video2World & prompt + 1 frame & 17.34 & 0.538 & 0.836 & 0.332 \\
        Cosmos-Predict1-7B-Video2World & prompt + 9 frames & \textbf{21.06} & \textbf{0.691} & 0.859 & 0.592 \\
        \midrule
        Cosmos-Predict1-14B-Video2World & prompt + 1 frame & 16.81 & 0.521 & 0.836 & 0.338 \\
        Cosmos-Predict1-14B-Video2World & prompt + 9 frames & 20.21 & 0.635 & 0.860 & \textbf{0.598} \\
        \midrule
        Cosmos-Predict1-4B & 1 frame & 17.91 & 0.486 & 0.827 & 0.394 \\
        Cosmos-Predict1-4B & 9 frames & 18.13 & 0.482 & 0.859 & 0.481 \\
        \midrule
        Cosmos-Predict1-5B-Video2World & prompt + 1 frame & 17.67 & 0.478 & 0.818 & 0.376 \\
        Cosmos-Predict1-5B-Video2World & prompt + 9 frames & 18.29 & 0.481 & 0.864 & 0.481 \\
        \midrule
        Cosmos-Predict1-12B & 1 frame & 17.94 & 0.486 & 0.829 & 0.395 \\
        Cosmos-Predict1-12B & 9 frames & 18.22 & 0.487 & \textbf{0.869} & 0.487 \\
        \midrule
        Cosmos-Predict1-13B-Video2World & prompt + 1 frame & 18.00 & 0.486 & 0.830 & 0.397 \\
        Cosmos-Predict1-13B-Video2World & prompt + 9 frames & 18.26 & 0.482 & 0.865 & 0.482 \\
        \bottomrule
    \end{tabular}}
    \label{tab:phyics_eval}
\end{table}

\noindent \textbf{Results.} Quantitative results on physical alignment are outlined in~\cref{tab:phyics_eval}. Based on quantitative and qualitative results, we make the following observations. Unsurprisingly, the models are able to better predict the overall object kinematics with more frames as conditioning input (which allows us to better infer 1st and 2nd order quantities such as speed and acceleration).

From the table, we also find that our diffusion WFMs perform better in pixel-level prediction than our autoregressive WFMs on the 9-frame conditional setting. This correlates with our visual observation that the diffusion-based WFMs render videos with higher visual quality. We also note that our results do not suggest that the larger model performs better on our physics alignment. While we observe larger models render videos with higher visual quality, all the WFMs equally struggle with physics adherence and require better data curation and model design.

More generally, we observe that the rigid-body simulations described above already test the limits of our WFMs, serving as valuable tools for identifying specific failure cases. These range from low-level issues like object impermanence (spontaneous appearance and disappearance of objects) and deformation (shape changes) to more complex problems such as implausible kinematics, violation of gravity, \etc. We believe such structured simulations offer a useful methodology to test physics alignment. We, therefore, intend to improve them over time by incorporating more complex scenarios, enhancing photorealism to bridge the sim-to-real gap (since WFM pre-training data consists of real videos), and refining our evaluation metrics for a more comprehensive assessment of physical understanding.

\section{Post-trained World Foundation Model}\label{sec::posttrained_world_models}

In this section, we demonstrate how our Cosmos WFMs can be fine-tuned to support diverse Physical AI applications. We include examples from post-training our WFM with camera control to achieve 3D navigable visual world generation, post-training our WFM with action control on two different robotic setups for two different robotic manipulation tasks, and post-training our WFM with multi-view support for training autonomous driving agents.

\begin{table}[ht]
    \setlength{\tabcolsep}{9.8pt} %
    \small
    \captionsetup{justification=centering}
    \caption{A map of Post-trained WFMs discussed in~\cref{sec::posttrained_world_models}.}
    \centering
    \begin{tabular}{l|l|c}
        \toprule
            \textbf{Section} & \textbf{Model} & \textbf {Condition(s)}   \\
        \midrule
        \cref{sec::camera} & Cosmos-Predict1-7B-Video2World-Sample-CameraCond & Text + Image + Cameras \\
        \midrule
        \cref{sec::robo} & Cosmos-Predict1-7B-Video2World-Sample-Instruction & Text + Video \\
        \cref{sec::robo} & Cosmos-Predict1-7B-Video2World-Sample-Instruction & Text + Video \\
        \cref{sec::robo} & Cosmos-Predict1-7B-Video2World-Sample-ActionCond & Action + Video \\
        \cref{sec::robo} & Cosmos-Predict1-7B-Video2World-Sample-ActionCond & Action + Video \\
        \midrule
        \cref{sec::av} & Cosmos-Predict1-7B-Text2World-Sample-MultiView & Text \\
        \cref{sec::av} & Cosmos-Predict1-7B-Text2World-Sample-MultiView-TrajectoryCond & Text + Trajectory \\
        \cref{sec::av} & Cosmos-Predict1-7B-Video2World-Sample-MultiView &  Text + Video \\
        \bottomrule
    \end{tabular}
    \label{tab:post_model_summarize}
\end{table}

\cref{tab:post_model_summarize} provides a list of the discussed post-trained WFMs in different subsections of this section. We also list the conditional inputs to highlight the operation mode. Note that for each model, we add ``-Sample`` to emphasize our goal is to provide sample applications of our pre-trained WFMs. Those models are by no means a complete system or a production model for any real-world applications. The developer would need to fine-tune the WFMs on their custom datasets for their Physical AI setups for their target applications.

\subsection{Post-training WFM for Camera Control}\label{sec::camera}

Through camera pose conditioning, we integrate camera control into Cosmos-Predict1-7B-Video2World, making it an effective 3D world simulator. We term the result post-trained WFM as Cosmos-Predict1-7B-Video2World-Sample-CameraCond. We focus on generating 3D worlds from a single reference input image, leveraging camera control to produce temporally coherent and 3D-consistent video simulations from the specified camera trajectories, where changes in perspective align with the underlying 3D structure of the scene.

\subsubsection{Dataset}

We use DL3DV-10K~\citep{ling2024dl3dv}, a large-scale video dataset of static scenes, for this task. As a preprocessing step, we chunk all videos into clips with 256 frames. To obtain camera pose annotations densely for all frames within a clip, we run structure-from-motion on the chunked clips using GLOMAP~\citep{pan2025global}. We set the camera pose of the first frame to be the identity transform and compute the relative camera poses for all subsequent frames. We also use a proprietary VLM to caption the videos to obtain text prompts that describe the videos as static scenes.

\subsubsection{Fine-tuning}

We add camera control conditioning by concatenating the sampled latent embeddings with Pl\"ucker embeddings~\citep{sitzmann2021light}, which has the same spatial dimensions as the latent embeddings. Specifically, given the camera pose, we compute the Pl\"ucker coordinates via
\begin{align}
    \label{eq:plucker}
    \mathbf{r} = (\mathbf{d}, \mathbf{m}) \in \mathbb{R}^6 \;\;\; \text{where} \;\; \mathbf{m} = \mathbf{c} \times \mathbf{d} \;,
\end{align}
where $\mathbf{c}$ is the camera center location and $\mathbf{d}$ is the unit ray direction of each latent pixel (where the latent embedding is treated as a downsampled image). All the camera poses are relative with respect to the initial frame. The Cosmos-Tokenize1-CV8$\times$8$\times$8-720p used by Cosmos-Predict1-7B-Video2World models has a temporal compression rate of $8\times$, and thus for every 8 frames, we use the Pl\"ucker embedding at the 4th frame to concatenate with the corresponding latent representation.

We resized the input frames of our training videos to $704\times1252$ and padded them to $704\times1280$ with reflection. We sample 57 frames during training. The training objective and other hyper-parameters are the same as the base Diffusion WFM training (\cref{sec::diffusion_model_training}).

\subsubsection{Evaluation}

We assume a single reference image of the world is given and generate the future rollout as a video from the input image. We compare against CamCo~\citep{xu2024camco}, the state-of-the-art model for camera-controllable video generation under this setup. For a fair comparison, we use the CamCo model that was also fine-tuned on the DL3DV-10K~\citep{ling2024dl3dv} training set. As our post-trained WFM generates 57 frames and CamCo can only generate 14 frames, we compare the same 57-frame trajectories where we temporally downsample by $4\times$ for CamCo. The video resolution from CamCo is limited to $256\times256$. We additionally maximally center-crop the input image and test frames for evaluation.

For the test data, we use the same 500 samples from the RealEstate10K~\citep{zhou2018stereo} test set previously described in~\cref{sec::eval_3d_consistency}. We use the initial frame as the reference image and camera trajectories provided by the dataset as the camera control input, which we additionally rescale such that the distance between two ends of trajectories is normalized to 1.

\noindent \textbf{Metrics.} Following~\citet{xu2024camco}, we evaluate the camera controllability of the post-trained world model in two aspects: video generation quality and 3D consistency. For video quality, we use the Fr\'{e}chet Inception Distance (FID)~\citep{FID} and the Fr\'{e}chet Video Distance (FVD)~\citep{FVD} to assess the qualities at the frame and video levels, respectively. We use the same test data as the reference videos to compute the metrics (note that they are not used for pixel-level comparisons).

For 3D consistency, we evaluate via the ability of structure-from-motion~\citep{schoenberger2016sfm,schonberger2016pixelwise,pan2025global} libraries to re-estimate the camera poses, and we compare the results against the input camera control trajectories. Given $N$ frames in the video, we quantify the camera trajectory error into two terms: the average rotation error $\epsilon_\text{rot}$ and translation error $\epsilon_\text{trans}$, defined respectively as
\begin{align}
    \label{eq:pose-error}
    \epsilon_\text{rot} = \frac{1}{N} \sum_{i=1}^N \cos^{-1}\!\left( \frac{\mathrm{trace}( \mathbf{\hat{R}}_i^\top \mathbf{R}_i ) - 1}{2} \right)
    \qquad \text{and} \qquad
    \epsilon_\text{trans} = \frac{1}{N} \sum_{i=1}^N \left\| \mathbf{\hat{t}}_i - \mathbf{t}_i \right\|_2 \;,
\end{align}
where $\mathbf{R}_i$ and $\mathbf{t}_i$ are the input rotation and translation of the $i$-th frame (serving as ground truth), and $\mathbf{\hat{R}}_i$ and $\mathbf{\hat{t}}_i$ are the re-estimated quantities. To account for ambiguities from camera pose estimation results up to a similarity transformation, we follow~\citet{lin2021barf} and run Procrustes analysis on the predicted camera trajectories to align against the ground truth.

\begin{figure}[!htp]
    \centering
    \includegraphics[width=\textwidth]{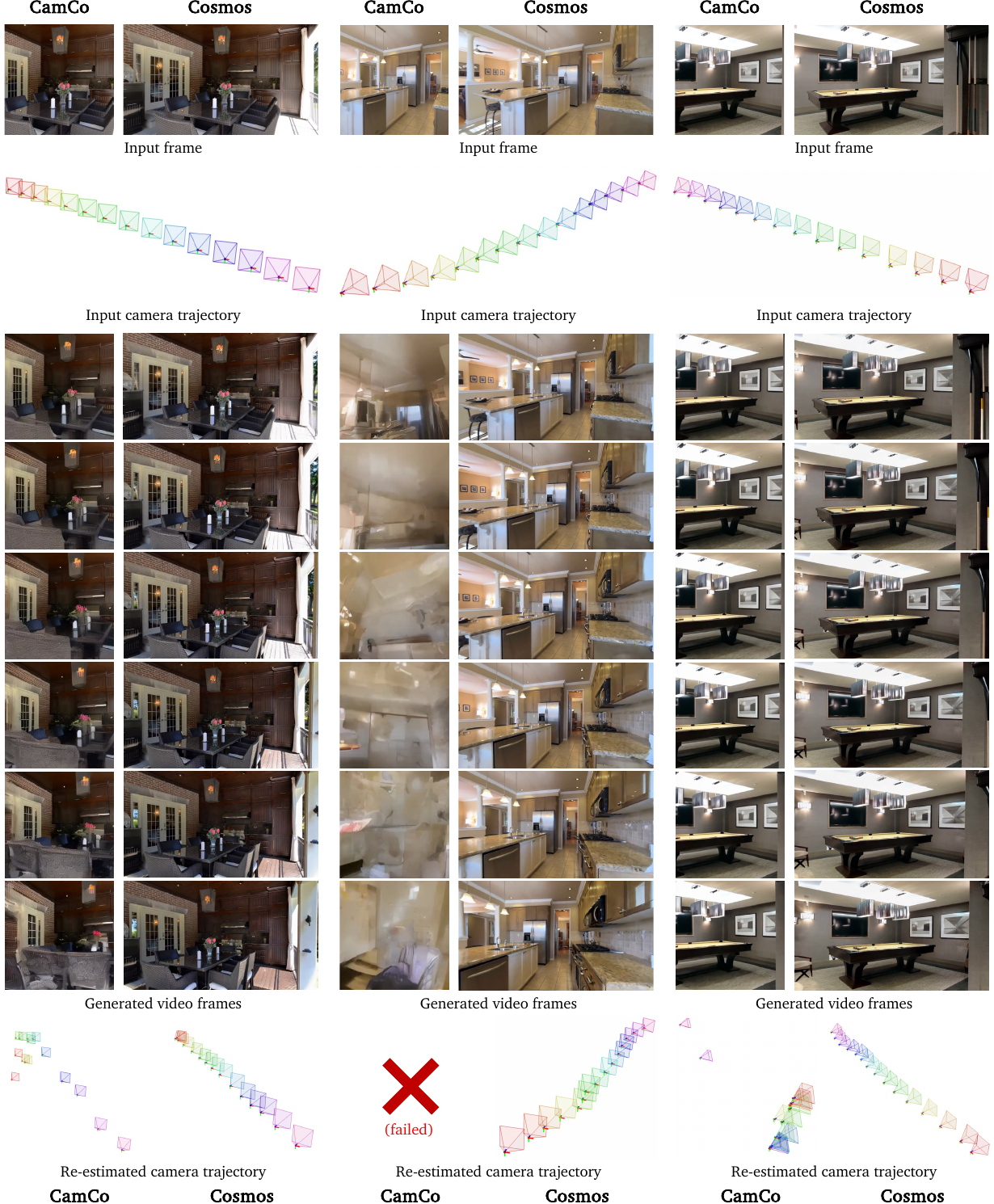}
    \caption{
        \textbf{Qualitative comparison of camera control models}. Given the input frame and camera trajectory (color-coded temporally from red to violet), we compare Cosmos-Predict1-7B-Video2World-Sample-CameraCond against CamCo~\citep{xu2024camco} on the generated future frames as well as the re-estimated camera poses. CamCo suffers from the data distribution shift and often generates inaccurate trajectories or even out-of-distribution image syntheses that lead to un-estimatable camera poses. In contrast, the Cosmos camera control model can successfully generate future frames aligned with the camera control input while also maintaining high video quality and 3D consistency.
    }
    \label{fig:camera-main}
\end{figure}

\begin{table}[h]
    \setlength{\tabcolsep}{4.7pt}
    \small
    \captionsetup{justification=centering}
    \caption{
        Quantitative comparison of post-trained WFM with camera control.
    }
    \centering
    \vspace{-1em}
    \setlength{\tabcolsep}{6pt}
    \begin{tabular}{rccccc}
        \toprule
        & \multicolumn{3}{c}{\bf Camera Trajectory Alignment} & \multicolumn{2}{c}{\bf Video Generation Quality} \\
        \cmidrule(r){2-4} \cmidrule(r){5-6}
        & Pose estimation & Rotation & Translation & \multirow{2}{*}{~~~~ FID $\downarrow$} & \multirow{2}{*}{FVD $\downarrow$} \\
        Method & success rate (\%) $\uparrow$ & error ($\degree$) $\downarrow$ &  error $\downarrow$ & & \\
        \midrule
        CamCo~\citep{xu2024camco} & 43.0\% & 8.277 & 0.185 & ~~~~ 57.49 & 433.24 \vspace{2pt} \\
        \diffusionsevenbvideotoworld- & \multirow{2}{*}{\bf 82.0\%} & \multirow{2}{*}{\bf 1.646} & \multirow{2}{*}{\bf 0.038} & \multirow{2}{*}{~~~~ \bf 14.30} & \multirow{2}{*}{\bf 120.49} \\
        Sample-CameraCond & & & & & \\
        \bottomrule
    \end{tabular}
    \label{tab:camera-results}
\end{table}

\noindent \textbf{Comparisons.} We present the results in~\cref{tab:camera-results}. First, our post-trained WFM can generate realistic and coherent 3D worlds. This is evidenced by the lower FID/FVD scores (higher visual quality) and the higher camera pose estimation success rate. Cosmos-Predict1-7B-Video2World-Sample-CameraCond demonstrates better camera control, as the camera trajectory re-estimation is significantly closer to the original control input.

We also provide visual comparisons in~\cref{fig:camera-main}. While CamCo struggles to generate content beyond the input image, Cosmos-Predict1-7B-Video2World-Sample-CameraCond effectively generates visuals that adhere to the structure of a 3D world. Note that both models were post-trained on DL3DV-10K and evaluated on the RealEstate10K dataset, which introduces a significant distribution shift between training and testing. The Cosmos model successfully overcomes this distribution shift while also demonstrating its capability to generalize to unseen input camera trajectories.

\begin{figure*}[!htp]
    \centering
    \begin{subfigure}[c]{0.18\textwidth}
        \centering
        \includegraphics[width=\textwidth]{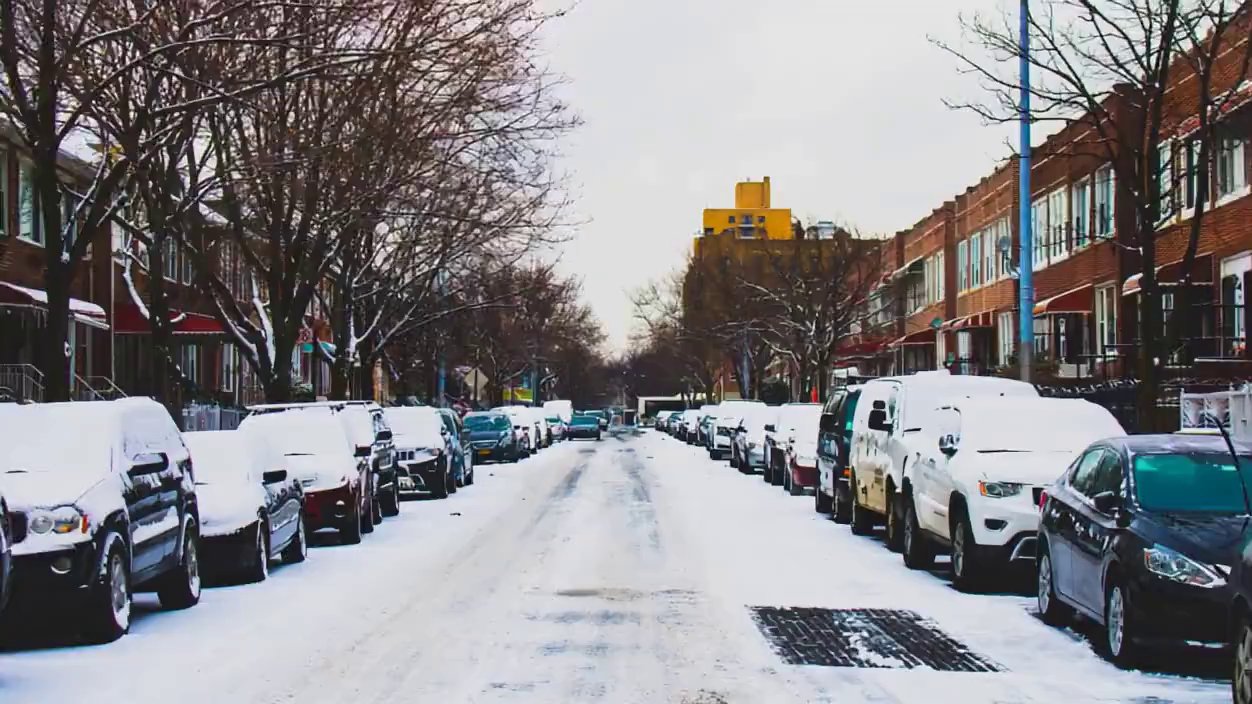}
    \end{subfigure}
    \hfill
    \begin{subfigure}[c]{0.04\textwidth}
        \centering
        \includegraphics[width=\textwidth]{images/6_post_training/3d_joystick_wasd/arrows/arrow_up.pdf}
    \end{subfigure}
    \hfill
    \begin{subfigure}[c]{0.72\textwidth}
        \centering
        \includegraphics[width=\textwidth]{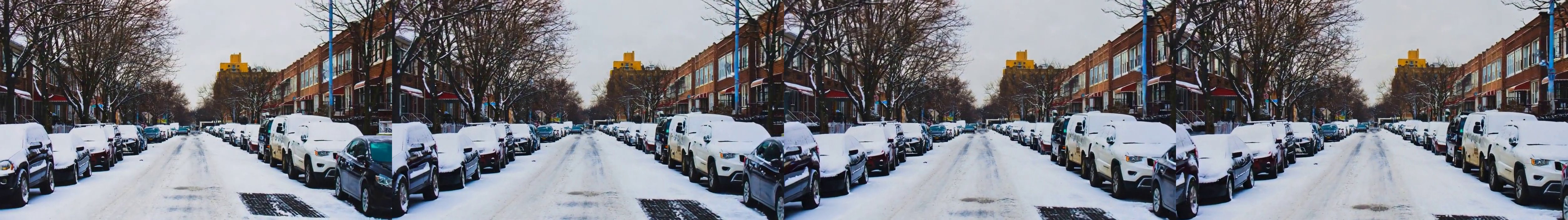}
    \end{subfigure}
    \\
    \begin{subfigure}[c]{0.18\textwidth}
        \centering
        \includegraphics[width=\textwidth]{images/6_post_training/3d_joystick_wasd/0002-0_0.jpg}
    \end{subfigure}
    \hfill
    \begin{subfigure}[c]{0.04\textwidth}
        \centering
        \includegraphics[width=\textwidth]{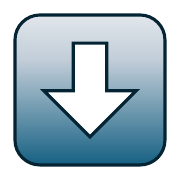}
    \end{subfigure}
    \hfill
    \begin{subfigure}[c]{0.72\textwidth}
        \centering
        \includegraphics[width=\textwidth]{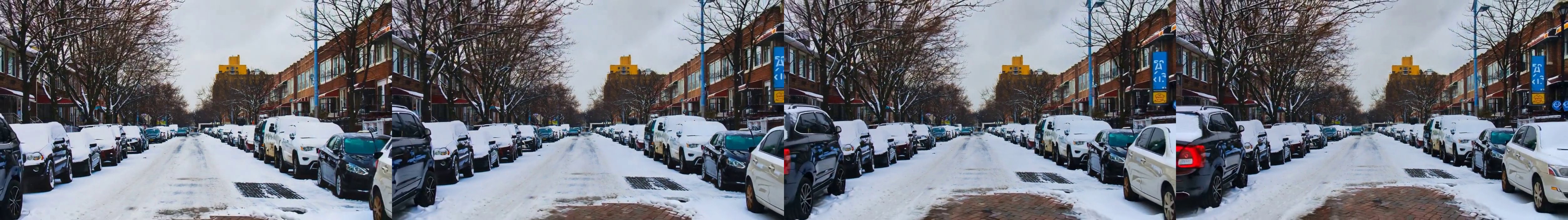}
    \end{subfigure}
    \\
    \begin{subfigure}[c]{0.18\textwidth}
        \centering
        \includegraphics[width=\textwidth]{images/6_post_training/3d_joystick_wasd/0002-0_0.jpg}
    \end{subfigure}
    \hfill
    \begin{subfigure}[c]{0.04\textwidth}
        \centering
        \includegraphics[width=\textwidth]{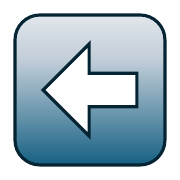}
    \end{subfigure}
    \hfill
    \begin{subfigure}[c]{0.72\textwidth}
        \centering
        \includegraphics[width=\textwidth]{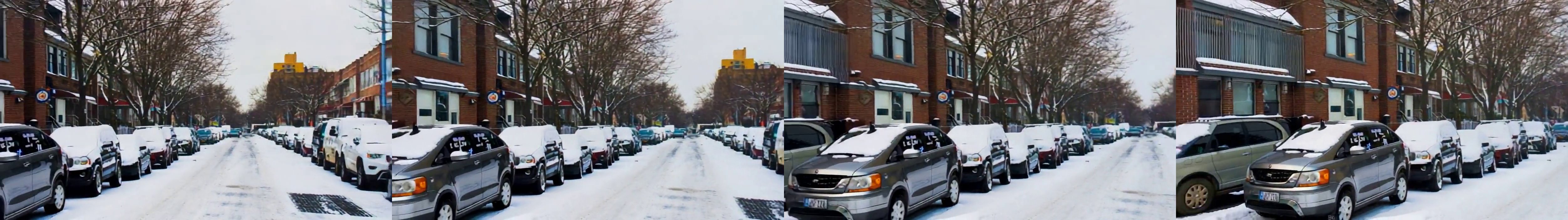}
    \end{subfigure}
    \\
    \begin{subfigure}[c]{0.18\textwidth}
        \centering
        \includegraphics[width=\textwidth]{images/6_post_training/3d_joystick_wasd/0002-0_0.jpg}
    \end{subfigure}
    \hfill
    \begin{subfigure}[c]{0.04\textwidth}
        \centering
        \includegraphics[width=\textwidth]{images/6_post_training/3d_joystick_wasd/arrows/arrow_right.pdf}
    \end{subfigure}
    \hfill
    \begin{subfigure}[c]{0.72\textwidth}
        \centering
        \includegraphics[width=\textwidth]{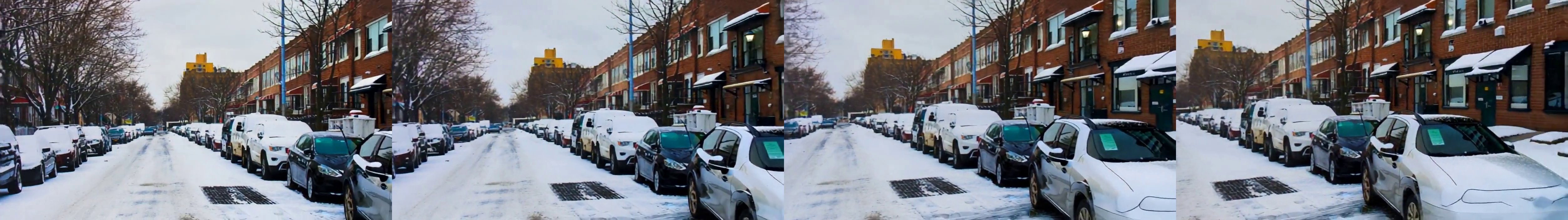}
    \end{subfigure}
    \\
    \begin{subfigure}[c]{0.18\textwidth}
        \centering
        \scriptsize{Input frame}
    \end{subfigure}
    \hfill
    \begin{subfigure}[c]{0.04\textwidth}
        \centering
        \scriptsize{Control}
    \end{subfigure}
    \hfill
    \begin{subfigure}[c]{0.72\textwidth}
        \centering
        \scriptsize{Generated video frames}
    \end{subfigure}
    \\
    \vfill
    \begin{subfigure}[c]{0.18\textwidth}
        \centering
        \includegraphics[width=\textwidth]{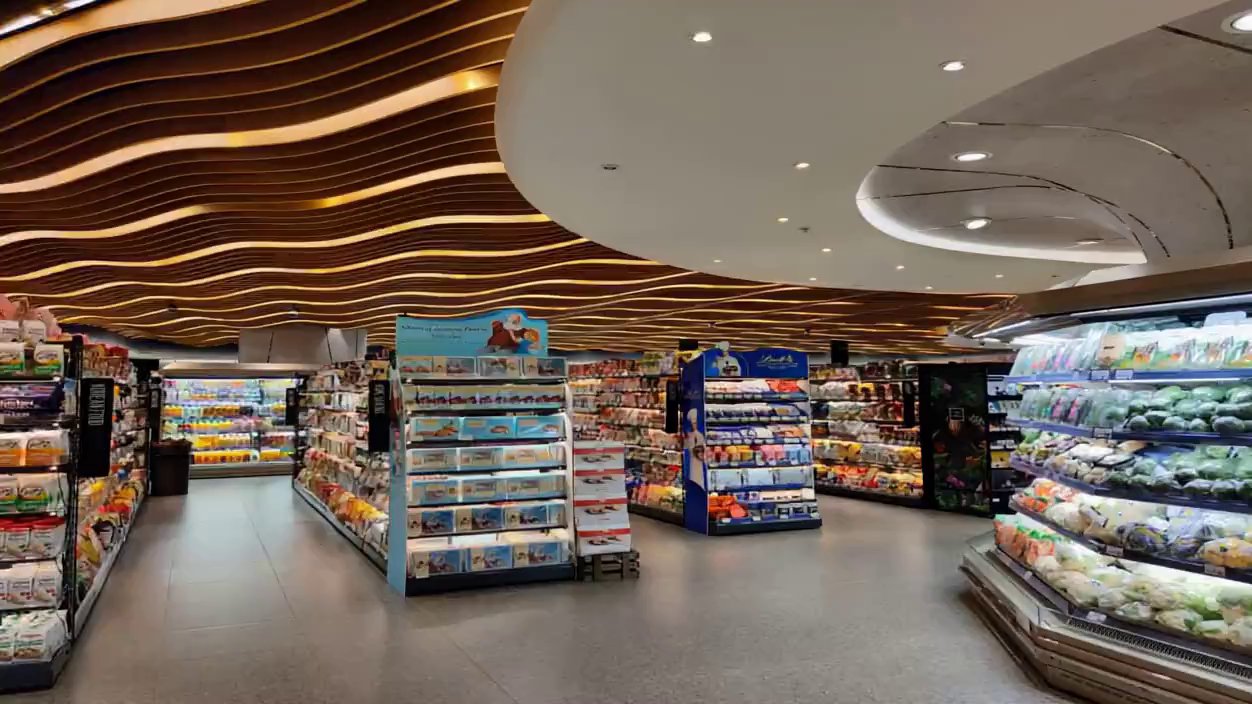}
    \end{subfigure}
    \hfill
    \begin{subfigure}[c]{0.04\textwidth}
        \centering
        \includegraphics[width=\textwidth]{images/6_post_training/3d_joystick_wasd/arrows/arrow_up.pdf}
    \end{subfigure}
    \hfill
    \begin{subfigure}[c]{0.72\textwidth}
        \centering
        \includegraphics[width=\textwidth]{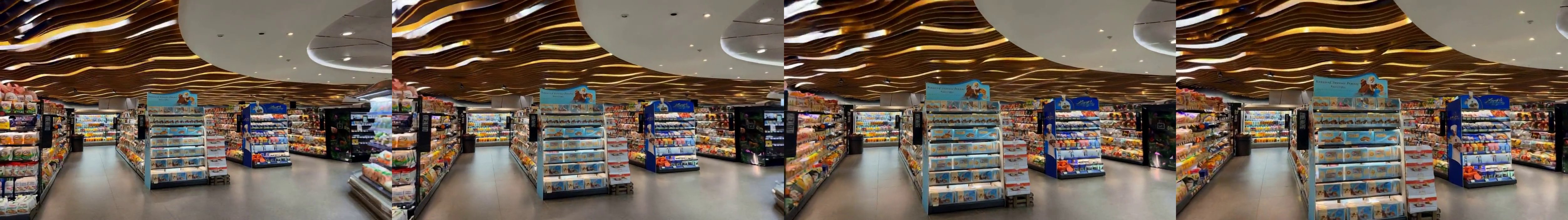}
    \end{subfigure}
    \\
    \begin{subfigure}[c]{0.18\textwidth}
        \centering
        \includegraphics[width=\textwidth]{images/6_post_training/3d_joystick_wasd/0014-0_0.jpg}
    \end{subfigure}
    \hfill
    \begin{subfigure}[c]{0.04\textwidth}
        \centering
        \includegraphics[width=\textwidth]{images/6_post_training/3d_joystick_wasd/arrows/arrow_down.pdf}
    \end{subfigure}
    \hfill
    \begin{subfigure}[c]{0.72\textwidth}
        \centering
        \includegraphics[width=\textwidth]{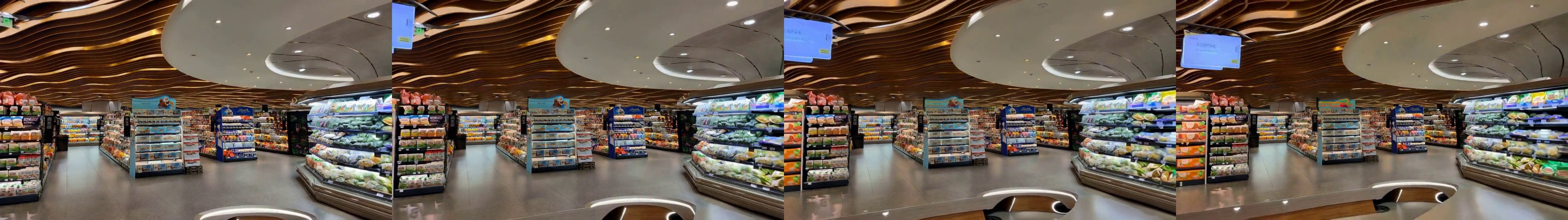}
    \end{subfigure}
    \\
    \begin{subfigure}[c]{0.18\textwidth}
        \centering
        \includegraphics[width=\textwidth]{images/6_post_training/3d_joystick_wasd/0014-0_0.jpg}
    \end{subfigure}
    \hfill
    \begin{subfigure}[c]{0.04\textwidth}
        \centering
        \includegraphics[width=\textwidth]{images/6_post_training/3d_joystick_wasd/arrows/arrow_left.pdf}
    \end{subfigure}
    \hfill
    \begin{subfigure}[c]{0.72\textwidth}
        \centering
        \includegraphics[width=\textwidth]{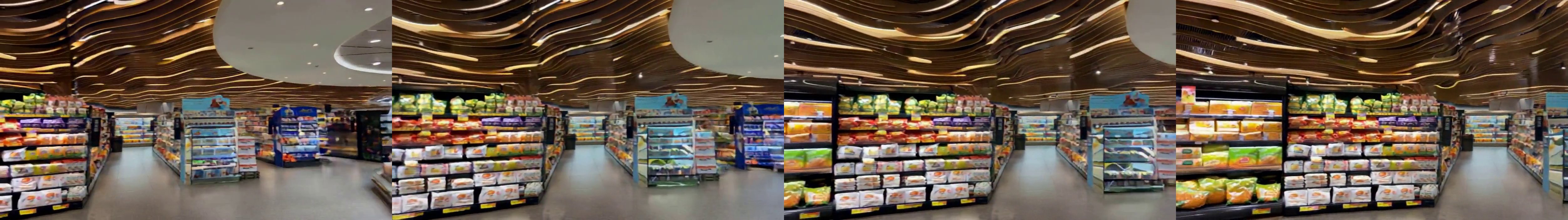}
    \end{subfigure}
    \\
    \begin{subfigure}[c]{0.18\textwidth}
        \centering
        \includegraphics[width=\textwidth]{images/6_post_training/3d_joystick_wasd/0014-0_0.jpg}
    \end{subfigure}
    \hfill
    \begin{subfigure}[c]{0.04\textwidth}
        \centering
        \includegraphics[width=\textwidth]{images/6_post_training/3d_joystick_wasd/arrows/arrow_right.pdf}
    \end{subfigure}
    \hfill
    \begin{subfigure}[c]{0.72\textwidth}
        \centering
        \includegraphics[width=\textwidth]{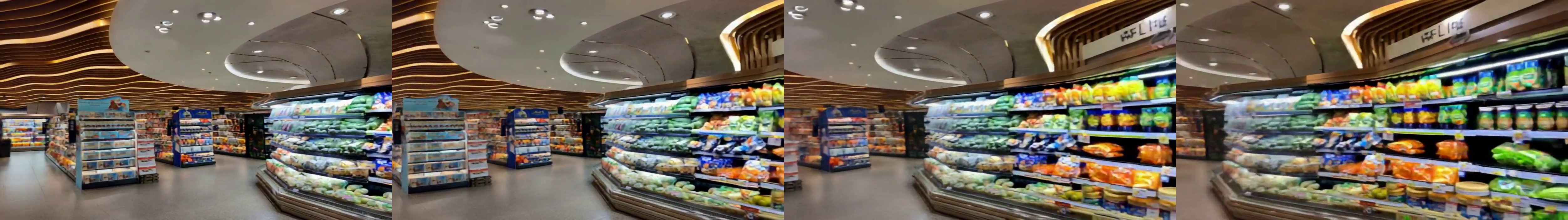}
    \end{subfigure}
    \\
    \begin{subfigure}[c]{0.18\textwidth}
        \centering
        \scriptsize{Input frame}
    \end{subfigure}
    \hfill
    \begin{subfigure}[c]{0.04\textwidth}
        \centering
        \scriptsize{Control}
    \end{subfigure}
    \hfill
    \begin{subfigure}[c]{0.72\textwidth}
        \centering
        \scriptsize{Generated video frames}
    \end{subfigure}
    \\
    \vfill
    \begin{subfigure}[c]{0.18\textwidth}
        \centering
        \includegraphics[width=\textwidth]{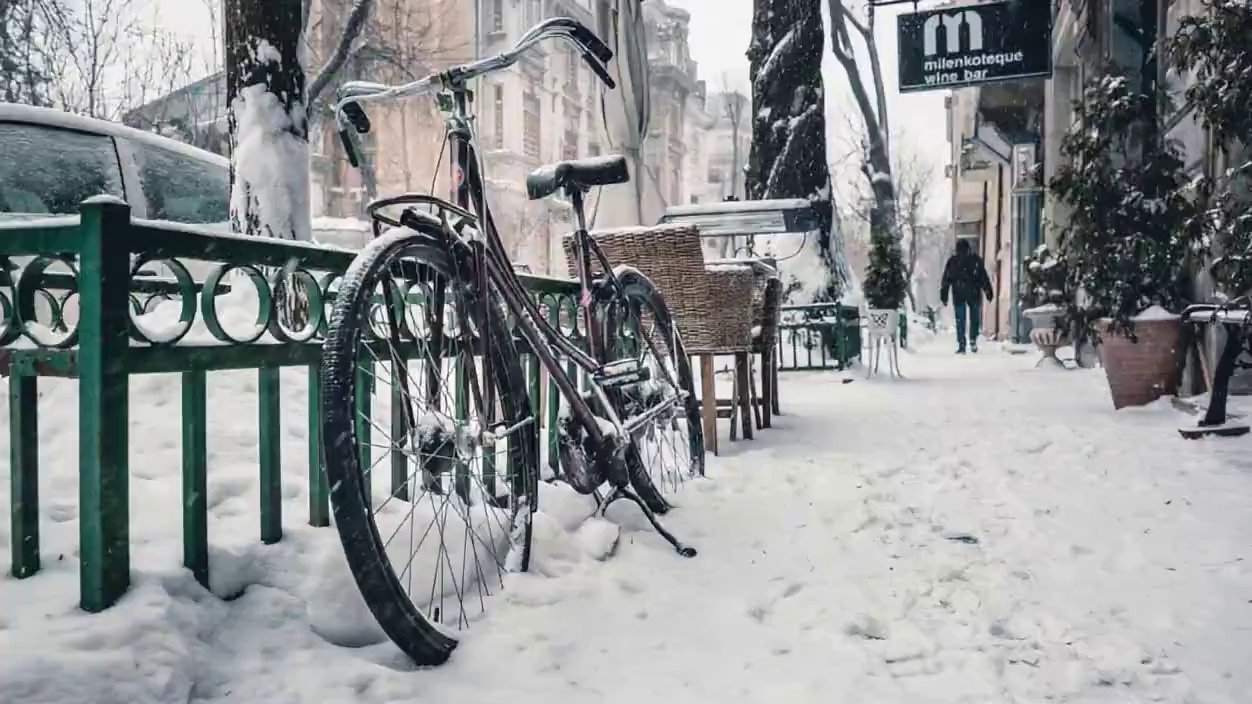}
    \end{subfigure}
    \hfill
    \begin{subfigure}[c]{0.04\textwidth}
        \centering
        \includegraphics[width=\textwidth]{images/6_post_training/3d_joystick_wasd/arrows/arrow_up.pdf}
    \end{subfigure}
    \hfill
    \begin{subfigure}[c]{0.72\textwidth}
        \centering
        \includegraphics[width=\textwidth]{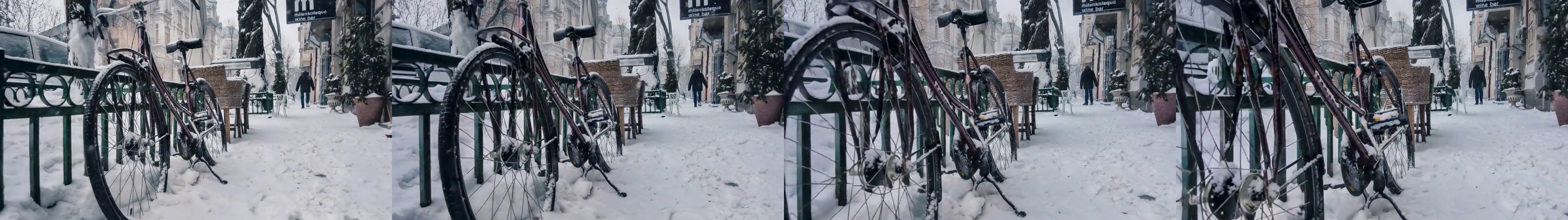}
    \end{subfigure}
    \\
    \begin{subfigure}[c]{0.18\textwidth}
        \centering
        \includegraphics[width=\textwidth]{images/6_post_training/3d_joystick_wasd/0001-0_0.jpg}
    \end{subfigure}
    \hfill
    \begin{subfigure}[c]{0.04\textwidth}
        \centering
        \includegraphics[width=\textwidth]{images/6_post_training/3d_joystick_wasd/arrows/arrow_down.pdf}
    \end{subfigure}
    \hfill
    \begin{subfigure}[c]{0.72\textwidth}
        \centering
        \includegraphics[width=\textwidth]{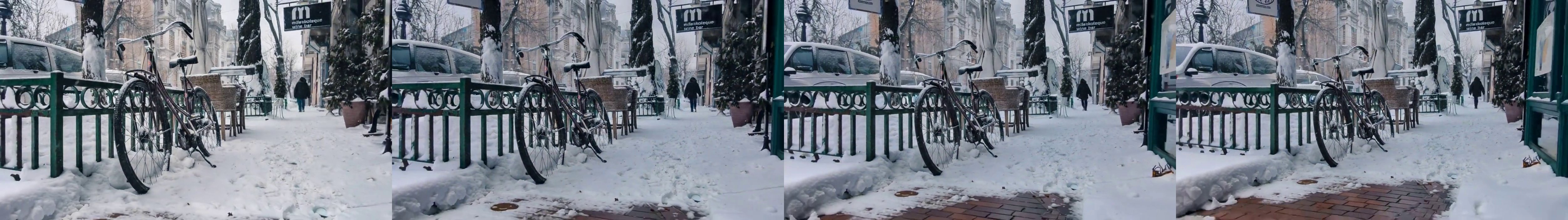}
    \end{subfigure}
    \\
    \begin{subfigure}[c]{0.18\textwidth}
        \centering
        \includegraphics[width=\textwidth]{images/6_post_training/3d_joystick_wasd/0001-0_0.jpg}
    \end{subfigure}
    \hfill
    \begin{subfigure}[c]{0.04\textwidth}
        \centering
        \includegraphics[width=\textwidth]{images/6_post_training/3d_joystick_wasd/arrows/arrow_left.pdf}
    \end{subfigure}
    \hfill
    \begin{subfigure}[c]{0.72\textwidth}
        \centering
        \includegraphics[width=\textwidth]{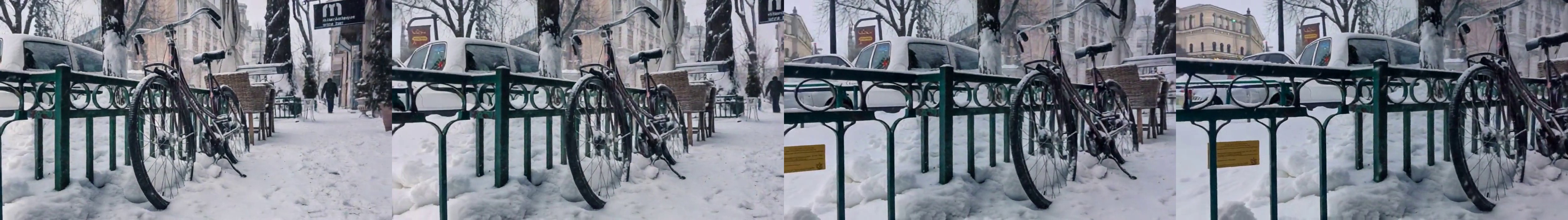}
    \end{subfigure}
    \\
    \begin{subfigure}[c]{0.18\textwidth}
        \centering
        \includegraphics[width=\textwidth]{images/6_post_training/3d_joystick_wasd/0001-0_0.jpg}
    \end{subfigure}
    \hfill
    \begin{subfigure}[c]{0.04\textwidth}
        \centering
        \includegraphics[width=\textwidth]{images/6_post_training/3d_joystick_wasd/arrows/arrow_right.pdf}
    \end{subfigure}
    \hfill
    \begin{subfigure}[c]{0.72\textwidth}
        \centering
        \includegraphics[width=\textwidth]{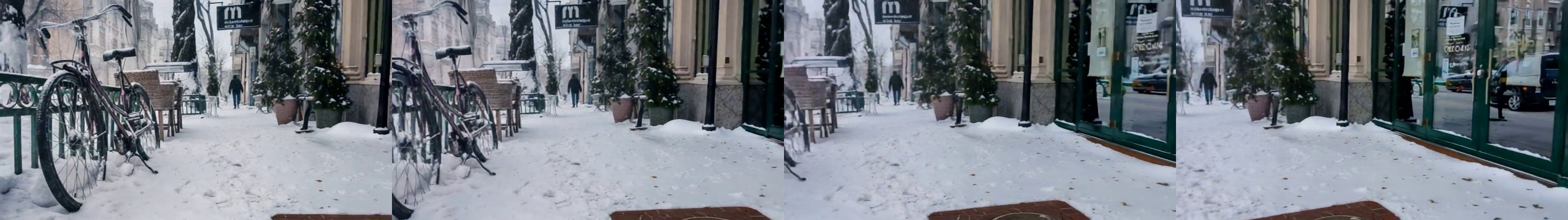}
    \end{subfigure}
    \\
    \begin{subfigure}[c]{0.18\textwidth}
        \centering
        \scriptsize{Input frame}
    \end{subfigure}
    \hfill
    \begin{subfigure}[c]{0.04\textwidth}
        \centering
        \scriptsize{Control}
    \end{subfigure}
    \hfill
    \begin{subfigure}[c]{0.72\textwidth}
        \centering
        \scriptsize{Generated video frames}
    \end{subfigure}

    \caption{
        \textbf{Cosmos-Predict1-7B-Video2World-Sample-CameraCond results with joystick control}. For each input frame (left-most column), we apply 4 different camera trajectories created with joystick-like control: \textit{moving forward}, \textit{moving backward}, \textit{rotating left}, and \textit{rotating right}. We visualize frames 14, 28, 42, and 57 from the generated videos.
    }
    \label{fig:camera-joystick-wasd}
\end{figure*}

\begin{figure*}[!htp]
    \centering
    \setlength{\tabcolsep}{1pt}
    \renewcommand{\arraystretch}{0.7}
    \begin{tabular}[t]{ccccccc}    %
        \includegraphics[width=0.16\textwidth]{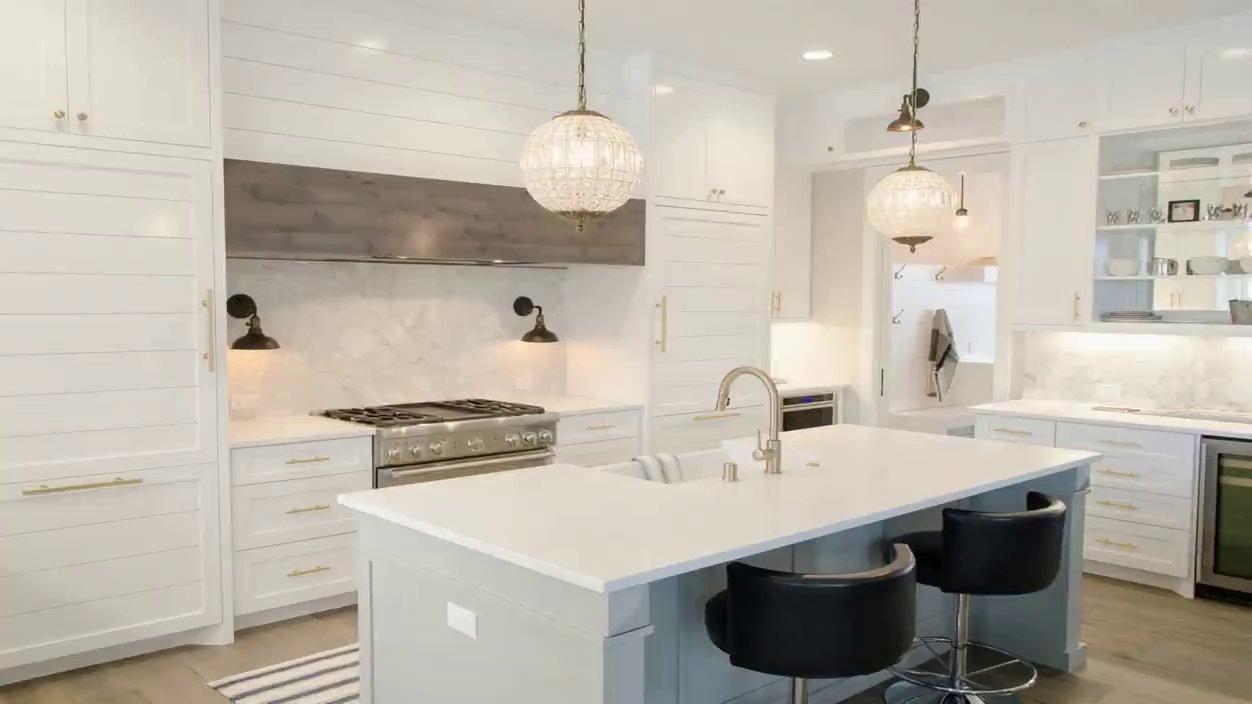} &
        \includegraphics[width=0.16\textwidth]{images/6_post_training/3d_joystick_seeds/0013-0_0.jpg} &
        \includegraphics[width=0.16\textwidth]{images/6_post_training/3d_joystick_seeds/0013-0_0.jpg} & &
        \includegraphics[width=0.16\textwidth]{images/6_post_training/3d_joystick_seeds/0013-0_0.jpg} &
        \includegraphics[width=0.16\textwidth]{images/6_post_training/3d_joystick_seeds/0013-0_0.jpg} &
        \includegraphics[width=0.16\textwidth]{images/6_post_training/3d_joystick_seeds/0013-0_0.jpg}
        \\
        \multicolumn{3}{c}{\scriptsize{Input frame}} & & \multicolumn{3}{c}{\scriptsize{Input frame}} \\
        \includegraphics[width=0.16\textwidth]{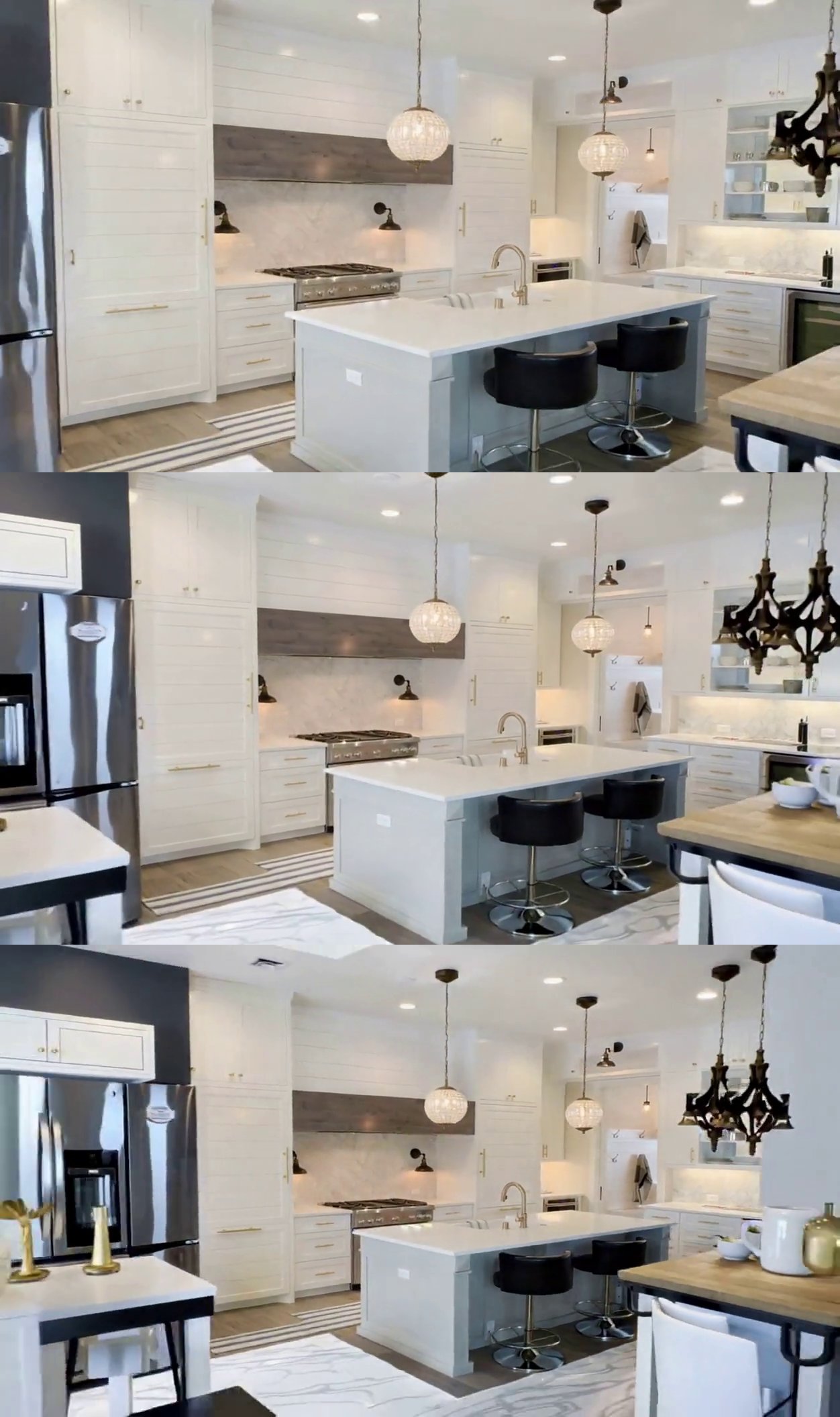} &
        \includegraphics[width=0.16\textwidth]{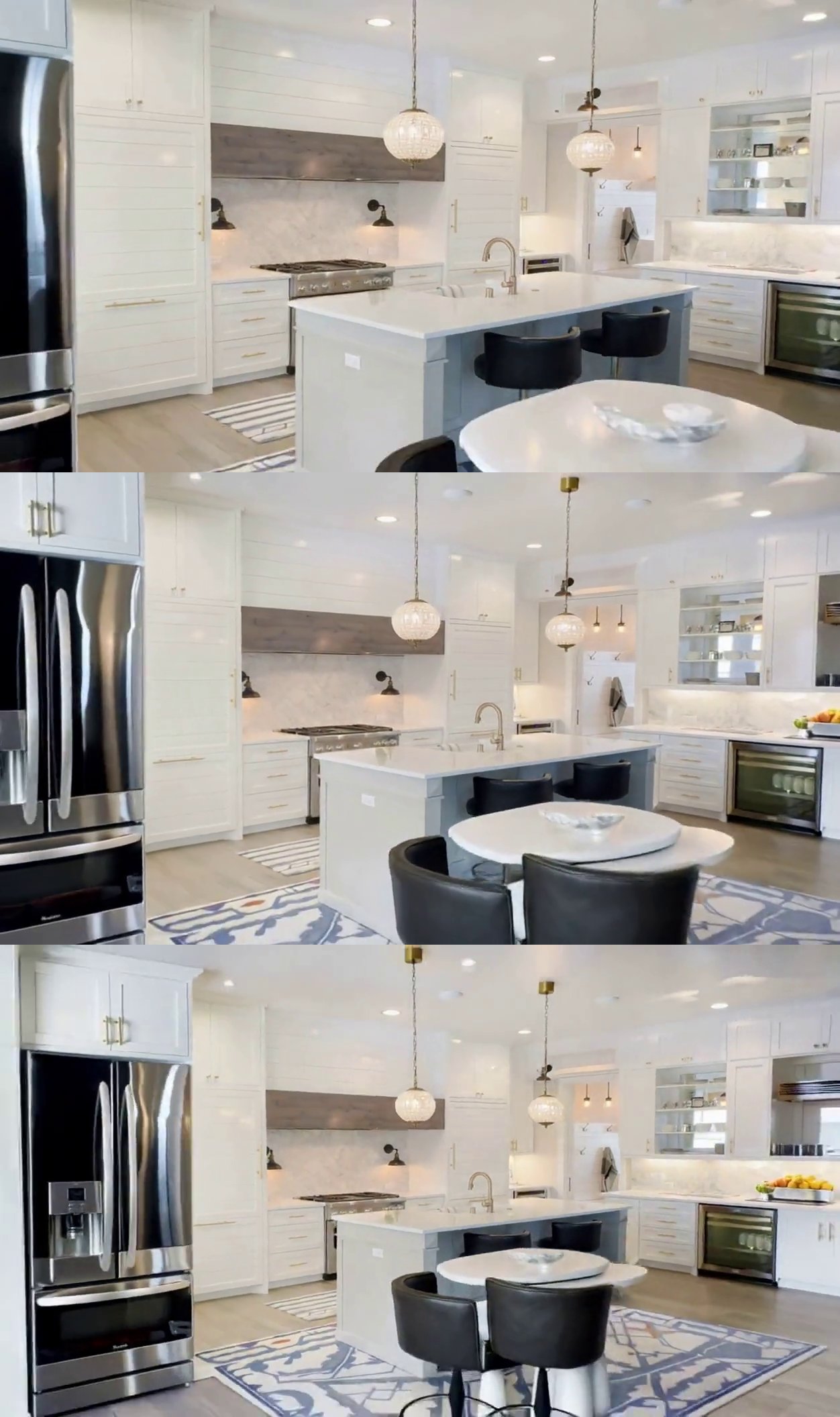} &
        \includegraphics[width=0.16\textwidth]{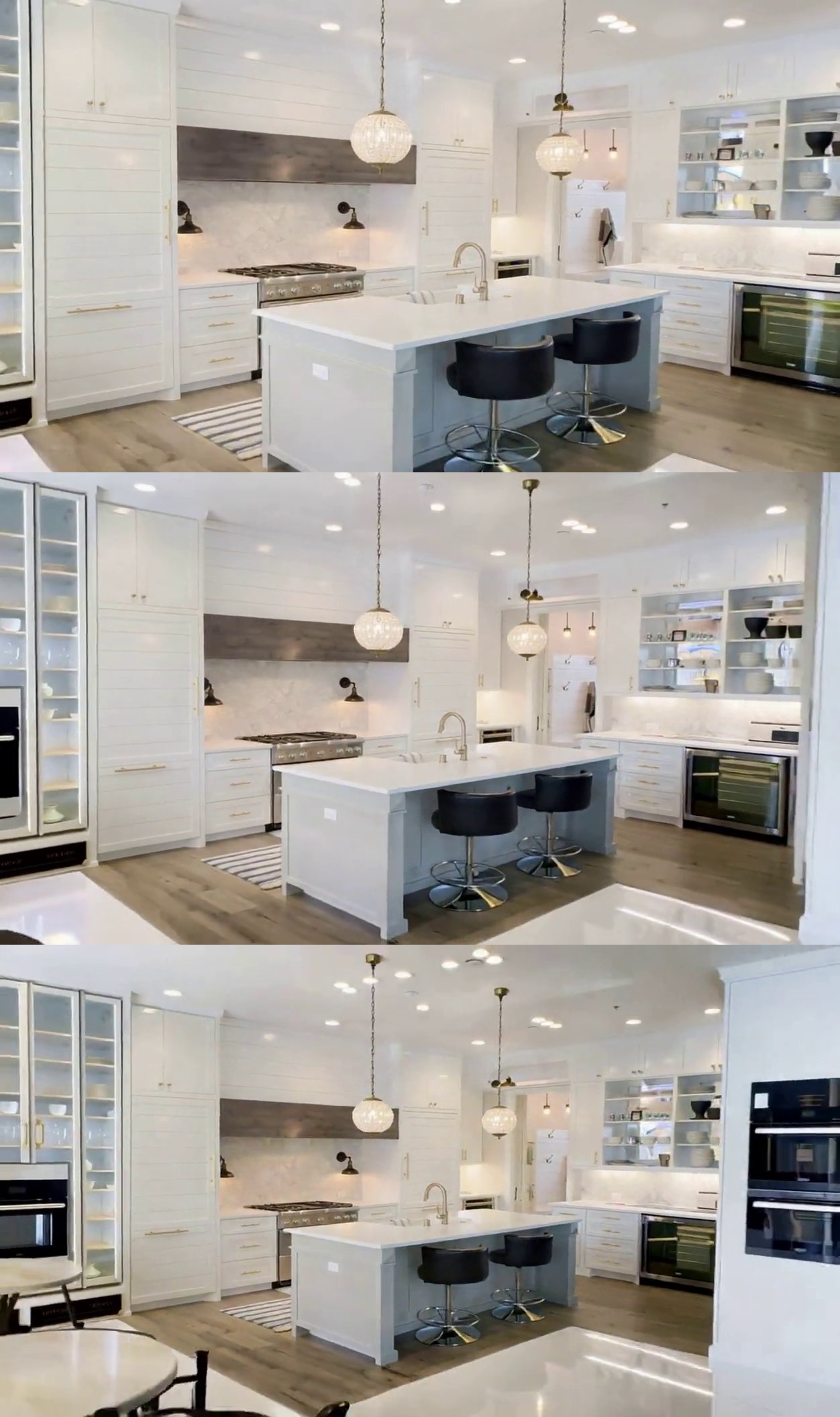} & &
        \includegraphics[width=0.16\textwidth]{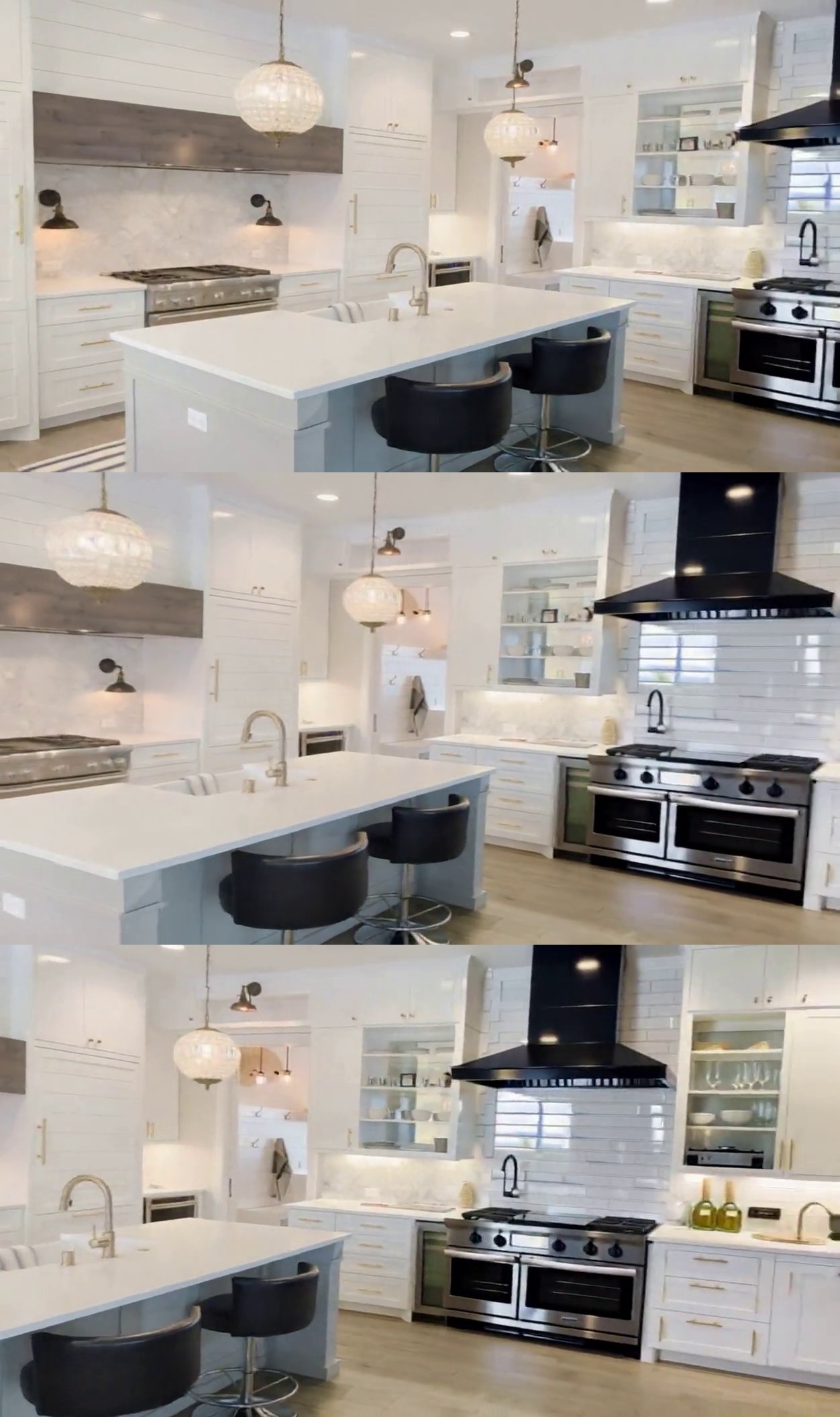} &
        \includegraphics[width=0.16\textwidth]{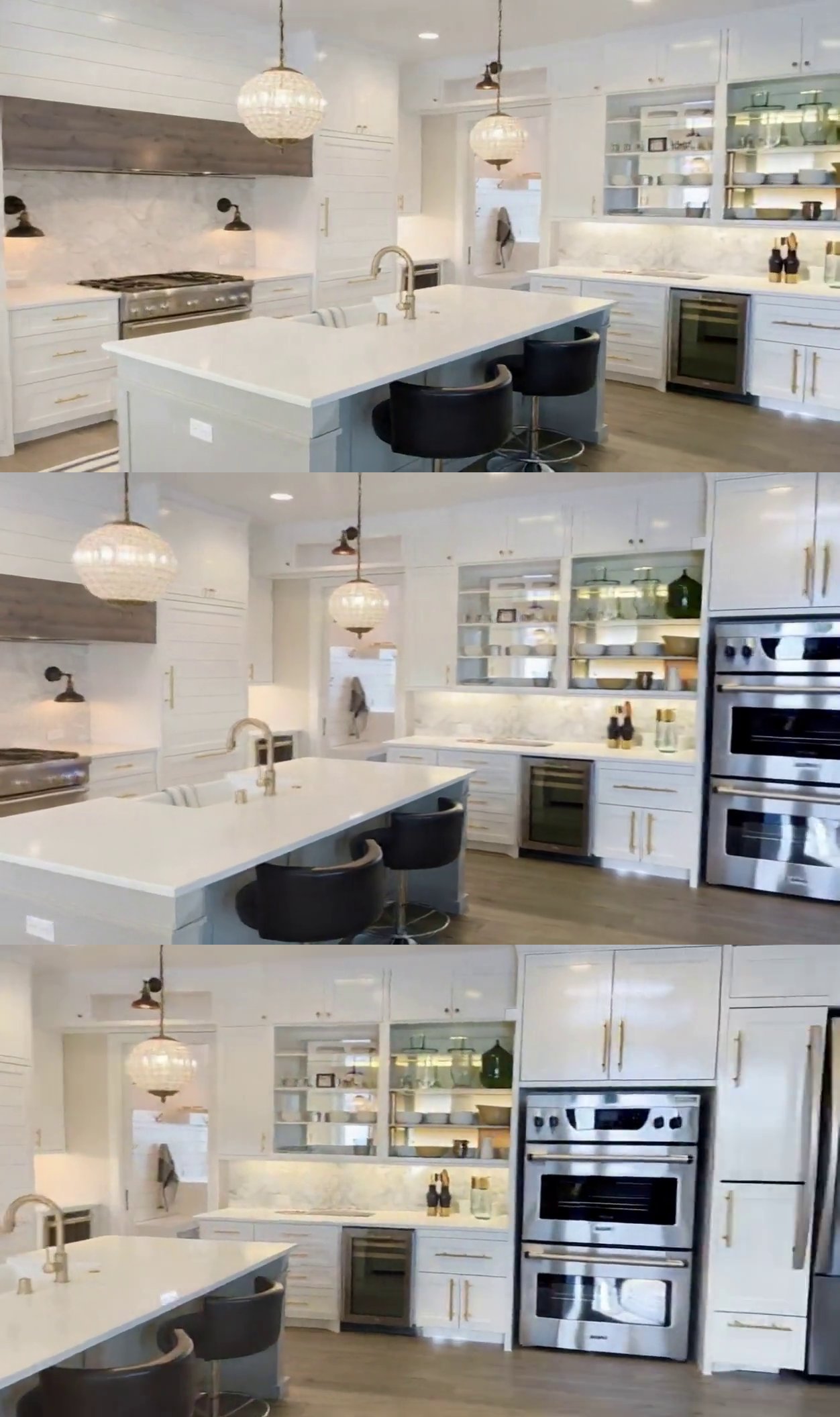} &
        \includegraphics[width=0.16\textwidth]{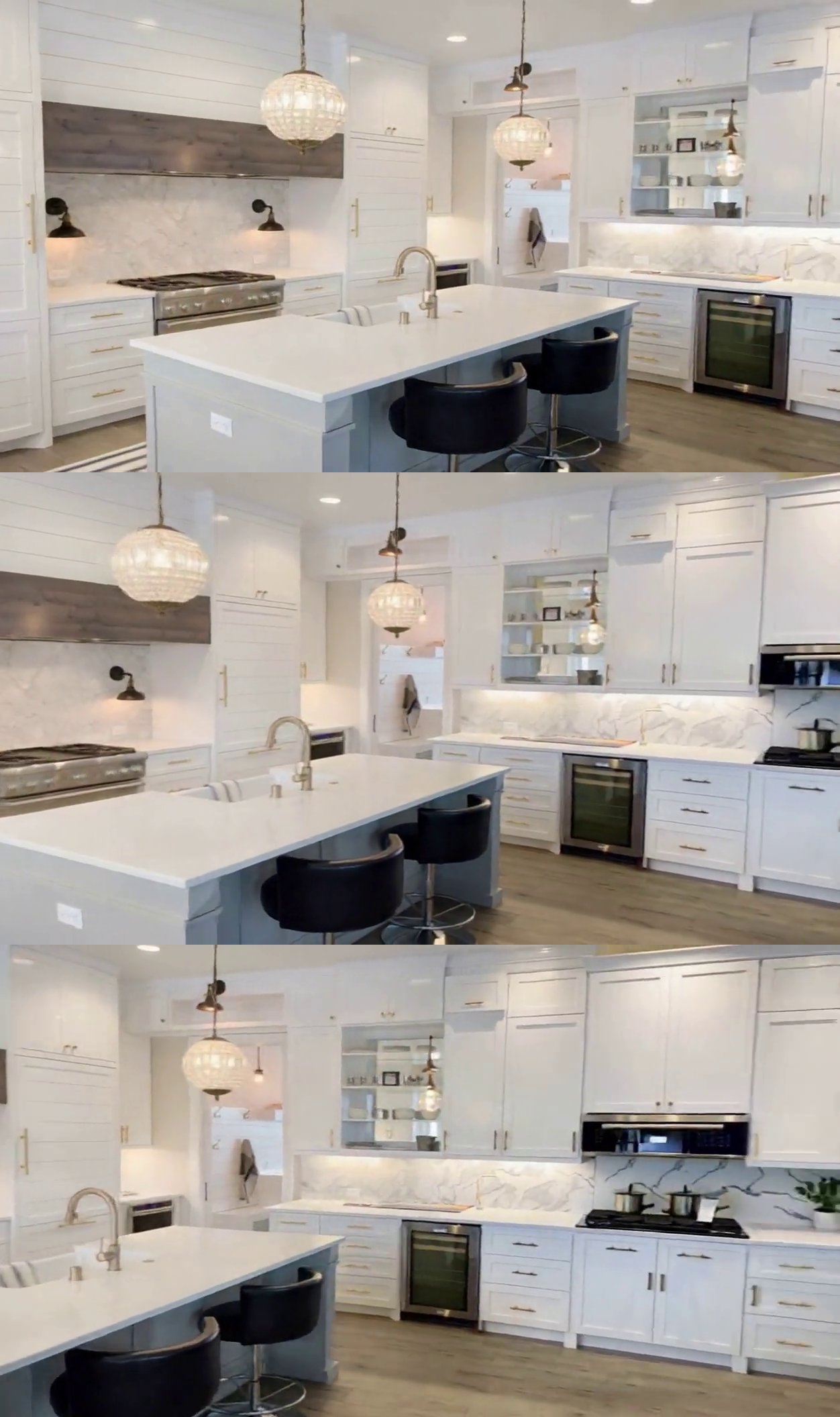}
        \\
        \multicolumn{3}{c}{\scriptsize{Generated frames with various seeds (moving backward)}} & & \multicolumn{3}{c}{\scriptsize{Generated frames with various seeds (rotating right)}} \\
        \arrayrulecolor{gray}\hline \\
        \includegraphics[width=0.16\textwidth]{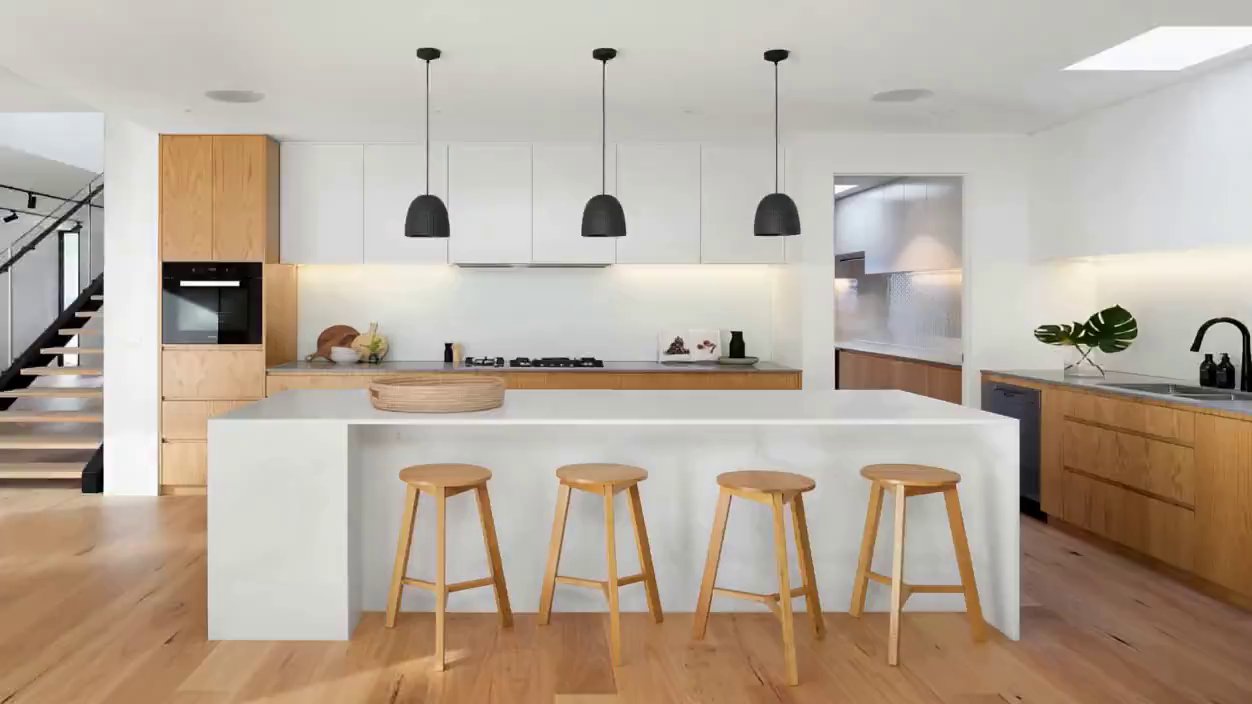} &
        \includegraphics[width=0.16\textwidth]{images/6_post_training/3d_joystick_seeds/0034-0_0.jpg} &
        \includegraphics[width=0.16\textwidth]{images/6_post_training/3d_joystick_seeds/0034-0_0.jpg} & &
        \includegraphics[width=0.16\textwidth]{images/6_post_training/3d_joystick_seeds/0034-0_0.jpg} &
        \includegraphics[width=0.16\textwidth]{images/6_post_training/3d_joystick_seeds/0034-0_0.jpg} &
        \includegraphics[width=0.16\textwidth]{images/6_post_training/3d_joystick_seeds/0034-0_0.jpg}
        \\
        \multicolumn{3}{c}{\scriptsize{Input frame}} & & \multicolumn{3}{c}{\scriptsize{Input frame}} \\
        \includegraphics[width=0.16\textwidth]{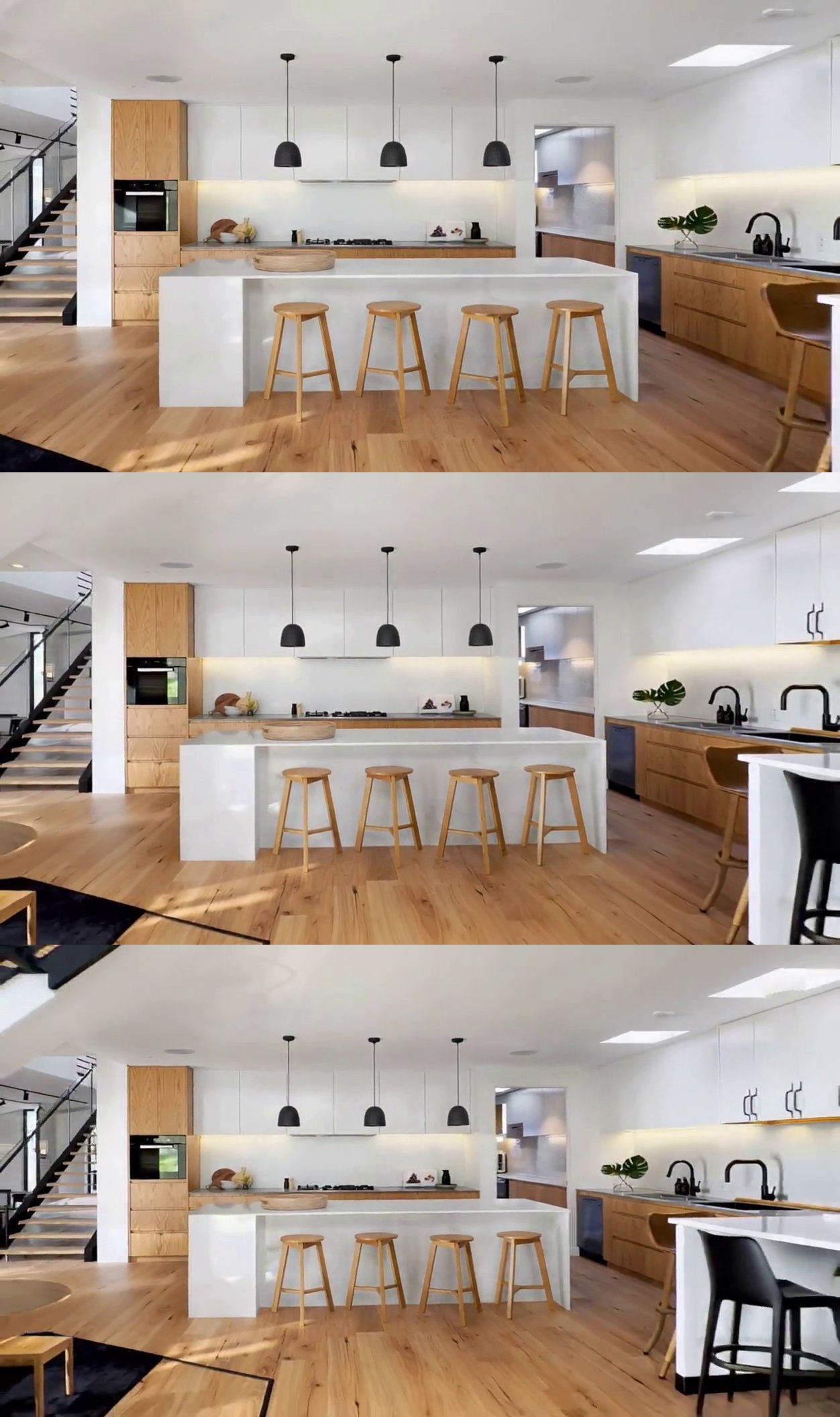} &
        \includegraphics[width=0.16\textwidth]{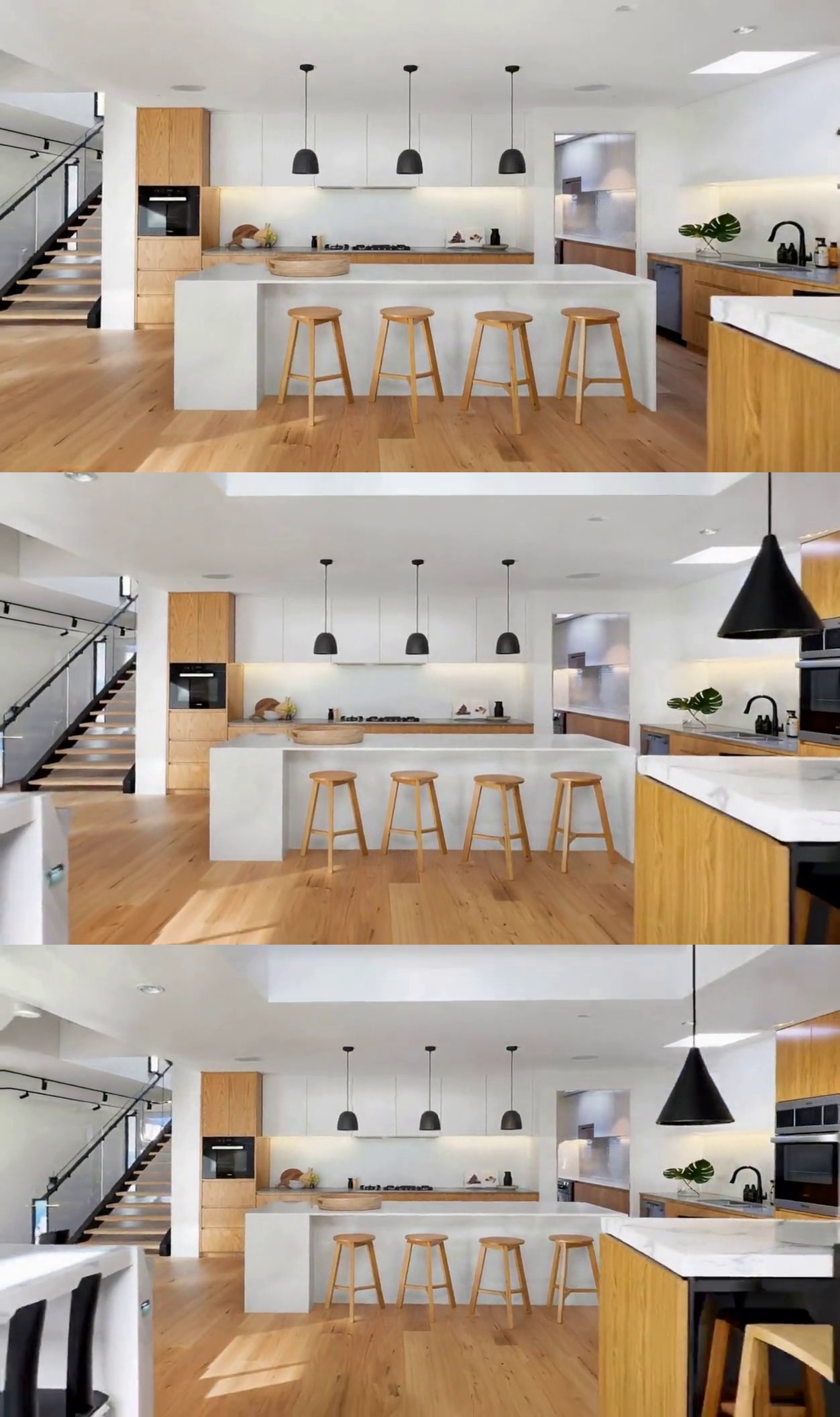} &
        \includegraphics[width=0.16\textwidth]{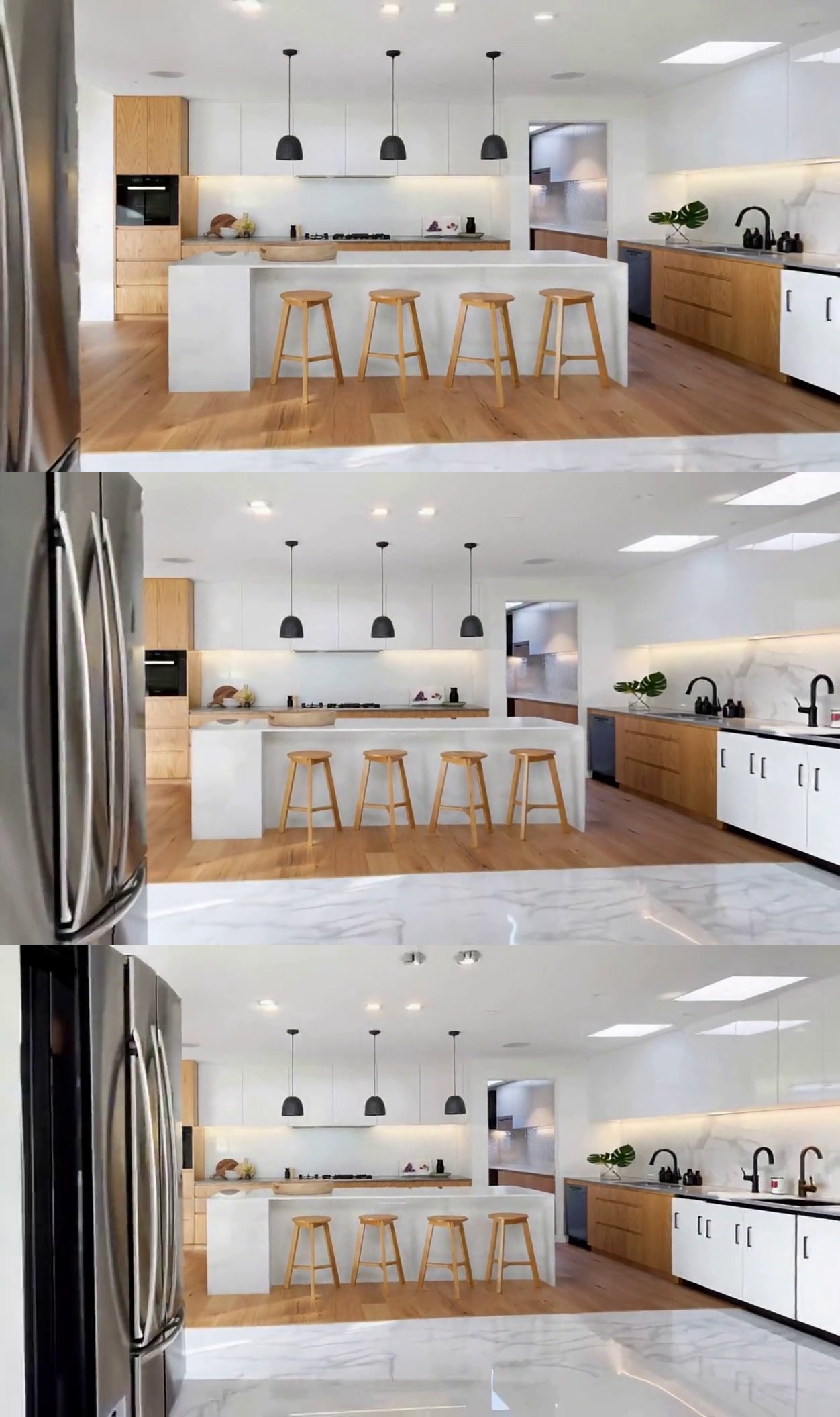} & &
        \includegraphics[width=0.16\textwidth]{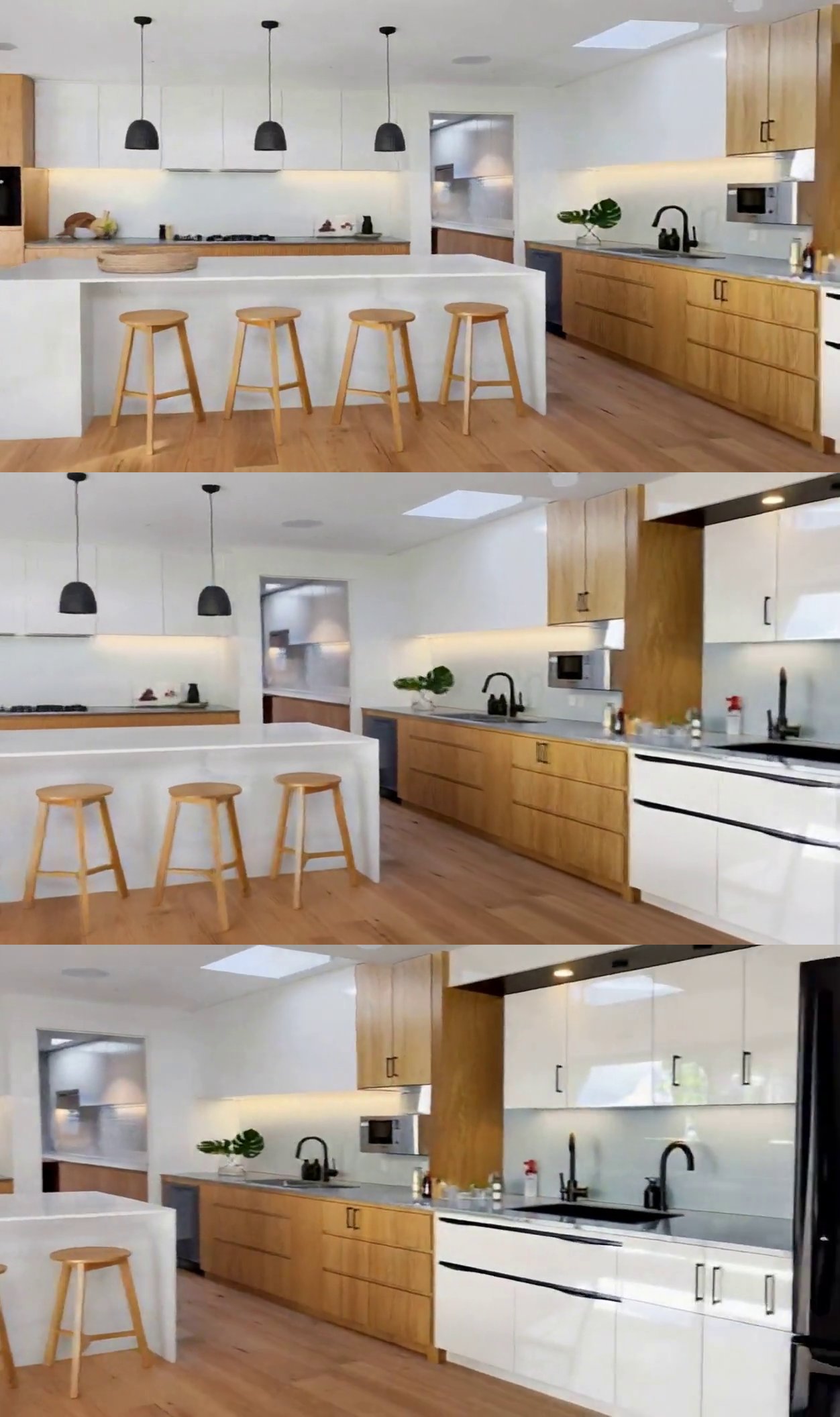} &
        \includegraphics[width=0.16\textwidth]{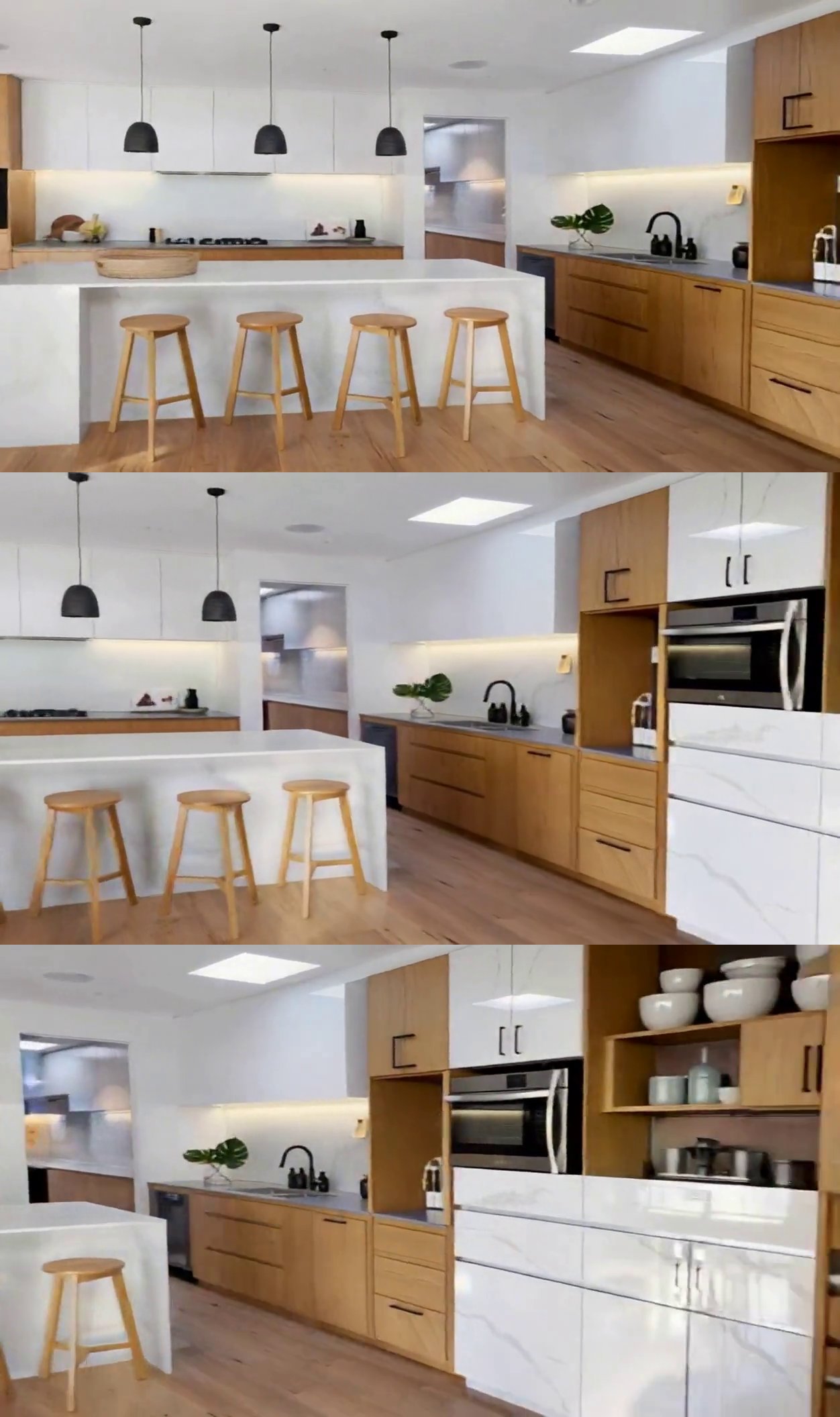} &
        \includegraphics[width=0.16\textwidth]{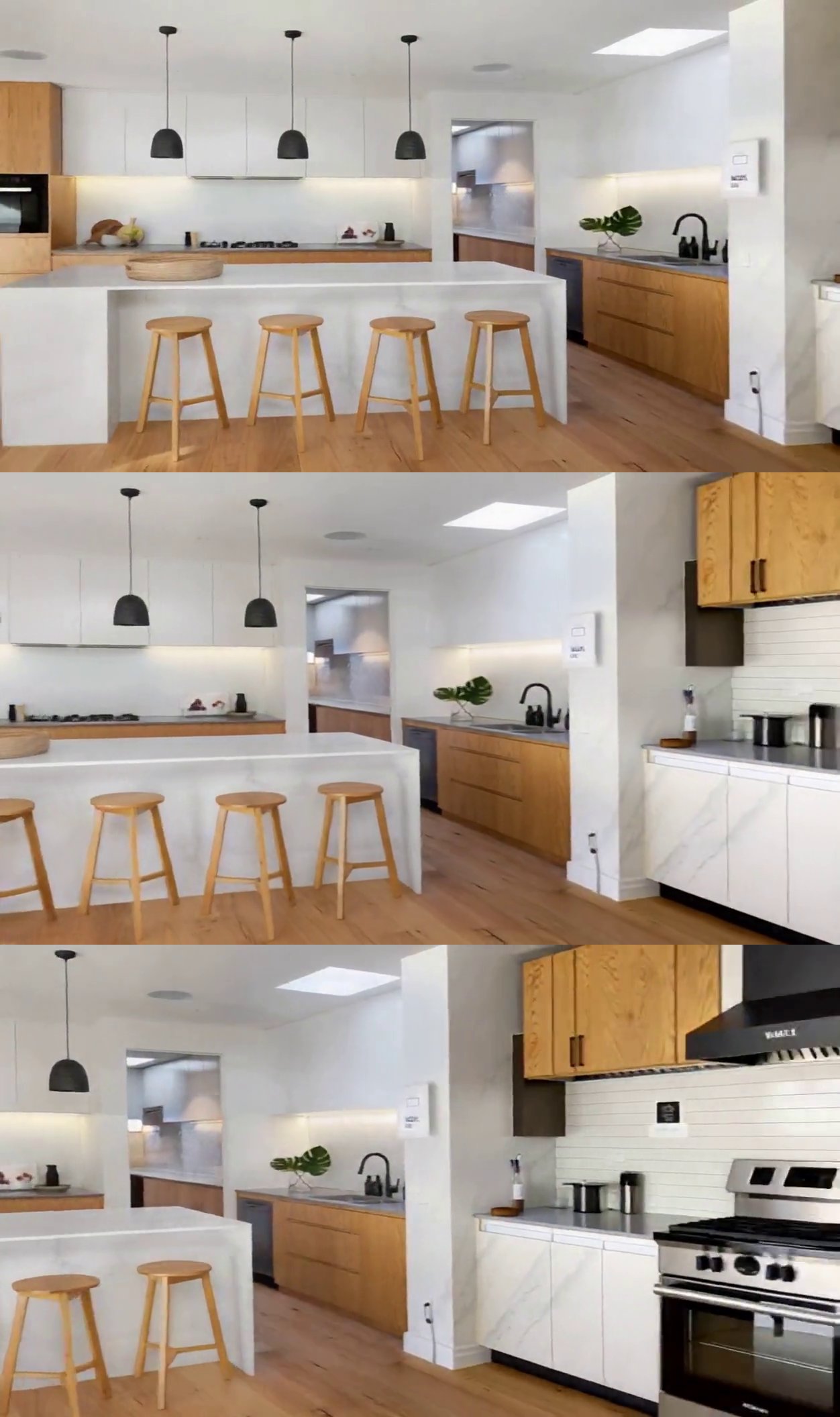}
        \\
        \multicolumn{3}{c}{\scriptsize{Generated frames with various seeds (moving backward)}} & & \multicolumn{3}{c}{\scriptsize{Generated frames with various seeds (rotating right)}} \\
        \arrayrulecolor{gray}\hline \\
        \includegraphics[width=0.16\textwidth]{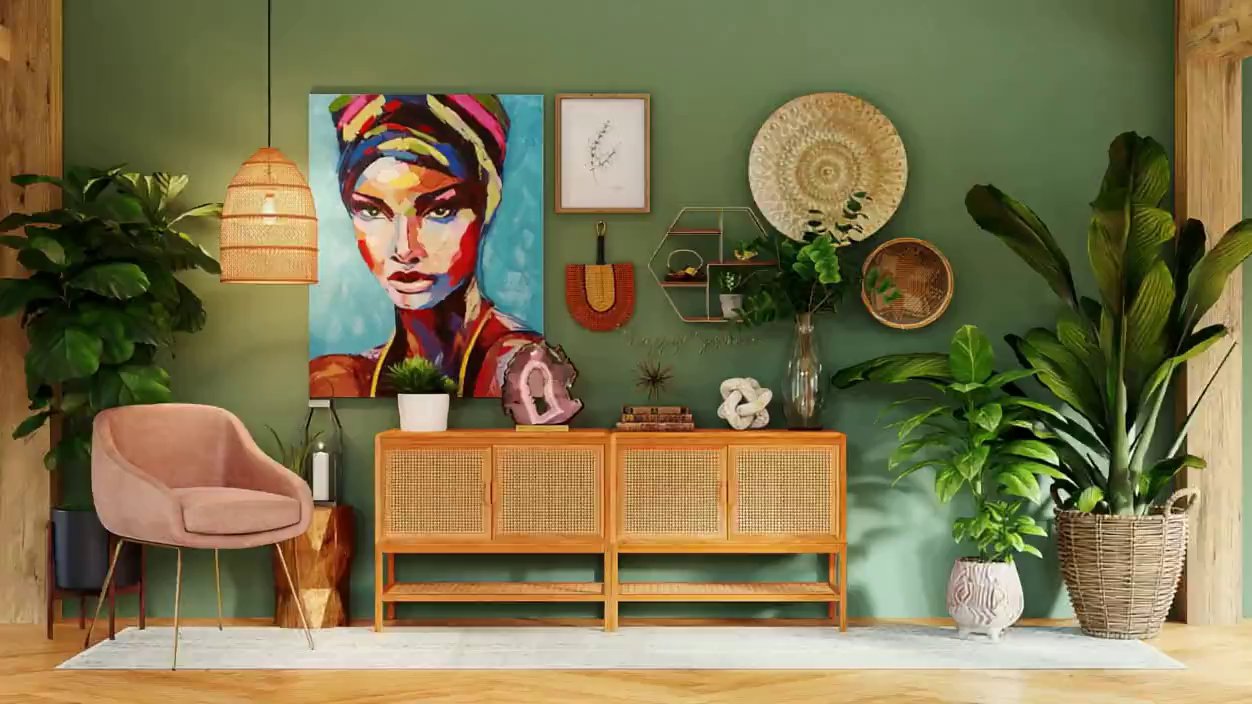} &
        \includegraphics[width=0.16\textwidth]{images/6_post_training/3d_joystick_seeds/0038-0_0.jpg} &
        \includegraphics[width=0.16\textwidth]{images/6_post_training/3d_joystick_seeds/0038-0_0.jpg} & &
        \includegraphics[width=0.16\textwidth]{images/6_post_training/3d_joystick_seeds/0038-0_0.jpg} &
        \includegraphics[width=0.16\textwidth]{images/6_post_training/3d_joystick_seeds/0038-0_0.jpg} &
        \includegraphics[width=0.16\textwidth]{images/6_post_training/3d_joystick_seeds/0038-0_0.jpg}
        \\
        \multicolumn{3}{c}{\scriptsize{Input frame}} & & \multicolumn{3}{c}{\scriptsize{Input frame}} \\
        \includegraphics[width=0.16\textwidth]{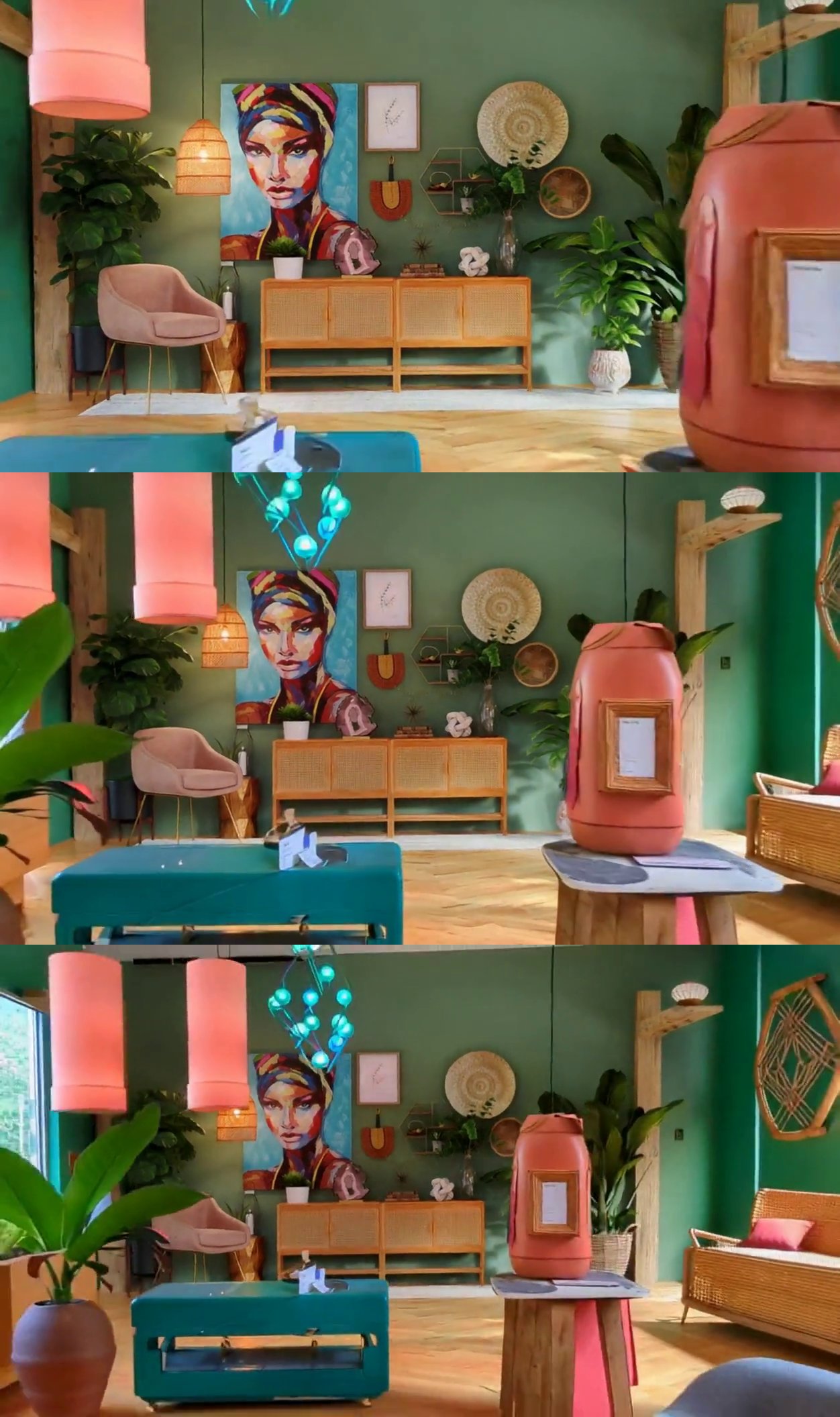} &
        \includegraphics[width=0.16\textwidth]{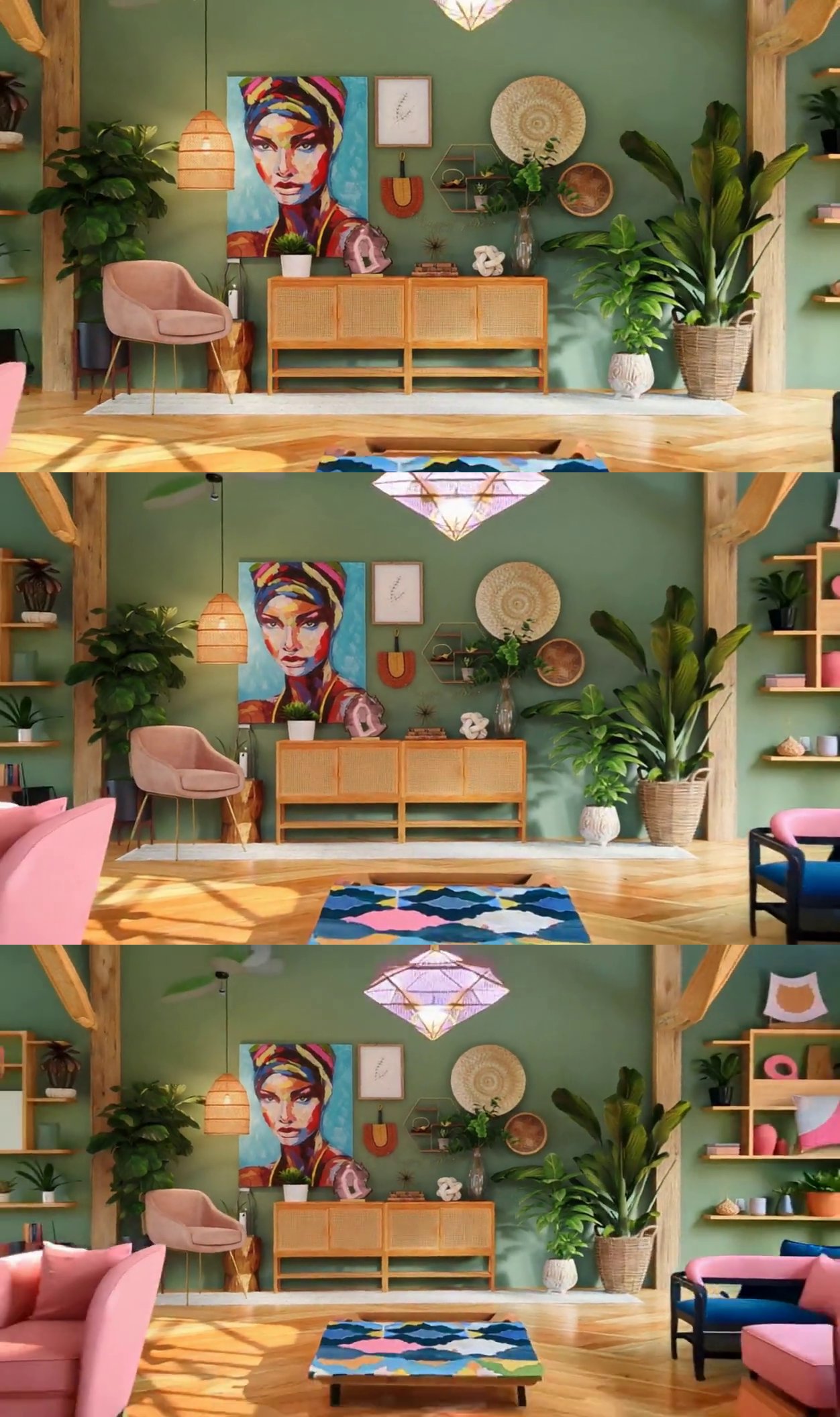} &
        \includegraphics[width=0.16\textwidth]{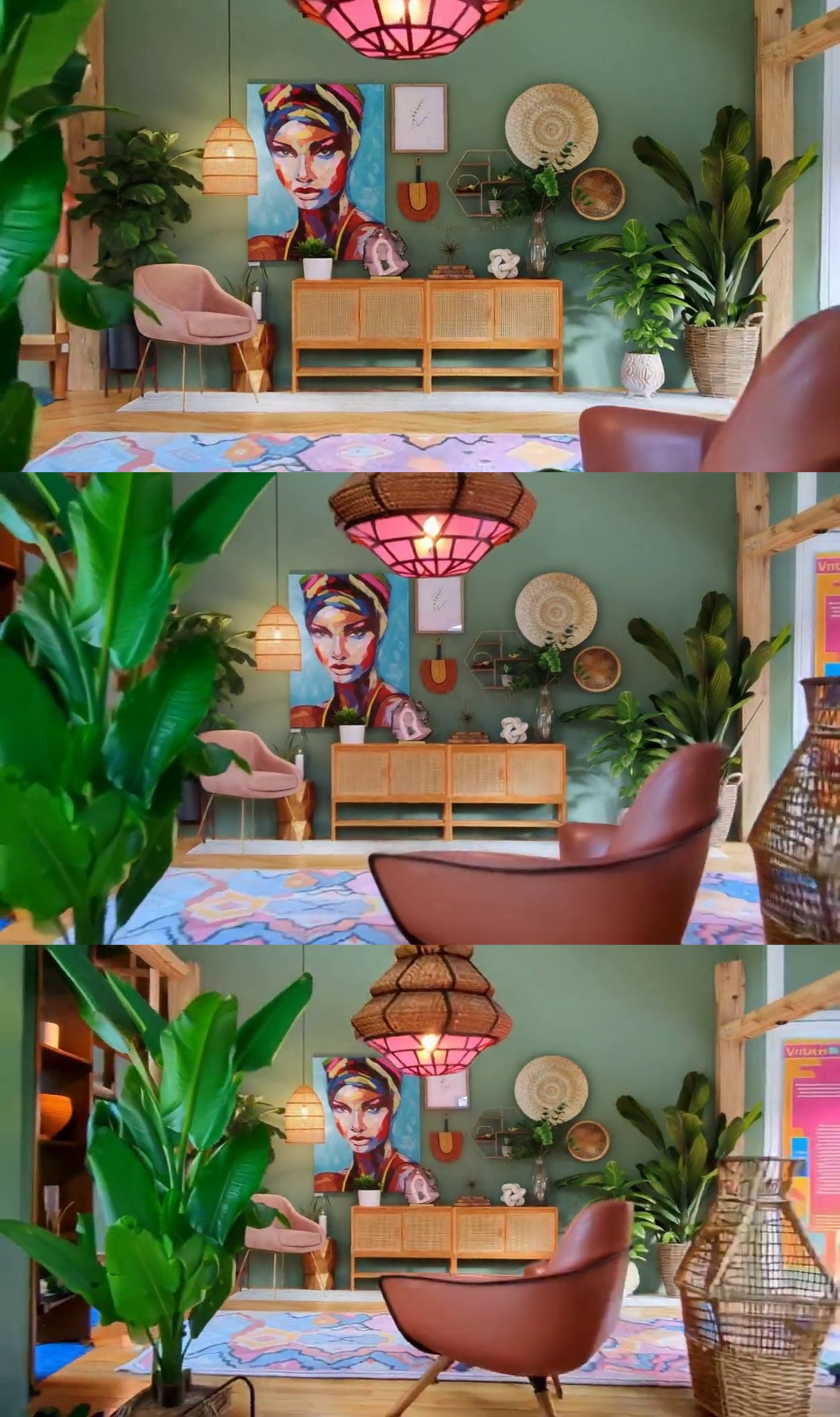} & &
        \includegraphics[width=0.16\textwidth]{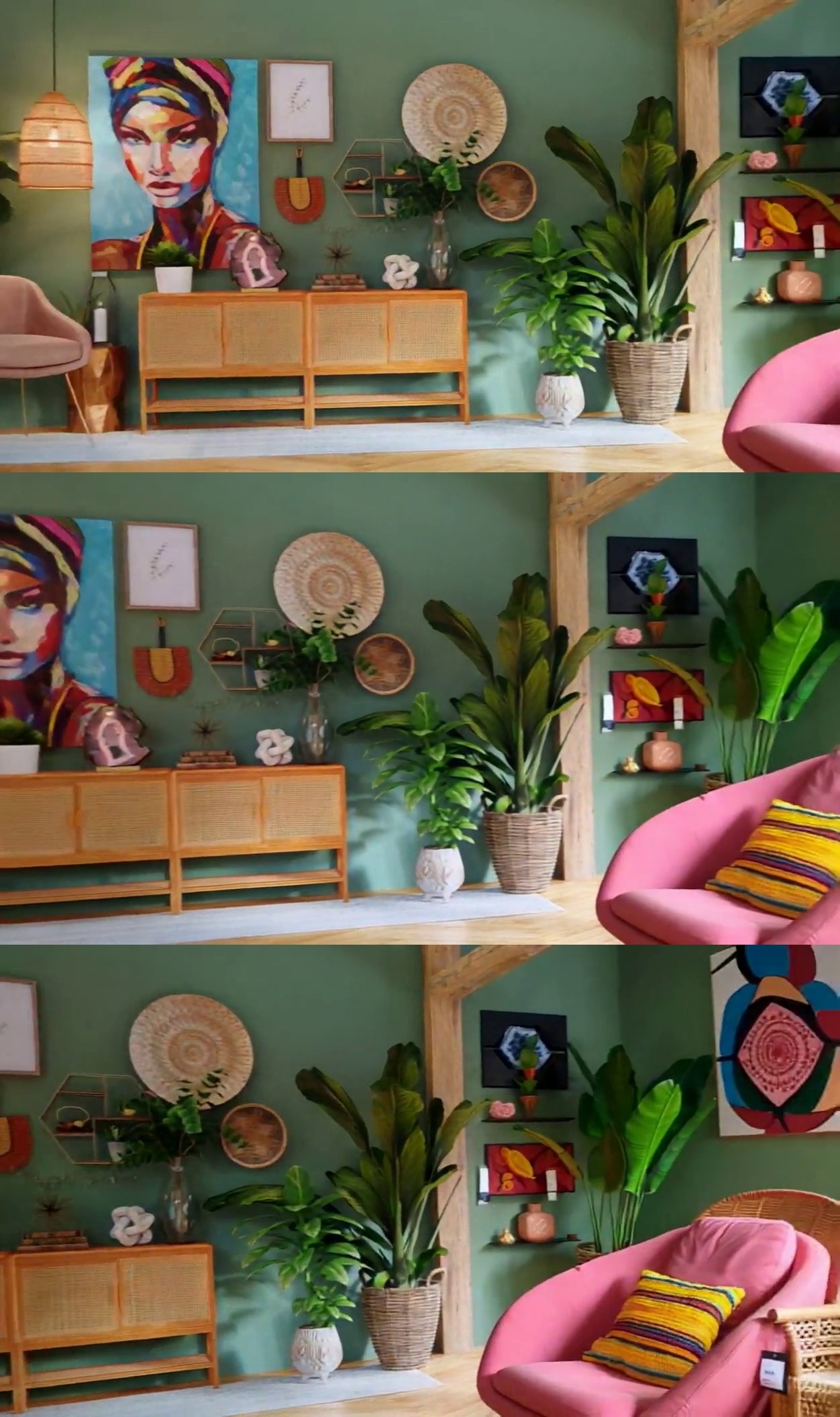} &
        \includegraphics[width=0.16\textwidth]{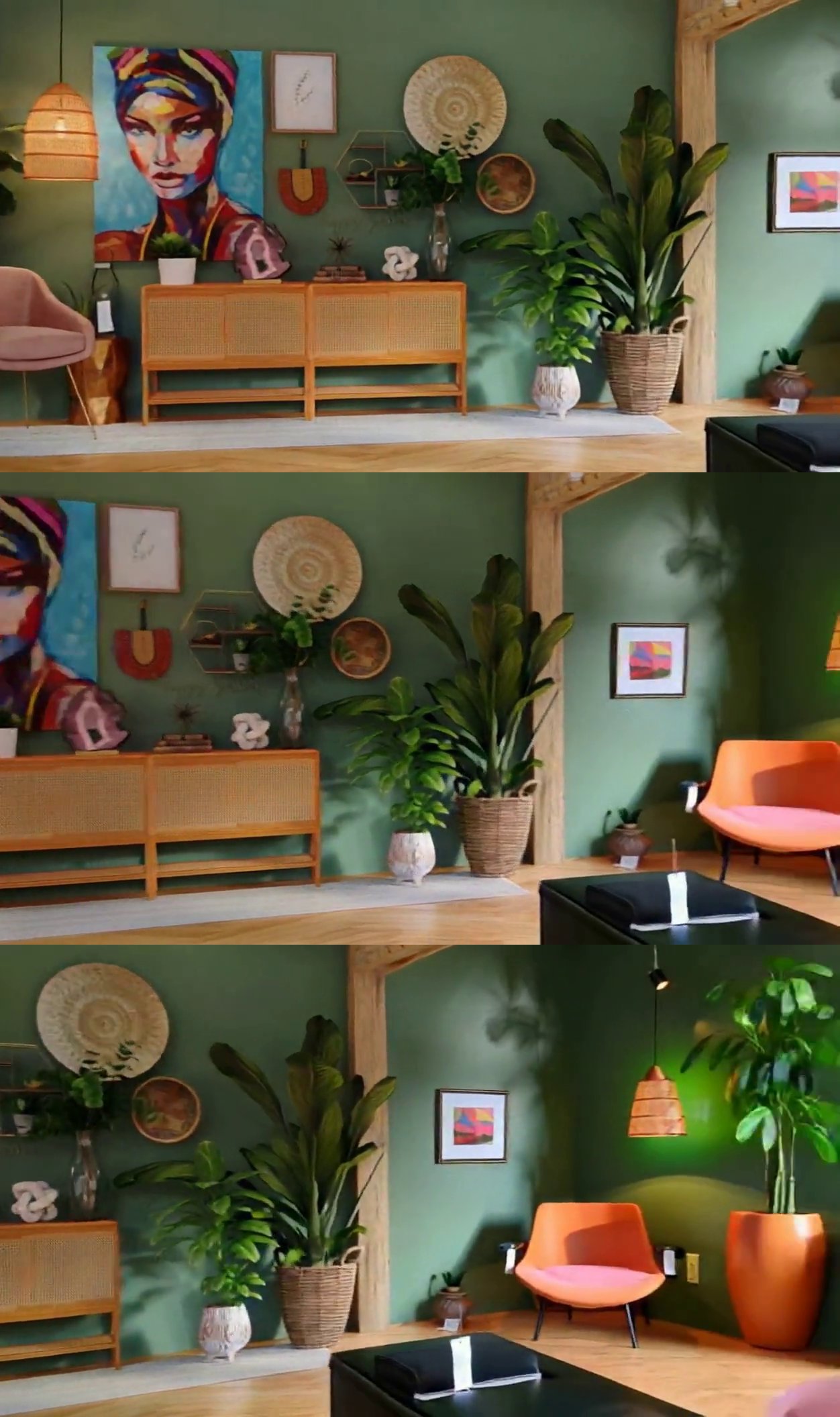} &
        \includegraphics[width=0.16\textwidth]{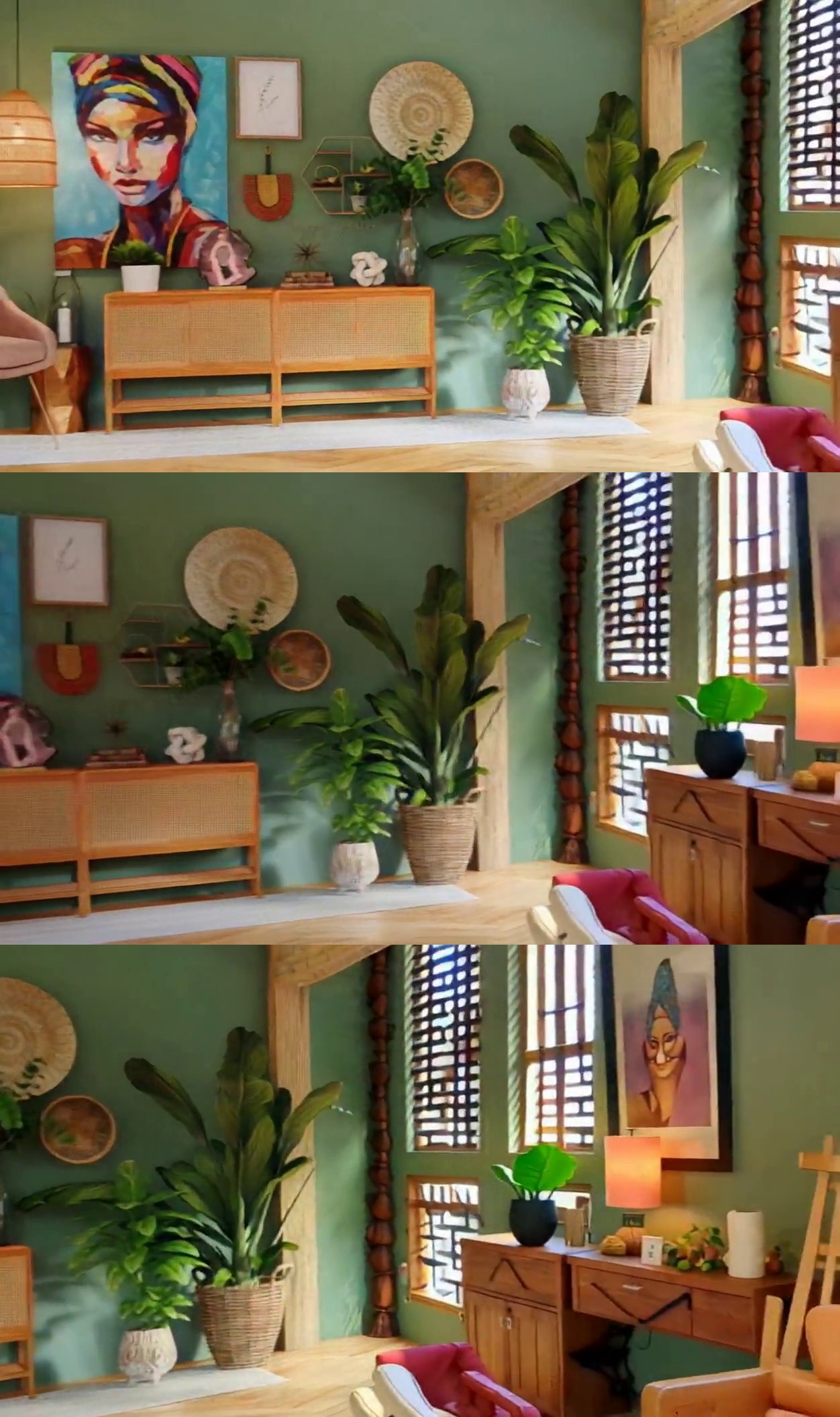}
        \\
        \multicolumn{3}{c}{\scriptsize{Generated frames with various seeds (moving backward)}} & & \multicolumn{3}{c}{\scriptsize{Generated frames with various seeds (rotating right)}}
    \end{tabular}

    \caption{
        \textbf{Cosmos-Predict1-7B-Video2World-Sample-CameraCond results with different seeds.} We show the capability of simulating diverse futures with our camera control model given the same input image and camera condition. For each group, we apply the same input frame and camera condition created with joystick. The first group shows \textit{moving backward} and second group shows \textit{rotating right}. Within each group, we show the generated videos with 3 different random seeds in each column. We visualize frames 19, 38, and 57 from the generated videos.
    }
    \label{fig:camera-joystick-seeds}
\end{figure*}

\noindent \textbf{Qualitative results.} \cref{fig:camera-joystick-wasd} shows our results from joystick-like control input on the camera, including \emph{moving forward}, \emph{moving backward}, \emph{rotating left}, and \emph{rotating right}. This demonstrates the use case where one can navigate the simulated world using a joystick to control the model in generating future video frames.
A Physical AI agent could also use such control to predict the future of the world under different scenarios.

To show the diversity of the generation, we show generation results from the same input image and camera control with different random seeds in~\cref{fig:camera-joystick-seeds}. Cosmos-Predict1-7B-Video2World-Sample-CameraCond is able to generate different worlds while still maintaining 3D spatial and temporal coherence in the videos. This could be used to simulate different possible futures given the current states.

\subsection{Post-training WFM for Robotic Manipulation}\label{sec::robo}

A world model has the potential to serve as a powerful planner and simulator for robotic manipulation. Here, we demonstrate how we fine-tune our pre-trained WFMs for two tasks: (1) instruction-based video prediction and (2) action-based next-frame generation. For \textbf{instruction-based video prediction}, the input is the current video frame of a robot as well as a text instruction, and the output is a predicted video of the robot following the instruction. For \textbf{action-based next-frame prediction}, the input is the current video frame of a robot as well as an action vector between the current and next frame, and the output is the predicted next frame showing the result of the robot performing the specified action. Given a sequence of actions,  the model can be run autoregressively to predict a video of the robot executing the given actions.

\subsubsection{Datasets}

We curate two datasets for the two tasks described above. For \textbf{instruction-based video prediction}, we created an internal dataset called the Cosmos-1X dataset. It comprises approximately 200 hours of egocentric videos captured by EVE, a humanoid robot from 1x.Tech~\citep{1xgpt} performing a variety of tasks, including navigation, folding clothes, cleaning tables, picking up objects, \etc. From the raw videos, we selected approximately $12{,}000$ episodes ranging from 1 to 9 seconds. Each episode is labeled with a one-sentence instruction, which is later upsampled with a proprietary VLM. The videos are captured at 30 FPS with a resolution of $512\times512$.

For \textbf{action-based next-frame generation}, we used a public dataset called Bridge~\citep{frederik22bridge}, with the same configuration as a prior work~\citep{zhu2024irasim} for comparison. The Bridge dataset includes approximately $20{,}000$ episodes of third-person views of a robot arm performing different tasks in a kitchen environment, with videos of $320\times256$ resolution captured at 5 FPS. For each video frame, the corresponding action is defined as a 7-dimensional vector in the gripper coordinate space $(\Delta x, \Delta y, \Delta z, \Delta \theta_r, \Delta \theta_p, \Delta \theta_y, \Delta\mbox{Gripper})$ as in OpenVLA~\citep{kim24openvla}.

\subsubsection{Fine-tuning}

We fine-tune both our Cosmos-Predict1-7B-Video2World (\cref{sec::diffusion_model}) and Cosmos-Predict1-5B-Video2World (\cref{sec::autoregress}) for instruction-based video prediction and action-based next-frame prediction tasks.

For \textbf{instruction-based video prediction}, we build two models based on the base WFMs. The first is called Cosmos-Predict1-7B-Video2World-Sample-Instruction, and the second is called Cosmos-Predict1-5B-Video2World-Sample-Instruction. We compute the T5 embedding of the instruction, which is added to the finetuing of the base model via cross-attention.

For \textbf{action-based next-frame prediction}, we also build two models based on the base WFMs. The first one is called Cosmos-Predict1-7B-Video2World-Sample-ActionCond, and the second one is called Cosmos-Predict1-5B-Video2World-Sample-ActionCond.

Since action is a new modality not encountered during pre-training, we introduce additional modules inside our models for conditioning. For Cosmos-Predict1-5B-Video2World-Sample-ActionCond, we add an action embedder MLP to project the action vector into a tensor, which is then incorporated into the model via cross-attention. For Cosmos-Predict1-7B-Video2World-Sample-ActionCond, we also add an action embedder MLP to predict the action into a tensor but instead, incorporate it into the model by adding it to the timestamp embedding of the DiT modules.

\subsubsection{Evaluation}

\begin{figure}[!ht]
    \centering
    \begin{subfigure}[t]{0.49\textwidth}
        \centering
        \includegraphics[width=\textwidth]{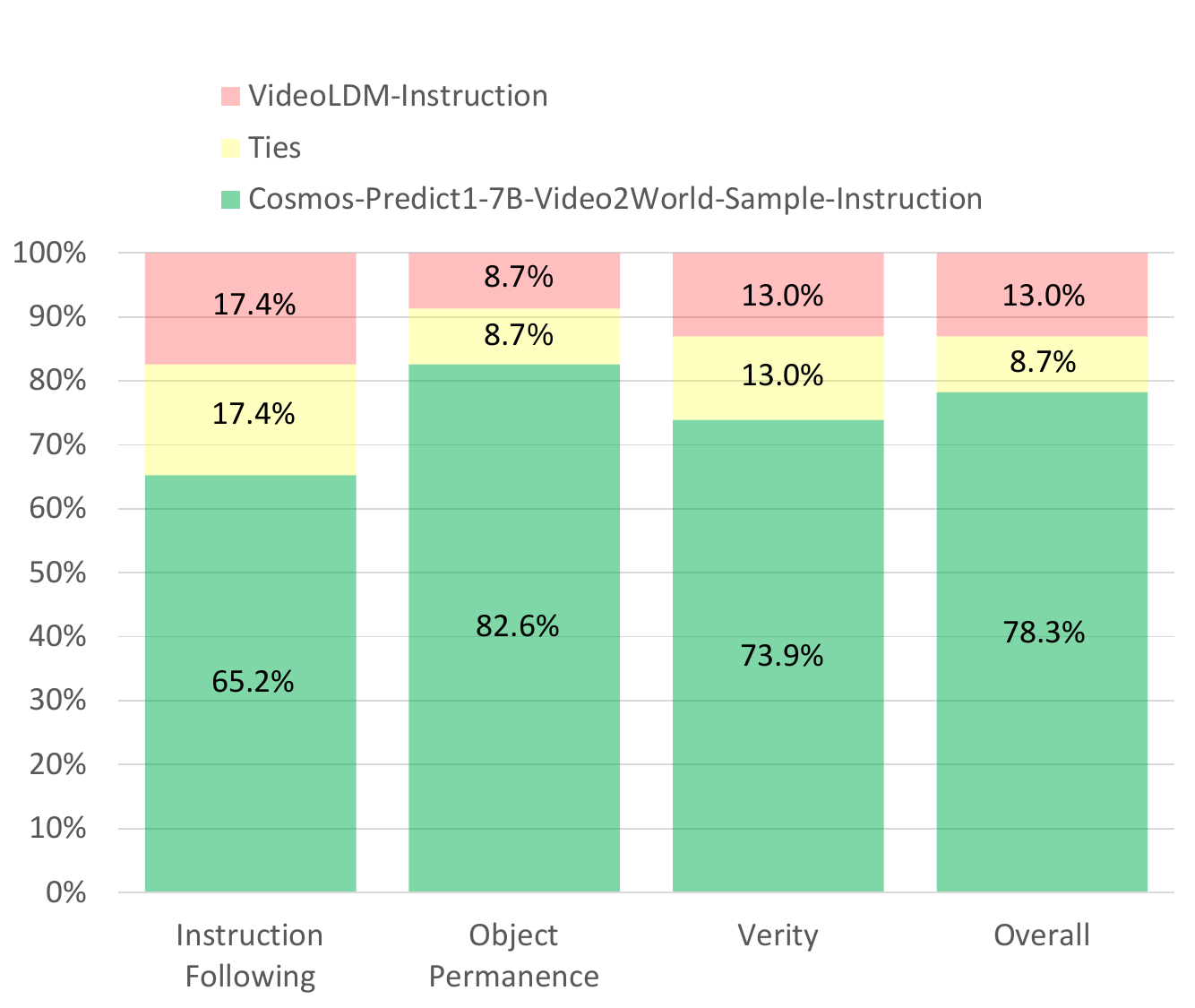}
        \caption{Cosmos-Predict1-7B-Video2World-Sample-Instruction \vs VideoLDM-Instruction.}
    \end{subfigure}
    \hfill
    \begin{subfigure}[t]{0.49\textwidth}
        \centering
         \includegraphics[width=\textwidth]{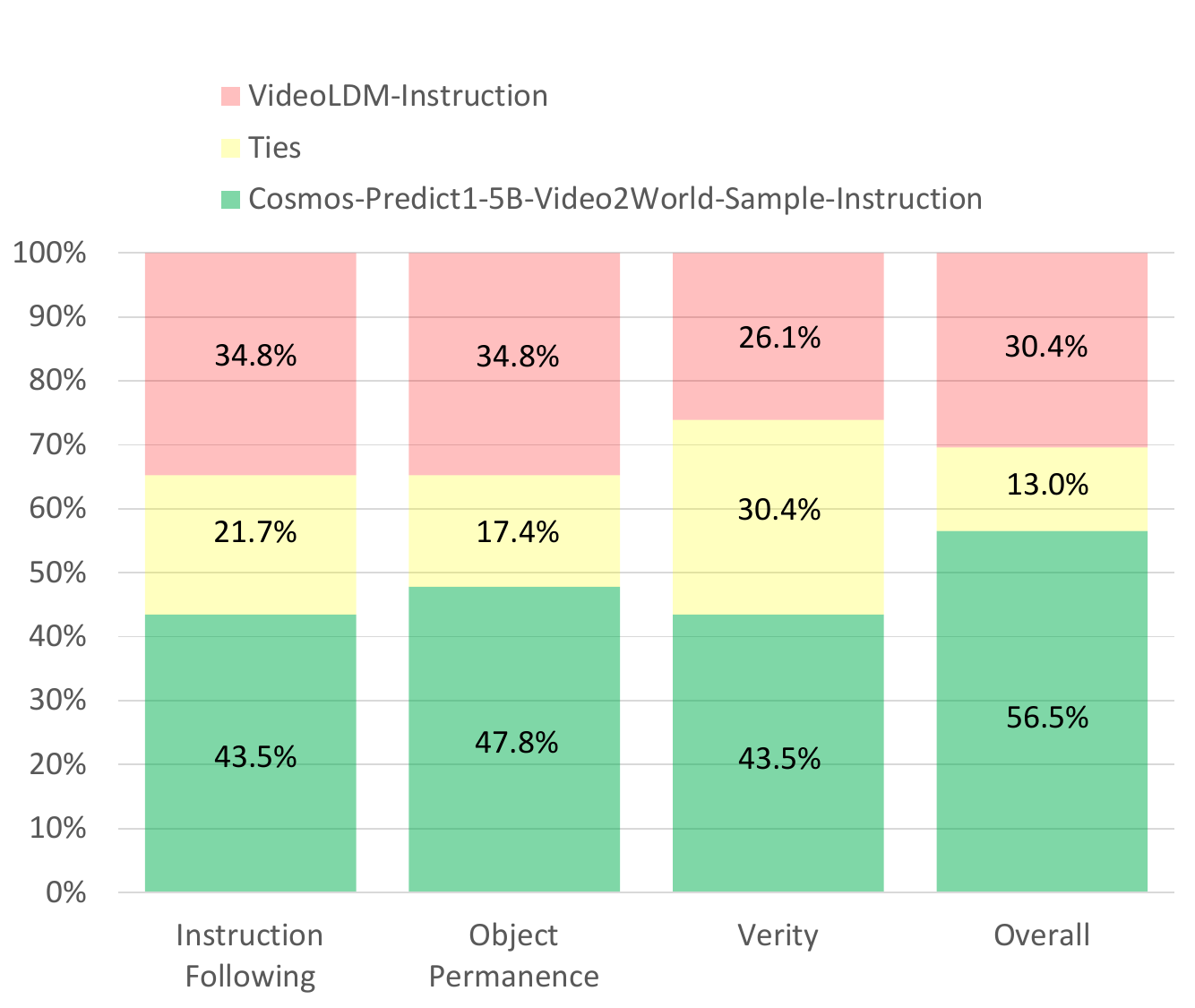}
        \caption{Cosmos-Predict1-5B-Video2World-Sample-Instruction \vs VideoLDM-Instruction.}
    \end{subfigure}
    \caption{\textbf{Human evaluation results for instruction-based video prediction on Cosmos-1X dataset}. The results show that compared with the baseline model (VideoLDM-instruction), our fine-tuned models (Cosmos-Predict1-7B-Video2World-Sample-Instruction and Cosmos-Predict1-5B-Video2World-Sample-Instruction) have higher preferences in the four evaluation dimensions.}
    \label{fig:instruction_conditioned_results_2x2}
\end{figure}

\begin{figure*}[ht]
    \centering
    \setlength{\tabcolsep}{1pt}
    \renewcommand{\arraystretch}{0.5}
    \resizebox{1.009\textwidth}{!}{%
    \begin{tabular}{ccccccccccc} %
        \footnotesize{Input frame} & & \multicolumn{3}{c}{\footnotesize Instruction-conditioned generation} & & \footnotesize{Input frame} & & \multicolumn{3}{c}{\footnotesize Instruction-conditioned generation} \\
        \includegraphics[width=0.12\textwidth]{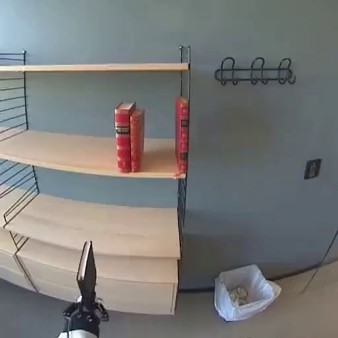} & & \includegraphics[width=0.12\textwidth]{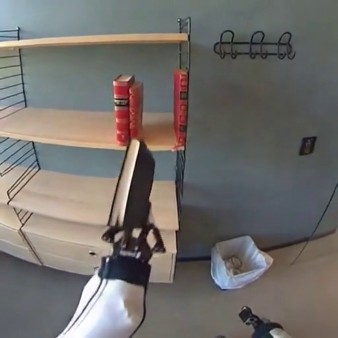} & \includegraphics[width=0.12\textwidth]{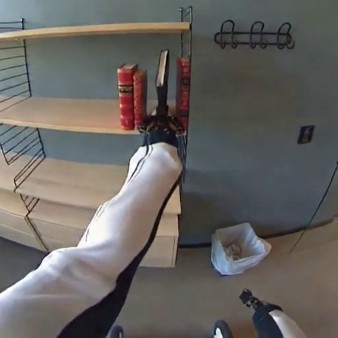} & \includegraphics[width=0.12\textwidth]{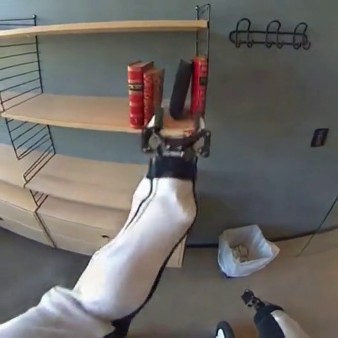} & & \includegraphics[width=0.12\textwidth]{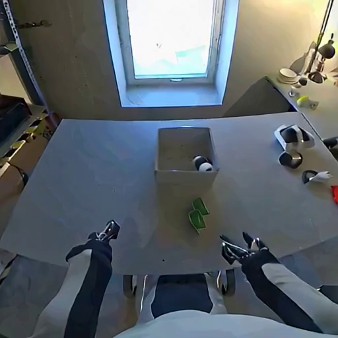} & & \includegraphics[width=0.12\textwidth]{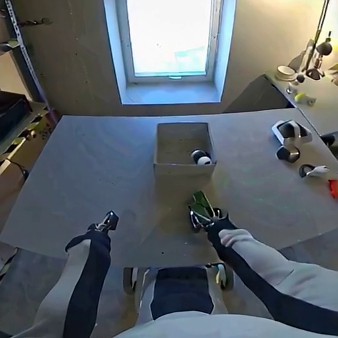} & \includegraphics[width=0.12\textwidth]{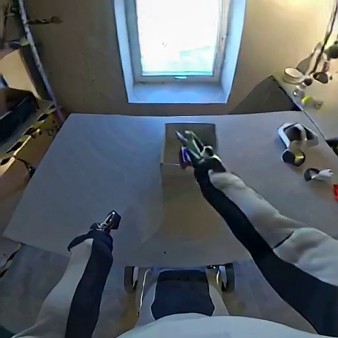} & \includegraphics[width=0.12\textwidth]{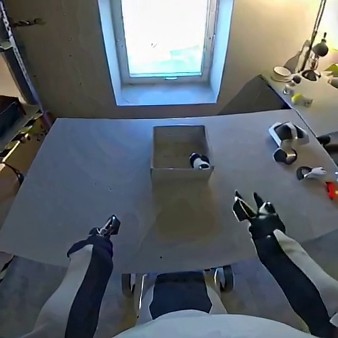} \\
        \multicolumn{5}{p{\dimexpr\textwidth/2\relax}}{
        \begin{multicoltextblockhalf}
        \prompttext{Prompt}: \prompt{Organize books by placing them vertically on a shelf.}
        \end{multicoltextblockhalf}
        } & & \multicolumn{5}{p{\dimexpr\textwidth/2\relax}}{
        \begin{multicoltextblockhalf}
        \prompttext{Prompt}: \prompt{Grip and elevate a green object from a box on a tidy worktable.}
        \end{multicoltextblockhalf}
        } \\
        & \\ [6pt]
         \includegraphics[width=0.12\textwidth]{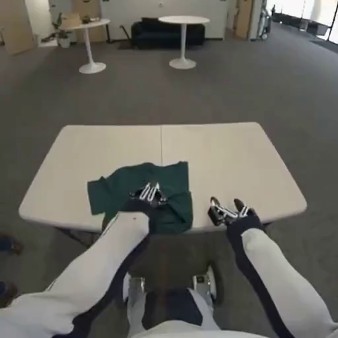} & & \includegraphics[width=0.12\textwidth]{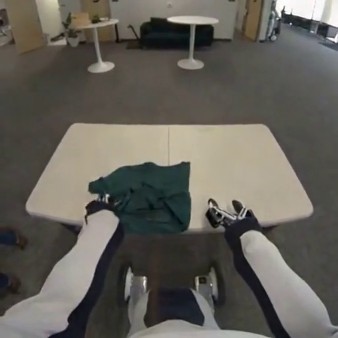} & \includegraphics[width=0.12\textwidth]{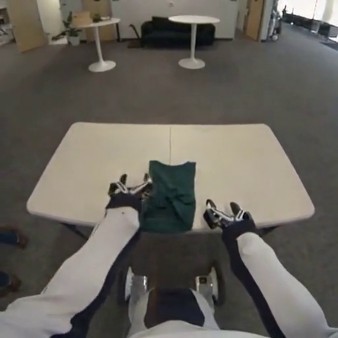} & \includegraphics[width=0.12\textwidth]{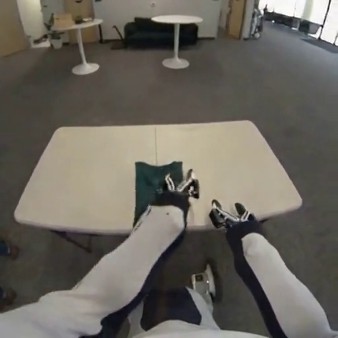} & & \includegraphics[width=0.12\textwidth]{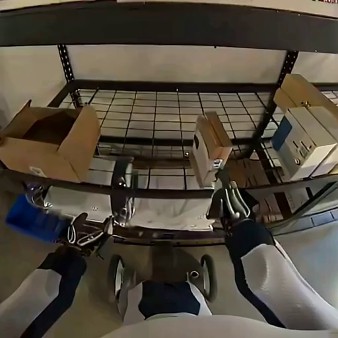} & & \includegraphics[width=0.12\textwidth]{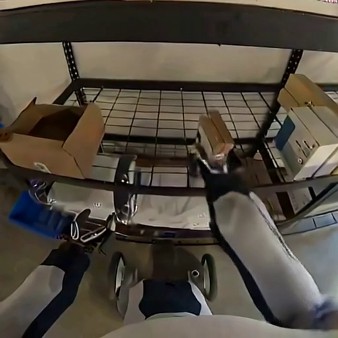} & \includegraphics[width=0.12\textwidth]{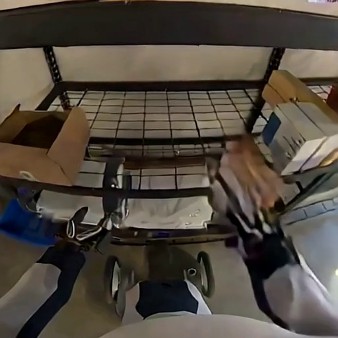} & \includegraphics[width=0.12\textwidth]{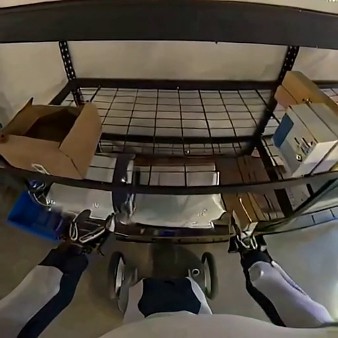} \\
        \multicolumn{5}{p{\dimexpr\textwidth/2\relax}}{
        \begin{multicoltextblockhalf}
        \prompttext{Prompt}: \prompt{Fold a green fabric item on a table.}
        \end{multicoltextblockhalf}
        } & & \multicolumn{5}{p{\dimexpr\textwidth/2\relax}}{
        \begin{multicoltextblockhalf}
        \prompttext{Prompt}: \prompt{Retrieve a box from a storage shelf using its articulated hands in a warehouse setting.}
        \end{multicoltextblockhalf}
        } \\
        \\
        \multicolumn{5}{c}{\footnotesize Cosmos-Predict1-7B-Video2World-Sample-Instruction} & & \multicolumn{5}{c}{\footnotesize Cosmos-Predict1-5B-Video2World-Sample-Instruction} \\
    \end{tabular}
    }
    \caption{\textbf{Instruction-based video prediction samples on the Cosmos-1X dataset}. The left are the results of Cosmos-Predict1-7B-Video2World-Sample-Instruction model, and the right are the results of Cosmos-Predict1-5B-Video2World-Sample-Instruction model.}
    \label{fig:robot_instruction_conditioned}
\end{figure*}

\begin{figure*}[th!]
    \centering
    \setlength{\tabcolsep}{1pt}
    \begin{tabular}{cccccccccccc} %
         & \footnotesize{Input frame} & & \multicolumn{3}{c}{\footnotesize Predicted frames} & & \footnotesize{Input frame} & & \multicolumn{3}{c}{\footnotesize Predicted frames} \\
         \rotatebox{90}{\footnotesize \hspace{9pt} Prediction} & \includegraphics[width=0.1171\textwidth]{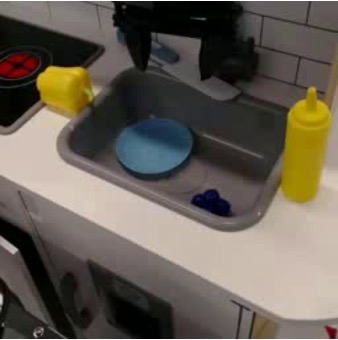} & & \includegraphics[width=0.1171\textwidth]{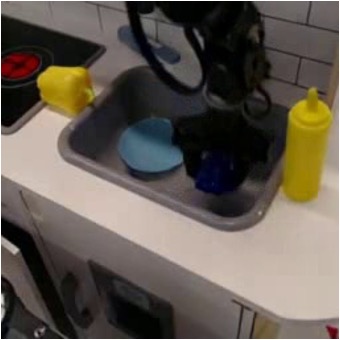} & \includegraphics[width=0.1171\textwidth]{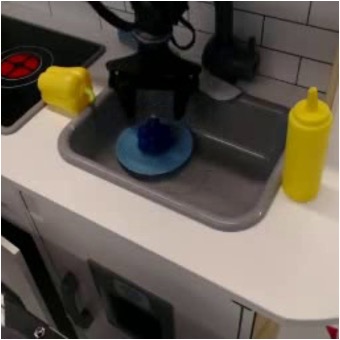} & \includegraphics[width=0.1171\textwidth]{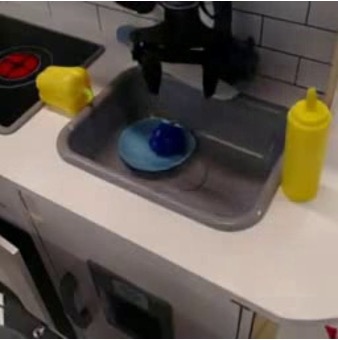} & & \includegraphics[width=0.1171\textwidth]{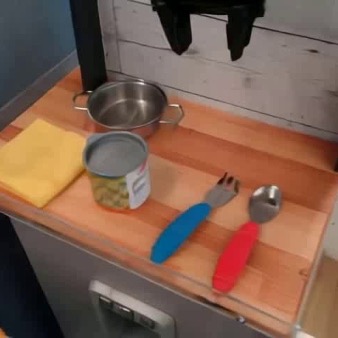} & & \includegraphics[width=0.1171\textwidth]{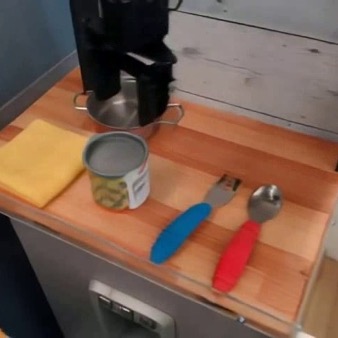} & \includegraphics[width=0.1171\textwidth]{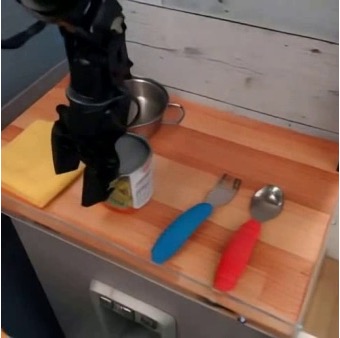} & \includegraphics[width=0.1171\textwidth]{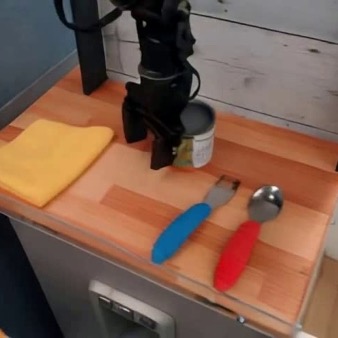} \\
         \rotatebox{90}{\footnotesize \hspace{22pt} GT} & \includegraphics[width=0.1171\textwidth]{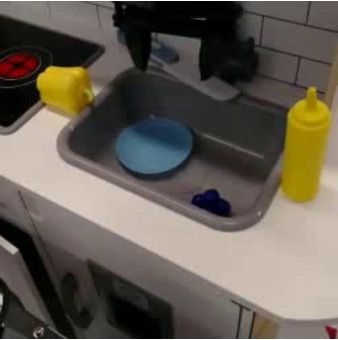} & & \includegraphics[width=0.1171\textwidth]{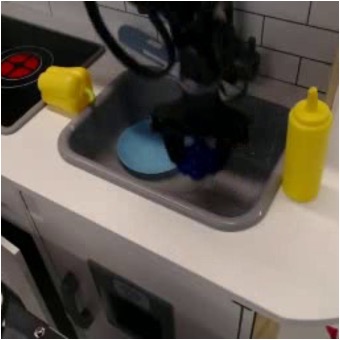} & \includegraphics[width=0.1171\textwidth]{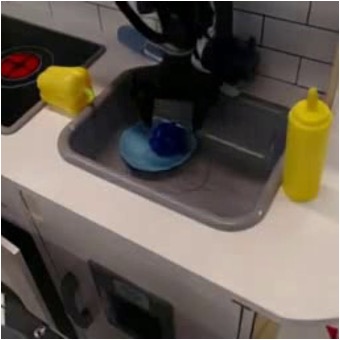} & \includegraphics[width=0.1171\textwidth]{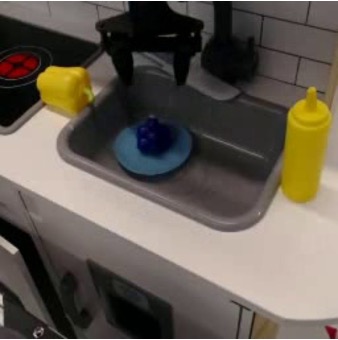} & & \includegraphics[width=0.1171\textwidth]{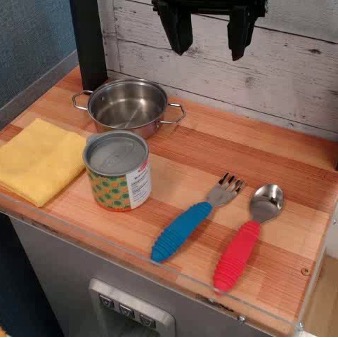} & & \includegraphics[width=0.1171\textwidth]{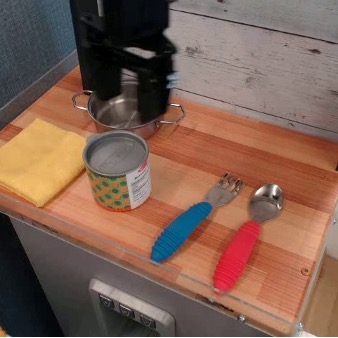} & \includegraphics[width=0.1171\textwidth]{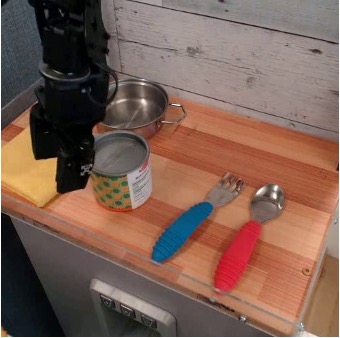} & \includegraphics[width=0.1171\textwidth]{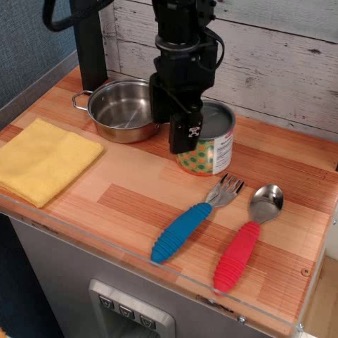} \\
         & \multicolumn{5}{c}{\footnotesize Cosmos-Predict1-7B-Video2World-Sample-ActionCond} & & \multicolumn{5}{c}{\footnotesize Cosmos-Predict1-5B-Video2World-Sample-ActionCond} \\
    \end{tabular}
    \caption{\textbf{Action-based next-frame prediction samples on the Bridge dataset}. The left is the results of the Cosmos-Predict1-7B-Video2World-Sample-ActionCond model, and the right is the results of the Cosmos-Predict1-5B-Video2World-Sample-ActionCond model. As shown, the predicted video frames closely match the GT video frames for both models.}
    \label{fig:robot_qualitative}
\end{figure*}

For \textbf{instruction-based video prediction}, we fine-tune VideoLDM~\citep{Blattmann2023Align} on the Cosmos-1X dataset and obtained VideoLDM-Instruction as a baseline for comparison. To evaluate the video generation performance of the models, we define the following dimensions:
\begin{itemize}
    \item \textbf{Instruction following}: Is the generated video aligned with the input language instruction?
    \item \textbf{Object permanence}: Do objects present in the scene remain throughout the generated video?
    \item \textbf{Verity}: Does the generated video faithfully represent the real world without unexpected imaginary objects?
    \item \textbf{Overall}: Is the generated video reasonable for the robot to plan accordingly?
\end{itemize}
Human evaluators are tasked to observe a pair of anonymous videos generated by different models but with the same language instruction and compare them along the dimensions listed above. A group of ten human evaluators performed the evaluation over 23 test episodes. The statistical results are summarized in \cref{fig:instruction_conditioned_results_2x2}.

As shown, we find that both Cosmos-Predict1-7B-Video2World-Sample-Instruction and Cosmos-Predict1-5B-Video2World-Sample-Instruction perform better than VideoLDM-Instruction along the four evaluation dimensions. Cosmos-Predict1-7B-Video2World-Sample-Instruction achieved $78.3\%$ overall preference compared to $13.0\%$ for VideoLDM-Instruction. Cosmos-Predict1-5B-Video2World-Sample-Instruction has also achieved better performance than diffusion-based VideoLDM-Instruction. Some predicted video frames for both fine-tuned WFMs are presented in \cref{fig:robot_instruction_conditioned}, which shows the quality of the predicted videos.

For \textbf{action-based next frame prediction}, we fine-tuned our models on the Bridge dataset. As a baseline, we fine-tune IRASim~\citep{zhu2024irasim} to derive an action-based next-frame prediction model IRASim-Action. We perform the next-frame prediction autoregressively to generate videos. To evaluate video generation quality, we compare the generated videos against ground truth videos over 100 episodes randomly selected from the official Bridge test set.

\begin{table}[ht]
    \setlength{\tabcolsep}{16.12pt} %
    \small
    \captionsetup{justification=centering}
    \caption{Evaluation of action-based next-frame prediction on Bridge dataset.}
    \centering
    \begin{tabular}{rcccc}
        \toprule
        Method & {PSNR} $\uparrow$ & {SSIM} $\uparrow$ & {Latent L2} $\downarrow$ & {FVD} $\downarrow$ \\
        \midrule
        IRASim-Action & 19.13 & 0.64 & 0.38 & 593 \\
        \midrule
        \makecell[r]{Cosmos-Predict1-5B-Video2World-\\Sample-ActionCond} & 19.95 & 0.80 & 0.36 & 434 \\
        \midrule
        \makecell[r]{Cosmos-Predict1-7B-Video2World-\\Sample-ActionCond} & \textbf{21.14} & \textbf{0.82} & \textbf{0.32} & \textbf{190} \\
        \bottomrule
    \end{tabular}
    \label{tab:robot_action_conditional_video_prediction}
\end{table}

The computed metrics are summarized in \cref{tab:robot_action_conditional_video_prediction}, including PSNR, SSIM, Latent L2~\citep{zhu2024irasim}, and FVD. As shown, both Cosmos-Predict1-5B-Video2World-Sample-ActionCond and Cosmos-Predict1-7B-Video2World-Sample-ActionCond models outperform the baseline model (IRASim-Action). Some predicted video frames are presented in~\cref{fig:robot_qualitative}, which shows the quality of the predicted videos compared to the ground truth.

\subsection{Post-training WFM for Autonomous Driving}\label{sec::av}

A world model for in-the-wild driving scenes has the potential to serve as a powerful simulation engine for training autonomous driving agents. As most autonomous vehicles are equipped with multiple cameras viewing different directions, an ideal world model for an autonomous vehicle should also be a multi-view one, preferably matching the precise setup of the sensors in the target vehicle. Here, we demonstrate how we fine-tune our pre-trained WFM to create a multi-view world model for autonomous driving tasks.

\subsubsection{Dataset}

We curate an internal dataset called the Real Driving Scene (RDS) dataset. It comprises approximately 3.6 million 20-second surround-view video clips (equivalent to approximately $20{,}000$ hours of data) captured using an NVIDIA internal driving platform. Each clip is recorded from six camera views: front, left, right, rear, rear-left, and rear-right. In addition, the dataset includes ego-motion information that we use to construct the trajectory data. We use the recorded timestamps of the front camera video to synchronize the frames of all other views.

This dataset was selected from a large labeled data corpus to match a target distribution of data attributes. The specific attribute tags include:
\begin{itemize}
  \item Contender vehicle density (\eg, none, low, medium, high)
  \item Weather (\eg, clear, raining, snowing, fog)
  \item Illumination (\eg, day, night)
  \item Ego vehicle speed (\eg, standing, low, local, highway speeds)
  \item Ego vehicle behavior (\eg, high, medium, low curvature trajectories and accelerations)
  \item Road type/population density (based on OpenStreetMap definitions: rural, residential, urban).
\end{itemize}
Additionally, the dataset was augmented through a second data-mining run to ensure a minimum number of clips containing rare road structures (\eg, tollbooths, bridges, tunnels, speed bumps, \etc). Finally, videos from each camera view are captioned separately, starting with a template text string: ``The video is captured from a camera mounted on a car. The camera is facing forward|left|right|backward|rear-left|rear-right.''

\subsubsection{Fine-tuning}

We fine-tune our Cosmos-Predict1-7B-Text2World (\cref{sec::diffusion_model}) into a multiple-view world model using the RDS dataset. To ensure consistent video generation across multiple views, we slightly modify the architectural design described in \cref{sec::diffusion_model} and fine-tune the WFM to generate videos from all six cameras simultaneously.

We build three multi-view world models, summarized in~\cref{tab:post_model_summarize}. The first one is called Cosmos-Predict1-7B-Text2World-Sample-MultiView, which is a multi-view world model that can generate six camera views based on a text prompt input. The second one is called Cosmos-Predict1-7B-Text2World-Sample-MultiView-TrajectoryCond. This model is built on top of Cosmos-Predict1-7B-Text2World-Sample-AV-MultiView and takes an additional trajectory input as the conditional input signal. The final model, Cosmos-Predict1-7B-Video2World-Sample-MultiView, is fine-tuned from the Diffusion-7B-Video2World-Sample-MultiView model to support video-based conditioning. It achieves this by incorporating previous frames into the generation process. Cosmos-Predict1-7B-Video2World-Sample-MultiView can take the video output from Cosmos-Predict1-7B-Text2World-Sample-MultiView and generate its extension. All three models output 6 views of 57 frames of video at a resolution of $848 \times 480$.

\noindent \textbf{View-independent positional embedding and view embedding.} Instead of extending the FPS-aware 3D RoPE Positional Embedding to include an additional view dimension, we opt to use the same positional embedding described in \cref{sec::diffusion_model} independently to each view. To represent view differences, we modify the denoising function $D_\theta$ to take an additional view embedding as input. That is, the camera view information is supplied through global view embeddings instead of positional embedding.

\noindent \textbf{View-dependent cross-attention.} In our multi-view setting, each of the six views of the same scene would have a different video description. While we treat all six views as a whole as the state of the diffusion process and perform self-attention among all the elements in the six views for denoising, we find it beneficial to employ view-dependent cross-attention for textual inputs. Specifically, the cross-attention operation for each view only attends to the textual description for the specific view. Note that each view has a different video description in our dataset. With the view embedding and view-dependent cross-attention, we derive Cosmos-Predict1-7B-Text2World-Sample-MultiView from fine-tuning Cosmos-Predict1-7B-Text2World.

\noindent \textbf{Trajectory control condition.} Optionally, in addition to the text condition, we fine-tune the model to produce videos that conform to the given future trajectory paths to enable more precise control of the agent. This enables the generation of unique driving scenarios that adhere to both driving trajectories recorded by real-world data and driving environments specified by the input text descriptions. The fine-tuned model is Cosmos-Predict1-7B-Text2World-Sample-MultiView-TrajectoryCond.

We define a trajectory as a sequence of 64 points in the 3D space, representing a sequence of translations of the agent from the initial position $(0,0,0)$ to the final destination, with each point separated by a 0.1-second interval. We compute the embedding of the trajectory input and make the result a conditional input to the denoiser of the fine-tuned Cosmos-Predict1-7B-Video2World model. We note that it is possible to achieve more fine-grained control signals by giving a per-interval action vector, following prior works~\citep{Kim2020_GameGan,Kim2021_DriveGAN,hu2023gaia} or as in the robotic manipulation task (\cref{sec::robo}). We leave such extensions for future work.

\begin{figure*}[ht]
    \centering
    \setlength{\tabcolsep}{0pt}
    \begin{tabular}{c@{\hspace{0.3cm}}c} %
        \includegraphics[width=0.49\textwidth]{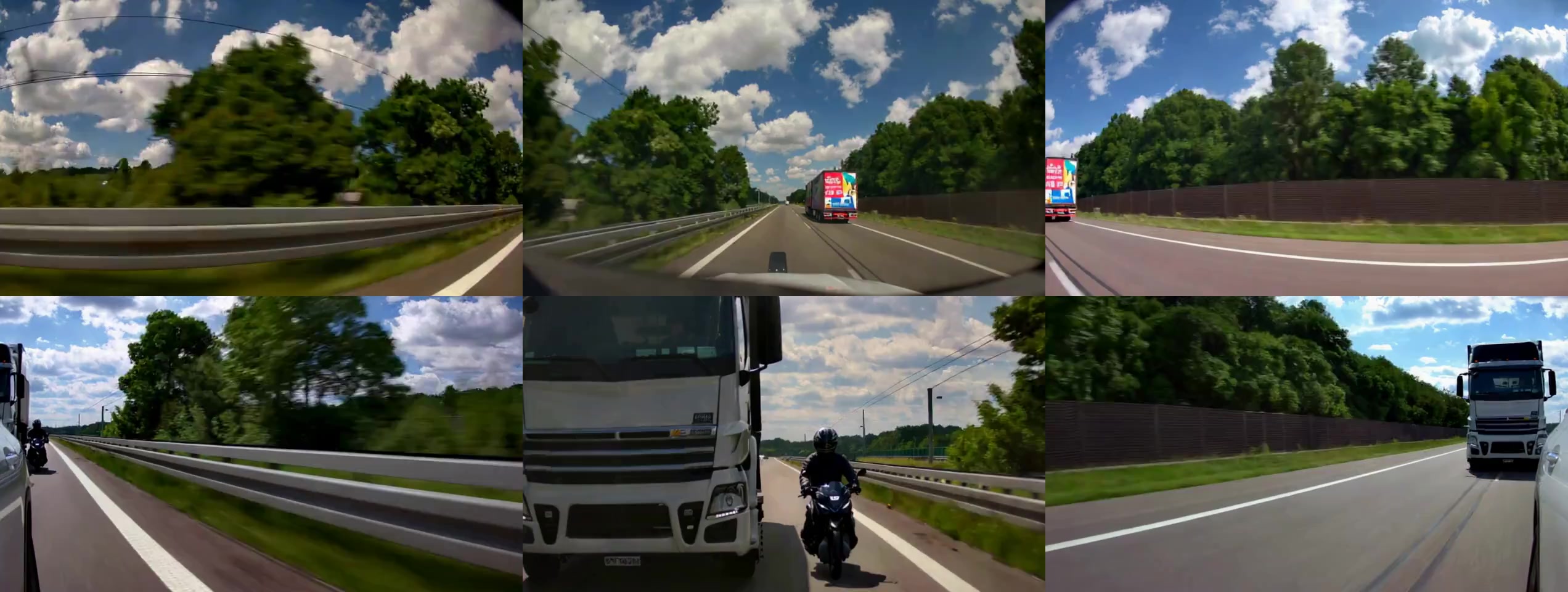} & \includegraphics[width=0.49\textwidth]{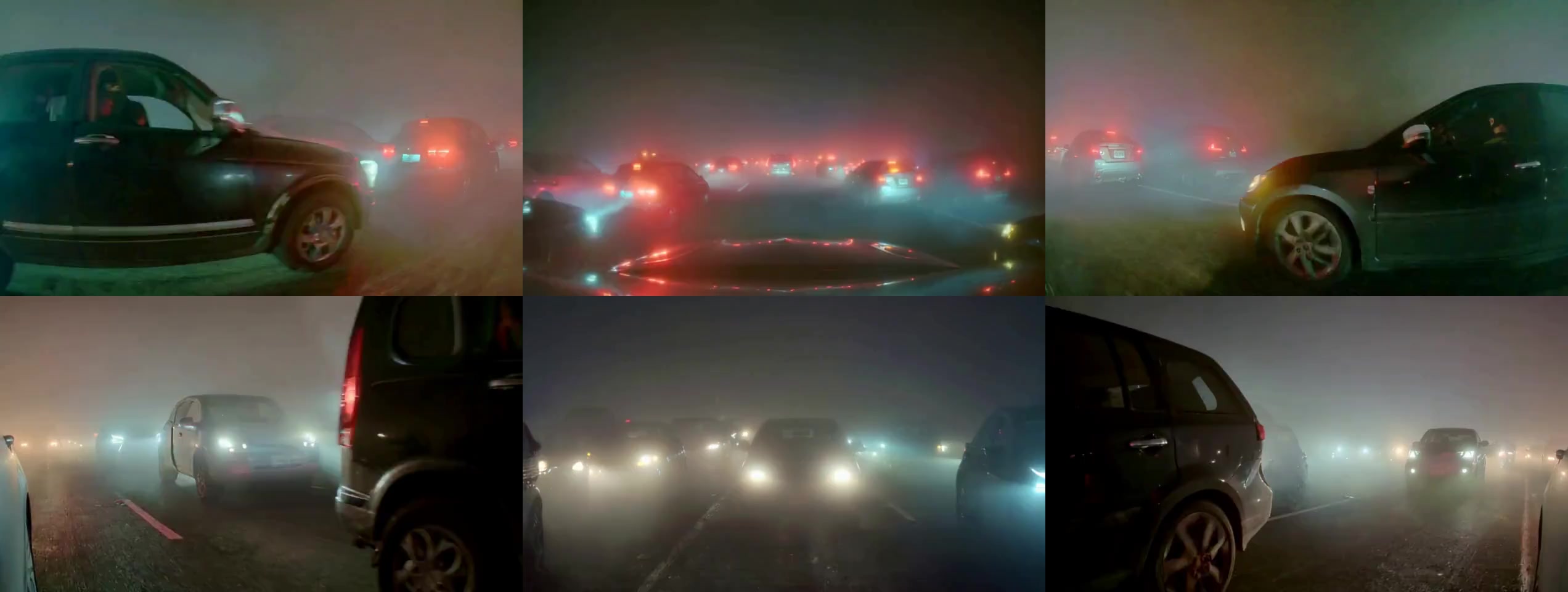} \\
        \includegraphics[width=0.49\textwidth]{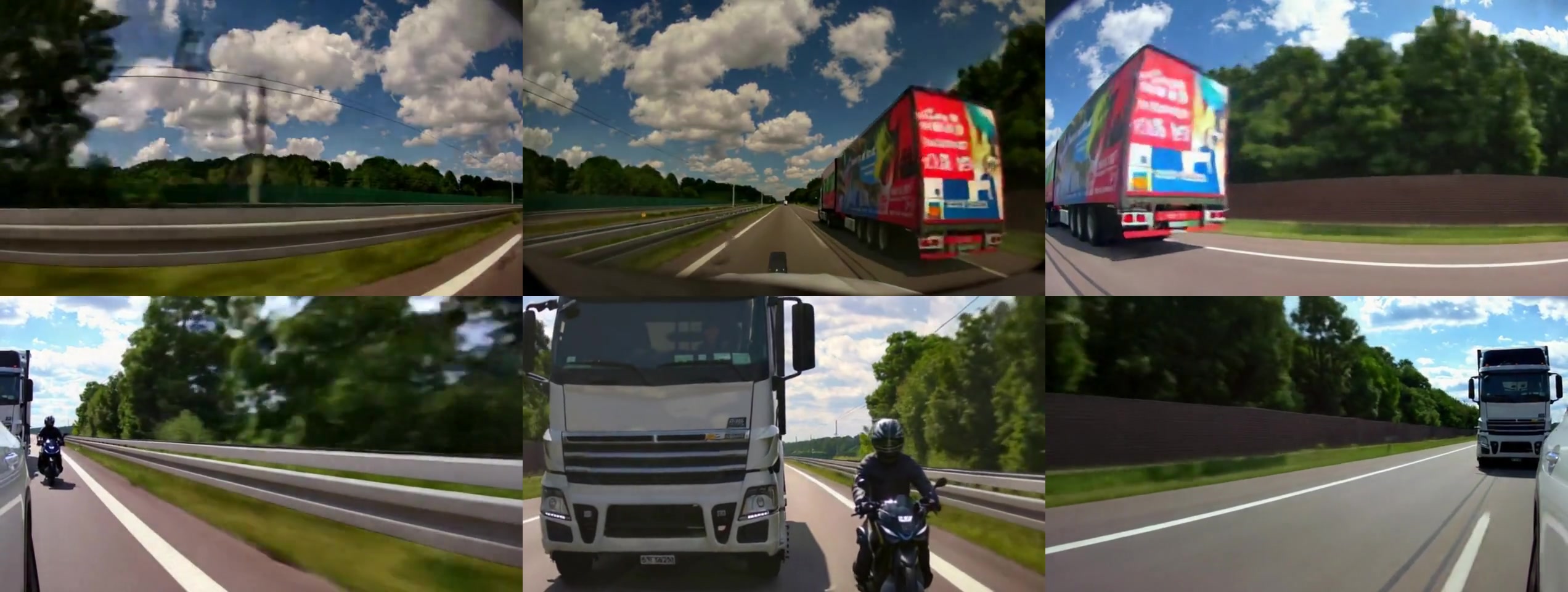} & \includegraphics[width=0.49\textwidth]{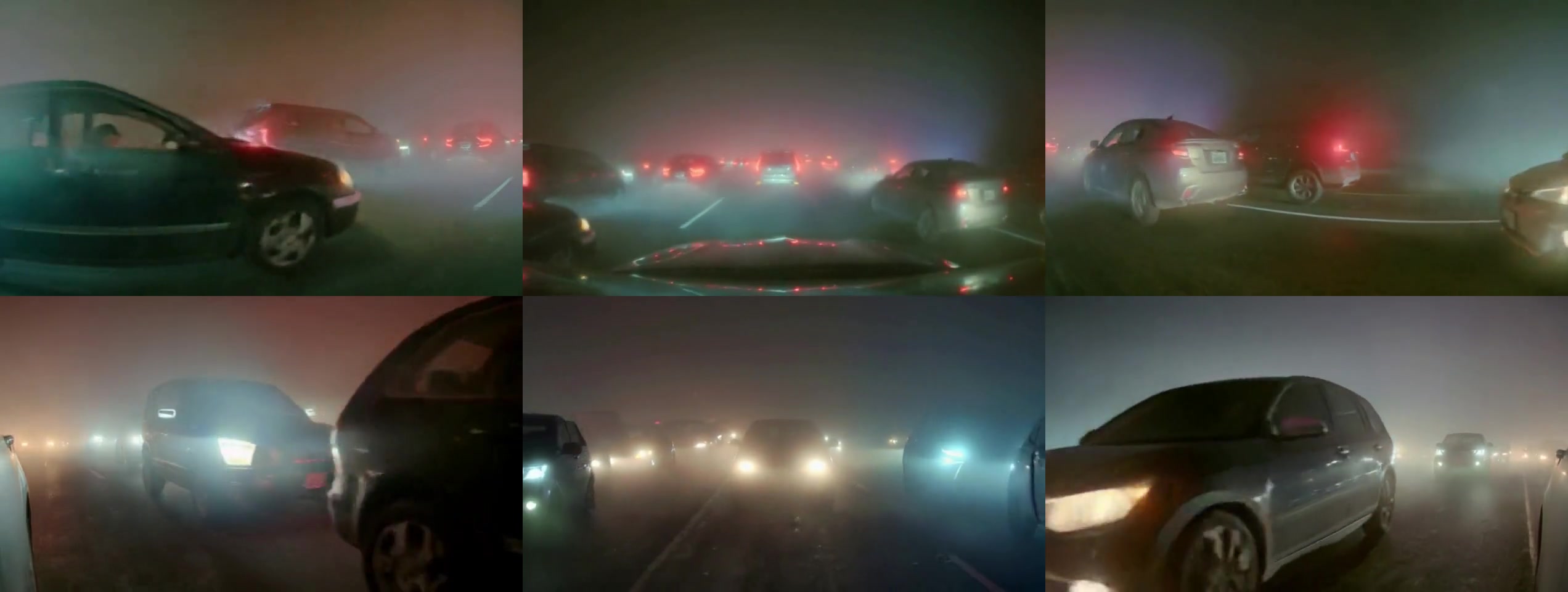} \\
        \includegraphics[width=0.49\textwidth]{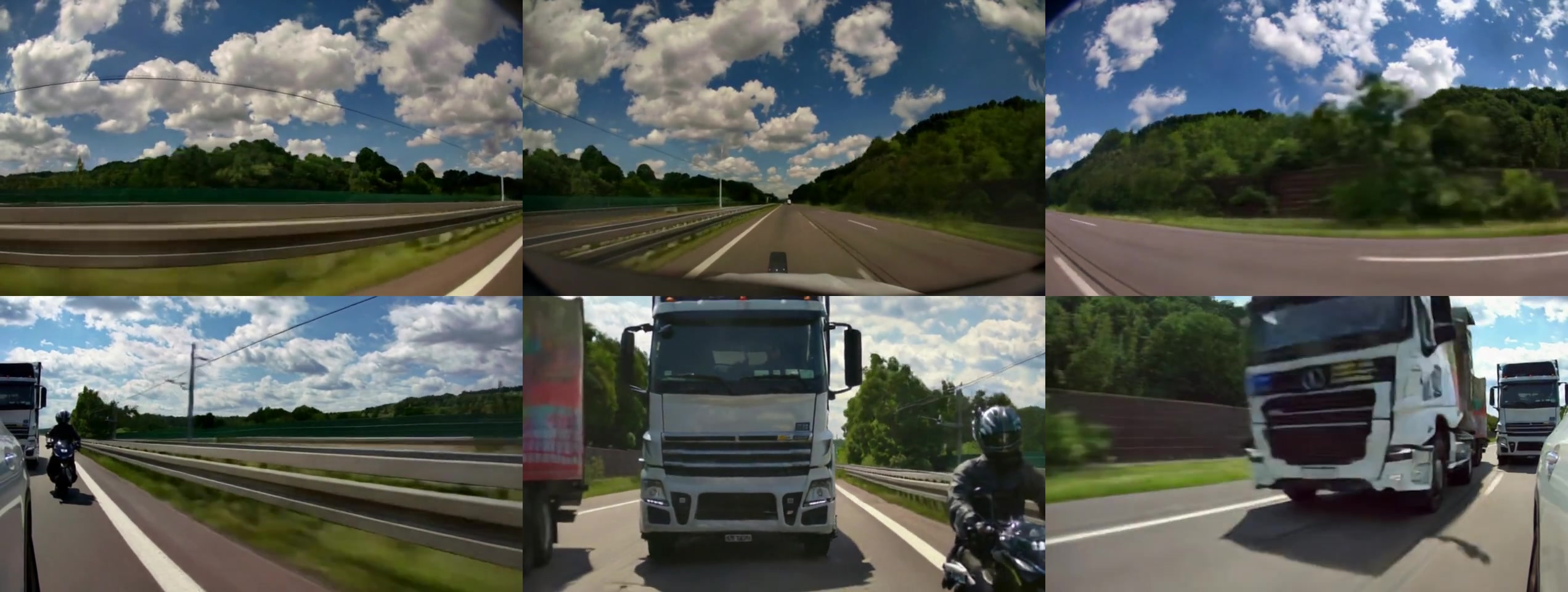} & \includegraphics[width=0.49\textwidth]{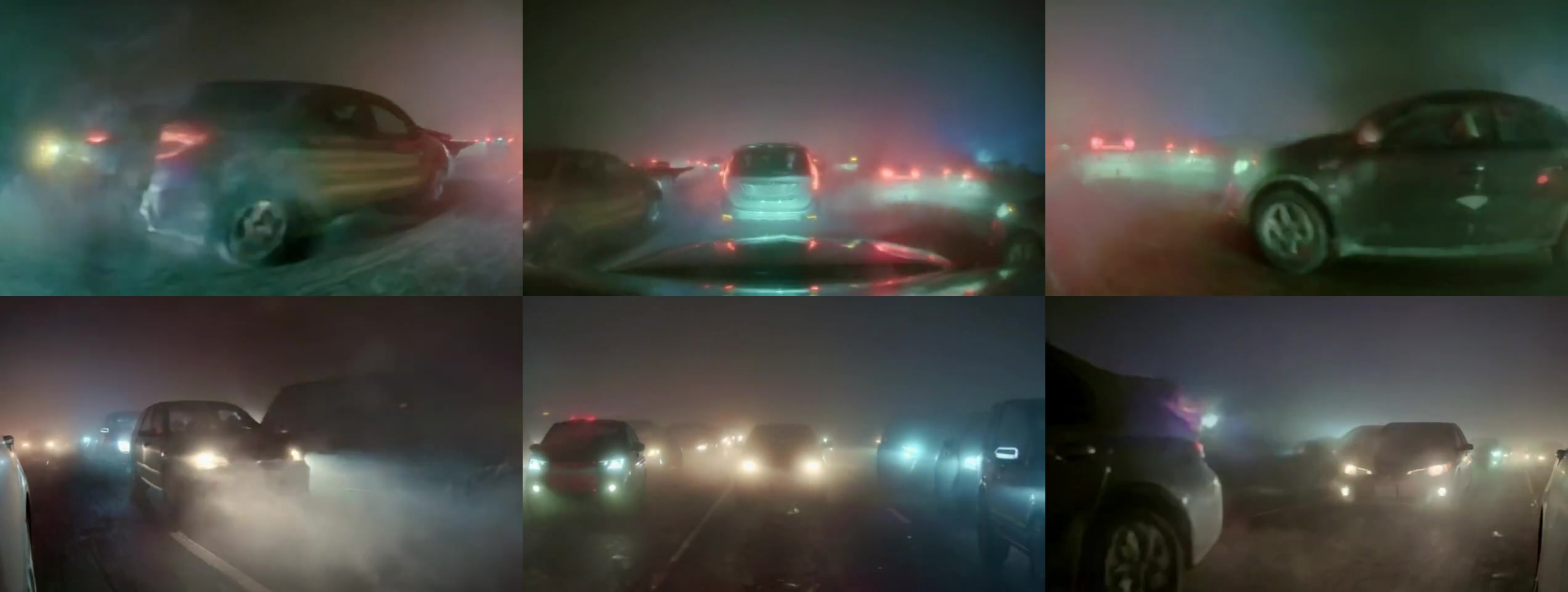} \\
        \begin{multicoltextblockhalf}
        \prompttext{Prompt}: \prompt{The video captures a highway scene with a white truck in the foreground, moving towards the camera. The truck has a large cargo area and is followed by a motorcyclist wearing a full-face helmet. The road is marked with white lines and has a metal guardrail on the right side. The sky is partly cloudy, and there are green trees and bushes visible on the roadside. The video is taken from a moving vehicle, as indicated by the motion blur and the changing perspective of the truck and motorcyclist.}
        \end{multicoltextblockhalf} &
        \begin{multicoltextblockhalf}
        \prompttext{Prompt}: \prompt{The footage shows a multi-car pile-up on a foggy highway. Visibility is severely reduced due to thick fog, with only the taillights of vehicles ahead visible. Suddenly, brake lights flash, and cars begin to swerve and stop abruptly. The highway is cluttered with stopped and crashed vehicles. The surroundings are obscured by fog, adding to the chaos and confusion of the scene.}
        \end{multicoltextblockhalf}
    \end{tabular}
    \caption{\textbf{Text-conditioned samples generated by Cosmos-Predict1-7B-Text2World-Sample-MultiView, extended to 8 seconds by Cosmos-Predict1-7B-Video2World-Sample-MultiView}. This figure visualizes all six camera views in a group, with each row corresponding to a specific timestamp. The left example depicts a highway scene where a motorcycle is riding alongside a large truck. The right example shows the ego car following a sedan as it takes a right turn in a heavy snowy day.}
    \label{fig:av_finetune1}
\end{figure*}

\begin{figure*}[ht]
    \centering
    \setlength{\tabcolsep}{0pt}
    \begin{tabular}{c@{\hspace{0.3cm}}c} %
        \includegraphics[width=0.49\textwidth]{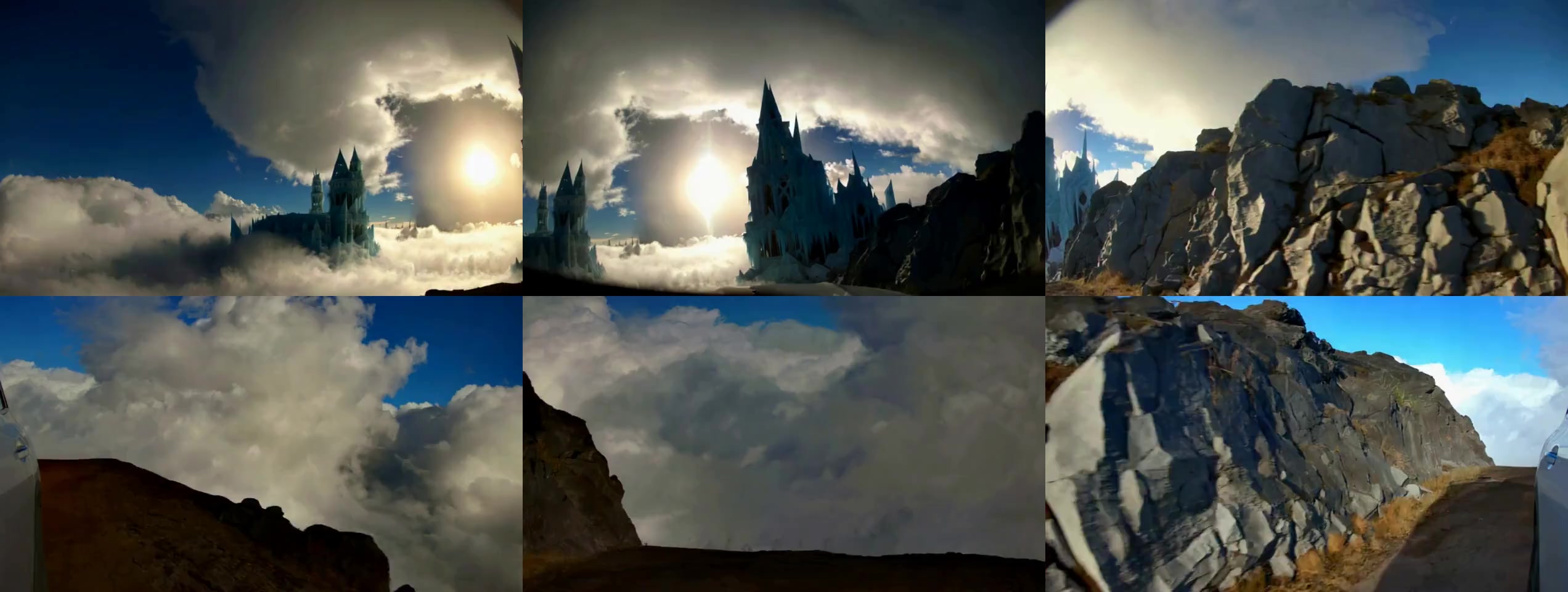} & \includegraphics[width=0.49\textwidth]{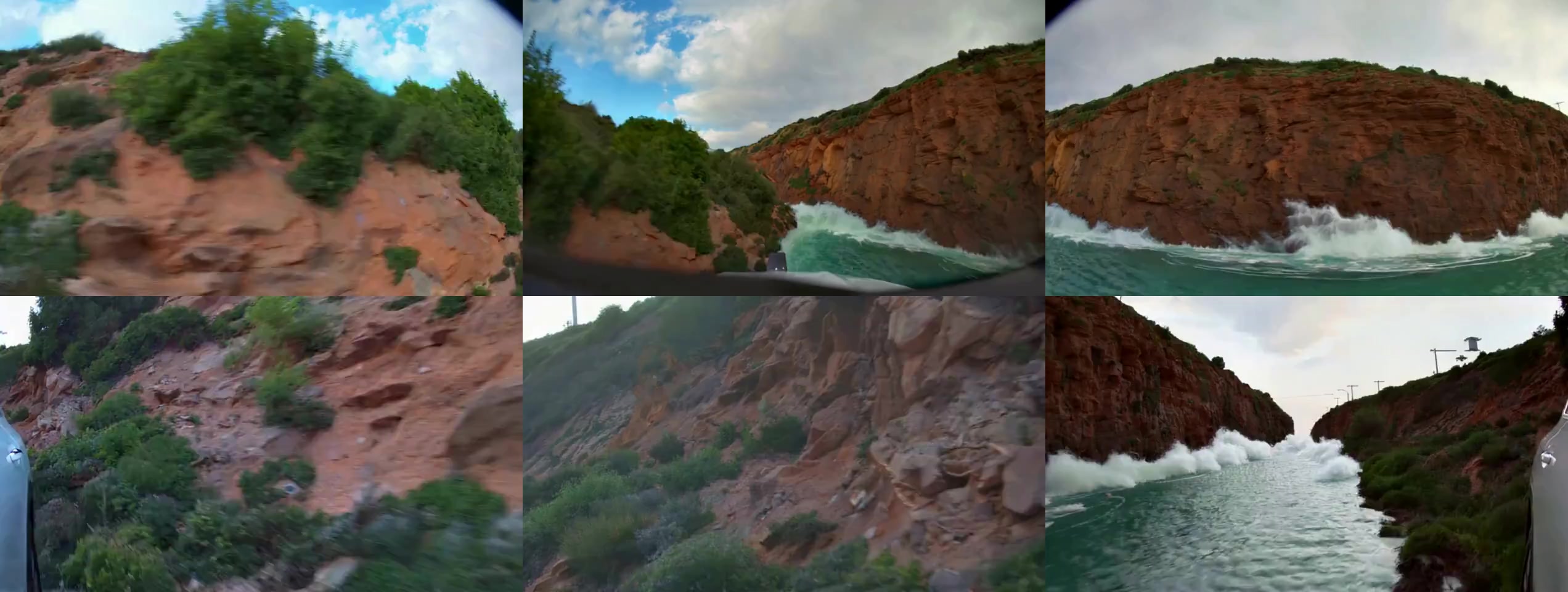} \\
        \includegraphics[width=0.49\textwidth]{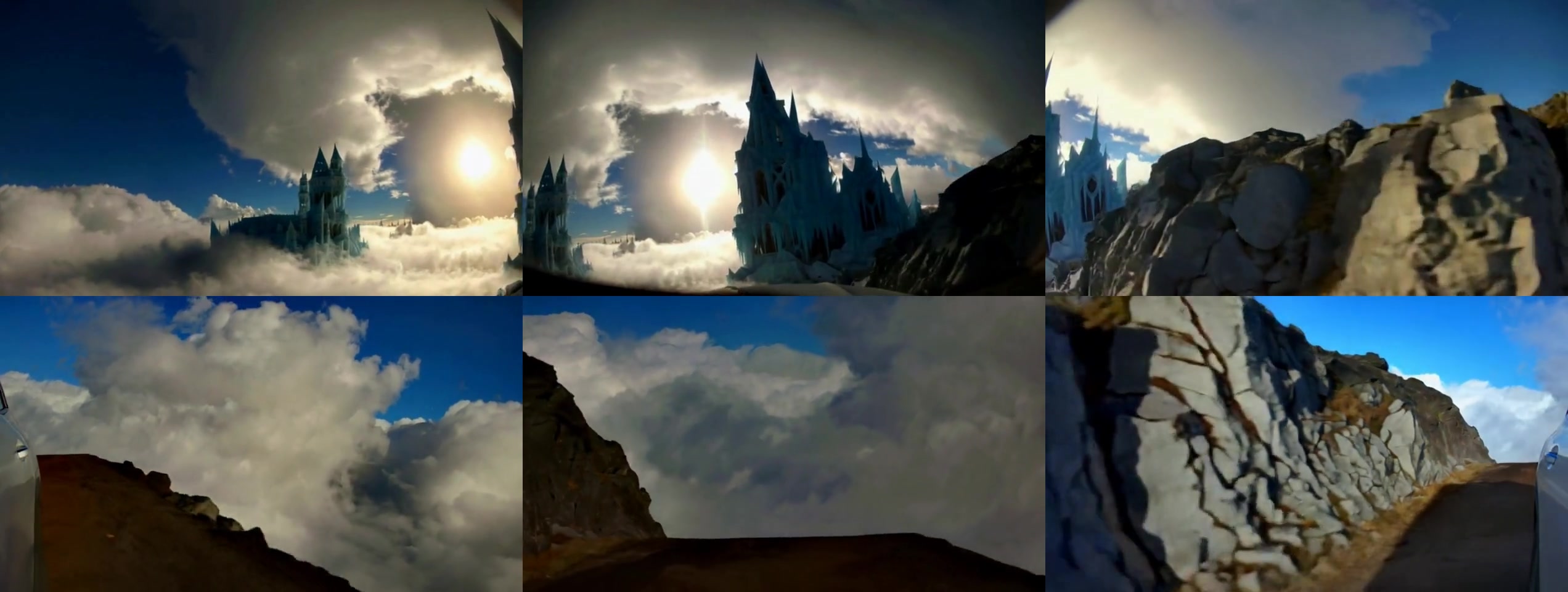} & \includegraphics[width=0.49\textwidth]{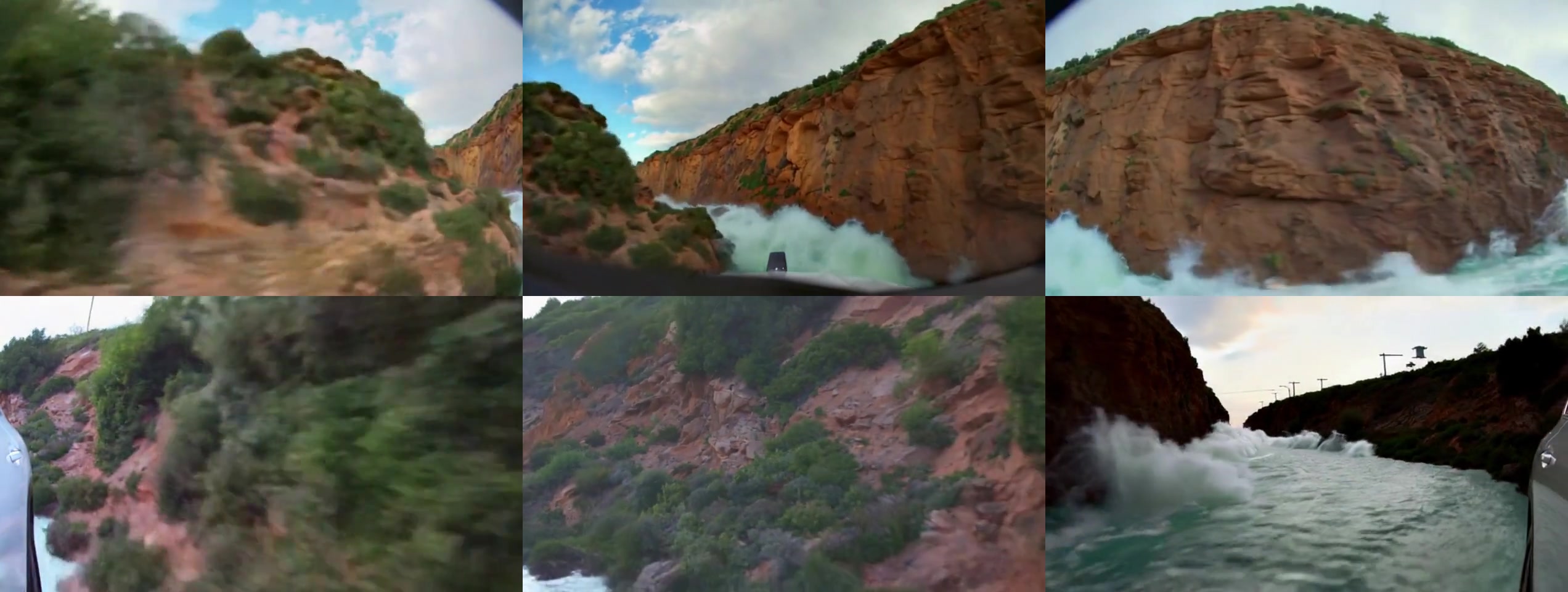} \\
        \includegraphics[width=0.49\textwidth]{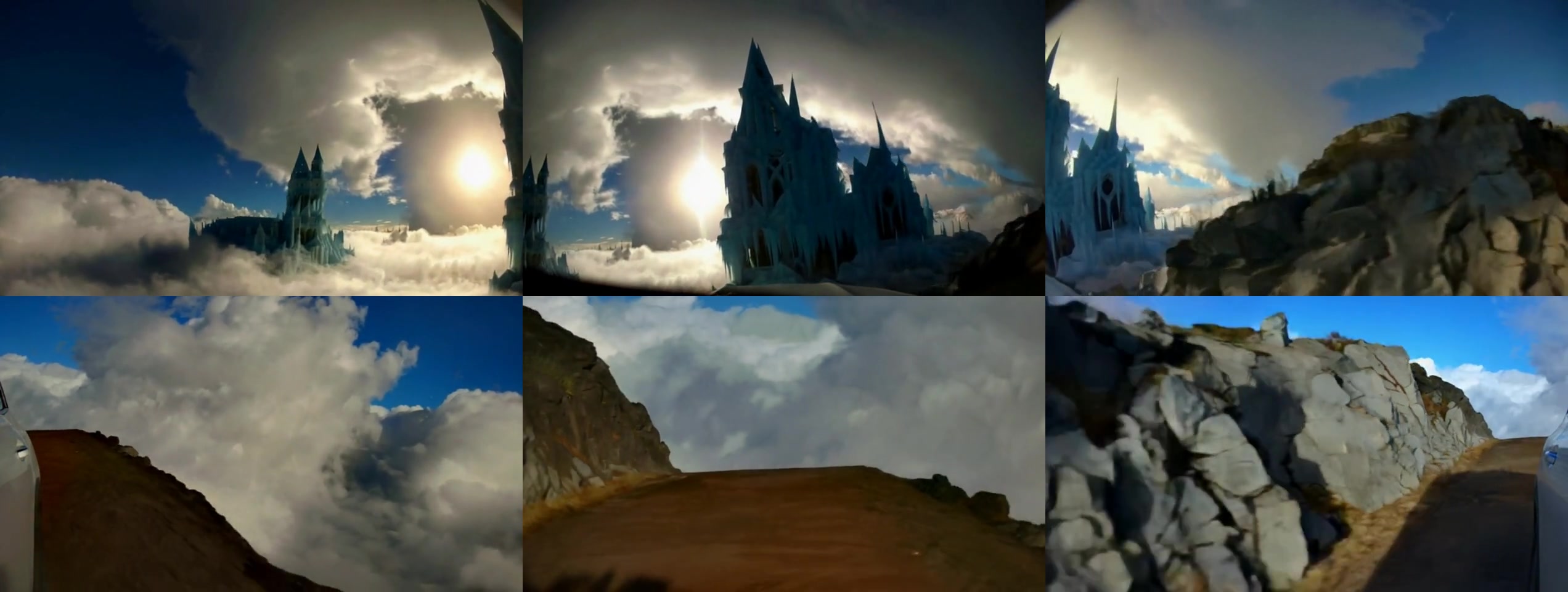} & \includegraphics[width=0.49\textwidth]{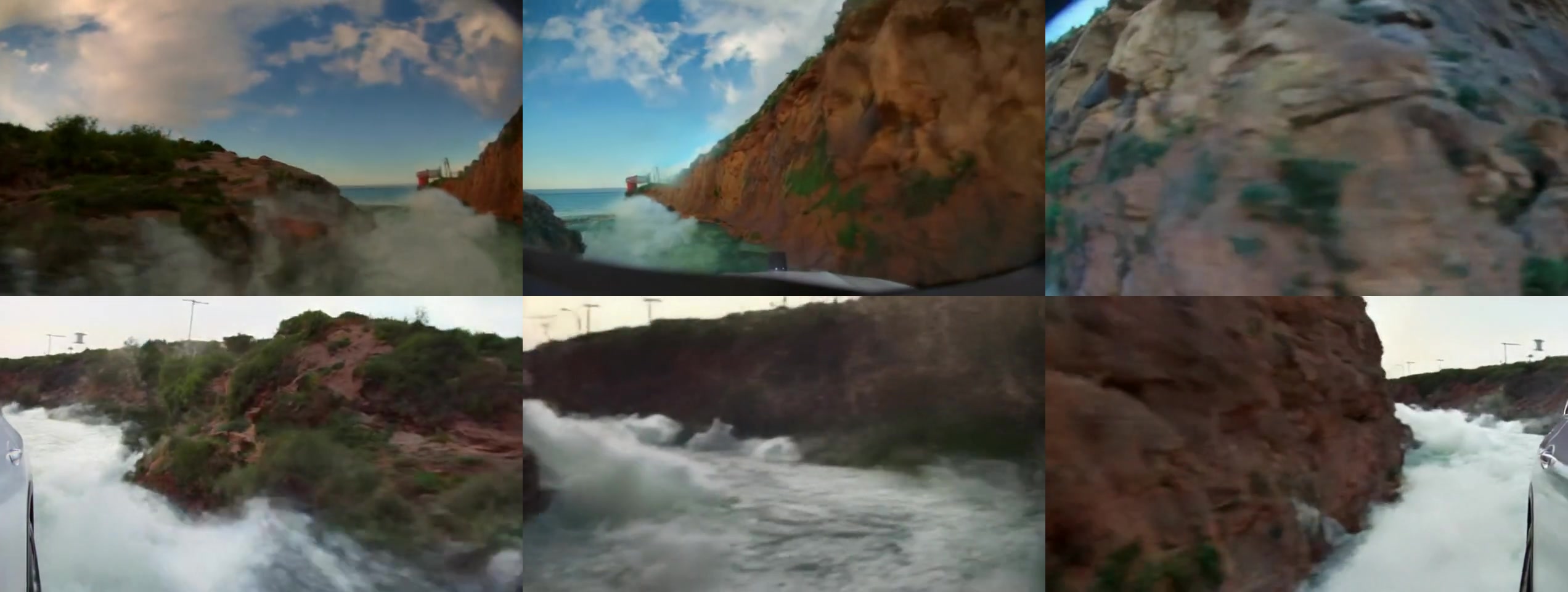} \\
        \begin{multicoltextblockhalf}
        \prompttext{Prompt}: \prompt{There are towering, intricately designed ice castle illuminated from within. Ahead, the sky showcases a vibrant sunset and a luminous moon positioned to the left of the castle, casting a blue hue across the scene. Fast-moving, dark, and dramatic clouds add to the otherworldly atmosphere. The 3D realistic art style focuses on lighting and texture, creating a striking visual effect.}
        \end{multicoltextblockhalf} &
        \begin{multicoltextblockhalf}
        \prompttext{Prompt}: \prompt{The footage captures a coastal area where the ocean meets a rugged, rocky cliff. Ahead, vibrant blue waves crash against the rocks, creating white foam. The cliff is a mix of brown and green hues, indicating vegetation and possibly moss or algae.}
        \end{multicoltextblockhalf}
    \end{tabular}
    \caption{\textbf{Text-conditioned samples generated by Cosmos-Predict1-7B-Text2World-Sample-MultiView are extended to 8 seconds by Cosmos-Predict1-7B-Video2World-Sample-MultiView}. The post-trained world model effectively preserves its generalization ability. In the left example, the ego car is driving towards an ice castle, while in the right example, the ego car is shown driving on a river.}
    \label{fig:av_finetune2}
\end{figure*}

\subsubsection{Evaluation}

\begin{table}[ht]
    \setlength{\tabcolsep}{1.5pt}
    \captionsetup{justification=centering}
    \caption{Evaluation on post-trained multi-view world models for multi-view driving video generation.}
    \centering
    \setlength{\tabcolsep}{19pt}    %
    \begin{tabular}{rcccc}
        \toprule
        & \multicolumn{2}{c}{\textbf{Generation Quality}} & \multicolumn{2}{c}{\textbf{Multi-View Consist.}} \\
        \cmidrule(r){2-3} \cmidrule(r){4-5}
        Method & FID $\downarrow$ & FVD $\downarrow$ & TSE $\downarrow$ & CSE $\downarrow$  \\
        \midrule
        VideoLDM-MultiView & 60.84 & 884.46 & 1.24 & 6.48  \\
        \midrule
        \makecell[r]{Cosmos-Predict1-7B-Text2World-\\Sample-MultiView} & \textbf{32.16} & \textbf{210.23} & 0.68 & 2.11 \\
        \midrule
        \makecell[r]{Cosmos-Predict1-7B-Text2World-\\Sample-MultiView-TrajectoryCond} &- & - & \textbf{0.59}  & \textbf{2.02} \\
        \midrule
       Real Videos (Reference) & -  & - & 0.69  & 1.71 \\
        \bottomrule
    \end{tabular}
    \label{tab::av-result}
\end{table}

\begin{table}[ht]
    \setlength{\tabcolsep}{4.5pt}
    \captionsetup{justification=centering}
    \caption{Trajectory consistency evaluation on post-trained multi-view world models for multi-view driving video generation. The numbers of TAE are scaled by $10^2$ for convenience, and the unit for TFE is cm.}
    \centering
    \setlength{\tabcolsep}{12.5pt}  %
    \begin{tabular}{rcccc}
        \toprule
        Method& TAE-ATE $\downarrow$ & TAE-RPE-R $\downarrow$ & TAE-RPE-t $\downarrow$ & TFE $\downarrow$  \\
        \midrule
        VideoLDM-MultiView & 0.88 & 22.94 & 0.77 & - \\
        \midrule
        \makecell[r]{Cosmos-Predict1-7B-Text2World-\\Sample-MultiView} & 0.77 & \textbf{4.25} & 0.29 & - \\
        \midrule
        \makecell[r]{Cosmos-Predict1-7B-Text2World-\\Sample-MultiView-TrajectoryCond} & \textbf{0.54} & 4.31 & \textbf{0.18} & \textbf{20.20} \\
        \midrule
       Real Videos (Reference) & 0.49 & 4.60 & 0.14 & 13.49 \\
        \bottomrule
    \end{tabular}
    \label{tab::av-result-trajectory}
\end{table}

\begin{figure*}[ht]
    \centering
    \setlength{\tabcolsep}{0.1pt}
    \renewcommand{\arraystretch}{0.2}
    \begin{tabular}{cccccc} %
        \footnotesize{Visualized Trajectory Input} &  & \footnotesize{Frame 25} & \footnotesize{Frame 50} & \footnotesize{Frame 75} & \footnotesize{Frame 100} \\
        \includegraphics[width=0.2\textwidth]{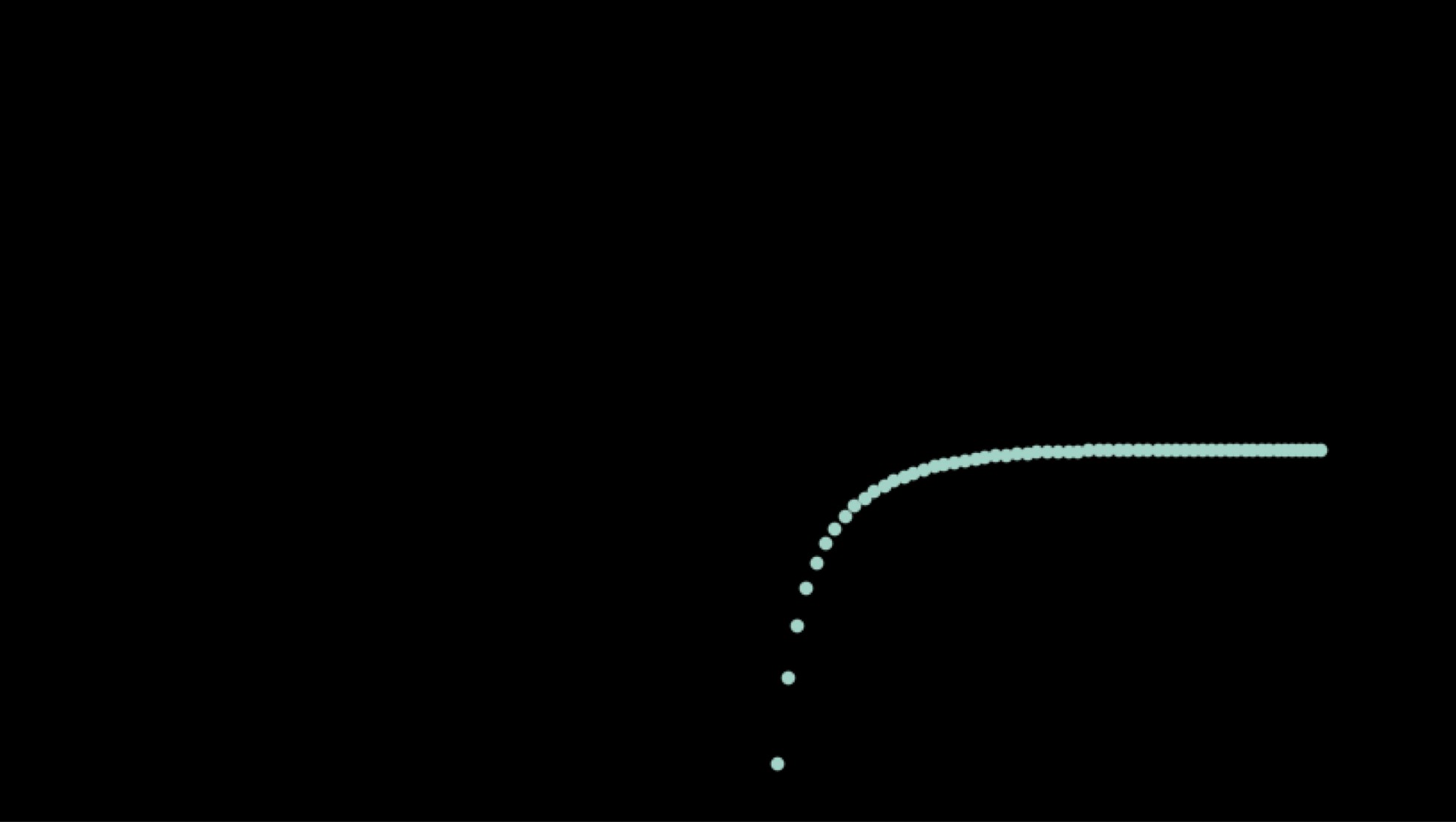} & &
        \includegraphics[width=0.2\textwidth]{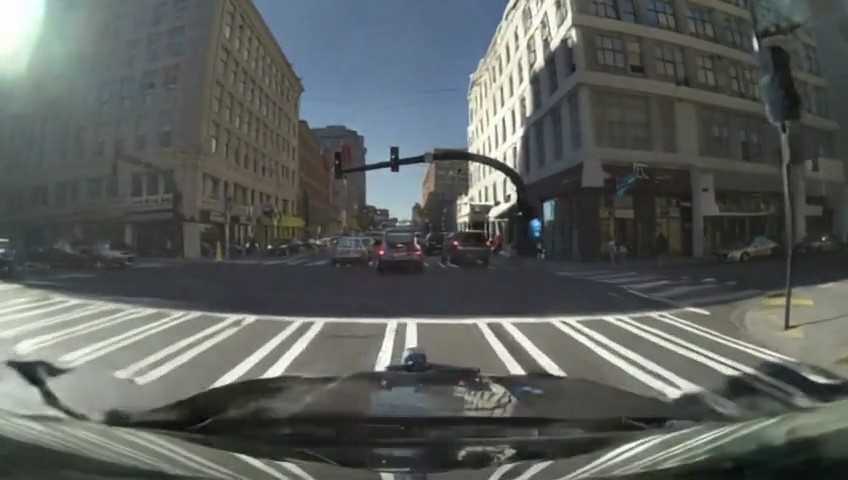} &
        \includegraphics[width=0.2\textwidth]{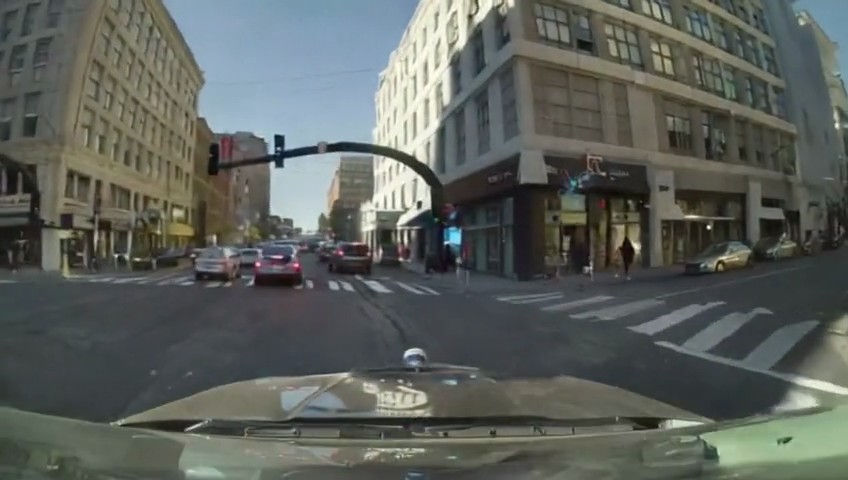} &
        \includegraphics[width=0.2\textwidth]{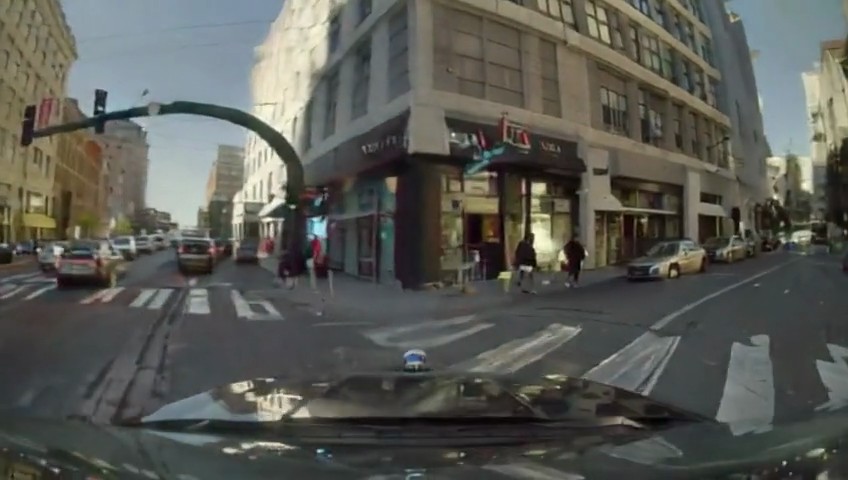} &
        \includegraphics[width=0.2\textwidth]{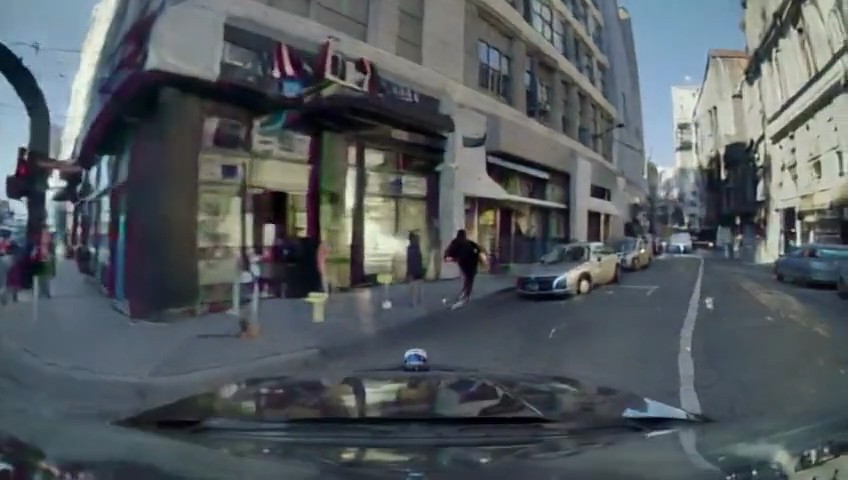} \\
        \includegraphics[width=0.2\textwidth]{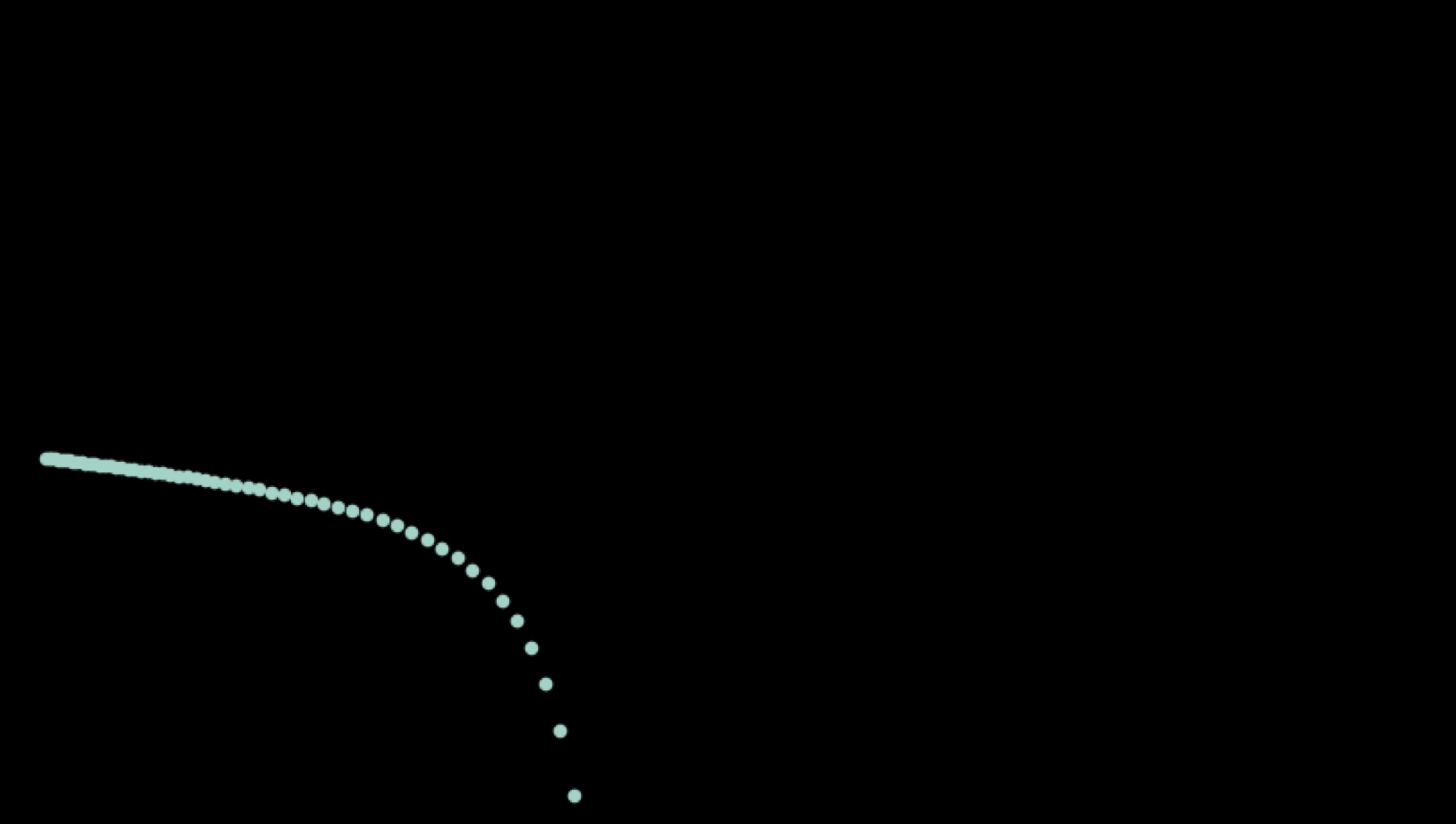} & &
        \includegraphics[width=0.2\textwidth]{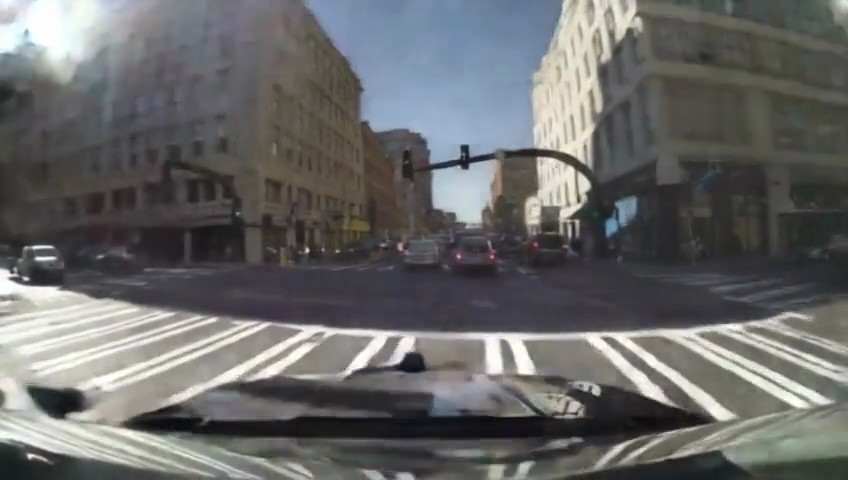} &
        \includegraphics[width=0.2\textwidth]{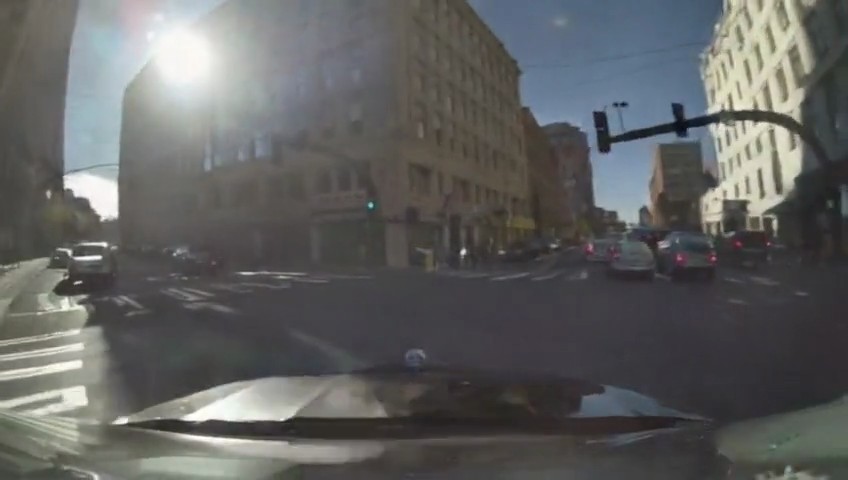} &
        \includegraphics[width=0.2\textwidth]{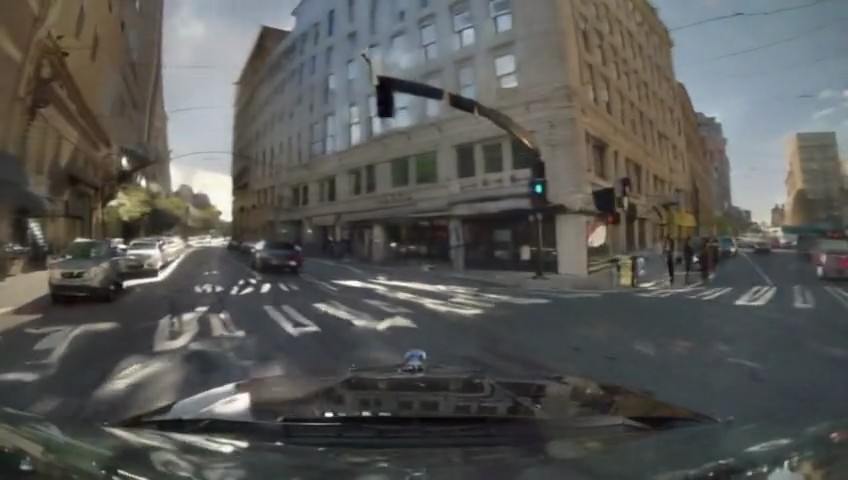} &
        \includegraphics[width=0.2\textwidth]{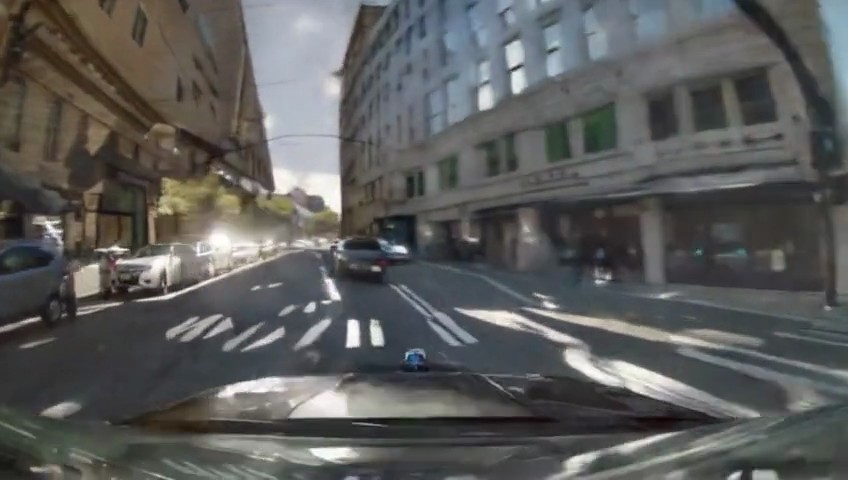} \\
        \includegraphics[width=0.2\textwidth]{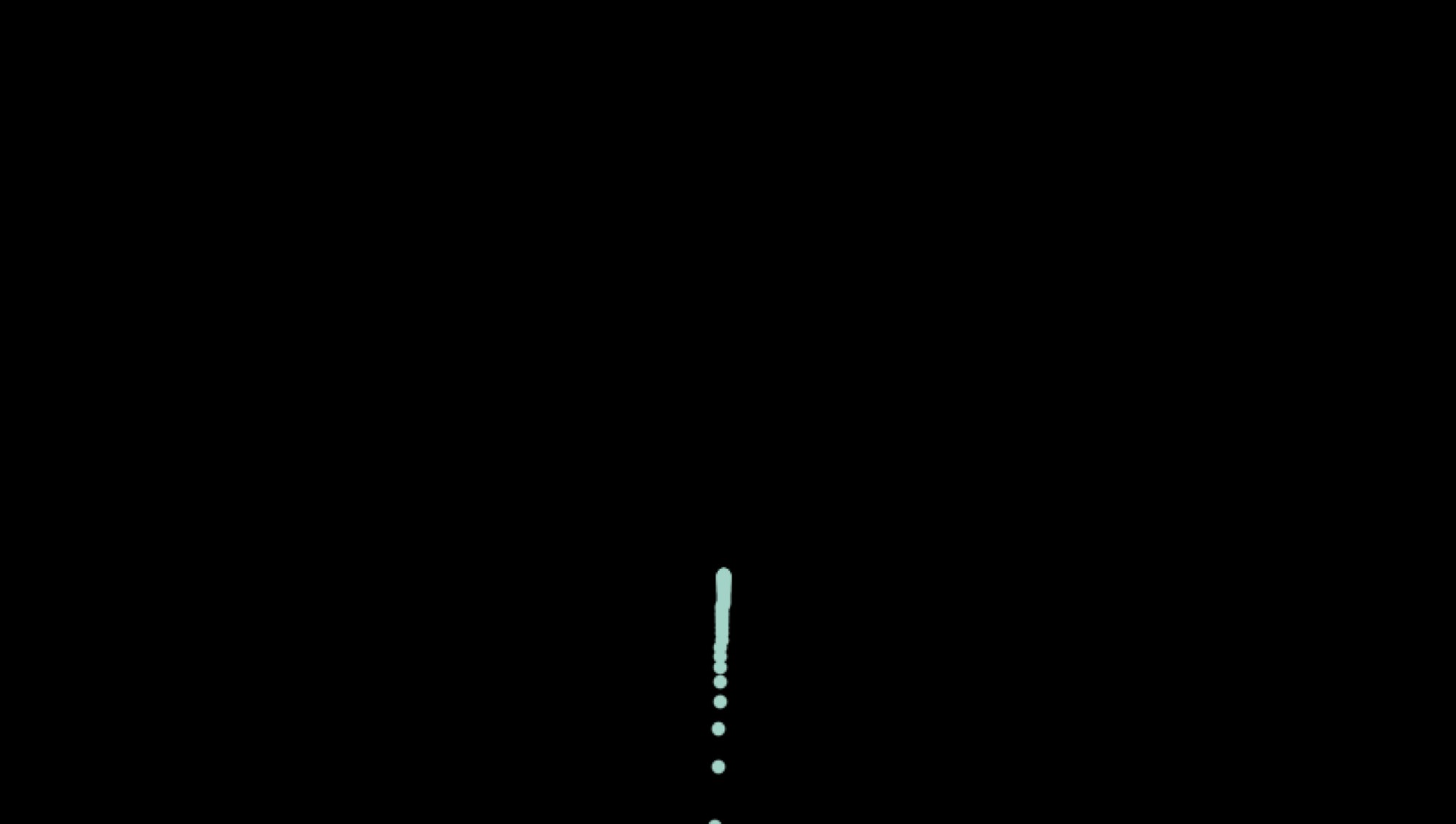} & &
        \includegraphics[width=0.2\textwidth]{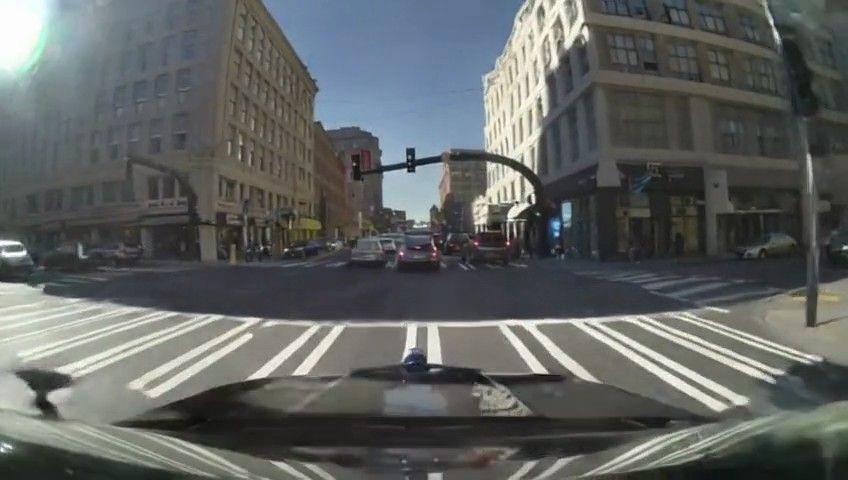} &
        \includegraphics[width=0.2\textwidth]{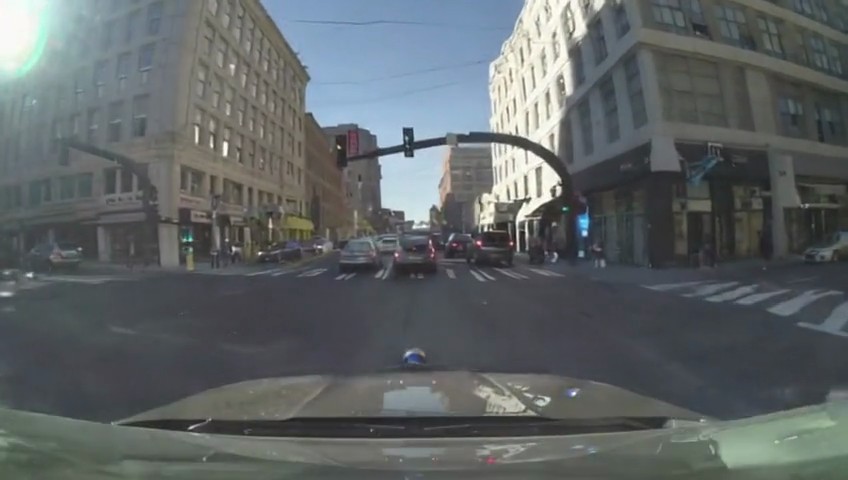} &
        \includegraphics[width=0.2\textwidth]{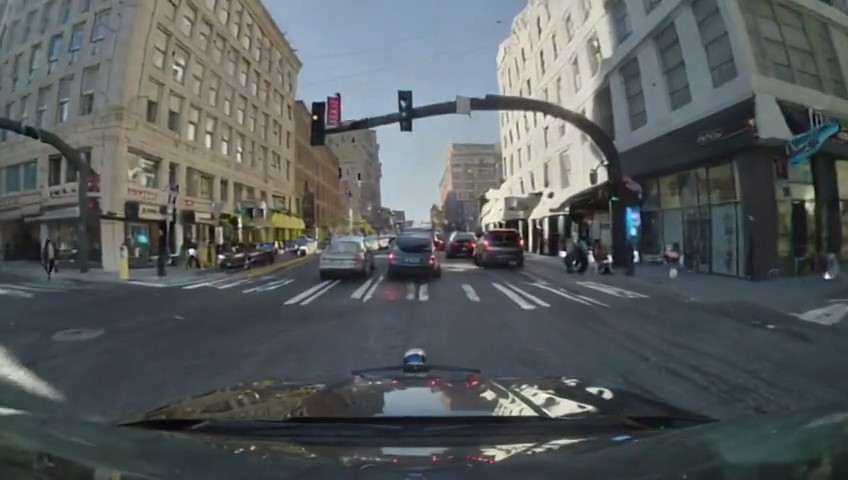} &
        \includegraphics[width=0.2\textwidth]{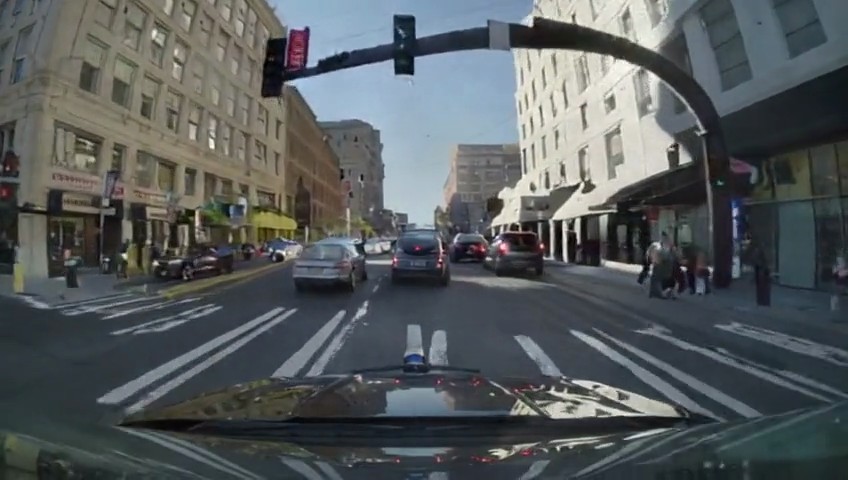} \\
        \\\\
        \includegraphics[width=0.2\textwidth]{images/5_3_3_results/right_turn.jpg} & &
        \includegraphics[width=0.2\textwidth]{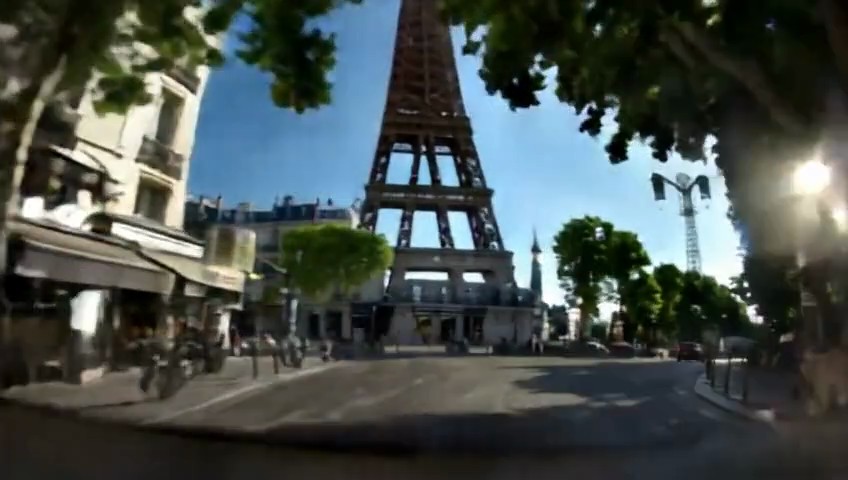} &
        \includegraphics[width=0.2\textwidth]{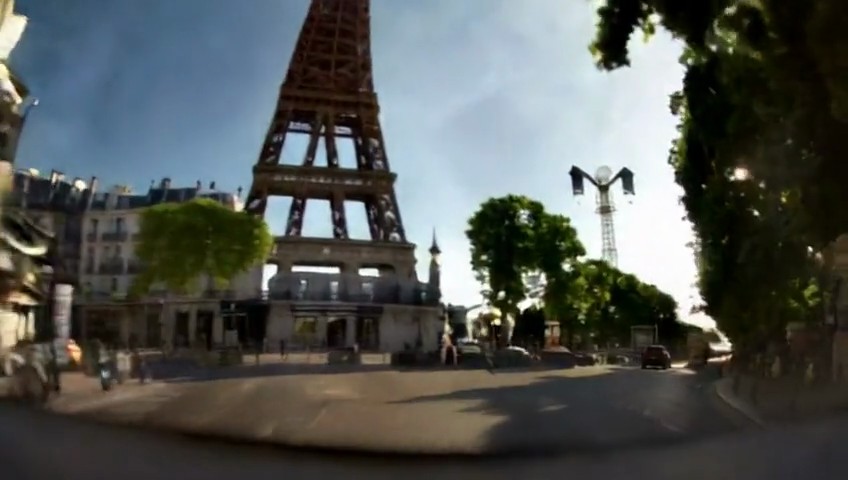} &
        \includegraphics[width=0.2\textwidth]{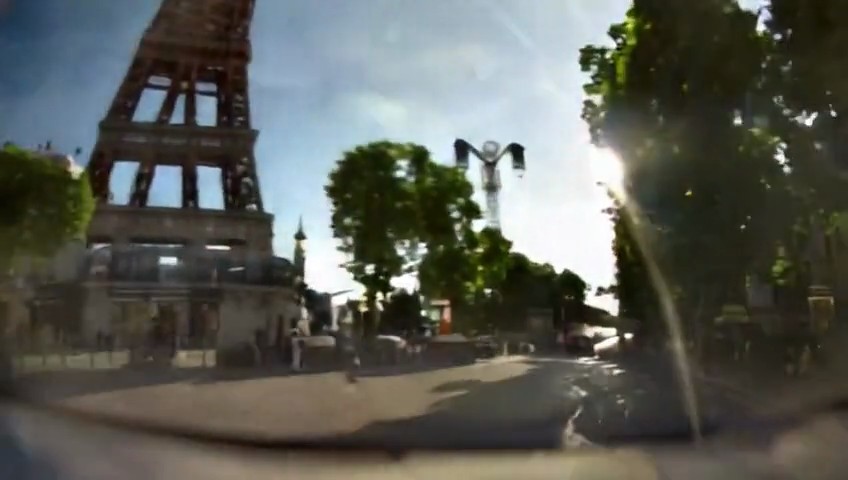} &
        \includegraphics[width=0.2\textwidth]{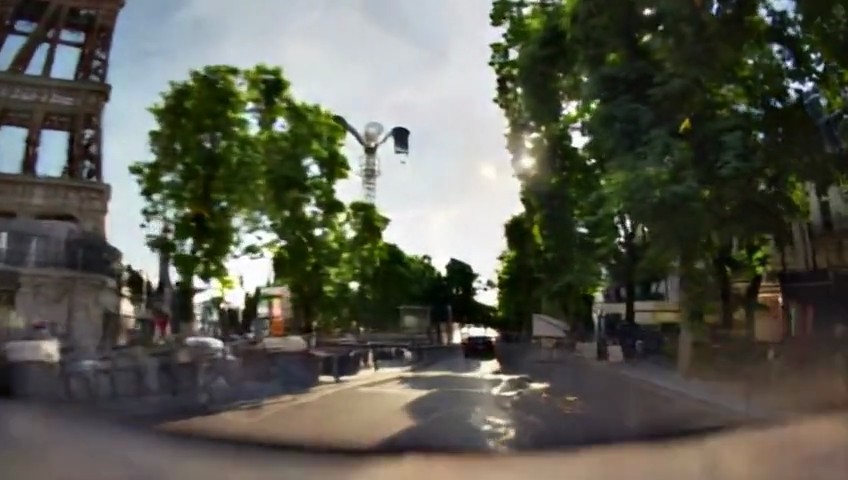} \\
        \includegraphics[width=0.2\textwidth]{images/5_3_3_results/left_turn.jpg} & &
        \includegraphics[width=0.2\textwidth]{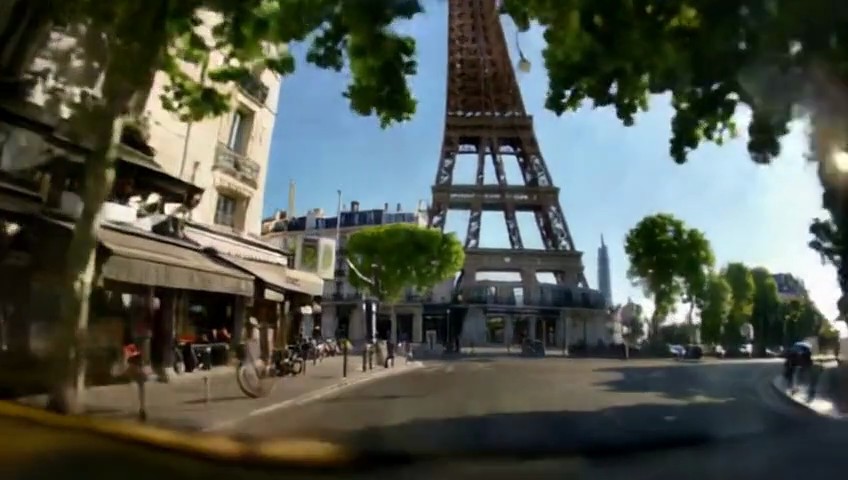} &
        \includegraphics[width=0.2\textwidth]{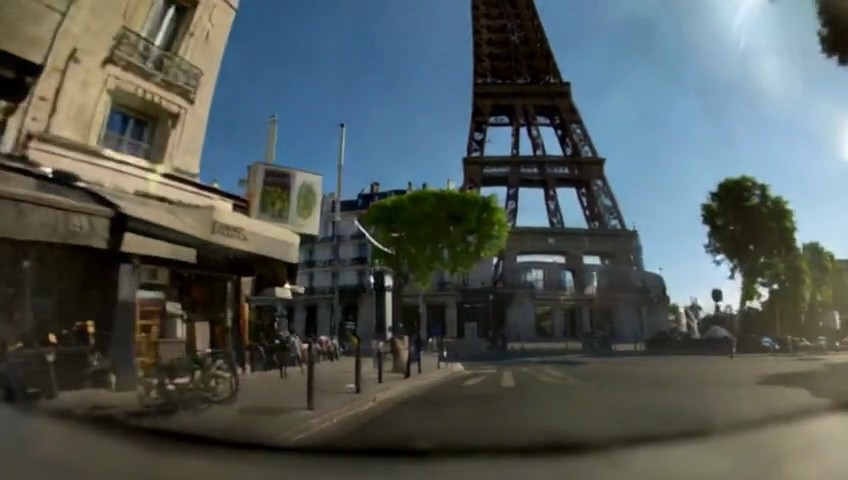} &
        \includegraphics[width=0.2\textwidth]{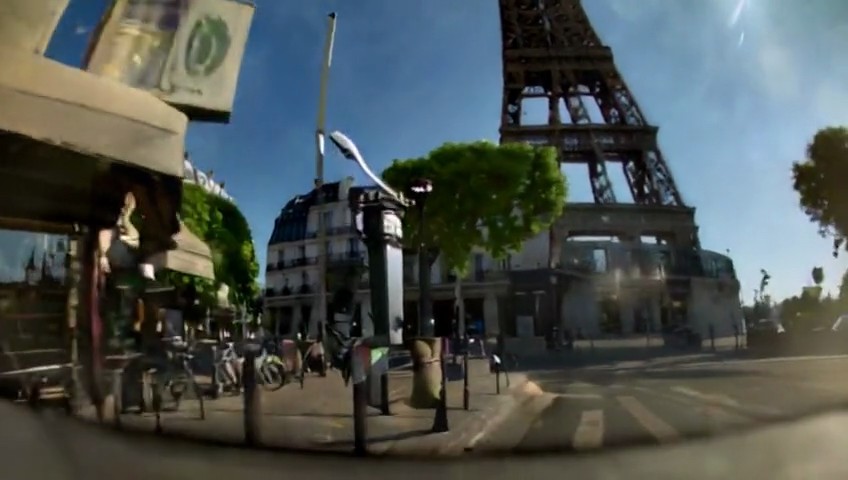} &
        \includegraphics[width=0.2\textwidth]{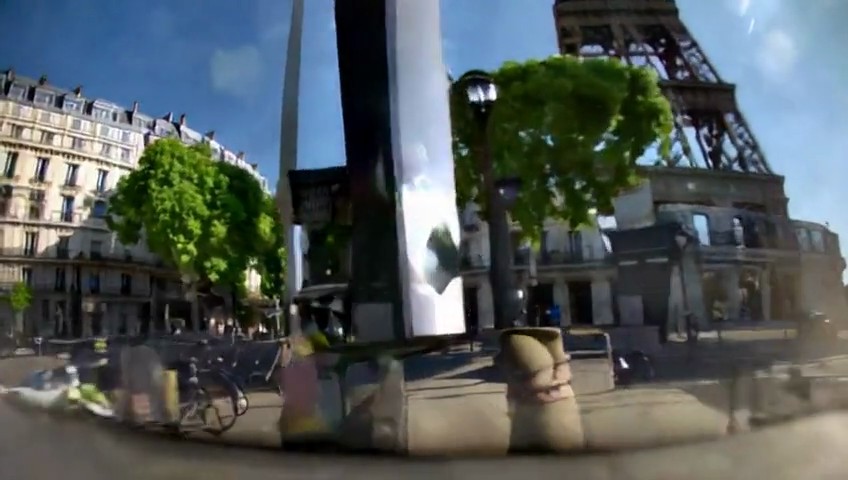} \\
        \includegraphics[width=0.2\textwidth]{images/5_3_3_results/straight.jpg} & &
        \includegraphics[width=0.2\textwidth]{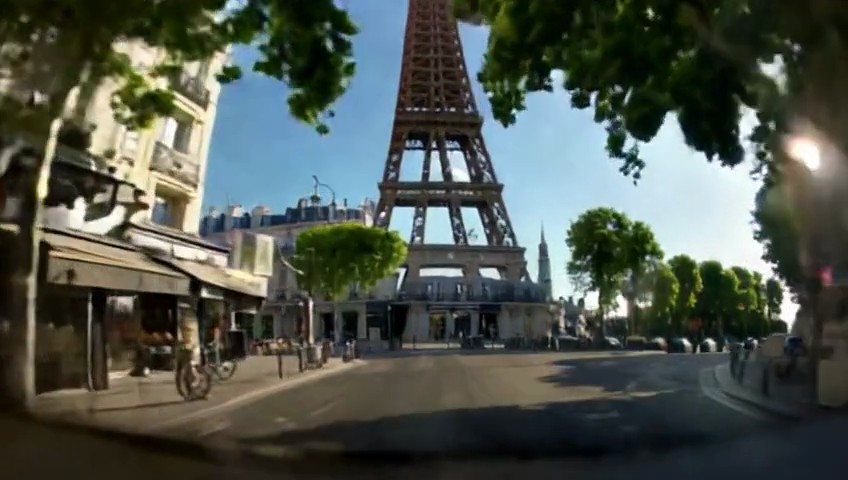} &
        \includegraphics[width=0.2\textwidth]{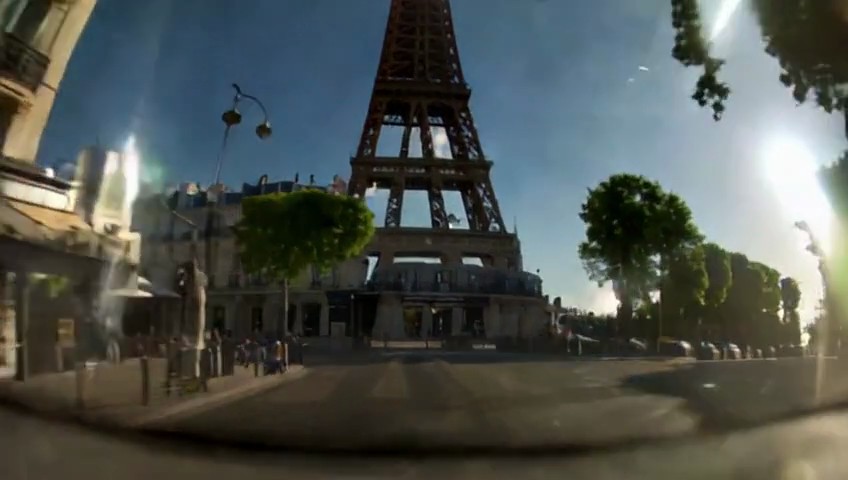} &
        \includegraphics[width=0.2\textwidth]{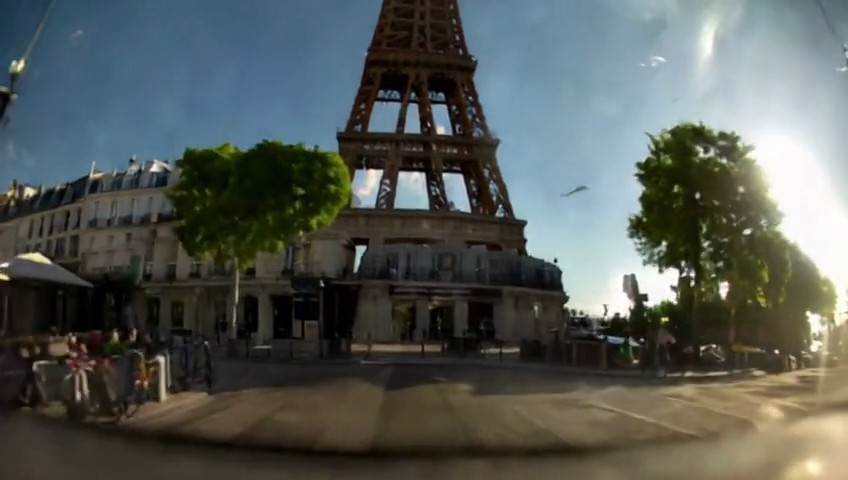} &
        \includegraphics[width=0.2\textwidth]{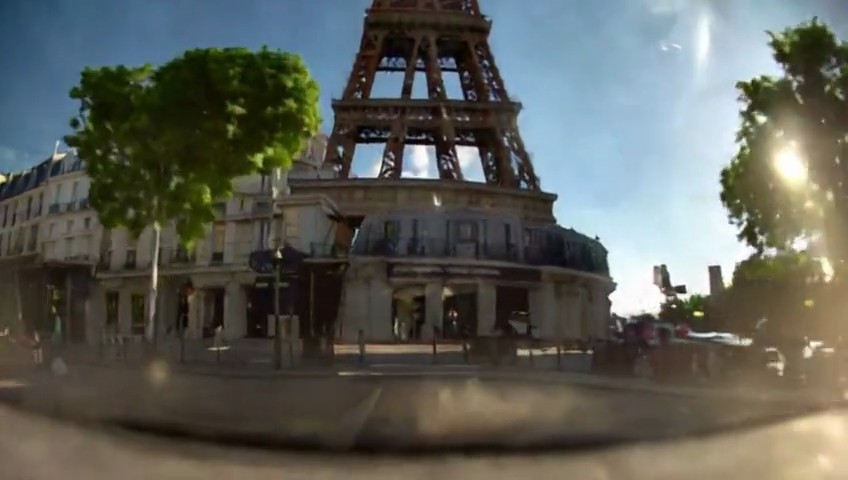}
    \end{tabular}
    \caption{\textbf{Trajectory-conditioned generated samples from Cosmos-Predict1-7B-Text2World-Sample-MultiView-TrajectoryCond}. Given the trajectory inputs on the left-most column, we generate multiview videos that follow the given trajectory. We visualize the front camera view in this figure.}
    \label{fig:traj_fig}
\end{figure*}

We first present text-conditioned qualitative results in \cref{fig:av_finetune1}. Using Cosmos-Predict1-7B-Text2World-Sample-MultiView, we generate a 57-frame video with six views, which is then extended to 201 frames using the Cosmos-Predict1-7B-Video2World-Sample-MultiView model. In \cref{fig:av_finetune2}, we demonstrate how the pre-trained world model enhances generalization, enabling the generation of rare or out-of-domain scenes from the RDS dataset, such as driving on a river. Lastly, \cref{fig:traj_fig} showcases the results from Cosmos-Predict1-7B-Text2World-Sample-MultiView-TrajectoryCond, where the ego car accurately follows the input trajectory.

For quantitative results, as a baseline, we followed the same fine-tuning recipe to fine-tune VideoLDM~\citep{Blattmann2023Align} to derive a multi-view world model called VideoLDM-MultiView. We use a set of evaluation metrics measuring video generation quality, multi-view consistency, and trajectory following accuracy. To evaluate video generation quality, we use 1000 samples to compute the scores. For consistency-related metrics, to better understand different models' behaviors under different scenarios, we categorize the ground-truth trajectories into four types: moving forward, turning left, turning right, and others (including static or complex movements). For each category, we gathered 200 samples and their corresponding prompts and conditions, totaling 800 samples. Below, we provide detailed descriptions of the metrics and the results.

\noindent \textbf{Generation quality.} We utilize Fréchet Inception Distance (FID)~\citep{FID} and Fréchet Video Distance (FVD)~\citep{FVD} to measure the quality of the generated videos relative to the real ones. We first calculate a score per view by extracting 16 frames from each video. We then report the average score across all views per method. As shown in~\cref{tab::av-result}, we find both Cosmos-Predict1-7B-Text2World-Sample-MultiView and Cosmos-Predict1-7B-Text2World-Sample-MultiView-TrajectoryCond significantly outperform VideoLDM-MultiView on both metrics, demonstrating the superior quality of our pre-trained 7B Diffusion-based WFM over the VideoLDM-MultiView baseline.

\noindent \textbf{Multi-view consistency.} We use an extended version of the Sampson error~\citep{sampson1982fitting,hartley2003multiple} formulated in \cref{sec::eval_3d_consistency} to quantify the geometry consistency of the generated multi-view videos. As the ground-truth videos in our RDS dataset share similar fisheye camera intrinsic parameters, we use the median calibration to undistort the keypoints to a regular pinhole camera with a uniform size of $960 \times 540$ and 120 degrees of horizontal FoV. Under this setting, two metrics are computed for the generated multi-view video:
\begin{enumerate}
    \item \emph{Temporal Sampson Error (\textbf{TSE})} measures whether the content generated for each camera is consistent over time. It is the median Sampson error of adjacent frames for each of the views.
    \item \emph{Cross-view Sampson Error (\textbf{CSE})} measures whether multi-view consistency is preserved over time. It is the Sampson error across different generated views averaged in time. The fundamental matrix used in CSE is estimated using the keypoints accumulated across all temporal frames.
\end{enumerate}

As shown in \cref{tab::av-result}, we find that both Cosmos-Predict1-7B-Text2World-Sample-MultiView and Cosmos-Predict1-7B-Text2World-Sample-MultiView-TrajectoryCond render better multi-view geometry consistency over VideoLDM-MultiView. The overall geometric plausibility of the generated videos is much better for a world model fine-tuned from our WFM. We also note that, with the trajectory control condition, such consistency is further improved thanks to the explicit 3D guidance as Cosmos-Predict1-7B-Text2World-Sample-MultiView-TrajectoryCond is ranked the best.

\noindent \textbf{Trajectory consistency: Trajectory Agreement Error (TAE).} We design a robust multi-view camera pose estimation pipeline similar to the one used in \citet{liang2024feed} based on the formulation of \citet{teed2021droid}. Such a pose estimation pipeline features an online dynamic mask generation module and a highly efficient dense bundle adjustment module, reaching a robust and real-time performance for estimating multi-view camera poses. We use this pipeline to estimate the camera poses of the front camera, separately using two multi-view camera configurations that consider ``front + left-front'' cameras and ``front + right-front'' cameras. We then calculate their trajectory errors to show their agreement, reflecting the consistency of the multi-view generation. Specifically, we compute the Absolute Trajectory Error (ATE) and Relative Pose Error for both the translational component (RPE-t) and the rotational component (RPE-R). We normalize the length of the trajectories to 1.0 for fair comparison and exclude the cases with minor camera movements (\eg, a car stopped at a red light).

As shown in \cref{tab::av-result-trajectory}, the results echo the findings from the multi-view geometry consistency, where the trajectory consistency of the world models fine-tuned from the Cosmos WFM is much better than that from the VideoLDM-MultiView. We note that the post-trained Cosmos world models have trajectory consistency that is close to real-world videos.

\noindent \textbf{Trajectory consistency: Trajectory Following Error (TFE).} Furthermore, for the model where we have trajectory control conditions being fed into the model, we use the same camera pose estimation pipeline used above to compute the poses of the front camera using multi-view information and compare the predicted trajectory with respect to the ground truth trajectory condition. This measures how well the model follows the given trajectory path. As shown in \cref{tab::av-result-trajectory}, the trajectory error estimated using the generated videos from our Cosmos post-trained world models is only $<$7cm less precise than the ground-truth oracle. Such a considerably slight margin shows that our model is able to accurately follow the given trajectory path, which is crucial for training autonomous driving agents.

\noindent \textbf{Objects tracking consistency.} Finally, we applied object detection and tracking using YOLOv11x~\citep{khanam2024yolov11} on the generated 8-second videos. Human annotators were tasked with identifying instances where the tracking algorithm misinterpreted physically impossible scenarios, such as two distinct objects (\eg, a person and a car) merging incorrectly into a single tracked entity. To evaluate this, we provided annotators with a random sample of 20 generated videos containing 157 objects. Remarkably, none of the 157 objects exhibited any physically impossible scenarios, demonstrating the physical consistency and object permanence of our generated driving videos.

\section{Guardrails}\label{sec::guardrails}

\begin{figure}[h!]
    \centering
    \includegraphics[width=0.99\textwidth]{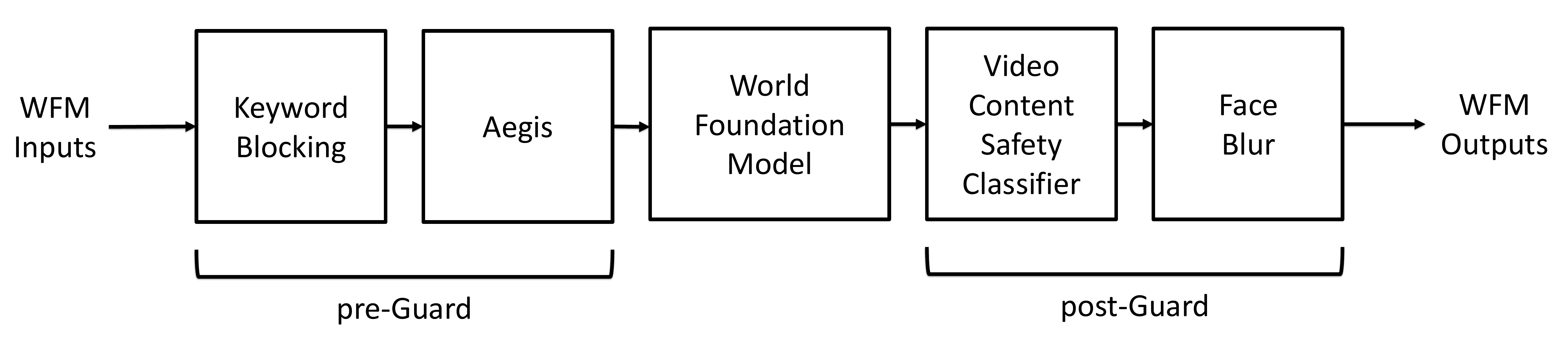}
    \caption{\textbf{Cosmos Guardrail overview}. Cosmos Guardrail contains pre-Guard and post-Guard, where pre-Guard blocks inputs based on Aegis~\citep{aegis} and keywords, while post-Guard blocks outputs based on a video content safety classifier and blurs output faces.}
    \label{fig:guardrail_pipeline}
\end{figure}

For the safe use of our WFMs, we develop a comprehensive guardrail system. It consists of two stages: the pre-Guard stage and the post-Guard stage. The pre-Guard stage leverages Aegis~\citep{aegis} and a keyword list to block harmful prompts. The post-Guard stage blocks harmful visual outputs using a video content safety classifier and a face blur filter. The pipeline is illustrated in \cref{fig:guardrail_pipeline}.

\subsection{Pre-Guard}\label{sec::pre-guard}

Our pre-Guard is a text-domain guardrail comprising an LLM-based guardrail for semantically complex prompts and a simple blocklist-based checker for explicitly unsafe keywords. 

\subsubsection{Keyword Blocking}\label{sec::blocklist}

The blocklist heuristic will act as the first line of defense to mitigate the risk of generating unsafe content. This is designed to block explicitly harmful generations by doing a keyword search on the prompt against a hard-coded blocklist of a large corpus of explicit and objectionable words. Input words are lemmatized using \emph{WordNetLemmatizer}, a tool that uses a lexical database of the English language \citep{wordnet} to extract the root word from its variants. For example, the root word of ``\textit{abacii}'' is ``\textit{abacus}''. These lemmatized words are then compared to the words in the hard-coded blocklist, and the entire prompt is rejected if any profanity is found. We use a comprehensive set of keywords to maximally protect our users.

\subsubsection{Aegis Guardrail}\label{sec::aegis}

As the second line of defense, we use \emph{Aegis-AI-Content-Safety-LlamaGuard-LLM-Defensive-1.0} ~\citep{aegis}, which is a fine-tuned version of \emph{Llama-Guard} \citep{llamaguard} trained on NVIDIA's Aegis Content Safety Dataset covering NVIDIA's broad taxonomy of 13 critical safety risk categories. There are two versions of AEGIS 1.0, the defensive version and the permissive version. The defensive version adopts a tighter permission boundary than the permissive version. Cosmos uses the defensive version of Aegis to block potentially harmful user prompts that attempt to generate harmful content. If the input prompt is categorized as unsafe by this prompt filter, the video is not generated, and an error message is displayed.  

For using Aegis as a prompt filter, we classify the prompt as unsafe if it falls into the following categories: violence, sexual, criminal planning, weapons, substance abuse, suicide, child sexual abuse material, hatred, harassment, threat, and profanity. Any prompt that does not fall into the above categories is considered safe from the prompt-filtering standpoint.

\subsection{Post-Guard}\label{sec::post-guard}
Our post-Guard is a vision-domain guardrail comprising a video content safety filter and a face blur filter for the generated output.

\subsubsection{Video Content Safety Filter}\label{sec::safety-classifier}

The Video Content Safety Filter is a frame-level multi-class classifier trained on our video dataset and generation results. Among the classes, some are considered safe, while others are unsafe. A major challenge in training the classifier is in balancing false positives, where safe content is mistakenly flagged as unsafe, and false negatives, where unsafe content is wrongly classified as safe. To minimize classification errors, we carefully balanced the data during training.

We collect three kinds of ground truth annotated data. First, we sample a large set of videos from our dataset, extract frames, and determine its class using a VLM. Next, we generate synthetic videos with our WFMs using a set of prompts to ensure coverage of corner cases and least-represented content categories. Finally, human annotators provide the ``gold standard'' labels for a portion of our dataset, adding a vital layer of validation and helping us continuously refine the accuracy of our classifier. We extract the SigLIP~\citep{siglip} embedding for each video frame and train a simple MLP classifier on the embeddings.
ing inference, we generate a SigLIP embedding for every frame and then apply the classifier. The entire video is flagged as unsafe if any frame is classified as unsafe

\subsubsection{Face Blur Filter}\label{sec::faceblur}

We use RetinaFace \citep{retinaface}, a state-of-the-art face detection model, to identify facial regions with high confidence scores. For any detected face region larger than $20 \times 20$ pixels, we apply pixelation to obscure the regions while preserving the overall scene composition for Physical AI applications.

\subsection{Red Team Effort}\label{sec::redteaming}

We employ a dedicated red team to actively probe the system using both standard and adversarial examples that are collected in an internal attack prompt dataset. These video outputs are annotated by a team of expert annotators, who were specially trained for our task, to classify the generated video on a scale of 1-5 on multiple categories of harm related to the taxonomy in \cref{sec::aegis}.  These annotations also specify the start and end-frames where the unsafe content is detected, thereby generating high-quality annotations. The red team also probed each guardrail component independently with targeted examples to identify weaknesses and improve performance in edge cases.  As of the date of publication, the red team has tested and annotated over $10{,}000$ distinct prompt-video pairs that were carefully crafted to cover a broad range of unsafe content.

\section{Related Work}\label{sec:related}

\noindent \textbf{World models.}
The concept of ``world models'' originated from the seminal work of~\citet{ha2018world}, which proposed learning a representation of the real world using neural network models to predict future states given current states and inputs. An accurate representation of the physical world model enables not only reliable prediction of future states but also informed decision-making. This concept of modeling the physical world is not new; traditional automation and robotics industries have long employed mathematical models based on physics laws and system identification in planning and control algorithms~\citep{murray2017mathematical}. However, these system-specific models, typically confined to low-dimensional state spaces, restrict generalization and knowledge transfer across different systems, limiting model reuse when applied to new tasks or environments. Recent advances in deep learning, particularly generative AI, have made it possible to learn world models directly from visual observations.

Modern world model pipelines can be categorized based on their backbone architecture. Most works~\citep{hafner2019dream,hafner2020mastering,Kim2020_GameGan,Kim2021_DriveGAN,hafner2023mastering,hansen2023td}, including the original paper~\citep{ha2018world} by Ha and Schmidhuber, employ a recurrent neural network to model system state evolution in a latent space learned via an autoencoder. More recent trends view world models as generative models in visual observation space, often in the form of conditional video generative models (\eg, action-to-video, text-to-video). These models can be either autoregressive~\citep{yang2023learning,micheli2022transformers,robine2023transformer,bruce2024genie,liu2024world} or diffusion-based~\citep{valevski2024diffusion,alonso2024diffusion,ding2024diffusion}, as considered in this work. Another promising approach is generative simulation~\citep{nasiriany2024robocasa,hua2024gensim2}, which combines generative AI and physical simulators to model the real world.

A well-trained world model can be applied in various ways, including verification~\citep{hu2023gaia}, planning-based model predictive control~\citep{hansen2023td,bar2024navigation}, and model-based reinforcement learning~\citep{yang2023learning,robine2023transformer,alonso2024diffusion,zhang2024storm,ding2024diffusion}. The effectiveness of world models has been demonstrated in domains such as computer games~\citep{hafner2020mastering,Kim2020_GameGan,bruce2024genie,valevski2024diffusion,alonso2024diffusion}, real-world robots~\citep{wu2023daydreamer,yang2023learning}, and autonomous driving~\citep{Kim2021_DriveGAN,Blattmann2023Align,hu2023gaia,zhao2024drivedreamer}. We envision that foundational world models will have transformative impacts on these industries.

\noindent \textbf{Video generative models.}
The field of video generative models has undergone rapid development in recent years. From the initial models that produced short, low-resolution videos, the field has evolved significantly, with video generative models now at the forefront of generative AI research~\citep{ho2022video,huang2024vbench}. Recent years have seen the emergence of impressive video generative models, such as Sora, Dream Machine, Gen 3 and Kling, capable of producing realistic, high-resolution videos~\citep{dreammachine,sora,kling,gen3}. These advancements have been made in just a few years since the release of the first video generative model.

Most existing work on video generative models focuses on text-to-video tasks, which generate videos based on text prompt inputs~\citep{yang2024cogvideox,ma2024latte,lin2024open,blattmann2023stable,ge2023preserve,girdhar2023emu}. These models enable users to create impressive videos using carefully designed text prompts. Other popular tasks include Image-to-Video that generates videos starting from a given image frame~\citep{blattmann2023stable,wang2024unianimate,ren2024consisti2v,wang2021one,mallya2022implicit,gururani2022SPACE}, Video-to-Video that generates new videos given a reference video~\citep{wang2018video, wang2019few, mallya2020world, ku2024anyv2v,liu2024stablev2v} and Action-to-Video that generates videos based on actions driven by the development of world models and embodied AI~\citep{bruce2024genie,valevski2024diffusion,alonso2024diffusion,tulyakov2018mocogan}.

The majority of video generative models adopt the diffusion model framework~\citep{blattmann2023stable,lin2024open,ge2023preserve,ma2024latte,yang2024cogvideox} to gradually transform noise into video sequences. Autoregressive models have also been employed for video generation, offering the advantage of handling video and other modalities in a unified manner~\citep{videopoet,deng2024autoregressive,liu2024mardini}. While autoregressive models have shown promise, diffusion-based video models still excel in terms of visual quality. Our goal is to help Physical AI developers advance their applications. We believe that the diffusion-based and autoregressive-based models both have their pros and cons. Diffusion-based models could render videos with better visual quality. Autoregressive-based models can better leverage all sorts of techniques developed by the LLM community. We build both diffusion-based (Cosmos-Diffusion) and autoregressive-based (Cosmos-Autoregressive) WFMs and make them available to the Physical AI builders.

\noindent \textbf{Video generation with camera control.} 3D-consistent video generation traces back to early works in view synthesis and 3D reconstruction, where the community sought to create 3D-consistent videos using neural rendering applied to various 3D representations~\citep{zhou2018stereo,mildenhall2020nerf,wang2021neus,li2023neuralangelo,kerbl20233d}. Within this line of research, single-image 3D view synthesis~\citep{tucker2020single,wiles2020synsin,yu2021pixelnerf,lin2023vision,charatan2024pixelsplat} is particularly challenging, typically requiring the learning of a strong 3D prior model from multi-view image datasets. As such 3D prior models often suffer to scale well, learning-based view synthesis has also been explored through a purely data-driven approach using scalable Transformer architectures~\citep{vaswani2017attention,dosovitskiy2021image}. This bypasses the need for explicit 3D prior knowledge~\citep{rombach2021geometry,sajjadi2022scene}: instead of relying on neural rendering applied to 3D representations, novel views are synthesized directly by neural networks conditioned on camera inputs~\citep{tatarchenko2016multi}. This paradigm has been successfully scaled up with diffusion models~\citep{liu2023zero}, finding broad applications in 3D asset generation~\citep{poole2023dreamfusion,lin2023magic3d,shi2023mvdream,qian2024magic123,li2024instant3d,nvidia2024edify3d}. Recent advances in video generation quality suggest the potential for achieving full 3D consistency through the scaling of training video data~\citep{videoworldsimulators2024}. Camera controllability on such models has since become an active area of investigation~\citep{he2024cameractrl,wang2024motionctrl,xu2024camco} for its great potential applications to robotics and autonomous navigation.

\noindent \textbf{Generative models for robotic control. }Recent advances in deep generative models have sparked significant interest in their application to robotic control. Several approaches have emerged, with one line of work directly employing diffusion models as visuomotor policies, demonstrating substantial improvements in imitation learning in various robotic tasks~\citep{chi2023diffusion,wang2024one,prasad2024consistency,ke20243d}. Two other threads more related to this work are the use of pre-trained image and video generation models as motion planners and the use of image and video data for generative pre-training. The generative motion planning approach ~\citep{kolearning,zhourobodreamer,finn2017deep,black2023zero,du2024learning} aims to enhance generalization to unseen environments by generating intermediate visual sub-goals rather than explicit action sequences. This visual representation strategy proves more robust, as image and video sub-goals can generalize across diverse environmental setups, unlike action sequences that are typically environment- and task-specific. The generative pre-training approach~\citep{gupta2024pre,cheang2024gr,he2024learning} leverages large-scale image and video datasets for pre-training. While \cite{gupta2024pre} extract and utilize features from pre-trained text-to-image diffusion models to guide subsequent policy learning, \cite{cheang2024gr} and \cite{he2024learning} use a two-stage framework: first pre-training the model to predict future frames, then fine-tuning it to jointly predict both actions and future frames.

\noindent \textbf{Generative models for autonomous driving.} Video generative models have the potential to revolutionize autonomous driving simulation by enabling the generation of realistic driving videos conditioned on diverse input modalities, such as text, images, trajectories, 3D data, or maps~\citep{Kim2021_DriveGAN,Blattmann2023Align,gao2024vista,wang2023drivedreamer,wang2024driving,lu2025wovogen,jia2023adriver,yang2024generalized,hu2023gaia,gao2023magicdrive,gao2024magicdrivedit}. Despite their potential, existing approaches have been limited by constraints in data scale~\citep{wang2023drivedreamer,wang2024driving,lu2025wovogen,jia2023adriver,gao2023magicdrive,gao2024magicdrivedit}, resolution~\citep{yang2024generalized,hu2023gaia}, and the number of camera views~\citep{Blattmann2023Align,gao2024vista}, restricting their effectiveness as comprehensive driving world simulators. To overcome these limitations, we leverage the capabilities of a powerful pre-trained WFM to develop a flexible and scalable driving simulator. Our model achieves high resolution, elevated frame rates, and multi-view consistency.

\noindent \textbf{Tokenizer.}
There has been a fairly long history of learning latent features that reproduce the input visual data~\citep{kingma2013auto,van2017neural,hinton1995wake,he2022masked}. Recently, such models, also known as tokenizers, have been widely incorporated as essential components to improve the efficiency of training large-scale generative models~\citep {rombach2022high,esser2021taming}.

Continuous visual tokenizers, often including Autoencoder (AE) and Variational Autoencoder (VAE), compress visual data into a continuous latent space where diffusion-based models can be efficiently trained on~\citep{song2020score,lipman2022flow,ho2020denoising}. At inference time, the generated latents are decoded back to RGB space with the tokenizer decoder. Various diffusion models have been trained in such a way for image~\citep{rombach2022high,DALLE2,betker2023improving,dai2023emu,podell2024sdxl,flux2024,gafni2022make} and video generation~\citep{Zeng2023Video,Blattmann2023Align,blattmann2023stable,ge2023preserve,videoworldsimulators2024,wang2023lavie,yu2023video,an2023latent,girdhar2023emu}.

Discrete visual tokenizers additionally involve a quantizer~\citep{van2017neural,zhao2024bsqvit,mentzer2023finite,yu2024language,lee2022rvq,yu2024titok} that further discretizes the continuous latents into a discrete space, allowing for easy integration into large language models (LLMs) and vision language models (VLMs) alongside other modalities, such as text and audio. Thus, discrete tokenizers are deployed in various visual understanding~\citep{wu2024vilau,chameleon,emu2,wang2024emu3} as well as image~\citep{esser2021taming,dalle,yu2022scaling,chang2022maskgit,llamagen} and video generation tasks~\citep{yan2021videogpt,villegas2022phenaki,hong2022cogvideo,wu2022nuwa,ge2022long,yu2023magvit,luo2024open_magvit2,videopoet}.

Cosmos tokenizers are extensively built based on previous studies, \eg, FSQ~\citep{mentzer2023finite} and causal architecture~\citep{yu2023magvit}, with the goal of creating a suite of efficient and high-quality tokenizers.

\section{Conclusions and Discussions}\label{sec:conclusion}

The Cosmos World Foundation Models mark a significant step towards building general-purpose simulators for the physical world. This work outlines our comprehensive approach, including the data curation pipeline, the design of continuous and discrete tokenizers, the architecture of diffusion and autoregressive world foundation models, and the fine-tuning process for diverse downstream Physical AI tasks. Notably, we demonstrate the adaptability of our pre-trained world models to critical applications, including 3D world navigation, robotic manipulation, and autonomous vehicle systems, which demand both 3D consistency and action controllability.

\noindent \textbf{Limitations.} Despite the progress, the development of world foundation models is still in the early stages. Current models, including ours, fall short as reliable simulators of the physical world. We observe that our models still suffer from issues, including the lack of object permanence, inaccuracies in contact-rich dynamics, and inconsistency in instruction following. Additionally, the realism of the generated videos does not always reflect adherence to fundamental physical principles, such as gravity, light interactions, and fluid dynamics.

Evaluation presents another significant challenge. Defining robust rubrics for humans to evaluate physical fidelity is hard as such assessments are often influenced by personal biases, backgrounds, and other subjective factors. Moreover, these evaluations may not align positively with metrics used in downstream Physical AI tasks. In order to address these challenges, promising directions include the development of automated evaluators powered by multi-modal LLMs and leveraging existing physical simulators to enable reproducible and interactive evaluation, thereby reducing dependence on human evaluation.

\noindent \textbf{Autoregressive \vs Diffusion WFMs.} Our evaluation results in 3D consistency (\cref{sec::eval_3d_consistency}) and video generation for robotics (\cref{sec::robo}) indicate that diffusion-based WFMs currently deliver better generation quality. Through fine-tuning, diffusion-based WFMs are able to incorporate diverse control signals, including camera pose, end-effector positions, or autonomous vehicle trajectories, and generate outputs of novel formats like multi-view videos. However, autoregressive-based WFMs possess significant untapped potential. They could (1) leverage pre-trained weights from large language models (LLMs) to inherit extensive world knowledge and (2) enable faster generation through the use of advanced inference optimization techniques designed for causal attention. If these capabilities are fully realized, autoregressive WFMs may become particularly well-suited for applications requiring interactive control or real-time processing, such as planning and simulation in robotics. Importantly, the boundary between diffusion and autoregressive models is not rigid. Recent advancements have shown that diffusion transformers with bidirectional attention can be distilled into student transformers with causal attention, enabling support for key-value caching during inference~\citep{yin2024slow}. Similarly, autoregressive models can incorporate locally bidirectional attention to generate images via diffusion heads~\citep{zhou2024transfusion}. Exploring these hybrid approaches and their trade-offs remains an active and promising area of research. We plan to investigate these formulations further and provide a comprehensive analysis in future work.

\clearpage
\appendix
\section{Contributors and Acknowledgements}
\label{sec:contributors}

\subsection{Core Contributors}

\vspace{8pt}
\begin{itemize}[leftmargin=14pt]
    \setlength\itemsep{8pt}
    
    \item \textbf{Data Curation} \\
    Jacob Huffman, Francesco Ferroni, Alice Luo, Niket Agarwal, Hao Wang, Jing Zhang, David Page, Vasanth Rao Naik Sabavat, Sriharsha Niverty, Erik Barker, Lindsey Pavao, Stella Shi, Prithvijit Chattopadhyay, Shitao Tang, Yin Cui, Yunhao Ge, Qianli Ma, Yifan Ding, Seungjun Nah, Siddharth Gururani, Jiashu Xu, Grace Lam, Tiffany Cai, Jibin Varghese, Pooya Jannaty, Jay Zhangjie Wu, Yuxuan Zhang, Huan Ling, Hanzi Mao, Heng Wang
    
    \item \textbf{Tokenizer} \\
    Jinwei Gu, Xian Liu, Songwei Ge, Ting-Chun Wang, Haoxiang Wang, Fitsum Reda
    
    \item \textbf{Diffusion-based World Foundation Model Pre-training} \\
    Qinsheng Zhang, Lin Yen-Chen, Xiaohui Zeng, Huan Ling, Shitao Tang, Maciej Bala, Ting-Chun Wang, Yu Zeng, Seungjun Nah, Qianli Ma, Hanzi Mao
    
    \item \textbf{Autoregressive-based World Foundation Model Pre-training} \\
    Haoxiang Wang, Yifan Ding, Xian Liu, Jiaojiao Fan, Xiaohui Zeng, Yogesh Balaji
    
    \item \textbf{Prompt Upsampler} \\
    Yunhao Ge, Haoxiang Wang, Jiashu Xu, Yin Cui
    
    \item \textbf{Diffusion Decoder} \\
    Huan Ling, Jiaojiao Fan, Fitsum Reda, Yogesh Balaji, Hanzi Mao, Qinsheng Zhang
    
    \item \textbf{3D Consistency Pre-training Evaluation} \\
    Jiahui Huang, Chen-Hsuan Lin
    
    \item \textbf{Physics Alignment Pre-training Evaluation} \\
    Francesco Ferroni, Prithvijit Chattopadhyay, Xinyue Wei, Qianli Ma, Gergely Klár, Chen-Hsuan Lin
    
    \item \textbf{Camera Control Post-training Evaluation} \\
    Xiaohui Zeng, Tsung-Yi Lin, Jingyi Jin, Chen-Hsuan Lin
    
    \item \textbf{Robotics Post-training Evaluation} \\
    Lin Yen-Chen, Wei-Cheng Tseng, Yunhao Ge, Xian Liu, Shitao Tang, Fangyin Wei, Lyne Tchapmi, Yu Zeng, Qingqing Zhao, Yin Cui, Zhaoshuo Li, Jinwei Gu
    
    \item \textbf{Autonomous Driving Post-training Evaluation} \\
    Seung Wook Kim, Jay Zhangjie Wu, Jiahui Huang, Francesco Ferroni, Michele Fenzi, Daniel Dworakowski, Despoina Paschalidou, Ed Schmerling, Shiyi Lan, Laura Leal-Taixe, Sanja Fidler, Huan Ling
    
    \item \textbf{Guardrail} \\
    Jibin Varghese, Arslan Ali, Grace Lam, Pooya Jannaty
    
    \item \textbf{Platform Architect} \\
    Ming-Yu Liu

\end{itemize}

\subsection{Contributors}
Anqi Li, Arsalan Mousavian, Artur Zolkowski, Bartosz Stefaniak, Dieter Fox, Ethan He, Kaichun Mo, Morteza Ramezanali, Przemek Tredak, Wei Yang, Xiaowei Ren, Yongxin Chen, Zeeshan Patel

\subsection{Acknowledgments}
We thank 1X Technologies for generously providing humanoid robot data and offering invaluable support for the post-training for robotic manipulation in this technical report.

We thank Aarti Basant, Akan Huang, Alex Qi, Alexis Bjorlin, Amanda Moran, Amol Fasale, Ankit Patel, Arash Vahdat, Aryaman Gupta, Ashna Khetan, Ashwath Aithal, Bor-Yiing Su, Bryan Catanzaro, Charles Hsu, Chris Pruett, Christopher Horvath, Clark Doan, Coulten Holt, Dane Aconfora, Deepak Narayanan, Dennis Chang, Dheeraj Kapur, Dong Ahn, Ebrar Erdem, Elmar Haussmann, Fuzhao Xue, Gandhi Vaithilingam, Henry Estela, Henry Vera, Herb Woodruff, Imad El Hanafi, Jashojit Mukherjee, Jason Sewall, Jensen Huang, John Dickinson, Jonah Alben, Jonah Philion, Josh Abbott, Jun Gao, Kumar Anik, Lee Ditiangkin, Ligeng Zhu, Linxi Fan, Luke Alonso, Madison Huang, Marek Dabek, Mark Arnold, Max Ehrlich, Michele Ferretti, Misbah Mubarak, Misha Smelyanskiy, Mohamed Fawzy, Mohammad Harrim, Mohammad Shoeybi, Omkar Mehta, Pallab Bhattacharya, Paniz Karbasi, Pasha Shamis, Raju Wagwani, Rick Izzo, Robert Hero, Sharon Clay, Song Han, Songyan Tang, Sophia Huang, Sridhar Bhuvanapalli, TJ Galda, Thomas Volk, Tobias Lasser, Vaibhav Ranglani, Vijay Anand Korthikanti, Yao Lu, Yazdan Aghaghiri, Yugi Guvvala, Yuke Zhu and Zekun Hao for their feedback and engineering support. 

We thank Iain Cunningham, Jim Fan, Marco Pavone, Meredith Price, Nikki Pope and Scott Reed for their feedback on the early draft of this technical report.

\clearpage
\setcitestyle{numbers}
\bibliographystyle{plainnat}
\bibliography{main}

\begin{thebibliography}{252}
\providecommand{\natexlab}[1]{#1}
\providecommand{\url}[1]{\texttt{#1}}
\expandafter\ifx\csname urlstyle\endcsname\relax
  \providecommand{\doi}[1]{doi: #1}\else
  \providecommand{\doi}{doi: \begingroup \urlstyle{rm}\Url}\fi

\bibitem[Abbas et~al.(2023)Abbas, Tirumala, Simig, Ganguli, and Morcos]{abbas2023semdedup}
Amro Abbas, Kushal Tirumala, D{\'a}niel Simig, Surya Ganguli, and Ari~S Morcos.
\newblock Semdedup: Data-efficient learning at web-scale through semantic deduplication.
\newblock \emph{arXiv preprint arXiv:2303.09540}, 2023.

\bibitem[Adler et~al.(2024)Adler, Agarwal, Aithal, Anh, Bhattacharya, Brundyn, Casper, Catanzaro, Clay, Cohen, et~al.]{adler2024nemotron}
Bo~Adler, Niket Agarwal, Ashwath Aithal, Dong~H Anh, Pallab Bhattacharya, Annika Brundyn, Jared Casper, Bryan Catanzaro, Sharon Clay, Jonathan Cohen, et~al.
\newblock Nemotron-4 340b technical report.
\newblock \emph{arXiv preprint arXiv:2406.11704}, 2024.

\bibitem[Agrawal et~al.(2024)Agrawal, Antoniak, Hanna, Bout, Chaplot, Chudnovsky, Costa, De~Monicault, Garg, Gervet, et~al.]{agrawal2024pixtral}
Pravesh Agrawal, Szymon Antoniak, Emma~Bou Hanna, Baptiste Bout, Devendra Chaplot, Jessica Chudnovsky, Diogo Costa, Baudouin De~Monicault, Saurabh Garg, Theophile Gervet, et~al.
\newblock Pixtral 12b.
\newblock \emph{arXiv preprint arXiv:2410.07073}, 2024.

\bibitem[{AI Image Lab, University of Modena}(2016)]{bbcplanetearthdataset}
{AI Image Lab, University of Modena}.
\newblock Bbc planet earth dataset, 2016.
\newblock URL \url{https://aimagelab.ing.unimore.it/imagelab/researchActivity.asp?idActivity=19}.
\newblock Accessed: 2024-10-17.

\bibitem[Alonso et~al.(2024)Alonso, Jelley, Micheli, Kanervisto, Storkey, Pearce, and Fleuret]{alonso2024diffusion}
Eloi Alonso, Adam Jelley, Vincent Micheli, Anssi Kanervisto, Amos Storkey, Tim Pearce, and Fran{\c{c}}ois Fleuret.
\newblock Diffusion for world modeling: Visual details matter in atari.
\newblock In \emph{NeurIPS}, 2024.

\bibitem[An et~al.(2023)An, Zhang, Yang, Gupta, Huang, Luo, and Yin]{an2023latent}
Jie An, Songyang Zhang, Harry Yang, Sonal Gupta, Jia-Bin Huang, Jiebo Luo, and Xi~Yin.
\newblock Latent-shift: Latent diffusion with temporal shift for efficient text-to-video generation.
\newblock \emph{arXiv preprint arXiv:2304.08477}, 2023.

\bibitem[Atzmon et~al.(2024)Atzmon, Bala, Balaji, Cai, Cui, Fan, Ge, Gururani, Huffman, Isaac, et~al.]{atzmon2024edify}
Yuval Atzmon, Maciej Bala, Yogesh Balaji, Tiffany Cai, Yin Cui, Jiaojiao Fan, Yunhao Ge, Siddharth Gururani, Jacob Huffman, Ronald Isaac, et~al.
\newblock Edify image: High-quality image generation with pixel space laplacian diffusion models.
\newblock \emph{arXiv preprint arXiv:2411.07126}, 2024.

\bibitem[Balaji et~al.(2022)Balaji, Nah, Huang, Vahdat, Song, Zhang, Kreis, Aittala, Aila, Laine, et~al.]{ediffI}
Yogesh Balaji, Seungjun Nah, Xun Huang, Arash Vahdat, Jiaming Song, Qinsheng Zhang, Karsten Kreis, Miika Aittala, Timo Aila, Samuli Laine, et~al.
\newblock ediff-i: Text-to-image diffusion models with an ensemble of expert denoisers.
\newblock \emph{arXiv preprint arXiv:2211.01324}, 2022.

\bibitem[Bar et~al.(2024)Bar, Zhou, Tran, Darrell, and LeCun]{bar2024navigation}
Amir Bar, Gaoyue Zhou, Danny Tran, Trevor Darrell, and Yann LeCun.
\newblock Navigation world models.
\newblock \emph{arXiv preprint arXiv:2412.03572}, 2024.

\bibitem[Betker et~al.(2023)Betker, Goh, Jing, Brooks, Wang, Li, Ouyang, Zhuang, Lee, Guo, et~al.]{betker2023improving}
James Betker, Gabriel Goh, Li~Jing, Tim Brooks, Jianfeng Wang, Linjie Li, Long Ouyang, Juntang Zhuang, Joyce Lee, Yufei Guo, et~al.
\newblock Improving image generation with better captions.
\newblock \emph{Computer Science. https://cdn. openai. com/papers/dall-e-3. pdf}, 2023.

\bibitem[Black et~al.(2023)Black, Nakamoto, Atreya, Walke, Finn, Kumar, and Levine]{black2023zero}
Kevin Black, Mitsuhiko Nakamoto, Pranav Atreya, Homer Walke, Chelsea Finn, Aviral Kumar, and Sergey Levine.
\newblock Zero-shot robotic manipulation with pre-trained image-editing diffusion models.
\newblock In \emph{NeurIPS Workshops}, 2023.

\bibitem[Blattmann et~al.(2023{\natexlab{a}})Blattmann, Dockhorn, Kulal, Mendelevitch, Kilian, Lorenz, Levi, English, Voleti, Letts, et~al.]{blattmann2023stable}
Andreas Blattmann, Tim Dockhorn, Sumith Kulal, Daniel Mendelevitch, Maciej Kilian, Dominik Lorenz, Yam Levi, Zion English, Vikram Voleti, Adam Letts, et~al.
\newblock Stable video diffusion: Scaling latent video diffusion models to large datasets.
\newblock \emph{arXiv preprint arXiv:2311.15127}, 2023{\natexlab{a}}.

\bibitem[Blattmann et~al.(2023{\natexlab{b}})Blattmann, Rombach, Ling, Dockhorn, Kim, Fidler, and Kreis]{Blattmann2023Align}
Andreas Blattmann, Robin Rombach, Huan Ling, Tim Dockhorn, Seung~Wook Kim, Sanja Fidler, and Karsten Kreis.
\newblock Align your latents: High-resolution video synthesis with latent diffusion models.
\newblock In \emph{CVPR}, 2023{\natexlab{b}}.

\bibitem[Brooks et~al.(2024)Brooks, Peebles, Holmes, DePue, Guo, Jing, Schnurr, Taylor, Luhman, Luhman, Ng, Wang, and Ramesh]{videoworldsimulators2024}
Tim Brooks, Bill Peebles, Connor Holmes, Will DePue, Yufei Guo, Li~Jing, David Schnurr, Joe Taylor, Troy Luhman, Eric Luhman, Clarence Ng, Ricky Wang, and Aditya Ramesh.
\newblock Video generation models as world simulators, 2024.
\newblock URL \url{https://openai.com/research/video-generation-models-as-world-simulators}.

\bibitem[Brown et~al.(2020)Brown, Mann, Ryder, Subbiah, Kaplan, Dhariwal, Neelakantan, Shyam, Sastry, Askell, et~al.]{gpt3}
Tom Brown, Benjamin Mann, Nick Ryder, Melanie Subbiah, Jared~D Kaplan, Prafulla Dhariwal, Arvind Neelakantan, Pranav Shyam, Girish Sastry, Amanda Askell, et~al.
\newblock Language models are few-shot learners.
\newblock In \emph{NeurIPS}, 2020.

\bibitem[Bruce et~al.(2024)Bruce, Dennis, Edwards, Parker-Holder, Shi, Hughes, Lai, Mavalankar, Steigerwald, Apps, et~al.]{bruce2024genie}
Jake Bruce, Michael~D Dennis, Ashley Edwards, Jack Parker-Holder, Yuge Shi, Edward Hughes, Matthew Lai, Aditi Mavalankar, Richie Steigerwald, Chris Apps, et~al.
\newblock Genie: Generative interactive environments.
\newblock In \emph{ICML}, 2024.

\bibitem[Cai et~al.(2024)Cai, Li, Geng, Peng, Lee, Chen, and Dao]{cai2024medusa}
Tianle Cai, Yuhong Li, Zhengyang Geng, Hongwu Peng, Jason~D. Lee, Deming Chen, and Tri Dao.
\newblock Medusa: Simple {LLM} inference acceleration framework with multiple decoding heads.
\newblock In \emph{ICML}, 2024.

\bibitem[Castellano(2024)]{PySceneDetect}
Brandon Castellano.
\newblock Pyscenedetect, 2024.
\newblock URL \url{https://www.scenedetect.com}.
\newblock Video Cut Detection and Analysis Tool.

\bibitem[Chang et~al.(2022)Chang, Zhang, Jiang, Liu, and Freeman]{chang2022maskgit}
Huiwen Chang, Han Zhang, Lu~Jiang, Ce~Liu, and William~T. Freeman.
\newblock Maskgit: Masked generative image transformer.
\newblock In \emph{CVPR}, 2022.

\bibitem[Charatan et~al.(2024)Charatan, Li, Tagliasacchi, and Sitzmann]{charatan2024pixelsplat}
David Charatan, Sizhe~Lester Li, Andrea Tagliasacchi, and Vincent Sitzmann.
\newblock pixelsplat: 3d gaussian splats from image pairs for scalable generalizable 3d reconstruction.
\newblock In \emph{CVPR}, 2024.

\bibitem[Cheang et~al.(2024)Cheang, Chen, Jing, Kong, Li, Li, Liu, Wu, Xu, Yang, et~al.]{cheang2024gr}
Chi-Lam Cheang, Guangzeng Chen, Ya~Jing, Tao Kong, Hang Li, Yifeng Li, Yuxiao Liu, Hongtao Wu, Jiafeng Xu, Yichu Yang, et~al.
\newblock Gr-2: A generative video-language-action model with web-scale knowledge for robot manipulation.
\newblock \emph{arXiv preprint arXiv:2410.06158}, 2024.

\bibitem[Chen et~al.(2016)Chen, Xu, Zhang, and Guestrin]{chen2016training}
Tianqi Chen, Bing Xu, Chiyuan Zhang, and Carlos Guestrin.
\newblock Training deep nets with sublinear memory cost.
\newblock \emph{arXiv preprint arXiv:1604.06174}, 2016.

\bibitem[Chen(2023)]{chen2023importance}
Ting Chen.
\newblock On the importance of noise scheduling for diffusion models.
\newblock \emph{arXiv preprint arXiv:2301.10972}, 2023.

\bibitem[Chen et~al.(2024)Chen, Siarohin, Menapace, Deyneka, Chao, Jeon, Fang, Lee, Ren, Yang, et~al.]{chen2024panda70m}
Tsai-Shien Chen, Aliaksandr Siarohin, Willi Menapace, Ekaterina Deyneka, Hsiang-wei Chao, Byung~Eun Jeon, Yuwei Fang, Hsin-Ying Lee, Jian Ren, Ming-Hsuan Yang, et~al.
\newblock Panda-70m: Captioning 70m videos with multiple cross-modality teachers.
\newblock In \emph{CVPR}, 2024.

\bibitem[Chi et~al.(2023)Chi, Feng, Du, Xu, Cousineau, Burchfiel, and Song]{chi2023diffusion}
Cheng Chi, Siyuan Feng, Yilun Du, Zhenjia Xu, Eric Cousineau, Benjamin Burchfiel, and Shuran Song.
\newblock Diffusion policy: Visuomotor policy learning via action diffusion.
\newblock \emph{RSS}, 2023.

\bibitem[Dai et~al.(2023)Dai, Hou, Ma, Tsai, Wang, Wang, Zhang, Vandenhende, Wang, Dubey, et~al.]{dai2023emu}
Xiaoliang Dai, Ji~Hou, Chih-Yao Ma, Sam Tsai, Jialiang Wang, Rui Wang, Peizhao Zhang, Simon Vandenhende, Xiaofang Wang, Abhimanyu Dubey, et~al.
\newblock Emu: Enhancing image generation models using photogenic needles in a haystack.
\newblock \emph{arXiv preprint arXiv:2309.15807}, 2023.

\bibitem[de~Br{\'e}bisson and Vincent(2016)]{de2016z}
Alexandre de~Br{\'e}bisson and Pascal Vincent.
\newblock The z-loss: a shift and scale invariant classification loss belonging to the spherical family.
\newblock \emph{arXiv preprint arXiv:1604.08859}, 2016.

\bibitem[Dehghani et~al.(2023)Dehghani, Djolonga, Mustafa, Padlewski, Heek, Gilmer, Steiner, Caron, Geirhos, Alabdulmohsin, et~al.]{dehghani2023scaling}
Mostafa Dehghani, Josip Djolonga, Basil Mustafa, Piotr Padlewski, Jonathan Heek, Justin Gilmer, Andreas~Peter Steiner, Mathilde Caron, Robert Geirhos, Ibrahim Alabdulmohsin, et~al.
\newblock Scaling vision transformers to 22 billion parameters.
\newblock In \emph{ICML}, 2023.

\bibitem[Deng et~al.(2024)Deng, Pan, Diao, Luo, Cui, Lu, Shan, Qi, and Wang]{deng2024autoregressive}
Haoge Deng, Ting Pan, Haiwen Diao, Zhengxiong Luo, Yufeng Cui, Huchuan Lu, Shiguang Shan, Yonggang Qi, and Xinlong Wang.
\newblock Autoregressive video generation without vector quantization.
\newblock \emph{arXiv preprint arXiv:2412.14169}, 2024.

\bibitem[Deng et~al.(2009)Deng, Dong, Socher, Li, Li, and Fei-Fei]{deng2009imagenet}
Jia Deng, Wei Dong, Richard Socher, Li-Jia Li, Kai Li, and Li~Fei-Fei.
\newblock Imagenet: A large-scale hierarchical image database.
\newblock In \emph{CVPR}, 2009.

\bibitem[Deng et~al.(2020)Deng, Guo, Zhou, Yu, Kotsia, and Zafeiriou]{retinaface}
Jiankang Deng, Jia Guo, Yuxiang Zhou, Jinke Yu, Irene Kotsia, and Stefanos Zafeiriou.
\newblock Retinaface: Single-stage dense face localisation in the wild.
\newblock \emph{CVPR}, 2020.

\bibitem[DeTone et~al.(2018)DeTone, Malisiewicz, and Rabinovich]{detone2018superpoint}
Daniel DeTone, Tomasz Malisiewicz, and Andrew Rabinovich.
\newblock Superpoint: Self-supervised interest point detection and description.
\newblock In \emph{CVPR Workshops}, 2018.

\bibitem[Ding et~al.(2024)Ding, Zhang, Tian, and Zheng]{ding2024diffusion}
Zihan Ding, Amy Zhang, Yuandong Tian, and Qinqing Zheng.
\newblock Diffusion world model: Future modeling beyond step-by-step rollout for offline reinforcement learning.
\newblock \emph{arXiv preprint arXiv:2402.03570}, 2024.

\bibitem[Dosovitskiy et~al.(2021)Dosovitskiy, Beyer, Kolesnikov, Weissenborn, Zhai, Unterthiner, Dehghani, Minderer, Heigold, Gelly, et~al.]{dosovitskiy2021image}
Alexey Dosovitskiy, Lucas Beyer, Alexander Kolesnikov, Dirk Weissenborn, Xiaohua Zhai, Thomas Unterthiner, Mostafa Dehghani, Matthias Minderer, Georg Heigold, Sylvain Gelly, et~al.
\newblock An image is worth 16x16 words: transformers for image recognition at scale.
\newblock In \emph{ICLR}, 2021.

\bibitem[Du et~al.(2024)Du, Yang, Dai, Dai, Nachum, Tenenbaum, Schuurmans, and Abbeel]{du2024learning}
Yilun Du, Sherry Yang, Bo~Dai, Hanjun Dai, Ofir Nachum, Josh Tenenbaum, Dale Schuurmans, and Pieter Abbeel.
\newblock Learning universal policies via text-guided video generation.
\newblock In \emph{NeurIPS}, 2024.

\bibitem[Dubey et~al.(2024)Dubey, Jauhri, Pandey, Kadian, Al-Dahle, Letman, Mathur, Schelten, Yang, Fan, et~al.]{dubey2024llama}
Abhimanyu Dubey, Abhinav Jauhri, Abhinav Pandey, Abhishek Kadian, Ahmad Al-Dahle, Aiesha Letman, Akhil Mathur, Alan Schelten, Amy Yang, Angela Fan, et~al.
\newblock The llama 3 herd of models.
\newblock \emph{arXiv preprint arXiv:2407.21783}, 2024.

\bibitem[Ebert et~al.(2022)Ebert, Yang, Schmeckpeper, Bucher, Georgakis, Daniilidis, Finn, and Levine]{frederik22bridge}
Frederik Ebert, Yanlai Yang, Karl Schmeckpeper, Bernadette Bucher, Georgios Georgakis, Kostas Daniilidis, Chelsea Finn, and Sergey Levine.
\newblock Bridge data: Boosting generalization of robotic skills with cross-domain datasets.
\newblock In \emph{RSS}, 2022.

\bibitem[Esser et~al.(2021)Esser, Rombach, and Ommer]{esser2021taming}
Patrick Esser, Robin Rombach, and Bjorn Ommer.
\newblock Taming transformers for high-resolution image synthesis.
\newblock In \emph{CVPR}, 2021.

\bibitem[Esser et~al.(2024)Esser, Kulal, Blattmann, Entezari, M{\"u}ller, Saini, Levi, Lorenz, Sauer, Boesel, et~al.]{esser2024scaling}
Patrick Esser, Sumith Kulal, Andreas Blattmann, Rahim Entezari, Jonas M{\"u}ller, Harry Saini, Yam Levi, Dominik Lorenz, Axel Sauer, Frederic Boesel, et~al.
\newblock Scaling rectified flow transformers for high-resolution image synthesis.
\newblock In \emph{ICML}, 2024.

\bibitem[Farneb{\"a}ck(2003)]{farneback2003two}
Gunnar Farneb{\"a}ck.
\newblock Two-frame motion estimation based on polynomial expansion.
\newblock In \emph{Scandinavian Conference on Image Analysis}, 2003.

\bibitem[Finn and Levine(2017)]{finn2017deep}
Chelsea Finn and Sergey Levine.
\newblock Deep visual foresight for planning robot motion.
\newblock In \emph{ICRA}, 2017.

\bibitem[{FLUX}(2024)]{flux2024}
{FLUX}.
\newblock {FLUX}.1: Image generation, 2024.
\newblock URL \url{https://huggingface.co/black-forest-labs/FLUX.1-dev}.

\bibitem[Fu et~al.(2023)Fu, Tamir, Sundaram, Chai, Zhang, Dekel, and Isola]{fu2023dreamsim}
Stephanie Fu, Netanel~Yakir Tamir, Shobhita Sundaram, Lucy Chai, Richard Zhang, Tali Dekel, and Phillip Isola.
\newblock Dreamsim: Learning new dimensions of human visual similarity using synthetic data.
\newblock In \emph{NeurIPS}, 2023.

\bibitem[Gadre et~al.(2024)Gadre, Ilharco, Fang, Hayase, Smyrnis, Nguyen, Marten, Wortsman, Ghosh, Zhang, et~al.]{gadre2023datacomp}
Samir~Yitzhak Gadre, Gabriel Ilharco, Alex Fang, Jonathan Hayase, Georgios Smyrnis, Thao Nguyen, Ryan Marten, Mitchell Wortsman, Dhruba Ghosh, Jieyu Zhang, et~al.
\newblock Datacomp: In search of the next generation of multimodal datasets.
\newblock In \emph{NeurIPS}, 2024.

\bibitem[Gafni et~al.(2022)Gafni, Polyak, Ashual, Sheynin, Parikh, and Taigman]{gafni2022make}
Oran Gafni, Adam Polyak, Oron Ashual, Shelly Sheynin, Devi Parikh, and Yaniv Taigman.
\newblock Make-a-scene: Scene-based text-to-image generation with human priors.
\newblock In \emph{ECCV}, 2022.

\bibitem[Gage(1994)]{gage1994new}
Philip Gage.
\newblock A new algorithm for data compression.
\newblock \emph{The C Users Journal}, 1994.

\bibitem[Gao et~al.(2024{\natexlab{a}})Gao, Hoogeboom, Heek, Bortoli, Murphy, and Salimans]{gao2025diffusionmeetsflow}
Ruiqi Gao, Emiel Hoogeboom, Jonathan Heek, Valentin~De Bortoli, Kevin~P. Murphy, and Tim Salimans.
\newblock Diffusion meets flow matching: Two sides of the same coin, 2024{\natexlab{a}}.
\newblock URL \url{https://diffusionflow.github.io/}.

\bibitem[Gao et~al.(2024{\natexlab{b}})Gao, Chen, Xiao, Hong, Li, and Xu]{gao2024magicdrivedit}
Ruiyuan Gao, Kai Chen, Bo~Xiao, Lanqing Hong, Zhenguo Li, and Qiang Xu.
\newblock Magicdrivedit: High-resolution long video generation for autonomous driving with adaptive control.
\newblock \emph{arXiv preprint arXiv:2411.13807}, 2024{\natexlab{b}}.

\bibitem[Gao et~al.(2024{\natexlab{c}})Gao, Chen, Xie, HONG, Li, Yeung, and Xu]{gao2023magicdrive}
Ruiyuan Gao, Kai Chen, Enze Xie, Lanqing HONG, Zhenguo Li, Dit-Yan Yeung, and Qiang Xu.
\newblock Magicdrive: Street view generation with diverse 3d geometry control.
\newblock In \emph{ICLR}, 2024{\natexlab{c}}.

\bibitem[Gao et~al.(2024{\natexlab{d}})Gao, Yang, Chen, Chitta, Qiu, Geiger, Zhang, and Li]{gao2024vista}
Shenyuan Gao, Jiazhi Yang, Li~Chen, Kashyap Chitta, Yihang Qiu, Andreas Geiger, Jun Zhang, and Hongyang Li.
\newblock Vista: A generalizable driving world model with high fidelity and versatile controllability.
\newblock In \emph{NeurIPS}, 2024{\natexlab{d}}.

\bibitem[Gatys et~al.(2016)Gatys, Ecker, and Bethge]{gatys2016image}
Leon~A Gatys, Alexander~S Ecker, and Matthias Bethge.
\newblock Image style transfer using convolutional neural networks.
\newblock In \emph{CVPR}, 2016.

\bibitem[Ge et~al.(2022)Ge, Hayes, Yang, Yin, Pang, Jacobs, Huang, and Parikh]{ge2022long}
Songwei Ge, Thomas Hayes, Harry Yang, Xi~Yin, Guan Pang, David Jacobs, Jia-Bin Huang, and Devi Parikh.
\newblock Long video generation with time-agnostic vqgan and time-sensitive transformer.
\newblock In \emph{ECCV}, 2022.

\bibitem[Ge et~al.(2023)Ge, Nah, Liu, Poon, Tao, Catanzaro, Jacobs, Huang, Liu, and Balaji]{ge2023preserve}
Songwei Ge, Seungjun Nah, Guilin Liu, Tyler Poon, Andrew Tao, Bryan Catanzaro, David Jacobs, Jia-Bin Huang, Ming-Yu Liu, and Yogesh Balaji.
\newblock Preserve your own correlation: A noise prior for video diffusion models.
\newblock In \emph{ICCV}, 2023.

\bibitem[Ge et~al.(2024)Ge, Zeng, Huffman, Lin, Liu, and Cui]{ge2024visual}
Yunhao Ge, Xiaohui Zeng, Jacob~Samuel Huffman, Tsung-Yi Lin, Ming-Yu Liu, and Yin Cui.
\newblock Visual fact checker: Enabling high-fidelity detailed caption generation.
\newblock In \emph{CVPR}, 2024.

\bibitem[Ghosh et~al.(2024)Ghosh, Varshney, Galinkin, and Parisien]{aegis}
Shaona Ghosh, Prasoon Varshney, Erick Galinkin, and Christopher Parisien.
\newblock Aegis: Online adaptive ai content safety moderation with ensemble of llm experts.
\newblock \emph{arXiv preprint arXiv:2404.05993}, 2024.

\bibitem[Girdhar et~al.(2023)Girdhar, El-Nouby, Liu, Singh, Alwala, Joulin, and Misra]{girdhar2023imagebind}
Rohit Girdhar, Alaaeldin El-Nouby, Zhuang Liu, Mannat Singh, Kalyan~Vasudev Alwala, Armand Joulin, and Ishan Misra.
\newblock Imagebind: One embedding space to bind them all.
\newblock In \emph{CVPR}, 2023.

\bibitem[Girdhar et~al.(2024)Girdhar, Singh, Brown, Duval, Azadi, Rambhatla, Shah, Yin, Parikh, and Misra]{girdhar2023emu}
Rohit Girdhar, Mannat Singh, Andrew Brown, Quentin Duval, Samaneh Azadi, Sai~Saketh Rambhatla, Akbar Shah, Xi~Yin, Devi Parikh, and Ishan Misra.
\newblock Emu video: Factorizing text-to-video generation by explicit image conditioning.
\newblock In \emph{ECCV}, 2024.

\bibitem[Grauman et~al.(2024)Grauman, Westbury, Torresani, Kitani, Malik, Afouras, Ashutosh, Baiyya, Bansal, Boote, et~al.]{grauman2024egoexo4d}
Kristen Grauman, Andrew Westbury, Lorenzo Torresani, Kris Kitani, Jitendra Malik, Triantafyllos Afouras, Kumar Ashutosh, Vijay Baiyya, Siddhant Bansal, Bikram Boote, et~al.
\newblock Ego-exo4d: Understanding skilled human activity from first-and third-person perspectives.
\newblock In \emph{CVPR}, 2024.

\bibitem[Gupta et~al.(2024{\natexlab{a}})Gupta, Yu, Sohn, Gu, Hahn, Fei-Fei, Essa, Jiang, and Lezama]{gupta2023photorealistic}
Agrim Gupta, Lijun Yu, Kihyuk Sohn, Xiuye Gu, Meera Hahn, Li~Fei-Fei, Irfan Essa, Lu~Jiang, and Jos{\'e} Lezama.
\newblock Photorealistic video generation with diffusion models.
\newblock In \emph{ECCV}, 2024{\natexlab{a}}.

\bibitem[Gupta et~al.(2024{\natexlab{b}})Gupta, Yadav, Gal, Batra, Kira, Lu, and Rudner]{gupta2024pre}
Gunshi Gupta, Karmesh Yadav, Yarin Gal, Dhruv Batra, Zsolt Kira, Cong Lu, and Tim~GJ Rudner.
\newblock Pre-trained text-to-image diffusion models are versatile representation learners for control.
\newblock In \emph{ICLR Workshops}, 2024{\natexlab{b}}.

\bibitem[Gururani et~al.(2023)Gururani, Mallya, Wang, Valle, and Liu]{gururani2022SPACE}
Siddharth Gururani, Arun Mallya, Ting-Chun Wang, Rafael Valle, and Ming-Yu Liu.
\newblock {SPACE: Speech-driven Portrait Animation with Controllable Expression}.
\newblock In \emph{ICCV}, 2023.

\bibitem[Ha and Schmidhuber(2018)]{ha2018world}
David Ha and J{\"u}rgen Schmidhuber.
\newblock World models.
\newblock \emph{arXiv preprint arXiv:1803.10122}, 2018.

\bibitem[Hafner et~al.(2019)Hafner, Lillicrap, Ba, and Norouzi]{hafner2019dream}
Danijar Hafner, Timothy Lillicrap, Jimmy Ba, and Mohammad Norouzi.
\newblock Dream to control: Learning behaviors by latent imagination.
\newblock \emph{arXiv preprint arXiv:1912.01603}, 2019.

\bibitem[Hafner et~al.(2021)Hafner, Lillicrap, Norouzi, and Ba]{hafner2020mastering}
Danijar Hafner, Timothy Lillicrap, Mohammad Norouzi, and Jimmy Ba.
\newblock Mastering atari with discrete world models.
\newblock In \emph{ICLR}, 2021.

\bibitem[Hafner et~al.(2023)Hafner, Pasukonis, Ba, and Lillicrap]{hafner2023mastering}
Danijar Hafner, Jurgis Pasukonis, Jimmy Ba, and Timothy Lillicrap.
\newblock Mastering diverse domains through world models.
\newblock \emph{arXiv preprint arXiv:2301.04104}, 2023.

\bibitem[Hansen et~al.(2024)Hansen, Su, and Wang]{hansen2023td}
Nicklas Hansen, Hao Su, and Xiaolong Wang.
\newblock Td-mpc2: Scalable, robust world models for continuous control.
\newblock In \emph{ICLR}, 2024.

\bibitem[Hartley and Zisserman(2003)]{hartley2003multiple}
Richard Hartley and Andrew Zisserman.
\newblock \emph{Multiple view geometry in computer vision}.
\newblock Cambridge university press, 2003.

\bibitem[He et~al.(2024{\natexlab{a}})He, Xu, Guo, Wetzstein, Dai, Li, and Yang]{he2024cameractrl}
Hao He, Yinghao Xu, Yuwei Guo, Gordon Wetzstein, Bo~Dai, Hongsheng Li, and Ceyuan Yang.
\newblock Cameractrl: Enabling camera control for text-to-video generation.
\newblock \emph{arXiv preprint arXiv:2404.02101}, 2024{\natexlab{a}}.

\bibitem[He et~al.(2024{\natexlab{b}})He, Bai, Pan, Zhang, Zhao, and Li]{he2024learning}
Haoran He, Chenjia Bai, Ling Pan, Weinan Zhang, Bin Zhao, and Xuelong Li.
\newblock Learning an actionable discrete diffusion policy via large-scale actionless video pre-training.
\newblock In \emph{NeurIPS}, 2024{\natexlab{b}}.

\bibitem[He et~al.(2022)He, Chen, Xie, Li, Doll{\'a}r, and Girshick]{he2022masked}
Kaiming He, Xinlei Chen, Saining Xie, Yanghao Li, Piotr Doll{\'a}r, and Ross Girshick.
\newblock Masked autoencoders are scalable vision learners.
\newblock In \emph{CVPR}, 2022.

\bibitem[Heusel et~al.(2017)Heusel, Ramsauer, Unterthiner, Nessler, and Hochreiter]{FID}
Martin Heusel, Hubert Ramsauer, Thomas Unterthiner, Bernhard Nessler, and Sepp Hochreiter.
\newblock Gans trained by a two time-scale update rule converge to a local nash equilibrium.
\newblock In \emph{NeurIPS}, 2017.

\bibitem[Hinton et~al.(1995)Hinton, Dayan, Frey, and Neal]{hinton1995wake}
Geoffrey~E Hinton, Peter Dayan, Brendan~J Frey, and Radford~M Neal.
\newblock The "wake-sleep" algorithm for unsupervised neural networks.
\newblock \emph{Science}, 1995.

\bibitem[Ho and Salimans(2022)]{ho2022classifier}
Jonathan Ho and Tim Salimans.
\newblock Classifier-free diffusion guidance.
\newblock \emph{arXiv preprint arXiv:2207.12598}, 2022.

\bibitem[Ho et~al.(2020)Ho, Jain, and Abbeel]{ho2020denoising}
Jonathan Ho, Ajay Jain, and Pieter Abbeel.
\newblock Denoising diffusion probabilistic models.
\newblock In \emph{NeurIPS}, 2020.

\bibitem[Ho et~al.(2022)Ho, Salimans, Gritsenko, Chan, Norouzi, and Fleet]{ho2022video}
Jonathan Ho, Tim Salimans, Alexey Gritsenko, William Chan, Mohammad Norouzi, and David~J Fleet.
\newblock Video diffusion models.
\newblock In \emph{NeurIPS}, 2022.

\bibitem[Hong et~al.(2023)Hong, Ding, Zheng, Liu, and Tang]{hong2022cogvideo}
Wenyi Hong, Ming Ding, Wendi Zheng, Xinghan Liu, and Jie Tang.
\newblock Cogvideo: Large-scale pretraining for text-to-video generation via transformers.
\newblock In \emph{ICLR}, 2023.

\bibitem[Hoogeboom et~al.(2023)Hoogeboom, Heek, and Salimans]{hoogeboom2023simple}
Emiel Hoogeboom, Jonathan Heek, and Tim Salimans.
\newblock Simple diffusion: End-to-end diffusion for high resolution images.
\newblock In \emph{ICML}, 2023.

\bibitem[Hoogeboom et~al.(2024)Hoogeboom, Mensink, Heek, Lamerigts, Gao, and Salimans]{hoogeboom2024simpler}
Emiel Hoogeboom, Thomas Mensink, Jonathan Heek, Kay Lamerigts, Ruiqi Gao, and Tim Salimans.
\newblock Simpler diffusion (sid2): 1.5 fid on imagenet512 with pixel-space diffusion.
\newblock \emph{arXiv preprint arXiv:2410.19324}, 2024.

\bibitem[Hu et~al.(2023)Hu, Russell, Yeo, Murez, Fedoseev, Kendall, Shotton, and Corrado]{hu2023gaia}
Anthony Hu, Lloyd Russell, Hudson Yeo, Zak Murez, George Fedoseev, Alex Kendall, Jamie Shotton, and Gianluca Corrado.
\newblock Gaia-1: A generative world model for autonomous driving.
\newblock \emph{arXiv preprint arXiv:2309.17080}, 2023.

\bibitem[Hu et~al.(2022)Hu, yelong shen, Wallis, Allen-Zhu, Li, Wang, Wang, and Chen]{hu2021lora}
Edward~J Hu, yelong shen, Phillip Wallis, Zeyuan Allen-Zhu, Yuanzhi Li, Shean Wang, Lu~Wang, and Weizhu Chen.
\newblock Lo{RA}: Low-rank adaptation of large language models.
\newblock In \emph{ICLR}, 2022.

\bibitem[Hua et~al.(2024)Hua, Liu, Macaluso, Lin, Zhang, Xu, and Wang]{hua2024gensim2}
Pu~Hua, Minghuan Liu, Annabella Macaluso, Yunfeng Lin, Weinan Zhang, Huazhe Xu, and Lirui Wang.
\newblock Gensim2: Scaling robot data generation with multi-modal and reasoning llms.
\newblock \emph{arXiv preprint arXiv:2410.03645}, 2024.

\bibitem[Huang et~al.(2024)Huang, He, Yu, Zhang, Si, Jiang, Zhang, Wu, Jin, Chanpaisit, et~al.]{huang2024vbench}
Ziqi Huang, Yinan He, Jiashuo Yu, Fan Zhang, Chenyang Si, Yuming Jiang, Yuanhan Zhang, Tianxing Wu, Qingyang Jin, Nattapol Chanpaisit, et~al.
\newblock Vbench: Comprehensive benchmark suite for video generative models.
\newblock In \emph{CVPR}, 2024.

\bibitem[Inan et~al.(2023)Inan, Upasani, Chi, Rungta, Iyer, Mao, Tontchev, Hu, Fuller, Testuggine, et~al.]{llamaguard}
Hakan Inan, Kartikeya Upasani, Jianfeng Chi, Rashi Rungta, Krithika Iyer, Yuning Mao, Michael Tontchev, Qing Hu, Brian Fuller, Davide Testuggine, et~al.
\newblock Llama guard: Llm-based input-output safeguard for human-ai conversations.
\newblock \emph{arXiv preprint arXiv:2312.06674}, 2023.

\bibitem[Jacobs et~al.(2023)Jacobs, Tanaka, Zhang, Zhang, Song, Rajbhandari, and He]{jacobs2023deepspeed}
Sam~Ade Jacobs, Masahiro Tanaka, Chengming Zhang, Minjia Zhang, Shuaiwen~Leon Song, Samyam Rajbhandari, and Yuxiong He.
\newblock Deepspeed ulysses: System optimizations for enabling training of extreme long sequence transformer models.
\newblock \emph{arXiv preprint arXiv:2309.14509}, 2023.

\bibitem[Jia et~al.(2023)Jia, Mao, Liu, Zhao, Wen, Zhang, Zhang, and Wang]{jia2023adriver}
Fan Jia, Weixin Mao, Yingfei Liu, Yucheng Zhao, Yuqing Wen, Chi Zhang, Xiangyu Zhang, and Tiancai Wang.
\newblock Adriver-i: A general world model for autonomous driving.
\newblock \emph{arXiv preprint arXiv:2311.13549}, 2023.

\bibitem[Jiang et~al.(2023)Jiang, Sablayrolles, Mensch, Bamford, Chaplot, de~las Casas, Bressand, Lengyel, Lample, Saulnier, Lavaud, Lachaux, Stock, Scao, Lavril, Wang, Lacroix, and Sayed]{jiang2023mistral}
Albert~Q. Jiang, Alexandre Sablayrolles, Arthur Mensch, Chris Bamford, Devendra~Singh Chaplot, Diego de~las Casas, Florian Bressand, Gianna Lengyel, Guillaume Lample, Lucile Saulnier, Lélio~Renard Lavaud, Marie-Anne Lachaux, Pierre Stock, Teven~Le Scao, Thibaut Lavril, Thomas Wang, Timothée Lacroix, and William~El Sayed.
\newblock Mistral 7b.
\newblock \emph{arXiv preprint arXiv:2310.06825}, 2023.

\bibitem[Kang et~al.(2024)Kang, Yue, Lu, Lin, Zhao, Wang, Huang, and Feng]{kang2024how}
Bingyi Kang, Yang Yue, Rui Lu, Zhijie Lin, Yang Zhao, Kaixin Wang, Gao Huang, and Jiashi Feng.
\newblock How far is video generation from world model? -- a physical law perspective.
\newblock \emph{arXiv preprint arXiv:2411.02385}, 2024.

\bibitem[Kaplan et~al.(2020)Kaplan, McCandlish, Henighan, Brown, Chess, Child, Gray, Radford, Wu, and Amodei]{kaplan2020scaling}
Jared Kaplan, Sam McCandlish, Tom Henighan, Tom~B Brown, Benjamin Chess, Rewon Child, Scott Gray, Alec Radford, Jeffrey Wu, and Dario Amodei.
\newblock Scaling laws for neural language models.
\newblock \emph{arXiv preprint arXiv:2001.08361}, 2020.

\bibitem[Karras et~al.(2020)Karras, Laine, Aittala, Hellsten, Lehtinen, and Aila]{karras2020analyzing}
Tero Karras, Samuli Laine, Miika Aittala, Janne Hellsten, Jaakko Lehtinen, and Timo Aila.
\newblock Analyzing and improving the image quality of stylegan.
\newblock In \emph{CVPR}, 2020.

\bibitem[Karras et~al.(2022)Karras, Aittala, Aila, and Laine]{Karras2022edm}
Tero Karras, Miika Aittala, Timo Aila, and Samuli Laine.
\newblock Elucidating the design space of diffusion-based generative models.
\newblock In \emph{NeurIPS}, 2022.

\bibitem[Karras et~al.(2024)Karras, Aittala, Lehtinen, Hellsten, Aila, and Laine]{Karras2024edm2}
Tero Karras, Miika Aittala, Jaakko Lehtinen, Janne Hellsten, Timo Aila, and Samuli Laine.
\newblock Analyzing and improving the training dynamics of diffusion models.
\newblock In \emph{CVPR}, 2024.

\bibitem[Ke et~al.(2024)Ke, Gkanatsios, and Fragkiadaki]{ke20243d}
Tsung-Wei Ke, Nikolaos Gkanatsios, and Katerina Fragkiadaki.
\newblock 3d diffuser actor: Policy diffusion with 3d scene representations.
\newblock In \emph{CoRL}, 2024.

\bibitem[Kerbl et~al.(2023)Kerbl, Kopanas, Leimk{\"u}hler, and Drettakis]{kerbl20233d}
Bernhard Kerbl, Georgios Kopanas, Thomas Leimk{\"u}hler, and George Drettakis.
\newblock 3d gaussian splatting for real-time radiance field rendering.
\newblock \emph{ACM Transactions on Graphics (TOG)}, 2023.

\bibitem[Khanam and Hussain(2024)]{khanam2024yolov11}
Rahima Khanam and Muhammad Hussain.
\newblock Yolov11: An overview of the key architectural enhancements.
\newblock \emph{arXiv preprint arXiv:2410.17725}, 2024.

\bibitem[Kim et~al.(2024)Kim, Pertsch, Karamcheti, Xiao, Balakrishna, Nair, Rafailov, Foster, Lam, Sanketi, Vuong, Kollar, Burchfiel, Tedrake, Sadigh, Levine, Liang, and Finn]{kim24openvla}
{Moo Jin} Kim, Karl Pertsch, Siddharth Karamcheti, Ted Xiao, Ashwin Balakrishna, Suraj Nair, Rafael Rafailov, Ethan Foster, Grace Lam, Pannag Sanketi, Quan Vuong, Thomas Kollar, Benjamin Burchfiel, Russ Tedrake, Dorsa Sadigh, Sergey Levine, Percy Liang, and Chelsea Finn.
\newblock {OpenVLA}: An open-source vision-language-action model.
\newblock \emph{arXiv preprint arXiv:2406.09246}, 2024.

\bibitem[Kim et~al.(2020)Kim, Zhou, Philion, Torralba, and Fidler]{Kim2020_GameGan}
Seung~Wook Kim, Yuhao Zhou, Jonah Philion, Antonio Torralba, and Sanja Fidler.
\newblock {Learning to Simulate Dynamic Environments with GameGAN}.
\newblock In \emph{CVPR}, 2020.

\bibitem[Kim et~al.(2021)Kim, , Philion, Torralba, and Fidler]{Kim2021_DriveGAN}
Seung~Wook Kim, , Jonah Philion, Antonio Torralba, and Sanja Fidler.
\newblock {DriveGAN: Towards a Controllable High-Quality Neural Simulation}.
\newblock In \emph{CVPR}, 2021.

\bibitem[Kingma(2013)]{kingma2013auto}
Diederik~P Kingma.
\newblock Auto-encoding variational bayes.
\newblock \emph{arXiv preprint arXiv:1312.6114}, 2013.

\bibitem[Ko et~al.(2024)Ko, Mao, Du, Sun, and Tenenbaum]{kolearning}
Po-Chen Ko, Jiayuan Mao, Yilun Du, Shao-Hua Sun, and Joshua~B Tenenbaum.
\newblock Learning to act from actionless videos through dense correspondences.
\newblock In \emph{ICLR}, 2024.

\bibitem[Kondratyuk et~al.(2024)Kondratyuk, Yu, Gu, Lezama, Huang, Schindler, Hornung, Birodkar, Yan, Chiu, et~al.]{videopoet}
Dan Kondratyuk, Lijun Yu, Xiuye Gu, Jos{\'e} Lezama, Jonathan Huang, Grant Schindler, Rachel Hornung, Vighnesh Birodkar, Jimmy Yan, Ming-Chang Chiu, et~al.
\newblock Videopoet: A large language model for zero-shot video generation.
\newblock In \emph{ICML}, 2024.

\bibitem[Kong et~al.(2024)Kong, Tian, Zhang, Min, Dai, Zhou, Xiong, Li, Wu, Zhang, et~al.]{kong2024hunyuanvideo}
Weijie Kong, Qi~Tian, Zijian Zhang, Rox Min, Zuozhuo Dai, Jin Zhou, Jiangfeng Xiong, Xin Li, Bo~Wu, Jianwei Zhang, et~al.
\newblock Hunyuanvideo: A systematic framework for large video generative models.
\newblock \emph{arXiv preprint arXiv:2412.03603}, 2024.

\bibitem[Korthikanti et~al.(2023)Korthikanti, Casper, Lym, McAfee, Andersch, Shoeybi, and Catanzaro]{korthikanti2023reducing}
Vijay~Anand Korthikanti, Jared Casper, Sangkug Lym, Lawrence McAfee, Michael Andersch, Mohammad Shoeybi, and Bryan Catanzaro.
\newblock Reducing activation recomputation in large transformer models.
\newblock \emph{Proceedings of Machine Learning and Systems}, 2023.

\bibitem[Ku et~al.(2024)Ku, Wei, Ren, Yang, and Chen]{ku2024anyv2v}
Max Ku, Cong Wei, Weiming Ren, Huan Yang, and Wenhu Chen.
\newblock Anyv2v: A tuning-free framework for any video-to-video editing tasks.
\newblock \emph{TMLR}, 2024.

\bibitem[KuaiShou(2024)]{kling}
KuaiShou.
\newblock Kling, 2024.
\newblock URL \url{https://klingai.com/}.

\bibitem[Lee et~al.(2022)Lee, Kim, Kim, Cho, and Han]{lee2022rvq}
Doyup Lee, Chiheon Kim, Saehoon Kim, Minsu Cho, and Wook-Shin Han.
\newblock Autoregressive image generation using residual quantization.
\newblock In \emph{CVPR}, 2022.

\bibitem[Lei~Ba et~al.(2016)Lei~Ba, Kiros, and Hinton]{lei2016layer}
Jimmy Lei~Ba, Jamie~Ryan Kiros, and Geoffrey~E Hinton.
\newblock Layer normalization.
\newblock \emph{arXiv preprint arXiv:1607.06450}, 2016.

\bibitem[Leviathan et~al.(2023)Leviathan, Kalman, and Matias]{leviathan2023fast}
Yaniv Leviathan, Matan Kalman, and Yossi Matias.
\newblock Fast inference from transformers via speculative decoding.
\newblock In \emph{ICML}, 2023.

\bibitem[Li et~al.(2024)Li, Tan, Zhang, Xu, Luan, Xu, Hong, Sunkavalli, Shakhnarovich, and Bi]{li2024instant3d}
Jiahao Li, Hao Tan, Kai Zhang, Zexiang Xu, Fujun Luan, Yinghao Xu, Yicong Hong, Kalyan Sunkavalli, Greg Shakhnarovich, and Sai Bi.
\newblock Instant3d: Fast text-to-3d with sparse-view generation and large reconstruction model.
\newblock In \emph{ICLR}, 2024.

\bibitem[Li et~al.(2023)Li, M\"uller, Evans, Taylor, Unberath, Liu, and Lin]{li2023neuralangelo}
Zhaoshuo Li, Thomas M\"uller, Alex Evans, Russell~H Taylor, Mathias Unberath, Ming-Yu Liu, and Chen-Hsuan Lin.
\newblock Neuralangelo: High-fidelity neural surface reconstruction.
\newblock In \emph{CVPR}, 2023.

\bibitem[Liang et~al.(2024)Liang, Ren, Mirzaei, Torralba, Liu, Gilitschenski, Fidler, Oztireli, Ling, Gojcic, et~al.]{liang2024feed}
Hanxue Liang, Jiawei Ren, Ashkan Mirzaei, Antonio Torralba, Ziwei Liu, Igor Gilitschenski, Sanja Fidler, Cengiz Oztireli, Huan Ling, Zan Gojcic, et~al.
\newblock Feed-forward bullet-time reconstruction of dynamic scenes from monocular videos.
\newblock \emph{arXiv preprint arXiv:2412.03526}, 2024.

\bibitem[Lin et~al.(2024{\natexlab{a}})Lin, Ge, Cheng, Li, Zhu, Wang, He, Ye, Yuan, Chen, et~al.]{lin2024open}
Bin Lin, Yunyang Ge, Xinhua Cheng, Zongjian Li, Bin Zhu, Shaodong Wang, Xianyi He, Yang Ye, Shenghai Yuan, Liuhan Chen, et~al.
\newblock Open-sora plan: Open-source large video generation model.
\newblock \emph{arXiv preprint arXiv:2412.00131}, 2024{\natexlab{a}}.

\bibitem[Lin et~al.(2021)Lin, Ma, Torralba, and Lucey]{lin2021barf}
Chen-Hsuan Lin, Wei-Chiu Ma, Antonio Torralba, and Simon Lucey.
\newblock Barf: Bundle-adjusting neural radiance fields.
\newblock In \emph{ICCV}, 2021.

\bibitem[Lin et~al.(2023{\natexlab{a}})Lin, Gao, Tang, Takikawa, Zeng, Huang, Kreis, Fidler, Liu, and Lin]{lin2023magic3d}
Chen-Hsuan Lin, Jun Gao, Luming Tang, Towaki Takikawa, Xiaohui Zeng, Xun Huang, Karsten Kreis, Sanja Fidler, Ming-Yu Liu, and Tsung-Yi Lin.
\newblock Magic3d: High-resolution text-to-3d content creation.
\newblock In \emph{CVPR}, 2023{\natexlab{a}}.

\bibitem[Lin et~al.(2024{\natexlab{b}})Lin, Yin, Ping, Molchanov, Shoeybi, and Han]{lin2024vilapretrainingvisuallanguage}
Ji~Lin, Hongxu Yin, Wei Ping, Pavlo Molchanov, Mohammad Shoeybi, and Song Han.
\newblock Vila: On pre-training for visual language models.
\newblock In \emph{CVPR}, 2024{\natexlab{b}}.

\bibitem[Lin et~al.(2023{\natexlab{b}})Lin, Lin, Lai, Lin, Shih, and Ramamoorthi]{lin2023vision}
Kai-En Lin, Yen-Chen Lin, Wei-Sheng Lai, Tsung-Yi Lin, Yi-Chang Shih, and Ravi Ramamoorthi.
\newblock Vision transformer for nerf-based view synthesis from a single input image.
\newblock In \emph{WACV}, 2023{\natexlab{b}}.

\bibitem[Lin et~al.(2014)Lin, Maire, Belongie, Hays, Perona, Ramanan, Doll{\'a}r, and Zitnick]{lin2015mscoco}
Tsung-Yi Lin, Michael Maire, Serge Belongie, James Hays, Pietro Perona, Deva Ramanan, Piotr Doll{\'a}r, and C~Lawrence Zitnick.
\newblock Microsoft coco: Common objects in context.
\newblock In \emph{ECCV}, 2014.

\bibitem[Lindenberger et~al.(2023)Lindenberger, Sarlin, and Pollefeys]{lindenberger2023lightglue}
Philipp Lindenberger, Paul-Edouard Sarlin, and Marc Pollefeys.
\newblock Lightglue: Local feature matching at light speed.
\newblock In \emph{ICCV}, 2023.

\bibitem[Ling et~al.(2024)Ling, Sheng, Tu, Zhao, Xin, Wan, Yu, Guo, Yu, Lu, et~al.]{ling2024dl3dv}
Lu~Ling, Yichen Sheng, Zhi Tu, Wentian Zhao, Cheng Xin, Kun Wan, Lantao Yu, Qianyu Guo, Zixun Yu, Yawen Lu, et~al.
\newblock Dl3dv-10k: A large-scale scene dataset for deep learning-based 3d vision.
\newblock In \emph{CVPR}, 2024.

\bibitem[Lipman et~al.(2022)Lipman, Chen, Ben-Hamu, Nickel, and Le]{lipman2022flow}
Yaron Lipman, Ricky~TQ Chen, Heli Ben-Hamu, Maximilian Nickel, and Matt Le.
\newblock Flow matching for generative modeling.
\newblock \emph{arXiv preprint arXiv:2210.02747}, 2022.

\bibitem[Liu et~al.(2024{\natexlab{a}})Liu, Li, Zhang, Lan, and Liu]{liu2024stablev2v}
Chang Liu, Rui Li, Kaidong Zhang, Yunwei Lan, and Dong Liu.
\newblock Stablev2v: Stablizing shape consistency in video-to-video editing.
\newblock \emph{arXiv preprint arXiv:2411.11045}, 2024{\natexlab{a}}.

\bibitem[Liu et~al.(2023{\natexlab{a}})Liu, Zaharia, and Abbeel]{liu2023ring}
Hao Liu, Matei Zaharia, and Pieter Abbeel.
\newblock Ring attention with blockwise transformers for near-infinite context.
\newblock \emph{arXiv preprint arXiv:2310.01889}, 2023{\natexlab{a}}.

\bibitem[Liu et~al.(2024{\natexlab{b}})Liu, Yan, Zaharia, and Abbeel]{liu2024world}
Hao Liu, Wilson Yan, Matei Zaharia, and Pieter Abbeel.
\newblock World model on million-length video and language with blockwise ringattention.
\newblock \emph{CoRR}, 2024{\natexlab{b}}.

\bibitem[Liu et~al.(2024{\natexlab{c}})Liu, Liu, Zhou, Xu, Xie, Han, P{\'e}rez, Liu, Kahatapitiya, Jia, et~al.]{liu2024mardini}
Haozhe Liu, Shikun Liu, Zijian Zhou, Mengmeng Xu, Yanping Xie, Xiao Han, Juan~C P{\'e}rez, Ding Liu, Kumara Kahatapitiya, Menglin Jia, et~al.
\newblock Mardini: Masked autoregressive diffusion for video generation at scale.
\newblock \emph{arXiv preprint arXiv:2410.20280}, 2024{\natexlab{c}}.

\bibitem[Liu et~al.(2023{\natexlab{b}})Liu, Wu, Van~Hoorick, Tokmakov, Zakharov, and Vondrick]{liu2023zero}
Ruoshi Liu, Rundi Wu, Basile Van~Hoorick, Pavel Tokmakov, Sergey Zakharov, and Carl Vondrick.
\newblock Zero-1-to-3: Zero-shot one image to 3d object.
\newblock In \emph{ICCV}, 2023{\natexlab{b}}.

\bibitem[Loshchilov and Hutter(2019)]{loshchilov2019decoupledweightdecayregularization}
Ilya Loshchilov and Frank Hutter.
\newblock Decoupled weight decay regularization.
\newblock In \emph{ICLR}, 2019.

\bibitem[Lu et~al.(2025)Lu, Huang, Yang, Zhang, and Zhang]{lu2025wovogen}
Jiachen Lu, Ze~Huang, Zeyu Yang, Jiahui Zhang, and Li~Zhang.
\newblock Wovogen: World volume-aware diffusion for controllable multi-camera driving scene generation.
\newblock In \emph{ECCV}, 2025.

\bibitem[Luma(2024)]{dreammachine}
Luma.
\newblock Dream machine, 2024.
\newblock URL \url{https://lumalabs.ai/dream-machine}.

\bibitem[Luo et~al.(2024)Luo, Shi, Ge, Yang, Wang, and Shan]{luo2024open_magvit2}
Zhuoyan Luo, Fengyuan Shi, Yixiao Ge, Yujiu Yang, Limin Wang, and Ying Shan.
\newblock Open-magvit2: An open-source project toward democratizing auto-regressive visual generation.
\newblock \emph{arXiv preprint arXiv:2409.04410}, 2024.

\bibitem[Ma et~al.(2024)Ma, Wang, Jia, Chen, Liu, Li, Chen, and Qiao]{ma2024latte}
Xin Ma, Yaohui Wang, Gengyun Jia, Xinyuan Chen, Ziwei Liu, Yuan-Fang Li, Cunjian Chen, and Yu~Qiao.
\newblock Latte: Latent diffusion transformer for video generation.
\newblock \emph{arXiv preprint arXiv:2401.03048}, 2024.

\bibitem[Mallya et~al.(2020)Mallya, Wang, Sapra, and Liu]{mallya2020world}
Arun Mallya, Ting-Chun Wang, Karan Sapra, and Ming-Yu Liu.
\newblock World-consistent video-to-video synthesis.
\newblock In \emph{ECCV}, 2020.

\bibitem[Mallya et~al.(2022)Mallya, Wang, and Liu]{mallya2022implicit}
Arun Mallya, Ting-Chun Wang, and Ming-Yu Liu.
\newblock {Implicit Warping for Animation with Image Sets}.
\newblock In \emph{NeurIPS}, 2022.

\bibitem[Mentzer et~al.(2023)Mentzer, Minnen, Agustsson, and Tschannen]{mentzer2023finite}
Fabian Mentzer, David Minnen, Eirikur Agustsson, and Michael Tschannen.
\newblock Finite scalar quantization: Vq-vae made simple.
\newblock \emph{arXiv preprint arXiv:2309.15505}, 2023.

\bibitem[Micheli et~al.(2023)Micheli, Alonso, and Fleuret]{micheli2022transformers}
Vincent Micheli, Eloi Alonso, and Fran{\c{c}}ois Fleuret.
\newblock Transformers are sample-efficient world models.
\newblock In \emph{ICLR}, 2023.

\bibitem[Mildenhall et~al.(2020)Mildenhall, Srinivasan, Tancik, Barron, Ramamoorthi, and Ng]{mildenhall2020nerf}
Ben Mildenhall, Pratul~P Srinivasan, Matthew Tancik, Jonathan~T Barron, Ravi Ramamoorthi, and Ren Ng.
\newblock Nerf: Representing scenes as neural radiance fields for view synthesis.
\newblock In \emph{ECCV}, 2020.

\bibitem[Miller(1995)]{wordnet}
George~A Miller.
\newblock Wordnet: a lexical database for english.
\newblock \emph{Communications of the ACM}, 1995.

\bibitem[Mistral and NVIDIA(2024)]{mistral_nemo_2024}
Mistral and NVIDIA.
\newblock Mistral-nemo-12b-instruct: A 12b parameter large language model, 2024.
\newblock URL \url{https://mistral.ai/news/mistral-nemo/}.

\bibitem[Moritz et~al.(2017)Moritz, Nishihara, Wang, Tumanov, Liaw, Liang, Paul, Jordan, and Stoica]{moritz2017ray}
Philipp Moritz, Robert Nishihara, Stephanie Wang, Alexey Tumanov, Richard Liaw, Eric Liang, William Paul, Michael~I. Jordan, and Ion Stoica.
\newblock Ray: {A} distributed framework for emerging {AI} applications.
\newblock \emph{CoRR}, abs/1712.05889, 2017.
\newblock URL \url{http://arxiv.org/abs/1712.05889}.

\bibitem[Murray et~al.(2017)Murray, Li, and Sastry]{murray2017mathematical}
Richard~M Murray, Zexiang Li, and S~Shankar Sastry.
\newblock \emph{A mathematical introduction to robotic manipulation}.
\newblock CRC press, 2017.

\bibitem[Nasiriany et~al.(2024)Nasiriany, Maddukuri, Zhang, Parikh, Lo, Joshi, Mandlekar, and Zhu]{nasiriany2024robocasa}
Soroush Nasiriany, Abhiram Maddukuri, Lance Zhang, Adeet Parikh, Aaron Lo, Abhishek Joshi, Ajay Mandlekar, and Yuke Zhu.
\newblock Robocasa: Large-scale simulation of everyday tasks for generalist robots.
\newblock \emph{arXiv preprint arXiv:2406.02523}, 2024.

\bibitem[NVIDIA(2024{\natexlab{a}})]{IsaacSim}
NVIDIA.
\newblock Isaac sim, 2024{\natexlab{a}}.
\newblock URL \url{https://developer.nvidia.com/isaac/sim}.

\bibitem[NVIDIA(2024{\natexlab{b}})]{Omniverse}
NVIDIA.
\newblock Omniverse, 2024{\natexlab{b}}.
\newblock URL \url{https://www.nvidia.com/en-us/omniverse/}.

\bibitem[NVIDIA(2024{\natexlab{c}})]{PhysX}
NVIDIA.
\newblock Physx, 2024{\natexlab{c}}.
\newblock URL \url{https://github.com/NVIDIA-Omniverse/PhysX}.

\bibitem[NVIDIA(2024{\natexlab{d}})]{nvidia2024edify3d}
NVIDIA.
\newblock Edify 3d: Scalable high-quality 3d asset generation.
\newblock \emph{arXiv preprint arXiv:2411.07135}, 2024{\natexlab{d}}.

\bibitem[NVIDIA(2024{\natexlab{e}})]{nvidia_transformer_engine}
NVIDIA.
\newblock Transformer engine, 2024{\natexlab{e}}.
\newblock URL \url{https://github.com/NVIDIA/TransformerEngine}.

\bibitem[OpenAI(2022)]{tiktoken}
OpenAI.
\newblock Tiktoken, 2022.
\newblock URL \url{https://github.com/openai/tiktoken}.

\bibitem[OpenAI(2024{\natexlab{a}})]{openai2024dalle3}
OpenAI.
\newblock Dall·e 3, 2024{\natexlab{a}}.
\newblock URL \url{https://openai.com/dall-e}.
\newblock Accessed: [Insert access date here].

\bibitem[OpenAI(2024{\natexlab{b}})]{sora}
OpenAI.
\newblock Sora, 2024{\natexlab{b}}.
\newblock URL \url{https://openai.com/sora/}.

\bibitem[Pan et~al.(2025)Pan, Bar{\'a}th, Pollefeys, and Sch{\"o}nberger]{pan2025global}
Linfei Pan, D{\'a}niel Bar{\'a}th, Marc Pollefeys, and Johannes~L Sch{\"o}nberger.
\newblock Global structure-from-motion revisited.
\newblock In \emph{ECCV}, 2025.

\bibitem[Paszke et~al.(2019)Paszke, Gross, Massa, Lerer, Bradbury, Chanan, Killeen, Lin, Gimelshein, Antiga, et~al.]{paszke2019pytorch}
Adam Paszke, Sam Gross, Francisco Massa, Adam Lerer, James Bradbury, Gregory Chanan, Trevor Killeen, Zeming Lin, Natalia Gimelshein, Luca Antiga, et~al.
\newblock Pytorch: An imperative style, high-performance deep learning library.
\newblock \emph{Advances in neural information processing systems}, 32, 2019.

\bibitem[Peebles and Xie(2023)]{peebles2023scalable}
William Peebles and Saining Xie.
\newblock Scalable diffusion models with transformers.
\newblock In \emph{ICCV}, 2023.

\bibitem[Peng and Quesnelle(2023)]{bloc97}
Bowen Peng and Jeffrey Quesnelle.
\newblock Ntk-aware scaled rope allows llama models to have extended (8k+) context size without any fine-tuning and minimal perplexity degradation, 2023.

\bibitem[Peng et~al.(2023)Peng, Quesnelle, Fan, and Shippole]{peng2023yarn}
Bowen Peng, Jeffrey Quesnelle, Honglu Fan, and Enrico Shippole.
\newblock Yarn: Efficient context window extension of large language models.
\newblock \emph{arXiv preprint arXiv:2309.00071}, 2023.

\bibitem[Perazzi et~al.(2016)Perazzi, Pont-Tuset, McWilliams, Van~Gool, Gross, and Sorkine-Hornung]{perazzi2016benchmark}
Federico Perazzi, Jordi Pont-Tuset, Brian McWilliams, Luc Van~Gool, Markus Gross, and Alexander Sorkine-Hornung.
\newblock A benchmark dataset and evaluation methodology for video object segmentation.
\newblock In \emph{CVPR}, 2016.

\bibitem[Podell et~al.(2024)Podell, English, Lacey, Blattmann, Dockhorn, M{\"u}ller, Penna, and Rombach]{podell2024sdxl}
Dustin Podell, Zion English, Kyle Lacey, Andreas Blattmann, Tim Dockhorn, Jonas M{\"u}ller, Joe Penna, and Robin Rombach.
\newblock {SDXL}: Improving latent diffusion models for high-resolution image synthesis.
\newblock In \emph{ICLR}, 2024.

\bibitem[Polyak et~al.(2024)Polyak, Zohar, Brown, Tjandra, Sinha, Lee, Vyas, Shi, Ma, Chuang, et~al.]{polyak2024movie}
Adam Polyak, Amit Zohar, Andrew Brown, Andros Tjandra, Animesh Sinha, Ann Lee, Apoorv Vyas, Bowen Shi, Chih-Yao Ma, Ching-Yao Chuang, et~al.
\newblock Movie gen: A cast of media foundation models.
\newblock \emph{arXiv preprint arXiv:2410.13720}, 2024.

\bibitem[Poole et~al.(2023)Poole, Jain, Barron, and Mildenhall]{poole2023dreamfusion}
Ben Poole, Ajay Jain, Jonathan~T Barron, and Ben Mildenhall.
\newblock Dreamfusion: Text-to-3d using 2d diffusion.
\newblock In \emph{ICLR}, 2023.

\bibitem[Prasad et~al.(2024)Prasad, Lin, Wu, Zhou, and Bohg]{prasad2024consistency}
Aaditya Prasad, Kevin Lin, Jimmy Wu, Linqi Zhou, and Jeannette Bohg.
\newblock Consistency policy: Accelerated visuomotor policies via consistency distillation.
\newblock \emph{arXiv preprint arXiv:2405.07503}, 2024.

\bibitem[Qian et~al.(2024)Qian, Mai, Hamdi, Ren, Siarohin, Li, Lee, Skorokhodov, Wonka, Tulyakov, et~al.]{qian2024magic123}
Guocheng Qian, Jinjie Mai, Abdullah Hamdi, Jian Ren, Aliaksandr Siarohin, Bing Li, Hsin-Ying Lee, Ivan Skorokhodov, Peter Wonka, Sergey Tulyakov, et~al.
\newblock Magic123: One image to high-quality 3d object generation using both 2d and 3d diffusion priors.
\newblock In \emph{ICLR}, 2024.

\bibitem[Raffel et~al.(2020)Raffel, Shazeer, Roberts, Lee, Narang, Matena, Zhou, Li, and Liu]{raffel2020exploring}
Colin Raffel, Noam Shazeer, Adam Roberts, Katherine Lee, Sharan Narang, Michael Matena, Yanqi Zhou, Wei Li, and Peter~J Liu.
\newblock Exploring the limits of transfer learning with a unified text-to-text transformer.
\newblock \emph{JMLR}, 2020.

\bibitem[Ramachandran et~al.(2017)Ramachandran, Zoph, and Le]{ramachandran2017searching}
Prajit Ramachandran, Barret Zoph, and Quoc~V Le.
\newblock Searching for activation functions.
\newblock \emph{arXiv preprint arXiv:1710.05941}, 2017.

\bibitem[Ramesh et~al.(2021)Ramesh, Pavlov, Goh, Gray, Voss, Radford, Chen, and Sutskever]{dalle}
Aditya Ramesh, Mikhail Pavlov, Gabriel Goh, Scott Gray, Chelsea Voss, Alec Radford, Mark Chen, and Ilya Sutskever.
\newblock Zero-shot text-to-image generation.
\newblock In \emph{ICML}, 2021.

\bibitem[Ramesh et~al.(2022)Ramesh, Dhariwal, Nichol, Chu, and Chen]{DALLE2}
Aditya Ramesh, Prafulla Dhariwal, Alex Nichol, Casey Chu, and Mark Chen.
\newblock Hierarchical text-conditional image generation with clip latents.
\newblock \emph{arXiv preprint arXiv:2204.06125}, 2022.

\bibitem[RAPIDS(2023)]{RAPIDS}
RAPIDS.
\newblock Rapids: Libraries for end to end gpu data science, 2023.
\newblock URL \url{https://rapids.ai}.

\bibitem[Ren et~al.(2024)Ren, Yang, Zhang, Wei, Du, Huang, and Chen]{ren2024consisti2v}
Weiming Ren, Huan Yang, Ge~Zhang, Cong Wei, Xinrun Du, Wenhao Huang, and Wenhu Chen.
\newblock Consisti2v: Enhancing visual consistency for image-to-video generation.
\newblock \emph{arXiv preprint arXiv:2402.04324}, 2024.

\bibitem[Robine et~al.(2023)Robine, H{\"o}ftmann, Uelwer, and Harmeling]{robine2023transformer}
Jan Robine, Marc H{\"o}ftmann, Tobias Uelwer, and Stefan Harmeling.
\newblock Transformer-based world models are happy with 100k interactions.
\newblock \emph{arXiv preprint arXiv:2303.07109}, 2023.

\bibitem[Rombach et~al.(2021)Rombach, Esser, and Ommer]{rombach2021geometry}
Robin Rombach, Patrick Esser, and Bj{\"o}rn Ommer.
\newblock Geometry-free view synthesis: Transformers and no 3d priors.
\newblock In \emph{ICCV}, 2021.

\bibitem[Rombach et~al.(2022)Rombach, Blattmann, Lorenz, Esser, and Ommer]{rombach2022high}
Robin Rombach, Andreas Blattmann, Dominik Lorenz, Patrick Esser, and Bj{\"o}rn Ommer.
\newblock High-resolution image synthesis with latent diffusion models.
\newblock In \emph{CVPR}, 2022.

\bibitem[Runway(2024)]{gen3}
Runway.
\newblock Gen 3, 2024.
\newblock URL \url{https://runwayml.com/research/introducing-gen-3-alpha}.

\bibitem[Sadat et~al.(2024)Sadat, Buhmann, Bradley, Hilliges, and Weber]{sadat2024litevae}
Seyedmorteza Sadat, Jakob Buhmann, Derek Bradley, Otmar Hilliges, and Romann~M Weber.
\newblock Litevae: Lightweight and efficient variational autoencoders for latent diffusion models.
\newblock \emph{arXiv preprint arXiv:2405.14477}, 2024.

\bibitem[Saharia et~al.(2022)Saharia, Chan, Saxena, Li, Whang, Denton, Ghasemipour, Gontijo~Lopes, Karagol~Ayan, Salimans, et~al.]{saharia2022photorealistic}
Chitwan Saharia, William Chan, Saurabh Saxena, Lala Li, Jay Whang, Emily~L Denton, Kamyar Ghasemipour, Raphael Gontijo~Lopes, Burcu Karagol~Ayan, Tim Salimans, et~al.
\newblock Photorealistic text-to-image diffusion models with deep language understanding.
\newblock In \emph{NeurIPS}, 2022.

\bibitem[Sajjadi et~al.(2022)Sajjadi, Meyer, Pot, Bergmann, Greff, Radwan, Vora, Lu{\v{c}}i{\'c}, Duckworth, Dosovitskiy, et~al.]{sajjadi2022scene}
Mehdi~SM Sajjadi, Henning Meyer, Etienne Pot, Urs Bergmann, Klaus Greff, Noha Radwan, Suhani Vora, Mario Lu{\v{c}}i{\'c}, Daniel Duckworth, Alexey Dosovitskiy, et~al.
\newblock Scene representation transformer: Geometry-free novel view synthesis through set-latent scene representations.
\newblock In \emph{CVPR}, 2022.

\bibitem[Sampson(1982)]{sampson1982fitting}
Paul~D Sampson.
\newblock Fitting conic sections to “very scattered” data: An iterative refinement of the bookstein algorithm.
\newblock \emph{Computer graphics and image processing}, 1982.

\bibitem[Sch{\"o}nberger et~al.(2016)Sch{\"o}nberger, Zheng, Frahm, and Pollefeys]{schonberger2016pixelwise}
Johannes~L Sch{\"o}nberger, Enliang Zheng, Jan-Michael Frahm, and Marc Pollefeys.
\newblock Pixelwise view selection for unstructured multi-view stereo.
\newblock In \emph{ECCV}, 2016.

\bibitem[Sch\"{o}nberger and Frahm(2016)]{schoenberger2016sfm}
Johannes~Lutz Sch\"{o}nberger and Jan-Michael Frahm.
\newblock Structure-from-motion revisited.
\newblock In \emph{CVPR}, 2016.

\bibitem[Schuhmann(2022)]{schuhmann2022}
Christoph Schuhmann.
\newblock {Improved Aesthetic Predictor}, 2022.
\newblock URL \url{https://github.com/christophschuhmann/improved-aesthetic-predictor}.

\bibitem[Shi et~al.(2023)Shi, Wang, Ye, Long, Li, and Yang]{shi2023mvdream}
Yichun Shi, Peng Wang, Jianglong Ye, Mai Long, Kejie Li, and Xiao Yang.
\newblock Mvdream: Multi-view diffusion for 3d generation.
\newblock \emph{arXiv preprint arXiv:2308.16512}, 2023.

\bibitem[Shoeybi et~al.(2019)Shoeybi, Patwary, Puri, LeGresley, Casper, and Catanzaro]{shoeybi2019megatron}
Mohammad Shoeybi, Mostofa Patwary, Raul Puri, Patrick LeGresley, Jared Casper, and Bryan Catanzaro.
\newblock Megatron-lm: Training multi-billion parameter language models using model parallelism.
\newblock \emph{arXiv preprint arXiv:1909.08053}, 2019.

\bibitem[Simonyan and Zisserman(2014)]{simonyan2014very}
Karen Simonyan and Andrew Zisserman.
\newblock Very deep convolutional networks for large-scale image recognition.
\newblock \emph{arXiv preprint arXiv:1409.1556}, 2014.

\bibitem[Sitzmann et~al.(2021)Sitzmann, Rezchikov, Freeman, Tenenbaum, and Durand]{sitzmann2021light}
Vincent Sitzmann, Semon Rezchikov, Bill Freeman, Josh Tenenbaum, and Fredo Durand.
\newblock Light field networks: Neural scene representations with single-evaluation rendering.
\newblock In \emph{NeurIPS}, 2021.

\bibitem[Song et~al.(2020)Song, Sohl-Dickstein, Kingma, Kumar, Ermon, and Poole]{song2020score}
Yang Song, Jascha Sohl-Dickstein, Diederik~P Kingma, Abhishek Kumar, Stefano Ermon, and Ben Poole.
\newblock Score-based generative modeling through stochastic differential equations.
\newblock \emph{arXiv preprint arXiv:2011.13456}, 2020.

\bibitem[Soucek and Lokoc(2024)]{soucek2020transnetv2}
Tom{\'a}s Soucek and Jakub Lokoc.
\newblock Transnet v2: An effective deep network architecture for fast shot transition detection.
\newblock In \emph{ACM MM}, 2024.

\bibitem[Su et~al.(2024)Su, Ahmed, Lu, Pan, Bo, and Liu]{su2024roformer}
Jianlin Su, Murtadha Ahmed, Yu~Lu, Shengfeng Pan, Wen Bo, and Yunfeng Liu.
\newblock Roformer: Enhanced transformer with rotary position embedding.
\newblock \emph{Neurocomputing}, 2024.

\bibitem[Sun et~al.(2024{\natexlab{a}})Sun, Jiang, Chen, Zhang, Peng, Luo, and Yuan]{llamagen}
Peize Sun, Yi~Jiang, Shoufa Chen, Shilong Zhang, Bingyue Peng, Ping Luo, and Zehuan Yuan.
\newblock Autoregressive model beats diffusion: Llama for scalable image generation.
\newblock \emph{arXiv preprint arXiv:2406.06525}, 2024{\natexlab{a}}.

\bibitem[Sun et~al.(2024{\natexlab{b}})Sun, Cui, Zhang, Zhang, Yu, Wang, Rao, Liu, Huang, and Wang]{emu2}
Quan Sun, Yufeng Cui, Xiaosong Zhang, Fan Zhang, Qiying Yu, Yueze Wang, Yongming Rao, Jingjing Liu, Tiejun Huang, and Xinlong Wang.
\newblock Generative multimodal models are in-context learners.
\newblock In \emph{CVPR}, 2024{\natexlab{b}}.

\bibitem[Tancik et~al.(2023)Tancik, Weber, Ng, Li, Yi, Wang, Kristoffersen, Austin, Salahi, Ahuja, et~al.]{tancik2023nerfstudio}
Matthew Tancik, Ethan Weber, Evonne Ng, Ruilong Li, Brent Yi, Terrance Wang, Alexander Kristoffersen, Jake Austin, Kamyar Salahi, Abhik Ahuja, et~al.
\newblock Nerfstudio: A modular framework for neural radiance field development.
\newblock In \emph{ACM SIGGRAPH}, 2023.

\bibitem[Tang et~al.(2018)Tang, Feng, Kuang, Chen, and Zhang]{tang2018fastvideoshottransition}
Shitao Tang, Litong Feng, Zhanghui Kuang, Yimin Chen, and Wei Zhang.
\newblock Fast video shot transition localization with deep structured models.
\newblock In \emph{ACCV}, 2018.

\bibitem[Tatarchenko et~al.(2016)Tatarchenko, Dosovitskiy, and Brox]{tatarchenko2016multi}
Maxim Tatarchenko, Alexey Dosovitskiy, and Thomas Brox.
\newblock Multi-view 3d models from single images with a convolutional network.
\newblock In \emph{ECCV}, 2016.

\bibitem[Team(2024{\natexlab{a}})]{chameleon}
Chameleon Team.
\newblock Chameleon: Mixed-modal early-fusion foundation models.
\newblock \emph{URL https://arxiv. org/abs/2405.09818}, 2024{\natexlab{a}}.

\bibitem[Team(2024{\natexlab{b}})]{gemma2}
Gemma Team.
\newblock Gemma 2: Improving open language models at a practical size, 2024{\natexlab{b}}.
\newblock URL \url{https://arxiv.org/abs/2408.00118}.

\bibitem[Technologies(2024)]{1xgpt}
1X~Technologies.
\newblock 1xgpt, 2024.
\newblock URL \url{https://github.com/1x-technologies/1xgpt}.

\bibitem[Teed and Deng(2020)]{teed2020raft}
Zachary Teed and Jia Deng.
\newblock Raft: Recurrent all-pairs field transforms for optical flow.
\newblock In \emph{ECCV}, 2020.

\bibitem[Teed and Deng(2021)]{teed2021droid}
Zachary Teed and Jia Deng.
\newblock Droid-slam: Deep visual slam for monocular, stereo, and rgb-d cameras.
\newblock In \emph{NeurIPS}, 2021.

\bibitem[Teng et~al.(2024)Teng, Shi, Liu, Ning, Dai, Wang, Li, and Liu]{teng2024accelerating}
Yao Teng, Han Shi, Xian Liu, Xuefei Ning, Guohao Dai, Yu~Wang, Zhenguo Li, and Xihui Liu.
\newblock Accelerating auto-regressive text-to-image generation with training-free speculative jacobi decoding.
\newblock \emph{arXiv preprint arXiv:2410.01699}, 2024.

\bibitem[Tucker and Snavely(2020)]{tucker2020single}
Richard Tucker and Noah Snavely.
\newblock Single-view view synthesis with multiplane images.
\newblock In \emph{CVPR}, 2020.

\bibitem[Tulyakov et~al.(2018)Tulyakov, Liu, Yang, and Kautz]{tulyakov2018mocogan}
Sergey Tulyakov, Ming-Yu Liu, Xiaodong Yang, and Jan Kautz.
\newblock {MoCoGAN}: Decomposing motion and content for video generation.
\newblock In \emph{CVPR}, 2018.

\bibitem[Unterthiner et~al.(2019)Unterthiner, van Steenkiste, Kurach, Marinier, Michalski, and Gelly]{FVD}
Thomas Unterthiner, Sjoerd van Steenkiste, Karol Kurach, Rapha{\"e}l Marinier, Marcin Michalski, and Sylvain Gelly.
\newblock Fvd: A new metric for video generation.
\newblock In \emph{ICLR Workshops}, 2019.

\bibitem[Valevski et~al.(2024)Valevski, Leviathan, Arar, and Fruchter]{valevski2024diffusion}
Dani Valevski, Yaniv Leviathan, Moab Arar, and Shlomi Fruchter.
\newblock Diffusion models are real-time game engines.
\newblock \emph{arXiv preprint arXiv:2408.14837}, 2024.

\bibitem[van~den Oord et~al.(2017)van~den Oord, Vinyals, and Kavukcuoglu]{van2017neural}
Aaron van~den Oord, Oriol Vinyals, and Koray Kavukcuoglu.
\newblock Neural discrete representation learning.
\newblock In \emph{NeurIPS}, 2017.

\bibitem[Vaswani et~al.(2017)Vaswani, Shazeer, Parmar, Uszkoreit, Jones, Gomez, Kaiser, and Polosukhin]{vaswani2017attention}
Ashish Vaswani, Noam Shazeer, Niki Parmar, Jakob Uszkoreit, Llion Jones, Aidan~N Gomez, \L~ukasz Kaiser, and Illia Polosukhin.
\newblock Attention is all you need.
\newblock In \emph{NeurIPS}, 2017.

\bibitem[Villegas et~al.(2023)Villegas, Babaeizadeh, Kindermans, Moraldo, Zhang, Saffar, Castro, Kunze, and Erhan]{villegas2022phenaki}
Ruben Villegas, Mohammad Babaeizadeh, Pieter-Jan Kindermans, Hernan Moraldo, Han Zhang, Mohammad~Taghi Saffar, Santiago Castro, Julius Kunze, and Dumitru Erhan.
\newblock Phenaki: Variable length video generation from open domain textual description.
\newblock In \emph{ICLR}, 2023.

\bibitem[Walke et~al.(2023)Walke, Black, Lee, Kim, Du, Zheng, Zhao, Hansen-Estruch, Vuong, He, Myers, Fang, Finn, and Levine]{walke2023bridgev2}
Homer Walke, Kevin Black, Abraham Lee, Moo~Jin Kim, Max Du, Chongyi Zheng, Tony Zhao, Philippe Hansen-Estruch, Quan Vuong, Andre He, Vivek Myers, Kuan Fang, Chelsea Finn, and Sergey Levine.
\newblock Bridgedata v2: A dataset for robot learning at scale.
\newblock In \emph{CoRL}, 2023.

\bibitem[Wang et~al.(2024{\natexlab{a}})Wang, Jiang, Yuan, Peng, Wu, and Jiang]{wang2024omnitokenizer}
Junke Wang, Yi~Jiang, Zehuan Yuan, Binyue Peng, Zuxuan Wu, and Yu-Gang Jiang.
\newblock Omnitokenizer: A joint image-video tokenizer for visual generation.
\newblock \emph{arXiv preprint arXiv:2406.09399}, 2024{\natexlab{a}}.

\bibitem[Wang et~al.(2021{\natexlab{a}})Wang, Liu, Liu, Theobalt, Komura, and Wang]{wang2021neus}
Peng Wang, Lingjie Liu, Yuan Liu, Christian Theobalt, Taku Komura, and Wenping Wang.
\newblock Neus: Learning neural implicit surfaces by volume rendering for multi-view reconstruction.
\newblock In \emph{NeurIPS}, 2021{\natexlab{a}}.

\bibitem[Wang et~al.(2024{\natexlab{b}})Wang, Bai, Tan, Wang, Fan, Bai, Chen, Liu, Wang, Ge, et~al.]{wang2024qwen2}
Peng Wang, Shuai Bai, Sinan Tan, Shijie Wang, Zhihao Fan, Jinze Bai, Keqin Chen, Xuejing Liu, Jialin Wang, Wenbin Ge, et~al.
\newblock Qwen2-vl: Enhancing vision-language model's perception of the world at any resolution.
\newblock \emph{arXiv preprint arXiv:2409.12191}, 2024{\natexlab{b}}.

\bibitem[Wang et~al.(2018)Wang, Liu, Zhu, Liu, Tao, Kautz, and Catanzaro]{wang2018video}
Ting-Chun Wang, Ming-Yu Liu, Jun-Yan Zhu, Guilin Liu, Andrew Tao, Jan Kautz, and Bryan Catanzaro.
\newblock Video-to-video synthesis.
\newblock In \emph{NeurIPS}, 2018.

\bibitem[Wang et~al.(2019)Wang, Liu, Tao, Liu, Kautz, and Catanzaro]{wang2019few}
Ting-Chun Wang, Ming-Yu Liu, Andrew Tao, Guilin Liu, Jan Kautz, and Bryan Catanzaro.
\newblock Few-shot video-to-video synthesis.
\newblock In \emph{NeurIPS}, 2019.

\bibitem[Wang et~al.(2021{\natexlab{b}})Wang, Mallya, and Liu]{wang2021one}
Ting-Chun Wang, Arun Mallya, and Ming-Yu Liu.
\newblock One-shot free-view neural talking-head synthesis for video conferencing.
\newblock In \emph{CVPR}, 2021{\natexlab{b}}.

\bibitem[Wang et~al.(2024{\natexlab{c}})Wang, Zhang, Gao, Wang, Zhou, Zhang, Yan, and Sang]{wang2024unianimate}
Xiang Wang, Shiwei Zhang, Changxin Gao, Jiayu Wang, Xiaoqiang Zhou, Yingya Zhang, Luxin Yan, and Nong Sang.
\newblock Unianimate: Taming unified video diffusion models for consistent human image animation.
\newblock \emph{arXiv preprint arXiv:2406.01188}, 2024{\natexlab{c}}.

\bibitem[Wang et~al.(2023{\natexlab{a}})Wang, Zhu, Huang, Chen, Zhu, and Lu]{wang2023drivedreamer}
Xiaofeng Wang, Zheng Zhu, Guan Huang, Xinze Chen, Jiagang Zhu, and Jiwen Lu.
\newblock Drivedreamer: Towards real-world-driven world models for autonomous driving.
\newblock \emph{arXiv preprint arXiv:2309.09777}, 2023{\natexlab{a}}.

\bibitem[Wang et~al.(2024{\natexlab{d}})Wang, Zhang, Luo, Sun, Cui, Wang, Zhang, Wang, Li, Yu, et~al.]{wang2024emu3}
Xinlong Wang, Xiaosong Zhang, Zhengxiong Luo, Quan Sun, Yufeng Cui, Jinsheng Wang, Fan Zhang, Yueze Wang, Zhen Li, Qiying Yu, et~al.
\newblock Emu3: Next-token prediction is all you need.
\newblock \emph{arXiv preprint arXiv:2409.18869}, 2024{\natexlab{d}}.

\bibitem[Wang et~al.(2023{\natexlab{b}})Wang, Chen, Ma, Zhou, Huang, Wang, Yang, He, Yu, Yang, et~al.]{wang2023lavie}
Yaohui Wang, Xinyuan Chen, Xin Ma, Shangchen Zhou, Ziqi Huang, Yi~Wang, Ceyuan Yang, Yinan He, Jiashuo Yu, Peiqing Yang, et~al.
\newblock Lavie: High-quality video generation with cascaded latent diffusion models.
\newblock \emph{arXiv preprint arXiv:2309.15103}, 2023{\natexlab{b}}.

\bibitem[Wang et~al.(2025)Wang, Li, Li, Yu, He, Chen, Pei, Zheng, Wang, Shi, et~al.]{wang2024internvideo2}
Yi~Wang, Kunchang Li, Xinhao Li, Jiashuo Yu, Yinan He, Guo Chen, Baoqi Pei, Rongkun Zheng, Zun Wang, Yansong Shi, et~al.
\newblock Internvideo2: Scaling foundation models for multimodal video understanding.
\newblock In \emph{ECCV}, 2025.

\bibitem[Wang et~al.(2024{\natexlab{e}})Wang, He, Fan, Li, Chen, and Zhang]{wang2024driving}
Yuqi Wang, Jiawei He, Lue Fan, Hongxin Li, Yuntao Chen, and Zhaoxiang Zhang.
\newblock Driving into the future: Multiview visual forecasting and planning with world model for autonomous driving.
\newblock In \emph{CVPR}, 2024{\natexlab{e}}.

\bibitem[Wang et~al.(2024{\natexlab{f}})Wang, Li, Mandlekar, Xu, Fan, Narang, Fan, Zhu, Balaji, Zhou, et~al.]{wang2024one}
Zhendong Wang, Zhaoshuo Li, Ajay Mandlekar, Zhenjia Xu, Jiaojiao Fan, Yashraj Narang, Linxi Fan, Yuke Zhu, Yogesh Balaji, Mingyuan Zhou, et~al.
\newblock One-step diffusion policy: Fast visuomotor policies via diffusion distillation.
\newblock \emph{arXiv preprint arXiv:2410.21257}, 2024{\natexlab{f}}.

\bibitem[Wang et~al.(2024{\natexlab{g}})Wang, Yuan, Wang, Li, Chen, Xia, Luo, and Shan]{wang2024motionctrl}
Zhouxia Wang, Ziyang Yuan, Xintao Wang, Yaowei Li, Tianshui Chen, Menghan Xia, Ping Luo, and Ying Shan.
\newblock Motionctrl: A unified and flexible motion controller for video generation.
\newblock In \emph{ACM SIGGRAPH}, 2024{\natexlab{g}}.

\bibitem[Weng et~al.(2023)Weng, Yang, Yu, Wang, Tang, Yang, and Zhang]{weng2023beware}
Qizhen Weng, Lingyun Yang, Yinghao Yu, Wei Wang, Xiaochuan Tang, Guodong Yang, and Liping Zhang.
\newblock Beware of fragmentation: Scheduling $\{$GPU-Sharing$\}$ workloads with fragmentation gradient descent.
\newblock In \emph{USENIX ATC}, 2023.

\bibitem[Wiles et~al.(2020)Wiles, Gkioxari, Szeliski, and Johnson]{wiles2020synsin}
Olivia Wiles, Georgia Gkioxari, Richard Szeliski, and Justin Johnson.
\newblock Synsin: End-to-end view synthesis from a single image.
\newblock In \emph{CVPR}, 2020.

\bibitem[Wortsman et~al.(2023)Wortsman, Liu, Xiao, Everett, Alemi, Adlam, Co-Reyes, Gur, Kumar, Novak, et~al.]{wortsman2023small}
Mitchell Wortsman, Peter~J Liu, Lechao Xiao, Katie Everett, Alex Alemi, Ben Adlam, John~D Co-Reyes, Izzeddin Gur, Abhishek Kumar, Roman Novak, et~al.
\newblock Small-scale proxies for large-scale transformer training instabilities.
\newblock \emph{arXiv preprint arXiv:2309.14322}, 2023.

\bibitem[Wu et~al.(2022)Wu, Liang, Ji, Yang, Fang, Jiang, and Duan]{wu2022nuwa}
Chenfei Wu, Jian Liang, Lei Ji, Fan Yang, Yuejian Fang, Daxin Jiang, and Nan Duan.
\newblock N{\"u}wa: Visual synthesis pre-training for neural visual world creation.
\newblock In \emph{ECCV}, 2022.

\bibitem[Wu et~al.(2023{\natexlab{a}})Wu, Zhang, Liao, Chen, Hou, Wang, Sun, Yan, and Lin]{wu2023exploring}
Haoning Wu, Erli Zhang, Liang Liao, Chaofeng Chen, Jingwen Hou, Annan Wang, Wenxiu Sun, Qiong Yan, and Weisi Lin.
\newblock Exploring video quality assessment on user generated contents from aesthetic and technical perspectives.
\newblock In \emph{ICCV}, 2023{\natexlab{a}}.

\bibitem[Wu et~al.(2023{\natexlab{b}})Wu, Escontrela, Hafner, Abbeel, and Goldberg]{wu2023daydreamer}
Philipp Wu, Alejandro Escontrela, Danijar Hafner, Pieter Abbeel, and Ken Goldberg.
\newblock Daydreamer: World models for physical robot learning.
\newblock In \emph{CoRL}, 2023{\natexlab{b}}.

\bibitem[Wu et~al.(2024)Wu, Zhang, Chen, Tang, Li, Fang, Zhu, Xie, Yin, Yi, et~al.]{wu2024vilau}
Yecheng Wu, Zhuoyang Zhang, Junyu Chen, Haotian Tang, Dacheng Li, Yunhao Fang, Ligeng Zhu, Enze Xie, Hongxu Yin, Li~Yi, et~al.
\newblock Vila-u: a unified foundation model integrating visual understanding and generation.
\newblock \emph{arXiv preprint arXiv:2409.04429}, 2024.

\bibitem[Wu and He(2018)]{wu2018group}
Yuxin Wu and Kaiming He.
\newblock Group normalization.
\newblock In \emph{ECCV}, 2018.

\bibitem[Xu et~al.(2024)Xu, Nie, Liu, Liu, Kautz, Wang, and Vahdat]{xu2024camco}
Dejia Xu, Weili Nie, Chao Liu, Sifei Liu, Jan Kautz, Zhangyang Wang, and Arash Vahdat.
\newblock Camco: Camera-controllable 3d-consistent image-to-video generation.
\newblock \emph{arXiv preprint arXiv:2406.02509}, 2024.

\bibitem[Xu et~al.(2019)Xu, Sun, Zhang, Zhao, and Lin]{xu2019understanding}
Jingjing Xu, Xu~Sun, Zhiyuan Zhang, Guangxiang Zhao, and Junyang Lin.
\newblock Understanding and improving layer normalization.
\newblock In \emph{NeurIPS}, 2019.

\bibitem[Xue et~al.(2024)Xue, Chen, Li, Hu, Zhu, Li, Fang, Tang, Yang, Liu, et~al.]{xue2024longvilascalinglongcontextvisual}
Fuzhao Xue, Yukang Chen, Dacheng Li, Qinghao Hu, Ligeng Zhu, Xiuyu Li, Yunhao Fang, Haotian Tang, Shang Yang, Zhijian Liu, et~al.
\newblock Longvila: Scaling long-context visual language models for long videos.
\newblock \emph{arXiv preprint arXiv:2408.10188}, 2024.

\bibitem[Yan et~al.(2021)Yan, Zhang, Abbeel, and Srinivas]{yan2021videogpt}
Wilson Yan, Yunzhi Zhang, Pieter Abbeel, and Aravind Srinivas.
\newblock Videogpt: Video generation using vq-vae and transformers.
\newblock \emph{arXiv preprint arXiv:2104.10157}, 2021.

\bibitem[Yang et~al.(2024{\natexlab{a}})Yang, Yang, Zhang, Hui, Zheng, Yu, Li, Liu, Huang, Wei, et~al.]{qwen2p5}
An~Yang, Baosong Yang, Beichen Zhang, Binyuan Hui, Bo~Zheng, Bowen Yu, Chengyuan Li, Dayiheng Liu, Fei Huang, Haoran Wei, et~al.
\newblock Qwen2.5 technical report.
\newblock \emph{arXiv preprint arXiv:2412.15115}, 2024{\natexlab{a}}.

\bibitem[Yang et~al.(2024{\natexlab{b}})Yang, Huang, Chai, Jiang, and Hwang]{yang2024samurai}
Cheng-Yen Yang, Hsiang-Wei Huang, Wenhao Chai, Zhongyu Jiang, and Jenq-Neng Hwang.
\newblock Samurai: Adapting segment anything model for zero-shot visual tracking with motion-aware memory.
\newblock \emph{arXiv preprint arXiv:2411.11922}, 2024{\natexlab{b}}.

\bibitem[Yang et~al.(2024{\natexlab{c}})Yang, Gao, Qiu, Chen, Li, Dai, Chitta, Wu, Zeng, Luo, et~al.]{yang2024generalized}
Jiazhi Yang, Shenyuan Gao, Yihang Qiu, Li~Chen, Tianyu Li, Bo~Dai, Kashyap Chitta, Penghao Wu, Jia Zeng, Ping Luo, et~al.
\newblock Generalized predictive model for autonomous driving.
\newblock In \emph{CVPR}, 2024{\natexlab{c}}.

\bibitem[Yang et~al.(2023)Yang, Du, Ghasemipour, Tompson, Schuurmans, and Abbeel]{yang2023learning}
Mengjiao Yang, Yilun Du, Kamyar Ghasemipour, Jonathan Tompson, Dale Schuurmans, and Pieter Abbeel.
\newblock Learning interactive real-world simulators.
\newblock \emph{arXiv preprint arXiv:2310.06114}, 2023.

\bibitem[Yang et~al.(2024{\natexlab{d}})Yang, Teng, Zheng, Ding, Huang, Xu, Yang, Hong, Zhang, Feng, et~al.]{yang2024cogvideox}
Zhuoyi Yang, Jiayan Teng, Wendi Zheng, Ming Ding, Shiyu Huang, Jiazheng Xu, Yuanming Yang, Wenyi Hong, Xiaohan Zhang, Guanyu Feng, et~al.
\newblock Cogvideox: Text-to-video diffusion models with an expert transformer.
\newblock \emph{arXiv preprint arXiv:2408.06072}, 2024{\natexlab{d}}.

\bibitem[Yin et~al.(2024)Yin, Zhang, Zhang, Freeman, Durand, Shechtman, and Huang]{yin2024slow}
Tianwei Yin, Qiang Zhang, Richard Zhang, William~T Freeman, Fredo Durand, Eli Shechtman, and Xun Huang.
\newblock From slow bidirectional to fast causal video generators.
\newblock \emph{arXiv preprint arXiv:2412.07772}, 2024.

\bibitem[Yu et~al.(2021)Yu, Ye, Tancik, and Kanazawa]{yu2021pixelnerf}
Alex Yu, Vickie Ye, Matthew Tancik, and Angjoo Kanazawa.
\newblock pixelnerf: Neural radiance fields from one or few images.
\newblock In \emph{CVPR}, 2021.

\bibitem[Yu et~al.(2020)Yu, Chen, Wang, Xian, Chen, Liu, Madhavan, and Darrell]{yu2020bdd100k}
Fisher Yu, Haofeng Chen, Xin Wang, Wenqi Xian, Yingying Chen, Fangchen Liu, Vashisht Madhavan, and Trevor Darrell.
\newblock Bdd100k: A diverse driving dataset for heterogeneous multitask learning.
\newblock In \emph{CVPR}, 2020.

\bibitem[Yu et~al.(2022)Yu, Xu, Koh, Luong, Baid, Wang, Vasudevan, Ku, Yang, Ayan, et~al.]{yu2022scaling}
Jiahui Yu, Yuanzhong Xu, Jing~Yu Koh, Thang Luong, Gunjan Baid, Zirui Wang, Vijay Vasudevan, Alexander Ku, Yinfei Yang, Burcu~Karagol Ayan, et~al.
\newblock Scaling autoregressive models for content-rich text-to-image generation.
\newblock \emph{TMLR}, 2022.

\bibitem[Yu et~al.(2023{\natexlab{a}})Yu, Cheng, Sohn, Lezama, Zhang, Chang, Hauptmann, Yang, Hao, Essa, and Jiang]{yu2023magvit}
Lijun Yu, Yong Cheng, Kihyuk Sohn, Jos{\'e} Lezama, Han Zhang, Huiwen Chang, Alexander~G Hauptmann, Ming-Hsuan Yang, Yuan Hao, Irfan Essa, and Lu~Jiang.
\newblock {MAGVIT}: Masked generative video transformer.
\newblock In \emph{CVPR}, 2023{\natexlab{a}}.

\bibitem[Yu et~al.(2024{\natexlab{a}})Yu, Lezama, Gundavarapu, Versari, Sohn, Minnen, Cheng, Gupta, Gu, Hauptmann, Gong, Yang, Essa, Ross, and Jiang]{yu2024language}
Lijun Yu, Jose Lezama, Nitesh~Bharadwaj Gundavarapu, Luca Versari, Kihyuk Sohn, David Minnen, Yong Cheng, Agrim Gupta, Xiuye Gu, Alexander~G Hauptmann, Boqing Gong, Ming-Hsuan Yang, Irfan Essa, David~A Ross, and Lu~Jiang.
\newblock Language model beats diffusion - tokenizer is key to visual generation.
\newblock In \emph{ICLR}, 2024{\natexlab{a}}.

\bibitem[Yu et~al.(2024{\natexlab{b}})Yu, Weber, Deng, Shen, Cremers, and Chen]{yu2024titok}
Qihang Yu, Mark Weber, Xueqing Deng, Xiaohui Shen, Daniel Cremers, and Liang-Chieh Chen.
\newblock An image is worth 32 tokens for reconstruction and generation.
\newblock \emph{arXiv preprint arXiv:2406.07550}, 2024{\natexlab{b}}.

\bibitem[Yu et~al.(2023{\natexlab{b}})Yu, Sohn, Kim, and Shin]{yu2023video}
Sihyun Yu, Kihyuk Sohn, Subin Kim, and Jinwoo Shin.
\newblock Video probabilistic diffusion models in projected latent space.
\newblock In \emph{CVPR}, 2023{\natexlab{b}}.

\bibitem[Zeng et~al.(2024)Zeng, Wei, Zheng, Zou, Wei, Zhang, and Li]{Zeng2023Video}
Yan Zeng, Guoqiang Wei, Jiani Zheng, Jiaxin Zou, Yang Wei, Yuchen Zhang, and Hang Li.
\newblock Make pixels dance: High-dynamic video generation.
\newblock In \emph{CVPR}, 2024.

\bibitem[Zhai et~al.(2023)Zhai, Mustafa, Kolesnikov, and Beyer]{siglip}
Xiaohua Zhai, Basil Mustafa, Alexander Kolesnikov, and Lucas Beyer.
\newblock Sigmoid loss for language image pre-training.
\newblock In \emph{ICCV}, 2023.

\bibitem[Zhang and Sennrich(2019)]{zhang2019root}
Biao Zhang and Rico Sennrich.
\newblock Root mean square layer normalization.
\newblock In \emph{NeurIPS}, 2019.

\bibitem[Zhang et~al.(2018)Zhang, Isola, Efros, Shechtman, and Wang]{zhang2018unreasonable}
Richard Zhang, Phillip Isola, Alexei~A Efros, Eli Shechtman, and Oliver Wang.
\newblock The unreasonable effectiveness of deep features as a perceptual metric.
\newblock In \emph{CVPR}, 2018.

\bibitem[Zhang et~al.(2024)Zhang, Wang, Sun, Yuan, and Huang]{zhang2024storm}
Weipu Zhang, Gang Wang, Jian Sun, Yetian Yuan, and Gao Huang.
\newblock Storm: Efficient stochastic transformer based world models for reinforcement learning.
\newblock In \emph{NeurIPS}, 2024.

\bibitem[Zhao et~al.(2024{\natexlab{a}})Zhao, Wang, Zhu, Chen, Huang, Bao, and Wang]{zhao2024drivedreamer}
Guosheng Zhao, Xiaofeng Wang, Zheng Zhu, Xinze Chen, Guan Huang, Xiaoyi Bao, and Xingang Wang.
\newblock Drivedreamer-2: Llm-enhanced world models for diverse driving video generation.
\newblock \emph{arXiv preprint arXiv:2403.06845}, 2024{\natexlab{a}}.

\bibitem[Zhao et~al.(2024{\natexlab{b}})Zhao, Xiong, and Kr{\"a}henb{\"u}hl]{zhao2024bsqvit}
Yue Zhao, Yuanjun Xiong, and Philipp Kr{\"a}henb{\"u}hl.
\newblock Image and video tokenization with binary spherical quantization.
\newblock \emph{arXiv preprint arXiv:2406.07548}, 2024{\natexlab{b}}.

\bibitem[Zhou et~al.(2024{\natexlab{a}})Zhou, Yu, Babu, Tirumala, Yasunaga, Shamis, Kahn, Ma, Zettlemoyer, and Levy]{zhou2024transfusion}
Chunting Zhou, Lili Yu, Arun Babu, Kushal Tirumala, Michihiro Yasunaga, Leonid Shamis, Jacob Kahn, Xuezhe Ma, Luke Zettlemoyer, and Omer Levy.
\newblock Transfusion: Predict the next token and diffuse images with one multi-modal model.
\newblock \emph{arXiv preprint arXiv:2408.11039}, 2024{\natexlab{a}}.

\bibitem[Zhou et~al.(2024{\natexlab{b}})Zhou, Du, Chen, LI, Yeung, and Gan]{zhourobodreamer}
Siyuan Zhou, Yilun Du, Jiaben Chen, YANDONG LI, Dit-Yan Yeung, and Chuang Gan.
\newblock Robodreamer: Learning compositional world models for robot imagination.
\newblock In \emph{ICML}, 2024{\natexlab{b}}.

\bibitem[Zhou et~al.(2018)Zhou, Tucker, Flynn, Fyffe, and Snavely]{zhou2018stereo}
Tinghui Zhou, Richard Tucker, John Flynn, Graham Fyffe, and Noah Snavely.
\newblock Stereo magnification: learning view synthesis using multiplane images.
\newblock \emph{ACM Transactions on Graphics (TOG)}, 2018.

\bibitem[Zhu et~al.(2024)Zhu, Wu, Guo, Liu, Cheang, and Kong]{zhu2024irasim}
Fangqi Zhu, Hongtao Wu, Song Guo, Yuxiao Liu, Chilam Cheang, and Tao Kong.
\newblock Irasim: Learning interactive real-robot action simulators.
\newblock \emph{arXiv preprint arXiv:2406.14540}, 2024.

\bibitem[Zhu et~al.(2023)Zhu, Huang, Xie, Liu, Deng, Zhang, Wang, and Liu]{zhu2023autoshot}
Wentao Zhu, Yufang Huang, Xiufeng Xie, Wenxian Liu, Jincan Deng, Debing Zhang, Zhangyang Wang, and Ji~Liu.
\newblock Autoshot: A short video dataset and state-of-the-art shot boundary detection.
\newblock In \emph{CVPR Workshops}, 2023.

\end{thebibliography}

\end{document}